\documentclass[fontsize=11pt, a4paper,twoside=semi, bibliography=totoc,usegeometry]{scrbook} 
\usepackage[bottom=3.5cm,top=3.5cm,left=3.5cm,right=3.5cm]{geometry}

\usepackage[hidelinks]{hyperref} 

\newcommand*{\doi}[1]{\texttt{DOI}: \href{http://doi.org/#1}{\nolinkurl{#1}}}
\usepackage{lmodern}
\usepackage{fancyhdr}

\usepackage[utf8]{inputenc}
\usepackage[english]{babel}
\usepackage[chapterstyle=madsen]{komaChapStyle}
\RedeclareSectionCommand[beforeskip=0pt]{chapter}
\usepackage{amsmath}
\usepackage{amssymb}
\usepackage{amsfonts}
\usepackage{float}
\usepackage{mathtools}
\usepackage{fancyhdr} 
\usepackage{bm}
\usepackage[round, sort]{natbib}
\usepackage{verbatim}
\usepackage{graphicx}
\usepackage{changepage} 
\usepackage{bbm}
\usepackage{tabularx}
\usepackage[refpage]{nomencl}
\usepackage{subfig}
\usepackage{centernot}
\usepackage{marvosym} 
\usepackage{booktabs}
\usepackage{etoolbox}

\usepackage{lscape}
\usepackage{algorithm}
\usepackage{tikz}
\usepackage{tkz-graph}
\usetikzlibrary{arrows,automata,positioning,shapes,backgrounds,intersections,arrows.meta}

\usepackage{algpseudocode}
\algdef{SE}{Begin}{End}{\textbf{begin}}{\textbf{end}}
\usepackage{array,multirow}

\usepackage[nameinlink,capitalize]{cleveref}

\usepackage{layouts}
\usepackage{quoting}
\usepackage{hhline}
\usepackage{bbm}
\usepackage{mathtools}
\usepackage{caption}
\usepackage{url}
\usepackage{enumerate}
\usepackage{siunitx}
\usepackage{bibentry}
\usepackage[inline]{enumitem}
\usepackage{tabu}
\usepackage{cancel}
 
\makeatletter
\def\thmt@refnamewithcomma #1#2#3,#4,#5\@nil{%
	\@xa\def\csname\thmt@envname #1utorefname\endcsname{#3}%
	\ifcsname #2refname\endcsname
	\csname #2refname\expandafter\endcsname\expandafter{\thmt@envname}{#3}{#4}%
	\fi
}
\newcommand{\indep}{\rotatebox[origin=c]{90}{$\models$}}
\makeatother

\def\independenT#1#2{\mathrel{\rlap{$#1#2$}\mkern2mu{#1#2}}}

\newcommand\independent{\protect\mathpalette{\protect\independenT}{\perp}}

\def\independenT#1#2{\mathrel{\rlap{$#1#2$}\mkern2mu{#1#2}}}

\newcommand\dsep[1]{{\perp\!\!\!\!\perp_{#1}}}

\renewcommand{\t}{\intercal}

\newcommand{\ep}{\varepsilon}
\renewcommand{\epsilon}{\varepsilon}

\renewcommand{\i}{\infty}

\renewcommand{\phi}{\varphi}

\newcommand{\Cov}{\mathrm{Cov}}
\newcommand{\Var}{\mathrm{Var}}
\newcommand{\Corr}{\mathrm{Corr}}

\newcommand\Jonas[1]{{\color{red}Jonas: #1}}

\newcommand\Martin[1]{{\color{blue}Martin: #1}}

\newcommand\Rm[1]{{\color{gray}Remove: #1}}

\newcommand\lra{%
	\mathrel{{\ooalign{\hss\raisebox{-0.5ex}{$\longrightarrow$}\hss\cr\raisebox{0.5ex}{$\dashleftarrow$}}}}
}

\newcommand{\tPA}[1]{{\widetilde{\mathrm{PA}}({#1})}}
\newcommand{\PA}[1]{{\mathrm{PA}({#1})}}
\DeclareMathOperator{\pa}{PA}
\newcommand{\PAg}[2]{{\mathrm{pa}^{#1}({#2})}}
\newcommand{\CHg}[2]{{\mathrm{ch}^{#1}({#2})}}
\newcommand{\DEg}[2]{{\mathrm{de}^{#1}({#2})}}
\newcommand{\ANg}[2]{{\mathrm{an}^{#1}({#2})}}
\newcommand{\NDg}[2]{{\mathrm{nd}^{#1}({#2})}}

\renewcommand{\root}[1]{\mathrm{rt}(#1)}

\newcommand{\lE}{\ell_\mathrm{E}}
\newcommand{\lG}{\ell_\mathrm{G}}
\newcommand{\lCE}{\ell_\mathrm{CE}}

\newcommand{\convp}{\stackrel{P}{\longrightarrow}}
\newcommand{\convas}{\stackrel{a.s.}{\longrightarrow}}
\newcommand{\convd}{\stackrel{\mathcal{D}}{\longrightarrow}}

\DeclareMathOperator*{\argmin}{arg\,min}
		
\newcommand{\eqd}{\stackrel{\mathcal{D}}{=}}
\newcommand{\eqas}{\stackrel{\mathrm{a.s}}{=}}

\DeclareMathOperator*{\plim}{P-lim}

\newcommand{\define}[4]{\expandafter#1\csname#3#4\endcsname{#2{#4}}}
\newcommand{\R}{\mathbb{R}}

\newcommand{\N}{\mathbb{N}}
\forcsvlist{\define{\newcommand}{\mathcal}{c}}{A,B,C,D,E,F,G,H,I,J,K,L,P,R,M,N,U,S,T,W,V,X,Y,Z,O}
\forcsvlist{\define{\newcommand}{\mathbb}{b}}{A,B,C,D,E,F,G,H,I,J,K,L,P,Q,M,N,Z,T,X,Y}
\forcsvlist{\define{\newcommand}{\mathbf}{f}}{A,B,C,D,E,F,G,H,I,J,K,L,M,N,O,P,Q,R,S,T,U,W,V,X,Y,Z,x}
\newcommand{\E}{\mathbb{E}}

\newcommand{\lp}{\left( }
\newcommand{\rp}{\right) }
\newcommand{\lb}{\left\lbrace}
\newcommand{\rb}{\right\rbrace}
\newcommand{\lf}{\left[}
\newcommand{\rf}{\right]}

\newcommand{\ra}{\rangle}
\newcommand{\la}{\langle}

\usepackage{newfloat}
\DeclareFloatingEnvironment[
    fileext=los,
    listname={List of Schemes},
    name=Running Example,
    placement=tbhp,
    within=none,
]{runex}

\numberwithin{equation}{section}

\DeclareOldFontCommand{\it}{\normalfont\itshape}{\mathit}
\DeclareOldFontCommand{\bf}{\normalfont\bfseries}{\mathbf}

\graphicspath{{./pulse/Plots/}{./c4c/Plots/}{./generalization/figures/}{./trees/Plots/}}

\definecolor{gray75}{gray}{0.75}

\usepackage{amsthm}
\usepackage{thmtools, thm-restate}
\newtheorem{theorem}{Theorem}
\newtheorem{assumption}{Assumption}
\newtheorem{proposition}{Proposition}

\newtheorem{lemma}{Lemma}
\newtheorem{definition}{Definition}

\declaretheoremstyle[
spaceabove=\topsep, spacebelow=\topsep,
headfont=\normalfont\bfseries,
notefont=\mdseries, notebraces={(}{)},
bodyfont=\normalfont,
postheadspace=1em,
qed=$\circ$ 
]{example}

\declaretheorem[style=example]{example}
\declaretheorem[style=example]{remark}

\usepackage{pifont}
\newcommand{\xmark}{\ding{55}}%
\newcommand{\bt}{\fontseries{b}\selectfont}
\newenvironment{proofenv}[1] {\noindent\textbf{Proof of {#1}:}}{\hfill$\square$\bigskip}

\usepackage{apptools}
\usepackage[toc,page]{appendix}
\crefname{assumption}{assumption}{assumptions}
\crefname{appsec}{appendix}{appendices}
\Crefname{appsec}{Appendix}{Appendices}

\counterwithin{assumption}{chapter}
\counterwithin{definition}{chapter}
\counterwithin{lemma}{chapter}
\counterwithin{theorem}{chapter}
\counterwithin{corollary}{chapter}
\counterwithin{example}{chapter}
\counterwithin{remark}{chapter}
\counterwithin{table}{chapter}
\counterwithin{figure}{chapter}
\counterwithin{equation}{chapter}
\counterwithin{proposition}{chapter}
\counterwithin{algorithm}{chapter}

\renewcommand{\thesection}{\arabic{chapter}.\arabic{section}}
\renewcommand{\theequation}{\arabic{chapter}.\arabic{equation}}
\renewcommand{\thetheorem}{\arabic{chapter}.\arabic{theorem}}
\renewcommand{\theassumption}{\arabic{chapter}.\arabic{assumption}}
\renewcommand{\theproposition}{\arabic{chapter}.\arabic{proposition}}
\renewcommand{\thecorollary}{\arabic{chapter}.\arabic{corollary}}
\renewcommand{\thelemma}{\arabic{chapter}.\arabic{lemma}}
\renewcommand{\theexample}{\arabic{chapter}.\arabic{example}}
\renewcommand{\theremark}{\arabic{chapter}.\arabic{remark}}
\renewcommand{\thedefinition}{\arabic{chapter}.\arabic{definition}}
\renewcommand{\thefigure}{\arabic{chapter}.\arabic{figure}}
\renewcommand{\thetable}{\arabic{chapter}.\arabic{table}}
\renewcommand{\thealgorithm}{\arabic{chapter}.\arabic{algorithm}}

\definecolor{TIDY}{RGB}{0,0,0}
\definecolor{FACTOR}{RGB}{81,81,81}
\usepackage{xfrac}

\renewcommand\thmcontinues[1]{Continued}

\usepackage{nicefrac} 
\usepackage{microtype}
\usepackage{array}
\usepackage{extpfeil}
\usepackage{placeins}
\usepackage{tkz-graph}
\usetikzlibrary{shapes.geometric}
\usetikzlibrary{backgrounds}
\usetikzlibrary{arrows.meta}
\usepackage[framemethod=TikZ]{mdframed}
\usepackage{colortbl}

\renewcommand\epsilon{\varepsilon}
\renewcommand\subset{\subseteq}

\renewcommand\P{\mathbb{P}}
\newcommand\var{\operatorname{Var}}
\newcommand\cov{\operatorname{Cov}}

\newcommand\PP{\mathcal{P}}

\newcommand\FF{\mathcal{F}}
\newcommand\GG{\mathcal{G}}
\newcommand\II{\mathcal{I}}
\newcommand\HH{\mathcal{H}}
\newcommand\MM{\mathcal{M}}

\newcommand\fs{f_{\diamond}}

\newcommand\gs{g_{\diamond}}

\newcommand\bes{b_{\diamond}}
\newcommand\beb{\bar{b}}

\usepackage{tcolorbox}
\newtcbox{\mybox}{nobeforeafter,colframe=black!50,colback=white,boxrule=1.8pt,arc=5pt,
	boxsep=0pt,left=6pt,right=6pt,top=6pt,bottom=6pt,tcbox raise base}

\newcommand{\B}[1]{\mathbf{#1}}
\newcommand{\given}{\, \vert \,}
\newcommand{\st}{\, : \,}
\newcommand{\iid}{\overset{\text{iid}}{\sim}}
\DeclarePairedDelimiterX{\card}[1]{\lvert}{\rvert}{#1}
\DeclarePairedDelimiterX{\norm}[1]{\lVert}{\rVert}{#1}
\DeclarePairedDelimiterX{\abs}[1]{\lvert}{\rvert}{#1}
\newcommand{\supp}{\mathrm{supp}}
\DeclareSymbolFont{largesymbolsA}{U}{txexa}{m}{n}
\DeclareMathSymbol{\varprod}{\mathop}{largesymbolsA}{16} %
\DeclareMathOperator{\sign}{sign}

\usepackage{nameref}

\newlength{\drop}
\author{Martin Emil Jakobsen}

\newcommand*{\titleTMB}{\begingroup
	\drop=0.1\textheight
	\centering
	\settowidth{\unitlength}{\Huge Causality and Generalizability}
	\vspace*{4\baselineskip}
	{\large\scshape Martin Emil Jakobsen}\\[\baselineskip]
	\rule{\unitlength}{1.6pt}\vspace*{-\baselineskip}\vspace*{2pt}
	\rule{\unitlength}{0.4pt}\\[\baselineskip]
	{\Huge Causality and Generalizability}\\[\baselineskip]
	{\LARGE \itshape Identifiability and Learning Methods}\\[0.2\baselineskip]
	\rule{\unitlength}{0.4pt}\vspace*{-\baselineskip}\vspace{3.2pt}
	\rule{\unitlength}{1.6pt}\\[\baselineskip]
	{\Large\scshape phd thesis}\\[\baselineskip]\vspace{1cm}
	{\small \scshape this thesis has been submitted to the phd school of \\the faculty of 	science, university of copenhagen}
	\vfill
	
	{\large\scshape Department of Mathematical Sciences \\ University of Copenhagen}\\[\baselineskip]
	{\small\scshape August 2021}\\ [\baselineskip]
	
	\vspace*{\drop}
	\endgroup}

\begin{document}

\frontmatter

\begin{titlepage}
	\centering
	\titleTMB
\end{titlepage}


Martin Emil Jakobsen  \par
\texttt{m.jakobsen@math.ku.dk}   \par
\texttt{martin.emil.jakobsen@gmail.com}  \par
Department of Mathematical Sciences \par
University of Copenhagen \par
Universitetsparken 5 \par
2100 Copenhagen \par
Denmark

\vspace{2cm}

\begin{minipage}[t]{0.25\linewidth}
    \begin{flushleft}
    	{\bf Thesis title:} \par
    	\, \vspace*{.3cm}\\
        {\bf Supervisor:} \par 
        \, \vspace*{.3cm} \\     
        {\bf Assessment} \par 
        {\bf committee:} \vspace*{.3cm}\\ 
        \, \par
        \, \vspace*{.3cm} \\
        \, \par
        \, \vspace*{.3cm} \\
        {\bf Date of} \par
        {\bf Submission:}\vspace*{.3cm}\\
        {\bf Date of} \par
        {\bf Defense:}\\

    \end{flushleft}
\end{minipage}%
\begin{minipage}[t]{0.75\linewidth}
    \begin{flushleft}
    	Causality and Generalizability: \par
    	Identifiability and Learning Methods\vspace*{.3cm}\\

        Professor Jonas Peters \par 
        University of Copenhagen \vspace*{.3cm} \\

        Associate Professor Trine Krogh Boomsma (chair) \par
        University of Copenhagen \vspace*{.3cm} \\
        
        Professor Søren Hauberg \par
        Technical University of Denmark  \vspace*{.3cm}  \\
        
        Professor Joris Mooij \par
        University of Amsterdam	 \vspace*{.3cm}\\
        August 31, \par 
        2021 \vspace*{.3cm}  \\
        November 4, \par
        2021
    \end{flushleft}
\end{minipage}%

\vfill
\noindent \textit{This thesis has been submitted to the PhD School of The Faculty of
Science,\\ University of Copenhagen. It was supported by the Carlsberg Foundation.}

\newpage


\addcontentsline{toc}{chapter}{Abstract}
\begin{center}
  \normalfont\usekomafont{disposition} \Large Abstract 
\end{center}
This Ph.D.\ thesis contains several contributions to the field of statistical causal modeling. Statistical causal models are statistical models embedded with causal assumptions that allow for the inference and reasoning about the behavior of stochastic systems affected by external manipulation (interventions). This thesis contributes to the research areas concerning the estimation of causal effects, causal structure learning, and distributionally robust (out-of-distribution generalizing) prediction methods. We present novel and consistent linear and non-linear causal effects estimators in instrumental variable settings that employ data-dependent mean squared prediction error regularization. Our proposed estimators show, in certain settings, mean squared error improvements compared to both canonical and state-of-the-art estimators. We show that recent research on distributionally robust prediction methods has connections to well-studied estimators from econometrics. This connection leads us to prove that general K-class estimators possess distributional robustness properties. We, furthermore,  propose a general framework for distributional robustness with respect to intervention-induced distributions. In this framework, we derive sufficient conditions for the identifiability of distributionally robust prediction methods and present impossibility results that show the necessity of several of these conditions. We present a new structure learning method applicable in additive noise models with directed trees as causal graphs. We prove consistency in a vanishing identifiability setup and provide a method for testing substructure hypotheses with asymptotic family-wise error control that remains valid post-selection.  Finally, we present heuristic ideas for learning summary graphs of nonlinear time-series models.

\vspace{0.3cm}
\begin{center}
  \normalfont\usekomafont{disposition}\Large Resumé
\end{center}
Denne Ph.D.\ afhandling indeholder flere bidrag til forskningsområdet for statistisk kausal modellering.  Statistiske kausale modeller er statistiske modeller med kausale antagelser, som muliggør inferens og ræsonnement omkring stokastiske systemers adfærd under ekstern manipulation. Denne afhandling bidrager til forskningsområderne vedrørende estimering af kausale effekter, kausale strukturer og  fordelingsrobuste prædiktionsmetoder. Vi præsenterer nye estimatorer for lineære og ikke-lineære kausale effekter i modeller med instrumentelle variabler. Disse estimatorer anvender dataafhængig regulering og viser forbedret gennemsnitlig kvadratfejl sammenlignet med anerkendte metoder. Vi viser, at nyere forskning, om fordelingsrobuste forudsigelsesmetoder har forbindelser til velkendte estimatorer fra økonometri. Vi beviser, at generelle K-klasse estimatorer besidder fordelingsrobuste prædiktions egenskaber. Vi foreslår endvidere en kausal tilgang til fordelingsrobuste prædiktionsmetoder. Vi udleder tilstrækkelige betingelser for identificering af fordelingsrobuste prædiktionsmetoder og viser endvidere nødvendigheden af flere af disse betingelser. Vi præsenterer en ny metode til at estimere kausale strukturer, der kan anvendes i modeller med additiv støj og orienterede træer som kausale grafer. Vi beviser, at metoden er konsistent, og fremstiller metoder til at teste hypoteser omkring den kausale struktur. Endelig præsenterer vi heuristiske ideer til at lære opsummeringsgrafer for ikke-lineære tidsseriemodeller.

\chapter*{Preface}
This thesis has been submitted in partial fulfillment of the requirements for the Ph.D.\ degree at the Department of Mathematical Sciences, Faculty of Science, University of Copenhagen. This work was written between August 2018 and August 2021 at the Copenhagen Causality Lab, Section for Statistics and Probability Theory. This research was funded by The Carlsberg Foundation. While the pandemic threw a wrench in the planned research visit abroad and resulted in approximately half of this work being written within the confines of my apartment, it has nonetheless been a great experience. 

\begin{center}
	\normalfont\usekomafont{disposition}Acknowledgments
\end{center}

First and foremost, I would like to thank my supervisor Jonas Peters. It has truly been a pleasure working under your excellent guidance. Your commitment to our projects and our frequent meetings have been invaluable. You always dropped whatever you had in your hands in order to consider and answer my countless questions, no matter how trivial or uninteresting they may have been. 

To all my co-authors, Peter Bühlmann, Rune Christiansen, Nicola Gnecco, Phillip Mogensen, Jonas Peters, Lasse Petersen, Niklas Pfister, Rajen Shah, Nikolaj Thams, Gherardo Varando, and Sebastian Weichwald, I thank you for the fruitful collaborations, exciting discussions, and uplifting company. To  all my colleagues at the department, thank you for making my time at the department enjoyable. I thank Steffen Lauritzen for helpful discussions about various mathematical problems. I thank all my teachers, in particular Ernst Hansen, Thomas Mikosch and Anders Rønn-Nielsen, who taught me the foundations on which this thesis is written. To my friend Mads Raad, thank you for almost ten years of mathematical null-set discussions. 

To all of my friends, who time and time again have been told that I was too busy to hang out, thank you for never stopping to care. I thank my friends and family for their encouraging words and for always taking an interest in my work. To my mother and father, who inspired me to push my boundaries and helped me realize the fun and beauty of mathematics, thank you for your unconditional support and love.\par 
\vspace{1cm} \par
\hfill Martin Emil Jakobsen \par
\hfill August, 2021 \par
\vspace{1cm}
\newpage
The thesis has been edited and minor typographical errors has been corrected in agreement with the official guidelines prior to printing. A version of this thesis containing additional corrections can be found at \texttt{http://arxiv.org/a/jakobsen\_m\_1}. \par 
\vspace{1cm} \par
\hfill Martin Emil Jakobsen \par
\hfill October, 2021 \par
\vspace{1cm}
This online version of the thesis contains additional corrections. \par 
\vspace{1cm} \par
\hfill Martin Emil Jakobsen \par
\hfill October, 2021 \par
\newpage 
\chapter*{Summary of Contributions}
\addcontentsline{toc}{chapter}{Contributions}
This thesis consists of one introductory and four main chapters. The main chapters aim to advance various areas of research within the field of statistical causal modeling. \Cref{ch:introduction} contains a general introduction to causal modeling and reasoning in the mathematical framework of statistical causal models. We, furthermore, introduce the research topics of later chapters and discuss and summarize our contributions in more detail. The main chapters and their corresponding appendices consist (up to minor corrections and aesthetic modifications) of previously published, forthcoming, or soon-to-be-submitted papers. The four main chapters correspond to the following papers:

\nobibliography*
\begin{enumerate}[label=\textbf{Chapter \arabic*}:, align=left,itemsep=1em]
	\setcounter{enumi}{1}
	\item \bibentry{jakobsen2020distributional}.
	\item \bibentry{TPAMI}.
	\item \bibentry{TREES}.
	\item \bibentry{PMLR}.
\end{enumerate}

\Cref{ch:PULSE} proposes a novel estimator, called the p-uncorrelated least squares estimator (PULSE), for linear causal effects in instrumental variable (IV) setups. The PULSE can be viewed as a data-dependent mean squared prediction error regularization of the two-stage least squares estimator. We prove that the estimator is consistent, and through simulations studies, we show that in, e.g., weak instrument settings, it is MSE superior to other competing IV causal effect estimators. Furthermore, we establish a connection between K-class estimators from econometrics and the recently proposed anchor regression estimators from the field of out-of-distribution generalizing prediction methods. Prediction methods are said to be distributionally robust (or out-of-distribution generalizing) with respect to a class of test distributions if it minimizes the worst-case risk over said class. We show that K-class estimators are  distributionally robust prediction methods with respect to bounded interventions on exogenous system variables.  

In \Cref{ch:Generalization}, we propose a general framework for analyzing distributional robustness with respect to test distributions generated by interventions. We provide sufficient conditions for out-of-distribution generalization and present several impossibility results showing the necessity of certain conditions. We propose a nonlinear instrumental variable estimator that uses the previously mentioned data-dependent mean squared prediction error regularization. A simulation study shows that it, in specific setups, is MSE superior to various state-of-the-art nonparametric  instrumental variable estimators.

In \Cref{ch:trees}, we contribute to the field of causal structure learning. We propose a method for learning the causal structure of systems with directed trees as causal graphs. We strengthen established identifiability results of causal graphs for restricted structural causal models. Furthermore, we provide an alternative analysis that proves that for Gaussian noise models, the identifiability of the causal graph is a purely local property of the underlying model. Our learning method does not require heuristic optimization algorithms to recover the causal graph, something that plagues virtually all structure learning methods that do not search for Markov equivalent structures. Furthermore, we prove consistency in an asymptotic setup with decreasing identifiability. We propose a method for testing causal substructure hypotheses. The proposed method has asymptotic family-wise error rate control that remains valid post-selection.

\Cref{ch:pmlr} presents the approaches for learning summary graphs of time-series that won the NeurIPS Causality 4 Climate competition. We articulate our heuristic learning approaches and discuss artifacts of simulated DAG models. \\\\

\tableofcontents

\mainmatter

\chapter[Introduction]{Introduction} \label{ch:introduction}

In many applications, we are interested in reasoning about the behavior of a stochastic system that is affected by external manipulation. For example, in a prediction setup, we may anticipate future external manipulation of the system of interest, such that differences emerge between training and test distributions. Alternatively, we may be interested in the expected changes to a system when we intervene (apply external manipulation) on a system variable. Statistical and probabilistic models are insufficient for such purposes, as they do not possess the formal language and tools to quantify such changes. For such purposes, we need to consider statistical causal models. These are statistical models embedded with causal assumptions that allow us to model and reason about how external manipulation affects the behavior of stochastic systems.

This chapter serves as an introduction to causal modeling and inference. We discuss certain fundamental causal concepts and problems, which hopefully will ease the reading of later chapters for the causally uninitiated reader. We summarize the contributions of the later chapters and explain how they fit within  established research in the statistical causal literature. 

In \Cref{Intro_Sec_CausalModels}, we discuss the difference between the statistical and causal models and introduce some graph terminology used in later chapters. Furthermore, we define  structural causal models and introduce the concept of interventions in connection with the assumption of autonomy.  In \Cref{sec:Intro_CausalInference}, we discuss the general difficulties with causal inference and explain the necessity of unfalsifiable causal assumptions when inferring causal quantities from observational data. \Cref{sec:LearningGraphsIntro} introduces the independence-based (also called constraint-based) and score-based approaches to causal structure learning and discusses the causal assumptions these approaches need. \Cref{sec:INTRO_LearningCausalEffects} introduces the concept of causal effects. Here, we discuss how sufficient knowledge of the underlying causal structure  enables the inference of causal effects  from observational data. We also introduce the instrumental variable method for inferring causal effects in the presence of hidden variables. In \Cref{sec:LearningGeneralizingFunctionsIntroduction}, we introduce the concept of generalizing prediction functions.

\section{Causal Models}\label{Intro_Sec_CausalModels}

Causal or statistical causal models are enhanced statistical (probabilistic) models which first and foremost specify a probability distribution over a system of random variables exactly as regular statistical models do. Furthermore, these models are enhanced with a preconceived notion of how the system acts under external manipulation. We further highlight the fundamental differences between statistical and causal models in the next section. 

In the rapidly increasing literature on statistical causal modeling, different frameworks exist for defining and manipulating causal models.  Some of the more popular frameworks are structural causal models \citep[][]{Pearl2009,Peters2017}, causal graphical models \citep[][]{spirtes2000causation}, and the potential outcomes framework \cite[][]{rubin1974estimating,Rubin2005}. They all render interventional and counterfactual questions well-defined, but since their construction differs, the underlying causal assumptions needed to infer answers to such questions also differs. Thus, depending on the application, one framework may present the causal assumptions in a manner that is more easily digested compared to other frameworks. In this thesis, we work under the framework of structural causal models. We define these models formally in \Cref{sec:IntroStructuralCausalModels}.

\subsection{Statistical and Causal Models}
First, consider a statistical model over the random variables $X$ and $Y$. For example, a typical specification of the association between $X$ and $Y$ in a linear regression model is given by
\begin{align} \label{eq:IntroductionThesisStatisticalModel}
	Y = \gamma X + \ep,
\end{align}
for some $\gamma\in \R$ with $X$ and $\ep$ being mutually independent standard normal distributed random variables. This statistical model specifies a simultaneous distribution over $(X,Y)$  given by 
\begin{align*}
	\begin{pmatrix}
		X \\ Y
	\end{pmatrix} \sim \cN\left( \begin{pmatrix}
		0\\0
	\end{pmatrix}, \begin{pmatrix}
		1 & \gamma \\
		\gamma & 1+\gamma^2
	\end{pmatrix} \right).
\end{align*}
Given independent and identically distributed (i.i.d.) data generated in accordance with the above specified statistical model, we may consistently estimate the statistical parameter $\gamma$ by, for example, the ordinary least squares estimator. Knowledge of the statistical model and the statistical parameter $\gamma$ fully specifies the simultaneous distribution over $(X,Y)$, allowing us to derive predictions for new i.i.d.\ observations. For example, we may derive the probability that $Y$ is positive given that we have observed that $X$ is positive, or the conditional expectation of $Y$ given an observed value of $X$, i.e.,  $E[Y|X=x] = \gamma x$.


The specification of the statistical model in \Cref{eq:IntroductionThesisStatisticalModel} may look as if $Y$ is generated by a process that adds noise to $\gamma X$. In which case, a natural interpretation is that if we were to increase $X$ artificially, we would see an increase in $Y$ if $\gamma$ is positive. Such interpretations are not valid as a statistical model only specifies an observational distribution.
More specifically, the above interpretation relies on a causal assumption of the observed system, i.e., a causal physical mechanism that outputs $Y$ from the input $X$ and that this physical mechanism does not change when artificially intervening on the input $X$. 

Statistical causal models give us the language and tools to specify and analyze such extended interpretations of statistical models. However, it is worth noting that causal interpretations always require causal assumptions. Without agreeing to certain unfalsifiable causal assumptions, one can never infer causal effects or relations from observational data. 

\subsection{Graphs}
Before we define structural causal models, we introduce some graph terminology used throughout this thesis. Graphs are vital in causal reasoning and inference; they allow us to analyze and visualize the causal relations between variables in a system.  

A \textit{directed graph} $\cG=(V,\cE)$ consists of $p\in \N_{>0}$ vertices (nodes)   $V=\{1,\ldots,p\}$ and a collection of directed edges $\cE\subset \{(j\to i) \equiv (j,i): i,j\in V, i\not = j\}$. We let $\PAg{\cG}{i}:= \{v\in V: \exists (v,i)\in \cE\}$ and $\CHg{\cG}{i} :=\{v\in V : \exists (j,v)\in \cE\}$ denote the \textit{parents} and \textit{children} of node $i\in V$ and we define root nodes $
\mathrm{rt}(\cG) := \{v\in V: \PAg{\cG}{i} =\emptyset \}$ as nodes with no parents (that is, no incoming edges). Two nodes are \textit{adjacent} if there exists an edge between them and  a \textit{$v$-structure} consists three nodes where one node is a child of two non-adjacent nodes.
A \textit{path} in $\cG$ between two nodes $i_1,i_k\in V$ consists of a sequence $(i_1, i_2,..., i_k)$ of adjacent nodes, i.e., a sequence of pairs of nodes such that for all $j \in \{1, \ldots, k-1\}$, we have either 
$(i_j \to i_{j+1}) \in \cE$
or
$(i_{j+1} \to i_{j}) \in \cE$.
A \textit{directed path} in $\cG$ between two nodes $i_1,i_k\in V$ consists of a sequence $(i_1, i_2,..., i_k)$ of pairs of nodes such that for all 
$j \in \{1, \ldots, k-1\}$, we have $(i_j \to i_{j+1}) \in \cE$. Furthermore, we let $\ANg{\cG}{i}$ and $\DEg{\cG}{i}$ denote the \textit{ancestors} and \textit{descendants} of node $i\in V$, consisting of all nodes $j\in V$ for which there exists a directed path to and from $i$, respectively. 

A \textit{directed acyclic graph} (DAG) is a directed graph that does not contain any directed cycles, i.e., directed paths visiting the same node twice.
We say that a graph is \textit{connected} if a path exists between any two nodes. A \textit{directed tree} is a connected DAG in which all nodes have at most one parent. More specifically, every node has a unique parent except the root node, which has no parent. The root node $\root{\cG}$ is the unique node such that there exists a directed path from $\root{\cG}$ to any other node in the directed tree. A directed tree is also called an \textit{arborescence}, a \textit{directed rooted tree} and a \textit{rooted out-tree} in graph theory.   We 	let $\cT_p$ denote the set of all directed trees of $p\in \N_{>0}$ nodes.  A graph $\cG'=(V',\cE')$ is a \textit{subgraph} of another graph $\cG=(V,\cE)$ if $V'\subseteq V$, $\cE'\subseteq \cE$. A subgraph is \textit{spanning} if  $V'=V$.

An \textit{undirected graph} $\cG=(V,\cE)$ consists of $p\in \N_{>0}$ nodes (vertices)   $V=\{1,\ldots,p\}$ and a collection of undirected edges $\cE\subseteq \{ \{j,i\}: i,j\in V, i\not = j\}$ and a \textit{partially directed graph} or \textit{mixed graph} $\cG=(V,\cE_u,\cE_d)$ has both a collection of undirected edges $\cE_u\subseteq  \{ \{j,i\}: i,j\in V, i\not = j\}$ and a collection of directed edges $\cE_d \subseteq \{(j,i): i,j\in V, i\not = j\}$. 
\subsubsection{D-separation} \label{sec:IntroductionDseparation}
Pearl's d-separation \citep[][]{Pearl2009} is a graphical notion that will allow us to deduce conditional and unconditional independence statements concerning system variables generated by a structural causal model by analyzing the corresponding causal graph. For now, we introduce it as a purely graphical definition concerning directed acyclic graphs. Suppose that we have a directed acyclic graph $\cG=(V,\cE)$.  We say that a path $(i_1,...,i_k)$ in $\cG$ between two nodes $i_1$ and $i_k$ is \textit{blocked} by a collection of nodes $C \subseteq V\setminus \{i_1,i_k\} $ if either
\begin{enumerate}
	\item[(i)] there exists $m\in\{2,...,k-1\}$ such that $i_m\in C$ and the path contains a subpath of the form $i_{m-1} \to i_m \to i_{m+1}$, $i_{m-1} \leftarrow i_m \leftarrow i_{m-1}$ or $i_{m-1} \leftarrow i_{m} \to  i_{m+1}$, or 
	\item[(ii)] there exists $m\in\{2,...,k-1\}$ for which neither the node $i_m$ nor any of its descendants are in $C$, i.e., $(\{i_m\}\cup \DEg{\cG}{i_m})\cap C=\emptyset$, and the path contains the subpath $i_{m-1} \to i_m \leftarrow i_{m+1}$.
\end{enumerate}
\begin{definition}[d-separation]
	Consider a directed acyclic graph $\cG=(V,\cE)$. Let $A,B,C\subset V$ be three distinct subsets of nodes. $A$ and $B$ are \textit{d-separated} by $C$ in $\cG$, written $
	A \dsep{\cG} B \, | \, C $ 
	if and only if all paths between any two nodes in $A$ and $B$ are blocked by $C$.
\end{definition}
\subsection{Structural Causal Models} \label{sec:IntroStructuralCausalModels}
Causal models allow one to specify an observational probability distribution over a system of variables (i.e., a statistical model) but also enable one to reason about interventional and counterfactual questions. This section introduces models with these properties from the framework of structural causal models (SCMs). Later in this section, we  introduce interventions, but we refrain from introducing counterfactual reasoning since this thesis does not contribute to this area of research.

\begin{definition}[Structural causal models] \label{def:acylicSCMIntrodutction} A structural causal model $M = (Q, \mathcal S)$ of dimension $p\in \N_{>0}$ consists of a noise distribution $Q$ on $\R^p$ with mutually independent marginals and $p$ structural assignments $\mathcal S$:
	\begin{align*}
		1\leq i \leq p : \quad 	X_i := f_i(X_{\PA{i}},N_i), 
	\end{align*}
	where $X_{\PA{i}} \subseteq X=(X_1,...,X_p)$ denotes the parents or direct causes of $X_i$ and $N=(N_1,...,N_p)\sim Q$. \end{definition}
The collection of functions $(f_i)_{1\leq i \leq p}$ and variables $N=(N_1,...,N_p)$,  present in the structural assignments, are called the causal functions and the noise innovations, respectively. Structural causal models are also known as structural equation models or simultaneous equation models in statistics and econometrics \citep[applied with varying degrees of causal interpretation, see, e.g.,][]{Pearl2012}. 

We distinguish between two fundamentally different SCM structures; those that are cyclic and those that are acyclic. Whether or not an SCM is cyclic or acyclic plays an essential role in constructing a solution, i.e., the induced random system of variables satisfying the structural assignments.
\begin{definition}[Acyclic and cyclic SCMs]
	A $p$-dimensional SCM $M=(Q,\cS)$ is acyclic if there exists a causal order $\pi$, i.e., a permutation $\pi:\{1,...,p\}\to \{1,...,p\}$, satisfying $\pi(j)<\pi(i)$ whenever $j\in\PA i$ for all $1\leq i \leq p$. An SCM called cyclic if it is not acyclic.
\end{definition}
Let $M=(Q,\cS)$ be an SCM and let  $N:(\Omega,\bF) \to \R^p$ and $X:(\Omega,\bF)\to \R^p$ be defined on a common probability space $(\Omega,\bF,P)$ such that $N\sim Q$. We say that the pair $(X,N)$ solves $M$ if 
\begin{align*}
	 X \, \stackrel{\text{a.s.}}{=} f(X,N),
\end{align*}
where $f(x,n):=(f_1(x_{\PA 1},n_1),...,f_p(x_{\PA p},n_p))$ are the structural assignments $\cS$ of $M$. 
We say that a random vector $X$ is induced or generated by an SCM $M=(Q,\cS)$ whenever there exists an $N\sim Q$ such that $(X,N)$ solves the SCM. An SCM-induced random vector is therefore only uniquely defined up to a $P$-null set.

It is, in general, not guaranteed that solutions exist to cyclic a SCM; see \cite{bongers2021foundations} for further information on the theoretical foundations of cyclic SCMs. An acyclic SCMs $M = (Q, \mathcal S)$ is, however, always solvable. 
 Suppose that we have a random vector $N=(N_1,...,N_p):(\Omega,\bF) \to \R^p$ with $N\sim Q$ and that $\pi$ is the causal order of the acyclic SCM. We can now define the random vector $X:(\Omega,\bF) \to \R^p$ in increasing order of $i\in\{1,...,p\}$, 
\begin{align*}
	X_{\pi^{-1}(i)}:= f_i(X_{\PA{\pi^{-1}(i)}},N_i),
\end{align*}
which by definition solves the SCM.
\begin{example} \label{ex:acylicSCM} Consider the acyclic structural causal model given by a noise innovation distribution $Q$ and structural assignments 
	\begin{align*}
		X_1 &:= f_1(N_1),\\
		X_2 &:= f_2(X_1,N_2), \\
		X_3 &:= f_3(X_1,X_2,N_3),
	\end{align*}
	where $N=(N_1,N_2,N_3)\sim Q$.  This SCM has a causal order given by $$(\pi(1),\pi(2),\pi(3))=(1,2,3),$$ so we can given a noise innovation $N$ iteratively define $X_1,X_2$ and finally $X_3$. 
\end{example}
We define the induced or observational distribution of a solvable SCM $M$ by the push-forward measure $P_X=X(P)$ on $\R^p$ for any solution $X$. Sometimes we also denote the observational distribution by $P_M$. The observational distribution is always uniquely defined.

\begin{example}\label{ex:introduction_linearassignments} Consider the SCM of \Cref{ex:acylicSCM}. Now suppose that $Q$ denotes the 3-dimensional standard multivariate normal distribution $N \sim Q = \cN(0,I_3)$ and that the structural assignments are linear and given by
	\begin{align*}
		X_1 & :=f_1(N_1) \equiv N_1, \quad 	X_2 :=f_2(X_1,N_2) \equiv \alpha X_1 + N_2, \\
		 X_3 &:= f_3(X_1,X_2,N_3)\equiv \gamma X_1 + \beta X_2 + N_3.
	\end{align*}
By subsitution we find that  $
		X_1  = N_1,$ $ 
		X_2  = \alpha N_1 + N_2$ and $ 
		X_3   = (\gamma+\beta\alpha) N_1 + \beta N_2 + N_3,$
from which the induced distribution of $M$ is easily found to be given by  $(X_1,X_2,X_3) 	\sim \cN\lp0, \Sigma \rp$ where 
	\begin{align*}
		\Sigma := \begin{pmatrix}
			1 & \alpha & \gamma+\beta \alpha \\
			\alpha & \alpha^2 +1 & \alpha(\gamma + \beta \alpha) + \beta \\
			\gamma+\beta \alpha & \alpha(\gamma + \beta \alpha) + \beta & (\gamma+\beta \alpha)^2 + \beta^2 + 1
		\end{pmatrix}.
	\end{align*}
\end{example}
 Henceforth, we assume that all solvable structural causal models have structurally minimal assignments. That is, for any structural assignment $X_i := f_i(X_{\PA{i}},N_i)$ there does not exist a $j\in \PA{i}$ and a measurable map $\tilde f_i$ such that $
f_i(X_{\PA{i}},N_i) = \tilde f_i(X_{\PA{i}\setminus \{j\}},N_i)$ almost surely.

\begin{figure}
	\begin{center}
\includegraphics[scale=1]{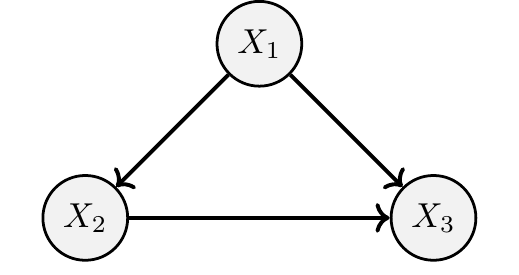}
	\end{center}
	\caption{The causal graph of the common confounder structural causal model in \Cref{ex:acylicSCM}.}
	\label{fig:introduction_commonconfoundersetup}
\end{figure}

\begin{definition}[Causal graph]
	The causal directed graph $\cG=(V,\cE)$ of an SCM $M = (Q,\mathcal S)$ is given by the vertex set $V:=\{1,...,p\}$ and direct edges drawn from each $j\in\PA i$ to $i$ for all $i\in V$, i.e.,
	\begin{align*}
		\cE = \{(j\to i): i\in V, j\in \PA{i}\}.
	\end{align*}
	That is, the causal graph is determined by letting $\PAg{\cG}{i}:= \PA{i}$ for all $i\in V$. 
\end{definition}
The causal graph of an acyclic SCM is, therefore, always a DAG. In \Cref{fig:introduction_commonconfoundersetup}, we have illustrated the causal graph of the acyclic structural causal model $M=(Q,\cS)$ from \Cref{ex:acylicSCM}.

%
%

In this thesis, we are mainly concerned with linear cyclic SCMs and general acyclic SCMs.  \Cref{rm:exampleLinearSCMsolIntroduction} highlights sufficient conditions for the existence and construction of solutions to linear cyclic SCMs.

\begin{example}[Linear cyclic SCMs.] \label{rm:exampleLinearSCMsolIntroduction}
	A linear cyclic SCM $M = (Q,\mathcal S)$ satisfies linear structural assignments. That is, for each $1\leq i \leq p$, the structural assignment is given by
	\begin{align*}
		X_i :=f_i(X_{\PA{i}},X_i) \equiv b_i^\t X_{\PA i} + N_i,
	\end{align*}
	for some $b_i\in \R^{|\PA i|}$. Now let $B\in \R^{p\times p}$ be a constant matrix such that $x=Bx+n$ conforms with the above structural assignments. If $\rho(B)$, the spectral radius of $B$, is strictly less than one, then we know that $(I-B)$ is invertible. Hence, $x=(I-B)^{-1}n$. Thus, given a noise innovation $N:(\Omega,\bF,P)\to \R^p$ with $N\sim Q$, define $X= (I-B)^{-1}N$ and note that $(X,N)$ solves the SCM,  since $X=BX+N$ holds $P$-almost surely.
\end{example}

For any structural causal model, the induced observational distribution satisfies the global Markov property with respect to the causal graph --- a one-way connection between the d-separation statements in the causal graph and conditional independencies in the induced distribution.
\begin{theorem}[\citealp{Pearl2009}, Theorem 1.4.1]
	Let $X=(X_{1},...,X_p)\in \R^p$ be random vector induced by an acyclic structural causal model $M$ with acyclic causal graph $\cG=(V,\cE)$. The induced distribution $P_X$ satisfies the global Markov property with respect to the causal graph. That is,
	\begin{align*}
		A\, \, \dsep{\cG} \, \,  B\, |\, C \implies 	X_A \independent X_B \, | \, X_C,
	\end{align*}
	for all disjoint subsets $A,B,C\subseteq V=\{1,...,p\}$.
\end{theorem}
Thus, the causal graph yields through $d$-separation a visual representation of conditional independence statements in the observational distribution of a structural causal model.

\subsubsection{Interventions}
\begin{figure}
	\begin{center}
	\includegraphics[scale=1]{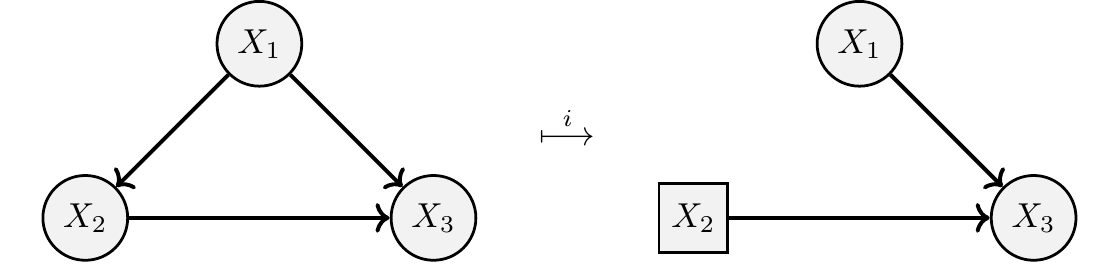}
	\end{center}
	\caption{Illustration of the original and post-intervention causal graph for the structural causal model and intervention considered in \Cref{ex:interventionIntroduction}.}
	\label{fig:introduction_commonconfoundersetupafterintervention}
\end{figure}

So far, the structural causal models only induce an observational distribution, i.e., a statistical model which only allows us to ask and answer questions about probabilistic associations. The main difference between a statistical model and a causal model is the ability to explain the behavior of a stochastic system of variables under external manipulation (intervention). In the search for a tractable behavior of systems under manipulation, one usually assumes autonomy, also called modularity, of the causal (physical) mechanisms of the system we are modeling.
\begin{assumption}[Autonomy of causal mechanisms; \citealp{Peters2017}] \label{ass:autonomyIntroduction}
	The causal generative process of a
	system’s variables is composed of autonomous modules that do not inform or influence
	each other.
\end{assumption}
The assumption of autonomous causal mechanisms yields the ability to conduct external manipulations of the generative process in selected parts of a system without affecting the generative processes of the remaining system. 
\begin{example}[Autonomy in a cause-effect system]
	Consider a bivariate cause-effect system where $X$ causes $Y$. Suppose that $f$ is the mechanism that produces $Y$ given the cause/input $X$, i.e., $
	Y:= f(X)$. 
	\Cref{ass:autonomyIntroduction} translates to independence between cause and mechanism. The assumption of autonomous causal mechanisms stipulates that any external manipulation of $X$ does not affect the mechanism $f$, which produces $Y$.
\end{example}
The assumption of autonomous causal mechanisms allows us to analyze the behavior of a system under external interventions in a tractable fashion. 
\begin{definition}
	An intervention $i$ is a map between structural causal models 
	\begin{align*}
		M = (Q, \mathcal S) \stackrel{i}{\mapsto}  (Q^i,\mathcal S^i),
	\end{align*}
	where $Q^i$ and $\mathcal S^i$ are the post-intervention noise distribution and structural assignments. We let $M(i) = (Q^i, \cS^i)$ denote the post-intervention structural causal model.
\end{definition}
In this introduction, we only concern ourselves with fairly simple interventions. Later chapters will introduce more general interventions as needed.  For example, an intervention on a single system variable amounts, by \Cref{ass:autonomyIntroduction}, to only changing the structural assignment of said variable; see \Cref{ex:interventionIntroduction} below.
\begin{example} \label{ex:interventionIntroduction}
	Consider the SCM  $M = (Q,\mathcal S)$ of \Cref{ex:acylicSCM}. Let $i$ be an intervention that randomizes $X_2$, i.e., forces it to obey a distribution $P^i$ independently of the outcome of its original direct cause $X_1$. That is, we change the structural assignments in the following way:
	\begin{center}
		\begin{tabular} {llr}
			$\mathcal S= \Bigg\{$
			\begin{tabular}{l}
				$X_1 := f_1(N_1),$\\
				$X_2 := f_2(X_1,N_2)$, \\
				$X_3 := f_3(X_1,X_2,N_3)$,
			\end{tabular}
			& $\stackrel{i}{\longmapsto}$ & $\mathcal S^i =\Bigg\{$
			\begin{tabular}{l}
				$ X_1 := f_1( N_1),$\\
				$ X_2 := \tilde N_2$, \\
				$ X_3 := f_3(X_1,X_2, N_3)$,
			\end{tabular}
		\end{tabular}
	\end{center}
	where $(N_1,N_2,N_3)\sim Q=Q_1\times Q_2 \times Q_3$ and $(N_1,\tilde N_2, N_3) \sim Q^i= Q_{1}\times P^i \times Q_{3}$.  In \Cref{fig:introduction_commonconfoundersetupafterintervention}, we have illustrated the corresponding changes to the causal graph. The edge from $X_1$ to $X_2$ is removed due to the effect breaking intervention.
\end{example}

Interventions need not break the direct link of the original causes; it can also simply change the causal mechanism which produces the variable from its causes. We denote such interventions on single system variables, say, $X_i$, by $$\mathrm{do}(X_i:=\tilde f_i(X_{\widetilde \pa(i)},\tilde N_i)),$$ where $\tilde f$ is a (possibly) new causal mechanism taking the new direct causes $\widetilde{\pa}(i)$ and noise innovation $\tilde N_i$ as inputs. For example, the intervention in \Cref{ex:interventionIntroduction} is denoted by $\mathrm{do}(X_2:=\tilde N_2)$ with $\tilde N_2 \sim P^i$. In the upcoming chapters, we use slightly different notations for intervention-induced distributions, i.e., the post-intervention simultaneous distribution of the system. For example, the intervention-induced distribution for the intervention $i=\mathrm{do}(X_2:=\tilde N_2)$ in an SCM $M$ may be denoted by
\begin{align*}
P_{M(i)}, \quad \text{or} \quad P_M^{\mathrm{do}(X_2:=\tilde N_2)}	, \quad \text{or} \quad  P^{\mathrm{do}(X_2:=\tilde N_2)},
\end{align*}
depending on whether  or not the underlying SCM $M$ and intervention $i$ is clear from the context. In the example below, we derive an intervention-induced distribution.
\begin{example}
	Consider the SCM $M=(Q,\cS)$ of \Cref{ex:introduction_linearassignments}. Suppose that we conduct the intervention $i=\mathrm{do}(X_2:=\tilde N_2)$ with $\tilde N_2 \sim \cN(0,1)$ independent from the original noise innovations of the system. The post-intervention structural assignments are now given by
		\begin{align*}
		X_1 & := N_1, \quad 	X_2 :=\tilde  N_2, \quad	X_3 := \gamma X_1 + \beta X_2 + N_3.
	\end{align*}
	Thus, $
	X_1  = N_1,$ $ 
	X_2  =  \tilde N_2$ and $ 
	X_3   = \gamma N_1 + \beta \tilde N_2 + N_3$, so the intervention-induced distribution is given by  $P_{M(i)}= \cN\lp0, \Sigma \rp$ where 
	\begin{align*}
		\Sigma := \begin{pmatrix}
			\, 1 \, & \, 0 \, & \gamma \\
			0 & 1 & \beta \\
			\gamma & \beta & \gamma^2 + \beta^2 + 1
		\end{pmatrix}.
	\end{align*}
\end{example}


\section{The Difficulties of Causal Inference} \label{sec:Intro_CausalInference}
Inferential targets in causal models can be statistical or causal quantities. For example, we may be interested in statistical targets, i.e., quantities defined in terms of the joint distribution of the system variables. Statistical targets include, for example, the correlation between  variables, conditional probabilities, or conditional expectations between certain  variables. Causal targets are non-statistical quantities defined in terms of a causal model \citep{Pearl2009}. Common causal targets include the causal graph (or parts thereof, e.g., the direct causes of a specific variable), causal effects, and general post-interventional probabilistic quantities of system variables, i.e., the post-intervention distribution or a derivative thereof.

However, as causal quantities are not defined in terms of the system's observational distribution, their inference from observational data will instead rely on causal assumptions about the system of interest. Such assumptions are, by definition, not  falsifiable by observational data and therefore purely rests on the practitioner's expert judgment \citep{Pearl2009}.

\begin{figure}
	\begin{center}
	\includegraphics[width=\textwidth]{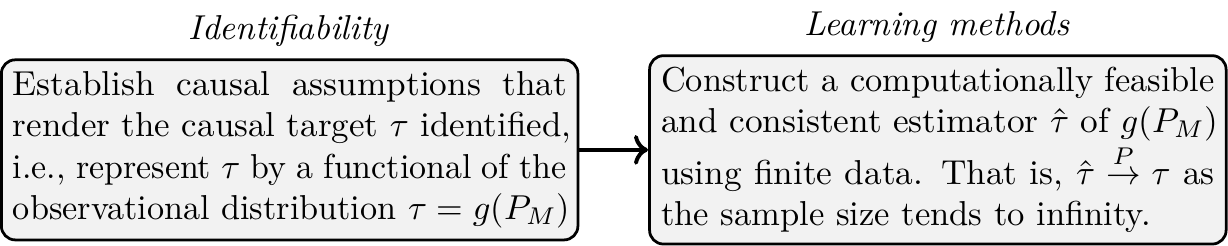}
	\end{center}
	\caption{Flowchart of causal inference from observational data.}
	\label{fig:introduction_flowchart}
\end{figure}

There are two main aspects to learning causal targets: identifiability and learning methods; see the flowchart in \Cref{fig:introduction_flowchart}. First, we have the aspect of identifiability; see \Cref{sec:Introduction_Identifiability}. Here we are concerned with the theoretical ability to infer the target from the observational distribution of the system. Second, in the affirmation of identifiability, we have the aspect of constructing learning methods (identification); see \Cref{sec:Introduction_LearningMethods}. Here we are concerned with estimating the causal target from finite data, similar to  regular inference of statistical quantities.

%

\subsection{Identifiability} \label{sec:Introduction_Identifiability}
In practice, most causal targets can be recovered by conducting specific interventions in a system and analyzing the observed changes. For example, it is possible to recover the average treatment effect of a drug by conducting a randomized controlled trial \citep{Peirce83} where one randomly assigns a patient the treatment or a placebo. The random assignment can be seen as an intervention in which the treatment indicator (i.e., whether the patients get the drug or a placebo) is externally manipulated to follow the outcome of a binary random variable that is independent of other system variables (e.g., patient covariates, etc.). However, due to either ethical, monetary or practical reasons, we may not be able to conduct the preferred system interventions that would enable us to quantify the causal targets. In this thesis, we are mainly concerned with the latter scenario where interventions are not possible.

In theory, there could be several distinct data-generating processes (causal models) that are observationally equivalent (induces identical observational distributions) but differ on the causal quantity of interest.  Hence, an essential aspect of causal modeling is specifying causal assumptions that allow us to infer the causal targets from the observational distribution alone. A causal target is said to be identified if we can theoretically infer it from the observational distribution. 

A lot of causal targets become identified once the causal graph of the causal model is known. Thus, we either have to resort to expert judgment on the causal structure or infer the structure from data. In \Cref{sec:LearningGraphsIntro}, we highlight some standard structure learning methods and detail the causal assumptions they rely on.

%
%

\subsection{Learning Methods} \label{sec:Introduction_LearningMethods}

The next problem in causal inference is inferring or learning the causal target of interest from finite data in a consistent and computationally feasible way. In the affirmation of identifiability, we know that the observational distribution uniquely determines the causal target.  Thus, in theory, we could infer the causal target given complete knowledge of the observational distribution. 

Under appropriate causal assumptions, some causal targets are given by quantities of the observational distribution (distributional features) for which inference has been well-studied in the statistical literature, e.g., conditional expectations or linear regression coefficients. In such cases, inference can be achieved by simply applying established statistical inference methods. However, sometimes the causal target is not a commonly studied quantity of the observational distribution. In these cases, inference requires new methods with accompanying theoretical large sample guarantees. 

\section{Learning Causal Graphs} \label{sec:LearningGraphsIntro}
The causal graph of a causal model is often of interest to practitioners due to the intrinsic value of knowing what system components cause a specific variable. Alternatively, one is interested in the causal structure since other causal targets become identified from the observational distribution once the causal graph is known; see, e.g., \Cref{sec:INTRO_LearningCausalEffects}.

We focus on the problem of inferring the causal structure from observational data. However, as we have previously mentioned, inference of causal quantities from observational data necessitates causal assumptions on the system of interest. That is, we need causal assumptions that make it theoretically possible to infer the causal graph of an acyclic SCM from its induced distribution.
%

Standard structure learning methods are classified as independence-based (also known as constraint-based), score-based, or mixed.  Structure learning methods that are independence-based rest on the nonparametric causal assumption of faithfulness; see \Cref{def:introFaithfulness}. Faithfulness renders parts of the causal structure identified through the independence constraints encoded in the observational distribution. On the other hand, score-based methods rest on causal assumptions on the causal mechanisms and noise innovations of the system of interest.

In \Cref{sec:IndependenceBasedStructureLearningIntro}, we introduce the causal assumptions for independence-based structure learning and briefly discuss established methods for inference. \Cref{sec:ScoreBasedStructureLearningIntro} introduces score-based approaches to causal structure learning, which is also the topic of \Cref{ch:trees}.

\subsection{Independence-based Structure Learning} \label{sec:IndependenceBasedStructureLearningIntro}

Independence-based structure learning methods infer parts of the causal graph by utilizing (conditional) independence constraints encoded in observational distribution. We have previously seen that the induced distribution of an acyclic SCM is Markov with respect to the causal graph. However, for learning the structure itself, this is a useless property as, for example, any SCM induced distribution is also Markov with respect to the fully connected graph. In general, without further causal assumptions, the (conditional) independence constraints encoded in the observational distribution do not yield any causal graph information. This problem leads us to the fundamental causal assumption on which independence-based structure learning methods rests; the assumption of faithfulness with respect to the causal graph.
\begin{definition}[Faithfulness] \label{def:introFaithfulness}
	Let $X=(X_1,...,X_p)\in \R^p$ be a random vector with distribution $P_X$ and let $\cG$ be a DAG with nodes $V=\{1,...,p\}$. The distribution $P_X$ is said to be faithful with respect to the graph $\cG$ if
	\begin{align*}
		X_A \independent X_B | X_C  \implies  A\dsep{\cG} B\, |\, C
	\end{align*}
	for all disjoint subsets $A,B,C\subseteq V=\{1,...,p\}$.
\end{definition}

Thus, if we assume that the induced distribution of an acyclic SCM is faithful to the causal graph, then by the global Markov property, we have a one-to-one correspondence between $d$-separations in the causal graph and conditional independence constraints encoded by the induced distribution. Independence-based structure learning methods exploit this correspondence: utilizing conditional independence testing, one draws inference on conditional independence statements that allow one to draw inference about the causal graph through the faithfulness assumption.

Faithfulness implies causal minimality (\citealp{Peters2017}, Proposition 6.35), i.e., if $P_X$ is faithful with respect to the causal graph $\cG$, then  $P_X$ it is not Markov with respect to any proper subgraph of $\cG$. Faithfulness is a causal assumption that is not satisfied in general; see \Cref{ex:faithfulnessIntroduction} below.

\begin{example}\label{ex:faithfulnessIntroduction}
	Consider the linear Gaussian SCM of \Cref{ex:introduction_linearassignments}, with causal graph is illustrated in \Cref{fig:introduction_commonconfoundersetup}. The structural assignments are given by
	\begin{align*}
		X_1 := N_1, \quad  X_2:= \alpha X_1+N_2, \quad X_3 := \gamma X_1 + \beta X_2+N_3,
	\end{align*}
	where $N=(N_1,N_2,N_3)\sim \cN(0,I)$. If $\alpha  \beta = - \gamma$, then $X_2 \independent X_3$. However,  $X_2$ is not $d$-separated from $X_3$ given the empty set, so faithfulness is not satisfied with respect to the causal graph.
\end{example}

\begin{figure}
	\begin{center}
	\includegraphics[]{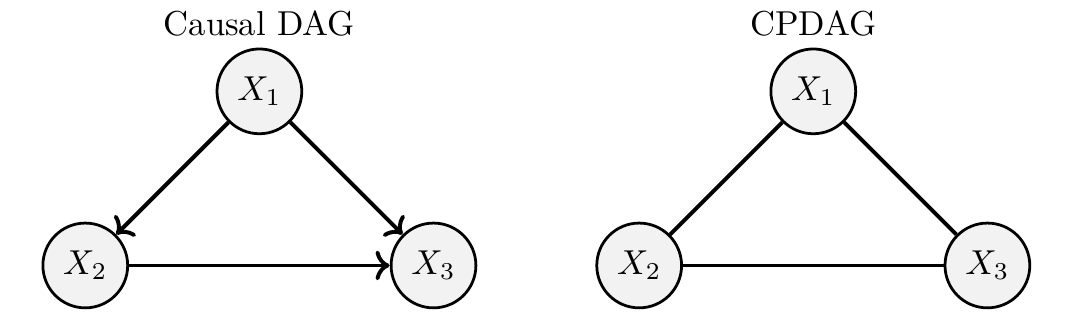}
	\end{center}
	\caption{The causal graph from the SCM of \Cref{ex:acylicSCM} and the corresponding CPDAG representing its Markov equivalence class.}
	\label{fig:introduction_CPDAG}
\end{figure}

Let us discuss what parts of the causal structure the assumption of faithfulness identifies. That is, we will discuss what it entails that we can infer all $d$-separation statements of a causal graph. 
%
To this end, we say that two graphs $\cG$ and $\tilde \cG$ are Markov equivalent if every probability distribution that is globally Markov with respect to $\cG$ is also globally Markov with respect to $\tilde \cG$ and vice versa. The Markov equivalence class (MEC) of a graph $\cG$, $\mathrm{MEC}(\cG)$, consists of all graphs that are Markov equivalent to $\cG$. It has been shown that $\mathrm{MEC}(\cG)=\{ \tilde \cG \text{ is a DAG}: \tilde \cG \text{ and } \cG \text{ share the same d-separations}\}$  \citep{verma1990causal}, so faithfulness implies that the Markov equivalence class of the causal graph is identified.
Finally, the following theorem quantifies the shared structure of all DAGs in the Markov equivalence class.
\begin{theorem}[\citealp{VermaEquivalence90}]
	Two DAGs are Markov equivalent if and only if they share the same skeleton and $v$-structures.
\end{theorem}
Thus, it is possible to represent the Markov equivalence class of a DAG $\cG$ by a unique partially directed acyclic graph (PDAG) known as the \textit{completed} PDAG (CPDAG) with the skeleton and directed edges that make up $v$-structures shared by all members. In \Cref{fig:introduction_CPDAG}, we have illustrated the causal graph and the corresponding CPDAG representing its Markov equivalence class of the SCM from \Cref{ex:acylicSCM}.

As for learning the CPDAG, we can use the popular PC-algorithm \citep{spirtes2000causation}. The contributions in this thesis do not add to the literature on independence-based structure learning, so we refer to \cite{spirtes2000causation} for further details on the algorithm. Nevertheless, given oracle knowledge on conditional independence statements, the PC-algorithm recovers the CPDAG whenever faithfulness is satisfied. However, when inferring the CPDAG from finite data, the conditional independence statements have to be inferred by successive conditional independence tests. One usually chooses a fixed significance level for the tests, but due to the successive testing, one loses the error quantification of the method as a whole. Furthermore, conditional independence tests can not have power against any alternative \citep{shah2020hardness} unless specific distributional assumptions are made, such as joint Gaussianity. Type I errors of the conditional independence tests can lead to the removal of causal edges and the inclusion of non-causal edges in the resulting CPDAG \citep{spirtes2000causation}.

\subsection{Score-based Structure Learning} \label{sec:ScoreBasedStructureLearningIntro}
Score-based approaches to causal structure learning use (parametric) assumptions on the structural causal model that allow for the construction of a scoring function for causal structures. That is, in the affirmation of identifiability of the causal graph (or parts thereof), we define a (population) score function $\ell$ 
that only attains its minimum in the causal graph
\begin{align} \label{eq:ScoreFunctionMinimizationIntroductionchapter}
	\cG = \argmin_{\tilde \cG \,: \,\tilde  \cG \text{ is a DAG}} \ell(\tilde \cG).
\end{align}
The greedy equivalence search \citep[GES,][]{chickering2002optimal} assumes faithfulness which renders the MEC identified.  Under the additional assumption of joint Gaussianity of the observed distribution, GES minimizes a BIC-penalized likelihood score function directly on the space of Markov equivalence classes. 

Causal system assumptions that guarantee identifiability of the causal graph itself have also been studied. For example, in SCMs with additive Gaussian noise and nonlinear causal functions, the causal graph is identified; see the introduction of \Cref{ch:trees} for an overview. However, in the pursuit of the causal graph, we stumble onto new computationally problematic issues. Even though the optimization problem in \Cref{eq:ScoreFunctionMinimizationIntroductionchapter} is guaranteed to have a unique minimum, the optimization problem is a combinatorial problem with a search space cardinality that grows super-exponentially in the number of system variables. Thus, for even moderately large systems, brute-force optimization (exhaustive search) becomes computationally infeasible. 

At the current state of the literature, no optimization procedure guarantees to solve the problem with computationally feasible time complexity for large systems. However, several heuristic optimization procedures have been proposed. For example, \cite{CAM} propose a greedy search technique on the space of DAGs, and \cite{zheng2018dags} propose an equivalent continuous albeit non-convex representation of the optimization problem in \Cref{eq:ScoreFunctionMinimizationIntroductionchapter}. These approaches do not guarantee to recover the causal graph. For example, the non-convex continuous optimization problem representation necessitates naive optimization approaches with no guarantees of not getting stuck in a local minimum. Moreover, it is currently being discussed whether the seemingly remarkable performance in simulation studies of \cite{zheng2018dags} is due to the exploitation of simulated DAG artifacts rather than successful naive optimization; see \cite{reisach2021beware} and \Cref{Intro:sec:Artifacts} below. In \Cref{sec:WhenGreedySearchFail}, we show an example where the greedy search of \cite{CAM} fails. Hence, there is currently no practical method that guarantees the recovery of the actual causal graph with probability tending to one in the large sample limit.
\subsubsection{Causal Structure Learning for Directed Trees}

In \Cref{ch:trees}, we take a slightly different approach to the computational problems associated with recovering the actual causal graph in score-based approaches. Instead of proposing another heuristic optimization procedure, we look at what relaxations in the system complexity allow for exact score-function minimization.

In particular, we restrict our attention to less complex systems with causal graphs given as directed trees and additive noise. While brute-force minimization over the space of directed trees is still computationally infeasible, i.e., the search space still grows super-exponentially in the system size, we show that the optimization is possible with polynomial time complexity. More specifically, we show that Chu--Liu--Edmonds' algorithm \citep[proposed independently by][]{chuliu1965,edmonds1967optimum} from graph theory solves the optimization problem. 

We show that the proposed method, called causal additive trees (CAT), is consistent under weak conditions. Moreover, due to the reasonably simple causal structure, we provide inference results to test causal substructure hypotheses. Our proposed hypothesis testing procedure retains its level-guarantees under post-selection hypothesis generation and multiple testing. 
Furthermore, we investigate the identifiability gap, i.e., the minimum score difference between the causal graph and any alternative graph. For Gaussian noise innovations, we provide a lower bound that depends only on local dependence properties. That is, the identifiability of the causal graph reduces to a purely local property for Gaussian additive noise models.

\subsubsection{When Greedy Searches Fail} \label{sec:WhenGreedySearchFail}
Greedy search techniques do not, in general, come with theoretical guarantees. We now present an example where the greedy search of  \cite{CAM}, called CAM, consistently fails to recover the causal graph, while our method CAT successfully recovers the causal graph as the sample size increases.   The following model is taken from  \cite{Peters2022}. Consider the following three node Gaussian additive structural causal model with causal graph $(X\to Y \to Z)$:
\begin{align} \label{eq:3nodesetup}
	X := N_X, \quad 
	Y := \frac{X^3}{\Var(X^3)} + N_Y, \quad 
	Z := Y+ N_Z,
\end{align}
where $N_X\sim \cN(0,1.5)$, $N_Y \sim \cN(0,0.5)$ and $N_Z \sim \cN(0,0.5)$ are mutually independent. Our method CAT has two variants: CAT.G and CAT.E using a Gaussian and entropy scoring function, respectively.  We simulate data from this model and estimate the causal graph by CAT.G, CAT.E and CAM. \Cref{fig:boxplot_3node_SHD} illustrates the results.  Even with increasing sample size, CAM does not converge to the correct answer. The reason is that it selects the wrong edge in the first step of the greedy search algorithm.
\begin{figure}[H]
	\begin{center}
		\includegraphics[width=\textwidth-20pt]{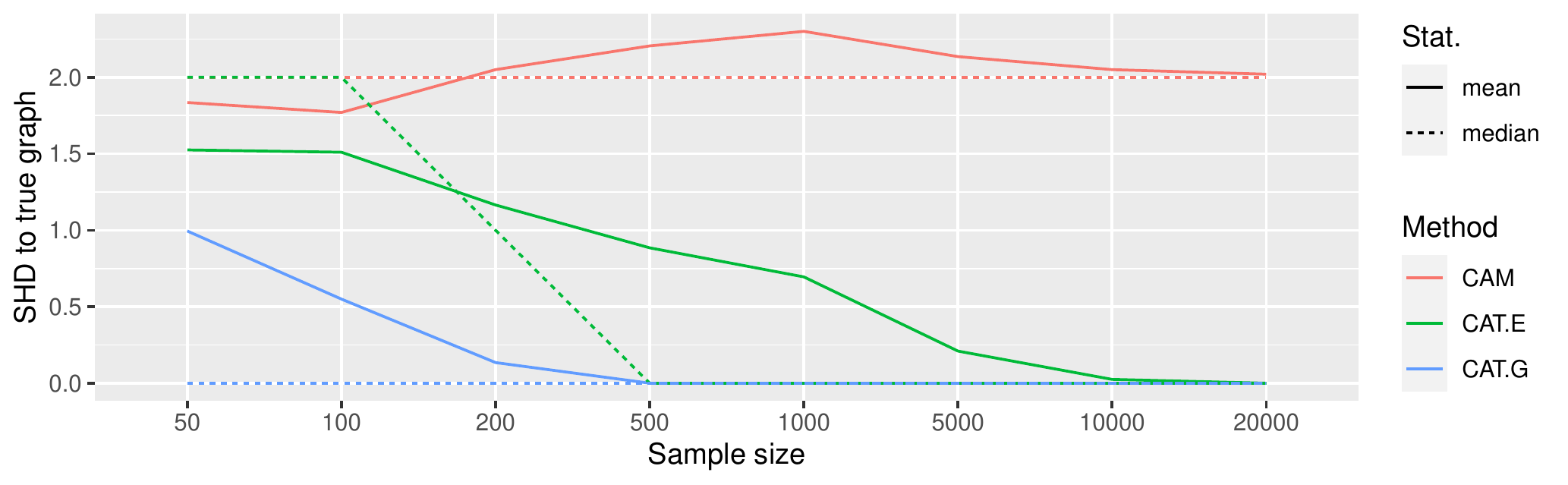}
	\end{center}
	\caption{Structural hamming distance \citep[SHD,][]{tsamardinos2006max} performance of CAT.G, CAT.E and CAM in the three node setup of \Cref{eq:3nodesetup}. The solid and dashed lines represent the mean and median SHD, respectively, based on 200 repetitions.}
	\label{fig:boxplot_3node_SHD}
\end{figure}

We now highlight why the greedy search fail. The following explanation relies on the theory presented in \Cref{ch:trees}, but for now it suffices to know that the score function evaluated in a graph $\tilde \cG= (\tilde \cE,V)$ is given by the sum of certain edge weights $w_\mathrm{G}(j\to i)$ for all edges in the graph. The greedy search technique of CAM iteratively selects the lowest scoring directed edge under the constraint that no cycles is introduced in the resulting graph. \Cref{fig:boxplot_3node_weights} shows
the estimated  Gaussian edge weights.  The smallest edge weight is given by the wrong edge $(Z\to Y)$ so the greedy search erroneously picks this edge. However, Chu--Liu--Edmonds' algorithm used by CAT correctly realizes that the full score of the correct graph $X\to Y \to Z$ is smaller than the full score of $Z\to Y \to X$ which is recovered by CAM.
\begin{figure}[h]
	\begin{center}
	\includegraphics{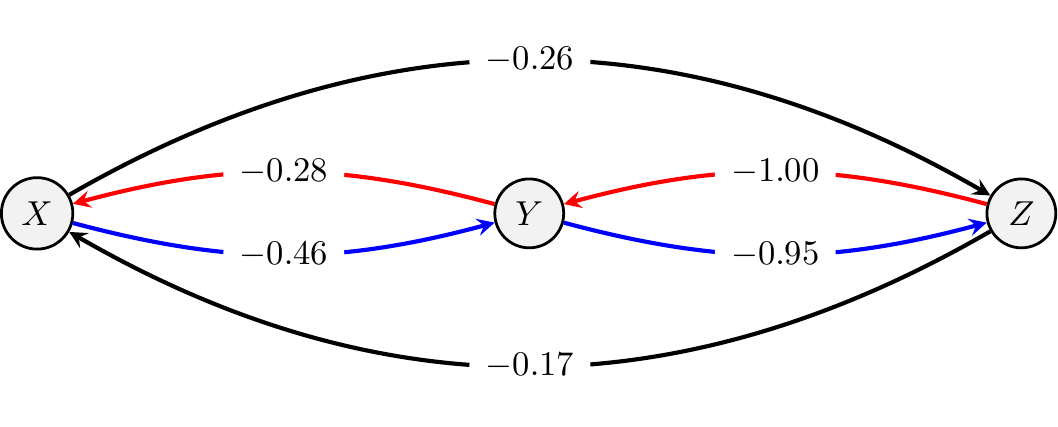}
	\end{center}
	\caption{Visualization of the edge weights  of the experiment in Section~\ref{sec:WhenGreedySearchFail}. Each edge label is the estimated Gaussian edge weight  as produced by the CAM scoring method based on 1000000 i.i.d.\ observations generated from the structural causal system of \Cref{eq:3nodesetup}. The red edges are recovered by the greedy search of CAM and the blue edges are recovered by  Chu--Liu--Edmonds' algorithm of CAT. We see that $-1.41=
		\hat 	w_\mathrm{G}(X \to Y) + \hat	w_\mathrm{G}(Y \to Z) < \hat	w_\mathrm{G}(Z \to Y)+ 	\hat w_\mathrm{G}(Y \to X)=-1.28$.}
	\label{fig:boxplot_3node_weights}
\end{figure}

\subsection{Learning Summary Graphs of Time Series}

In \Cref{ch:pmlr}, we consider the problem of learning summary graphs of time-homoge-neous stochastic processes. The paper is the culmination of the authors' participation and victory in the NeurIPS 2019 Causality 4 Climate (C4C) competition.\footnote{https://causeme.uv.es/neurips2019}  Here, teams were given finite sample data of different simulated $d$-dimensional time series and then tasked with inferring the underlying summary graph.  The summary graph is a simplification of the (infinite) causal graph. It consists of $d$ nodes with an edge from node $j$ to node $i$ if and only if any past values of the $j$'th coordinate process enter the structural assignment of the $i$'th coordinate process. For each data set, the participants could upload a weighted adjacency matrix $A$ corresponding to the summary graph where each entry held the belief or score that an edge is present. The online platform, to which the weighted adjacency matrix was uploaded, then scored the method by the area under the curve of the receiver operating characteristic (AUC-ROC) metric.

The receiver operating characteristic is a function $\text{ROC}:[0,1]\to [0,1]^2$ which for a binary classifier system takes a threshold $t\in[0,1]$ and yields $\text{ROC}(t)=(\text{FPR}(t),\text{TPR}(t))$ where $\text{FPR}(t)$ and $\text{TPR}(t)$, are the false positive rate and true positive rate of the classifier system using a threshold of $t$. In our setting, for a fixed threshold $t\in[0,1]$, we convert the weighted adjacency matrix $A$ to a binary adjacency matrix $A^*(t)$, where $A^*(t)_{ji} = 1_{[A_{ji}/\max_{ji}A_{ji},1]}(t)$. The true positive rate (TPR) using the threshold $t$ is then given by calculating the fraction of correct edges in $A^*(t)$ over the number of true edges in the underlying summary graph. The false positive rate (FPR) is given by the number of incorrect edges in $A^*(t)$ over the total number of absent edges in the underlying summary graph.

In the paper, we detail our algorithms and present heuristic justifications for our choices. Two important observations are that: 1) our methods using linear regression to capture causal effects seems to work well even though the true causal mechanisms are nonlinear, and 2) the size of the estimated linear coefficients seemed to work better than using an associated test-statistics for a test of vanishing linear effect. 
We now present a heuristic justification for why linear methods can still be used to discover nonlinear causal effects. In \Cref{Intro:sec:Artifacts}, we discuss why using the size of linear regression coefficients can outperform methods using corresponding test sizes for tests of vanishing linear effect.

Consider a simple (single-lag) time-homogeneous discrete-time stochastic process $(X(t))_{t\in \N_+}$, where for each time step $t \geq 1$ the process $X(t)\in \R^d$ is driven by past values according to $
X(t) := F(X(t-1))+N(t)$, for $t\geq 1$, some fixed function $F=(F_1,...,F_d):\R^{d}\to \R^d$, noise innovations $(N(t))_{t\geq 1}$ and some initial distribution $X_0$. As such consider the parameter $		\theta_{ji}(t) = \E| \partial_j F_i(X(t))|$. When the process $(X(t))_{t\in \N_+}$ is strictly stationary, this parameter does not depend on $t$, and it is clear that when there is an edge in the summary graph from $j$ to $i$, then $\theta_{ji}>0$ and $\theta_{ji}=0$ otherwise. In order to detect regions with non-zero gradients of $F$, we create random bootstrap samples $\cD_1,...,\cD_B$ of the observed time series. We then obtain (possibly penalized) linear regression coefficients each bootstrap sample. The idea is that, if there is no link in the summary graph, then all the bootstrap coefficients are likely small. On the other hand, if $\theta_{ji}>0$, then there might be at least one large absolute coefficient. We then use the average of the absolute regression coefficients over the $B$ bootstrap samples as a proxy for $\theta_{ji}$. We average the absolute coefficients to avoid possible cancellation. This estimate does not contain any information about whether there is a positive or negative effect from $X_{j}(t-1)$ to $X_{i}(t)$, nor can it be used for prediction purposes. It solely serves as a score or belief in the existence of a cause-effect mechanism between past values of $X_j$ onto $X_i$.
%
%
%
\subsubsection{Artifacts in DAG Models} \label{Intro:sec:Artifacts}
In the above learning framework we were only interested in the belief of a causal link, i.e., only quantifying that a linear coefficient is nonvanishing. An immediate question is now: why do we not use, for example, the T-statistic corresponding to the test for the hypothesis that the regression coefficients are zero instead of the absolute size of the corresponding coefficient? The answer is that our proposed algorithms are to some extend tailored towards maximizing the AUC-ROC on the simulated time series data. We explicitly saw a drop in performance when changing to test statistics or p-values. As shown in the simulation experiment, such behavior is also seen in general DAG models where the marginal variance tends to increase the further down the  causal order we go. 

We exploited this in our methods, but this is not a desirable feature of general-purpose structure learning algorithms, since we generally have no evidence or a priori belief that real-world systems exhibit such behavior.  \cite{reisach2021beware} further investigated these observations. They argue that for simulated linear additive noise DAG models, it is very easy to, unknowingly, construct models for which the marginal variance increases with the causal order. For example, they show that the benchmark setup of, e.g., \cite{zheng2018dags} and \cite{Golem} is highly affected by this increasing variance artifact. The problem with such benchmarking setups is that heuristic score-based approaches like  \cite{zheng2018dags} can exhibit remarkable performance that is superior to other more canonical and well-studied structure learning methods. This performance superiority is immediately lost when data is properly standardized.

\section{Learning Causal Effects} \label{sec:INTRO_LearningCausalEffects}

The previous section discussed causal structure learning methods that enable us to learn the existence of cause-effect relationships in stochastic systems. We may also be interested in knowing how a system variable behaves under external manipulation (interventions) on the causes of said variable.

Consider, for example, a binary treatment indicator $T\in\{0,1\}$ indicating whether a patient is administered a specific treatment or not. Suppose that we want to quantify the effect of said treatment on a response variable $Y$, e.g., a post-treatment indicator of a specific disease or some other biochemical marker of interest. One way to quantify this effect is to consider the treatment's average causal effect (or average treatment effect) on the response variable. That is, we may consider the difference in the expected response variable under two different interventions:
\begin{align*}
	\text{ATE}:=	\E^{\text{do}(T:=1)}[Y] - \E^{\text{do}(T:=0)}[Y].
\end{align*}
We may also be interested in quantifying how much a response variable $Y$ is affected by interventions on a continuous system variable $X$. For example, the expected behavior of $Y$ under interventions that fix $X$ at specific values, i.e., the function $x \mapsto \E^{\text{do}(X:=x)}[Y]$ or its derivative $
x\mapsto D_x \,  \E^{\text{do}(X:=x)}[Y]$. These quantities provide information about whether the response, on average, will decrease or increase due to applying external manipulation, which artificially increases the continuous variable $X$. 

For certain models where $X$ is a direct cause of the response $Y$ the inferential target quantifying the causal effects becomes the causal coefficients (in linear SCMs) and causal functions (in nonlinear SCMs) appearing in the structural assignments of $Y$; see \Cref{example:into:lineareffects} and \Cref{example_into_nonlinearCausalEffect} below.

\begin{example}[Causal effects in linear models] \label{example:into:lineareffects}
	Consider a linear additive structural causal model $(Q,\cS)$ over $(Y,X,H)$ with $Y\in \R$, $X\in \R^d$ and $H\in \R^r$ with structural assignments given by
	\begin{align*} 
		\begin{bmatrix}
			Y & X^\t & H^\t \end{bmatrix} 
		:=  \begin{bmatrix}
			Y &X^\t  & H^\t  \end{bmatrix} B   + N^\t,
	\end{align*}
	for some strictly lower triangular constant matrix $B$ and noise innovation vector $N\sim Q$ with zero mean. Assume w.l.o.g.\ that the first column of $B$ is given by $(0,\gamma ,\delta)$ such that the structural equation of $Y$ becomes $
	Y := \gamma^\t X + \delta^\t H + N_Y.$
	Since $B$ is strictly lower triangular, we know that the variables $H$ act as possible confounders of the causal effect from $X$ to $Y$, i.e., the causal effect is not mediated by $H$. As such, they are unaffected by interventions on $X$. Now consider the intervention $\mathrm{do}(X:=x)$ for some constant $x\in \R^d$ and note that  $
	\E^{\mathrm{do}(X:=x)}[Y]= \E^{\mathrm{do}(X:=x)}[\gamma^\t x + \delta^\t H + N_Y] = \gamma^\t x$. Hence, the causal effect $
	D_x \, \E^{\mathrm{do}(X:=x)}[Y] = \gamma,$ 
	is constant and given by the structural parameters $\gamma$.
\end{example}

\begin{example}[Causal effects in nonlinear additive models] \label{example_into_nonlinearCausalEffect} 
	Consider a possibly nonlinear structural causal model $(Q,\cS)$ over $(Y,X,H)$ with $Y\in \R, X\in \R^d$ and $H\in \R^r$ with the structural assignments given by
	\begin{align*}
		Y := f(X)+ g_1(H,N_Y),\quad X := g_2(H,N_X), \quad H:= N_H,
	\end{align*}
	for some functions $f:\R^d \to \R$, $g_1:\R^r\to \R$,  $g_2:\R^r \to \R^d$. Now notice that $\E^{\mathrm{do}(X:=x)}[Y] = f(x) + \E[g_1(H,N_Y)]$ from which we get that $D_x \E^{\mathrm{do}(X:=x)}[Y] = D_x f(x)$. Thus, the problem reduces to finding $x \mapsto D_x f(x)$ or $x\mapsto f(x)$, i.e., the causal function $f$.
	
\end{example}

Causal effects (and other causal targets) are given by distributional features of post-interventional distributions. Hence, inference should be possibly by observing said interventions and analyzing the resulting data.  
However, given sufficient knowledge of the causal structure it is possible, in certain settings, to infer the interventional distribution (and derivatives thereof) from the observational distribution. For example, if we in \Cref{example_into_nonlinearCausalEffect} have that $X:= g_2(N_X)$, i.e., that $X$ and $Y$ are not confounded, then intervening coincides with conditioning. That is, the inferential target reduces to $
	\E^{\text{do}(X:=x)}[Y]= \E[Y|X=x]$, for which inference from observational data is a well-studied statistical problem. The next section introduces adjustment formulas that allow for a similar translation when $X$ and $Y$ are confounded.

\subsection{Adjustment Formulas}
Adjustment formulas allow one to derive intervention distributions in terms of the observational distribution, given that we have sufficient knowledge of the underlying causal structure of the system. The adjustment formulas are known in the different causal modeling frameworks as truncated factorization \citep[][]{Pearl2009}, the G-computation formula \citep[][]{Robins1986}, and the manipulation theorem \citep[][]{spirtes2000causation}.
If all relevant densities exist, we say that a set of variables $Z$ is a valid adjustment set for the causal effect from $X$ to $Y$ if it holds that
\begin{align*}
	p_Y^{\text{do}(X:=x)}(y) = \int p_{Y|X,Z}(y|x,z)p_Z(z) \, dz,
\end{align*}
where $p_Y^{\text{do}(X:=x)}$ is the post-intervention density of $Y$ under the intervention  $\text{do}(X:=x)$, $p_Z$ is a density of $Z$ and $p_{Y|Z,X}$ is a conditional density of $Y$ given $Z$ and $X$, both under the observational distribution. Thus, a valid adjustment set allows for the interventional distribution to be represented solely by the observational distribution. Various graphical criteria exist to check whether a set $Z$  is a valid adjustment set for the causal effect from $X$ to $Y$. For example,
\begin{enumerate}
	\item[$\bullet$] \textit{Parent adjustment:} Suppose that $Y$ is not a parent of $X$, $Y\not \in \PA{X}$. It holds that the collection of all parents of $X$, $Z:=\PA{X}$,  is a valid adjustment set. 
	\item[$\bullet$] \textit{Backdoor adjustment:} Suppose that $Z$ does not contain $X$ or $Y$ and that (i) $Z$ contains no descendant of $X$ and (ii) $Z$ blocks (see, \Cref{sec:IntroductionDseparation}) all paths between $X$ and $Y$ with an edge incoming edge into $X$.
\end{enumerate}
See \cite{Peters2017} for further characterizations of valid adjustment sets. However, whenever there are hidden (latent) variables, i.e., variables present in the system but not observed, we might not be able to find a valid adjustment set. We discuss this further in the next section.

\subsection{Inference in the Presence of Hidden Variables}

Latent variables further complicates the inference of causal effects. That is, the valid adjustment sets may overlap with the latent variables rendering the use of adjustment formulas to compute causal effects infeasible. In fact, the presence of hidden variables might render the causal effect unidentified. Even when the causal structure and the form of the structural assignments are known a priori, there might be multiple distinct structural causal models that generate identical observational distributions over the observed variables; see \Cref{example_introduction_non_identified}. 

\begin{figure}[t]
	\begin{center}
	\includegraphics{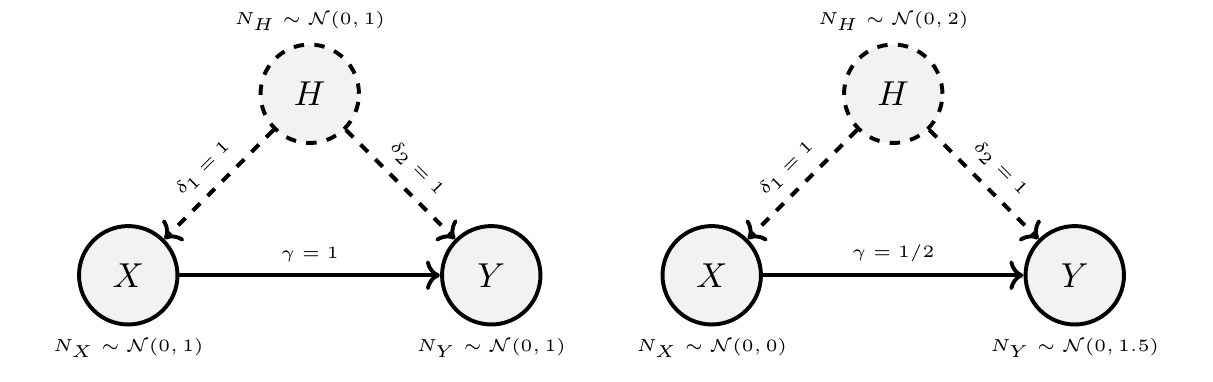}
	\end{center}
	\caption{Specifications of two linear structural causal models with the same causal graph but different causal coefficients and noise innovation variances; see \Cref{example_introduction_non_identified}. In the two linear SCMs, the causal effect from $X$ to $Y$ differs, but the induced observational distributions over $(X,Y)$ coincide.}
	\label{fig:introduction_nonidentified}
\end{figure}

\begin{example}[Hidden confounding models] \label{example_introduction_non_identified}
	Consider a linear SCM $M$ over $(Y,X,H)$ with $Y\in \R, X\in \R$ and $H\in \R$ where the $H$ denotes a hidden variable, i.e., a variable which can not be observed.
	Suppose that the structural assignments are given by
	\begin{align*}
		Y := \gamma X + \delta_1 H + N_Y, \quad X := \delta_2 H + N_X, \quad H:= N_H
	\end{align*}
	$N_Y,N_X,N_H$ being mutually independent noise innovations. In \Cref{fig:introduction_nonidentified}, two structural causal models with the above structural assignments are specified. They induce the same observational distribution over $X$ and $Y$, but the causal effects from $X$ to $Y$ differ. This example clearly illustrates that the causal effect $\gamma$ is not identified, as it is impossible to infer it from the observational distribution.	
\end{example}

In the presence of hidden confounding, we may still be able to identify causal effects by the instrumental variable method.

\subsubsection{The Instrumental Variable Method} \label{sec:IVMethodIntroduction}
\begin{figure}
	\begin{center}
	\includegraphics{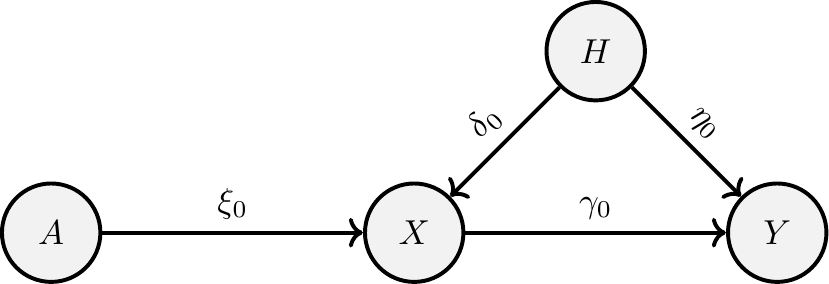}
	\end{center}
\caption{The causal graph of the one-dimensional instrumental variable setup.} \label{fig:introduction_IVModel}
\end{figure}

The instrumental variable method \citep[][]{wright1928tariff,theil1953repeated} is a method for identifying and estimating causal effects in the presence of hidden confounding. Suppose that we want to estimate the causal effect from $X$ to $Y$. The method assumes the existence of system variables $A$, called instruments, which satisfies the following two criteria \citep{Pearl2009}: 
\begin{enumerate}
	\item[(i)] \textit{Relevance:} $A$ is dependent on the predictors $X$.
	\item[(i)] \textit{Exogeneity}: $A$ is independent of all variables (including noise innovations) that influence $Y$ which is not mediated by $X$. That is, $A$ is independent of $Y$ when $X$ is held fixed:  $A\independent Y$ under distributions induced by interventions of the form $\text{do}(X:=x)$ that breaks the dependence between $A$ and $X$.
\end{enumerate}
For simplicity we introduce the method of instrumental variables in a linear setting. Suppose that $(A,X,H,Y)$, with $H$ unobserved, is generated by a linear SCM of the form
\begin{align*}
	A &:= N_A, \quad H:= N_H,\\
	X&:= \xi_0^\t A + \delta_0^\t H + N_X,\\
	Y&:= \gamma_0^\t X + \eta_0^\t H + N_Y
\end{align*}
for some mutually independent noise innovations $N_A,N_H,N_X,N_Y$ and structural coefficients $\xi_0,\delta_0,\eta_0,\gamma_0 \not = 0$. The causal graph for this setup, corresponding to $A,H,X,Y\in \R$, is illustrated in \Cref{fig:introduction_IVModel}

Suppose, furthermore, that the covariance matrices  $\text{Var}(A)$ and $\text{Var}(X)$ are positive definite.  For notational simplicity, let $U := \eta_0^\t H + N_Y$ denote the unobserved variables entering the structural assignment of $Y$. Note that $A$ satisfies the criteria of relevancy and exogeneity for being instruments for the causal effect from $X$ to $Y$. The ordinary least squares method, in general, fails to be a consistent estimator of the causal effect $\gamma$ from $X$ to $Y$, i.e., the population OLS coefficient given by
\begin{align*}
	\gamma_{\mathrm{OLS}} := \E[X X^\t]^{-1}\E[X Y] = \gamma_0 +  \E[X X^\t]^{-1}\E[XU^\t]\not = \gamma_0,
\end{align*}
as $\E[XU]\not =0$ due to the hidden confounding. On the other hand, if $\E[A X^\t]$ is of full column rank (known as the rank condition for identification which requires that $|A|\geq |X|$), then we realize that the population two-stage least squares (TSLS) coefficient
\begin{align*}
	&\gamma_{\mathrm{TSLS}}:=(\E[XA^\t]\E[AA^\t]^{-1} E[AX^\t])^{-1}\E[XA^\t]\E[AA^\t]^{-1} E[AY]\\
	&= \gamma_0+	(\E[XA^\t]\E[AA^\t]^{-1} E[AX^\t])^{-1}\E[XA^\t]\E[AA^\t]^{-1} E[AU] = \gamma_0,
\end{align*}
coincides with the causal effect from $X$ to $Y$, as $E[AU]=0$ by exogeneity. Thus, under the existence of instruments, the causal effect becomes identified from the observational distribution in the presence of hidden confounding. The name two-stage least squares come from the empirical counterpart to the population two-stage least squares coefficient coincides with the estimate resulting from a two-stage ordinary least squares procedure, where one first regresses $X$ on $A$ followed by a regression of $Y$ on the first stage predicted values of $X$.  The TSLS estimator can also be seen as a special case of the generalized method of moments (GMM), exploiting the moment restriction $\E[A(Y-\gamma^\t X)]=0$ if and only if $\gamma = \gamma_0$ \citep[see, e.g.,][]{hall2005generalized}.

The instrumental variable method is also applicable in nonlinear structural causal models; In \Cref{ch:Generalization}, we, for example, utilize that the existence of instruments can identify nonlinear causal functions. See \Cref{sec:IVconditions} for further discussion and references on nonlinear and nonparametric instrumental variable regression.

\subsubsection{The P-Uncorrelated Least Squares Estimator}
In \Cref{ch:PULSE}, we propose a novel estimator in the linear instrumental variable setting called the p-uncorrelated least squares estimator (PULSE), which has the intuitive interpretation of minimizing the mean squared prediction error over a confidence region for the causal parameter.  We show through simulation studies that our estimator, which can also be seen as a data-driven regularized TSLS regression, suffers from less variability than TSLS and other competing estimators while maintaining consistency.  We continue our summary of the PULSE using the linear SCM setup of \Cref{sec:IVMethodIntroduction}.

The two-stage least squares estimator is very unstable, especially in weak instrument settings (the effect from $A$ to $X$ is weak; see \Cref{sec:WeakInst} for further details). The TSLS estimator does not have moments of any order in the just-identified setup ($|A|=|X|$); see, e.g., \cite{mariano2001simultaneous}.

Under certain identifiability conditions, the null-hypothesis $\cH_0(\alpha): \text{Corr}(A,Y-X\alpha)=0$ is only satisfied by the causal coefficient, i.e., the causal effect from $X$ to $Y$. The TSLS estimator sets the sample covariance between instruments and the regression residuals to zero in the just-identified setup. Intuitively, this restriction might be too strong as the sample covariance, even for the true causal coefficients, is likely to be small but non-zero. On the other hand, the OLS estimate is known to be biased but fairly stable with moments of any order for sufficiently large sample sizes \citep[see, e.g., ][]{mariano1972existence}. The idea  of the p-uncorrelated least squares estimator (PULSE) is to minimize the mean squared prediction error constrained to a finite-sample acceptance region $\cA_n$ of a test for uncorrelatedness, $\cH_0(\alpha)$. That is, we propose an estimator of the form
\begin{align} \label{eq:PULSE_introduction}
	\hat{\gamma}_{\text{PULSE}}^n :=  \begin{array}{ll}
		\argmin_{\gamma} & \frac{1}{n} \sum_{k=1}^n (Y_k-\gamma^\t X_k)^2  \\
		\text{subject to} & \gamma \in \cA_n.
	\end{array}
\end{align}
In \Cref{ch:PULSE}, we propose a class of asymptotically valid hypothesis tests for $\cH_0(\alpha)$. While the test has desirable properties the resulting minimization in  \Cref{eq:PULSE_introduction} becomes a non-convex optimization problem. However, through careful analysis and dual theory, we show that the estimator can be efficiently computed as
\begin{align} \label{eq:IntroductionPULSEDual}
	\hat{\gamma}_{\text{PULSE}}^n:=& l_{\mathrm{OLS}}^n(\gamma) + \lambda^* l_{\mathrm{IV}}^n(\gamma),
\end{align}
where $l_{\mathrm{OLS}}^n$ and $l_{\mathrm{IV}}^n(\gamma)$ is the empirical ordinary and two-stage least squares loss functions (i.e., the OLS and TSLS estimators minimizes these functions, respectively) and $\lambda^*$ is a data-dependent regularization parameter that can be approximated with arbitrary precision. This representation also reveals that the PULSE estimator belongs to a special class of estimators known as K-class estimators \citep{theil1953repeated}.

In an identified setup, the PULSE estimator consistently estimates the causal coefficient $\gamma_0$. In other words, the data-dependent $\lambda^*$ is guaranteed to tend to infinity as the sample size increases. Hence, the data-driven mean squared prediction error (MSPE) regularization vanishes in the large sample limit. The PULSE estimator is also well-defined in the under-identified setup $(|A|<|X|)$, which renders the causal effect unidentified. In the under-identified setup, the empirical objective is still to find the best predictive model among all coefficients that do not reject uncorrelatedness. Here, however, the target is not the causal coefficient but the coefficient in the TSLS solution space (all coefficients that render the instruments independent of residuals), which minimizes the MSPE. 

Extensive simulation studies show that there are settings where the PULSE estimator indeed outperforms the TSLS and other competing instrumental variable estimators in terms of mean squared error (MSE). Weak instruments and weak endogeneity roughly characterize these settings.  The MSPE regularization increases the bias in these settings, but the corresponding decrease in variance yields an MSE superior estimator.
Furthermore, in \Cref{ch:Generalization}, we extend this data-dependent MSPE regularization idea to nonlinear instrumental variable setups. The  proposed estimator NILE likewise shows an MSE performance gain compared to various state-of-the-art nonparametric instrumental variable estimators.


\section{Learning Generalizing Functions} \label{sec:LearningGeneralizingFunctionsIntroduction}
Suppose that we are interested in learning prediction methods that minimize a particular loss function over the observational distribution. For example, it is common to construct a prediction method that minimizes the mean squared prediction error (MSPE) over the observational distribution $
\argmin_{f_\diamond } \E[(Y-f_\diamond (X))^2]$, 
which we know coincides with the conditional expectation function of $Y$ given $X$, but other loss functions may be reasonable too.

However, in many applications, one may wish to employ the prediction method on future system instances. For some systems, it may be reasonable to expect that future instances are subject to change. Alternatively, one may wish to employ a prediction method to entirely new systems  known to differ from the system on which the method was trained.  These problems are known under slight variations as, for example, covariate shift, domain generalization/adaption, and  out-of-distribution generalization/prediction. We refer the reader to \Cref{sec:intro} of \Cref{ch:Generalization} for numerous references in this area of research. Common to these research areas is that the distribution of the training instance $P_{\mathrm{train}}$ differs from the class of possible test distributions $\cP$ on which the prediction method is to be applied.

If one has a priori knowledge of the likelihood that each possible test distribution is to appear, one could, for example, try to  minimize a weighted average of the MSPE over all possible test distributions. Alternatively, we may consider the problem of learning a prediction method $f^*$ that seeks to minimize the worst-case MSPE;
\begin{align*}
	f^* \in \argmin_{f_\diamond } 	\sup_{P_{\mathrm{test}}\in \cP} \E_{P_{\mathrm{test}}}[(Y-f_\diamond (X))^2],
\end{align*}
where $\E_{P_{\mathrm{test}}}$ denotes the expectation with respect to the distribution $P_{\mathrm{test}}$. In this thesis, we concentrate on the latter objective. We say that a prediction method $f^*$ is distributionally robust, a generalizing function, or a minimax solution with respect to a class of distributions $\cP$ if it minimizes the worst-case prediction risk over all distributions in $\cP$

%
%

In order to learn such a generalizing prediction method, we first must specify the class $\cP$ of possible test distributions.  A common approach is to say that the test distributions are slight variations of the training distribution in the sense that $P_{\mathrm{test}}\in B_{\rho}(P_{\mathrm{train}},\ep)$, i.e., that the test distribution lies within an $\epsilon$-ball of the training distribution $P_{\mathrm{train}}$, for some metric $\rho$ on the space of probability measures, e.g., the Wasserstein metric.  While this framework aims to guard against test distributions that arise from small perturbations in training distribution with respect to some probability metric, one may argue that it may be more natural for many applications that the test distributions arise from external manipulation of the original system.


In \Cref{ch:Generalization}, we consider the problem of learning generalizing functions with respect to test distributions that are induced by interventions. That is, the set of possible test distributions $\cP= \{P_{M(i)}:i\in \cI\}$ are given by intervention-induced distributions in the underlying structural causal model $M$. Here, $\cI$ denotes a class of interventions. We consider a framework where $M$ belongs to a fairly general class of models $\cM$ which both contains the response variable $Y$, predictors $X$, latent variables $H$, and exogenous variables $A$. We allow for certain well-behaved interventions on $X$ and $A$ and aim to find generalizing prediction functions within some pre-specified function class $\cF$. That is, we aim to learn
\begin{align*}
	f^*\in \argmin_{f_\diamond \in \cF} \sup_{i\in \cI} \E_{M(i)}[(Y-f_\diamond(X))^2],
\end{align*}
where $\E_{M(i)}$ denotes the expectation with respect to the interventional distribution induced by the intervention $i$ in the model $M$. Such generalizing prediction functions depend, among other things, on the function class $\cF$, the model class $\cM$ and the class of interventions $\cI$.




It is well-known that when $\cI$ contains all possible hard interventions of the form $\cI= \{\text{do}(X:=x): x\in \R^d\}$ then the causal function $f$ solves the minimax problem \citep[see, e.g., ][]{Rojas2016}. Conversely, we may also consider $\cI$ to be a singleton consisting of the trivial intervention which does nothing, in which case $x \mapsto \E[Y|X=x]$ is a minimax solution. 

We show, for example, that the causal function is a minimax solution even for singleton interventions that are confounding-removing, i.e., interventions that break the confounding between the predictors $X$ and the target $Y$. Furthermore, we show that minimax solutions that differ from the causal function are highly susceptible to misspecifications of the intervention class. While the causal function is minimax whenever $\cI$ contains at least one confounding-removing intervention, alternative non-causal minimax solutions may perform worse than the causal function if the intervention class is misspecified.

In practical scenarios, the underlying model $M$ is unknown, and we do therefore not have access to the intervention induced-distributions $P_{M(i)}$ for $i\in \cI$. Thus, similar to the hurdles plaguing the inference of causal effects from observational data, we can not identify and learn generalizing functions from observational data without further causal assumptions. There may exist an alternative model $\tilde M\in \cM$ with identical observational distribution, $P_{\tilde M}=P_{M}$ but which differs on intervention distributions. As such, we say that distribution generalization is possible if there exists a function $f^*$ which is minimax optimal for all observationally equivalent models within the model class $\cM$.

We present sufficient conditions for distribution generalization in terms of restrictions on the observational distribution $P_M$, the intervention class $\cI$, and the model class $\cM$. Furthermore, we provide several impossibility theorems which illustrate the necessity of some of these restrictions.

\subsection{PULSE and NILE}

We know that when the intervention class contains arbitrarily strong interventions on $X$ or at least one confounding-removing intervention then the causal function is a generalizing function. As such, any learning method for the causal function is equivalently learning a generalizing prediction function. The PULSE estimator, for example, consistently estimates a generalizing linear prediction function.

Similar considerations hold for nonlinear and nonparametric instrumental variable estimators as long as the intervention class is not support extending. That is, as long as the interventions do not extend the support of $X$. Since instrumental variable estimators can only recover the causal function on the support of the observational distribution this restriction is necessary without further causal assumptions. If, however, the interventions are  support extending, then further causal assumptions are needed to extrapolate the estimate outside the support of $X$. 

In \Cref{ch:Generalization}, we present a nonlinear instrumental variable estimator which explicitly incorporates causal assumptions that the causal functions extrapolate linearly outside the support of observational distribution. We call this the nonlinear intervention-robust linear extrapolator (NILE). The linear extrapolation is not of importance --- any extrapolation scheme which is uniquely determined by the on-support behavior works equally well. The NILE also uses the data-driven MSPE regularization ideas introduced for the PULSE.
\subsection{Anchor Regression and K-class Estimators} \label{sec:IntroARandKclassRobustness}
For linear SCMs \cite{rothenhausler2018anchor} show that among linear prediction functions, there exist functions that are minimax solutions but do not coincide with the causal functions whenever the intervention class $\cI$ consists of bounded interventions on exogenous  variables. That is, they show that for linear SCMs with exogenous variables $A$ (called anchors), endogenous variables $X$, and a target $Y$, the anchor regression coefficient with regularization parameter $\lambda$ is distributionally robust. More specifically, this linear prediction of $Y$ from $X$ is distributionally robust with respect to interventions on the exogenous variables $A$ up to a certain strength that depends on $\lambda$.

The results of \Cref{ch:PULSE} shows that anchor regression is closely related to K-class estimators \citep[][]{theil1953repeated}, which are parameterized by a real-valued parameter $\kappa$. The K-class estimators contain several well-known linear effect estimators: the ordinary least squares estimator for $\kappa=0$, the two-stage least squares estimator for $\kappa=1$, and for specific data-driven $\kappa$ one can recover the limited information maximum likelihood \citep[][]{anderson1949estimation} and Fuller estimators \citep[][]{fuller1977some}. 

Using the ideas  of \cite{rothenhausler2018anchor}, we extend the distributional robustness property of anchor regression to general K-class estimators with fixed $\kappa \in [0,1)$. Namely, we show that for a fixed $\kappa \in[0,1)$ the K-class estimator $\hat \alpha^n_{\mathrm{K}}(\kappa,Y,Z,A)$ for regressing $Y$ onto $Z\subseteq(X,A)$ using that $A$ are exogenous variables, converges in probability towards a population quantity that is minimax prediction optimal among all linear predictors. That is, 
\begin{align*}
	\hat \alpha^n_{\mathrm{K}}(\kappa,Y,Z,A) \convp_n \argmin_{\alpha} \sup_{v\in C(\kappa)} \E^{\text{do}(A:=v)}[(Y-\alpha^\t Z)^2],
\end{align*}
as the sample size $n$ tends to infinity, where the intervention class is given by $
C(\kappa) := \{v:\Omega \to \R^q:\E[vv^\t] \preceq (1-\kappa)^{-1}\E[AA^\t]\}.$

\chapter[Distributional Robustness of K-class Estimators and the PULSE]{Distributional Robustness of K-class Estimators and the PULSE} \label{ch:PULSE}

{ \textsc{Joint work with
		\begin{quote}
			Jonas Peters
\end{quote}}}
\vspace{0.75cm}

\begin{quoting}[leftmargin=0.5cm]
	\begin{center}
		\textbf{Abstract}
	\end{center}
	
	{  While causal models are robust in that they are prediction optimal under arbitrarily 
		strong interventions, 
		they may not be optimal when the interventions are bounded.
		We prove that the classical K-class estimator satisfies such optimality by establishing a connection between 
		K-class estimators and anchor regression. 
		This connection 
		further motivates a novel estimator in
		instrumental variable settings 
		that minimizes
		the mean squared prediction
		error subject to the constraint that the estimator lies in an
		asymptotically valid confidence region of the causal coefficient. We call
		this estimator PULSE (p-uncorrelated least squares estimator), relate it to work on invariance,
		show
		that it can be computed efficiently as a data-driven K-class estimator, even though the underlying optimization problem is non-convex, and prove consistency.
		We evaluate 
		the estimators on real data and 
		perform simulation experiments illustrating 
		that PULSE suffers from less variability. There are several settings including 
		weak instrument settings, where it outperforms other estimators. }
\end{quoting}
\textbf{Keywords:} Causality, distributional robustness, instrumental variables

\Crefname{enumi}{}{}

\section{Introduction}
Learning causal parameters from data has been 
a key challenge in many scientific fields 
and has been a long-studied problem in 
econometrics
\citep[e.g.][]{	goldberger1972structural, koopmans1953studies,wold1954causality}.
Many years after the groundbreaking work by
\citet{Fisher35} and \citet{Peirce83},
causality plays again an increasingly important role 
in machine learning and statistics,
two research areas that
are most often considered part of mathematics or computer science 
\citep[e.g.,][]{Imbens2015, Pearl2009, Peters2017,spirtes2000causation}.
Even though the current
developments 
in mathematics, computer science on the one
and econometrics on the other hand
do not forego independently, 
we believe that there is a lot of
potential for more
fruitful
interaction 
between these two fields.
Differences in the 
language have emerged, which can 
make 
communication difficult, but
the target of inference,
the underlying principles, and 
the methodology 
in both fields are closely related.
This paper 
establishes
a link between two developments 
in these fields:
K-class estimation 
which aims at 
estimation of causal parameters with good 
statistical properties 
and 
invariance principles
that are used to build methods that 
are robust with respect to distributional shifts. 
This connection allows us to prove distributional robustness guarantees for K-class estimators and motivates a new estimator, PULSE.
We summarize our main results in Section~\ref{sec:summary}.

\subsection{Related Work}
Given causal background knowledge, 
causal parameters 
can be estimated 
when taking into account 
confounding effects between treatment and outcome.
Several related 
techniques have been suggested to tackle that problem,
including 
variable adjustment \citep[][]{Pearl2009}, 
propensity score matching \citep[][]{Rosenbaum83},
inverse probability weighting \citep[][]{Horvitz1952} or 
G-computation \citep[][]{Robins1986}.

If some of the relevant variables have not been observed, 
one may instead 
use exogenous variation in the data to 
infer causal parameters, 
e.g., in the setting of instrumental variables
\citep[e.g.,][]{angrist1995identification,newey2013nonparametric,Wang2016,wright1928tariff}. 
Limited information estimators leverage instrumental variables 
to conduct single equation inference. 
%
An example of such methods is the two-stage 
least squares estimators (TSLS) developed by \citet{theil1953repeated}. 
Instead of minimizing the 
residual sum of squares as 
done by the
ordinary least square (OLS) estimator, 
the TSLS
minimizes
the sample-covariance between the instruments and regression residuals. 
TSLS estimators are consistent, but
are known to have suboptimal finite sample properties, e.g., they only have moments op to the degree of  over-identification \citep[][]{mariano1972existence}.  
\citet{Kadane1971} shows that under suitable conditions, the mean squared error of TSLS might even be larger than the one of OLS 
if the sample size is small (more precisely and using the notation introduced below, if $0\leq n-q\leq 2(3-(q_2-d_1))$, where $q_2-d_1$ is the degree of overidentification). This result is another indication that under certain conditions, it might be beneficial to use the OLS for regularization.
Another method of inferring causal parameters in structural equation models is the limited information maximum likelihood (LIML) estimator due to \citet{anderson1949estimation}. %
\citet{theil1958economic} introduced K-class estimators, which contain OLS, TSLS and the LIML estimator as special cases. This class of estimators is parametrized by a deterministic or stochastic parameter $\kappa\in[0,\i)$ that depends on the observational data. Under mild regularity conditions a member of this class is consistent and asymptotically normally distributed if 
$(\kappa-1)$ and 
$\sqrt{n}(\kappa-1)$ 
converge, respectively,
to zero in probability when 
$n$ tends to infinity; see, e.g.\ \citet{MARIANO1975},  \citet{mariano2001simultaneous}. While the LIML does not have moments of any order, it shares the same asymptotic normal distribution with TSLS.
Based on simulation studies,
\citet{anderson_hildenbrand_1983} argues that,
in many 
practically relevant
cases,
the normal approximation to a 
finite-sample estimator 
is inadequate for TSLS but 
a useful approximation in the case of LIML.
Using Monte Carlo simulations, 
\citet{hahn2004estimation} 
recommend that the no-moment estimator LIML should not be used in weak instrument situations, where Fuller estimators have a substantially smaller MSE.
The Fuller estimators
\citep[][]{fuller1977some} form
a subclass of the K-class estimators based on a modification to the LIML, which fixes the no-moment problem while maintaining consistency and asymptotic normality. 
\citet{Kiviet2020} proposes a modification to the OLS estimator that makes use of explicit knowledge of the partial correlation between the covariates and the unobserved noise in $Y$. 
\citet{AndrewsArmstrong2017} propose an unbiased estimator 
that is based on knowledge of the sign of the first stage regression and the variance the reduced form errors and that is less dispersed than TSLS, for example. 
\citet{judge2012minimum} consider an affine combination of the OLS and TSLS estimators, which, again, yields
a modification in the space of estimators.
We prove that our proposed estimator, PULSE, can also be written as a data driven K-class estimator. 
As such, it minimizes a convex combination of the OLS and TSLS loss functions and can, in general, not be written as a convex combination of the estimators.

All of the above methods exploit 
background knowledge, e.g., 
in form of exogeneity of some of the variables.  
If no such background knowledge is available, 
it may still be possible,
under additional assumptions, 
to infer the
causal structure, e.g., represented by a graph, from observational (or observational and interventional) data. 
This problem is sometimes referred to as 
causal discovery.
Constraint-based methods 
assume that the underlying distribution is Markov and faithful with respect to the causal graph and perform 
conditional independence tests to infer (parts of) the graph; see, e.g.\ \citet{spirtes2000causation}.
Score-based methods 
assume a certain statistical model and optimize (penalized) likelihood scores; see, e.g.\ \citet{chickering2002optimal}.
Some methods exploit a simple form of causal assignments, such as 
additive noise
(e.g.,  \citealp{peters2014causal}, and \citealp{shimizu2006linear})
and others are based on exploiting invariance statements \citep[e.g.,][]{Peters2016jrssb, Meinshausen2016}. 
Many of such methods assume causal sufficiency, i.e., 
that all causally relevant variables have been observed, but some versions exist that
allow for hidden variables; see, e.g.\ \citet{Claassen2013} and \citet{Spirtes1995}.

Recent works in the fields of machine learning and computational statistics \citep[e.g.][]{HeinzeDeml17,Pfister2019pnas,Scholkopf2012}
investigate whether
causal ideas can help to make machine learning
methods more robust.
The reasoning is that causal models are robust against any intervention in the 
following sense. 
Consider a target 
or response
variable $Y$ and
covariates $X_1, \ldots, X_p$. 
If we regress $Y$ on the set 
$X_S$, 
$S \subseteq \{1, \ldots, p\}$,
of direct causes,
then this regression function 
$x \mapsto E[Y|X_S = x]$
does not change 
when intervening on any of the covariates (which is sometimes referred to as `invariance').  
This statement 
can be proved using the 
local Markov property \citep[][]{Lauritzen1996}, 
for example,
but the underlying fundamental principle 
has been discussed already several decades ago;
most prominently
using the terms 
`autonomy' or `modularity' (\citealp{Haavelmo1944}, and \citealp{Aldrich1989}).
As a result, causal models 
of the form 
$x \mapsto E[Y|X_S = x]$
may perform well
in prediction tasks, 
where, in the test distribution,
the covariates have been intervened on.
If, however, training and
test distributions
coincide, a model 
focusing only on prediction and the estimand 
%
$x \mapsto E[Y|X = x]$
may outperform a causal approach.

The two models described above (OLS and the causal model) formally solve a minimax problem on
distributional robustness. 
Consider therefore an 
acyclic linear structural equation model (SEM)  over $(Y,X)$ with observational distribution $F$. Details on SEMs and interventions can be found in
\Cref{sec:simsem}.
Assume that the assignment for $Y$ equals
$Y = \gamma_0^\t X + \ep_Y$
for some $\gamma_0\in \R^{d}$. The
variables corresponding to non-zero entries
in $\gamma_0^\t X$
are called
the parents of $Y$, and
$\ep_Y$ is assumed to be independent of these parents.
Then, the 
mean squared prediction error 
when considering the observational
distribution is not necessarily minimized by
$\gamma_0$, that is, in general, we have
$\gamma_0 \neq \gamma_{\text{OLS}} := \argmin_{\gamma} E_F\left[ (Y-\gamma^\t X)^2\right]$. 
Intuitively, 
we may improve the prediction of $Y$
by including other variables than the parents of $Y$, such as its descendants.
When considering distributional robustness, we 
are interested in finding a $\gamma$ that minimizes
the worst case expected squared prediction error
over a class of distributions, $\mathcal{F}$, 
that is, 
\begin{align}
	\label{eq:optimF}
	\argmin_\gamma \sup_{F\in \cF} E_F\left[ (Y-\gamma^\t X)^2\right].
\end{align}
If we observe data from 
all different distributions in $\mathcal{F}$ 
(and know which data point comes from which distribution), we can 
tackle 
this optimization 
directly
\citep[][]{meinshausen2015maximin}.
But estimators of~\Cref{eq:optimF}
may be available even
if we do not observe data 
from each distribution in $F$.
The true causal coefficient $\gamma_0$, for example, 
minimizes \Cref{eq:optimF}
when $\cF$ is the set of all possible 
(hard)
interventions on $X$ \citep[e.g.,][]{rojas2018invariant}. 
The OLS solution is optimal when $\mathcal{F}$ only contains the training distribution.
In this sense,
the OLS solution and the true causal coefficient 
constitutes the end points of a spectrum of
estimators that are prediction optimal under a 
certain class of distributions.

Intuitively, models trading off causality and predictability may perform
well in situations, where the test distribution is only moderately
different from the training distribution. 
%
Anchor regression by \citet{rothenhausler2018anchor}, see Section~\ref{sec:AR} for details, is 
one approach formalizing this intuition
in a linear setup.
Similarly to an instrumental variable setting, 
one assumes the existence of exogenous variables that are called $A$ (for anchor)
which may or may not act directly on the target $Y$.
The proposed estimator minimizes a convex
combination of the residual sum of squares and the TSLS loss function
and is shown to be prediction optimal 
in the sense of
\Cref{eq:optimF} 
for a 
class $\mathcal{F}$
containing interventions on the covariates up to 
a certain strength; this strength depends on a regularization parameter: the weight that is used in the convex combination of anchor regression. 
Other approaches \citep[][]{pfister2019stabilizing, rojas2018invariant,Magliacane2018} search over different subsets $S$ and
aim to choose sets that are both invariant and predictive.

\subsection{Summary and Contributions} \label{sec:summary}

This paper contains two main contributions: A distributional robustness property of K-class estimators with fixed $\kappa$-parameter and a novel estimator for causal coefficients called the p-uncorrelated least squares estimator (PULSE). The following two sections 
summarize our contributions.
\subsubsection{Distributional Robustness of K-class Estimators.}
\label{sec:intro:robustness}

In \Cref{SEC:ROBUSTNESS} we show
that anchor regression is closely related to 
K-class estimators. 
In particular,
we prove that for a restricted subclass of models K-class estimators
can be written as 
anchor regression estimators.
For this subclass, this directly implies a distributional
robustness property of K-class estimators.
We then 
prove a similar robustness
property for general
K-class estimators with a fixed penalty parameter, 
and show that these properties hold even if 
the model is misspecified. 

Consider a possibly cyclic linear SEM over the variables
$(Y,X,H,A)$ of the form
\begin{align*} 
	\begin{bmatrix}
		Y & X^\t & H^\t \end{bmatrix} 
	:=  \begin{bmatrix}
		Y &X^\t  & H^\t  \end{bmatrix} B   +  A^\t M+ \ep^\t,
\end{align*}
subject to regularity conditions that ensure that the distribution of $(Y,X,H,A)$ is well-defined.
Here, B and M are constant matrices, the random vectors $A$ and $\ep$ are defined on a common probability space $(\Omega, \cF,P)$, $Y$ is the endogenous target for the single equation inference, $X$ are the observed endogenous variables, $H$ are hidden endogenous variables and $A$ are exogenous variables independent from the unobserved noise innovations $\ep$. 

SEMs allow for the notion of interventions, i.e., modeling external manipulations of the system. In this work, we are only concerned with interventions on the exogenous variables $A$ of the form $\text{do}(A:=v)$. 
Because $A$ is exogeneous, these interventions can be defined as follows: they change the distribution of $A$ to that of a random vector $v$. The
interventional distribution of the variables $(Y,X,H,A)$ under the intervention $\text{do}(A:=v)$ is given by the simultaneous distribution of $(X_v, Y_v,H_v,v)$ generated by the SEM 

\begin{align*}
	\begin{bmatrix}
		Y_v & X_v^\t & H_v^\t \end{bmatrix} := \begin{bmatrix}
		Y_v & X_v^\t & H_v^\t \end{bmatrix} B   + v^\t M+ \ep.
\end{align*}
Thus, the intervention does not change any of the original structural assignments of the endogenous variables. 
Instead, the change in the distribution of the exogeneous variable propagates through the system. We henceforth let $E^{\mathrm{do}(A:=v)}$ denote the expectation with respect to the interventional distribution of the system under the intervention $\mathrm{do}(A:=v)$. More details on interventions can be found \Cref{sec:simsem}

  Let $(\fY,\fX,\fH,\fA)$ consist of $n$ row-wise independent and identically distributed copies of the random vector $(Y,X,H,A)$ and consider the single equation of interest
\begin{align*} 
	\fY = \fX \gamma_0 + \fA \beta_0  + \fH \eta_0 + \bm{\ep}_Y = \fX \gamma_0 + \fA \beta_0  + \tilde{\fU}_Y.
\end{align*}
The K-class estimator with parameter $\kappa$ using non-sample information that only $\fZ_* \subset [\fX \, \, \fA]$ have non-zero coefficients in the target equation of interest is given by
\begin{align*}
	\hat{\alpha}_{\text{K}}^n(\kappa) =  (\fZ_*^\t (I-\kappa P_\fA^\perp)\fZ_*)^{-1} \fZ_*^\t(I-\kappa P_\fA^\perp)\fY,
\end{align*}
where $P_\fA^\perp$ is the projection onto the orthogonal complement of the column space of $\fA$. 
For a fixed $\kappa \in [0,1)$ K-class estimators can be represented by a penalized regression problem $\hat{\alpha}_{\text{K}}^n(\kappa)  = \argmin_{\alpha} l_{\mathrm{OLS}}^n(\alpha)+ \kappa/(1-\kappa)l_{\mathrm{IV}}^n(\alpha)$, where $l_{\mathrm{OLS}}^n$ and $l_{\mathrm{IV}}^n$ are the empirical OLS and TSLS loss functions, respectively. This representation and the ideas of \cite{rothenhausler2018anchor} allow us to prove that K-class estimator converges to a coefficient that is minimax optimal when considering all distributions induced by a certain set of interventions of $A$. More specifically, we show that for a fixed $\kappa$ and regardless of identifiability,
\begin{align*}
	\hat{\alpha}_{\text{K}}^n(\kappa) \overset{P}{\underset{n\to\i }{\longrightarrow}} \argmin_{\alpha}\sup_{v\in C(\kappa)}  E^{\mathrm{do}(A:=v)}\left[ (Y - \alpha^\t Z_*   )^2 \right],
\end{align*}
where $
C(\kappa) := \{ v:\Omega \to \R^{q}:  \mathrm{Cov}(v,\ep)=0, \,   E    [ vv^\t ]  \preceq \frac{1}{1-\kappa} E[AA^\t] \}$. 
The argmin on the right-hand side 
minimizes the worst case prediction error when considering 
interventions up to a certain strength (measured by the set $C(\kappa)$). 
This objective becomes relevant when we consider a response variable with several covariates and
aim to minimize the mean squared prediction error of future realizations of the system of interest that do not follow the training distribution. 
The above result says that 
if the new realizations correspond to (unknown)
interventions on the exogenous variables that are of bounded strength, K-class estimators with fixed $\kappa\in (0,1)$ 
minimize the worst case prediction performance and, in particular, outperform the true causal parameter and the least squares solution (see also \Cref{fig:DistRobustness} in \Cref{app:DistributionalRobustness}).
For $\kappa$ approaching one, we recover the guarantee of the causal solution and for $\kappa$ approaching zero, the set of distributions contains the training distribution.
The above minimax property therefore adds to the discussion 
whether 
non-consistent
K-class estimators with 
penalty parameter not 
converging to 
one
can be useful; see, e.g.\ \cite{dhrymes1974}.


\subsubsection{The PULSE Estimator} \label{sec:summarypuls}
\Cref{SEC:PULSE} contains the second main contribution in this work. We
propose a novel data driven K-class estimator for causal coefficients,
which we call the p-uncorrelated least square estimator (PULSE). 
As above, we consider
a single endogenous target 
in an SEM (or simultaneous equation model)
and aim to predict it 
from observed predictors that are with a priori (non-sample) information known to be either endogenous or exogenous. The PULSE estimator can be written in several equivalent forms. It can, first, be seen as a data-driven K-class estimator 
\begin{align*}
	\hat{\alpha}^n_{\mathrm{K}}( \lambda^\star_n/(1+\lambda^\star_n)  ) = \argmin_{\alpha} l_{\mathrm{OLS}}^n(\alpha) + \lambda^\star_n l_{\mathrm{IV}}^n(\alpha) ,
\end{align*}
where
\begin{align*}
\lambda^\star_n
:= \inf \left\{ \lambda > 0\,:\,  \begin{tabular}{c}
		\text{testing }\text{Corr}$(A,Y-Z\hat{\alpha}_{\text{K}}^n(\lambda/(1+\lambda)))=0$ \\
		 \text{yields a }p\text{-value }$ \geq p_{\min}$
	\end{tabular}
 \right\},
\end{align*}
for some pre-specified level of the hypothesis test $p_{\min}\in(0,1)$. 
In words, the PULSE estimator outputs the K-class estimator closest to the OLS while maintaining a non-rejected test of uncorrelatedness.
In principle, PULSE can be used with any testing procedure.
The choice of test, however, may influence the difficulty of the 
resulting optimization problem. In this paper, we investigate PULSE in connection with a specific class of hypothesis tests that, for example, contain the test of \cite{anderson1949estimation}.  For these hypothesis tests we develop an efficient 
and 
provably correct optimization method, that is based on binary line search and quadratic programming.

We show that our estimator can, second, 
be written as the solution to a constrained optimization problem. To that end, define the primal problems
\begin{align*} 
\hat{\alpha}_{\text{Pr}}^n(t) :=  \begin{array}{ll}
\argmin_{\alpha} & l_{\mathrm{OLS}}^n(\alpha)  \\
\text{subject to} & l_{\mathrm{IV}}^n(\alpha)\leq t.
\end{array}
\end{align*}
For the choice $t^\star_n := \sup\{t\,:\, \text{testing } \mathrm{Corr}(A,Y-Z\hat{\alpha}_{\text{Pr}}^n(t))=0$ $\text{yields a}$ $p\text{-value} \geq p_{\min}\}$,
we provide a detailed analysis proving that  
$\hat{\alpha}^n_{\mathrm{K}}
(\lambda^\star_n/(1+\lambda^\star_n) ) 
= \hat{\alpha}_{\text{Pr}}^n(t^\star_n)$. 

For the 
testing procedure proposed in this paper, we show that, third, 
PULSE can be written as 
	\begin{align*} 
	\begin{array}{ll}
		\text{argmin}_\alpha & l_{\mathrm{OLS}}^n(\alpha;\fY,\fZ)  \\
		\text{subject to} & \alpha \in \cA_n(1-p_{\min}),
	\end{array}
\end{align*}
where $\cA_n(1-p_{\min})$ is the non-convex acceptance region for our test of uncorrelatedness.

This third formulation allows for a simple interpretation of our estimator: among all coefficients (not restricted to K-class estimators) that do not yield a rejection of uncorrelatedness, we choose the one that yields the best prediction. 
If the acceptance region is empty it outputs a warning indicating a possible model misspecification or an assumption violation to the user (in that case, one can formally output another estimator such as TSLS or Fuller, yielding PULSE well-defined).

In the just-identified setup, the TSLS estimator solves a 
normal equation which is equivalent to setting a sample covariance between the instruments and the resulting prediction residuals to zero; it then corresponds to $t=0$. 
For this (and the over-identified) setting, we prove that PULSE is a consistent estimator for the causal coefficient. 

The TSLS does not have a finite variance if there is insufficient degree of overidentification, for example. 
In particular for weak instruments, this usually comes with poor finite sample performance. In such cases, however, the acceptance region of uncorrelatedness is usually large. This yields a weak constraint in the optimization problem and the PULSE will be closer to the OLS, which in certain settings suffers from less variability \citep[see, e.g.,][]{hahn2004estimation,HAHN05}. 
In simulations we indeed see that, similarly to other data-driven K-class estimators that are pulled towards the OLS, such as Fuller estimators, the PULSE comes with beneficial finite sample properties compared to TSLS and LIML.


Unlike other estimators such as LIML or the classical TSLS, the PULSE is well-defined in under-identified settings, too. Here, its objective is still to find the best predictive solution among all parameters that do not reject uncorrelatedness. 
Uncorrelatedness to the exogeneous variable is sometimes referred to as invariance. 
The idea of choosing the best predictive among all invariant models has been investigated in several works \citep[e.g.][]{pfister2019stabilizing, rojas2018invariant, Magliacane2018}  with the motivation to 
find models that generalize well (in particular, with respect to interventions on the exogenous variables). 
Existing methods, however, focus on selecting subsets of variables and then consider least squares regression of the response variable onto the full subset. 
PULSE can recover such type of solutions if they are indeed optimal. 
But it also allows to search over coefficients that are different from least squares regression for sets of variables. 
Consequently, PULSE allows us to find solutions in situations, where the above methods would not find any invariant subsets, which may often be the case if there are hidden variables (see \Cref{app:underidentifiedexperiment} for an example).

We show in a simulation study that there are several settings in which 
PULSE outperforms existing estimators both in terms of MSE ordering and several one-dimensional scalarizations of the MSE. More specifically, we show that PULSE can outperform the TSLS and Fuller estimators in weak instrument situations, for example, where Fuller estimators are known to
have good MSE properties; see, e.g.\ \citet{hahn2004estimation} and \citet{stock2002survey}.

Implementation of  PULSE and code for experiments (R) are available on GitHub.\footnote{
	\url{https://github.com/MartinEmilJakobsen/PULSE}}

\section{Robustness Properties of K-class Estimators} \label{SEC:ROBUSTNESS}
\setcounter{equation}{0}
In this section 
we consider
 K-class 
estimators (\citealp{theil1958economic}, and \citealp{nagar1959bias})
and show a connection with anchor regression  of \citet{rothenhausler2018anchor}. 
In \Cref{sec:KclassInIVModelCoincidesWithAr} we establish the connection 
in models where we use \textit{a priori} information that there are no included exogenous variables in the target equation of interest. 
In \Cref{sec:KclassAsPenalizedRegression} we then show that general
K-class estimators can be written as the solution to a penalized regression problem. In \Cref{sec:intervrobustnessofKclass} we utilize this representation and the ideas of \citet{rothenhausler2018anchor} to prove a distributional robustness guarantee of general K-class estimators with fixed $\kappa\in[0,1)$,
even under model misspecification and non-identifiability. 	Proofs of results in this section can be found in \Cref{sec:RobustnessProofs}.

\subsection{Setup and Assumptions} \label{sec:SetupAndAssumptionsRobustness}
Denote the random vectors $Y\in \R, X\in \R^{d}, A\in \R^{q},H\in \R^r$  and $\ep\in \R^{d+1+r}$ 
by the target, endogenous regressor, anchors, hidden and noise variables, respectively.
Let further 
$(Y,X,H)$ be generated by the possibly cyclic  structural equation model (SEM)
\begin{align} \label{ARModel}
	\begin{bmatrix}
		Y & X^\t & H^\t \end{bmatrix} 
	:=  \begin{bmatrix}
		Y &X^\t  & H^\t  \end{bmatrix} B   +  A^\t M+ \ep^\t,
\end{align}
for some random vectors $\ep\independent A$ and constant matrices $B$ and $M$.  Let $(\fY,\fX,\fH,\fA)$ consist of $n\geq \min\{d,q\}$ row-wise independent and identically distributed copies of the random vector $(Y,X,H,A)$. Solving for the endogenous variables we get the structural and reduced form equations $[\,\fY\,\, \fX\,\, \fH\,]\, \Gamma = \fA M + \bm{\ep}$ and $[\,\fY\,\, \fX\,\, \fH\,] = \fA \Pi + \bm{\ep} \Gamma^{-1}$, 
where $\Gamma := I-B$ and  $\Pi := M \Gamma^{-1}$.  Assume without loss of generality that $\Gamma$ has a unity diagonal, such that the target equation of interest is given by
\begin{align} \label{eq:StructuralEquationOfInterest}
	\fY = \fX \gamma_0 + \fA \beta_0  + \fH \eta_0 + \bm{\ep}_Y = %
	\fZ \alpha_0 + \tilde{\fU}_Y,
\end{align}
where $(1, -\gamma_0,- \eta_0)\in \R^{(1+d+r)}$, $ \beta_0\in \R^q$ and $\bm{\ep}_Y$  are the first columns of $\Gamma$, $M$ and $\bm{\ep}\in \R^n$ respectively, $\fZ := [
\fX \, \, \fA]$, $\alpha_0 = (\gamma_0, \beta_0)\in \R^{d+q}$ and $\tilde{\fU}_Y := \fH\eta_0 + \bm{\ep}_Y$.  

The possible dependence between the noise $\tilde \fU_Y$ and the endogenous variables, i.e., the influence by hidden variables, generally,
renders the standard OLS approach for estimating $\alpha_0$ inconsistent. 
Instead, one can make use of the components in $A$ that have vanishing coefficient in 
\Cref{eq:StructuralEquationOfInterest} 
for consistent estimation.  In the remainder of this work, we disregard any \textit{a priori} (non-sample) 
information not concerning the target equation.
The question of identifiability of $\alpha_0$ has been studied extensively \citep[][]{frisch38,Haavelmo1944,measuringtheequationsystems} and more recent overviews can be found in, e.g., 
\citet{Didelez2010}, \citet{fisher1966identification}, and \citet{greene2003econometric}.

We will use the following assumptions concerning the structure of the SEM:

\setcounter{assumption}{0}
\begin{assumption}[Global assumptions] \label{ass:global}
	\begin{enumerate*}[label=(\alph*),ref=\ref{ass:global}.(\alph*)]
		\item 
		$(Y,X,H,A)$ is generated in accordance with the SEM in \Cref{ARModel}; 
		\label{ass:linearSEM}
				\item $\rho(B)<1$ where $\rho(B)$ is the spectral radius of $B$;  \label{ass:SpectralRadiusOfBLessThanOne}
				\item $\ep$ has jointly independent marginals $\ep_1, \ldots, \ep_{d+1+r}$;\label{ass:epIndependentMarginals}
				\item $A$ and $\ep$ are independent; \label{ass:AindepEp}
				\item No variable in $Y$, $X$ and $H$ is an ancestor of $A$, that is, $A$ is exogenous;
				 \label{ass:Aexogenous}
				\item $E[\|\ep\|^2_2]$, $E[\|A\|^2_2] < \i$; \label{ass:SecondMomentEp}
			 \label{ass:SecondMomentA}
				\item $E[\ep] =0$. \label{ass:EpMeanZero}
					\item $\text{Var}(A) \succ 0$, i.e., the variance matrix of $A$ is positive definite; \label{ass:VarianceOfAPositiveDefinite}
						\item $\fA^\t \fA$ is almost surely of full rank;  \label{ass:AtAfullRank}
			\end{enumerate*}
\end{assumption}
\begin{assumption}[Finite sample  assumptions] \label{ass:finiteass}
	\begin{enumerate*}[label=(\alph*),ref=\ref{ass:finiteass}.(\alph*)]
		\item 	$\fZ_*^\t \fZ_*$ is almost surely of full rank;  \label{ass:ZtZfullRank}
		\item $\fA^\t \fZ_*$ is almost surely of full column rank. \label{ass:AtZfullRank}
				\item $\fX^\t \fX$ is almost surely of full rank; \label{ass:XtXfullRank}
	\end{enumerate*}
\end{assumption}
\begin{assumption}[Population assumptions] \label{ass:popass}
	\begin{enumerate*}[label=(\alph*),ref=\ref{ass:popass}.(\alph*)]
		\item $\text{Var}(Z_*) \succ 0$, i.e., the variance matrix of $Z_*$ is positive definite;\label{ass:VarianceOfZPositiveDefinite}
		\item 	$E[AZ_*^\t]$ is of full column rank. \label{ass:EAZtFullColumnRank}
	\end{enumerate*}
\end{assumption}
We will henceforth assume that \Cref{ass:global} always holds. This assumption ensure that the SEM and that the TSLS objectives are well-defined.
In the above assumptions, $Z_*$ and $\fZ_*$ are 
generic placeholders for a subset of endogenous and exogenous variables from 
$[
X^\t \, \,  A^\t
]^\t$ 
and
$[
\fX \, \,  \fA
]$, respectively,
which should be clear from the context in which they are used.  Both Assumption\Cref{ass:AtAfullRank} and Assumption\Cref{ass:XtXfullRank} hold if $X$ and $A$ have density with respect to Lebesgue measure, which in turn is guaranteed by Assumption\Cref{ass:AindepEp} if $A$ and $\ep$ have density with respect to Lebesgue measure.
Assumption\Cref{ass:VarianceOfAPositiveDefinite} and \Cref{ass:AtAfullRank} implies that the instrumental variable objective functions introduced below is almost surely well-defined and Assumption\Cref{ass:XtXfullRank} yields that the ordinary least square solution is almost surely well-defined. Assumption\Cref{ass:SecondMomentA,ass:SecondMomentEp} implies that $Y,X$ and $H$ all have finite second moments. 
	For Assumption\Cref{ass:EAZtFullColumnRank} and \Cref{ass:AtZfullRank} it is necessary that $q\geq \mathrm{dim}(Z_*)$, i.e., that the setup must be just- or over-identified; see \Cref{sec:SetupAndAssumptionsPULSE} below.
\subsection{Distributional Robustness of Anchor Regression} \label{sec:AR}
 \citet{rothenhausler2018anchor} proposes a method, called anchor regression,
for predicting 
the endogenous target variable $Y$ 
from the endogenous variables $X$. %
The collection of exogenous variables $A$,
called anchors, are not included in that prediction model. 
Anchor regression trades off predictability and invariance
by considering a convex combination of 
the ordinary least square (OLS) loss function and
the two-stage least square (IV) loss function using the anchors as instruments.
More formally, we define 
	\begin{align} \label{eq:lossPop}
	l_{\mathrm{OLS}}(\gamma ;Y,X) &:= E(Y-\gamma^\t X)^2, \\
	l_{\mathrm{IV}}(\gamma ;Y,X,A) 
	&:=E(A(Y-\gamma^\t X))^\t   E(A A^\t)^{-1} E(A(Y- \gamma^\t X)), \notag \\ \label{eq:lossEmp}
	l^n_{\mathrm{OLS}}(\gamma;\fY,\fX) &:=  n^{-1} (\fY-\fX \gamma)^\t (\fY-\fX \gamma),\\
	l^n_{\mathrm{IV}}(\gamma;\fY,\fX, \fA) &:=n^{-1} (\mathbf{Y}-\mathbf{X}\gamma)^\t  P_\fA (\mathbf{Y}-\mathbf{X}\gamma),
	\end{align}
the population and finite sample versions of the loss functions.
$P_\fA = \fA(\fA^\t \fA)^{-1}\fA^\t$
is the orthogonal projection onto the column space of $\fA$. 
To simplify notation, we omit the dependence on $Y$, $X$, $A$, $\fA$, $\fX$ or $\fY$ when they are clear from a given context.
%
%
%
%
%
%
For a penalty parameter $\lambda> -1$, 
the anchor regression coefficients are defined as
	\begin{align} 
	\gamma_{\mathrm{AR}}(\lambda)&:= 
	\argmin_{\gamma\in \R^d} \{l_{\text{OLS}}(\gamma)+\lambda l_{\text{IV}}(\gamma) \},	\quad
	\label{eq:EmpiricalARestimator}
	\hat{\gamma}_{\mathrm{AR}}^n(\lambda):= 
	\argmin_{\gamma\in \R^d} \{l_{\text{OLS}}^n(\gamma)+\lambda l_{\text{IV}}^n (\gamma)\}.
	\end{align}
The estimator $\hat{\gamma}_{\text{AR}}^n(\lambda)$ consistently estimates the population estimand $\gamma_{\text{AR}}(\lambda)$
and
minimizes prediction error while simultaneously penalizing a transformed sample covariance between the anchors and the resulting prediction residuals. 
Unlike the TSLS estimator, for example, the anchor regression estimator is 
 almost surely well-defined under the rank condition of Assumption\Cref{ass:XtXfullRank}, even if the model is under-identified, that is, there are less exogenous than endogenous variables. The solution to the empirical minimization problem of anchor regression is given by
\begin{align} \label{eq:ARsolution}
\hat{\gamma}_{\text{AR}}^n(\lambda) = [\fX^\t (I+\lambda P_\fA)\fX]^{-1}\fX^\t(I+\lambda P_\fA)\fY,
\end{align}
which follows from solving the normal equation of \Cref{eq:EmpiricalARestimator}.

The motivation of anchor regression
is not to %
infer a causal parameter.
Instead, for a fixed penalty parameter 
$\lambda$,
the estimator is shown to
possess a distributional or interventional robustness 
property: the estimator is 
optimal when predicting under interventions on the exogenous 
variables that are below a certain intervention strength. 
%
%
%
%
%
%
%
%
%
%
%
%
By Theorem 1 of \citet{rothenhausler2018anchor} it holds that $$
\gamma_{\text{AR}}(\lambda) = \argmin_{\gamma\in \R^d} \sup_{v\in C(\lambda)} E^{\text{do}(A:=v)}\left[ \lp Y-\gamma^\t X \rp^2	\right],$$ 
where $
C(\lambda) := \lb  v:\Omega \to \R^q : \text{Cov}(v,\ep)=0, E(v v^\t) \preceq (\lambda+1) E(A A^\t )  \rb.$ 

%
\subsection{Distributional Robustness of K-class Estimators} \label{sec:RobustnessOfKclass}
We now introduce the limited information estimators known as K-class estimators  (\citealp{theil1958economic}, and \citealp{nagar1959bias}) used
for single equation inference. 
%
Suppose that we are given non-sample information about which components of $\gamma_0$ and $\beta_0$, of \Cref{eq:StructuralEquationOfInterest}, are zero. 
We can then 
partition $\fX = [
\fX_* \, \, \fX_{-*}
] \in \R^{n\times(d_1+d_2)}$, $\fA = [\fA_{*} \, \,  \fA_{-*}]\in \R^{n\times(q_1+q_2)}$ and $\fZ = [
\fZ_* \, \, \fZ_{-*}
] = [
\fX_{*} \, \,  \fA_* \, \,  \fX_{-*}  \, \,  \fA_{-*}
]$
with $ \fZ \in \R^{n\times((d_1+q_1)+(d_2+q_2))}$, where $\fX_{-*}$ and $\fA_{-*}$ corresponds to the variables for which 
our non-sample information states that 
the components of $\gamma_0$ and $\beta_0$ are zero, respectively. 
We call the variables corresponding to $\fA_*$ included exogenous variables. 
Similarly, we write
$\gamma_{0} =(\gamma_{0,*},\gamma_{0,-*}) $, $\beta_0 = (\beta_{0,*},\beta_{0,-*})$ and  $\alpha_0 = (
\alpha_{0,*}, \alpha_{0,- *}) =(\gamma_{0,*},\beta_{0,*},\gamma_{0,-*},\beta_{0,-*})$.  The structural equation of interest then reduces to $
\fY = \fX_* \gamma_{0,*} + \fX_{-*} \gamma_{0,-*} +  \fA_* \beta_{0,*} +\fA_{-*} \beta_{0,-*} + \tilde{\fU}_Y = \fZ_* \alpha_{0,*} + \fU_Y$, 
where $\fU_Y = \fX_{-*} \gamma_{0,-*} + \fA_{-*} \beta_{0,-*}+ \fH\eta_0 + \bm{\ep}_Y$.
In the case that the non-sample information is indeed correct, we have that $\fU_Y = \tilde{\fU}_Y= \fH\eta_0 + \bm{\ep}_Y$. When well-defined, the K-class estimator with 
parameter $\kappa\in \R$ for a simultaneous estimation of $\alpha_{0,*}$ is given by

\begin{align} \label{eq:KclassEstimatorInSingleLine}
\hat{\alpha}_{\text{K}}^n(\kappa;\fY,\fZ_*,\fA) =  (\fZ_*^\t (I-\kappa P_\fA^\perp)\fZ_*)^{-1} \fZ_*^\t(I-\kappa P_\fA^\perp)\fY,
\end{align}
where 
$I-\kappa P_\fA^\perp = I- \kappa(I-P_\fA) = (1-\kappa)I + \kappa P_\fA$. 
Comparing \Cref{eq:ARsolution,eq:KclassEstimatorInSingleLine}
 suggests a close connection 
between anchor regression 
and K-class estimators for inference of structural equations with no included exogenous variables. In the following subsections, we establish this connection and subsequently extend the distributional robustness
property to general K-class estimators.

\subsubsection{K-class Estimators in Models with no Included Exogenous Variables}

 \label{sec:KclassInIVModelCoincidesWithAr}
Assume that,
in addition to \Cref{ass:global}, we have the 
non-sample information that $\beta_0=0$, that is, no exogenous variable in $A$ directly affects the target variable $Y$. 
By direct comparison we see that the K-class estimator for $\kappa<1$ coincides with the anchor regression estimator with penalty parameter $\lambda =\kappa/(1-\kappa)$, i.e., $\hat{\gamma}_{\text{K}}^n(\kappa)=  \gamma_{\mathrm{AR}}^n \lp\frac{\kappa }{1-\kappa} \rp$.
Equivalently, we have $\gamma_{\mathrm{AR}}^n \lp \lambda \rp =  \gamma_{\text{K}}^n \lp\lambda /(1+\lambda) \rp$ for any  $\lambda >-1$. As such, the K-class estimator, for a fixed $\kappa$,  inherits the following distributional robustness property:
\begin{align} \label{eq:robIV}
\gamma_{\text{K}}(\kappa) &=  \gamma_{\mathrm{AR}} \lp\frac{\kappa }{1-\kappa} \rp = \argmin_{\gamma\in \R^d} \sup_{v\in C(\kappa/(1-\kappa))} E^{\mathrm{do}(A:=v)} \left[ \lp Y-\gamma^\t X\rp^2 \right],
\end{align}
where $
C(\kappa/(1-\kappa)) = \{  v:\Omega \to \R^q : \text{Cov}(v,\ep)=0, E[v v^\t] \preceq \frac{1}{1-\kappa} E[A A^\t]  \}$. 
This statement holds by Theorem 1 of \citet{rothenhausler2018anchor}. 

In an identifiable model with $P \lim_{n\to \i}\kappa =1$ we have that $\hat{\gamma}^n_{\text{K}}(\kappa)$
consistently estimates the causal parameter; see  e.g.\ \citet{mariano2001simultaneous}. 
For such
a choice of $\kappa$, the robustness above is just a weaker version of what the causal coefficient can guarantee. 
However, the above result in~\Cref{eq:robIV}
establishes a robustness property for fixed $\kappa <1$, even in cases where the model is not identifiable. 
Furthermore,
since we did not use that the non-sample information that $\beta_0=0$ was true, 
the robustness property is resilient to model misspecification in terms of excluding included exogenous variables from the target equation which generally also breaks identifiability. 

\subsubsection{The K-class Estimators as Penalized Regression Estimators} \label{sec:KclassAsPenalizedRegression}
We now show that 
general K-class estimators 
can be written as
solutions to penalized regression problems. 
The first appearance of such a representation is, to the best of our knowledge, due to \citet{mcdonald1977k} building upon previous work of \citet{basmann1960finite,basmann1960asymptotic}. Their representation, however, 
concerns only the endogenous part $\gamma$.
We require a slightly different statement and
will show that the entire K-class estimator of $\alpha_{0,*}$, i.e., the simultaneous estimation of $\gamma_{0,*}$ and $\beta_{0,*}$,
 can be written as a penalized regression problem. 
Let therefore $l_{\mathrm{IV}}(\alpha;\fY,\fZ_*,\fA)$, $l_{\mathrm{IV}}^n(\alpha;\fY,\fZ_*,\fA)$ and $l_{\mathrm{OLS}}(\alpha;\fY,\fZ_*)$, $l_{\mathrm{OLS}}^n(\alpha;\fY,\fZ_*)$ denote the population and empirical TSLS and OLS loss functions as defined in \Crefrange{eq:lossPop}{eq:lossEmp}. That is, the TSLS loss function for regressing $\fY$ on the included endogenous and exogenous variables $\fZ_*$ using the exogeneity of $\fA$ and $\fA_{-*}$ as instruments and the OLS loss function for regressing $\fY$ on $\fZ_*$. 
We define the K-class population and finite-sample loss functions as 
an affine combination of the two loss functions above. That is, 
\begin{align} \label{KclassLossFunctionPop}
l_{\mathrm{K}}(\alpha;\kappa,Y,Z_*,A)&= (1-\kappa)l_{\mathrm{OLS}}(\alpha;Y,Z_*) + \kappa l_{\mathrm{IV}}(\alpha;Y,Z_*,A), \\ \label{KclassLossFunctionEmp}
l_{\mathrm{K}}^n(\alpha;\kappa,\fY,\fZ_*,\fA)&= (1-\kappa)l_{\mathrm{OLS}}^n(\alpha;\fY,\fZ_*) + \kappa l_{\mathrm{IV}}^n(\alpha;\fY,\fZ_*,\fA). 
\end{align}
\begin{restatable}[]{proposition}{PenalizedKClassSolutionUniqueAndExists}
\label{lm:PenalizedKClassSolutionUniqueAndExists}%
   Consider one of the following scenarios: 1) $\kappa <1$ and \Cref{ass:ZtZfullRank} holds, or 2) $\kappa = 1$ and \Cref{ass:AtZfullRank} holds.
The estimator minimizing the empirical loss function of \Cref{KclassLossFunctionEmp} is almost surely well-defined  and coincides with the K-class estimator of \Cref{eq:KclassEstimatorInSingleLine}. That is, it almost surely holds that
	\begin{align} \label{eq:kclassminim}
	\hat{\alpha}_{\mathrm{K}}^n(\kappa;\fY,\fZ_*,\fA) =\argmin_{\alpha\in \R^{d_1+q_1}} l_{\mathrm{K}}^n(\alpha;\kappa,\fY,\fZ_*,\fA).
	\end{align}
\end{restatable}

Assuming $\kappa\not =1$, we can rewrite \Cref{eq:kclassminim} to
\begin{align}
\label{eq:KclassLossFunctionAsPenalizedOLS}
	\hat{\alpha}_{\text{K}}^n(\kappa;\fY,\fZ_*,\fA) = \argmin_{\alpha\in \R^{d_1+q_1}} \{  l^n_{\mathrm{OLS}}(\alpha;\fY,\fZ_*) + \frac{\kappa}{1-\kappa} l^n_{\mathrm{IV}}(\alpha;\fY,\fZ_*,\fA) \}.
\end{align}
Thus, 
K-class estimators
seek to minimize the ordinary least squares loss for regressing $\fY$ on  $\fZ_*$, while simultaneously penalizing the strength of a transform on the sample covariance between  
the prediction residuals and  
collection of 
exogenous variables $\fA$.

 In the following section, we 
 consider a population version of the above quantity.
If we replace the finite sample \Cref{ass:finiteass}
with the 
corresponding population \Cref{ass:popass}, we get that the minimization estimator of the empirical loss function of \Cref{KclassLossFunctionEmp} is asymptotically well-defined. Furthermore, 
we now prove that
whenever the population assumptions are satisfied, then,
for any fixed $\kappa \in [0,1]$, 
 $\hat{\alpha}_{\text{K}}^n(\kappa;\fY,\fZ_*,\fA)$ 
converges in probability towards the population K-class estimand.

\begin{restatable}[]{proposition}{PopulationPenalizedKClassSolutionUniqueAndExists}
	\label{lm:PopulationPenalizedKClassSolutionUniqueAndExists}%
	Consider one of the following scenarios: 1) $\kappa \in [0,1)$ and \Cref{ass:VarianceOfZPositiveDefinite} holds, or 2) $\kappa = 1$ and \Cref{ass:EAZtFullColumnRank} holds.
It holds that  $(\hat{\alpha}_{\text{K}}^n(\kappa;\fY,\fZ_*,\fA))_{n\geq 1}$ is an asymptotically well-defined sequence of estimators. 
Furthermore, the sequence consistently estimates the well-defined population K-class estimand. That is,

	$$
	\hat{\alpha}_{\mathrm{K}}^n(\kappa;\fY,\fZ_*,\fA) \overset{P}{\underset{n\to\i }{\longrightarrow}} \alpha_{\mathrm{K}}(\kappa;Y, Z_*,A) := \argmin_{\alpha\in \R^{d_1+q_1}}l_{\mathrm{K}}(\alpha;\kappa,Y,Z_*,A).
	$$
\end{restatable}

\subsubsection{Distributional Robustness of General K-class Estimators}
\label{sec:intervrobustnessofKclass}
We are now able to prove that the 
general K-class estimator
possesses a robustness property similar to the 
statements above.
It is prediction optimal under a set of interventions, now 
including interventions on all
exogenous $A$ up to a certain strength. 
%
%
%
%
%
%
%
%
%
%
%
%
%
%
%
%
\begin{restatable}[]{theorem}{TheoremIntRobustKclass}
	\label{sthm:TheoremIntRobustKclas}%
    Let \Cref{ass:global} hold. 
    For any fixed $\kappa \in[0,1)$ and $Z_*=(X_*, A_*)$ with $X_*\subseteq X$ and $A_* \subseteq A$, we have, whenever the population K-class estimand is well-defined, that 
	\begin{align*}
	\alpha_{\mathrm{K}}(\kappa;Y,Z_*,A)&= \argmin_{\alpha\in \R^{d_1+q_1}}\sup_{v\in C(\kappa)}  E^{\mathrm{do}(A:=v)}\left[ (Y - \alpha^\t Z_*   )^2 \right], 
	\end{align*}
	where $
	C(\kappa) := \lb v:\Omega \to \R^{q}:  \mathrm{Cov}(v,\ep)=0, \,   E    [ vv^\t ]  \preceq \frac{1}{1-\kappa} E[AA^\t] \rb$.
\end{restatable}
Here, $E^{\text{do}(A:=v)}$ denotes the expectation with respect to the distribution entailed under the intervention $\text{do}(A:=v)$ 
(see Section~\ref{sec:intro:robustness} and \Cref{sec:simsem})
and $(\Omega,\cF,P)$ is the common background probability space on which $A$ and $\ep$ are defined.

In words, 
among all linear prediction methods of $Y$ using $Z_*$ as predictors, 
the 
K-class estimator 
with parameter $\kappa$
has the lowest possible worst case mean squared prediction  error when considering 
all interventions on the exogenous variables $A$ contained in $C(\kappa)$. 
As $\kappa$ approaches one, the
estimator is prediction optimal under 
a class of arbitrarily strong
interventions 
in the direction of the variance of $A$. (Here, $\kappa$ is arbitrary but fixed; the statement does not cover data-driven choices of $\kappa$, such as LIML or Fuller.)
The above result is a consequence of the relation between anchor regression and 
K-class estimators.
The special case $A_* = \emptyset$ 
is a consequence of Theorem~1 by \citet{rothenhausler2018anchor}. 
Our 
proof 
follows similar arguments but 
additionally allows for $A_* \not = \emptyset$.
%

The property in Theorem~\ref{sthm:TheoremIntRobustKclas}
has a decision-theoretic interpretation (see \citet{Chamberlain2007} for an application of decision theory in IV models based on another loss function).
Consider a response $Y$, covariates $Z_*$ and a distribution (specified by $\theta$) over $(Y, Z_*)$,
and the squared loss $\ell(Y, Z, \alpha) := (Y-\alpha^\top Z_*)^2$.
Then, assuming finite variances, for each distribution the risk 
$E_{\theta}[(Y-\alpha^\top Z_*)^2]$ is minimized by the (population) OLS solution
$\alpha=\alpha_\theta := \mathrm{cov}_\theta(Z_*)^{-1}\mathrm{cov}_\theta(Z_*,Y)$.
In the setting of Theorem~\ref{sthm:TheoremIntRobustKclas}, we are given a distribution over $(Y,Z_*)$, specified by $\theta$,  but we are 
interested in minimizing the risk
$E_{\theta, v}[(Y-\alpha_{\theta}^\top Z_*)^2]$ 
for another 
 distribution that is induced by an intervention and specified by $(\theta, v)$. 
The above result states that the K-class estimator minimizes a worst-case risk when considering all $v \in C(\kappa)$.

Theorem~\ref{sthm:TheoremIntRobustKclas}
makes use of the language of SEMs in that it yields the notion of interventions.\footnote{In particular, we have not considered the SEM as a model for counterfactual statements.}
As such,  the result can be rephrased using other
causal frameworks. The crucial assumptions are the exogeneity of $A$ and the linearity of the system. 
Furthermore, the result is robust with respect to 
several types of model misspecifications that breaks identifiability of $\alpha_0$, such as excluding included endogenous or exogenous predictors or the existence of
latent variables; see \Cref{rm:ModelMispecification} in \Cref{app:AddRemarks}.

\section{The P-Uncorrelated Least Square Estimator} \label{SEC:PULSE}
\setcounter{equation}{0}
We now introduce 
the p-uncorrelated least square estimator (PULSE). 
As discussed in \Cref{sec:summary}, PULSE allows for different representations. In this section we start with the third representation and show the equivalence of the other representations afterwards.

Consider predicting the target $Y$ from endogenous and possibly exogenous regressors $Z$. Let therefore
$\cH_0(\alpha)$ denote the hypothesis that the prediction residuals using $\alpha$ as a regression coefficient is simultaneously uncorrelated with every exogenous variable, that is, $\cH_0(\alpha) : \text{Corr}(A,Y-\alpha^\t Z) =0$.
This hypothesis is in some models under certain conditions equivalent to the hypothesis that $\alpha$ is the true causal coefficient. One of these conditions is the rank condition \Cref{ass:EAZtfullrank} introduced below, also known as the rank condition for identification; \citet{wooldridge2010econometric}.

The two-stage least square (TSLS) estimator exploits the equivalence between the causal coefficient and the zero correlation between the instruments and the regression residuals. Here, one minimizes
a sample covariance between the instruments and the regression residuals: we can write
$l^n_{\mathrm{IV}}(\alpha;\fY,\fZ, \fA) 
= \| \widehat{\text{Cov}}_n(A,Y-\alpha^\t Z) \|^2_{(n^{-1}\fA^\t \fA)^{-1}}$ when $A$ is mean zero.\footnote{$\|\cdot\|_{(n^{-1}\fA^\t \fA)^{-1}}$ is the norm induced by the inner product 
	$\la x,y\ra = x^\t (n^{-1}\fA^\t \fA)^{-1} y$.}
In the just-identified setup 
the TSLS estimator yields a sample covariance that is exactly zero and is known to be 
unstable, in that it has no moments of any order. 
Intuitively, the constraint of vanishing 
sample covariance 
may be too strong. 

 Let $T(\alpha;\fY,\fZ,\fA)$ be a finite sample test statistic for testing  the hypothesis $\cH_0(\alpha)$ 
and let $\text{p-value}(T(\alpha;\fY,\fZ,\fA))$ denote the p-value associated with the test of $\cH_0(\alpha)$.
 We then define the p-uncorrelated least square estimator (PULSE) as
\begin{align} \label{eq:PULSEfirstEQ}
	\hat{\alpha}^n_{\mathrm{PULSE}}(p_{\min}) = \begin{array}{ll}
		\text{argmin}_\alpha & l_{\mathrm{OLS}}^n(\alpha;\fY,\fZ)  \\
		\text{subject to} & \text{p-value}(T(\alpha;\fY,\fZ,\fA)) \geq p_{\min},
	\end{array}
\end{align}
where  $p_{\min}$ is a pre-specified level of the hypothesis test.  In words, we aim to minimize the mean squared prediction error
among all coefficients which yield a
p-value for testing $\cH_0(\alpha)$ 
that does not fall below some pre-specified level-threshold $p_{\min} \in (0,1)$, such as $p_{\min}= 0.05$. 
That is, the minimization is constrained to the acceptance 
region of the test, i.e., a confidence region for the causal coefficient in the identified setup. 
Among these coefficient, 
we choose the solution that is `closest' to the OLS
solution.\footnote{  
	Here, closeness is measured in the
	OLS distance:
	We define the OLS norm via
	$\|\alpha\|_{\text{OLS}}^2 := 
	l_{\mathrm{OLS}}^n(\alpha + \hat{\alpha}^n_{\mathrm{OLS}})
	- 
	l_{\mathrm{OLS}}^n(\hat{\alpha}^n_{\mathrm{OLS}})
	= \alpha^\top \mathbf{Z}^T \mathbf{Z} \alpha$, where $\hat{\alpha}^n_{\mathrm{OLS}}$ is the OLS estimator.
	This defines a norm (rather than a semi-norm) 
	if $\mathbf{Z}^T \mathbf{Z}$ is
	non-degenerate. Minimizing 
	$l_{\mathrm{OLS}}^n(\alpha)=\|\fY-\fZ\alpha\|_2^2 = (\alpha-\hat{\alpha}^n_{\mathrm{OLS}})^\t \fZ^\t \fZ (\alpha- \hat{\alpha}^n_{\mathrm{OLS}})+\|\fY-\fZ \hat{\alpha}^n_{\mathrm{OLS}}\|_2^2$ is equivalent 
	to 
	minimizing
	$\|\alpha - \hat{\alpha}^n_{\mathrm{OLS}}\|_{\text{OLS}}^2$.
}

Thus, PULSE allows for an intuitive interpretation. We will see in the experimental section that it has good finite sample performance, in particular for weak instruments.
Unlike other estimators, such as LIML, the above estimator is well-defined in the under-identified setup, too.\footnote{
	The PULSE estimator is defined for finite samples, but the 
	following deliberation may help to build intuition:
	In an under-identified IV setting, minimizing $l_{\mathrm{OLS}}(\gamma)$ under the constraint that $l_{\mathrm{IV}}(\gamma) = 0$, 
	can be seen as choosing,
	under all causal models compatible with the distribution,
	the model with the least amount confounding -- when using 
	$E(Y-\gamma^\top X)^2 - E(Y-\gamma_{\mathrm{OLS}}^\top X)^2$
	as a measure for confounding.
} 
In such cases, PULSE extends on existing literature that aims to trade-off predictability and invariance but that so far has been restricted to search over subsets of variables 
(see \Cref{sec:summarypuls} and \Cref{app:underidentifiedexperiment}).
To maintain consistency of the estimator the chosen test must have asymptotic power of one.

In this paper,
we propose a class of significance tests, that contains, e.g., the Anderson-Rubin test \citep[][]{anderson1949estimation}.
While the objective function in \Cref{eq:PULSEfirstEQ}
is quadratic in $\alpha$, the resulting constraint is, 
in general, non-convex. 
In Section~\ref{sec:DualPULSE}, we develop a computationally efficient procedure that 
provably solves the optimization problem at low computational cost. 
Other choices of tests are possible,  
too, but may result in even harder optimization problems.

In \Cref{sec:SetupAndAssumptionsPULSE}, we briefly introduce the setup and assumptions.
In \Cref{sec:VanishCorr}, we specify a class of asymptotically consistent tests for $\cH_0(\alpha)$. 
In \Cref{sec:PULSEdefi} we formally define the PULSE estimator.
In \Cref{sec:PrimalPULSE}, we show that the PULSE estimator is well-defined by proving that it is equivalent to a solvable convex quadratically constrained quadratic program which we denote by the primal PULSE. %
In \Cref{sec:DualPULSE}, we utilize duality theory and derive an alternative representation which we denote by the dual PULSE. 
This representation 
yields a computationally feasible algorithm 
and shows that the PULSE estimator is a K-class estimator
with a 
data-driven 
$\kappa$. 
Proofs of results in this section can be found in \Cref{sec:RemainingProofsOfSecPULSE} unless stated otherwise. 

\subsection{Setup and Assumptions}\label{sec:SetupAndAssumptionsPULSE}
	In the following sections we again let $(\fY,\fX,\fH,\fA)$ consist of $n\geq \min\{d,q\}$ row-wise independent and identically distributed copies of $(Y,X,H,A)$ generated in accordance with the SEM in \Cref{ARModel}.	The structural equation of interest is  $Y = \gamma^\t_0 X + \eta^\t_0  H + \beta^\t_0 A +  \ep_{Y}$.  
Assume that we have some non-sample information about which $d_2=d-d_1$ and $q_2=q-q_1$ coefficients of $\gamma_0$ and $\beta_0$, respectively, are zero. As in \Cref{SEC:ROBUSTNESS}, 
we let the subscript $*$ denote the variables and coefficients that are non-zero according to the non-sample information but to simplify notation, we 
drop the $*$ subscript from $Z$, $\fZ$ and $\alpha_0$; that is, we write
$Z =[X_*^\t \; A_*^\t]^\t \in \R^{d_1+q_1}$, 
$\fZ=[
\fX_* \, \, \fA_*
]\in \R^{n\times(d_1+q_1)}$ and $\alpha_0 :=(\gamma_{0,*}^\t,\beta_{0,*}^\t)^\t:\in \R^{d_1+q_1}$.  That is, $Y = \alpha^\t_{0} Z + U_Y$,
where $U_Y = \alpha_{0,-*}^\t Z_{-*}+\eta^\t_0  H + \ep_Y$. If the non-sample information is true, then $U_Y = \eta^\t_0  H + \ep_Y$.

We define a setup as being under- just- and over-identified by the degree of over-identification $q_2-d_1$ being negative, equal to zero and positive, respectively. That is, the number of excluded exogenous variables $A_{-*}$ being less, equal or larger than the number of included endogenous variables $X_{*}$ in the target equation.

We assume that the global assumptions of \Cref{ass:global} 
from \Cref{sec:SetupAndAssumptionsRobustness}  still hold. 
Furthermore, we will 
make use of the following situational assumptions
\begin{assumption} \label{ass:AIndepUYandMeanZeroA}
	\begin{enumerate*}[label=(\alph*),ref=\ref{ass:AIndepUYandMeanZeroA}.(\alph*)]
		\item $A \independent U_Y$; \label{ass:AIndepUy}
		\item 	$E[A]=0$.  \label{ass:MeanZeroA}
	\end{enumerate*}
\end{assumption}
\begin{assumption}
	$\ep$ has non-degenerate marginals.\label{ass:NonDegenYNoise}
\end{assumption}
\begin{assumption} \label{ass:ZtZfullrankandAtZfullrank}
	\begin{enumerate*}[label=(\alph*),ref=\ref{ass:ZtZfullrankandAtZfullrank}.(\alph*)]
		\item 	$\fZ^\t \fZ$ is of full rank; \label{ass:ZtZfullrank}
		\item 	$\fA^\t \fZ$ is of full rank. \label{ass:AtZfullrank}
	\end{enumerate*}
\end{assumption}
\begin{assumption}
	$[\fZ \, \, \fY]$ is of full column rank. \label{ass:ZYfullcolrank}
\end{assumption}
\begin{assumption}
	$E[AZ^\t]$ is of full rank. \label{ass:EAZtfullrank}
\end{assumption}

	Assumption\Cref{ass:AIndepUy} holds if our non-sample information is true, and the instrument set $A$ is independent of all unobserved endogenous variables $H_i$ which directly affect the target $Y$. This holds, for example, if the latent variables are source nodes, that is, they have no parents in the causal graph of the corresponding SEM.
Assumption\Cref{ass:MeanZeroA} can be achieved by centering the data. Strictly speaking, this introduces a weak dependence structure in the
observations, which is commonly ignored. Alternatively, one can perform sample splitting. For more details on this assumption and the possibility of relaxing it, see \Cref{rm:AssumptionMeanZero}. 
Assumption\Cref{ass:ZtZfullrank} ensures that K-class estimators for $\kappa < 1$ are
well-defined, regardless of the over-identification degree. 
In the under-identified setup, Assumption\Cref{ass:AtZfullrank} yields that there exists a subspace of solutions minimizing $l_{\text{IV}}^n(\alpha)$. In the just- and over-identified setup this assumption ensures
that $l_{\text{IV}}^n(\alpha)$ has a unique minimizer 
given by the two-stage least squares estimator $\hat{\alpha}_{\text{TSLS}}^n := (\fZ^\t P_\fA \fZ)^{-1} \fZ^\t P_\fA \fY$. 
\Cref{ass:ZYfullcolrank} is used to ensure that the ordinary least square objective function $l_{\text{OLS}}^n(\alpha;\fY,\fZ)$ is strictly positive, such that division by this function is always well-defined.  \Cref{ass:NonDegenYNoise,ass:EAZtfullrank} ensure that various limiting arguments are valid. In the just- and over-identified setup \Cref{ass:EAZtfullrank} is known as the rank condition for identification.

\subsection{Testing for Vanishing Correlation} \label{sec:VanishCorr}

We now introduce a class of tests for the null hypothesis 
$\cH_0(\alpha) : \text{Corr}(A,Y-Z\alpha) =0$
that have point-wise asymptotic level and 
pointwise asymptotic power. 
These tests
will allow us to define the corresponding PULSE estimator. 
When \Cref{ass:ZYfullcolrank} holds we can define  $T_n^c:\R^{d_1+q_1} \to \R$  by
\begin{align*}
T_n^c(\alpha) := c(n) \frac{l_{\text{IV}}^n(\alpha)}{l_{\text{OLS}}^n(\alpha)} =  
c(n) \frac{\|P_\fA (\fY- \fZ \alpha) \|_2^2}{\|\fY- \fZ \alpha\|_2^2},
\end{align*}
where $c(n)$ is a function that will typically scale linearly in $n$.
Let us denote the $1-p$ quantile of the central Chi-Squared distribution with $q$ degrees of freedom by
$Q_{\chi^2_{q}}(1-p)$.
By 
standard limiting theory
we can test $\cH_0(\alpha)$ in the following manner. 
\begin{restatable}[Level and power of the test]{lemma}{TheoremTestingVanishingCorr}
	\label{prop:TestingVanishingCorr}
Let \Cref{ass:AIndepUYandMeanZeroA,ass:NonDegenYNoise,ass:ZYfullcolrank} hold
and assume that $c(n) \sim n$ as $n\to\i$. For any $p\in (0,1)$ 
and any fixed $\alpha$, the statistical test rejecting the null hypothesis $\cH_0(\alpha)$ if $T_n^c(\alpha) > Q_{\chi^2_{q}}(1-p),$  has point-wise 
asymptotic level $p$ and point-wise asymptotic power of 1 against all alternatives as  $n \rightarrow \infty$.
\end{restatable}

\begin{remark} \label{rm:AssumptionMeanZero}
	\textnormal{Assumption\Cref{ass:MeanZeroA}, $E[A]=0$, is important for the test statistic to be asymptotic pivotal under the null hypothesis, 
that is, 
to ensure that the asymptotic distribution of $T_n^c(\alpha)$ 
does not depend on the 
model parameters except for $q$. 
We can drop this assumption 
if we change the null hypothesis to $\cH_0(\alpha):E[A(Y-Z^\t\alpha)]=0$ and add the assumption that $E[U_Y]=0$. 	Furthermore, if we are in the just- or over-identified setup and \Cref{ass:EAZtfullrank} holds, both of these hypotheses are under their respective assumptions equivalent to $\tilde{\cH}_0(\alpha): \alpha=\alpha_0$. That is, the test in \Cref{prop:TestingVanishingCorr} becomes an asymptotically consistent test for the causal coefficient.}
\end{remark}
Depending on the choice of $c(n)$,
this class contains several tests, some of which are well known. 
With $c(n) = n-q+Q_{\chi^2_q}(1-p_{\min})$,
for example, one
recovers a test that is equivalent to the 
asymptotic version of the 
Anderson-Rubin test  (\citealp{anderson1950asymptotic}).
We make this connection precise 
in \Cref{rm:ConnectionToAndersonRubinCI}
in \Cref{app:AddRemarks}. 
The Anderson-Rubin test is robust to weak instruments in the sense that the limiting distribution of the test-statistic under the null-hypothesis is not affected by weak instrument asymptotics; see, e.g.\ \citet{staiger1997instrumental}  and \citet{stock2002survey}.\footnote{Weak instrument asymptotics is a model scheme where the instrument strength tends to zero at a rate of $n^{-1/2}$, i.e., the reduced form structural equation for the endogenous variables is given by $\fX = \fA n^{-1/2} \Pi_X +\bm{\ep} \Gamma^{-1}_X$.} For weak instruments, the confidence region may be unbounded with large probability; see \cite{dufour1997some}.
\cite{moreira2009tests} show that the test suffers 
from loss of power in the over-identified setting.

To simplify notation, we will from now on work with the choice 
$c(n) = n$
and define the acceptance region with level $p_{\min}\in(0,1)$
as
$\cA_n(1-p_{\min}) := \{\alpha \in \R^{d_1+q_1}: T_n(\alpha) \leq Q_{\chi^2_{q}}(1-p_{\min})\}$,
where $T_n(\alpha)$ corresponds to the choice $c(n) = n$.

\subsection{The PULSE Estimator} \label{sec:PULSEdefi}
For any level $p_{\min}\in(0,1)$, we formally define the PULSE estimator of \Cref{eq:PULSEfirstEQ} by letting the feasible 
set
be given by the acceptance region $\cA_n(1-p_{\min})$ of $\cH_0(\alpha)$ using the test of \Cref{prop:TestingVanishingCorr}. That is, we consider
\begin{align}  \label{eq:PULSE}
\hat{\alpha}^n_{\mathrm{PULSE}}(p_{\min}) := \begin{array}{ll}
\argmin_\alpha & l_{\mathrm{OLS}}^n(\alpha)  \\
\text{subject to} & T_n(\alpha) \leq Q_{\chi^2_{q}}(1-p_{\min}).
\end{array}
\end{align}
In general, 
this 
is a non-convex optimization problem (\citealp{boyd2004convex})
as the constraint function is non-convex,
see the blue contours in 
\Cref{fig:LevelsetsTestAndOLS}(left).
From \Cref{fig:LevelsetsTestAndOLS}(right) we 
see that 
in the given example the problem nevertheless has a unique and well-defined solution: 
the smallest level-set of $l_{\text{OLS}}^n$ with a non-empty intersection of the acceptance region $\{\alpha : T_n(\alpha) \leq Q_{\chi^2_{q}}(1-p_{\min})\}$ intersects with the latter region in a unique point. 
In \Cref{sec:PrimalPULSE}, we prove that this is not a coincidence: 
\Cref{eq:PULSE}
has a unique solution that coincides with the solution of a strictly convex, quadratically constrained quadratic program (QCQP) with a data-dependent constraint bound. 
In \Cref{sec:DualPULSE}, we further derive 
an equivalent Lagrangian dual problem.
This has two important implications.
(1) It allows us to construct a 
computationally efficient procedure to compute a solution of the non-convex problem above, and (2), 
it shows that the PULSE estimator can be written as K-class estimators.

Estimators with similar constraints albeit different optimization objective have been studied by \cite{gautier2011high}. In \Cref{rm:Pretest} in \Cref{app:AddRemarks} we briefly discuss the connection to pre-test estimators. 
Furthermore, any method for inverting the test, see, e.g., \cite{davidson2014confidence}, yields a valid confidence set including the proposed point estimator (given that the method outputs the point estimator when the acceptance region is empty).

\subsection{Primal Representation of PULSE} 
\label{sec:PrimalPULSE}
We now derive a
QCQP representation of the PULSE problem, 
which we call the primal PULSE. 
For all  $t\geq 0$ define the 
empirical primal minimization problem (Primal.$t.n$) by
\begin{align}  \label{PR.t.n}
\begin{array}{ll}
\text{minimize}_\alpha & l_{\mathrm{OLS}}^n(\alpha;\fY,\fZ)  \\
\text{subject to} & l_{\mathrm{IV}}^n(\alpha;\fY,\fZ,\fA)\leq t.
\end{array}
\end{align}
We drop the dependence of $\fY$, $\fZ$ and $\fA$ and refer to the objective and constraint functions as $l_{\text{OLS}}^n(\alpha)$ and $l_{\text{IV}}^n(\alpha)$. The following lemma shows that under suitable assumptions these problems are solvable, strictly convex QCQP problems satisfying Slater's condition.

\begin{restatable}[Unique solvability of the primal]{lemma}{LemmaPrimalUniqueSolution}		\label{lm:PrimalUniqueSolAndSlatersConditions}
	Let \Cref{ass:ZtZfullrankandAtZfullrank} hold. It holds that $\alpha \mapsto l_{\mathrm{OLS}}^n(\alpha)$ and $\alpha \mapsto l_{\mathrm{IV}}^n(\alpha)$ are strictly convex and convex, respectively. Furthermore, for any $t > \inf_{\alpha}l_{\mathrm{IV}}^n(\alpha)$ it holds that the constrained minimization problem (Primal$.t.n$) has a unique solution and satisfies Slater's condition. In the under- and just-identified setup the constraint bound requirement	
	is equivalent to $t>0$ and in the over-identified setup to $t> l_{\mathrm{IV}}^n(\hat{\alpha}^n_{\mathrm{TSLS}})$, where $\hat{\alpha}^n_{\mathrm{TSLS}}= (\fZ^\t P_\fA \fZ)^{-1}\fZ^\t P_\fA \fY $.
\end{restatable}

\begin{figure}[!ht] 
	\centering
	\includegraphics[width=\linewidth-0pt]{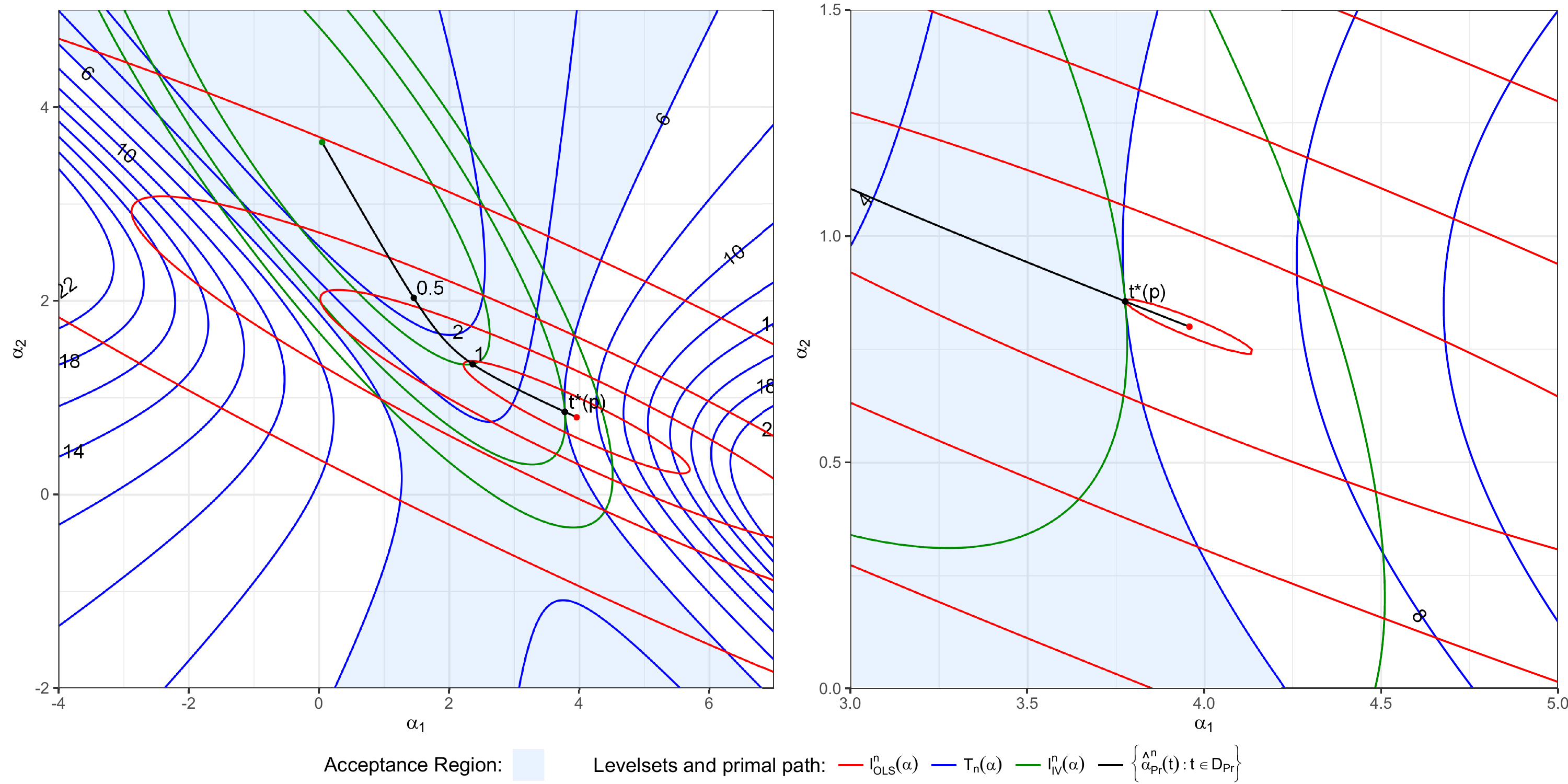}
	\caption{ Illustrations of the level sets of $l_{\mathrm{OLS}}^n$ (red contours), the proposed test-statistic $T_n$ (blue contours) and $l_{\text{IV}}^n$ (green contours) in a just-identified setup. The example is generated with a two dimensional anchor $A = (A_1,A_2)$, one of which is included, and one included endogenous variable $X$, i.e., $Y = \alpha_1 X + \alpha_2 A_1 + H+ \ep_Y$ with $(\alpha_1,\alpha_2)=(1,1)$. Both illustrations 
	show level sets from the same setup, but they use different scales. 
	The black text denotes the level of the test-statistic contours. In this setup, the PULSE constraint bound, the rejection threshold of the test with $p_{\min}=0.05$, is $Q_{\chi^2_{2}}(0.95) \approx 5.99$. The blue level sets of $T_n$ are 
	non-convex. The sublevel set of the test, corresponding to the acceptance region, is illustrated by the blue area. In the right plot, we see that the smallest level set of $l_{\text{OLS}}^n$ that has a non-empty intersection with the $Q_{\chi^2_{q}}(1-p_{\min})$-sublevel set of $T_n$ is a singleton (black dot, $t^*(p)$). This shows that 
	in this example
	the PULSE problem is solvable and has a unique solution. The $l_{\text{IV}}^n$ level set that intersects this singleton   is exactly the $t_n^\star(p_{\min})$-level set of $l_{\text{IV}}^n$, illustrating the statement of \Cref{thm:pPULSESolvesPULSE} in that the primal PULSE 
with that choice of $t$	
	solves the PULSE problem. The black line visualizes the solutions	
	$\{\hat{\alpha}_{\text{Pr}}^n(t)  : t \in D_{\text{Pr} }\}$. The black points and corresponding text labels indicates which constraint bound $t$ yields the specific point. In general, the class of primal solutions 
		does not coincide with the class of convex combinations of the OLS and the TSLS estimators.} %
	 \label{fig:LevelsetsTestAndOLSAndIV} \label{fig:LevelsetsTestAndOLS}
\end{figure}

We restrict the constraint bounds to $D_{\text{Pr}}:=(\inf_\alpha l_{\mathrm{IV}}^n(\alpha), l_{\text{IV}}^n(\hat{\alpha}_{\text{OLS}}^n)]$. Considering $t$ that are larger than 
$\inf_\alpha l_{\mathrm{IV}}^n(\alpha)$
ensures that the problem 
(Primal$.t.n$) is uniquely solvable and
furthermore that Slater's condition is satisfied (see \Cref{lm:PrimalUniqueSolAndSlatersConditions} above). 
Slater's condition 
will play a role 
in \Cref{sec:DualPULSE} when establishing a sufficiently strong connection with its corresponding dual problem for which we can 
derive a (semi-)closed form solution.
Constraint bounds greater than or  equal to $l_{\text{IV}}^n(\hat{\alpha}_{\text{OLS}}^n)$ yield identical solutions. 
Whenever 
well-defined, let $\hat{\alpha}_{\text{Pr}}^n:D_{\text{Pr}}\to \R^{d_1+q_1}$ denote the constrained minimization estimator given by the solution to the (Primal$.t.n$) problem
\begin{align} \label{eq:PrimalProblemSolutionDef}
\hat{\alpha}_{\text{Pr}}^n(t) :=  \begin{array}{ll}
\argmin_{\alpha} & l_{\mathrm{OLS}}^n(\alpha)  \\
\text{subject to} & l_{\mathrm{IV}}^n(\alpha)\leq t.
\end{array}
\end{align}
We now prove that 
for a specific choice of $t$, the
PULSE and the primal PULSE yield the same solutions. 
Define
$t_n^\star(p_{\min})$ as the data-dependent constraint bound given by
\begin{align} \label{eq.Def.t.star.p}
t_n^\star(p_{\min}) := \sup \{ t \in ( \inf_{\alpha}l_{\mathrm{IV}}^n(\alpha),l_{\mathrm{IV}}^n(\hat{\alpha}_{\mathrm{OLS}}^n) ] : T_n(\hat{\alpha}_{\mathrm{Pr}}^n(t))\leq Q_{\chi^2_{q}}(1-p_{\min})\}.
\end{align}
If $t^\star_n(p_{\min})>-\i$ or equivalently $t^\star_n(p_{\min})\in D_{\text{Pr}}$ we define the primal PULSE problem and its solution by (Primal$.t^\star_n(p_{\min}).n$) and $\hat{\alpha}_{\mathrm{Pr}}^n(t_n^\star(p_{\min}))$. 
The following theorem yields conditions for when the solutions to the primal PULSE and PULSE problems coincide.
\begin{restatable}[Primal representation of PULSE]{theorem}{pPULSESolvesPULSE}
	\label{thm:pPULSESolvesPULSE} 
	Let $p_{\min}\in(0,1)$ and \Cref{ass:ZtZfullrankandAtZfullrank,ass:ZYfullcolrank} hold and assume that  $t_n^\star(p_{\min}) >-\i$. If $T_n(\hat{\alpha}_{\mathrm{Pr}}^n(t_n^\star(p_{\min})))\leq Q_{\chi^2_{q}}(1-p_{\min})$, then the PULSE problem has a unique solution given by the primal PULSE solution. That is, 
	$
	\hat{\alpha}^n_{\mathrm{PULSE}}(p_{\min}) = \hat{\alpha}_{\mathrm{Pr}}^n(t_n^\star(p_{\min})).$
\end{restatable}
We show that $t_n^\star(p_{\min}) >-\i$ is a sufficient to guarantee that $T_n(\hat{\alpha}_{\mathrm{Pr}}^n(t_n^\star(p_{\min})))\leq Q_{\chi^2_{q}}(1-p_{\min})$ in the proof of \Cref{thm:PULSEpPULSEdPULSEEequivalent}. The sufficiency of $t_n^\star(p_{\min}) >-\i$ is postponed to the latter proof as it easily follows
from the dual representation.
Hence, 
we have shown that finding the PULSE estimator, i.e., 
finding a solution to the non-convex PULSE problem, is equivalent to solving the convex QCQP primal PULSE for a data dependent choice of $t_n^\star(p_{\min})$.\footnote{Given that value, we can use a numerical QCQP solver to calculate the PULSE estimate.
} However,
$t_n^\star(p_{\min})$ is still unknown. \Cref{fig:LevelsetsTestAndOLSAndIV} 
shows
an example of the equivalence in \Cref{thm:pPULSESolvesPULSE}.  \Cref{fig:LevelsetsTestAndOLSAndIV}(right) shows that the level set of $l_{\text{IV}}(\alpha) = t^{\star}(p_{\min})$ intersects the optimal level curve of $l_{\text{OLS}}^n(\alpha)$ in the same point given by minimizing over the constraint $T_n(\alpha) \leq Q_{\chi^2_q}(1-p_{\min})$.
The set of solutions to the primal problem
$\{\hat{\alpha}_{\text{Pr}}^n(t)  : t \in D_{\text{Pr} }\}$ can in the just- and over-identified setup be visualized as an (in general) non-linear path in $\R^{d_1+q_1}$ between the TSLS estimator $(t= l_{\mathrm{IV}}^n(\hat{\alpha}_{\text{TSLS}}^n))$ and the OLS estimator $(t= l_{\text{IV}}^n(\hat{\alpha}_{\text{OLS}}^n))$ \citep[see also][]{rothenhausler2018anchor}.  \Cref{thm:pPULSESolvesPULSE} yields that the PULSE estimator ($t=t_n^\star(p_{\min})$) then seeks the estimator 'closest' to the OLS estimator along this path that does not yield a rejected test of simultaneous vanishing correlation between the resulting prediction residuals and the exogenous variables $A$, see  \Cref{fig:LevelsetsTestAndOLSAndIV}. 
The path of possible solutions is not necessarily a straight line (see black line); thus, in general, the PULSE estimator is different from the affine combination of OLS and TSLS estimators studied by e.g.\ \cite{judge2012minimum}.

In the under-identified setup, the TSLS end point corresponding to $t= \min_\alpha l_{\mathrm{IV}}^n(\alpha)$ is instead given by the point in the IV solution space $\{\alpha \in \R^{d_1+q_1}: l_{\text{IV}}^n(\alpha)=0\}$ with the smallest mean squared prediction residuals. 

%
%
%
%
%
%

\subsection{Dual Representation of PULSE} \label{sec:DualPULSE}
In this section, we derive a dual representation of the primal PULSE problem which we will denote the dual PULSE problem. 
This specific dual representation allows for the construction of a binary search algorithm for the PULSE estimator and yields that PULSE is a member of the K-class estimators with stochastic $\kappa$-parameter. 

For any penalty parameter $\lambda \geq 0$ 
we define the dual problem 
(Dual$.\lambda.n$)  by
\begin{align} \label{K.lambda.n} 
\begin{array}{ll}
\text{minimize} & l_{\mathrm{OLS}}^n(\alpha) + \lambda l_{\mathrm{IV}}^n(\alpha).
\end{array}
\end{align} 
Whenever Assumption\Cref{ass:ZtZfullrank} holds, i.e., $\fZ^\t \fZ$ is of full rank, then for any $\lambda\geq 0$ the solution to (Dual$.\lambda.n$)  coincides with the 
K-class estimator with $\kappa = \lambda/(1+\lambda)\in[0,1)$, see \Cref{lm:PenalizedKClassSolutionUniqueAndExists}. That is, 
\begin{align*}
\hat{\alpha}_{\text{K}}^n(\kappa)
= (\fZ^\t (\fI+\lambda P_\fA)\fZ)^{-1} \fZ^\t(\fI+\lambda P_\fA)\fY 
\end{align*}
solves (Dual$.\lambda.n$). Henceforth,  let $\hat{\alpha}_{\mathrm{K}}^n (\lambda)$ denote the solution to (Dual$.\lambda.n$), i.e.,
in a slight abuse of notation we will 
denote 
the solution to~(Dual$.\lambda.n$)  by
$\hat{\alpha}_{\text{K}}^n(\lambda)$, such that
$\hat{\alpha}_{\text{K}}^n(\lambda)=
\hat{\alpha}_{\text{K}}^n(\kappa)$
for
$\kappa = \lambda/(1+\lambda)$. We refer to these two representations as the K-class estimator with penalty parameter $\lambda$ and parameter $\kappa$, respectively. The usage of 
$\kappa$ or $\lambda$ as argument should clarify which notation we refer to.

Under Assumption\Cref{ass:AtZfullrank} we have that the minimum of $l_{\text{IV}}^n(\alpha)$ is attainable (see the proof of \Cref{lm:PrimalUniqueSolAndSlatersConditions}). Hence, let the solution space for the minimization problem $\min_\alpha l_{\text{IV}}^n(\alpha)$ be given by
\begin{align}\label{eq:Miv}
\cM_{\mathrm{IV}} := \argmin_\alpha l_{\mathrm{IV}}^n(\alpha) =  \{\alpha \in \R^{d_1+q_1} : l_{\mathrm{IV}}^n(\alpha)= \min_{\alpha'}l_{\mathrm{IV}}^n(\alpha')\}.
\end{align}
In the under-identified setup $(q_2<d_1)$, 
$\cM_{\text{IV}}$ is a  $(d_1-q_2)$-dimensional subspace of $\R^{d_1+q_1}$ and in the just- and over-identified setup it holds that $\cM_{\text{IV}}= \{\hat{\alpha}_{\text{TSLS}}^n\}$.

We now prove that, in the generic case, K-class 
estimators 
for $\lambda \in [0,\infty)$
are different from the TSLS estimator. This result may not come as a surprise, but we include it as we need the result
later and have not found it elsewhere.
\begin{restatable}[K-class estimators and TSLS differ]{lemma}{KclassNotEqualToTwoSLS}
	\label{lm:KclassNotEqualToTwoSLS}
	Assume that we are in the just- or over-identified setup and $n>q$. Furthermore, assume that 	$\ep$ has density with respect to Lebesgue measure and that the coefficient matrix $B$ of the SEM in \Cref{ARModel} is lower triangular. If the rank conditions of \Cref{ass:ZtZfullrankandAtZfullrank} hold almost surely, then it  
	almost surely holds, that all K-class estimators with penalty parameter $\lambda\in[0,\infty)$ differ from the 
	TSLS estimator, i.e., $\hat{\alpha}_{\mathrm{TSLS}}^n  \not\in \{\hat{\alpha}_{\mathrm{K}}^n(\lambda): \lambda \geq 0 \}$. 
\end{restatable}
We conjecture that the corresponding statement 
holds in the under-identified setup and without the lower triangular assumption on B, too. That is,  $\cM_{\text{IV}}\cap \{\hat{\alpha}_{\mathrm{K}}^n(\lambda): \lambda \geq 0 \}=\emptyset$ holds almost surely. We therefore introduce this as an assumption.
\begin{assumption}
	No K-class estimator $\hat{\alpha}_{\mathrm{K}}^n(\kappa)$ with $\kappa\in[0,1)$, is a member of $\cM_{\text{IV}}$. \label{ass:KclassNotInIV}
\end{assumption}

Furthermore, when imposing that \Cref{ass:KclassNotInIV} holds we also have that the K-class estimators differ from each other. 
\begin{restatable}[K-class estimators differ]{corollary}{KclassSolutionsDistinct}
	\label{cor:KclassSolutionsDistinct}
	Let 
	\Cref{ass:ZtZfullrankandAtZfullrank,ass:KclassNotInIV} hold.
	If $\lambda_1,\lambda_2\geq 0$ with $\lambda_1\not = \lambda_2$, then $\hat{\alpha}_{\mathrm{K}}^n(\lambda_1) \not = \hat{\alpha}_{\mathrm{K}}^n(\lambda_2)$.
\end{restatable}

 The above corollary is proven as  \Cref{cor:KclassSolutionsDistinctApp} in \Cref{sec:SomeProofsOfSecPULSE}.
We now show that the class of K-class estimators with penalty parameter $\lambda \geq 0$ , i.e., $\kappa\in[0,1)$, coincides with the class of constrained minimization-estimators that minimize the primal problems with constraint bounds $t> \min_{\alpha}l_{\mathrm{IV}}^n(\alpha)$. 
\begin{restatable}[Connecting the primal and dual]{lemma}{EquivalenceBetweenKlikeAndPrimal}	\label{lm:EquivalenceBetweenKlikeAndPrimal}
	If 
	\Cref{ass:ZtZfullrankandAtZfullrank,ass:ZYfullcolrank,ass:KclassNotInIV} hold, then both of the following statements hold.  \textit{(a)} For any $t \in D_{\mathrm{Pr}}$, there exists a unique $\lambda(t) \geq 0$ such that (Primal$.t.n$) and (Dual$.\lambda(t).n$) have the same unique solution. \textit{(b)} For any $\lambda \geq 0$, there exists a unique $t(\lambda) \in D_{\mathrm{Pr}}$ such that (Primal$.t(\lambda).n$) and (Dual$.\lambda.n$) have the same unique solution.
\end{restatable}
 \Cref{lm:EquivalenceBetweenKlikeAndPrimal} tells us that, under appropriate assumptions, $
\{\hat{\alpha}_{\text{K}}^n(\kappa ): \kappa \in[0,1)\}= 	\{\hat{\alpha}_{\text{K}}^n(\lambda):\lambda \geq 0\}  = \{\hat{\alpha}_{\text{Pr}}^n(t): t\in D_{\text{Pr}} \}.
$
In words, we have recast the K-class estimators with $\kappa \in [0,1)$ as the class of solutions to the primal problems previously introduced. That the minimizers of $l_{\text{IV}}^n(\alpha)$ are different from all the K-class estimators with penalty $\lambda \geq 0$ (or $\kappa\in[0,1)$) guarantees that when representing a K-class  problem in terms of a constrained optimization problem it satisfies Slater's condition.

We are now able to show the main result of this section.
The PULSE estimator $\hat{\alpha}^n_{\text{PULSE}}(p_{\min})$ 
solves a K-class problem
(Dual$.\lambda.n$)
and can therefore be seen as a K-class estimator with 
a data-dependent parameter. To see this, let us define 
the dual PULSE penalty parameter, i.e., the dual analogue of the primal PULSE constraint $t_n^\star(p_{\min})$ as
	\begin{align} \label{eq.Def.lambda.star.p}
	\lambda_n^\star(p_{\min}) := \inf\{\lambda \geq 0 : T_n(\hat{\alpha}_{\mathrm{K}}^n (\lambda))\leq Q_{\chi^2_{q}}(1-p_{\min}) \}.
	\end{align}
If $\lambda^\star_n(p_{\min})<\i$, we define 
the
dual PULSE problem by (Dual.$\lambda^\star_n(p_{\min}).n)$ with solution
$
\hat{\alpha}_{\text{K}}^n (\lambda_n^\star(p)) = \argmin_{\alpha\in \R^{d_1+q_1}} l_{\text{OLS}}^n(\alpha) + \lambda_n^\star(p_{\min}) l_{\text{IV}}^n(\alpha).
$
	\begin{restatable}[Dual representation of PULSE]{theorem}{PrimalDualConnectionPvalConstraint}	\label{thm:PULSEpPULSEdPULSEEequivalent}
		Let $p_{\min}\in(0,1)$ and \Cref{ass:ZtZfullrankandAtZfullrank,ass:ZYfullcolrank,ass:KclassNotInIV} hold. %
		If $\lambda_n^\star(p_{\min}) <\i $, then it holds that $t_n^\star(p_{\min})>-\i$ and $		\hat{\alpha}_{\mathrm{K}}^n(\lambda_n^\star(p_{\min})) = \hat{\alpha}_{\mathrm{Pr}}^n(t_n^\star(p_{\min})) =	\hat{\alpha}_{\mathrm{PULSE}}^n(p_{\min})$. 
	\end{restatable}
Thus, the PULSE estimator seeks to minimize the K-class penalty $\lambda$, i.e., to pull the estimator along the K-class path $\{\hat{\alpha}_{\text{K}}^n(\lambda):\lambda \geq 0\}$ as close to the ordinary least square estimator as possible.
Furthermore, the statement implies that the PULSE estimator is a K-class estimator with data-driven penalty $\lambda_n^\star(p_{\min})$ or, equivalently, parameter $\kappa = \lambda_n^\star(p_{\min})/(1+\lambda_n^\star(p_{\min}))$.
Given a finite dual PULSE penalty parameter $\lambda_n^\star(p_{\min})$ we can, by utilizing the closed form solution of the K-class problem, represent the PULSE estimator in the following form:
\begin{align*}
\hat{\alpha}_{\mathrm{PULSE}}^n(p_{\min}) &= \hat{\alpha}_{\mathrm{K}}^n(\lambda_n^\star(p_{\min})) \\&= (\fZ^\t (\fI+\lambda_n^\star(p_{\min}) P_\fA)\fZ)^{-1} \fZ^\t(\fI+\lambda_n^\star(p_{\min}) P_\fA)\fY.
\end{align*}
However, to the best of our knowledge,
$\lambda_n^\star(p_{\min})$ 
has no known closed form, so the above expression cannot be computed in closed-form either.
In \Cref{sec:BinarySearch}, we prove that the PULSE penalty parameter $\lambda_n^\star(p_{\min})$ can be approximated with arbitrary precision by a 
simple binary search procedure.

The following lemma provides a necessary and sufficient (in practice checkable) condition
 for when the PULSE penalty parameter $\lambda_n^\star(p_{\min})$ 
is finite. 

\begin{restatable}[Infeasibility of the dual representation]{lemma}{LambdaStarFinite}
	\label{lm:LamdaStarFiniteIFF}
	Let $p_{\min}\in(0,1)$ and \Cref{ass:ZtZfullrankandAtZfullrank,ass:ZYfullcolrank,ass:KclassNotInIV} hold. In the under- and just-identified setup we have that $\lambda_n^\star(p_{\min})<\i$. In the over-identified setup it holds that
$
 \lambda^\star_n(p_{\min}) < \i \iff T_n(\hat{\alpha}_{\mathrm{TSLS}}^n)< Q_{\chi^2_q}(1-p_{\min}).
$
This is not guaranteed to hold as the event that $\cA_n(1-p_{\min})= \emptyset$ can have positive probability.
\end{restatable}

Thus, under suitable regularity assumptions \Cref{lm:LamdaStarFiniteIFF} yields that our dual representation of the PULSE estimator always holds in the under- and just-identified setup. It furthermore yields a sufficient and necessary condition for the dual representation to be valid in the over-identified setup, namely that the TSLS is in the interior of the acceptance region. Furthermore, this condition is possibly violated in the over-identified setup with non-negligible probability. 
\subsubsection{Binary Search for the Dual Parameter} \label{sec:BinarySearch}
The key insight allowing for a binary search procedure for $\lambda_n^\star(p_{\min})$ is
that
the mapping
$\lambda \mapsto T_n(\hat{\alpha}_{\text{K}}^n(\lambda))$
is monotonically decreasing.

\begin{restatable}[Monotonicity of the losses and test statistic]{lemma}{MonotonicityOfTestOfLambda}
	\label{lm:OLSandIV_Monotonicity_FnctOfPenaltyParameterLambda}
	When\\ Assumption\Cref{ass:ZtZfullrank} holds the maps $
	[0,\i)\ni \lambda \mapsto  l_{\mathrm{OLS}}^n(\hat{\alpha}_{\mathrm{K}}^n (\lambda) )$ and $ [0,\i)\ni  \lambda \mapsto l_{\mathrm{IV}}^n(\hat{\alpha}_{\mathrm{K}}^n (\lambda) ) $
	are monotonically increasing and monotonically decreasing, respectively. Consequently, if \Cref{ass:ZYfullcolrank} holds, we have that the map 
$
	[0,\i)\ni \lambda \mapsto T_n (\hat{\alpha}_{\mathrm{K}}^n (\lambda) ) 
$
	is monotonically decreasing. Furthermore, if \Cref{ass:KclassNotInIV} also holds, these monotonicity statements can be strengthened to strictly decreasing and strictly increasing.
\end{restatable}

 The above lemma is proven as  \Cref{lm:OLSandIV_Monotonicity_FnctOfPenaltyParameterLambdaApp} in \Cref{sec:SomeProofsOfSecPULSE}. If the OLS solution is not strictly feasible in the PULSE problem, then $\lambda_n^\star(p_{\min})$ indeed is the smallest penalty parameter for which the test-statistic reaches a p-value of exactly $p_{\min}$; see \Cref{lm:TestInAlphaLambdaStarEqualsQuantileApp} in \Cref{sec:SomeProofsOfSecPULSE}. 

We propose the binary search algorithm
presented in Algorithm~\ref{Binary.Search.Lambda.Star} in \Cref{app:algo},
that can approximate a finite $\lambda_n^\star(p_{\min})$ with arbitrary precision.
We terminate the binary search (see line 2) if $\lambda^\star_n(p_{\min})$ is not finite, in which case we have no computable representation of the PULSE estimator. It is possible to improve this algorithm in the under- and just-identified setup, by initializing $\ell_{\max}$ as the quantity given by \Cref{eq:LambdaEquality} in the proof of \Cref{lm:LamdaStarFiniteIFF}. This initialization removes the need for the first while loop in (lines 4--6).
We now prove that Algorithm~\ref{Binary.Search.Lambda.Star} 
achieves the required precision and 
is asymptotically correct.

\begin{restatable}[]{lemma}{BinarySearchLambdaStarConverges}
	\label{lm:BinarySearchLambdaStarConverges} 
	Let $p_{\min}\in(0,1)$ and \Cref{ass:ZtZfullrankandAtZfullrank,ass:ZYfullcolrank} hold. If it holds that $\lambda_n^\star(p_{\min})<\i$, then $\lambda_n^\star(p_{\min})$ can be approximated with arbitrary precision by the binary search \Cref{Binary.Search.Lambda.Star}, that is, $	\mathrm{Binary.Search}(N,p_{\min}) - \lambda_n^\star(p_{\min}) \to 0,$ as $N\to\i$.
\end{restatable}

\subsection{Algorithm and Consistency} \label{sec:AlgoAndConsistency}
The dual representation of the PULSE estimator is not guaranteed to be well-defined in the over-identified setup.
In particular, it is not well-defined if the TSLS is outside the interior of the acceptance region (which corresponds to a
p-value of less than or equal to $p_{\min}$). In this case, we propose to output a warning. 
This can be 
helpful information for the user since it may indicate a model misspecification.
For example, if the true relationship is in fact nonlinear, and one considers
an over-identified 
case (e.g., by 
constructing different 
transformations of the  instrument),
even the TSLS may be rejected when 
erroneously considering a linear model; see \citet{Keane2010} and \citet{Mogstad2010}.
For any $p_{\min}\in(0,1)$ we 
can still define an always well-defined
modified PULSE estimator 
$\hat{\alpha}_{\text{PULSE}+}^n(p_{\min})$ 
as
$\hat{\alpha}_{\mathrm{PULSE}}^n(p_{\min})$ if the dual representation is feasible %
and some other 
consistent estimator $\hat{\alpha}^n_{\text{ALT}}$ (such as  
the TSLS, LIML or Fuller estimator) otherwise. 
That is, we define
\begin{align*}
\hat{\alpha}_{\text{PULSE}+}^n(p_{\min}) := \left\{\begin{array}{ll}
\hat{\alpha}_{\mathrm{PULSE}}^n(p_{\min}), & \text{if } T_n(\hat{\alpha}_{\text{TSLS}}^n)<Q_{\chi^2_{q}}(1-p_{\min}) \\
\hat{\alpha}^n_{\text{ALT}}, & \text{otherwise}.
\end{array} \right. 
\end{align*}

Similarly to the case of an empty rejection region, 
we also output a warning 
for the case when the OLS estimator is accepted. This may, but does not have to, indicate weak instruments. 
Thus, we have the algorithm presented as \Cref{alg:2} in \Cref{algo:pulseplus} for computing the PULSE$+$ estimator.

We now prove that the PULSE$+$ estimator consistently estimates the causal parameter in the just- and over-identified setting. 
Assume that we choose 
a consistent
estimator
$\hat{\alpha}_{\text{ALT}}^n$
(under standard regularity assumptions, this is satisfied for the TSLS).\footnote{This holds as {\scriptsize
$
\hat{\alpha}_{\text{TSLS}}^n 
= \alpha_0 +  (n^{-1}\fZ^\t \fA (n^{-1}\fA^\t \fA)^{-1} n^{-1}\fA^\t \fZ)^{-1} n^{-1}\fZ^\t \fA (n^{-1}\fA^\t \fA)^{-1} n^{-1}\fA^\t \fU_Y 
$.}}
We can then show that, under mild conditions, the PULSE$+$ estimator, too, is a consistent estimator of $\alpha_0$.

\begin{restatable}[Consistency of PULSE$+$]{theorem}{ConsistencyOfPULSE}
	\label{thm:ConsistencyOfPULSE}
	Consider the just- or over-identified setup and let $p_{\min}\in (0,1)$.  If \Cref{ass:AIndepUYandMeanZeroA,ass:ZtZfullrankandAtZfullrank,ass:ZYfullcolrank,ass:EAZtfullrank,ass:KclassNotInIV} hold almost surely for all $n\in \N$ and $\hat{\alpha}_{\mathrm{ALT}}^n$ consistently estimates $\alpha_0$, then $\hat{\alpha}_{\mathrm{PULSE}+}^n(p_{\min})\convp \alpha_0$, when $n\to \i$.
\end{restatable}
We believe that a similar statement also holds in the under-identified setting, see \Cref{app:underidentifiedexperiment}.

\section{Simulation Experiments}
In \Cref{sec:Experiments} we conduct an extensive simulation study investigating the finite sample behaviour of the PULSE estimator. The concept of weak instruments is central to our analysis. An introduction to weak instruments can be found in \Cref{sec:WeakInst}. Here we give a brief overview of the study and the observations. 

\subsection{Distributional Robustness}
The 
theoretical results on distributional robustness 
proved in \Cref{SEC:ROBUSTNESS}
translate to finite data. The experiments of \Cref{app:DistributionalRobustness} shows that 
even for small sample sizes, K-class estimators outperform 
both OLS and TSLS for a certain range of interventions, matching the theoretical predictions with increasing sample size. In \cref{app:underidentifiedexperiment}, we furthermore consider an under-identified setting.

\subsection{Estimating Causal Effects}
When focusing on the estimation of a 
causal effect in an identified setting,
our simulations show 
 that there are several settings where PULSE outperforms the Fuller and TSLS estimators in terms of mean squared error (MSE).
 In univariate simulation experiments, such settings are 
 characterized by
 weakness of instruments and weak confounding (endogeneity). 
The characterization becomes more  involved 
in multivariate settings, 
but is similar in that PULSE outperforms all other methods for small confounding strengths, an effect amplified by the weakness of instruments. 
Below we detail the univariate simulation setup and refer the reader to \Cref{sec:Experiments} for further details and the multivariate simulation experiments mentioned above.

\subsubsection{Univariate Model.} \label{sec:SimUnivariate}
We first
compared performance measures of the estimators in a univariate instrumental variable model. As seen in \citet{hahn2002new} and \citet{hahn2004estimation}, we consider structural equation models of the form
\begin{align*}
	A := A \in \mathbb{R}^q,\quad 
	X :=  A^\t  \bar \xi  +  U_X \in \R, \quad
	Y :=  X \gamma + U_Y \in \R,
\end{align*}
where $A\sim \mathcal{N}(0,I)$ and $A\independent (U_X,U_Y) $ with 
$
\begin{psmallmatrix}
	U_X \\ U_Y
\end{psmallmatrix}\sim \mathcal{N}\left(\begin{psmallmatrix}
	0 \\ 0
\end{psmallmatrix}, \begin{psmallmatrix}
	1 & \rho \\
	\rho & 1
\end{psmallmatrix}\right).$
Furthermore, we let $\gamma =1$ and $\bar{\xi}^\t=(\xi,....,\xi)\in \R^q$, where $\xi>0$ is chosen according to the theoretical $R^2$-coefficient. We consider the following simulation scheme: for each $q\in \{1,2,3,4,5,10,20,30\}$, $\rho\in\{0.1,0.2,...,0.9\}$, $R^2\in\{0.0001,0.001,0.01,0.1,0.3\}$ and $n\in\{50,100,150\}$, we simulate $n$-samples from the above system and calculate the OLS, TSLS, Fuller(1), Fuller(4) and PULSE ($p_{\min}= 0.05$) estimates; see \Cref{sec:ExpResultsEstCausPerformanceMeasures}.

\begin{figure}[h] 
	\centering\includegraphics[width=\linewidth]{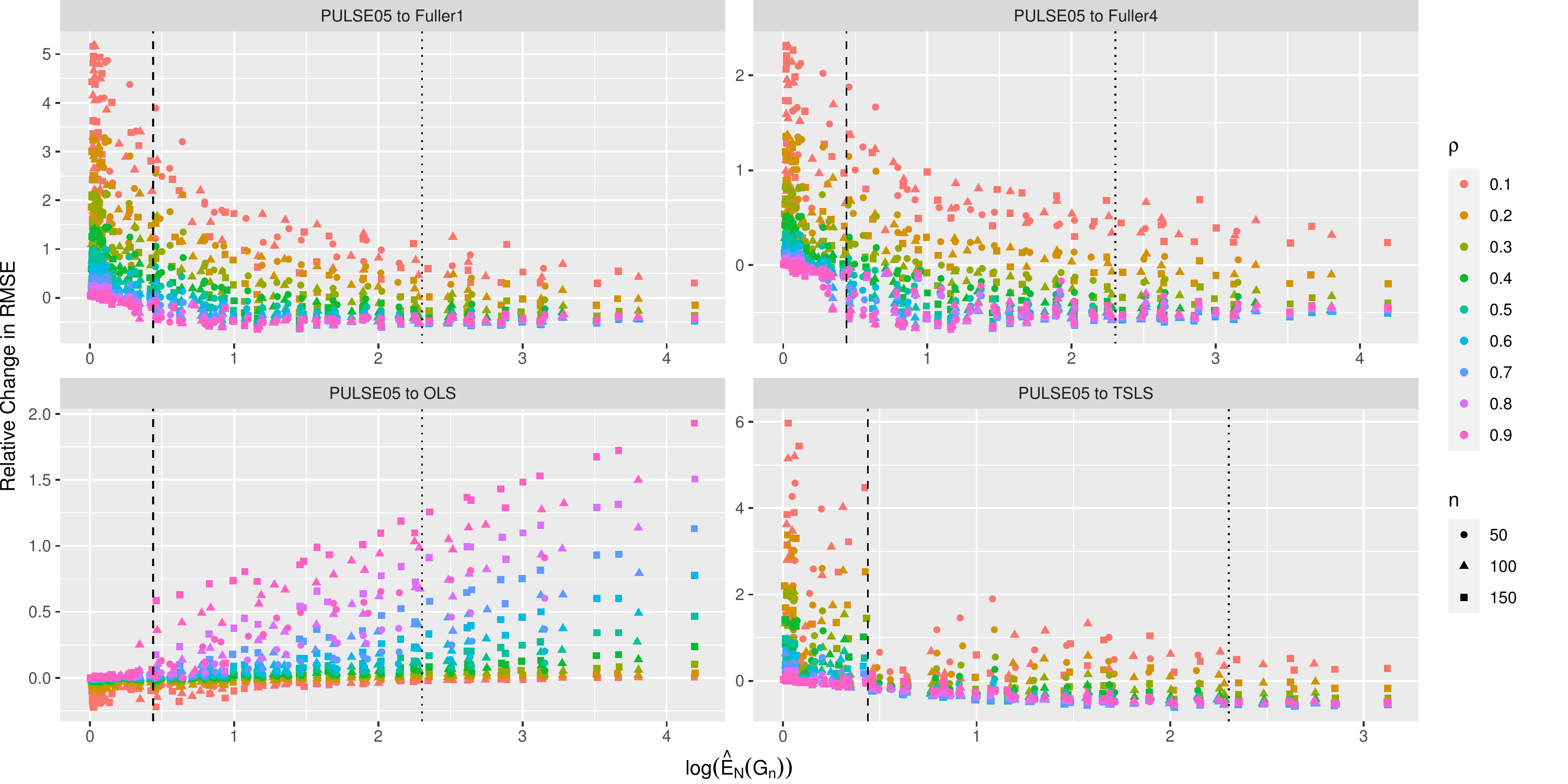}
	\caption{  \normalsize Illustrations of the relative change in  RMSE.}\label{fig:HahnExpRMSE}
\end{figure}

 \Cref{fig:HahnExpRMSE} contains illustrations of the relative change in square-root mean squared error (RMSE) estimated from $15 000$ repetitions. On the horizontal axis we have plotted the average first stage F-test as a measure of weakness of instruments; see \Cref{sec:WeakInst} for further details. A test for $H_0:\bar \xi=0$, i.e.,  for the relevancy of instruments, 
at a significance level of 5\%,	
has different rejection thresholds in the range $[1.55,4.04]$ depending on $n$ and $q$. The vertical dashed line corresponds to the smallest rejection threshold of 1.55 and the dotted line corresponds to the `rule of thump' threshold of 10.  Note that the lowest possible negative relative change is $-1$ and a positive relative change means that PULSE is better.

	In \Cref{sec:AppFigs}, further illustrations of e.g. the relative change in mean bias and variance of the estimators are presented. 
We also conducted the simulations for setups with combinations of $\gamma\in\{-1,0\}$, components of $\bar \xi$ chosen negatively, with random flipped sign in each coordinate and for negative $\rho$ (not shown but available in the folder 'Plots' in the code repository). The results with respect to MSE are similar to those shown in  \Cref{fig:HahnExpRMSE}, while the bias comparison changes depending on the setup.

We observe that there are settings, in which the PULSE is superior to TSLS, Fuller(1) and Fuller(4) in terms of MSE.
This is particularly often the case in weak instrument settings ($\hat E_N(G_n)<10$) for low confounding strength $(\rho\leq 0.2)$. Furthermore, as we tend towards the weakest instrument setting considered, we also see a gradual shift in favour of PULSE for higher confounding strengths. In these settings with weak instruments and low confounding we also see that OLS is superior to the PULSE in terms of MSE. However, for large confounding setups PULSE is superior to OLS in terms of both bias and MSE and this superiority increases as the instrument strength increases. 
The PULSE is generally more biased than the Fuller and TSLS estimators but less biased than OLS. However, in the settings with weak instruments and low confounding the bias of PULSE and OLS is comparable.
In summary, the PULSE is in these settings more biased but its variance is so small that it is MSE superior to the Fuller and TSLS estimators. 

\section{Empirical Applications}

We now consider three classical 
instrumental variable applications (see \citet{albouy2012colonial} and \citet{buckles2013season} for discussions on the underlying assumptions). 
\begin{itemize}
	\item[\textit{(i)}] ``Does compulsory school attendance affect schooling and earnings?'' by \cite{angrist1991does}. This paper investigates the effects of education on wages. The endogenous effect of education on wages are remedied by instrumenting education on quarter of birth indicators.
	\item[\textit{(ii)}] ``Using geographic variation in college proximity to estimate the return to schooling'' by \cite{card1993using}. This paper also investigates the effects of education on wages. In this paper education is instrumented by proximity to college indicator.
	\item[\textit{(iii)}] ``The colonial origins of comparative development: An empirical investigation'' by \cite{acemoglu2001colonial}. This paper investigates the effects of extractive institutions (proxied by protection against expropriation) on the gross domestic product (GDP) per capita. The endogeneity of the explanatory variables are remedied by instrumenting protection against expropriation on early European settler mortality rates.
\end{itemize}

We have
applied the different estimators OLS, TSLS, PULSE, and Fuller to the classical data sets
\cite{acemoglu2001colonial}, \cite{angrist1991does} and  \cite{card1993using}.
All models considered in \cite{angrist1991does} and  \cite{card1993using}, where we estimate the effect on years of education on wages, using quarter of birth and proximity to colleges as instruments, respectively, 
the OLS estimates are not rejected by our test statistic and 
PULSE outputs the OLS estimates; see \Cref{app:EmpricalApp} for futher details.
This may be either due to weak endogeneity (weak confounding), or that the test has insufficient power to reject the OLS estimates due to either weak instruments or severe over-identification.

\subsection{\cite{acemoglu2001colonial}} \label{sec:mainColonialApplication}
The dataset of \cite{acemoglu2001colonial} consists of 64 
observations,
each corresponding to a different country for which mortality rate estimates encountered by the first European settlers are available. The endogenous target of interest is log GDP per capita (in 1995). The main endogenous regressor in the dataset is an  index of expropriation protection (averaged over 1985--1995), i.e., protection against expropriation of private investment by the respective governments. The average expropriation protection is instrumented by the settler mortality rates. We consider eight models M1--M8 which correspond to the models presented in column (1)--(8) in Table 4 of \cite{acemoglu2001colonial}. Model M1 is given by the reduced form structural equations
\begin{align*}
	\log \mathrm{GDP} = \text{avexpr}\cdot \gamma + \mu_1 + U_1, \quad \text{avexpr} = \log \mathrm{em4}\cdot \delta + \mu_2 + U_2,
\end{align*}
where avexpr is the average expropriation protection, em4 is the settler mortality rates, $\mu_1$ and $\mu_2$ are intercepts and $U_1$ and $U_2$ are possibly correlated, unobserved noise variables. In model M2 we additionally introduce an included exogenous regressor describing the country latitude. In model M3 and M4 we fit model M1 and M2, respectively, on a dataset where we have removed Neo-European countries, Australia, Canada, New Zealand and the United States. In model M5 and M6 we fit model M1 and M2, respectively, on a dataset where we have removed observations from the continent of Africa. In model M7 and M8 we again fit model M1 and M2, respectively, but now also include three exogenous indicators for the continents Africa, Asia and other.

\Cref{tbl:SettlerMortalityCoefficients} shows the OLS and TSLS estimates 
(which replicate the values from the study), 
as well as the Fuller(4) and PULSE estimates for the linear effect of the average expropriation protection on log GDP. In model M1, for example, we see that the PULSE estimate suggests that the average expropriation risk linear effect on log GDP is 0.6583 which is 26\% larger than the OLS estimate but 34\% smaller than TSLS estimate. In models M5--M8, the  OLS estimates are not rejected by the Anderson-Rubin test, so the PULSE estimates coincide with the OLS estimates.

We can also use this example to
illustrate the robustness property of K-class estimators; 
see \Cref{sthm:TheoremIntRobustKclas}. 
Even though interventional data are not available, 
we can consider the mean squared prediction error when 
holding out the observations with the most extreme values of the instrument.
Depending on the degree of generalization, we indeed see that 
the PULSE and Fuller tend to outperform OLS or TSLS in terms of mean squared prediction error on the held out data; see \Cref{sec:Colonial} for further details.

\begin{table}[H]
	\caption{\label{tbl:SettlerMortalityCoefficients}The estimated return of expropriation protection on log GDP per capita. }
	\begin{center}{\small
		\begin{tabu}to \linewidth {r c c c c c c c}
			\toprule \toprule
			Model & OLS & TSLS & FUL & PULSE & Message & Test & Threshold \\
			\midrule
			M1 &0.5221 & 0.9443 & 0.8584 & 0.6583 & --  & 5.991 & 5.991\\
			M2 &0.4679 & 0.9957 & 0.8457 & 0.5834 & -- & 7.815 & 7.815\\
			M3 &0.4868 & 1.2812 & 0.9925 & 0.7429 & -- & 5.991 & 5.991\\
			M4 &0.4709 & 1.2118 & 0.9268 & 0.6292 & -- & 7.815 & 7.815\\
			M5 &0.4824 & 0.5780 & 0.5573 & 0.4824 & OLS Accepted & 1.180 & 5.991\\
			M6 &0.4658 & 0.5757 & 0.5476 & 0.4658 & OLS Accepted & 1.155 & 7.815\\
			M7 &0.4238 & 0.9822 & 0.7409 & 0.4238 & OLS Accepted & 10.772 & 11.071\\
			M8 &0.4013 & 1.1071 & 0.7059 & 0.4013 & OLS Accepted & 9.755 & 12.592\\
			\bottomrule\bottomrule
		\end{tabu}
	}
	\end{center}
	\footnotesize
	\renewcommand{\baselineskip}{11pt}
	\textbf{Note:} 
	Point estimates for the return of expropriation protection on log GDP per capita. 
	The OLS and TSLS values coincide with the ones shown in
	\cite{acemoglu2001colonial}.
	The right columns show the values of the test statistic (evaluated in the PULSE estimates) and the test rejection thresholds.
	The `--' indicates that OLS is not accepted and TSLS is not rejected.
\end{table}

%
%

\section{Summary and Future Work}
We have proved that a distributional robustness property similar to the one shown for anchor regression \citep[][]{rothenhausler2018anchor} 
fully extends to general K-class estimators of possibly non-identifiable structural parameters in a general linear structural equation model that allows for latent endogenous variables.
We have further proposed a novel estimator for structural parameters in linear structural equation models. This estimator, called PULSE, is derived as the solution to a minimization problem, where we seek to minimize mean squared prediction error constrained to a 
confidence region 
for the causal parameter. Even though this region is non-convex, we have shown 
that the corresponding optimization problem allows for a computationally efficient algorithm 
that approximates the above parameter with arbitrary precision using a simple binary search procedure.
In the under-identified setting, this estimator extends existing work in the machine learning literature that considers invariant subsets or the best predictive sets among them: PULSE is applicable even in situations when no invariant subsets exist.
We have proved that this estimator can also be written as a K-class estimator with data-driven 
$\kappa$-parameter, which lies between zero and one. 
Simulation experiments show that 
in various settings 
with weak instruments and weak confounding,
PULSE  
outperforms other estimators such as the Fuller(4) estimator. 
We thus regard PULSE as an 
interesting alternative for estimating causal effects
in instrumental variable settings.
It is easy to interpret and automatically provides the user feedback in case that the OLS is accepted (which may be an indication that the instruments are too weak)
or that the TSLS is outside the acceptance region (which may indicate a model misspecification).
We have applied the different estimators
to classical data sets and have seen that, indeed, K-class estimators tend to be more distributionally robust than OLS or TSLS.

There are several further directions that we consider worthwhile investigating. 
This 
includes better understanding of finite sample properties and for the identified setups, the study of loss functions other than MSE. It would be helpful, in particular with respect to real world applications, to understand to which extent similar principles can be applied to models allowing for a time structure of the error terms.
We believe that the simple primal form of PULSE could make it applicable for model classes that are more complex than linear models \citep[see also][]{TPAMI}.
Our procedure can be combined with other tests and it could furthermore be interesting to find efficient optimization procedures for tests that are robust with respect to weak instruments, such as  
Kleibergen's K-statistic \citep[][]{Kleibergen2002},
for example. 
In an under-identified setting, the causal parameters are not identified but the solutions obtained by optimizing predictability under invariance might be promising candidates for models that generalize well to distributional shifts.

\section*{Acknowledgements}
We are grateful to 
Trine 
Boomsma,
Peter 
B\"uhlmann, 
Rune Christiansen, 
Steffen 
Lauritzen, 
Nicolai 
Meinshausen, 
Whitney 
Newey,  
Cosma 
Shalizi, and
Nikolaj 
Thams for helpful 
discussions.
We thank the editor and two anonymous referees for 
helpful and constructive comments.  MEJ and JP were supported by the Carlsberg Foundation; JP was, in addition, supported by a research grant
(18968) from VILLUM FONDEN.

\chapter[A Causal Framework for Distribution Generalization]{A Causal Framework for Distribution Generalization} \label{ch:Generalization}

{\small \textsc{Joint work with
		\begin{quote}
			Rune Christiansen, Niklas Pfister, Nicola Gnecco and Jonas Peters
\end{quote}}}
\vspace{0.75cm}

\begin{quoting}[leftmargin=0.5cm]
	\begin{center}
		\textbf{Abstract}
	\end{center}
	
	{\small  We consider the problem of predicting a 
		response $Y$ from a set of covariates $X$
		when test and training distributions differ. 
		Since such differences may 
		have causal explanations,
		we
		consider
		test distributions that 
		emerge from interventions in a structural causal model,
		and 
		focus 
		on 
		minimizing
		the worst-case risk.
		Causal 
		regression
		models, which regress the
		response on its direct causes, 
		remain 
		unchanged
		under
		arbitrary interventions on the covariates, but they are not always optimal in the above sense.
		For example, for linear models
		and bounded
		interventions,  
		alternative solutions
		have been shown to be minimax prediction optimal.
		We introduce the formal framework 
		of distribution generalization 
		that allows us to analyze the above problem
		in partially observed nonlinear models
		for both direct interventions on $X$ and 
		interventions that occur indirectly via exogenous variables $A$.
		It takes into account that, in practice, 
		minimax solutions need to be identified from data.  
		Our framework allows us to characterize under which 
		class of interventions
		the causal function is minimax optimal.  
		We prove sufficient conditions for distribution 
		generalization and present corresponding impossibility results.
		We propose a
		practical method, 
		NILE, that 
		achieves distribution   
		generalization
		in a nonlinear IV
		setting
		with linear extrapolation. 
		We prove consistency and present empirical results.}
\end{quoting}
\textbf{Keywords:} Distribution generalization, causality, worst-case risk, distributional robustness, invariance, domain adaptation

\section{Introduction} \label{sec:intro}
Large-scale learning systems, particularly those focusing on
prediction tasks, have been successfully applied in various domains of
application. 
Since inference is usually done during training time,
any difference between training and test distribution poses a
challenge for prediction methods \citep[][]{Candela2009, Pan2010,
	Csurka2017, Arjovsky2019}.  Dealing with these differences
is of great importance in 
several
fields such as 
environmental sciences,
where methods need to extrapolate
both in space and time.
Tackling this problem requires restrictions on how the distributions may
differ, since, clearly, generalization becomes impossible if the test
distribution may be arbitrary.  Given a response $Y$ and some
covariates~$X$, several existing procedures aim to find a 
minimax
function
$f$ which minimizes the worst-case risk
$\sup_{P \in \mathcal{N}} \mathbb{E}_P [(Y - f(X))^2]$ across
distributions contained in a small neighborhood $\mathcal{N}$ of the
training distribution. The neighborhood~$\mathcal{N}$ should be
representative of the difference between the training and test
distributions, and often mathematical tractability is taken into
account, too \citep{abadeh2015distributionally, sinha2017certifying}.
A typical approach is to define a $\rho$-ball of distributions
$\cN_\rho(P_0) := \{P: D(P, P_0) \leq \rho\}$ around the (empirical)
training distribution $P_0$, with respect to some divergence measure
$D$,
such
as the Kullback-Leibler divergence
\citep{bagnell2005robust, hu2013kullback}.  
While 
some
divergence
functions only consider distributions with the same support as $P_0$,
the Wasserstein distance allows %
for a neighborhood of distributions around $P_0$ with possibly
different supports \citep{abadeh2015distributionally,
	sinha2017certifying,esfahani2018data,blanchet2019data}.

In our
analysis, we do not start from a divergence measure, but instead
model the difference between training and test distribution using the 
concept of 
interventions \citep{Pearl2009, Peters2017}.
We believe that for many problems this provides a useful description of distributional changes.
We will see that, depending on the considered setup, this approach allows to find models that perform well even on test distributions which would be considered far away from the training distribution in any commonly used metric.
For this class of distributions, causal regression models 
appear naturally because of the following
well-known observation.
A prediction
model, which uses only the direct causes of the response $Y$ as
covariates, is 
invariant under interventions on variables
other than $Y$:
the conditional distribution of $Y$ given its causes does not change
(this principle is known, e.g., as invariance, autonomy or modularity)
\citep{Aldrich1989, Haavelmo1944, Pearl2009}.  Such a causal regression %
model yields the minimal worst-case risk when considering
all interventions on variables other than $Y$
\citep[e.g.,][Theorem~1,
Appendix]{Rojas2016}. It has therefore been suggested to use causal
models in problems of distributional shifts
\citep{Scholkopf2012, Rojas2016, HeinzeDeml17, Magliacane2018,
	Meinshausen2018, Arjovsky2019, pfister2019stabilizing}.
In practice, however, not all relevant causal variables might be observed.
One may further argue that
causal methods are too
conservative in that the interventions 
which induce the test distributions may not
be arbitrarily strong.
Instead,
methods which 
focus on a
trade-off between predictability and
causality
have been proposed for linear models
\citep{rothenhausler2018anchor, Pfister2019pnas}, see also Section~\ref{sec:existingmethods}.
Anchor regression \citep{rothenhausler2018anchor} is shown to be predictive optimal under a set of bounded interventions.

In this work, we introduce the general framework of distribution
generalization, which permits a unifying perspective on the potentials
and limitations of applying causal concepts to the problem of generalizing 
regression models from training to test distribution.
In particular, we use it to characterize the
relationship between a minimax optimal solution and the causal
function, and to classify settings under which the minimax solution is
identifiable from the 
training
distribution.

\subsection{Further Related Work}

The field of
distributional robustness or out-of-distribution generalization
aims to develop procedures that are robust to 
changes between training and test distribution. 
This problem has been actively studied from an empirical perspective in
machine learning research, for example, in image
classification by using
adversarial attacks,
where small 
digital 
\citep{goodfellow2014explaining}
or
physical
\citep{evtimov2017robust}
perturbations of pictures 
can deteriorate 
the performance of a model. Arguably, 
these procedures are not yet fully understood theoretically. A more
theoretical perspective 
is given by the previously mentioned
minimization of a worst-case risk across distributions contained in a
neighborhood of the training distribution,
in our case, distributions generated by interventions.

Our framework includes the problems of multi-task
learning, domain generalization and transfer learning
\citep{Baxter2000, Candela2009, Caruana1997, Mansour2009} (see
Section~\ref{sec:intro_focus} for more details), with a focus on
minimizing the worst-case risk.
In settings of covariate shift
\citep[e.g.,][]{Shimodaira2000, Sugiyama2005, Sugiyama2008}, one
usually assumes that the training and test distribution of the
covariates are different, while the conditional distribution of the
response given the covariates remains invariant
\citep{daume2006domain, bickel2009discriminative, Ben-David2010,
	muandet2013domain}.  Sometimes, it is additionally assumed that the
support of the training distribution covers that of the test
distribution \citep{Shimodaira2000}.  In this work, the
conditional distribution of the response given the covariates is
allowed to change between interventions, due to the existence of 
hidden confounders,
and we consider settings where the test
observations lie outside the training support.

Data augmentation methods have become successful techniques,
e.g.\ in image classification, to adapt prediction procedures
to such types of distribution shifts. These methods %
increase the diversity of the training data
by changing the geometry and the color of the images (e.g., by
rotation, cropping or changing saturation) \citep{zhang2017mixup,
	shorten2019survey}. This allows the user to create models that
generalize better to unseen
environments %
\citep[e.g.,][]{volpi2018adversarial}.  We view these approaches as 
ways to enlarge the support of the covariates, which, as our results
show, comes with theoretical advantages, see
Section~\ref{sec:generalizability}.  

Minimizing the worst-case risk is considered in robust methods \citep{Ghaoui03,
	Kim2005}, too.  It can also be formulated in terms of minimizing the
regret in a multi-armed bandit problem \citep{lai1985asymptotically,
	auer2002finite, bartlett2008high}.  In that setting, the agent can
choose
the distribution which generates the data.
In our setting, though, we do not assume to have 
control over the interventions, 
and, hence, neither over the distribution of the sampled data.

\subsection{Contribution and Structure}
This work contains four main contributions:
(1) 
A novel framework for analyzing the problem of generalization from 
training to test distribution, 
using the notion of distribution generalization (Section~\ref{sec:framework}).
(2) Results elucidating the relationship between a causal function and a minimax solution (Section~\ref{sec:robustness}).
(3) Sufficient conditions which ensure distribution
generalization, along with corresponding impossibility results (Section~\ref{sec:generalizability}).
(4) A practical method, called NILE (`Non-linear Intervention-robust Linear Extrapolator'), which 
learns a minimax solution from i.i.d.\ observational data (Section~\ref{sec:learning}). 

Our framework describes how structural causal models can be used as 
technical devices for modeling
plausible test distributions. It further allows us to formally 
define distribution generalization, which describes the ability
to identify
generalizing
regression models (i.e., minimax solutions) from the observational distribution.
While it is well known that
the causal function is minimax optimal under the set of all
interventions on the covariates \citep[e.g.,][]{Rojas2016}, we extend
this result in several ways, for example, by allowing for hidden variables and
by characterizing more general sets of interventions under which the causal
function is minimax optimal.
We further derive conditions on the
model class, the observational distribution and the family of
interventions under which distribution generalization is
possible, and present impossibility results proving the 
necessity of some of these conditions. 
For example, we show that 
strong assumptions on the 
functional relationship between $X$ and $Y$ 
are 
needed whenever the interventions extend the
training support of $X$. %
An example of such an assumption is to consider the class of 
differentiable functions that
linearly extrapolate outside
the support of $X$. For that model class, we propose the explicit method NILE, 
which obtains distribution generalization by exploiting a nonlinear instrumental variables setup.
We show that our method learns a
minimax solution which corresponds to the causal
function.  
We prove consistency and compare our algorithm to
state-of-the art approaches empirically.

We believe that our results 
shed
some light on the potential merits of 
using causal concepts in 
the context 
of generalization. 
The framework allows us to make first steps 
towards answering
when it can be beneficial to use non-causal functions for prediction under interventions,
and 
what might happen under misspecification of the intervention class.
Our results also formalize in which sense
methods that generalize
in the linear case -- such as IV and anchor regression \citep{rothenhausler2018anchor} --
can be extended to nonlinear settings. %
Further, 
our framework 
implies impossibility statements for multi-task learning that relate to existing 
results
\citep{Ben-David2010}.

Our code
is available as an \texttt{R}-package at
\url{https://runesen.github.io/NILE}; scripts generating all our
figures and results can be found at the same url. Additional
supporting material is given in the online appendix.
Appendix~\ref{sec:causal_relations_X} shows how to represent several
causal models
in our framework. Appendix~\ref{sec:IVconditions} summarizes %
existing results on identifiability in IV models.
Appendix~\ref{sec:test_statistic} provides details on the test
statistic that we use for NILE.
Appendix~\ref{sec:additional_experiments} contains additional
experiments. All proofs are provided in Appendix~\ref{app:proofs}.

\section{Framework} \label{sec:framework}
For a real-valued response 
$Y\in\R$ and predictors $X \in \R^d$,
we consider the problem
of
identifying
a regression function that works
well not only on the training data, but also under 
perturbed distributions that we will model by
interventions.

\subsection{Modeling Intervention-induced Distributions}\label{sec:modeling_int_ind_distr}
We
require a model that is able to 
model 
an observational 
distribution of $(X,Y)$ 
(as training distribution)
and the distribution of $(X,Y)$
under a
class of interventions on (parts of) $X$ (as test distribution).
We will do so by means of a
structural causal model (SCM) \citep{Bollen1989, Pearl2009}.
More precisely, denoting by $H \in \R^q$ some additional (unobserved) 
variables, we consider the~SCM
\begin{equation}
	\label{eq:SCMmodelreduced}
	H \coloneqq \ep_H, \,\,\,
	X \coloneqq h_2(H, \epsilon_X), \,\,\,
	Y \coloneqq f(X) + h_1(H, \epsilon_Y),
\end{equation}
where the assignments for $H$, $X$ and $Y$ consist of $q$, $d$ and $1$
coordinate(s), respectively.
Here, $f$, $h_1$ and $h_2$ are measurable functions, and the innovation
terms $\ep_X$, $\ep_Y$ and $\ep_H$ are independent vectors with
possibly dependent coordinates. 
Two comments
are in order. First, the joint distribution of $(X, Y)$ is constrained only by requiring
that $X$ and $h_1(H, \ep_Y)$ enter the assignment for $Y$
additively. This constraint affects the allowed conditional
distributions of $Y$ given $X$, but does not make any restriction on
the marginal distributions of either $X$ or $Y$. 
Second, we only use the above SCM as a technical device for
modeling training and test distributions, by considering
interventions on $X$ or $A$ (introduced in Section~\ref{sec:interventions}), for which we are
analyzing the predictive performance of different models -- similarly
to how one could have considered a ball around the training
distribution. 
We therefore only require the SCM to 
correctly (a) model the 
training-distribution, 
and (b) induce the
test-distributions through interventions.
Any other
causal implications of the SCM, such as causal orderings between
variables, causal effects or counterfactual statements, are not
assumed to be correctly specified.
As such, our framework includes a wide range of cases, 
including situations where training and test distribution come from interventions in 
an SCM with a different structure than \eqref{eq:SCMmodelreduced}, where, 
for example, some of the variables in $X$ are not ancestors but descendants of $Y$.
To see whether our framework applies, one needs to check if
the considered training and test distributions can be equivalently expressed as 
interventions in a model of our form. If the structure of the true 
data generating SCM is known, this can be done 
by directly 
transforming the SCM and the interventions. The following remark shows 
an example of such a transformation
and may be
interesting to readers with a special interest in causality.  It can
be skipped at first reading.

\begin{remark}[Transforming causal models] \label{rem:model}
	Assume that the training distribution is induced by the following SCM
	\begin{equation*}
		X_1 \coloneqq \ep_1, \quad
		X_2 \coloneqq k(Y) + \epsilon_2, \quad
		Y \coloneqq f(X_1) + \epsilon_3,
	\end{equation*}
	with $(\epsilon_1,\epsilon_2, \ep_3)\sim Q$, and that we consider 
	test distributions arising from shift interventions on $X_2$. This 
	set of training and test distributions can be equivalently modeled 
	by 
	the 
	reduced SCM
	\begin{equation*}
		H \coloneqq \ep_3, \quad
		X \coloneqq h_2(H,(\ep_1,\ep_2)), \quad
		Y \coloneqq f(X_1) + H,
	\end{equation*}
	with $(\epsilon_1,\epsilon_2, \ep_3)\sim Q$, and where
	$h_2$ is defined by $h_2(H,(\ep_1, \ep_2))\coloneqq(\ep_1, k(f(\ep_1)+H) + \ep_2)$. 
	Both SCMs induce the same observational distribution over $(X_1,X_2, Y)$ and
	shift interventions on $X_2$ in the original SCM correspond to shift interventions on 
	$X = (X_1, X_2)$ in the reduced SCM (where only the second coordinate is shifted). 
	Our framework can then be used, for example, to give sufficient conditions under 
	which generalization (formally defined below) is possible, see Proposition~\ref{prop:genX_intra}~and~\ref{prop:genX_extra}.
	It is not always possible to transform an SCM into our reduced form, and it might also happen that 
	the transformed interventions are not covered by our framework.
	For example, we do not allow for direct interventions on $Y$ in the original model. In 
	other cases, where the original SCM may contain additional hidden variables, even
	interventions on (parts of) $X$ in the original SCM may translate into interventions
	on $H$ in the reduced SCM, and are therefore not covered. 
	Details and a more general treatment %
	are provided in
	Appendix~\ref{sec:causal_relations_X}. 
\end{remark}

Sometimes, the vector of covariates $X$
contains variables, which are independent of $H$, that enter into the assignments of the other
covariates additively and cannot be used
for the prediction (e.g., because they are not observed during
testing).
If such covariates exist, 
it can be useful to explicitly distinguish them from the 
remaining predictors.
We will denote them by $A$ and call them exogenous variables. 
Such
variables are interesting for several reasons.
(i) 
We will see that in general, 
interventions on $A$ lead to intervention distributions with desirable properties for distribution generalization, see Section~\ref{sec:int_onA}.
(ii) 
Some of our results rely on 
the function $f$ being identifiable from the observational distribution, see Assumption~\ref{ass:identify_f} below.
The variables $A$ can be used to state explicit conditions for identifiability.
Under additional assumptions, for example,
they 
can be used as
instrumental variables \citep[e.g.,][]{Bowden1985, greene2003econometric}, 
a well-established tool for 
recovering $f$
from the observational distribution of 
$(X,Y,A)$.
(iii) The variable $A$ can be used to model a 
covariate that is not observed under testing. 
It can also be used to index tasks (which we discuss at the end of Section~\ref{sec:intro_focus}). 
In the remainder of this work, we therefore 
consider a slightly larger class of SCMs that also includes exogenous variables $A$. 
It contains the SCM \eqref{eq:SCMmodelreduced} as a special 
case.\footnote{This follows from choosing $A$ as an independent noise variable and a constant $g$.}
We derive results for settings with and without exogenous variables $A$.

\subsection{Model}
Formally, we consider a response $Y \in \R^1$, covariates $X\in \R^d$,
exogenous variables $A\in \R^r$, and unobserved variables
$H \in \R^q$.  Let further $\FF\subseteq\{f:\R^d\rightarrow\R\}$,
$\GG\subseteq\{g:\R^r\rightarrow\R^d\}$,
$\HH_1\subseteq\{h_1:\R^{q+1}\rightarrow\R\}$ and
$\HH_2\subseteq\{h_2:\R^{q+d}\rightarrow\R^d\}$ be fixed sets of
measurable functions. Moreover, let $\mathcal{Q}$ be a collection of
probability distributions on $\R^{d+1+r+q}$, such that for all
$Q\in\mathcal{Q}$ it holds that if
$(\epsilon_X,\epsilon_Y, \ep_A, \ep_H)\sim Q$, then
$\epsilon_X,\epsilon_Y, \ep_A$ and $\ep_H$ are jointly independent, and
for all $h_1\in\HH_1$ and $h_2\in\HH_2$ it holds that
$\xi_Y := h_1(\ep_H, \epsilon_Y)$ and
$\xi_X := h_2(\ep_H, \epsilon_X)$ have mean zero.\footnote{ This can
	be assumed w.l.o.g.\ if $\FF$ and $\GG$ are closed
	under addition and scalar multiplication, and contain the
	constant~function.}  Let
$\mathcal{M}\coloneqq
\FF\times\GG\times\HH_1\times\HH_2\times\mathcal{Q}$ denote the model
class. Every model $M=(f, g, h_1, h_2, Q)\in \cM$ then specifies an
SCM by\footnote{ For an appropriate choice of $h_2$, the model
	includes settings in which (parts of) $A$ directly influence
	$Y$.}\\
\begin{minipage}{0.45\columnwidth}
	\begin{align*}
		A &\coloneqq \ep_A 	\\
		H &\coloneqq \ep_H 	\\
		X &\coloneqq g(A) + h_2(H, \epsilon_X)\\
		Y &\coloneqq f(X) + h_1(H, \epsilon_Y)
	\end{align*}
\end{minipage}%
\begin{minipage}{0.55\columnwidth}
	\centering
	\vspace{1em} \resizebox{0.8\columnwidth}{!}{
		\begin{tikzpicture}[scale=1.6]
			\tikzstyle{VertexStyle} = [shape = circle, minimum width = 3em,
			fill=lightgray] \Vertex[Math,L=Y,x=1,y=0]{Y}
			\Vertex[Math,L=X,x=-1,y=0]{X} \Vertex[Math,L=H,x=0,y=1.5]{H}
			\Vertex[Math,L=A,x=-2.25,y=1]{A}
			\tikzset{EdgeStyle/.append style = {-Latex, line width=1}}
			\Edge[label=$f$](X)(Y) \Edge[label=$h_2$](H)(X)
			\Edge[label=$h_1$](H)(Y) \Edge[label=$g$](A)(X)
	\end{tikzpicture}} \vspace{1em}
\end{minipage}

\noindent with $(\epsilon_X,\epsilon_Y, \ep_A, \ep_H)\sim Q$, 
where the assignments for $A$, $H$ , $X$ and $Y$ consist of 
$r$, $q$, $d$ and~$1$ coordinate(s), respectively.
For each model $M = (f,g,h_1,h_2,Q) \in\mathcal{M}$, we refer to 
$f$ as the
\textit{causal function} (for the pair $(X,Y)$), 
and denote by $\P_M$ the joint distribution over the observed variables $(X, Y, A)$.
We assume that this distribution has finite second moments.
If no exogenous
variables $A$ exist, one can think of the function $g$ as being 
constant. A model $M$ that correctly models 
the training and test distributions will be referred to as the `true model'. 

\subsection{Interventions}\label{sec:interventions}
Each SCM $M\in\mathcal{M}$ can now be modified by the concept of
interventions \citep[e.g.,][]{Pearl2009,Peters2017}. An
intervention corresponds to replacing one or more of the structural
assignments of the SCM 
(see Section~\ref{sec:types_of_interventions} for details on the types of interventions considered in this paper). 
For example, we intervene on some of the covariates
$X$ by replacing the corresponding assignments with, e.g.,
a Gaussian random vector that is independent of the other noise variables.
Importantly, an
intervention on some of the variables does not change the assignment
of any other variable.  In particular, an intervention on $X$ does not
change the conditional distribution of $Y$, given $X$ and $H$ (this is
an instance of the invariance property mentioned in
Section~\ref{sec:intro}) 
but it may change the conditional distribution of $Y$, given $X$.  

The problems addressed in this work require us to simultaneously consider 
several different SCMs that are all subject to the same (set of) interventions.
Formally, we therefore regard an intervention $i$
as a mapping from the model class $\MM$ into a (possibly larger) set of SCMs,
which takes as input a model $M \in \MM$ and outputs another model $M(i)$
over variables $(X^i, A^i, Y^i, H^i)$, the intervened model.
We do not %
need to assume
that the intervened
model $M(i)$ belongs to the model class $\cM$, 
but we require that $M(i)$ induces a joint distribution over
$(X^i,Y^i,A^i,H^i)$\footnote{If the context does not allow for any
	ambiguity, we omit the superscript $i$.} with finite second
moments.
We denote the 
corresponding distribution over the observed $(X^i, Y^i, A^i)$ by $\P_{M(i)}$, 
and 
use $\cI$ for a collection of interventions.
In our work, the test distributions are modeled as distributions
generated by these types of intervened models, and the set
$\mathcal{I}$ therefore indexes 
the set of 
test distributions. We
will be 
interested in the mean squared prediction error on each test
distribution $i$, formally written as $\E_{M(i)}[(Y-f(X))^2]$.
(In this work, we consider a univariate $Y$, but writing
$\E [\| Y - \fs(X) \|_{\R^d}^2] = \sum_{j=1}^d \E [(Y_j - f_{\diamond, j}(X))^2]$, 
most 
our results extend straight-forwardly to a $d$-dimensional response.)

The support of random variables under interventions will play
an important role for the analysis of distribution generalization. 
Throughout this paper, $\supp^{M}(Z)$ denotes the support
of the random variable $Z\in\{A, X, H, Y\}$ under the distribution
induced by the SCM $M\in\mathcal{M}$. 
Moreover,
$\supp_{\cI}^{M}(Z)$ denotes the union of $\supp^{M(i)}(Z)$ over all
interventions $i\in\cI$. We call a collection of interventions on $Z$ 
\textit{support-reducing} (w.r.t.\ $M$) if $\supp_\cI^M(Z)\subseteq \supp^M(Z)$ and
\textit{support-extending} (w.r.t.\ $M$)
if
$\supp_\cI^M(Z)\not \subseteq \supp^M(Z)$. 
Whenever it is clear from the context which model is considered, we
may drop the indication of $M$ altogether and simply write $\supp(Z)$.

\subsection{Distribution Generalization} \label{sec:intro_focus}
Let $\MM$ be a fixed model class, 
let $M = (f,g,h_1, h_2, Q) \in \MM$
and let $\cI$ be a class of interventions.
In this work, we aim to find a function $f^*:\R^d\rightarrow\R$, 
such that the predictive model $\hat Y = f^*(X)$ has low worst-case
risk over all test distributions induced by the interventions $\cI$ in model $M$.
We therefore consider, for the true $M$, the optimization problem 
\begin{equation} \label{eq:minimax_problem}
	\argmin_{\fs \in \FF} \sup_{i \in \cI} \E_{M(i)} \big[ (Y - \fs(X))^2 \big],
\end{equation}
where $\E_{M(i)}$ is the expectation 
in the intervened model
$M(i)$. In general, this optimization problem is neither guaranteed to
have a solution, 
nor is the solution, if it exists, ensured
to be unique. 
Whenever a solution $f^*$ to~\eqref{eq:minimax_problem} exists, we refer to it as a
\textit{minimax solution} (for model $M$ w.r.t.\ ($\mathcal{F},\cI$)).

Depending on the model class $\MM$, there may be several models $\tilde{M} \in \MM$ that induce the observational distribution $\P_M$, that is, the same distribution over the observed variables $A$, $X$ and $Y$, but do not agree with $M$ on all intervention distributions %
induced by 
$\cI$. 
Thus, each such model induces a potentially different minimax problem with different solutions. 
Given knowledge only of
$\P_M$, it is therefore generally not possible to identify a solution
to \eqref{eq:minimax_problem}.
In this paper, we study conditions
on $\MM$, $\P_M$ and $\cI$, under which this becomes possible. More
precisely, we aim to characterize under which conditions $(\P_M, \mathcal{M})$
admits distribution generalization to $\mathcal{I}$.

\begin{definition}[Distribution generalization]
	\label{defi:general}
	$(\P_{M},\mathcal{M})$ is said to 
	\emph{admit distribution generalization to} $\cI$, 
	or simply to \emph{admit generalization to} $\cI$,
	if for every
	$\epsilon > 0$ there exists a function $f^{*}_\varepsilon\in\mathcal{F}$ such
	that, for all models $\tilde{M}\in\mathcal{M}$ with
	$\P_{\tilde{M}}=\P_{M}$, it holds that %
	\begin{align} \label{eq:def_generalization}
		\begin{split}
			&\ \left \vert \sup_{i\in\cI}\E_{\tilde{M}(i)}\big[(Y-f_\varepsilon^*(X))^2\big] \right.
			\left.- \inf_{\fs\in\FF}\,
			\sup_{i\in\cI}\E_{\tilde{M}(i)}\big[(Y-\fs(X))^2\big] \right \vert \leq \epsilon.
		\end{split}
	\end{align}
\end{definition}
Distribution generalization does not require the existence of a minimax solution in
$\FF$ (which would 
require
further assumptions on the function class $\FF$)
and instead focuses on whether an approximate solution can be
identified based only on the observational distribution $\P_{M}$.
If, however, there exists a function $f^* \in \FF$ which, for every
$\tilde{M}\in\MM$ with $\P_{\tilde{M}} = \P_M$, is a minimax solution for $\tilde{M}$
w.r.t.\ $(\FF, \cI)$, then, in particular,
$(\P_{M},\mathcal{M})$ admits generalization to $\cI$.

Our framework also includes several settings of multitask learning
(MTL) 
and domain adaptation \citep{Candela2009}, where one often assumes to observe different training tasks. 
In MTL,
one is then interested in using the 
different tasks to improve the predictive performance on either one or all training tasks -- this is often referred to as asymmetric and symmetric MTL, respectively.
In our framework, 
such a setup
can be modeled using a categorical
variable $X$. 
If, however, 
one is interested in predicting on an unseen task
or if one does not know which of the observed tasks the new test data come from, one may instead use a categorical $A$ 
with support-extending or support-reducing interventions,
respectively.

\section{Minimax Solutions and the Causal Function} \label{sec:robustness}
To address the question of distribution generalization, we first 
study properties of the minimax optimization problem \eqref{eq:minimax_problem}. 
In the simplest case, where
$\cI$ consists only of the trivial intervention, that
is, $\P_M = \P_{M(i)}$, we are looking for the best predictor on the
observational distribution.  In that case, the minimax solution is 
attained at
any conditional mean function,
$f^*: x\mapsto\E[Y\vert X=x]$ (provided that 
$f^*\in\mathcal{F}$).  For larger classes of interventions, however, the
conditional mean may become sub-optimal in terms of prediction.
To see this, it is instructive to decompose the risk under 
an intervention.
Since the structural assignment for $Y$ remains unchanged 
for all interventions that we consider in this work, it holds
for all $\fs \in \FF$ and all interventions $i$ on either $A$ or $X$ that
\begin{align*}
	\E_{M(i)}[(Y - \fs(X))^2] = &\, \E_{M(i)}[(f(X) - \fs(X))^2]+\E_{M}[\xi_Y^2] \\&\quad +2\E_{M(i)}[\xi_Y(f(X)-\fs(X))].
\end{align*}
Here, the middle term does not depend on $i$ since
$\xi_Y = h_1(H, \ep_Y)$ remains fixed. 
We call the intervention $i$
\begin{align*}
	\begin{array}{ll}
		&\quad\text{\emph{confounding-removing}}\phantom{ii}\qquad
		\begin{array}{ll}
			\text{if for all models } M \in \mathcal{M} \text{ it holds that }\\
			X \indep H, \text{ under } M(i).
		\end{array}
	\end{array}
\end{align*} 
For such an intervention, we have that $\xi_Y\indep X$ under $\P_{M(i)}$, and hence, since 
$\E_{M}[\xi_Y] = 0$, the last term in the above equation
vanishes.
Therefore, if 
$\cI$ consists only of confounding-removing
interventions, the causal function is a solution to the
minimax
problem \eqref{eq:minimax_problem}. 
The
following proposition shows that an even stronger statement holds:
The causal function is already a minimax solution if $\cI$ contains
at least one confounding-removing intervention on~$X$.

\begin{proposition}[Confounding-removing interventions on $X$]
	\label{prop:minimax_equal_causal}
	Let $\cI$ be a set of interventions on $X$ or $A$ 
	such that
	there
	exists at least one $i\in\cI$ 
	that is confounding-removing.
	Then, the minimal worst-case risk is attained at a
	confounding-removing intervention, and the causal function $f$ is a
	minimax solution. %
\end{proposition}
We now prove that, 
in a linear setting, 
the causal
function is also minimax optimal
if the
interventions create unbounded variability in all directions of the covariance matrix of $X$.
\begin{proposition}[Unbounded interventions on $X$ with linear $\mathcal{F}$]
	\label{prop:shift_interventions}
	Let $\FF$ be the class of all linear functions, and let $\cI$ be a
	set of interventions on $X$ or $A$ s.t.\
	$\sup_{i\in \cI} \lambda_{\min}\big(\E_{M(i)}\big[XX^\top\big]\big) =\infty$,
	where $\lambda_{\min}$ denotes the smallest eigenvalue.
	Then, the
	causal function $f$ is the unique minimax solution. 
\end{proposition}
The unbounded eigenvalue condition above is satisfied 
if $\cI$ is the set of all shift interventions on $X$.
These interventions, formally defined in Section~\ref{subsec:intallowgen},
appear in linear IV models and recently gained further attention in the causal community \citep{rothenhausler2018anchor, Sani2020}. 
The proposition above considers a linear function class $\mathcal{F}$; in this way, shift interventions are related to linear models.

Even if the 
causal function $f$ does not 
solve the minimax problem~\eqref{eq:minimax_problem},
the difference
between the minimax solution and the causal function
cannot
be arbitrarily large.  
The following proposition shows that the worst-case 
$L_2$-distance between $f$ and any function $\fs$ that performs better than $f$ (in terms of worst-case risk) can be bounded by a term which is related to the strength of the confounding. 
\begin{proposition}[Difference between causal function and minimax solution]
	\label{prop:difference_to_causal_function}
	Let $\cI$ be a set of interventions on $X$ or $A$. Then, for any
	function $\fs\in\FF$ which satisfies that
	\begin{equation*}
		\sup_{i\in\cI}\E_{M(i)}[(Y-\fs(X))^2]\leq\sup_{i\in\cI}\E_{M(i)}[(Y-f(X))^2],
	\end{equation*}
	it holds that
	\begin{equation*}
		\sup_{i\in\cI}\E_{M(i)}[(f(X)-\fs(X))^2]\leq 4\var_M[\xi_Y].
	\end{equation*}
\end{proposition}
Even though the difference can be bounded, it may be non-zero, and one may %
benefit
from choosing a function that differs from the causal function $f$.
This choice, however, comes at a cost: it relies on the fact that we know the class of interventions $\mathcal{I}$. 
In general, being a minimax solution is not entirely
robust with respect to misspecification of $\mathcal{I}$. 
In particular, if the 
set $\cI_2$ of interventions
describing the test distributions is misspecified by a set
$\cI_1\neq\cI_2$, then the considered minimax solution with respect
to $\cI_1$ may perform worse than the causal function on the test
distributions. 

\begin{proposition}[Properties of the minimax solution under mis-specified
	interventions]
	\label{prop:misspecification_minimax}
	Let $\cI_1$ and $\cI_2$ be any two sets of interventions on
	$X$, and let $f_1^*\in\mathcal{F}$ be a minimax solution w.r.t.\
	$\cI_1$.  Then, if $\cI_2\subseteq\cI_1$, it holds that
	\begin{equation*}
		\sup_{i\in\cI_2}\E_{M(i)}\big[(Y-f_1^*(X))^2\big] \leq \sup_{i\in\cI_2}\E_{M(i)}\big[(Y-f(X))^2\big].
	\end{equation*}
	If $\cI_2\not\subseteq\cI_1$, however, it can happen (even if
	$\mathcal{F}$ is linear) that
	\begin{equation*}
		\sup_{i\in\cI_2}\E_{M(i)}\big[(Y-f_1^*(X))^2\big] > \sup_{i\in\cI_2}\E_{M(i)}\big[(Y-f(X))^2\big].
	\end{equation*}
\end{proposition}
The second part of 
the proposition should be understood as 
a non-robustness
property of non-causal minimax solutions.
Improvements on the causal 
function are possible 
in situations, where 
one has reasons to believe that 
the test distributions do not stem from 
a set of interventions that is much larger than the specified set.

\section{Distribution Generalization}\label{sec:generalizability}
As described in Section~\ref{sec:intro_focus},
we consider a fixed model class $\mathcal{M}$ containing the true (but unknown) 
model $M$, and let $\mathcal{I}$ be a class of interventions.
By definition, the optimizer of the minimax problem~\eqref{eq:minimax_problem}
depends on the true model $M$.
Section~\ref{sec:robustness} relates 
this optimizer
to the causal function $f$, 
whose knowledge, too, requires knowing $M$.
In practice, however, we do not have access to the true model $M$, but only to its observational distribution $\P_M$. 
This motivates the notion of distribution generalization, see~\eqref{eq:def_generalization}. 
In words, it states that approximate minimax solutions (which depend on the intervention distributions $\P_{M(i)}$, $i \in \cI$)
are identified from the observational distribution $\P_M$. This holds true, in particular, if the 
intervention distributions themselves are identified from $\P_M$.
\begin{proposition}[Sufficient conditions for distribution generalization]
	\label{prop:suff_general}
	Assume that for all $\tilde{M} \in \MM$ it holds that
	\begin{equation*}
		\P_{\tilde{M}} = \P_M \quad \Rightarrow \quad \P^{(X,Y)}_{\tilde{M}(i)} = \P^{(X,Y)}_{M(i)} \quad \forall i \in \cI,
	\end{equation*}
	where $\P^{(X,Y)}_{M(i)}$ 
	is the joint distribution of $(X, Y)$ under
	$M(i)$. Then, $(\P_M, \MM)$ admits generalization to~$\cI$.
\end{proposition}
Proposition~\ref{prop:suff_general} provides verifiable conditions for
distribution generalization, and 
can be used to prove
possibility statements. 
It is, however, not a necessary condition.
Indeed, we will see that, under certain types of interventions, 
distribution generalization becomes possible even in cases
where the interventional marginal of~$X$ is not identified.

In this section, we study conditions on $\MM$, $\P_M$ and $\cI$ which ensure 
generalization, and present 
corresponding impossibility results proving the necessity of some of these conditions. 
Two aspects will be of central importance. The first is related
to causal identifiability, i.e., whether the causal function $f$ is sufficiently identified from the observational 
distribution $\P_M$ (Section~\ref{sec:causal_identifiability}). The other aspect is related to the 
types of interventions (Section~\ref{sec:types_of_interventions}). 
We consider interventions on $X$ in
Section~\ref{sec:gen_int_on_X} and interventions on $A$ in
Section~\ref{sec:int_onA}.  
Parts of our results are summarized in Table~\ref{tab:generalizability}.
\begin{table}
	\centering
	{%
		\scriptsize
		\renewcommand{\arraystretch}{1.15}
		\begin{tabular}{>{\centering\arraybackslash}p{2.5cm}>{\centering\arraybackslash}p{2.5cm}>{\centering\arraybackslash}p{4cm}|>{\centering\arraybackslash}p{3cm}}
			\toprule    
			Intervention on & $\supp_{\cI}(X)$ & Assumptions & Result  \\
			\toprule
			$X$ (well-behaved) & $ \subseteq \supp(X)$ & Assumption~\ref{ass:identify_f} & Proposition~\ref{prop:genX_intra} \\
			$X$ (well-behaved) & $ \not \subseteq \supp(X)$ &Assumptions~\ref{ass:identify_f}~and \ref{ass:gen_f} & Proposition~\ref{prop:genX_extra}\\
			$A$ & $ \subseteq \supp(X)$ & Assumptions \ref{ass:identify_f} and \ref{ass:identify_g} & Proposition \ref{prop:genA}\\
			$A$ & $\not \subseteq \supp(X)$ & Assumptions~\ref{ass:identify_f},~\ref{ass:gen_f}~and~\ref{ass:identify_g} & Proposition \ref{prop:genA}\\
			\arrayrulecolor{black}\bottomrule
		\end{tabular}
	}
	\caption{Summary of conditions under which generalization is
		possible. Corresponding impossibility results are shown in
		Propositions~\ref{prop:impossibility_interpolation},~\ref{prop:impossibility_extrapolation} and~\ref{prop:impossibility_intA}.}
	\label{tab:generalizability}
\end{table}

\subsection{Identifiability of the Causal Function} \label{sec:causal_identifiability}
For specific types of interventions, the causal function $f$ is itself a minimax solution,
see Propositions~\ref{prop:minimax_equal_causal}~and~\ref{prop:shift_interventions}. 
If, in addition,
these interventions are 
support-reducing, 
generalization is directly implied by the following assumption.

\begin{assumption}[Identifiability of $f$ on the support of $X$]
	\label{ass:identify_f}
	For all
	$\tilde{M}=(\tilde{f}, \dots) \in \mathcal{M}$ with $\P_{\tilde{M}} = \P_{M}$, it holds
	that $\tilde{f}(x) = f(x)$ for all $x \in \supp(X)$.
\end{assumption}

Assumption~\ref{ass:identify_f} will play a central role in proving distribution generalization
even in situations where the causal function is not a minimax solution.
We 
use it as a starting point for 
most
of our results. 
The assumption
is
violated, for example, in a linear Gaussian setting with a single covariate $X$ (without $A$). Here, in general, we cannot identify $f$ and distribution generalization does not hold. 
Assumption~\ref{ass:identify_f}, however, is not necessary for  generalization.
In Section, \ref{sec:int_onA} we discuss 
a linear setting
where 
distribution generalization is possible, even if Assumption~\ref{ass:identify_f} does not hold.

The question of causal identifiability
has received a lot of attention in the literature. In linear instrumental
variables settings, for example, one assumes that the functions $f$ and
$g$ are linear and 
identifiability follows if
the product moment between $A$ and $X$ has rank at
least the dimension of $X$
\cite[e.g.,][]{wooldridge2010econometric}.
In
linear non-Gaussian models, %
one can identify the function
$f$ even 
if there are no instruments
\citep{Hoyer2008b}. %
For nonlinear models, restricted SCMs can be
exploited, too. In that case, Assumption~\ref{ass:identify_f} holds
under regularity conditions if $h_1(H, \epsilon_Y)$ is independent of
$X$ \citep{Zhang2009, peters2014causal, Peters2017}
and first attempts 
have been made
to extend such results to
non-trivial confounding cases 
\citep{Janzing2009uai}.  
The nonlinear IV setting \citep[e.g.,][]{amemiya1974nonlinear,newey2013nonparametric,newey2003instrumental}
is discussed in more detail in 
Appendix~\ref{sec:IVconditions}, where we give a brief overview of
identifiability
results for linear, parametric and non-parametric function classes. 
Assumption~\ref{ass:identify_f} states that $f$ is identifiable, 
even on $\P_M$-null sets, which 
is usually achieved by placing further 
constraints on the function class, such as smoothness.
Even though this issue seems 
technical, it becomes important when considering hard interventions
that
set $X$ to a fixed value, for example.

\subsection{Types of Interventions} \label{sec:types_of_interventions}
Whether distribution generalization is admitted depends on
the intervention class $\mathcal{I}$.
In this work,
we only consider interventions on the covariates $X$ and $A$. 
Each of these types of interventions can be characterized by a measurable 
function $\psi^i$,
which determines the structural assignment of the intervened variable, and a (possibly degenerate) random vector $I^i$,
which serves as an independent noise innovation. More formally, 
for an intervention on $X$, the pair $(\psi^i, I^i)$ defines the intervention which
maps the input model $M=(f,g,h_1,h_2,Q)\in \cM$ to the intervened model $M(i)$ given by the assignments
\begin{align*}
	A^i &:= \ep_A^i, \quad H^i := \ep_H^i, \\
	X^i &:= \psi^i(g, h_2, A^i,H^i, \ep^i_X ,I^i), \\
	Y^i &:= f(X^i) + h_1(H^i,\ep_Y^i).
\end{align*}
Similarly, for an intervention on $A$, 
$(\psi^i, I^i)$ specifies the intervention which outputs %
\begin{align*}
	A^i &:= \psi^i(I^i, \ep_A^i), \quad  H^i := \ep_H^i, \\
	X^i &:= g(A^i) + h_2(H^i, \ep_X^i), \\
	Y^i &:= f(X^i) + h_1(H^i,\ep_Y^i).
\end{align*}
In both cases, $(\ep_X^i, \ep_Y^i, \ep_A^i,\ep_H^i) \sim Q$ and $I^i \indep (\ep_X^i, \ep_Y^i, \ep_A^i,\ep_H^i)$.
We will see below that this
class of interventions is rather flexible.
It does, however, not allow for arbitrary manipulations of $M$. 
For example, it does not allow for changes in the structural assignments for $Y$ or $H$, 
or for the noise variable $\ep_Y^i$ to enter the assignment of the intervened variable. 
As the following section highlights, further constraints on the types of interventions are 
necessary to ensure distribution generalization.

\subsubsection{Impossibility of Generalization Without Constraints on the Interventions}

Let $\mathcal{Q}$ be a class of product 
distributions on $\R^4$, such that for all $Q \in \mathcal{Q}$, 
the coordinates of $Q$ are non-degenerate, zero-mean with finite second moment. 
Let $\MM$ be the class of all models of the form 
\begin{equation*}
	A\coloneqq \epsilon_A, \quad %
	H\coloneqq\sigma\epsilon_H, \quad %
	X\coloneqq \gamma A + \epsilon_X + \tfrac{1}{\sigma}H, \quad %
	Y\coloneqq \beta X + \epsilon_Y + \tfrac{1}{\sigma}H,
\end{equation*}
with $\gamma, \beta \in \R$, $\sigma > 0$ and $(\epsilon_A, \epsilon_X,\epsilon_Y,\epsilon_H) \sim Q \in \mathcal{Q}$. 
Assume that $\P_M$ is induced by some model $M = M(\gamma, \beta, \sigma, Q)$ from the above model class
(here, we slightly adapt the notation from Section~\ref{sec:framework}). 
The following proposition shows that, without constraining the set of interventions $\cI$, 
distribution generalization is not always ensured.

\begin{proposition}[Impossibility of generalization without constraining the class of interventions]
	\label{prop:impossibility_interpolation}
	Assume that $\MM$ is given as defined above, 
	let $\cI \subseteq \R_{>0}$ be 
	a 
	compact, non-empty
	set and define the interventions on $X$
	by $\psi^i(g, h_2, A^i, H^i, \epsilon_X^i, I^i) = iH$, for
	$i \in \cI$.
	Then, $(\P_M, \cM)$ does not admit generalization to $\cI$
	(even if Assumption~\ref{ass:identify_f} is satisfied).
	In addition,
	any prediction model other than the causal model
	may perform arbitrarily bad under the interventions $\cI$. 
	That is, for any $b \neq \beta$ and any $c > 0$,
	there exists a model $\tilde{M}\in\cM$
	with $\P_{\tilde{M}} = \P_{M}$, such that
	\begin{equation*}
		\abs[\Big]{\sup_{i\in\cI}\E_{\tilde{M}(i)}\big[(Y-bX)^2\big]
			- \inf_{\bes \in \R}\,
			\sup_{i\in\cI} \E_{\tilde{M}(i)}\big[(Y-\bes X)^2\big]} \geq c.
	\end{equation*}
\end{proposition}
We now give some intuition about the above result.
By definition, distribution generalization is ensured if there exist prediction functions
that are (approximately) minimax optimal for all models which induce the same observational
distribution as $M$. Since, in the above example, the distribution of $(X,Y,A)$ does not depend on $\sigma$, 
this includes all models of the form $M_{\tilde{\sigma}} = M(\gamma, \beta, \tilde{\sigma}, Q)$
for some $\tilde{\sigma} > 0$. However, while agreeing on the observational distribution, each of these 
models induces fundamentally different intervention distributions
(under $M_{\tilde{\sigma}}(i)$, $(X,Y)$ is equal in distribution to $(i \epsilon_H, (\beta i + \frac{1}{\tilde{\sigma}}) \epsilon_H)$)
and results in different (approximate) minimax solutions. 
Below, we introduce two types of interventions which ensure distribution generalization 
in a wide range of settings by constraining the influence of $H$ on $X$.

\subsubsection{Interventions Which Allow for Generalization} \label{subsec:intallowgen}
In Section~\ref{sec:robustness}, we already introduced confounding-removing interventions, 
which break the dependence between $X$ and $H$.
For an intervention set $\cI$ which contains at least one confounding-removing intervention, 
the causal function $f$ is always a minimax solution (see Proposition~\ref{prop:minimax_equal_causal})
and, in the case of support-reducing interventions, distribution generalization is therefore achieved by requiring Assumption~\ref{ass:identify_f} to hold.
The intervention $i$ 
with intervention map $\psi^i$
is called 
\begin{align*}
	\begin{array}{c}
			\emph{confounding-preserving}\text{ if there exists a map } \phi^i, \text{ such that }\\
			\psi^i(g, h_2, A^i ,H^i, \ep^i_X ,I^i) = \phi^i(A^i,g(A^i),h_2(H^i, \ep^i_X) ,I^i).
		\end{array}
\end{align*}
Confounding-preserving interventions contain, e.g., \textit{shift interventions} on $X$,
which linearly shift the original assignment by $I^i$, that is,
$\psi^i(g, h_2, A^i,H^i, \ep^i_X ,I^i) = g(A^i) + h_2(H^i, \ep_X^i) +
I^i$. 
The name 
`confounding-preserving'
stems from the fact that the confounding
variables $H$ only enter the intervened structural assignment of $X$
via the term $h_2(H^i, \ep^i_X)$,
which is the same as in the original model.
(This property fails to hold true for the interventions in Proposition~\ref{prop:impossibility_interpolation}.)
If $\cI$ consists only of confounding-preserving
interventions, the causal function is generally not a
minimax solution. 
However, we will see that, under Assumption~\ref{ass:identify_f}, these types 
of interventions lead to identifiability of the intervention distributions $\P_{M(i)}$, $i \in \cI$, and therefore
ensure 
generalization via Proposition~\ref{prop:suff_general}.

Some interventions are 
both
confounding-removing and confounding-preserving, but
not every confounding-removing
intervention is 
confounding-preserving. For example, the
intervention 
$\psi^i(g, h_2, A^i ,H^i, \ep^i_X ,I^i)=\ep^i_X$
is confounding-removing but,
in general,
not confounding-preserving.
Similarly, not all confounding-preserving interventions
are confounding-removing. %
We call a set of interventions $\cI$
\textit{well-behaved} either if it consists only of
confounding-preserving interventions or if it contains at least one
confounding-removing intervention.

\subsection{Generalization to Interventions on $X$} \label{sec:gen_int_on_X}
We now formally prove in which sense the two types of interventions defined above 
allow for distribution generalization. We will see that this question is closely linked to the
relation between the support of $\P_M$ and the support of the
intervention distributions.  Below, we therefore distinguish between
support-reducing and support-extending interventions on~$X$.

\subsubsection{Support-reducing Interventions} \label{sec:supp_reducing_onX}
For support-reducing interventions, Assumption~\ref{ass:identify_f} is sufficient 
for distribution generalization even in nonlinear settings, under a large class of interventions.
\begin{proposition}[Generalization to support-reducing interventions on $X$]
	\label{prop:genX_intra}
	Let $\cI$ be a well-behaved set of 
	interventions on $X$, and
	assume that $\supp_{\cI}(X)\subseteq\supp(X)$. 
	Then, under Assumption~\ref{ass:identify_f}, 
	$(\P_{M}, \cM)$ admits generalization to the interventions $\cI$.
	If one of the interventions is confounding-removing, then the causal function is a minimax solution.
\end{proposition}
In the case of support-extending interventions, further assumptions are required to ensure distribution generalization.

\subsubsection{Support-extending Interventions}\label{sec:support_extending_onX}

If the interventions in $\cI$ extend the support of $X$, i.e.,
$\supp_\cI(X) \not \subset \supp(X)$, Assumption~\ref{ass:identify_f}
is not sufficient for ensuring distribution generalization. 
This is because there may exist a model $\tilde{M} \in \MM$ which agrees with $M$ on the observational distribution, 
but whose corresponding causal function $\tilde{f}$ differs from $f$ outside of the support of $X$.
In that case, a
support-extending intervention on~$X$ may result in different
dependencies between $X$ and $Y$ in the two models, and therefore
potentially induce a different set of minimax solutions. 
The following assumption on the model class $\FF$
ensures that any $f\in \cF$ is uniquely determined by its values
on $\supp(X)$.
\begin{assumption}[Extrapolation of $\FF$] \label{ass:gen_f} For all
	$\tilde{f}, \bar{f} \in \FF$ with $\tilde{f}(x) = \bar{f}(x)$ for all
	$x \in \supp(X)$, it holds that $\tilde{f} \equiv \bar{f}$.
\end{assumption}
We will see that this assumption is sufficient
(Proposition~\ref{prop:genX_extra}) for generalization to well-behaved interventions on $X$. Furthermore, it is also necessary
(Proposition~\ref{prop:impossibility_extrapolation}) if $\cF$ is sufficiently flexible.
The following proposition  can be seen as an extension of Proposition~\ref{prop:genX_intra}. 
\begin{proposition}[Generalization to support-extending interventions on $X$]
	\label{prop:genX_extra}
	Let $\cI$ be a well-behaved set of 
	interventions on $X$.
	Then, under Assumptions~\ref{ass:identify_f}~and~\ref{ass:gen_f}, 
	$(\P_{M}, \cM)$ admits generalization to $\cI$.
	If one of the interventions is confounding-removing, then the causal function is a minimax solution. %
\end{proposition}
Because the interventions 
may change the marginal distribution of~$X$, 
the preceding proposition 
includes examples, 
in which 
distribution generalization is possible
even if 
some of the  considered
joint (test) distributions
are arbitrarily far 
from the training distribution, 
in terms of any reasonable divergence measure over
distributions, such as Wasserstein distance or $f$-divergence.

Proposition~\ref{prop:genX_extra} relies on Assumption~\ref{ass:gen_f}. Even though this assumption
is restrictive, it is satisfied by several reasonable function classes, which therefore allow for generalization
to any set of well-behaved interventions.  
Below, we give two examples
of such 
function classes.

\paragraph*{Sufficient conditions for generalization} 

Assumption~\ref{ass:gen_f} states that every function in $\FF$ is globally identified
by its values on $\supp(X)$. This is, for example, satisfied if 
$\mathcal{F}$ is a linear space of functions with domain $\mathcal{D} \subset \R^d$
which are linearly independent on $\supp(X)$. More precisely, 
$\mathcal{F}$ is linearly closed, i.e.,
\begin{align}
	f_1, f_2 \in \mathcal{F}, c \in \R, \implies
	f_1 + f_2 \in \mathcal{F}, cf_1 \in \mathcal{F},
	\label{eq:linear}
\end{align}
and $\mathcal{F}$ is linearly independent on $\supp(X)$, i.e.,
\begin{align}
	f_1(x) = 0 \quad \forall x \in \supp(X) \;\implies \;f_1(x) = 0 \quad \forall x \in \mathcal{D}.
	\label{eq:linearind}
\end{align}
Examples of such classes include (i) globally linear parametric
function classes, i.e., $\FF$ is of the form
\begin{equation*}
	\mathcal{F}^1 \coloneqq\{\fs:\mathcal{D}\rightarrow\R \given \exists\gamma \in \R^k \text{ s.t. } 
	\forall
	x\in\mathcal{D} \st
	\fs(x)=\gamma^\top \nu (x) \},
\end{equation*}
where $\nu = (\nu_1, \dots, \nu_k)$ consists of real-valued, linearly
independent functions satisfying that $\E_M[\nu(X) \nu(X)^\top]$ is
strictly positive definite, and (ii) the class of differentiable
functions that extend linearly outside of $\supp(X)$, that is, $\FF$
is of the form
{\small
\begin{equation*}
	\FF^2 := \left\lbrace \fs:\mathcal{D}\rightarrow\R \, \bigg\vert \,
	\begin{tabular}{@{}l@{}}
		$\fs \in C^1 \text{ and } \forall x\in\mathcal{D} \setminus  \supp(X):$
		$\fs(x)=\fs(x_b)+\nabla\fs(x_b)(x-x_b)$
	\end{tabular}
	\right\rbrace
\end{equation*}}
where $x_b\coloneqq\argmin_{z\in\supp(X)}\norm{x-z}$ 
and
$\supp(X)$ is assumed to be closed
with non-empty interior.
Clearly, both of the above function classes are linearly closed. To
see that $\FF^1$ satisfies \eqref{eq:linearind}, let $\gamma \in \R^k$
be s.t.\ $\gamma^\top \nu (x) = 0$ for all $x \in \supp(X)$. Then, it
follows that
$0 = \E_M[(\gamma^\top \nu(X))^2] = \gamma^\top \E_M[\nu(X)
\nu(X)^\top] \gamma$ and hence that $\gamma = 0$. To see that $\FF^2$
satisfies \eqref{eq:linearind}, let $\fs \in \FF^2$ and assume that
$\fs(x) = 0$ for all $x \in \supp(X)$.  Then,  $\fs(x) = 0$ for
all $x \in \mathcal{D}$ and thus  $\FF^2$ uniquely defines the
function on the entire domain $\mathcal{D}$.

By Proposition~\ref{prop:genX_extra}, generalization with respect to
these model classes is possible for any 
well-behaved
set of 
interventions. 
In practice, it may often be more realistic to impose bounds on the
higher order derivatives of the functions in $\FF$.  We now prove that
this still allows for 
what we will call
approximate distribution generalization, 
see Propositions~\ref{prop:extrapolation_bounded_deriv_cr} and~\ref{prop:extrapolation_bounded_deriv}.

\paragraph*{Sufficient conditions for approximate generalization} 
For differentiable functions, exact generalization cannot always be achieved. 
Bounding the first derivative, however, allows us to achieve
approximate generalization. We therefore consider 
the following
function class
\begin{equation}\label{eq:boundedder}
	\FF^2 := \{  \fs:\mathcal{D} \rightarrow \R \given 
	\fs \in C^1 \text{ with } \norm{\nabla\fs}_{\infty} \leq K \}
\end{equation}
for some fixed $K<\infty$, where $\nabla\fs$ denotes the 
gradient
and
$\mathcal{D}\subseteq\R^d$. We then have the following result.

\begin{proposition}[Approx. generalization with bdd. derivatives (confounding-removing)]
	\label{prop:extrapolation_bounded_deriv_cr}
	Let $\mathcal{F}$ be as defined in \eqref{eq:boundedder}.
	Let $\cI$ be a set of 
	interventions on $X$ containing at least one confounding-removing
	intervention, and assume that Assumption~\ref{ass:identify_f} holds
	true.
	(In this case, the causal function $f$ is a minimax solution.)
	Then, for all $f^*$
	with $f^*=f$ on $\supp(X)$ and all
	$\tilde{M}\in\mathcal{M}$ with $\P_{\tilde{M}}=\P_{M}$, it holds that
	\begin{align*}\label{eq:gen_cond}
		 \abs[\Big]{\sup_{i\in\cI}\E_{\tilde{M}(i)}\big[(Y-f^*(X))^2\big]
			- \inf_{\fs\in\FF}\,
			\sup_{i\in\cI}\E_{\tilde{M}(i)}\big[(Y-\fs(X))^2\big]}
		\\\leq
		4\delta^2K^2+4\delta K \sqrt{\var_M(\xi_Y)},
	\end{align*}
	where
	$\delta :=
	\sup_{x\in\supp_{\cI}^{M}(X)}\inf_{z\in\supp^{M}(X)}\norm{x-z}$. 
	If $\cI$ consists only of confounding-removing interventions, 
	the same statement holds when replacing
	the bound
	by
	$4\delta^2K^2$.
\end{proposition}
Proposition~\ref{prop:extrapolation_bounded_deriv_cr} states that the
deviation of the worst-case generalization 
error from the best possible value 
is bounded by a term that grows with the
square of $\delta$. 
Intuitively, this means that under the function
class defined in~\eqref{eq:boundedder}, approximate generalization is
reasonable only for interventions that are close to the support of
$X$. 
We now prove a
similar result 
for
cases in which the minimax
solution is not necessarily the causal function. The following
proposition bounds the worst-case generalization error for
arbitrary confounding-preserving interventions. 
Here, the bound 
additionally accounts for the approximation to the minimax solution.
\begin{proposition}[Approx. generalization with bdd. derivatives
	(confounding-preserving)]
	\label{prop:extrapolation_bounded_deriv}
	Let $\mathcal{F}$ be as defined in \eqref{eq:boundedder}. Let $\cI$
	be a set of confounding-preserving interventions on $X$, and assume
	that Assumption~\ref{ass:identify_f} is satisfied. 
	Let $\epsilon > 0$ and let $f^{*}\in\mathcal{F}$ be such that, 
	\begin{align*}
		\left\vert\sup_{i\in\cI}\E_{M(i)}\big[(Y-f^*(X))^2\big]\right.
		\left. - \inf_{\fs\in\FF}\,
		\sup_{i\in\cI}\E_{M(i)}\big[(Y-\fs(X))^2\big]\right\vert \leq \epsilon.
	\end{align*}
	Then, for all $\tilde{M}\in\mathcal{M}$ with
	$\P_{\tilde{M}}=\P_{M}$, it holds that
	\begin{align*}
		&\abs[\Big]{\sup_{i\in\cI}\E_{\tilde{M}(i)}\big[(Y-f^*(X))^2\big]
			- \inf_{\fs\in\FF}\,
			\sup_{i\in\cI}\E_{\tilde{M}(i)}\big[(Y-\fs(X))^2\big]} \\
		&\quad 
		\leq \ep + 12 \delta^2 K^2 + 32 \delta K \sqrt{\var_M(\xi_Y)} +
		4 \sqrt{2} \delta K \sqrt{\ep}
	\end{align*}
	where
	$\delta :=
	\sup_{x\in\supp_{\cI}^{M}(X)}\inf_{z\in\supp^{M}(X)}\norm{x-z}$. 
\end{proposition}
We 
can take $f^*$ to be the minimax
solution if it exists. In that case, the terms involving $\ep$ 
disappear from the bound, which 
then becomes more similar to the one in
Proposition~\ref{prop:extrapolation_bounded_deriv_cr}.

\paragraph*{Impossibility of generalization without constraints on $\FF$}
If we do not constrain the function class $\FF$, generalization is
impossible. 
Even if
we consider the set
of all continuous functions $\FF$, 
we cannot generalize to
interventions outside the support of $X$.
This statement holds
even if
Assumption~\ref{ass:identify_f} is satisfied.

\begin{proposition}[Impossibility of extrapolation]
	\label{prop:impossibility_extrapolation}
	Assume that
	$\FF = \{\fs: \R^d \to \R \mid \fs\text{ is continuous}\}$.  Let $\cI$
	be a well-behaved set of  support-extending  interventions on $X$,
	such that $\supp_{\cI}(X) \setminus \supp(X)$ has non-empty
	interior.  Then, $(\P_{M}, \cM)$ does not admit generalization to $\cI$, 
	even if Assumption~\ref{ass:identify_f} is
	satisfied.  In particular, for any function $\bar{f} \in\FF$ and any
	$c > 0$, there exists a model $\tilde{M} \in \MM$, with
	$\P_{\tilde{M}} = \P_{M}$, such that
	\begin{equation*}
		\abs[\Big]{\sup_{i\in\cI}\E_{\tilde{M}(i)}\big[(Y-\bar{f}(X))^2\big]
			- \inf_{\fs \in \FF}\,
			\sup_{i\in\cI} \E_{\tilde{M}(i)}\big[(Y-\fs(X))^2\big]} \geq c.
	\end{equation*}
\end{proposition}
The above impossibility result is visualized in Figure~\ref{fig:impossibility} (left).

\subsection{Generalization to Interventions on $A$} \label{sec:int_onA}

We will see that,
for interventions on $A$, 
parts of the analysis simplify.
Since $A$ influences the system only via 
the covariates $X$, any such intervention
may, in terms of its effect on $(X,Y)$, be 
equivalently expressed as an intervention on $X$
in which
the structural assignment
of $X$ is altered in a way that depends on 
the functional relationship $g$ between $X$ and $A$. 
We can therefore employ several of the results from Section~\ref{sec:gen_int_on_X} by 
imposing an additional assumption on the identifiability of $g$.

\begin{assumption}[Identifiability of $g$] \label{ass:identify_g}
	For all $\tilde{M} = (\tilde{f}, \tilde{g}, \tilde{h}_1, \tilde{h}_2, \tilde{Q}) \in \MM$ 
	with $\P_{\tilde{M}} = \P_M$, it holds that $\tilde{g}(a) = g(a)$ for all $a \in \supp(A) \cup \supp_\cI(A)$. 
\end{assumption}
Since $g(A)$ is a conditional mean for $X$ given $A$, the values of
$g$ are identified from $\P_M$ for $\P_M$-almost all $a$. If
$\supp_\cI(A) \subset \supp(A)$, Assumption~\ref{ass:identify_g} 
therefore holds if, for example, 
$\GG$ contains continuous functions only. 
The
pointwise identifiability of $g$ is necessary, for example, if some of the test
distributions are induced by 
hard interventions on $A$, which
set $A$ to some fixed value $a \in \R^r$. In the case where the
interventions $\cI$ extend the support of $A$, we additionally require
the function class $\GG$ to extrapolate from $\supp(A)$ to
$\supp(A) \cup \supp_\cI(A)$; this is similar to the
conditions on $\FF$ which we made in
Section~\ref{sec:support_extending_onX}
and requires further restrictions on $\GG$.
Under
Assumption~\ref{ass:identify_g}, we obtain a result corresponding to
Propositions~\ref{prop:genX_intra}~and~\ref{prop:genX_extra}.
\begin{proposition}[Generalization to interventions on $A$]
	\label{prop:genA}
	Let $\cI$ be a set of interventions on $A$ and assume
	Assumption~\ref{ass:identify_g} is satisfied. Then, $(\P_{M}, \cM)$
	admits generalization to $\cI$ if either $\supp_{\cI}(X)\subseteq \supp(X)$
	and Assumption~\ref{ass:identify_f} is satisfied or if both
	Assumptions~\ref{ass:identify_f}~and~\ref{ass:gen_f} are satisfied.
\end{proposition}
As becomes clear from the proof of this proposition, in general, the causal function does not need to be a minimax solution.
Further,
Assumption~\ref{ass:identify_f} is not necessary for
generalization. 
In the case where $\FF$, $\GG$, $\HH_1$ and $\HH_2$ consist of linear functions,
 \cite{rothenhausler2018anchor} (anchor regression) and  \cite{jakobsen2020distributional} (K-class estimators)
consider certain sets of interventions on $A$
which render minimax solutions identifiable (and estimate them consistently)
even if 
Assumption~\ref{ass:identify_f} does not hold. 
Similarly, if for a
categorical A, we have $\supp_\cI(A) \subseteq \supp(A)$, it is possible to drop Assumption~\ref{ass:identify_f}.

\subsubsection{Impossibility of Generalization Without Constraining $\mathcal{G}$}
\begin{figure}
	\centering
	\includegraphics[width=\linewidth]{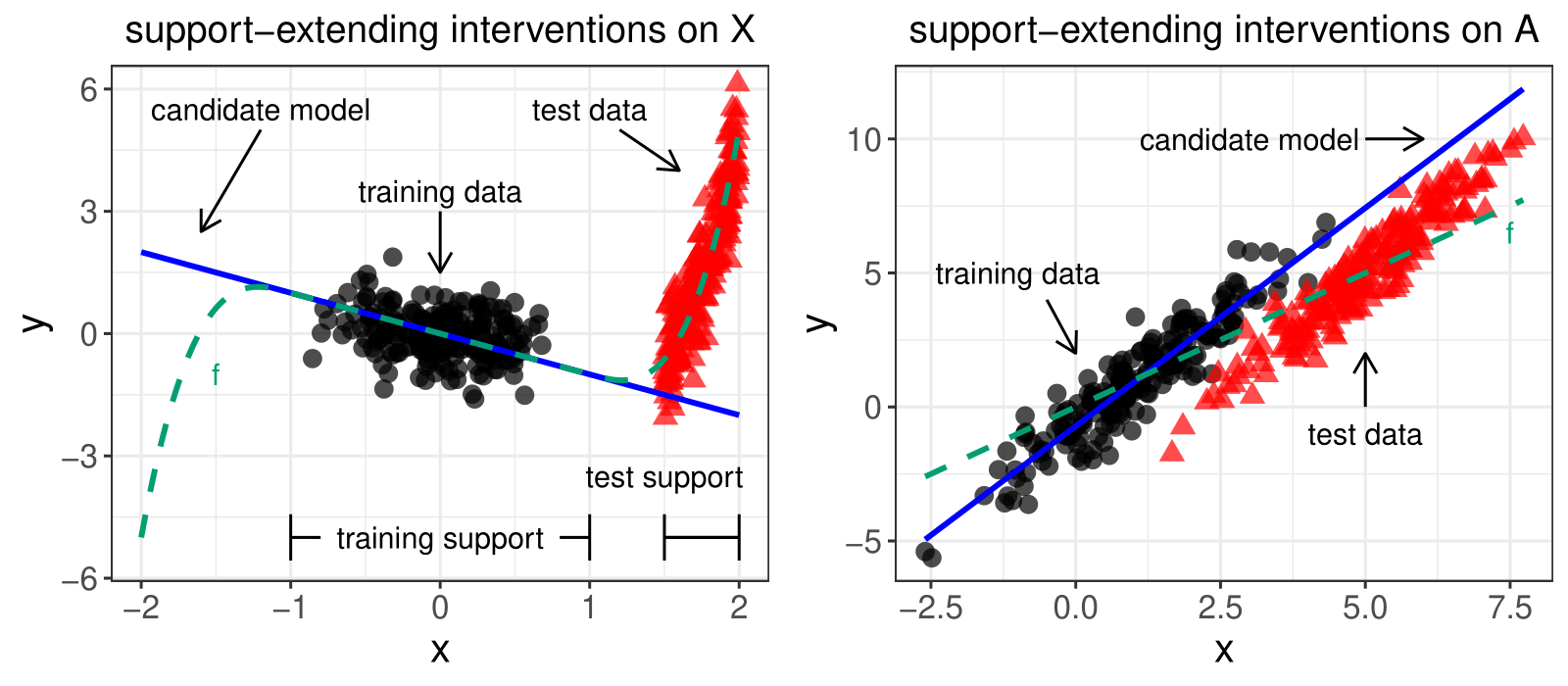} 
	\caption{
		Plots illustrating the straight-forward idea behind 
		the impossibility results in Proposition~\ref{prop:impossibility_extrapolation} (left)
		and Proposition~\ref{prop:impossibility_intA} (right).
		Both plots visualize the case of univariate variables.
		Under well-behaved interventions on $X$ (left; here using confounding-removing interventions)
		which extend the support of $X$, generalization 
		is impossible without further restrictions on the function class $\FF$. 
		This holds true even if Assumption~\ref{ass:identify_f} is
		satisfied. Indeed, although the candidate model (blue line) coincides
		with the causal model (green dashed curve) on the support of $X$, it
		may perform arbitrarily bad on test data generated under
		support-extending interventions. Under interventions on $A$ 
		(right)
		generalization is impossible even under strong assumptions on the
		function class $\FF$ (here, $\FF$ is the class of all linear
		functions). Any support-extending intervention on $A$ shifts the
		marginal distribution of $X$ by an amount which depends on the
		(unknown) function $g$, resulting in a distribution of $(X,Y)$ which,
		in general,
		cannot be identified from the observational distribution. Without
		further restrictions on the function class $\GG$, any candidate model
		apart from the causal model may result in arbitrarily large worst-case
		risk.
	}
	\label{fig:impossibility}
\end{figure}

Without restrictions on the model class
$\GG$, generalization to interventions on $A$ is impossible.
This holds true even under strong assumptions on the true
causal function (such as $f$ is known to be linear).  Below, we give a
formal impossibility result for hard interventions on $A$, which set $A$ to some fixed value,
and
where $\GG$ is the set of all
continuous functions.
\begin{proposition}[Impossibility of generalization to interventions on $A$] \label{prop:impossibility_intA}
	Assume that $\FF = \{\fs: \R^d \to \R \given \fs \text{ is linear} \}$ and
	$\GG = \{\gs : \R^r \to \R^d \given \gs \text{ is continuous} \}$.
	Let $\mathcal{A} \subset \R^r$ be bounded, and let $\cI$ denote the
	set of all hard interventions which set $A$ to some fixed value from
	$\mathcal{A}$.  Assume that $\mathcal{A} \setminus \supp(A)$ has
	nonempty interior. Assume further that $\E_M[\xi_X \xi_Y] \neq 0$
	(this excludes the case of no hidden confounding).  Then, $(\P_M, \MM)$
	does not admit generalization to $\cI$. 
	In addition,
	any
	function other than $f$ may perform arbitrarily bad under the
	interventions in $\cI$.  That is, for any $\bar{f} \neq f$ and $c > 0$, 
	there exists a model $\tilde{M} \in \MM$ with
	$\P_{\tilde{M}} = \P_M$ such that
	\begin{equation*}
		\abs[\Big]{\sup_{i\in\cI}\E_{\tilde{M}(i)}\big[(Y-\bar{f}(X))^2\big] - \inf_{\fs \in \FF}\,
			\sup_{i\in\cI} \E_{\tilde{M}(i)}\big[(Y-\fs(X))^2\big]} \geq c.
	\end{equation*}
\end{proposition}
The above impossibility result is visualized in Figure~\ref{fig:impossibility} (right).
This proposition 
is part of the argument 
showing that anchor regression
\citep{rothenhausler2018anchor} can be extended to nonlinear
settings  only under strong assumptions;
the setting of a linear class $\mathcal{G}$ and a potentially nonlinear class $\mathcal{F}$ is covered in Section~\ref{sec:support_extending_onX}, 
by rewriting interventions on $A$ as interventions on $X$.

An impossibility result similar to the 
proposition above can be shown if
$A$ is categorical. As long as not all categories have been observed
during training it is possible that the intervention which sets $A$
to a previously unseen category can result in a support-extending
distribution shift on $X$. Using
Proposition~\ref{prop:impossibility_extrapolation}, it therefore
follows that generalization can become impossible. 
Since a categorical $A$ can encode 
settings of multi-task learning 
and domain generalization (see Section~\ref{sec:intro_focus}),
this result then 
complements well-known impossibility results for these problems, even under the covariate shift assumption \citep[e.g.,][]{Ben-David2010}.

\section{Learning Generalizing Models from Data} \label{sec:learning}
So far, our focus has been on the possibility to generalize, that is,
we have investigated under which conditions it is possible to identify
generalizing models from the observational distribution. In practice,
generalizing models need to be estimated from finitely many data.
This task is challenging for several reasons. First, analytical
solutions to the minimax problem \eqref{eq:minimax_problem} are only
known in few cases.  Even if generalization is possible, the
inferential target thus often remains a complicated object, given as a
well-defined but unknown function of the observational
distribution. Second, we have seen that the ability to generalize
depends strongly on whether the interventions extend the support of
$X$, see Propositions~\ref{prop:genX_extra} and
\ref{prop:impossibility_extrapolation}.  In a setting with a finite
amount of data, the empirical support of the data lies within some
bounded region, and suitable constraints on the function class $\FF$
are necessary when aiming to achieve empirical generalization outside
this region, even if $X$ comes from a distribution with full
support. 
As we show in our simulations in  Section~\ref{sec:experiments} (see figures), constraining the function class can  also improve the prediction performance at the boundary of the  support.

In Section~\ref{sec:existingmethods}, we survey existing methods for learning
generalizing models. Often, these methods assume either a globally linear model class $\FF$
or are completely non-parametric and therefore do not generalize outside the empirical 
support of the data. Motivated by this observation, we introduce in Section~\ref{sec:nile}
a novel estimator, which exploits an instrumental variable setup and a particular extrapolation 
assumption to learn a globally generalizing model.

\subsection{Existing Methods}
\label{sec:existingmethods}
As discussed in Section~\ref{sec:intro}, a wide range of
methods have been proposed to guard against various types of
distributional changes. Here, we review methods that fit into
the causal framework in the sense that the distributions 
that in the minimax formulation the supremum is taken over %
are induced by interventions.
\begin{table*}
	\centering
	\scriptsize
	\renewcommand{\arraystretch}{1.35}
	\begin{tabular}{>{\centering\arraybackslash}p{1.75cm}>{\centering\arraybackslash}p{3cm}>{\centering\arraybackslash}p{1.5cm}>{\centering\arraybackslash}p{1.75cm}|>{\centering\arraybackslash}p{3.5cm}}
		\toprule
		model class & interventions & $\supp_{\cI}(X)$ & assumptions & algorithm  \\ \toprule
		$\mathcal{F}$ linear & on $X$ or $A$\newline of which at least one is
		confounding-removing & -- &
		Ass.~\ref{ass:identify_f}
		& linear
		IV\newline
		(e.g., two-stage
		least
		squares,
		K-class or
		PULSE
		\cite{theil1958economic,jakobsen2020distributional})\\\midrule
		$\FF, \GG$ linear & on $A$ & bounded  strength
		& -- & anchor regression
		\cite{rothenhausler2018anchor} and \newline K-class \cite{jakobsen2020distributional}\\\midrule
		$\mathcal{F}$ smooth & on $X$ or $A$\newline  of which at least one
		is confounding-removing & support-reducing &
		Ass.~\ref{ass:identify_f}
		&
		nonlinear
		IV\newline %
		(e.g.,
		NPREGIV
		\cite{NPREGIV-CRAN}, 
		Deep IV \citep{hartford2017deep},
		Sieve IV \citep{newey2003instrumental, chen2018optimal},
		Kernel IV \citep{singh2019kernel})\\\midrule
		$\mathcal{F}$ smooth \newline and linearly extrapolates & on
		$X$ or $A$\newline of which at least
		one is confounding-removing & -- &
		Ass.~\ref{ass:identify_f}
		& \textbf{NILE}\newline
		(Section~\ref{sec:nile})\\\bottomrule
	\end{tabular}
	\caption{List of algorithms to learn the generalizing function from
		data,  the considered model class,
		types of interventions, support under interventions,
		and additional
		model assumptions. Sufficient conditions for Assumption~\ref{ass:identify_f}
		are given, for example, in the IV literature by generalized rank
		conditions, see
		Appendix~\ref{sec:IVconditions}.
	}
	\label{tab:learnability}
\end{table*}

For well-behaved interventions on $X$ which contain at least one 
confounding-removing intervention, estimating minimax solutions
reduces to the well-studied problem of estimating causal relationships.
One class of algorithms for this task 
is given by linear instrumental variable
(IV) approaches. They assume that $\mathcal{F}$ is linear and require
identifiability of the causal function
(Assumption~\ref{ass:identify_f}) via a rank condition on the
observational distribution, see
Appendix~\ref{sec:IVconditions}. 
Their target of inference is to
estimate the causal function, which by
Proposition~\ref{prop:minimax_equal_causal} will coincide with the
minimax solution if the set $\cI$ consists of well-behaved
interventions with at least one of them being confounding-removing. 
A basic estimator for
linear IV models is the two-stage least squares (TSLS) estimator, 
which minimizes the norm of the prediction residuals
projected onto the subspace spanned by the observed instruments (TSLS objective).
TSLS
estimators are consistent but do not come  with strong finite sample guarantees; e.g., they do
not have finite moments in a just-identified setup  
\cite[e.g.,][]{mariano2001simultaneous}.
K-class estimators
\citep{theil1958economic} have been proposed to overcome some of these issues.
They minimize a linear combination of the residual sum of squares (OLS objective) and
the TSLS objective. 
K-class estimators can be seen as utilizing a bias-variance trade-off. 
For fixed and non-trivial relative weights, 
they have,
in a Gaussian setting,  finite moments up to a certain order that depends on the sample-size and the number of predictors used.
If the weights are such that 
the OLS objective is ignored asymptotically, 
they consistently estimate the causal parameter \citep[e.g.,][]{mariano2001simultaneous}.
More recently, PULSE has been proposed
\citep{jakobsen2020distributional}, a 
data-driven procedure for choosing the
relative weights 
such that the prediction residuals `just' pass a test
for simultaneous uncorrelatedness with the instruments.

In cases where the minimax solution does not coincide with the
causal function, only few algorithms exist. 
Anchor regression \citep{rothenhausler2018anchor} is a procedure that
can be used when $\mathcal{F}$ and $\mathcal{G}$ are linear and $h_1$
is additive in the noise component. It finds
the minimax solution if the set $\cI$ consists of all 
interventions on $A$ up to a fixed intervention strength,
and is applicable even if
Assumption~\ref{ass:identify_f} is not necessarily satisfied.

In a linear setting, where the
regression coefficients differ between different environments, it is
also possible to minimize the worst-case risk among the observed
environments \citep{meinshausen2015maximin}.  In its current
formulation, this approach does not quite fit into the above
framework, as it does not allow for changing distributions of
the covariates.
A summary of the
mentioned methods and their assumptions
is given in Table~\ref{tab:learnability}.

If $\mathcal{F}$ is a nonlinear or non-parametric class of functions, the task of finding 
minimax solutions becomes more difficult. In cases where the causal function is among 
such solutions, this problem has been studied in the econometrics community. 
For example, \cite{newey2013nonparametric,newey2003instrumental} treat the 
identifiability and estimation of causal functions in non-parametric function classes.
Several non-parametric IV procedures exists, e.g., NPREGIV \citep{NPREGIV-CRAN}
contains modified implementations of \cite{horowitz2011applied} and \cite{darolles2011nonparametric}, 
which we will refer to as NPREGIV-1 and NPREGIV-2, respectively. 
Other procedures include 
Deep IV \citep{hartford2017deep},
Sieve IV \citep{newey2003instrumental, chen2018optimal} and
Kernel IV \citep{singh2019kernel}. 
Identifiability and estimation of the causal function using nonlinear IV methods in parametric function classes is discussed in Appendix~\ref{sec:IVconditions}.
Unlike in the linear case, most of the
methods do not aim to extrapolate 
and only recover the causal function inside the support of $X$,
that is, they cannot be used to predict interventions
outside of 
this domain.
In the following section, we propose a procedure that is able to
extrapolate when $\mathcal{F}$ consists of functions which
extend linearly outside of the support of $X$. 
In principle, any other extrapolation rule may be employed here, 
as long as all functions from $\FF$ are uniquely determined by their values
on the support of $X$, that is, Assumption~\ref{ass:gen_f} is satisfied.

In our simulations, we see that our method can improve the prediction performance on the boundary of the support and outperforms other methods when comparing the estimation on the support. 

\subsection{NILE} \label{sec:nile}
We have seen in Proposition~\ref{prop:impossibility_extrapolation}
that in order to generalize to interventions which extend the support
of $X$, we require additional assumptions on the function class $\FF$.
In this section, we start from such assumptions and verify both
theoretically and practically that they allows us to perform
distribution generalization in the considered setup.  Along the way,
several choices can be made and usually several options are
possible.  We will see that our choices yield a method with
competitive performance, but we do not claim optimality of our
procedure.  Several of our choices were partially 
made to keep the theoretical exposition simple and the method
computationally efficient.  
We first consider the
univariate case (i.e., $X$ and $A$ are real-valued) and comment
later on the possibility to extend the methodology to higher
dimensions.  
Unless specific
background knowledge is given, it might be reasonable to assume that
the causal function extends linearly outside a fixed interval $[a,b]$. 
By
additionally imposing differentiability on $\FF$, any function from
$\FF$ is uniquely defined by its values within $[a,b]$, see also
Section~\ref{sec:support_extending_onX}.  Given an estimate $f$ on $[a,b]$,
the linear extrapolation property then yields a global estimate
on the whole of $\mathbb{R}$.
In
principle, any class of differentiable functions can be used. 
Here, 
we assume that, on the interval $[a,b]$, the causal function $f$
is contained in the linear span of a B-spline basis. 
More formally, let $B = (B_1, ..., B_k)$ be a fixed B-spline basis on $[a,b]$, and define $\eta := (a,b,B)$. 
Our procedure assumes that the true causal function $f$ belongs to the function class 
$\FF_{\eta} := \{f_\eta(\cdot; \theta) \st \theta \in \R^k\}$, 
where for every $x \in \R$ and $\theta \in \R^k$, $f_\eta(x; \theta)$ is given as
\begin{equation} \label{eq:f_theta}
	f_{\eta}(x;\theta) :=
	\begin{cases}
		B(a)^\top  \theta +  B^\prime (a)^\top  \theta (x - a) & \text{ if } x < a \\
		B(x)^\top  \theta & \text{ if } x \in [a, b] \\
		B(b)^\top  \theta + B^\prime (b)^\top \theta (x - b) & \text{ if } x >b,\\
	\end{cases}
\end{equation}
where 
$B^\prime := (B_1^\prime, \dots, B_k^\prime)$
denotes
the component-wise derivative of $B$.
In our algorithm, $\eta = (a, b, B)$ is a hyper-parameter, which can be set manually, 
or be chosen from data.

\subsubsection{Estimation Procedure} \label{sec:estimation}
We now introduce our estimation procedure for fixed choices of all
hyper-parameters.
Section~\ref{sec:algorithm} describes how these can
be chosen from data in practice.
Let $(\B{X}, \B{Y}, \B{A}) \in \R^{n \times 3}$ be 
$n$ i.i.d.\ realizations sampled from 
a distribution over
$(X,Y,A)$,
let $\eta = (a,b,B)$ be fixed and assume that $\supp(X)\subset [a,b]$. 
Our algorithm aims to learn the causal function $f_\eta(\cdot ; \theta^0) \in \FF_\eta$,
which is determined by
the linear causal parameter $\theta^0$ of 
a $k$-dimensional vector of covariates $(B_1(X), \dots, B_k(X))$. From standard linear IV theory, 
it is known that at least $k$ instrumental variables are required to identify the $k$ causal parameters, see Appendix~\ref{sec:IVconditions}. 
We therefore artificially generate such instruments by nonlinearly transforming $A$, by using another
B-spline basis $C = (C_1, \dots, C_k)$. 
The parameter $\theta^0$ 
can then be identified from the observational distribution under
appropriate rank conditions, see
Section~\ref{sec:consistency}. 
In that case, the hypothesis
$H_0(\theta) : \theta= \theta^0$ is equivalent to the hypothesis
$\tilde H_0(\theta) : \E[C(A)(Y - B(X)^\top \theta)] =
0$.
Let
$\B{B} \in \R^{n \times k}$ and $\B{C} \in \R^{n \times k}$ be the
associated design matrices, for each $i \in \{1, \dots, n\}$,
$j \in \{1, \dots, k\}$ given as $\B{B}_{ij} = B_j(X_i)$ and
$\B{C}_{ij} = C_{j}(A_i)$. 
A straightforward choice would be to construct the standard TSLS estimator, i.e.,  
$\hat \theta$ as the minimizer of
$\theta \mapsto \norm{\B{P} (\B{Y} - \B{B} \theta)}_2^2$, where
$\B{P}$ is the projection matrix onto the columns of $\B{C}$;
see also \cite{hall2005generalized}.  Even though this procedure may result in an
asymptotically consistent estimator, there are several reasons why it
may 
be suboptimal
in a finite
sample setting.  First, the above estimator can have large finite
sample bias, in particular if $k$ is large. Indeed, in the extreme
case where $k = n$, and assuming that all columns in $\B{C}$ are
linearly independent, $\B{P}$ is equal to the identity matrix, and
$\hat \theta$ coincides with the OLS estimator.  Second, since
$\theta$ corresponds to the linear parameter of a spline basis, it
seems reasonable to impose constraints on $\theta$ which enforce
smoothness of the resulting spline function. Both of these points can
be addressed by introducing additional penalties into the estimation
procedure.  Let therefore
$\B{K} \in \R^{k \times k}$ 
and $\B{M} \in \R^{k \times k}$ be the matrices that are, for each $i,j \in \{1, \dots, k\}$, 
defined as $\B{K}_{ij} = \int B^{\prime \prime}_i(x) B^{\prime \prime}_j(x) dx$ and 
$\B{M}_{i j} = \int C^{\prime \prime}_{i}(a) C^{\prime \prime}_{j}(a) da$, 
and let $\gamma, \delta > 0$ be the respective penalties associated with $\B{K}$ and $\B{M}$. 
For $\lambda \geq 0$ and with $\mu := (\gamma, \delta, C)$, we then define the estimator 
\begin{equation} \label{eq:thetahat}
	\hat \theta^n_{\lambda, \eta, \mu}
	:= \argmin_{\theta \in \R^{k}} 
	\norm{\B{Y} - \B{B} \theta }_2^2 + \lambda \norm{\B{P}_\delta(\B{Y} - \B{B} \theta)}_2^2 + \gamma \theta^\top \B{K} \theta, \\
\end{equation}
where $\B{P}_\delta := \B{C} (\B{C}^\top \B{C} + \delta \B{M})^{-1} \B{C}^\top$ is the `hat'-matrix 
for a 
penalized regression onto the columns of $\B{C}$. 
By choice of $\B{K}$, the term $\theta^\top \B{K} \theta$ is equal to the integrated squared curvature of the spline
function parametrized by $\theta$. 
The regularization induced by the second summand in~\eqref{eq:thetahat} 
is similar to the one from K-class estimators in linear settings \citep{theil1958economic}.
The function class~\eqref{eq:f_theta}
enforces 
linear extrapolation.
In principle, the above approach extends to situations where $X$ and $A$ are higher-dimensional,
in which case
$B$ and $C$ consist of multivariate functions. 
For example, \cite{fahrmeir2013regression} propose the use of tensor product splines, 
and introduce multivariate smoothness penalties based on pairwise first- or second order
parameter differences of basis functions which are close-by with respect to some suitably chosen metric. 
Similarly to \eqref{eq:thetahat}, such penalties result in a convex optimization problem. 
However, due to the large number of involved variables, the optimization procedure 
becomes computationally burdensome already in small dimensions.

Within the function class $\FF_\eta$, the above defines the global
estimate $f_\eta(x; \hat \theta^n_{\lambda, \eta, \mu})$, 
for every $x \in \R$, given by
\begin{equation} \label{eq:fhat_theta}
	f_\eta(x; \hat \theta^n_{\lambda, \eta, \mu}) :=
	\begin{cases}
		B(a)^\top \hat \theta^n_{\lambda, \eta, \mu} +  B^\prime (a)^\top \hat \theta^n_{\lambda, \eta, \mu} (x - a) & \text{ if } x < a \\
		B(x)^\top \hat \theta^n_{\lambda, \eta, \mu} & \text{ if } x \in [a,b] \\
		B(b)^\top \hat \theta^n_{\lambda, \eta, \mu} + B^\prime (b)^\top \hat \theta^n_{\lambda, \eta, \mu} (x - b) & \text{ if } x > b. \\
	\end{cases}
\end{equation}
We deliberately distinguish between three different groups of
hyper-parameters $\eta$, $\mu$ and $\lambda$.  The parameter
$\eta = (a,b,B)$ defines the function class to which the causal
function $f$ is assumed to belong. To prove consistency of our
estimator, we require this function class to be correctly specified.
In turn, the parameters $\lambda$ and
$\mu=(\gamma, \delta, C)$ are algorithmic parameters that do not
describe the statistical model. Their values only affects the finite
sample behavior of our algorithm, whereas consistency is ensured as
long as $C$ satisfies certain rank conditions, see Assumption~\ref{ass:RankCondition}
in Section~\ref{sec:consistency}.
In practice,
$\gamma$ and $\delta$ are chosen via a cross-validation procedure, see
Section~\ref{sec:algorithm}. The parameter $\lambda$ determines the
relative contribution of the OLS and TSLS losses to the objective
function.
To choose $\lambda$ from data, we use an idea 
similar to the PULSE \citep{jakobsen2020distributional}.

\subsubsection{Algorithm} \label{sec:algorithm}
Let for now $\eta, \mu$ be fixed. 
In the limit $\lambda \to \infty$, our estimation procedure becomes
equivalent to minimizing the
TSLS loss $\theta \mapsto \norm{\B{P}_\delta (\B{Y} - \B{B}\theta)}_2^2$, 
which may be interpreted as searching for the 
parameter $\theta$ which complies `best' with the hypothesis $\tilde{H}_0(\theta) : \E[C(A)(Y - B(X)^\top \theta)] = 0$. 
For finitely many data,
following the idea introduced in \citep{jakobsen2020distributional}, 
we propose to choose the value for $\lambda$ such that $\tilde{H}_0(\hat \theta^n_{\lambda, \eta, \mu})$ is 
just accepted (e.g., at a significance level $\alpha = 0.05$). 
That is, 
among all $\lambda \geq 0$ which result in an estimator that is not rejected as a 
candidate for the causal parameter, we chose the one which yields maximal contribution of the OLS 
loss to the objective function. 
More formally, let 
for every $\theta \in \R^k$, $T(\theta) = (T_n(\theta))_{n \in \N}$ be
a statistical test 
at (asymptotic) level $\alpha$ for $\tilde{H}_0(\theta)$ with rejection 
threshold $q(\alpha)$. That is, $T_n(\theta)$ does not reject
$\tilde{H}_0(\theta)$ if and only if $T_n(\theta) \leq q(\alpha)$. The penalty
$\lambda^\star_n$ is then chosen in the following data-driven way
\begin{align*}
	\lambda^\star_n := \inf \{\lambda \geq 0 : T_n(\hat \theta^n_{\lambda, \eta, \mu})\leq q(\alpha)\}.
\end{align*}
In general, $\lambda^\star_n$ is not guaranteed to be finite for an arbitrary test statistic $T_n$. 
Even for a reasonable test statistic it might happen 
that $T_n(\hat \theta^n_{\lambda, \eta, \mu} ) > q(\alpha)$ 
for all $\lambda \geq 0$; see \cite{jakobsen2020distributional} for further details. 
We can remedy the problem by reverting 
to another well-defined and consistent estimator, such as
the TSLS (which minimizes the TSLS loss above) 
if $\lambda^\star_n$ 
is not finite.  
Furthermore, if
$\lambda \mapsto T_n(\hat \theta^n_{\lambda, \eta, \mu})$ is
monotonic, $\lambda^\star_n$ can be computed efficiently by a binary
search procedure.  In our algorithm, the test statistic $T$
and rejection threshold $q$ can be supplied by the user. 
Conditions on $T$ that are sufficient to yield a consistent estimator $f_\eta(\cdot , \hat \theta_{\lambda_n^\star, \mu, \eta})$, 
given that $\mathcal{F}_\eta$ is correctly specified,
are presented in Section~\ref{sec:consistency}.
Two choices of test statistics which are implemented in our code package
can be found in Appendix~\ref{sec:test_statistic}.

For every $\gamma \geq 0$, let $\B{Q}_\gamma = \B{B} (\B{B}^\top \B{B} + \gamma \B{K})^{-1} \B{B}^\top$ be the `hat'-matrix 
for the penalized regression onto $\B{B}$. Our algorithm then proceeds as follows. \\
\begin{algorithm}
	\caption{NILE (``Nonlinear Intervention-robust Linear Extrapolator'')} \label{alg:nile}
\begin{algorithmic}[1]
	\State \textbf{input}: data $(\B{X}, \B{Y}, \B{A}) \in \R^{n \times 3}$\;
	\State \textbf{options}: $k$,  $T$, $q$, $\alpha$\; %
	\Begin
	\State $a \leftarrow \min_i X_i$, $b \leftarrow \max_i X_i$\; 
	\State construct cubic B-spline bases $B = (B_1, \dots, B_{k})$ and $C = (C_1, \dots, C_k)$ at equidistant knots, 
	with boundary knots at respective extreme values of $\B{X}$ and $\B{A}$\;
	\State define $\hat \eta \leftarrow (a,b,B)$\;
	\State choose $\delta^n_{\text{CV}}>0$ by 10-fold CV to minimize the out-of-sample mean squared error of $\hat{\B{Y}} = \B{P}_{\delta} \B{Y}$\;
	\State choose $\gamma^n_{\text{CV}}>0$ by 10-fold CV to minimize the out-of-sample mean squared error of $\hat{\B{Y}} = \B{Q}_{\gamma} \B{Y}$\;
	\State define $\mu^n_{\text{CV}} \leftarrow (\delta^n_{\text{CV}}, \gamma^n_{\text{CV}}, C)$\;
	\State approx. $\lambda^\star_n = \inf\{\lambda \geq 0 : T_n(\hat \theta^n_{\lambda, \mu^n_{\text{CV}}, \hat \eta}) \leq q(\alpha)\}$ by binary search\;
	\State update $\gamma^n_{\text{CV}} \leftarrow (1 + \lambda^\star_n) \cdot \gamma^n_{\text{CV}}$\;
	\State compute $\hat \theta^n_{\lambda_n^\star, \mu^n_{\text{CV}}, \hat \eta}$ using \Cref{eq:thetahat}\;
	\End
	\State \textbf{output}: $\hat{f}^n_{\text{NILE}}  := f_{\hat \eta}(\, \cdot \, ; \hat \theta^n_{\lambda_n^\star, \mu^n_{\text{CV}}, \hat \eta})$ defined by \Cref{eq:fhat_theta}\;
\end{algorithmic}
\end{algorithm}

The penalty parameter $\gamma^n_{\text{CV}}$ is chosen to minimize the
out-of-sample mean squared error
of the prediction model
$\hat{\B{Y}} = \B{Q}_{\gamma} \B{Y}$, which corresponds to the
solution of \eqref{eq:thetahat} for $\lambda = 0$.  
After choosing $\lambda_n^\star$, the objective function in \eqref{eq:thetahat} increases 
by the term $\lambda_n^\star \norm{\B{P}_{\delta_{\text{CV}}^n}(\B{Y} - \B{B} \theta)}_2^2$.
In order for the penalty term $\gamma \theta^\top \B{K} \theta$ to impose the same degree of
smoothness in the altered optimization problem, the penalty parameter $\gamma$ needs to be 
adjusted accordingly. 
The heuristic update in our algorithm is motivated by the simple observation that for all $\delta, \lambda \geq 0$,
$\norm{\B{Y} - \B{B} \theta}_2^2 + \lambda \norm{\B{P}_\delta (\B{Y} - \B{B} \theta)}_2^2 \leq (1 + \lambda) \norm{\B{Y} - \B{B} \theta}_2^2$.

\subsubsection{Asymptotic Generalization (consistency)} \label{sec:consistency}
We now prove consistency of our estimator in the case where
the hyper-parameters $(\eta, \mu)$ are fixed (rather than data-driven), and the
function class $\FF_{\eta}$ is correctly specified.
Fix any $a<b$
and a basis $B=(B_1, \dots, B_k)$. Let $\eta_0 = (a,b,B)$ and let the
model class be given by
$\MM = \FF_{\eta_0} \times \GG \times \HH_1 \times \HH_2 \times
\mathcal{Q}$, where $\FF_{\eta_0}$ is as described in
Section~\ref{sec:nile}. Assume that the data-generating model
$M = ( f_{\eta_0}(\, \cdot \,; \theta^0),g,h_1,h_2,Q) \in \MM$ 
induces
an observational distribution $\P_M$ such that
$\supp^M(X) \subseteq (a,b)$.
Let further $\cI$ be a set of 
interventions on $X$ or $A$,
and let $\alpha \in (0,1)$ be a fixed significance level.

We prove asymptotic generalization (consistency) for an idealized
version of the NILE estimator which utilizes $\eta_0$, rather than the
data-driven values. Choose any $\delta,\gamma \geq 0$ 
and basis
$C=(C_1,...,C_k)$ and let $\mu=(\delta,\gamma,C)$. 
We will make use of
the following assumptions.
\begin{enumerate}[label=(B\arabic*),ref=(B\arabic*)]
	\item 
	For all $\tilde{M}\in \cM$ with $\bP_M = \bP_{\tilde M}$ it holds that
	$\sup_{i\in\cI}\E_{\tilde{M}(i)} [X^2] < \infty$ and \newline
	$\sup_{i\in\cI} \lambda_{\max}(\E_{\tilde{M}(i)} [B(X)B(X)^\top ])<\infty.$ \label{ass:MaximumEigenValueBounded} 
	\item The matrices $\E_M[ B(X)B(X)^\top ]$, $\E_M[ C(A)C(A)^\top ]$ and
	$\E_M[ C(A)B(X)^\top  ]$ have full rank.\label{ass:RankCondition}
\end{enumerate}
\begin{enumerate}[label=(C\arabic*),ref=(C\arabic*)]
	\item  $T(\theta)$ has 
	uniform asymptotic power on any compact set of alternatives. \label{ass:ConsistentTestStatistic}
	\item  $\lambda^\star_n := \inf\{\lambda\geq 0 : T_n(\hat \theta^n _{\lambda,\eta_{0},\mu}) \leq q(\alpha)\}$ is almost surely finite. \label{ass:LambdaStarAlmostSurelyFinite}
	\item 
	$\lambda \mapsto T_n(\hat \theta^n _{\lambda,\eta_0,\mu})$ is weakly decreasing and $\theta \mapsto T_n(\theta)$ is continuous.
	\label{ass:MonotonicityAndContinuityOfTest}
\end{enumerate}
Assumptions \ref{ass:MaximumEigenValueBounded}--\ref{ass:RankCondition} ensure consistency of the estimator as long as $\lambda^\star_n$ tends to infinity. 
Intuitively, in this case, we can apply arguments similar to those
that prove consistency of the TSLS estimator. 
Assumptions \ref{ass:ConsistentTestStatistic}--\ref{ass:MonotonicityAndContinuityOfTest} ensure
that consistency is achieved when choosing $\lambda^\star_n$ in the data-driven fashion described in Section~\ref{sec:algorithm}. 
In Assumption \ref{ass:MaximumEigenValueBounded}, $\lambda_{\max}$ denotes the largest eigenvalue. 
In words, the assumption states that, 
under each model $\tilde M\in \cM$ with $\bP_M = \bP_{\tilde M}$, 
there exists a finite upper bound on the variance of any linear combination of the basis functions $B(X)$, 
uniformly over all distributions induced by $\cI$. 
The first two rank conditions of \ref{ass:RankCondition} 
enable certain limiting arguments to be valid and they guarantee that estimators are asymptotically well-defined. 
The last rank condition of \ref{ass:RankCondition} is the so-called rank condition for identification. 
It guarantees that $\theta^0$ is identified from the observational distribution in the sense 
that the hypothesis $H_0(\theta):\theta=\theta^0$ becomes equivalent with $\tilde{H}_0(\theta) : \E_M[C(A)(Y-B(X)^\top \theta)]=0$. 
\ref{ass:ConsistentTestStatistic} 
means that for any compact set $K\subseteq \R^k$ with $\theta^0 \not \in K$ it holds that $\lim_{n\to \infty} P(\inf_{\theta\in K} T_n(\theta) \leq q(\alpha)) =0$.
If the considered test 
has, in addition, a level guarantee, 
such as pointwise asymptotic level,
the 
interpretation of the finite sample estimator 
discussed in Section~\ref{sec:algorithm}
remains valid 
(such level guarantee may potentially yield improved
finite sample performance, too).
\ref{ass:LambdaStarAlmostSurelyFinite} 
is made to simplify the consistency proof. 
As previously discussed in Section~\ref{sec:algorithm}, if \ref{ass:LambdaStarAlmostSurelyFinite} is not satisfied, 
we can output
another well-defined and consistent estimator on the event $(\lambda^\star_n=\infty)$, ensuring that consistency still holds. 

Under these conditions, we have the following asymptotic generalization guarantee.

\begin{proposition}[Asymptotic generalization]\label{thm:consis}
	Let $\cI$ be a set of 
	interventions on $X$ or $A$ of which at
	least one is confounding-removing.  If
	assumptions~\ref{ass:MaximumEigenValueBounded}--\ref{ass:RankCondition}
	and
	\ref{ass:ConsistentTestStatistic}--\ref{ass:MonotonicityAndContinuityOfTest}
	hold true, then, for any $\tilde{M} \in \MM$ with
	$\P_{\tilde{M}} = \P_M$, and any $\epsilon > 0$, it holds that
{\small	\begin{align*} 
		\P_{M} &\left( \big\vert \sup_{i\in\cI}\E_{\tilde{M}(i)}\big[(Y-f_{\eta_0}(X;\hat{\theta}^n_{\lambda^\star_n,\eta_0,\mu}))^2\big] \right.
		- \left. \inf_{\fs\in\FF_{\eta_0}}\,
		\sup_{i\in\cI}\E_{\tilde{M}(i)}\big[(Y-\fs(X))^2\big] \big\vert \leq \epsilon \right) \to 1,
	\end{align*}}
	as $n\to \infty$. In the above event, only
	$\hat{\theta}^n_{\lambda^\star_n,\eta_0,\mu}$ is stochastic.
\end{proposition}

\subsubsection{Experiments} \label{sec:experiments}
\begin{figure}
	\centering
	\includegraphics[width=\linewidth]{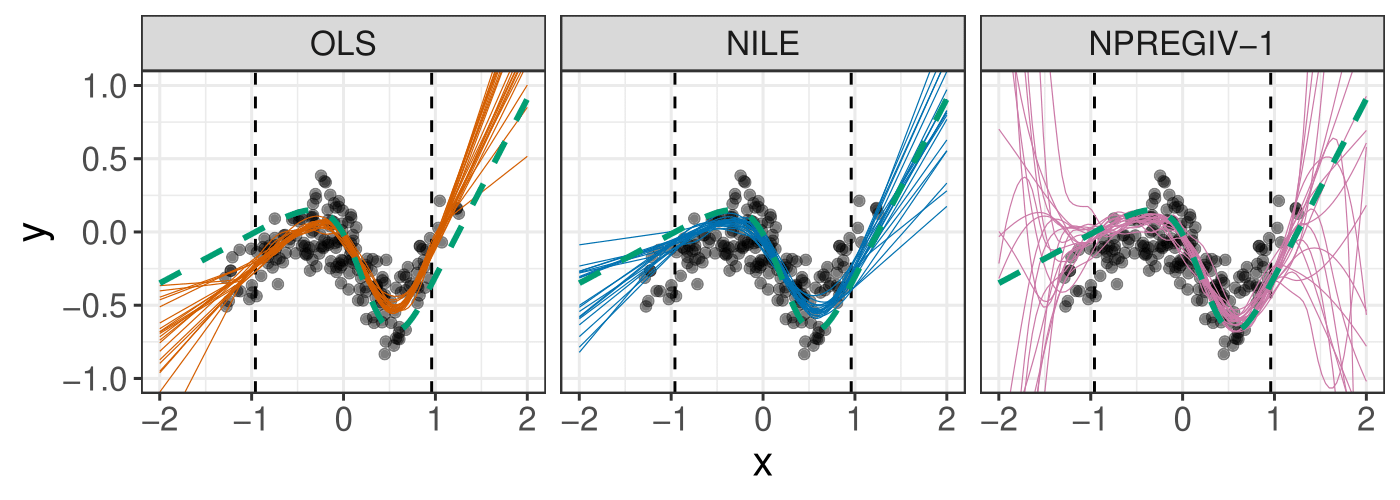}
	\caption{A sample dataset from the model \eqref{eq:sim_model} with 
		$\alpha_A = \sqrt{1/3}$, $\alpha_H = \sqrt{2/3}$, $\alpha_\epsilon = 0$. The true causal function
		is indicated by a green dashed line. For each method, we show 20 estimates of 
		this function, each based on an independent sample from \eqref{eq:sim_model}. 
		For values within the support of the training data (vertical dashed lines 
		mark the inner 90\% quantile range), 
		NPREGIV-1 correctly estimates the causal function well. 
		As expected, when moving
		outside the support of $X$, 
		the estimates become unreliable, 
		and we gain an increasing advantage by exploiting the linear extrapolation assumed by
		the NILE.}
	\label{fig:overlay_estimates_and_varying_confounding}
\end{figure}

We now investigate the empirical performance of our proposed
estimator, the NILE, with $k = 50$ spline basis functions.
To choose $\lambda_n^\star$, we use the test statistic
$T_n^2$, 
which tests the slightly stronger hypothesis $\bar{H}_0$, 
see Appendix~\ref{sec:test_statistic}. In all experiments
use the significance level $\alpha = 0.05$. 
We include two other approaches as baseline: (i) the method NPREGIV-1 
(using its default options)
introduced in Section~\ref{sec:existingmethods}, and (ii) a linearly
extrapolating estimator of the ordinary regression of $Y$ on $X$ (which
corresponds to the NILE with $\lambda^\star \equiv 0$).  In all
experiments, we generate data sets of size $n=200$ as independent
replications from %
\begin{align} \label{eq:sim_model}
	\begin{split}
		A:= \epsilon_A, \quad H:= \epsilon_H, 
		\quad X :=  \alpha_A A + \alpha_H H + \alpha_\epsilon \epsilon_X, \quad
		Y := f (X) + 0.3  H + 0.2  \epsilon_Y,
	\end{split}
\end{align}
where $(\epsilon_A, \epsilon_H, \epsilon_X, \epsilon_Y)$ are jointly 
independent with $\text{Uniform}(-1,1)$ marginals. To make results 
comparable across different parameter settings, we impose the constraint
$\alpha_A^2 + \alpha_H^2  +\alpha_\epsilon^2 = 1$, which ensures that
in all models, $X$ has variance $1/3$. The function $f$ is drawn from 
the linear span of a basis of four natural cubic splines with knots placed equidistantly 
within the $90\%$ inner quantile range of $X$. 
By well-known properties of 
natural splines, any such function extends linearly 
outside the boundary knots. 
Figure~\ref{fig:overlay_estimates_and_varying_confounding} (left) shows an example
data set from \eqref{eq:sim_model}, 
where the causal function is indicated in green. 
We additionally display estimates obtained by each of the considered methods, based
on 20 i.i.d.\ datasets. Due to the confounding variable $H$, the OLS estimator is 
clearly biased. NPREGIV-1 exploits $A$ as an instrumental variable and obtains good results 
within the support of the observed data. Due to its non-parametric nature, however, 
it cannot extrapolate outside this domain. The NILE estimator exploits the linear 
extrapolation assumption on $f$ to produce global estimates. 

\begin{figure}
	\centering
	\includegraphics[width=\linewidth]{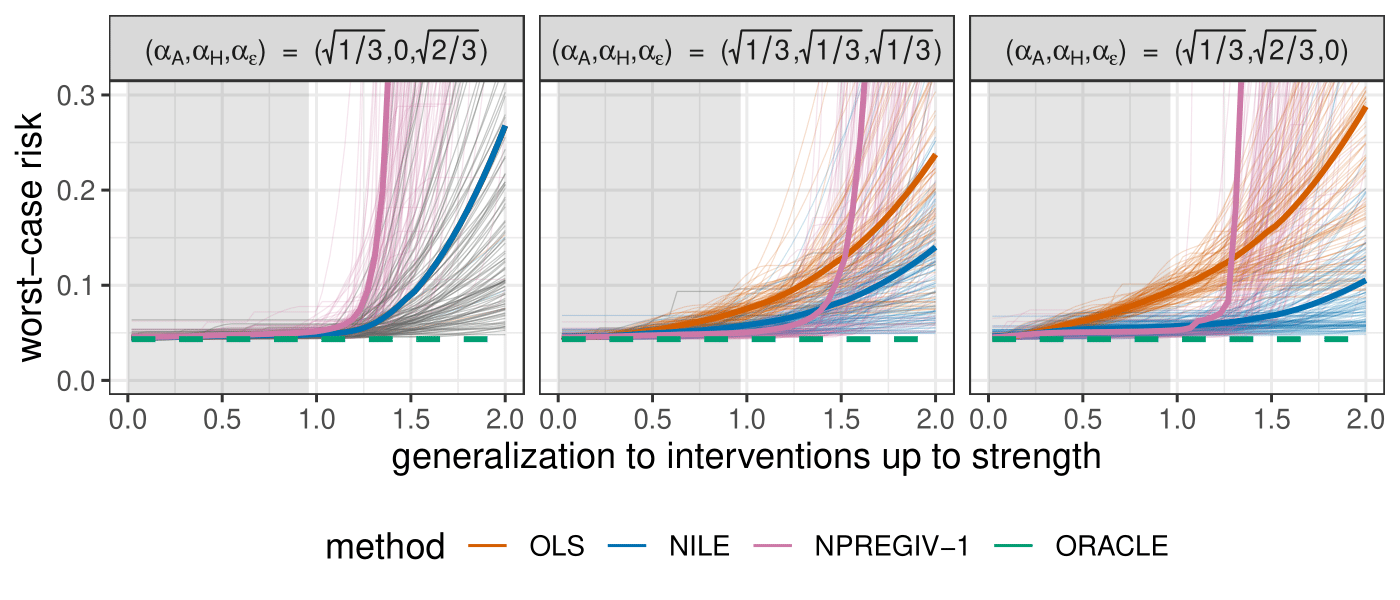}
	\caption{
		Predictive performance under confounding-removing interventions on $X$
		for different confounding- and intervention strengths 
		(see alpha values in the grey panel on top).
		The right panel corresponds to the same parameter setting as in Figure~\ref{fig:overlay_estimates_and_varying_confounding}. 
		The plots in each panel are based on data sets of size $n=200$, 
		generated from $N = 100$ different models of the form \eqref{eq:sim_model}. 
		For each model, we draw a different function $f$, 
		resulting in a different minimax solution (see Appendix~\ref{sec:additional_experiments} for details on the sampling procedure). 
		The performances under individual models are shown by thin lines; the average performance \eqref{eq:experiments_risk} 
		across all models is indicated by thick lines.
		In all considered models, the optimal prediction error (green dashed
		line) is equal to $\E[\xi_Y^2]$ 
		(by consistency, for any fixed
		function $f$, NILE's worst-case risk   
		converges pointwise to this value for increasing sample size).
		The grey area indicates the inner 90 \% quantile range 
		of $X$ in the training distribution; the white area can be seen as an area of generalization.
	}
	\label{fig:overlay_estimates_and_varying_confounding2}
\end{figure}

We further investigate the empirical worst-case risk %
across several different models of the form \eqref{eq:sim_model}. That is, for a fixed set of parameters
$(\alpha_A, \alpha_H, \alpha_\ep)$, we construct several models $M_1, \dots, M_N$ of the form 
\eqref{eq:sim_model} by randomly sampling causal functions $f_1, \dots, f_N$ 
(see Appendix~\ref{sec:additional_experiments} for further details on the sampling procedure). 
For every $x \in [0,2]$, let $\cI_x$ denote the set of hard interventions
which set $X$ to some fixed value in $[-x,x]$. 
We then characterize the performance of each method using the average (across different models) 
worst-case risk (across the interventions in $\cI_x$), i.e., for each estimator $\hat{f}$, we consider 
\begin{align} \label{eq:experiments_risk}
	\begin{split}
		\frac{1}{N} \sum_{j=1}^N \sup_{i \in \cI_x} \E_{M_j(i)} \big[ (Y - \hat{f}(X))^2 \big]
		= \E[\xi_Y^2] + \frac{1}{N} \sum_{j=1}^N \sup_{\tilde{x} \in [-x,x]}  (f_j(\tilde{x}) - \hat{f}(\tilde{x}))^2,
	\end{split}
\end{align}
where $\xi_Y :=  0.3  H + 0.2  \epsilon_Y$ is the noise term for $Y$ (which is fixed across all experiments).
In practice, we evaluate the functions $\hat{f}$, $f_1, \dots, f_N$ on a fine grid on $[-x,x]$ to approximate
the above supremum. 
Figure~\ref{fig:overlay_estimates_and_varying_confounding2}
plots the average worst-case risk versus intervention strength 
for varying degree of confounding ($\alpha_H$).
The optimal worst-case risk $\E[\xi_Y^2]$ is indicated by a green dashed line. 
The results show that the linear extrapolation
property of the NILE estimator is beneficial in particular for strong interventions. 
In the case of no confounding ($\alpha_H = 0$), the minimax solution coincides with the 
regression of $Y$ on $X$, hence even the OLS estimator yields good predictive performance. 
In this case, the hypothesis $\bar{H}_0(\hat \theta^n_{\lambda, \delta^n_{\text{CV}}, \gamma^n_{\text{CV}}})$ 
is accepted already for small values of $\lambda$
(in this experiment, the empirical average of $\lambda^\star_n$ equals 0.015),
and the 
NILE estimator becomes indistinguishable from the OLS. As the confounding strength increases, the OLS 
becomes increasingly biased, and the NILE objective function 
differs more notably
from the OLS 
(average $\lambda^\star_n$ of 2.412 and 5.136, respectively). The method NPREGIV-1 slightly 
outperforms the NILE inside the support of the observed data, but drops in performance 
for stronger interventions. 
We believe that the increase in extrapolation performance of the NILE for stronger 
confounding (increasing $\alpha_H$) 
might stem from
the fact that, as the $\lambda_n^\star$ increases, also the smoothness
penalty $\gamma$ increases, see Algorithm~\ref{alg:nile}. While this results in slightly worse 
in-sample prediction, it seems beneficial for extrapolation (at least for the particular function class 
that we consider). We do not claim that our algorithm has theoretical guarantees which 
explain this increase in performance.

\begin{figure}
	\centering
	\includegraphics[width=\linewidth]{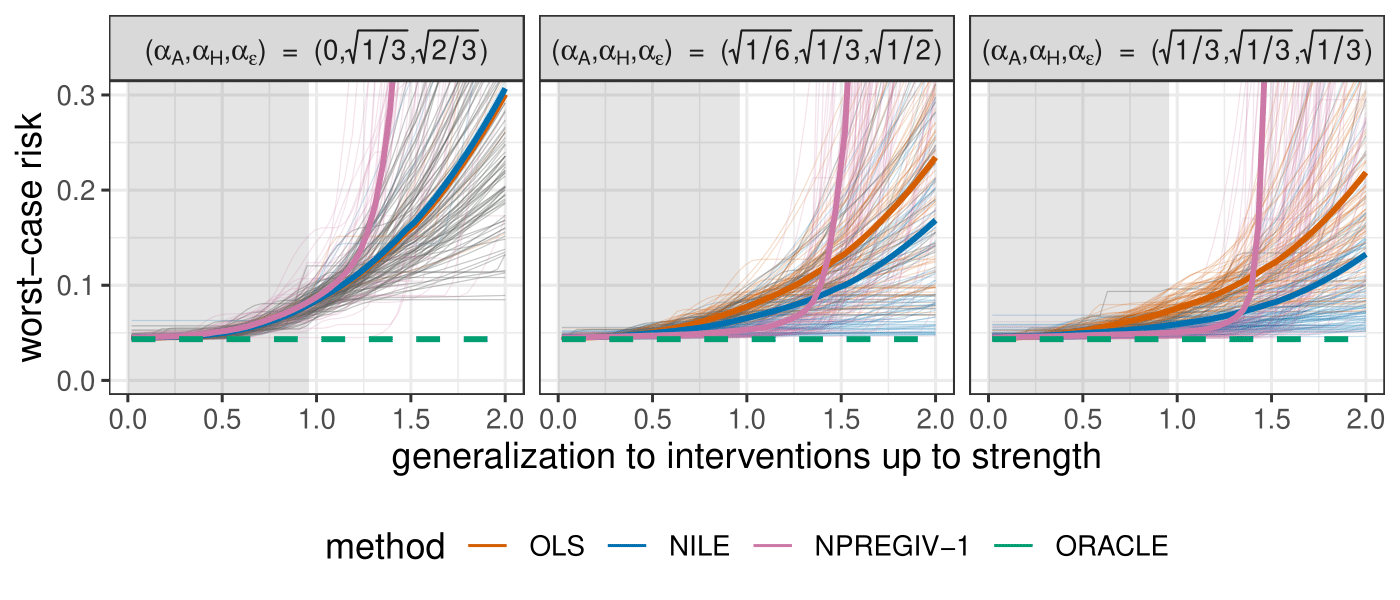}
	\caption{Predictive performance for varying instrument strength. 
		If the
		instruments have no influence on $X$ ($\alpha_A = 0$), the second
		term in the objective function \eqref{eq:thetahat} is effectively
		constant in $\theta$, and the NILE therefore coincides with the OLS
		estimator (which uses $\lambda=0$).  This guards the NILE against
		the large variance which most IV estimators suffer from in a weak
		instrument setting.  For increasing influence of $A$, it clearly
		outperforms both alternative methods for large intervention
		strengths.  %
	}
	\label{fig:weak_instruments}
\end{figure}

Figure~\ref{fig:weak_instruments} shows the worst-case risk for varying instrument strength ($\alpha_A$).
In the case where all exogenous noise comes from the unobserved variable $\epsilon_X$ (i.e., $\alpha_A$ = 0),
the NILE coincides with the OLS estimator. In such settings, standard IV methods are known to perform poorly, 
although also the NPREGIV-1 method seems 
robust to such scenarios. As the instrument strength
increases, the NILE %
clearly outperforms OLS and NPREGIV-1 for interventions on $X$ which 
include values outside the training data.

\begin{figure}
	\centering
	\includegraphics[width=.7\columnwidth]{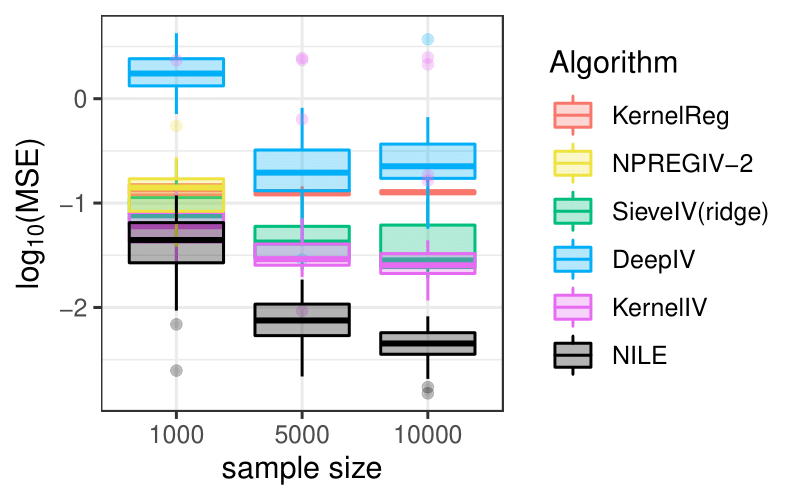}
	\caption{
		Comparison between the NILE and several alternative procedures for learning 
		a nonlinear causal function, 
		based on the same
		experimental setup as in \cite{singh2019kernel}. 
		The estimated functions are evaluated
		on the support (no generalization).
		NILE outperforms
		the competing methods.
	}
	\label{fig:kernelIV}
\end{figure}

We further 
compare NILE's ability to estimate the causal function 
on the support of the covariate $X$
in a nonlinear IV setting 
and compare it with the results from
other state-of-the-art procedures for nonlinear IV estimation, 
following the experimental setup by \cite{singh2019kernel}. Here, the authors consider a predictor variable $X \sim \text{Uniform}(0,1)$ which causally 
influences the target variable $Y$ via the structural assignment $Y := f(X) + \xi_Y$, where $f$ is the 
nonlinear causal function $f(x) = \log( \vert 16 x- 8 \vert +1) \cdot \text{sgn}(x-1/2)$, and $\xi_Y$
is an additive error term which is correlated with $X$. 
They compare their proposed procedure Kernel IV %
to the methods NPREGIV-2 (\cite{singh2019kernel} refer to this method as `Smooth IV'), 
Sieve IV and Deep IV (see Section~\ref{sec:existingmethods}). 
As a baseline, they also include a method for standard kernel ridge regression (`Kernel Reg') \citep{saunders1998ridge}, 
which ignores the existence of hidden confounders. 
Each procedure yields a different estimator $\hat{f}$.
Based on $40$ independent simulations, 
the 
estimators are then compared in terms of 
the average squared distance between $f$ and $\hat{f}$ across 1000 equidistant points in the interval $[0,1]$.
We refer to \cite[][Appendix~A.11]{singh2019kernel} %
for a precise description of the experimental setup. 
Figure~\ref{fig:kernelIV} shows the results of the above experiment 
(corresponding to Figure~2 in \citep{singh2019kernel}), where we have also included the NILE.
Our method outperforms all other procedures, in particular for large
sample sizes. 
There is slight
difference in the way the different algorithms use the available data.
In order to reduce finite sample bias, \cite{singh2019kernel} use sample splitting, 
where the first and second step of the two-stage-least-squares
procedure are performed on disjoint data sets. The NILE, in contrast, uses
all of the data at once. However, even when running our procedure on only half 
of the data, we still outperform the other procedures by a distinct margin, see Figure~\ref{fig:nile50_vs_kernelIV}.
We believe that the superior MSE performance of NILE could be due to 
the different approaches of regularization. For example, 
NILE uses causal regularization similar to that of PULSE, i.e., a data-driven K-class regularization;
in linear IV settings,
this type of regularization 
often yields a 
smaller MSE than 
standard IV methods such as TSLS
\citep{jakobsen2020distributional}.

\section{Discussion and Future Work}
In many real world
problems, the test distribution may differ from the training
distribution.  This requires statistical methods that come with a
provable guarantee in such a setting.  It is possible to characterize
robustness by considering predictive performance for distributions that
are close to the training distribution in terms of standard
divergences or metrics, such as KL divergences or Wasserstein
distance.  
As an alternative view point, we have introduced a novel
framework that formalizes the task of distribution generalization
when
considering distributions that are induced by a set of interventions.
Based on the concept of modularity,
interventions modify parts of the joint distribution and leave other
parts invariant.  
Thereby, they impose constraints on the changes of the distributions that are qualitatively different from considering balls in the above metrics. 
As such, we see them as a useful
language to describe realistic changes between training and test
distributions.

Our framework is general in that it allows us to
model a wide range of causal models and interventions, which do not
need to be known beforehand. 
We have proved several generalization
guarantees, some of which show robustness for distributions that are
not close to the training distribution by considering almost any of
the standard metrics.  
Here, generalization can be obtained by causal functions, but also by non-causal functions; in general, however, the minimizer changes when the 
intervention class is altered (or misspecified).
We have further
proved impossibility results that indicate the limits of what is
possible to learn from the training distribution.  In particular, in
nonlinear models, strong assumptions are required for
distribution generalization to a different support of the
covariates.
As such,
methods such as
anchor regression cannot be expected
to work in nonlinear models,
unless strong restrictions are placed on the function class $\GG$.

Our work can be extended into several directions.  
It
may, for example, be worthwhile to investigate the sharpness of the
bounds we provide in Section~\ref{sec:support_extending_onX} and other
extrapolation assumptions on $\mathcal{F}$. 
Our results make use of the form of the squared loss and it remains an open question to which extent they hold for general convex loss functions.
While our results can be applied to situations where causal background
knowledge is available, via a transformation of SCMs, our analysis is
deliberately agnostic about such information.  It would be interesting
to see whether stronger theoretical results can be obtained by
including causal background information.  We showed that the
type of the interventions play a crucial role in determining whether
the causal function is a minimax optimal solution. Building on this,
it would be interesting to find empirical procedures which test
whether an intervention is confounding-removing,
confounding-preserving or neither.
Finally, 
it could be worthwhile to investigate whether NILE,
which outperforms existing approaches with 
respect to extrapolation,
can be combined with non-parametric 
methods to further improve in-sample performance.
While our current framework already contains certain settings of multi-task learning and domain generalization, it could be instructive to additionally include the possibility to model unlabeled data in the test task.
Finally, our results concern the infinite sample case, but we believe that they can 
form the basis for a corresponding analysis involving rates
or even finite sample results.

We view our work as a step towards understanding the problem of
distribution generalization.
We hope that considering the concepts of
interventions may help to shed further light 
into the question 
of generalizing 
knowledge that was
acquired during training to a different test distribution.

\section*{Acknowledgments}
We thank Thomas Kneib for helpful discussions and two anonymous reviewers for valuable comments. 
RC and JP were supported by 
a research grant (18968) from VILLUM FONDEN; MEJ and JP were supported by the Carlsberg Foundation. 

\chapter[Structure Learning for Directed Trees]{Structure Learning for Directed Trees} \label{ch:trees}

{\small \textsc{Joint work with
    \begin{quote}
       Rajen Shah, Peter Bühlmann and Jonas Peters
    \end{quote}}}
\vspace{0.75cm}
  
\begin{quoting}[leftmargin=0.5cm]
  \begin{center}
    \textbf{Abstract}
  \end{center}
  
  {\small 		Knowing the causal structure of a system is of fundamental interest in many areas of science and can aid the design of prediction algorithms that work well
  	under manipulations to the system. The causal structure becomes identifiable from the
  	observational distribution under certain restrictions. To
  	learn the structure from data, score-based methods evaluate different
  	graphs according to the quality of their fits. However, for large  nonlinear models, these rely on heuristic optimization approaches with no general guarantees
  	of recovering the true causal structure. In this paper, we consider
  	structure learning of directed trees.
  	We propose a fast and scalable method based on
  	Chu–Liu–Edmonds’ algorithm we call causal additive trees (CAT).
  	For the case of Gaussian errors, we prove consistency in an asymptotic regime
  	with a vanishing identifiability gap. We also introduce a method for testing substructure
  	hypotheses with asymptotic family-wise error rate control that is valid post-selection and in unidentified settings.
  	Furthermore, we study the identifiability gap, which quantifies how much better the true causal model fits the observational
  	distribution, and prove that it is lower bounded by local properties of the causal model.
  	Simulation studies demonstrate the favorable performance of CAT compared to competing
  	structure learning methods.}
\end{quoting}
\textbf{Keywords:} 	Causality, restricted causal models, structure learning, directed trees, hypothesis testing.
\section{Introduction}
Learning the underlying causal structure of a stochastic system involving the random vector $X=(X_1,\ldots,X_p)$ is 
an important problem in economics, industry, and science. Knowing the causal structure allows researchers to understand 
whether $X_i$ causes $X_j$ (or vice versa)
and how a system reacts under an intervention. 
However, it is not generally possible to learn the causal structure (or parts thereof) from the observational data of a system alone. Without further restrictions on the system of interest there might exist another system with a different causal structure inducing the same observational distribution, i.e., the structure might not be identifiable from observed data.

Common structure learning methods using observational data are constraint-based \citep[e.g.,][]{Pearl2009, spirtes2000causation}, score-based \citep[e.g.,][]{chickering2002optimal}, or a mix thereof \citep[e.g.,][]{nandy2018high}. Each of these approaches requires different assumptions to ensure 
identifiability of the causal structure and consistency of the approach.  
In structural causal models, one assumes that there are (causal) functions $f_1, \ldots, f_p$ such that for all
$$1\leq i \leq p: \qquad X_i := f_i(X_{\PA i}, N_i),$$ 
for subsets $\PA i \subset \{1, ..., p\}$ and jointly independent noise variables $N=(N_1, ..., N_p)\sim P_N$ (see \Cref{def:StructuralCausalAdditiveTreeModel} for a precise definition including further restrictions). 
The causal graph is constructed as follows: for each variable $X_i$ one adds directed edges from its direct causes or parents $\PA i$ into $i$. For such models, system assumptions concerning the causal functions can make the causal graph identified from the observational distribution. Specific assumptions  that guarantee identifiability of the causal graph have been studied for, e.g., linear Gaussian models with equal noise variance \citep{peters2014identifiability}, linear non-Gaussian models \citep{shimizu2006linear}, nonlinear additive noise models \citep{hoyer2008nonlinear,peters2014causal}, partially-linear additive Gaussian models \citep{rothenhausler2018causal} and discrete models \citep{Peters2011tpami}.

Score-based structure learning 
usually starts with 
a function $\ell$ assigning a population score to causal structures.
Depending on the assumed model class, this function is minimized by the true structure. 
For example, when considering directed acylic graph (DAGs), the true causal DAG $\cG$ satisfy
\begin{align} \label{eq:ScoreFunctionMinimizationIntroduction}
	\cG = \argmin_{\tilde \cG \,: \,\tilde  \cG \text{ is a DAG}} \ell(\tilde \cG).
\end{align}
The idea is then to estimate the score from a finite sample and minimize the empirical score over all DAGs.
As the cardinality of the space of all DAGs grows super-exponentially in the number of nodes $p$ \citep{chickering2002optimal}, brute-force minimization becomes computationally infeasible even for moderately large systems.\footnote{For example, there are over $10^{275}$ distinct directed acyclic graphs over 40 nodes \citep{SizeOfDags}.}

For linear Gaussian models, %
assuming the Markov conditions and
faithfulness,
one can recover the correct Markov equivalence class (MEC) of $\cG$,
which can be represented by a unique completed partially directed acyclic graph (CPDAG) \citep{Pearl2009}.  
The optimization can be done greedily over MECs or 
DAGs
\citep{chickering2002optimal, tsamardinos2006max}
and in the former case, the method is known to be consistent
\citep{chickering2002optimal}.
In the nonlinear case, \citet{CAM} show that nonparametric maximum-likelihood estimation consistently estimates the correct causal order. However, the greedy search algorithm 
minimizing the score function does not come with any theoretical guarantees. 
Recently, methods  have been proposed that
perform continuous, non-convex optimization 
\citep{zheng2018dags} but such methods are without guarantees and it is currently debated whether they exploit some artifacts in simulated data \citep{reisach2021beware}.
Thus, for nonlinear models, there is currently no score-based method that guarantees recovery of the true causal graph with high probability.

This paper focuses on models of reduced complexity, namely models with directed trees as causal graphs. We will show that this complexity reduction allow for computationally feasible minimization of the score-function using the Chu--Liu--Edmonds' algorithm \citep[proposed independently by][]{chuliu1965,edmonds1967optimum}. 
Our method is called causal additive trees (CAT). The method is easy to implement and consists of two steps. In the first step, we employ user-specified (univariate) regression methods to estimate the pairwise conditional means of each variable given all other variables. We then use these to construct edge weights 
as inputs to the Chu--Liu--Edmonds' algorithm. This algorithm then outputs a directed tree with minimal edge weight, corresponding to a directed tree minimizing the score in \Cref{eq:ScoreFunctionMinimizationIntroduction}.

\subsection{Contributions}
We now highlight four main contributions of the paper:

\textit{(i) Computational feasibility:} 
Assuming an identifiable model class, such as additive noise, allows us 
to infer the causal DAG by minimizing   \Cref{eq:ScoreFunctionMinimizationIntroduction} for a suitable score function.
However, even for trees, the cardinality of the search space grows super-exponentially in the number of variables $p$. Hence, brute-force minimization (exhaustive search) in \Cref{eq:ScoreFunctionMinimizationIntroduction}
remains computationally infeasible for large systems.
We propose the score-based method CAT and prove that it recovers the causal tree with a run-time complexity of $\cO(p^2)$.

\textit{(ii) Consistency:} We prove that CAT is pointwise consistent in an identified Gaussian noise setup. That is, we recover the causal directed tree with probability tending to one as the sample size increases. Consistency only requires that the regression methods for estimating the conditional mean functions have mean squared prediction error converging to zero in probability. This property that is satisfied by many nonparametric regression methods such as nearest neighbors, neural networks, or kernel methods \citep[see e.g.][]{gyorfi2002distribution}. Moreover, the vanishing estimation error is only required for causal edges for which the conditional means coincide with the causal functions. 
We also derive sufficient conditions that ensure consistency in an asymptotic setup with vanishing identifiability.
Specifically, we show that consistency is retained even when the identifiability gap decreases at a rate $q_n$ with $q_n^{-1}=o(\sqrt{n})$ as long as the conditional expectation mean squared prediction error %
corresponding to the causal edges vanishes at a rate $o_p(q_n)$.

\textit{(iii) Hypothesis testing:} 
We provide an algorithm for performing hypothesis tests concerning the presence and absence of substructures, such as particular edges, in the true causal graph. The type I error is controlled asymptotically when the mean squared prediction error of the regression corresponding to the true causal edges decays at a relatively slow $o_p(n^{-1/2})$ rate. The tests are valid post-selection, that is, the hypotheses to be tested may be chosen after the graph has been estimated, and when multiple tests are performed, the family-wise error rate is controlled for any number of tests.  In the non-identified setting where multiple minimizers of the population score exist, the inferences derived are valid for the set of minimizers, so one can for instance test whether a particular edge is present in all graphs minimizing the score.

\textit{(iv) Identifiability analysis:} We analyze the identifiability gap, that is,  the smallest population score difference between an alternative graph and the causal graph. The reduced system complexity, due to the restriction to trees, allows us to derive simple yet informative lower bounds. 
For Gaussian additive models, for example, the lower bound can be computed using only local properties of the 
underlying model: it is based on a first term that considers the minimal score gap between individual edge reversals and a second term involving the minimal mutual information of two neighboring nodes, when conditioning on another neighbor of the parent node.

\subsection{Related Constraint-based Approaches}

As an alternative to score-based methods, constraint-based methods such as PC or FCI \citep{spirtes2000causation}
test for conditional independences statements in $P_X$ 
and use these results to infer (parts of) the causal structure. Such methods usually assume that $P_X$ is both Markov and faithful with respect to the causal graph $\cG$. 
Under these assumptions,
the Markov equivalence class of the causal graph $\cG$  %
is identified.
In a jointly Gaussian setting,
consistency of constraint-based approaches relies on faithfulness, whereas uniform consistency 
requires strong faithfulness \citep[see, e.g.,][]{ZhangStrongFaithful,kalisch2007estimating} -- a condition that 
has been shown to be strong
\citep{uhler2013geometry}. 
In nonlinear settings, corresponding guarantees do not exist.
This may at least partially be due to the fact that 
conditional independence testing is known to be a hard statistical problem
\citep{shah2020hardness}.

Constraint-based methods 
have also been studied for 
polytrees. A polytree is a DAG whose undirected graph is a tree. Polytrees, unlike directed trees, allow for multiple root nodes as well as nodes with multiple parents. 
\cite{DBLP:conf/uai/RebaneP87}, inspired by the work of \cite{chow1968approximating}, propose a constraint-based structure learning method for polytrees over discrete variables that can identify the correct skeleton and causal basins, structures constructed from nodes with at least two parents. %
More precisely, the skeleton is determined by the maximum weight spanning tree (MWST) algorithm with mutual information measure weights, while the directionality of edges is inferred by conditional independence constraints implied by the observed distribution.  
In the case of causal trees this constraint-based structure learning method cannot direct any edges because causal basins do not exist \citep{DBLP:conf/uai/RebaneP87}. 
\cite{dominguez2013gaussian} and \cite{ouerd2000learning} extend the \cite{DBLP:conf/uai/RebaneP87} algorithm for causal discovery to multivariate Gaussian polytree distributions.  
In this work, we employ  Chu--Liu--Edmonds' algorithm, a directed analogue of the MWST algorithm, to not only recover the skeleton but also the direction of all edges in the causal graph. This is possible since we consider restricted causal models, e.g., nonlinear additive Gaussian noise models.
(When discarding information that allows us to infer directionality of the edges, 
one recovers the mutual information weights of \cite{DBLP:conf/uai/RebaneP87}, see \Cref{rmk:ConditionalEntropyRebane} in \Cref{sec:AppDetails} for details.) %

\subsection{Organization of the Paper}
In \Cref{sec:ScoreBasedStructureLearning}, we define the setup and relevant score functions. We further
strengthen existing identifiability results for nonlinear additive noise models.
In \Cref{sec:Method},
we propose CAT, an algorithm solving the score-based structure learning problem that is based on 
Chu--Liu--Edmonds' algorithm. We prove consistency of CAT for a fixed distribution and for a setup with vanishing identifiability.
In \Cref{sec:HypothesisTest}, we provide results on asymptotic normality of the scores, construct confidence regions and propose feasible testing procedures. \Cref{sec:ScoreGap}, we analyzes the identifiability gap. \Cref{sec:Simulations} shows the results of various simulation experiments. All proofs can be found in \Cref{app:treesproofs}.

\section{Score-based Learning and Identifiability of Trees
} \label{sec:ScoreBasedStructureLearning}
In the remainder of this work we use of the following graph terminology (a more detailed introduction can be found in \Cref{app:graphs}, see also \citealp{Koller2009}).
A {directed graph} $\cG=(V,\cE)$ consists of $p\in \N_{>0}$ vertices (or nodes)   $V=\{1,\ldots,p\}$ and a collection of directed edges $\cE\subset \{(i\to j) \equiv (i,j): i,j\in V, i\not = j\}$. A {directed acyclic graph} (DAG) is a directed graph that does not contain any directed cycles. 
A {directed tree} is a connected DAG in which all nodes have at most one parent. The unique node of a directed tree $\cG$ with no parents is called the root node and is denoted by $\root{\cG}$.  We 	let $\cT_p$ denote the set of directed trees over $p\in \N_{>0}$ nodes.

\subsection{Identifiability of Causal Additive Tree Models}
We now revisit and strengthen known identifiability results on restricted structural causal models. 
Consider a distribution that is induced by a structural causal model (SCM) with additive noise. 
Then, there are only special cases (such as linear Gaussian models) for which alternative models with a different causal structure exist that generate the same distribution
\citep[see][for an overview]{Peters2017}. 
To state and strengthen these results formally, we introduce the following notation. 

For any $k\in \N$ we define the following classes of functions from $\R$ to $\R$: 
$\cM$ denotes all measurable functions,
$\cD_k$ denotes the set of all $k$ times differentiable functions and
$\cC_k$ denotes the $k$ times continuously differentiable functions. 
We let
$
\cP$ denote the set of mean zero probability measures on $\R$ that have a density with respect to Lebesgue measure. 
$\cP_{+}\subset \cP$ denotes the subset for which a density is strictly positive. 
For any function class $\cF\subseteq \{f|f:\R\to \R\}$, 
$\cP_{\cF}\subset \cP$ denotes the subset with a density function in $\cF$.  As a special case, we let  $\cP_{\text{G}}\subset \cP_{+\cC_\i}:=\cP_{+}\cap \cP_{\cC_\i}$ denote the subset of Gaussian probability measures. 
For any set $\cP$ of probability measures, 
$\cP^p$
denotes all $p$-dimensional product measures on $\R^p$ with marginals in $\cP$.

We now 
define structural causal additive tree models as
SCMs with a tree structure. 
\begin{definition}[Structural causal additive tree models] \label{def:StructuralCausalAdditiveTreeModel}
	Consider a class $\cT_p \times \cM^p \times \cP^p$.
	Any tuple 
	$(\cG,(f_i),P_N)\in \cT_p \times \cM^p \times \cP^p$
	induces a structural causal model 
	over $X=(X_1,\ldots,X_p)$ given by the following structural assignments
	\begin{align*}
		X_i := f_i(X_{\PAg{\cG}{i}})+ N_i, \quad \text{for all } 1\leq i \leq p,
	\end{align*}
	where $f_{\mathrm{rt}(\cG)}\equiv 0$ and $N=(N_1,\ldots,N_p)\sim P_N$,
	which we call a structural causal additive tree model. 
	By slight abuse of notation, we write 
	$Q\in \cT_p \times \cM^p \times \cP^p$ 
	for a probability distribution that is induced by a structural causal additive tree model.
\end{definition}

Furthermore, we define 
the set of 
restricted structural causal additive tree models. 
We will see 
later 
that for these models, the causal graph is %
identifiable from
the observable distribution of the system. 
When the causal graph of a sufficiently nice additive noise SCM is 
not identifiable,
then certain differential equations must hold (see the proof of \Cref{thm:UniqueGraph} for details). 
The definition of restricted structural causal additive tree models ensures that this does not happen.
\begin{definition}[Restricted structural causal additive tree models] \label{def:RestrictedSEMGeneralCase}
	The collection of restricted structural causal additive tree models  
	(or causal additive tree models, for short)
	$\Theta_R\subset \cT_p \times \cD_3^p \times \cP_{+\cC_3}^p$ is given by all models $\theta = (\cG,(f_i),P_N)\in \cT_p \times \cD_3^p \times \cP_{+\cC_3}^p$ satisfying the following conditions for all $i\in \{1,\ldots,p\}\setminus \{\root{\cG}\}$:
	\begin{itemize}
		\item[(i)]  %
		$f_i$ is nowhere constant, i.e., it is not constant on any non-empty open set, and %
		\item[(ii)] the induced log-density $\xi$ of $X_{\PAg{\cG}{i}}$, noise log-density $\nu$ of $N_i$ and causal function $f_i$ are such that there exists $x,y\in \R$ with $\nu''(y-f_i(x))f'_i(x)\not =0$ such that
		\begin{align} \label{eq:DifferentialEquationIdentifiabilityMainText}
			\xi''' \not = \xi''\lp\frac{f_i''}{f_i'} -\frac{\nu''' f_i'}{\nu''} \rp -2\nu''f_i''f_i'+\nu'f_i'''+\frac{\nu'\nu'''f_i''f_i'}{\nu''}-\frac{\nu'(f_i''')^2}{f'},
		\end{align}
		where the derivatives of $\xi,\nu$ and $f_i$ are evaluated in $x$, $y-f_i(x)$ and $x$, respectively.
	\end{itemize} 
\end{definition}

The following lemma, due to \cite{hoyer2008nonlinear}, 
shows that for additive Gaussian noise models, the differential equation constraints of \Cref{def:RestrictedSEMGeneralCase} simplify.\footnote{For completeness, we include the proof of~\Cref{lm:RestrictedModelConditionGaussian} in \Cref{app:treesproofs}, 
	using the approach of \cite{Zhang2009} but expressed in our notation.}

\begin{restatable}[]{lemma}{RestrictedModelConditionGaussian} \label{lm:RestrictedModelConditionGaussian}
	Let $\theta = (\cG,(f_i),P_N)\in \cT_p \times \cD_3^p \times \cP_{\mathrm{G}}^p$.  Assume that for all $i\in\{1,\ldots,p\}\setminus \{ \root{\cG}\}$  the following two conditions hold	
	(a) $f_i$ is nowhere constant 
	and (b) $f_i$ is not linear.
	Then, $\theta\in \Theta_R$.
\end{restatable}

Existing identifiability results for causal graphs in restricted SCMs \citep{hoyer2008nonlinear,peters2014causal} are stated and proven in terms of the ability to distinguish the induced distributions of two restricted structural causal models: For all $\theta= ( \cG,\ldots)\in\Theta_R$ and $\tilde \theta =(\tilde \cG,\ldots)\in \Theta_R$, 
if $\cG \not = \tilde \cG$, then $\cL(X_{\theta}) \not = \cL(X_{\tilde \theta})$, that is, $X_{\theta}$ and $X_{\tilde \theta}$ do not have the same distribution.
We now prove a stronger identifiability result that does not assume that $\tilde \theta$ is a restricted causal model.
\begin{restatable}[Identifiability of causal additive tree models]{proposition}{UniqueGraph}
	\label{thm:UniqueGraph}
	Suppose that $X_\theta$ and $X_{\tilde \theta}$ are generated by the SCMs $\theta = (\cG,(f_i),P_N)\in \Theta_R\subset \cT_p \times \cD_3^p \times \cP_{+\cC_3}^p$  and $\tilde \theta= (\tilde \cG,(\tilde f_i), \tilde P_N)\in \cT_p \times \cD_1^p \times \cP_{\cC_0}^p$, respectively. It holds that
	\begin{align*}
		\cL(X_\theta) = \cL(X_{\tilde \theta}) \implies \cG = \tilde \cG.
	\end{align*}
\end{restatable}
We prove \Cref{thm:UniqueGraph} using the techniques by \cite{peters2014causal}. 
While we prove the statement only 
for causal additive tree models, which suffices for this work, we conjecture that a similar extension holds for restricted structural causal DAG models. 
The extension of \Cref{thm:UniqueGraph} is important for the following reason. Given a finite data set, practical methods usually assume that the true distribution is induced by an underlying restricted SCM. One can then fit different causal structures and output the structure that fits the data best. The above extension accounts for the fact that regression methods hardly represent all such restrictions: e.g., most nonlinear regression techniques can also fit linear models.

\subsection{Score Functions} \label{sec:scorefunctions}
We now
define population score functions which are later used to recover the causal tree. 
We henceforth assume that $X:(\Omega,\cF,P)\to (\R^p,\cB(\R^p))$ is a random vector with distribution $P_X=X(P)$ generated by a %
causal additive tree model
$\theta = (\cG,(f_i),P_N)\in \Theta_{R}\subset \cT_p \times \cD_3^p \times \cP_{+\cC_3}^p$ with $\cG=(V,\cE) \in \cT_p$ such that $\E\|X\|_2^2<\i$. Thus, $\cG$ denotes
the causal tree. We use $\tilde \cG\in \cT_p$ to denote an arbitrary, different (directed) tree.
For the remainder of this paper, we assume that for any $i\not = j$ it holds that $X_i-\E[X_i|X_j]$ has a density with respect to Lebesgue measure.\footnote{This ensures that the entropy score function introduced in \Cref{def:ScoreFunctions} below is well-defined and that the analysis of the identifiability gap in \Cref{sec:ScoreGap} is valid.} 
We often refer to one of the following two scenarios: either, \textit{(i)}, we have limited a priori information that $P_N\in \cP_{+\cC_3}^p$, or, \textit{(ii)}, we know that the noise innovations are Gaussian, that is, $P_N\in \cP_{\mathrm{G}}^p$. Whenever the data-generating noise distributions are Gaussian, 
we refer to this model as a Gaussian setup (or setting or model), even though the full distribution is not.

\begin{definition}  \label{def:ScoreFunctions}
	For any graph $\tilde \cG \in \cT_p$ we define for each node $i\in V$ the
	\begin{enumerate}[label=(\roman*)]
		\item local Gaussian score as $\lG(\tilde \cG,i) :=   \log\lp \Var \lp X_i-  \E\lf X_i|X_{\PAg{\tilde \cG}{i}} \rf \rp  \rp/2$,
		\item local entropy score as $\lE(\tilde \cG,i) := h\lp X_i- \E\lf X_i|X_{\PAg{ \tilde \cG}{i}} \rf\rp$,
		\item local conditional entropy score as $\lCE(\tilde \cG,i) := h\lp X_i| X_{\PAg{\tilde \cG}{i}} \rp$.
	\end{enumerate}	
	Here, we use the convention that $\E(X_i|\emptyset)= 0$ and $h(X_i|\emptyset)=h(X_i)$; the functions $h(\cdot)$, $h(\cdot|\cdot)$, and $h(\cdot,\cdot)$ (used below) denote the differential entropy, conditional entropy, and cross entropy, respectively.
	The Gaussian, entropy  and conditional entropy score of $\tilde \cG$ are, respectively, given by the sum of local scores:
	\begin{align*}
		\lG(\tilde \cG) :=  \sum_{i=1}^p \lG(\tilde \cG,i),\quad \lE(\tilde \cG) := \sum_{i=1}^p \lE(\tilde \cG,i), \quad 
		\lCE(\tilde \cG) :=\sum_{i=1}^p  \lCE(\tilde \cG,i).
	\end{align*}
\end{definition}
(See \cite{polyanskiy2014lecture} or \cite{CoverThomasInformationTheory} for more details on the basic information-theoretic concepts used in this paper.)

The following lemma shows that the Gaussian score of the graph $\tilde \cG \in \cT_p$ arises naturally as a translated  infimum cross entropy between $P_X$ and all $Q$ induced by Gaussian SCMs.
Similarly, the entropy score can be seen as an infimum cross entropy between $P_X$ and all $Q$ induced by another class of SCMs.
\begin{restatable}[]{lemma}{EntropyScore}
	\label{lm:EntropyScore}
	For any $\tilde \cG\in \cT_p$ it holds that
	\begin{align*}
		\lG(\tilde \cG) =\inf_{Q\in \{\tilde \cG\} \times \cD_1^p \times \cP_{\mathrm{G}}^p} h(P_X,Q) -p\log(\sqrt{2\pi e}).
	\end{align*}
	Furthermore, with $\cF(\tilde \cG) := (\cF_i(\tilde \cG))_{1\leq i \leq p}$, where $\cF_i(\tilde \cG) := \{x\mapsto \E[X_i|X_{\PAg{\tilde \cG}{i}}=x]\}$ for all $1\leq i\leq p$, it holds that
	\begin{align*}
		\lE(\tilde \cG) =\inf_{Q\in \{\tilde \cG\} \times \cF(\tilde \cG) \times \cP^p} h(P_X,Q).
	\end{align*}
\end{restatable}

Score-based methods identify the underlying structure by
evaluating the score functions (or estimates thereof) on different graphs and choosing the best scoring graph.
The difference between the score 
$\ell_{\cdot}(\cG)$
of the true graph
and the score $\ell_{\cdot}(\tilde \cG)$
of the 
best scoring 
alternative graph $\tilde \cG$
is an important property of the problem: 
e.g., if it would be zero, we could not identify the true graph from the scores. 
We, therefore, refer to 
expressions of the form
$\min_{\tilde \cG \in \cT_p \setminus \{\cG\}}\ell_{\cdot}(\tilde \cG)- \ell_{\cdot}(\cG)$
as the identifiability gap. 

In the remainder of this paper, we work under the assumption that the identifiability gap is strictly positive (see also Section~\ref{sec:ScoreGap}).%

\begin{assumption} \label{ass:identifiabilityOfConditionalMeanScores}
	If $\theta\in \Theta_R\subset   \cT_p \times \cD_3^p \times \cP_{\mathrm{G}}^p$ or $\theta\in   \Theta_R\subset \cT_p \times \cD_3^p \times \cP_{+\cC_3}^p$ it holds 
	that
	\begin{align} \label{eq:GaussianInfimumKLDivergence}
		\min_{\tilde \cG \in \cT_p \setminus \{\cG\}}\lG(\tilde \cG)- \lG(\cG) > 0 \quad \text{or} \quad \min_{\tilde \cG \in \cT_p \setminus \{\cG\}}\lE(\tilde \cG)- \lE(\cG) > 0,
	\end{align}
	respectively.
\end{assumption}

\Cref{ass:identifiabilityOfConditionalMeanScores}
does not trivially follow from the results further above.
By arguments similar to those in \Cref{lm:EntropyScore} we have that, 
if the true data-generating model is a restricted Gaussian additive tree model, $\theta \in \Theta_R \subset \cT_p \times \cD_3^p \times \cP_{\mathrm{G}}^p$, then $\lG(\cG) = h(P_X)-p\log(\sqrt{2\pi e})$. Hence, the Gaussian score gap between $\tilde \cG$ and the causal graph $\cG$ equals
\begin{align*}
	\lG(\tilde \cG)- \lG(\cG) = \inf_{Q\in \{\tilde \cG\} \times \cD_1^p \times \cP_{\mathrm{G}}^p} h(P_X,Q) - h(P_X) =  \inf_{Q\in\{\tilde \cG\} \times \cD_1^p \times \cP_{\mathrm{G}}^p}	D_{\mathrm{KL}}(P_X\| Q),
\end{align*}
where $D_{\mathrm{KL}}$ denotes the Kullback-Leibler divergence measure.
\Cref{thm:UniqueGraph} implies that 
$$
\forall \tilde \cG \not = \cG, \quad 
\forall Q\in\{\tilde \cG\} \times \cD_1^p \times \cP_{\mathrm{G}}^p: D_{\mathrm{KL}}(P_X\| Q)>0.
$$
However, this does not immediately imply
that the 
identifiability gap (where we take the infimum over such $Q$)
is 
strictly positive. 
Similar  considerations\footnote{In fact, \Cref{thm:UniqueGraph} does  not immediately imply that $D_{\mathrm{KL}}(P_X\| Q)>0$ for $Q\in \{\tilde \cG\} \times \cF(\tilde \cG) \times \cP^p$ as it does not necessarily hold that the causal functions in $\cF(\tilde \cG)$ are differentiable or that the noise innovation densities in $\cP^p$ are continuous.}
hold for the entropy score gap
\begin{equation*} 
	\lE(\tilde \cG) - \lE(\cG) 
	= \inf_{Q\in \{\tilde \cG\} \times \cF(\tilde \cG) \times \cP^p} D_{\mathrm{KL}}(P_X\| Q).
\end{equation*}

In \Cref{sec:ScoreGap} we derive informative lower bounds on the Gaussian and entropy score gaps (i.e., the infimum KL-divergence) of \Cref{eq:GaussianInfimumKLDivergence}.
It is also possible to enforce \Cref{ass:identifiabilityOfConditionalMeanScores} indirectly by the assumptions and modifications detailed in the following remark.
\begin{remark} \label{rm:IdentifiabilityRemark}
	If $\theta\in \Theta_R\subset \cT_p \times \cD_3^p \times \cP_{\mathrm{G}}^p$, such that for all $i\not =j$ it hold that $x\mapsto \E[X_i|X_j=x]$ has a differentiable version, 
	then the Gaussian identifiability gap is strictly positive, so 
	the first part of \Cref{ass:identifiabilityOfConditionalMeanScores}  holds.
	If $\theta\in \Theta_R\subset \cT_p \times \cD_3^p \times \cP_{+\cC_3}^p$ and, in addition to the above condition it holds that
	for all $i\not = j$, 
	$X_i-\E[X_i|X_j]$ has a continuous density, then %
	then the entropy identifiability gap 
	is strictly positive, 
	so the in second part of \Cref{ass:identifiabilityOfConditionalMeanScores} holds.
	
	\Cref{ass:identifiabilityOfConditionalMeanScores} can also be enforced by adopting the model restrictions of \cite{CAM}. Assume that $\Theta_R\subset \cT_p \times \cD_3^p \times \cP_{\mathrm{G}}^p$ satisfies the further restriction  that for all causal edges $(j\to i)\in \cE$  the causal functions $f_i$ are contained within a function class $\cF_i\subset \cD_1$ that is closed with respect to the $L^2(P_{X_{j}})$-norm. Now consider a modified Gaussian score function $\ell_{\mathrm{G.mod}}:\cT_p \to \R$ that coincides with $\lG$ except that the conditional expectation function is replaced with $\argmin_{f'\in \cF_i}\E[(X_i-f'(X_j))^2]\in \cF_i$. It now follows that
	$$
	\ell_{\mathrm{G.mod}}(\tilde \cG) - \ell_{\mathrm{G.mod}}(\cG) = \inf_{Q\in \{\tilde \cG\}\times (\cF_i)_{1\leq i \leq p} \times \cP_{\mathrm{G}}^p} D_{\mathrm{KL}}(P_{X}\| Q) >0,
	$$	
	where the strict inequality follows from \Cref{thm:UniqueGraph} as the infimum is attained for some $Q^*\in \{\tilde \cG\}\times (\cF_i)_{1\leq i \leq p} \times \cP_{\mathrm{G}}^p $.  Our theory and subsequent results transfer effortlessly to these modifications. 
\end{remark}
We can now use the score functions to identify the true causal graph of a restricted structural model. In the Gaussian case, for example, we have, by virtue of \Cref{ass:identifiabilityOfConditionalMeanScores},
\begin{align} \label{eq:CausalGraphMinimizesGaussianScore}
	\cG	= \argmin_{\tilde \cG \in \cT_p} \lG(\tilde \cG).
\end{align}
In practice, we consider estimates of the above quantities and optimize the corresponding empirical loss function.
Solving \Cref{eq:CausalGraphMinimizesGaussianScore}
(or its empirical counterpart)
using exhaustive search is computationally intractable already for moderately large choices of $p$.\footnote{In the context of linear Gaussian models, \citet{chickering2002optimal} proves consistency of greedy equivalent search towards the correct Markov equivalence class. 
	This, however,  does not imply that the optimization problem in \Cref{eq:CausalGraphMinimizesGaussianScore} is solved: for a given sample, the method is not guaranteed to find the optimal scoring graph (but the output will converge to the correct graph).} We now introduce CAT, a computationally efficient method that solves the optimization exactly.

\section{Causal Additive Trees (CAT)} \label{sec:Method}
We 
introduce the population version of our algorithm CAT
in Section~\ref{sec:oracle} and discuss its finite sample version and asymptotic properties in Sections~\ref{sec:algor} and~\ref{sec:Consistency}.
\subsection{An Oracle Algorithm} \label{sec:oracle}
Similarly as for the case of
DAGs, the problem in \Cref{eq:CausalGraphMinimizesGaussianScore} is a combinatorial optimization problem, for which the cardinality of the search space grows super-exponentially with $p$. 
Indeed, the number of undirected trees on $p$ labelled nodes is $p^{p-2}$ \citep{cayley1889theorem} and therefore $p^{p-1}$ is the corresponding number of labelled trees. 
For the class of DAGs (which includes directed trees), existing structure learning 
such as \cite{CAM} propose a greedy search technique that iteratively selects the lowest scoring directed edge under the constraint that no cycles is introduced in the resulting graph. 
In general, greedy search procedures do not come with any guarantees and
there are indeed situations in which they fail \citep{Peters2022}.
By exploiting the assumption of a tree structure, 
we will see that
the 
optimization problem of \Cref{eq:CausalGraphMinimizesGaussianScore} can be solved computationally efficiently without the need for heuristic optimization techniques.

Provided with a
connected directed graph with edge weights, 
Chu--Liu--Edmonds'  algorithm
finds a minimum edge weight directed spanning tree, given that such a tree exists.
That is, for a connected directed graph $\cH = (V,\cE_{\cH})$ on the nodes $V=\{1,\ldots,p\}$ with edge weights  $\mathbf{w} :=\{w(j\to i): j \not = i\}$, Chu--Liu--Edmonds' algorithm recovers a minimum edge weight spanning directed tree subgraph  of $\cH$, 
\begin{align*}
	\argmin_{\tilde \cG =(V,\tilde \cE) \in \cT_p\cap \cH} \sum_{(j\to i)\in \tilde \cE} w(j\to i),
\end{align*}
where $\cT_p\cap \cH$ denotes all directed spanning trees of $\cH$.
The runtime of the original algorithms of \cite{chuliu1965} and \cite{edmonds1967optimum} for a pre-specified root node is $\cO(|\cE_{\cH}|\cdot p) = \cO(p^3)$.
\cite{tarjan1977finding} devised a modification of the algorithm that for dense graphs $\cH$ and an unspecified root node has runtime $\cO(p^2)$. 
In our experiments, 
we use the  \verb!C++! implementation of Tarjans modification by \cite{Edmondscpp} which is contained in the R-package \verb!RBGL! \citep[][]{carey2011package}.

The causal graph recovery problem in \Cref{eq:CausalGraphMinimizesGaussianScore} is equivalently solved by finding a minimum edge weight directed tree, i.e., a minimum edge weight directed spanning tree of the fully connected graph on the nodes $V$. For example, finding the minimum of the Gaussian score function is equivalent to minimizing a translated version of the Gaussian score function
\begin{align*}
	\argmin_{\tilde \cG \in \cT_p} \lG(\tilde \cG)
	&	= \argmin_{\tilde \cG \in \cT_p}  \sum_{i=1}^p \frac{1}{2}\log( \Var ( X_i-  \E[ X_i|X_{\PAg{\tilde \cG}{i}} ] )  ) - \sum_{i=1}^p \frac{1}{2}\log(\Var(X_i)) \\
	&= \argmin_{\tilde \cG \in \cT_p}  \sum_{i=1}^p \frac{1}{2} \log\lp \frac{ \Var ( X_i-  \E[ X_i|X_{\PAg{\tilde \cG}{i}} ]) }{\Var(X_i)} \rp.
\end{align*}
Because the summand for the root note equals zero,
we only need to sum over all nodes with an incoming edge in $\tilde \cG$:
\begin{align*}
	\cG =	\argmin_{\tilde \cG \in \cT_p} \lG(\tilde \cG) &=  \argmin_{\tilde \cG =(V,\tilde \cE)\in \cT_p} \sum_{(j \to i )\in \tilde \cE} w_{\mathrm{G}}(j \to i) ,
\end{align*}
for a Gaussian data-generating model. That is, the causal directed tree is given by the minimum edge weight directed tree with respect to the Gaussian edge weights  $\mathbf{w}_{\mathrm{G}}:=\{w_{\mathrm{G}}(j\to i):j\not = i\}$ given by
\begin{align} \label{eq:GaussianPopulationWeights}
	w_{\mathrm{G}}(j\to i) :=\frac{1}{2} \log\lp \frac{ \Var ( X_i-  \E[ X_i|X_{j} ]) }{\Var(X_i)} \rp
\end{align}
for all $j\not = i$. Similarly, the minimum of the entropy score function is given by the minimum edge weight directed tree with respect to the entropy edge weights  
$\mathbf{w}_{\mathrm{E}}:= \{w_{\mathrm{E}}(j \to i):j\not = i\}$ given by $w_{\mathrm{E}}(j\to i) :=  h(X_i-\E[X_i|X_j]) - h(X_i)$, 
for all $j\not = i$. We will henceforth denote the method where we apply  Chu--Liu--Edmonds' algorithm on Gaussian and entropy edge weights  as CAT.G and CAT.E, respectively.
\subsection{Finite Sample Algorithm} \label{sec:algor}

Given an $n \times p$ data matrix 
$\mathbf{X}_n$, representing $n$ i.i.d.\ copies of $X= (X_1, \ldots, X_p)$, we estimate the edge weights  by simple plug-in estimators. 
Let us denote the conditional expectation function and its estimate 
by
\begin{align} \label{def:phi}
	\phi_{ji}(x):=\E[X_i|X_j=x],\quad \quad \hat \phi_{ji}(x):=\hat \E[X_i|X_j=x],
\end{align}
for any $j \not = i$.  The estimated Gaussian edge weights  are then given by
\begin{align} \label{eq:estiamtededgeweights}
	\hat w_{\mathrm{G}}(j\to i) := \frac{1}{2}\log \lp\frac{ \widehat \Var(X_i - \hat \phi_{ji}(X_j))} {\widehat \Var(X_i)} \rp,
\end{align}
for all $i\not = j$, where  $\widehat \Var(\cdot)$ denotes a  variance estimator using the sample $\fX_n$.  We now propose to combine the  Chu--Liu--Edmonds' algorithm described above with the Gaussian score as detailed in \Cref{alg:Edmonds}. 
\begin{algorithm}[h] \caption{Causal additive trees (CAT)} \label{alg:Edmonds}
	\begin{algorithmic}[1]
		\Procedure{CAT}{$\fX_n$, regression method}
		\State For each 
		combination of $(i,j)$ with $j\not = i$, run %
		regression method to obtain 
		$\hat \phi_{ji}$.
		\State Compute empirical edge weights  $ \mathbf{\hat w}_{\mathrm{G}}:=(\hat w_{\mathrm{G}}(j\to i))_{j\not = i}$,  
		see~\Cref{eq:estiamtededgeweights}.
		\State Apply Chu--Liu--Edmonds' algorithm to the empirical edge weights.
		\State \textbf{return} minimum edge weight directed tree $\hat \cG$.
		\EndProcedure
	\end{algorithmic} 
\end{algorithm}

By default we suggest to use the estimated Gaussian edge weights  as described in \Cref{alg:Edmonds}.
However, it is also possible to run Chu--Liu--Edmonds' algorithm on estimated entropy edge weights  given by 
\begin{align*}
	\hat 	w_{\mathrm{E}}(j\to i) &:=  \hat h( X_i - \hat \phi_{ji}(X_j) ) - \hat h(X_i),
\end{align*}
for all $j\not =i$, where $\hat h(\cdot)$ denotes a user-specific entropy estimator using the observed data $\fX_n$. Estimating differential entropy is a difficult statistical problem but we will later in \Cref{sec:Simulations} demonstrate by simulation experiments that it can be beneficial to use the estimated entropy edge weights   when the additive noise distributions are highly non-Gaussian.

Under suitable conditions on the (possibly nonparametric) regression technique, we now show that the proposed algorithm consistently recovers the true causal graph in Gaussian settings using estimated Gaussian edge weights. 

\subsection{Consistency} \label{sec:Consistency}
We study a version of the CAT.G algorithm applied to a Gaussian noise model where the regression estimates are trained on auxiliary data, simplifying the theoretical analysis. We believe that consistency without sample splitting holds but may be more difficult to prove. As such, we only view the sample splitting as a theoretical device for simplifying proofs but we do not recommend it in practical applications. For each $n$ we let $\fX_n=((X_{1,i})_{1\leq i \leq p},\ldots,(X_{n,i})_{1\leq i \leq p})$ and $\tilde{\fX}_n=( (\tilde X_{1,i})_{1\leq i \leq p},\ldots, (\tilde X_{n,i})_{1\leq i \leq p})$ denote independent datasets each consisting of $n$ i.i.d.\ copies of $X=(X_1,...,X_p)\in \R^p$. We suppose that the regression estimates $\hat{\varphi}_{ji}$ have been trained on $\tilde{\fX}_n$ and then compute the edge weights using $\fX_n$ as in step 3 of Algorithm~\ref{alg:Edmonds}:
\begin{align} \label{eq:samplesplitEdgeWeight}
	\hat w_{\mathrm{G}}(j \to i):=\hat w_{ji}(\fX_n,\tilde \fX_n) := \frac{1}{2} \log \left(  \frac{\frac{1}{n}\sum_{k=1 }^n \left( X_{k,i} - \hat \phi_{ji} (X_{k,j}) \right)^2}{\frac{1}{n}\sum_{k=1}^n X_{k,i}^2 - (\frac{1}{n}\sum_{k=1}^n X_{k,i})^2} \right).
\end{align}

The following result shows pointwise consistency of CAT.G  whenever the conditional mean estimation is weakly consistent.

\begin{restatable}[Pointwise consistency]{theorem}{Consistency}
	\label{thm:consistency}
	Suppose that for all $j \not = i$ 
	the following two conditions hold:
	\begin{enumerate}[label=(\alph*)]
		\item if $(j \to i )\in \cE$, $\E[(\hat \phi_{ji}(X_j)-\phi_{ji}(X_j))^2|\tilde \fX_n] \convp_n 0$;
		\item if $(j\to i) \not \in \cE$, $\E[(\hat \phi_{ji}(X_j)-\tilde \phi_{ji}(X_j))^2|\tilde \fX_n] \convp_n 0$ for some fixed $\tilde \phi_{ji}:\R\to \R$,
	\end{enumerate} 
	where $\phi_{ji}$ and $\hat \phi_{ji}$ are defined in \Cref{def:phi}.
	In the large sample limit, we recover the causal graph with probability one, that is
	\begin{align*}
		P(\hat{\mathcal{G}} = \mathcal{G})
		\to_n 1,
	\end{align*}
	where $\hat \cG$ is the output of \Cref{alg:Edmonds} using weights $\hat w_{\mathrm{G}}(j\to i)$  given by \Cref{eq:samplesplitEdgeWeight}.
\end{restatable}
The assumptions of \Cref{thm:consistency} only require weakly consistent estimation of the conditional means for edges that are  present in the causal graph; these represent causal relationships and are often assumed to be smooth. 
This distinction allow us to employ regression techniques that are consistent only for those function classes that we consider reasonable for modeling the causal mechanisms.
For non-causal edges, $(j\to i)\not \in \cE$, 
the estimator $\hat \phi_{ji}$ only needs to converge to a function $\tilde \phi_{ji}$, which does
not necessarily need to be the conditional mean.
\subsubsection{Consistency under Vanishing Identifiability}
We now consider an asymptotic regime involving a sequence $(\theta_n)_{n\in \N}$ of SCMs with potentially changing conditional mean functions $\varphi_{ji}$ and a vanishing identifiability gap. We have the following result.
\begin{restatable}[Consistency under vanishing identifiability]{theorem}{ConsistencyVanishingIdentifiability}
	\label{thm:ConsistencyVanishing}
	Let $(\theta_n)_{n\in \N}$ be a sequence of SCMs  on $p\in \N$ nodes all with the same causal directed tree $\cG=(V,\cE)$ such that 
	\begin{enumerate}[label=(\roman*)]
		\item  for $q_n := \min_{\tilde \cG\in \cT_p\setminus \{\cG\}} \lG(\cG) - \lG(\tilde \cG)$ (the gap of model $\theta_n$), we have $q_n^{-1}=o(\sqrt{n})$; 
		\item for all $(j\to i)\in \cE$ and $\ep >0$, 
		$ P_{\theta_n}\lp q_n^{-1}\E_{\theta_n}\lf (\phi_{ji}(X_j)- \hat \phi_{ji}(X_j))^2   | \tilde \fX_n  \rf > \ep \rp\to_n 0$;
		\item for all $j\not = i$ and $\ep>0$, $
		P_{\theta_n}\lp \frac{q_n^{-2}}{n}\E_{\theta_n}\lf (\phi_{ji}(X_j)- \hat \phi_{ji}(X_j))^4   | \tilde \fX_n  \rf > \varepsilon  \rp   \to_n 0$; and
		\item there exists $C>0$ such that for all $j\not = i$ $
		\inf_{n} P_{\theta_n}(\Var_{\theta_n}(X_i|X_j)\leq C)=1$ and 
		$\sup_{n} \E_{\theta_n}  \|X\|_2^4<\i$.
	\end{enumerate}
	Then it holds that
	\begin{align*}
		P(\hat{\mathcal{G}} = \mathcal{G}) \to_n 1.
	\end{align*}
\end{restatable}
Condition (i) asks that the identifiability gap $q_n$ goes to zero more slowly than the standard convergence rate $1/\sqrt{n}$ of estimators in regular parametric models. Such a requirement would be necessary in almost any structure identification problem. Condition (ii) requires the mean squared error of the regression estimates corresponding to true causal edges to be $o_P(q_n)$. We regard this as a fairly mild assumption: indeed, the minimax rate of estimation of regression functions in H\"older balls with smoothness $\beta$ is $n^{-2\beta / (2\beta + 1)}$ \citep{Tsybakov}. Thus, we can expect that if the causal regression functions have smoothness $\beta \geq 1/2$ and all lie in a H\"older ball, (ii) can be satisfied for any $q_n$ satisfying (i). Condition (iii) allows the fourth moments of the estimation errors to increase at any rate slower than  $n q_n^{2} \to \infty$; of course, we would typically expect this error to decay, at least for the causal edges.
\section{Hypothesis Testing} \label{sec:HypothesisTest} 
This section presents a 
procedure to test any substructure hypothesis regarding the causal directed tree of a Gaussian additive noise model. We continue  our analysis using the sample split estimators of \Cref{eq:samplesplitEdgeWeight},  where the conditional expectations are estimated on an auxiliary dataset. Our approach makes use of the fact that the estimated weights in \Cref{eq:samplesplitEdgeWeight} are logarithms of ratios of i.i.d.\ quantities, and thus the joint distribution of the estimated edge weights should, with appropriate centering and scaling, be asymptotically Gaussian; see \Cref{thm:asymptoticnormalityedgecomponents} in \Cref{app:treesproofs} for 
the precise statement.
This allows us to create a (biased) confidence region of the true edge weights, which in turn gives a  confidence set for the true graph. This confidence set of graphs is not necessarily straightforward to compute and list. However, we show that it can be queried to test hypotheses of interest, such as the presence or absence of a particular edge. As these hypothesis tests are derived from a confidence region, they are valid even when the hypothesis to test has been chosen after examining the data.

Similar to the results in the previous sections, we avoid making assumptions on the performance of regressions corresponding to non-causal edges. Unlike the consistency analysis, however, here we do not require identifiability of the true graph, but in the non-identified case all assumptions and conclusions below involving the `true graph' should be interpreted as involving the set of all population score minimizing graphs. In order to state our results and assumptions, we introduce the following notation. 

For any collection $(K_{ji})_{j\not = i}$,  we let $
K_{i} := (K_{1i}, \ldots,K_{(i-1)i},K_{(i+1)i},\ldots,K_{pi})^\t \in \R^{p-1}$, furthermore, for any collection $(K_i)_{1\leq i \leq p}$, we let $
K  := (K_{1},\ldots,K_{p})^\t$. With this notation, let, for all $k\in\{1,...,n\}$, the vectors of squared residuals and squared centered observations 
be given by
\[
\hat{M}_k := \{(X_{k,i} - \hat{\varphi}_{ji}(X_{k,j}))^2\}_{j \neq i} \in \R^{p(p-1)}, \qquad \hat{V}_k= \bigg\{\bigg(X_{k,i} - \frac{1}{n}\sum_{m=1}^n X_{m,i} \bigg)^2\bigg\}_{1\leq i \leq p} \in \R^p.
\]
Further let
\[
\hat{\mu} := \frac{1}{n} \sum_{k=1}^n \hat{M}_k, \qquad \hat{\nu} =: \frac{1}{n} \sum_{k=1}^n \hat{V}_k.
\]
Note that with this notation, the estimated Gaussian edge weight for $j \to i$ is given by $\log(\hat{\mu}_{ji} / \hat{\nu}_i) / 2$. Let us denote by $\widehat{\Sigma}_M \in \R^{p(p-1) \cdot p(p-1)}$, $\widehat{\Sigma}_V \in \R^{p \cdot p}$ and $\widehat{\Sigma}_{MV} \in \R^{p(p-1) \cdot p}$, the empirical variances of the $\hat{M}_k$ and $\hat{V}_k$ and their empirical covariance respectively, so
\[
\begin{pmatrix}
	\widehat \Sigma_{M} & \widehat \Sigma_{MV} \\ \widehat \Sigma_{MV}^\t & \widehat \Sigma_{V}  
\end{pmatrix}:=\frac{1}{n}\sum_{k=1}^n \begin{pmatrix}
	\hat M_{k}
	\hat M_{k}^\t -\hat \mu\hat \mu^\t & \hat M_{k}  \hat V_{k}^\t- \hat \mu \hat \nu^\t  \\
	\hat V_k \hat M_k^\t- \hat \nu \hat \mu^\t &  V_{k}V_{k}^\t- \hat \nu \hat \nu^\t
\end{pmatrix}.
\]
With this, we may now present our construction of confidence intervals for the edge weights. (For simplicity, all proofs in this section assume the variables to have mean zero.)

\subsection{Confidence Region for the Causal Tree} \label{sec:Confidence}

We use the delta method to estimate the variances of the $\hat{w}_{ji}$, and a simple Bonferroni correction to ensure simultaneous coverage of the confidence intervals we develop. Writing $z_{\alpha}$ for the upper $\alpha / \{2p(p-1)\}$ quantile of a  standard normal distribution, we set
\begin{align*}
	\hat u_{ji},\, 	\hat l_{ji}   := & \, \frac{1}{2}\log\lp\frac{\hat \mu_{ji}}{\hat \nu_{i}}\rp \pm z_\alpha \frac{\hat \sigma_{ji}}{2\sqrt{n}},
\end{align*}
where
\begin{align*}
	\hat \sigma_{ji}^2 := \frac{\widehat \Sigma_{M,ji,ji}}{\hat \mu_{ji}^2} + \frac{\widehat \Sigma_{V,i,i}}{\hat \nu_{i}^2} - 2 \frac{\widehat \Sigma_{MV,ji,i}}{\hat \mu_{ji} \hat \nu_{i}}.
\end{align*}
We treat $[\hat{l}_{ji}, \hat{u}_{ji}]$ as a confidence interval for the true edge weight $w_{\mathrm{G}}(j\to i)$
and define the following region of directed trees formed of minimizers of the score with edge weights in the confidence hyperrectangle:
\begin{align*}
	\hat{C} := \bigg\{ \argmin_{\tilde \cG=(V,\tilde \cE)\in \cT_p} \sum_{(j \to i )\in \tilde \cE}w_{ji}', :\, \,  &\forall j\not = i , w_{ji}'\in[\hat l_{ji},\hat u_{ji}]\bigg\}.
\end{align*}
We have the following coverage guarantee for $\hat{C}$.
\begin{restatable}[Confidence region]{theorem}{thmconfidence}
	\label{thm:Confidence}
	Suppose the following conditions hold:
	\begin{itemize}
		\item[(i)] there exists $\xi > 0$ such that $\E \|X\|^{4 + \xi} < \infty$;
		\item[(ii)] there exists $\xi > 0$ such that for all $j \neq i$,  $\E[|\hat \phi_{ji}(X_{j})- \phi_{ji}(X_{j})|^{4+\xi}|\tilde \fX_n] = O_p(1)$;
		\item[(iii)] $\Var( (\hat{M}_1^\t, \hat{V}_1^\t)^\t |\tilde \fX_n )\convp_n \Sigma$, where $\Sigma$ is constant with strictly positive diagonal;
		\item[(iv)] for $(j\to i)\in \cE $, $\sqrt{n}\E[ (\hat \phi_{ji}(X_{k,j})- \phi_{ji}(X_{k,j}))^2 |\tilde \fX_n ]\convp 0$. \label{cond:sqrtconv}
		
	\end{itemize}
	Then
	\begin{align*}
		\liminf_{n\to \i} P\lp \cG \in \hat{C} \rp \geq 1-\alpha.
	\end{align*}
\end{restatable}
The second condition requires little more than 4th moments for the absolute errors in the regression (they do not need to converge to zero). 
Condition (iv) requires that the mean squared prediction errors corresponding to the true causal edges decay faster than a relatively slow $1/\sqrt{n}$ rate. 
If the causal graph is unidentifiable, then when (iv) holds for all edges corresponding to population score minimizing graphs, $\hat{C}$ will cover every such graph with a probability of at least $1-\alpha$.

\subsection{Testing of Substructures} \label{sec:testing}
Whilst the confidence region $\hat{C}$ has attractive coverage properties, it will typically not be possible to compute it in practice (due to the ranges of $w_{ji}'$ one would need to try). We now introduce a computationally feasible scheme for querying whether $\hat{C}$ 
satisfies certain constraints such as 
containing or not containing a given substructure.
A substructure $\cR= (\cE_\cR, \cE_\cR^{\text{miss}},r)$ on the nodes $V$ contains specified sets $\cE_\cR$ and $\cE_\cR^{\text{miss}}$ of existing and missing edges, respectively,
and/or a specific root node $r$; for example, this could be a specific directed tree or a single edge (such as $X_1 \to X_2$) or a single missing edge (such as $X_1 \not \to X_2$). 
Our approach allows us to conclude that  
at least one of the constraints in $\mathcal{R}$ does \emph{not} hold for the true graph. 
More precisely, we propose a test for the null hypothesis
\begin{align*}
	\cH_0(\cR) : \cE_\cR \setminus \cE = \emptyset, \;  \cE\setminus \cE^\text{miss}_\cR  = \emptyset,\; r=\root{\cG},
\end{align*}
i.e, that all constraints in $\cR$ are satisfied in the causal graph. %

In order to present our method, we introduce some notation.
Let 
$s(w)$ 
be the score attained by the minimum edge weight directed tree recovered by  Chu--Liu--Edmonds'  algorithm with input edge weights  
$w:=(w_{ji})_{j\not = i}$. 
Let $\cT(\cR)\subset \cT_p$ be the set of all directed trees 
satisfying the constraints $\cR$. Furthermore, let 
$s_{\cT(\cR)}(w)$ 
be the score attained by the minimum edge weight directed tree in $\cT(\cR)$. Now suppose that the causal directed tree $\cG$ satisfies the constraints $\cR$. If $[\hat l, \hat u]:=\prod_{j\not =i} [\hat l_{ji}, \hat u_{ji}]$ was an asymptotically valid confidence region for the Gaussian population edge weights 
$\mathbf{w}_{\mathrm{G}}$ 
defined in \Cref{eq:GaussianPopulationWeights}, we have 
with probability tending to at least $1-\alpha$ that
\begin{align*}
	s_{\cT(\cR)}(\hat{l})  \leq s_{\cT(\cR)}(\mathbf{w}_{\mathrm{G}}) = s(\mathbf{w}_{\mathrm{G}}) \leq s(\hat{u}).
\end{align*}
We may thus set as our test function
\[
\psi_{\cR} := \mathbbm{1}_{\{ s_{\cT(\cR)}(\hat{l}) > s(\hat{u})\}}.
\]

The expressions $s_{\cT(\cR)}(\hat{l})$ and $s(\hat{u})$ 
can be computed from the data. For
$s_{\cT(\cR)}(\hat{l})$, we perform the following steps: we apply Chu--Liu--Edmonds'  algorithm on the edge weights  $\hat{l}$ where, for any $(j\to i)\in \cE_\cR$, we remove all other  edges into $i$ from the edge pool (or set the corresponding edge weight to sufficiently large values) while for a specified root node $r \in \cR $ we remove all incoming edges into $r$ from the edge pool.
Edges $(j\to i)\in \cE_\cR^\text{miss}$ are removed from the edge pool, too.

Formalizing a line of reasoning similar  to the above, taking into account that $[\hat l, \hat u]$ is in fact a biased confidence region that may not necessarily contain the population edge weights with increasing probability, we have the following result. 
\begin{restatable}[Pointwise asymptotic level]{theorem}{thmtestlevel}
	\label{thm:testlevel}
	Suppose that the conditions of \Cref{thm:Confidence} are satisfied and let $\cR_1,\cR_2,\ldots$ be any collection of potentially data-dependent constraints. For any level $\alpha\in(0,1)$, we have that
	\[
	\limsup_{n\to \i} P\left(\bigcup_{k \,:\, \mathcal{H}_0(\cR_k) \text{ is true}} \{\psi_{\cR_k}=1\}\right) \leq  \alpha.
	\]
\end{restatable}

\section{Bounding the Identifiability Gap} \label{sec:ScoreGap}
We have seen that 
the identifiability gap, 
that is, the smallest score  
difference 
between 
the causal tree
$\cG$ and 
any alternative graph $\tilde \cG \in \cT_p\setminus \{\cG\}$,
plays an important role when 
identifying causal trees from data.
It provides information about whether the causal graph is identifiable by means of the corresponding score function, and it 
affects 
how quickly the estimation error needs to vanish in order to guarantee consistency, see \Cref{thm:ConsistencyVanishing}.
E.g., for the entropy score, the identifiability gap is given by
\begin{align}  \notag
	\min_{\tilde \cG\in \cT_p\setminus \{\cG\}}	\lE(\tilde \cG) - \lE(\cG) &= \min_{\tilde \cG\in \cT_p\setminus \{\cG\}} \sum_{i=1}^p \lE(\tilde \cG,i) - \lE(\cG,i) \\
	&= \inf_{Q\in \{\tilde \cG\} \times \cF(\tilde \cG) \times \cP^p} D_{\mathrm{KL}}(P_X\| Q), \label{eq:ScoreGapTermWise}
\end{align}
see \Cref{sec:scorefunctions}.

We now analyze the identifiability gap for the entropy score and the Gaussian score in more detail. 
More specifically, we will derive a 
lower bound for the identifiability gap that 
is based on local properties of the underlying structural causal models (such as the ability to reverse edges).
We first consider the special cases of 
bivariate models (Section~\ref{sec:Bivariate}) and
multivariate Markov equivalent trees (Section~\ref{sec:ScoreGapMarkovEquivalentGraphs}) and then turn to general trees (\Cref{sec:ScoreGapGeneralGraphs}).
However, before we venture into the derivation of the specific lower bounds we first examine the connection between the identifiability gaps associated with the different score 
functions.

In this section, 
we 
assume that $X\sim P_X$ is generated by a structural causal additive tree model with $\E\|X\|^2<\i$ such that the local Gaussian, entropy and conditional entropy scores are well-defined. We neither
assume that $\theta$ is a restricted structural causal additive model, i.e., $\theta\in \Theta_R$, nor strict positivity of the identifiability  gap, i.e.,  \Cref{ass:identifiabilityOfConditionalMeanScores}. 
The following result shows that the local node-wise score gaps associated with the different score functions are ordered.
\begin{restatable}[]{lemma}{ScoreOrderings}
	\label{lm:ScoreOrderings}
	For any $\tilde \cG\in \cT_p$ 
	and	for all $i\in V$
	\begin{align} \label{eq:EntropyLocalGapLargerThanCondEntropyLocalGap}
		\lCE(\tilde \cG,i)-\lCE( \cG,i) \leq \lE(\tilde \cG,i)-\lE(\cG,i).
	\end{align}
	If the underlying model is a Gaussian noise model, then 
	\begin{align} \label{eq:GaussianLocalGapLargerThanEntropyLocalGap}
		\lE(\tilde \cG,i)-\lE(\cG,i)  \leq \lG(\tilde \cG,i) -\lG(\cG,i).
	\end{align}	
\end{restatable}
It follows that the full graph score gaps and identifiability gaps associated with the different score functions satisfy a similar ordering. 
Thus, given that the underlying model is Gaussian, a strictly positive entropy identifiability gap implies that the Gaussian identifiability gap is strictly positive. It is, however, not possible to establish strict positivity of the conditional entropy identifiability gap; see \Cref{rmk:ConditionalEntropyRebane} in \Cref{sec:AppDetails}.  Therefore, we focus on establishing a lower bound for the entropy identifiability gap that is tighter than that given by the conditional entropy identifiability gap.

In general, 
we cannot use node-wise comparisons of 
the scores of two graphs to bound 
the 
identifiability gap
(the reason is that in general a node receives a better score in a graph, where it has a parent,
compared to a graph, where it does not;
see \Cref{ex:negativelocalscoregap} in \Cref{sec:AppDetails} for a formal argument).
We start by analyzing the identifiability gap in models with two variables.

\subsection{Bivariate Models} \label{sec:Bivariate}
We now consider two nodes $V=\{X,Y\}$, 
and 
graphs $\cT_2 = \{(X\to Y), (Y\to X)\}$. 
Without loss of generality assume that $(X,Y)\in \cL^2(P)$ is generated by an additive noise SCM $\theta =(\cG,(f_i),P_N)$ with causal graph $\cG= (X\to Y)\in \cT_2$ to which the only alternative graph is  $\tilde \cG= (Y\to X)$. That is,
\begin{align} \label{eq:bivariatesetup}
	X:= N_X, \quad 	Y :=f(X)+N_Y,
\end{align}
where $(N_X,N_Y)\sim P_N \in \cP^2$. %
The bivariate entropy identifiability gap,
which we will later refer to as the edge reversal entropy score gap, 
is defined as
\begin{align*}
	\Delta \lE ( X \lra Y) :&= \lE(\tilde \cG)- \lE(\cG)  \\
	&= h(Y) + h(X-\E[X|Y]) - h(X)-h(Y-\E[Y|X]),
\end{align*}
where the fully drawn arrow symbolizes the true causal relationship and the dashed arrow the alternative. The following lemma simplifies the bivariate entropy identifiability gap to a single mutual information between the  effect and the residual of the minimum mean squared prediction error regression of cause on the effect.
\begin{restatable}[]{lemma}{EdgeReversal}
	\label{lm:EdgeReversal} Consider the bivariate setup of \Cref{eq:bivariatesetup} and assume that $f(X)$ has density. It holds that	\begin{align*}
		\Delta \lE ( X \lra Y) = I(X-\E[X|Y];Y)\geq 0.
	\end{align*}
\end{restatable}
Thus, the causal graph is identified in a bivariate setting if one maintains dependence between the predictor and minimum mean squared error regression residual in the anti-causal direction. This result is in accordance with the previous identifiability results. For example, in the linear Gaussian case, 
$I(X-\E[X|Y];Y) = 0$. 
Consequently, 
the causal graph is not identified from the entropy score function.

Whenever the conditional mean in the anti-causal direction vanishes, e.g., with symmetric causal function and symmetric noise distribution, it is possible to derive a more explicit lower bound with more intuitive sufficient conditions for identifiability of the causal graph. 
\begin{restatable}[]{proposition}{EdgeReversalSymmetric} \label{lm:EdgeReversalSymmetric}
	Consider the bivariate setup of \Cref{eq:bivariatesetup} and assume that $f(X)$ has density. If the reversed direction conditional mean $\E[X|Y]$ almost surely vanishes (e.g., because $f$, $X$ and $N_Y$ are symmetric), then
	\begin{align*}
		\Delta \lE ( X \lra Y) = I(X;f(X)+N_Y),
	\end{align*}
	which is strictly positive if and only if $X\not \independent f(X)+N_Y$. 
	In addition, we have the following statements.
	\begin{itemize}
		\item[(a)] Let $f(X)^\mathrm{G}$ and $N_Y^{\mathrm{G}}$ be independently normally distributed with the same mean and variance as $f(X)$ and $N_Y$, respectively. If $D_{\mathrm{KL}}(f(X)\|f(X)^\mathrm{G}) \leq D_{\mathrm{KL}}(N_Y\|N_Y^\mathrm{G})$, 
		then
		\begin{align*}
			\Delta \lE ( X \lra Y) \geq \frac{1}{2}\log\lp 1+ \frac{\Var(f(X))}{\Var(N_Y)}\rp.
		\end{align*}
		\item[(b)] If the density of $f(X)+N_Y$ 
		is log-concave, then 
		\begin{align*}
			\Delta \lE ( X \lra Y) \geq \frac{1}{2}\log \lp \frac{2}{\pi e} + \frac{2}{\pi e} \frac{\Var(f(X))}{\Var(N_Y)} \rp.
		\end{align*}
		This lower bound is 
		non-trivial only if $$\Var(f(X)) > (\frac{\pi e}{2}-1) \Var(N_Y) \approx 3.27 \Var(N_Y).$$
	\end{itemize}
\end{restatable}

Thus, if the conditional mean $\E[X|Y]$ in the anti-causal direction vanishes, then under certain conditions, the causal direction is identified by the entropy score function (as long as $\Var(f(X))$ is sufficiently large relative to $\Var(N_Y)$).
The edge reversal score gap for the Gaussian score is given by
\begin{align*}
	\Delta \lG( X \lra Y) :=& \, \frac{1}{2}\log \lp \frac{\Var(X-\E[X|Y])}{\Var(X)} \rp - \frac{1}{2}\log \lp\frac{\Var(Y-\E[Y|X])}{\Var(Y)} \rp \\
	=&\, \frac{1}{2}\log \lp \frac{\Var(X-\E[X|Y])}{\Var(X)} \rp + \frac{1}{2}\log \lp 1 + \frac{\Var(f(X))}{\Var(N_Y)} \rp,
\end{align*}
which reduces to the lower bound in point (a) of \Cref{lm:EdgeReversalSymmetric} if the conditional mean $\E[X|Y]$ in the anti-causal direction vanishes.	
\begin{example} Consider the bivariate setup of \Cref{eq:bivariatesetup}.
	Suppose that the causal function $f$ is a quadratic function $f(x)=\alpha x^2 + \beta$ for some $\alpha,\beta,\in\R$ and that $ N_X\sim \cN(0,\sigma_X^2)$ and $N_Y\sim \cN(0,\sigma_Y^2)$. It holds that $E[X|Y]$ vanishes, and the bivariate Gaussian identifiability gap reduces to \begin{align*}
		\Delta \lG( X \lra Y) = \frac{1}{2}\log \lp 1 + 2\alpha^2\frac{ \sigma_X^4}{\sigma_Y^2} \rp.
	\end{align*}
\end{example}
\subsection{Multivariate Markov Equivalent Trees} \label{sec:ScoreGapMarkovEquivalentGraphs}
Two Markov equivalent trees 
differ in precisely  one directed path 
that is reversed in one graph relative to the other.\footnote{To see this, note that any two directed trees are Markov equivalent if and only if they satisfy the exact same $d$-separations or equivalently they share the same skeleton (there are no v-structures in directed trees). Distinct directed trees sharing the same skeleton must have distinct root nodes. Consequently, there exist a directed path in $\cG$ from $\root{\cG}$ to $\root{\tilde \cG}$ that is reversed in $\tilde \cG$; see also \Cref{lm:MarkovEquivTreesPathReversal}}
The entropy score gap of 
Markov equivalent trees 
therefore reduces to the binary case. 
\begin{restatable}[]{proposition}{MarkovEquivTreesScoreGap}
	\label{lm:MarkovEquivTreesScoreGap}
	Consider any $ \tilde \cG \in \cT_p\setminus \{\cG\}$ that is Markov equivalent to the causal tree~$\cG$. 
	Let 
	$c_1\to \cdots \to c_r$ be the unique directed path in $\cG$ that is reversed in $\tilde \cG$. 
	Then
	\begin{align*}
		\lE(\tilde \cG) - \lE(\cG) & = \sum_{i=1}^{r-1} \Delta \lE (c_i\lra c_{i+1}) \geq \min_{1\leq i \leq r-1} \Delta \lE (c_i\lra c_{i+1}).
	\end{align*}
\end{restatable}
Thus, a lower bound of the entropy score gap that holds uniformly over the Markov equivalence class is given by the smallest possible edge reversal in the causal directed graph:
\begin{align*}
	\min_{\tilde \cG \in \mathrm{MEC}(\cG)\setminus \{\cG\}} \lE( \tilde \cG) - \lE(\cG) \geq \min_{(j\to i)\in \cE} \Delta \lE ( j \lra i).
\end{align*}

\subsection{General Multivariate Trees } \label{sec:ScoreGapGeneralGraphs}
We now derive a lower bound of the entropy identifiability gap, i.e., a lower bound of the entropy score gap that holds uniformly over all alternative trees $\cT_p\setminus \{\cG\}$. 
To do so, we exploit a graph reduction technique \citep[introduced by][]{peters2014causal} which 
enables us to reduce the analysis %
to 
three distinct scenarios.  This graph reduction works as follows.
Fix any alternative graph $\tilde \cG\in \cT_p\setminus \{\cG\}$, 
and iteratively remove any node (from both $\cG$ and $\tilde \cG$)
that has no children and the same parents in both $\cG$ and $\tilde \cG$. 
The score gap is unaffected by the graph reduction.%
\footnote{All removed nodes $V\setminus V_R$ have identical incoming edges in both graphs and therefore have identical local scores. That is, for any loss function $l\in\{\lCE,\lE,\lG\}$ we have that
	$l(\tilde \cG) - \ell(\cG) = \sum_{i\in V_R} \ell(\tilde \cG,i) - \ell(\cG,i) + \sum_{i\in V\setminus V_R} \ell(\tilde \cG,i) - \ell(\cG,i) 
	= \sum_{i\in V_R} \ell(\tilde \cG,i) - \ell(\cG,i) = \ell(\tilde \cG_R) -l(\cG_R)$.
}

\begin{figure}[t] 
	\begin{center}
		\begin{tabular} {lr}
			\begin{tabular}{l}
				\begin{tikzpicture}[node distance = 1.15cm, roundnode/.style={circle, draw=black, fill=gray!10, thick, minimum size=7mm},
					roundnode/.style={circle, draw=black, fill=gray!10, thick, minimum size=7mm},
					outer/.style={draw=gray,dashed,fill=black!1,thick,inner sep=2pt}
					]
					\node[roundnode] (W) [] {W};
					\node[roundnode] (Y) [right = of W] {Y};
					\node[roundnode] (Z) [right = of Y] {Z};
					\node[roundnode] (L) [below = 0.8cm of Y] {L};
					\node[outer] (ANW) [above = 0.8cm of W] {$\ANg{\cG_R}{W}$};
					\node[outer] (ANY) [above = 0.8cm of Y] {$\ANg{\cG_R}{Y}$};
					\node[outer] (ANZ) [above = 0.8cm of Z] {$\ANg{\cG_R}{Z}$};
					\node[roundnode,fill opacity= 0, draw  opacity =0] (dummy2) [below = of L] {};
					\node[above,font=\large ] at (current bounding box.north) {subgraph of  $\cG_R$};
					
					\draw[->, line width=0.4mm] (W) -- (L);
					\draw[->, line width=0.4mm] (Y) -- (L);
					\draw[->, line width=0.4mm] (Z) -- (L);
					\draw[->, line width=0.4mm] (ANW) -- (W);
					\draw[->, line width=0.4mm] (ANY) -- (Y);
					\draw[->, line width=0.4mm] (ANZ) -- (Z);
					
				\end{tikzpicture}
			\end{tabular}
			&
			\begin{tabular}{l}
				\begin{tikzpicture}[node distance = 1.15cm, roundnode/.style={circle, draw=black, fill=gray!10, thick, minimum size=9mm},
					roundnode/.style={circle, draw=black, fill=gray!10, thick, minimum size=7mm,},
					outer/.style={draw=gray,dashed,fill=black!1,thick,inner sep=2pt}
					]
					\node[roundnode] (D) [] {D};
					\node[] (dummy) [right = of D] {};
					\node[roundnode] (L) [right = of D] {L};
					\node[roundnode] (Z) [right = of L] {Z};
					\node[roundnode,fill opacity= 0, draw  opacity =0] (dummy2) [below = 0.8cm of L] {};
					\node[roundnode] (Y) [left = of dummy2] {Y};
					\node[roundnode,label=center:$O_1$] (O1) [below = 0.8cm of L] {};
					\node[roundnode,label=center:$O_k$] (Ok) [right = of O1] {};
					\node[outer] (AND) [above  = 0.8cm of D] {$\ANg{\tilde \cG_R}{D}$};
					\node[outer] (ANZ) [above = 0.8cm of Z] {$\ANg{\tilde \cG_R}{Z}$};
					\node[outer] (DEY) [below = 0.8cm of Y] {$\DEg{\tilde \cG_R}{Y}$};
					\node[outer] (DEO1) [below = 0.8cm of O1] {$\DEg{\tilde \cG_R}{O_1}$};
					\node[outer] (DEOk) [below = 0.8cm of Ok] {$\DEg{\tilde \cG_R}{O_k}$};
					\node[above,font=\large ] at (current bounding box.north) {subgraph of  $\tilde \cG_R$};
					\draw[->, line width=0.4mm] (D) -- (L);
					\draw[->, line width=0.4mm] (Z) -- (L);
					\draw[->, line width=0.4mm] (L) -- (Y);
					\draw[->, line width=0.4mm] (L) -- (O1);
					\draw[->, line width=0.4mm] (L) -- (Ok);
					\draw[->, line width=0.4mm] (AND) -- (D);
					\draw[->, line width=0.4mm] (ANZ) -- (Z);
					\draw[->, line width=0.4mm] (Y) -- (DEY);
					\draw[->, line width=0.4mm] (O1) -- (DEO1);
					\draw[->, line width=0.4mm] (Ok) -- (DEOk);
					\draw[line width=2pt, line cap=round, dash pattern=on 0pt off 5\pgflinewidth,shorten <=5pt] (O1) -- (Ok);
				\end{tikzpicture} 
			\end{tabular}
		\end{tabular}
	\end{center}
	\caption{Schematic illustration of parts of two reduced graphs produced by the graph reduction technique described in \Cref{sec:ScoreGapGeneralGraphs}. 
		Consider a sink node $L$ in $\cG_R$. Its parent (in $\cG_R$) must either be 
		a parent in $\tilde{\cG}_R$, too,
		it must be a child in $\tilde{\cG}_R$,
		or it is unconnected to $L$ in $\tilde{\cG}_R$.
		Thus, exactly one of the sets 
		$Z$, $Y$, and $W$ is non-empty.
		This case distinction is used to compute the three bounds in \Cref{lm:thm:scoreGapEntropy}.
		$D$, $O_1, \ldots, O_k$ denote further (possibly existing) nodes in $\tilde{\cG}_R$. 
	}
	\label{fig:reducedSubgraphs}
\end{figure}
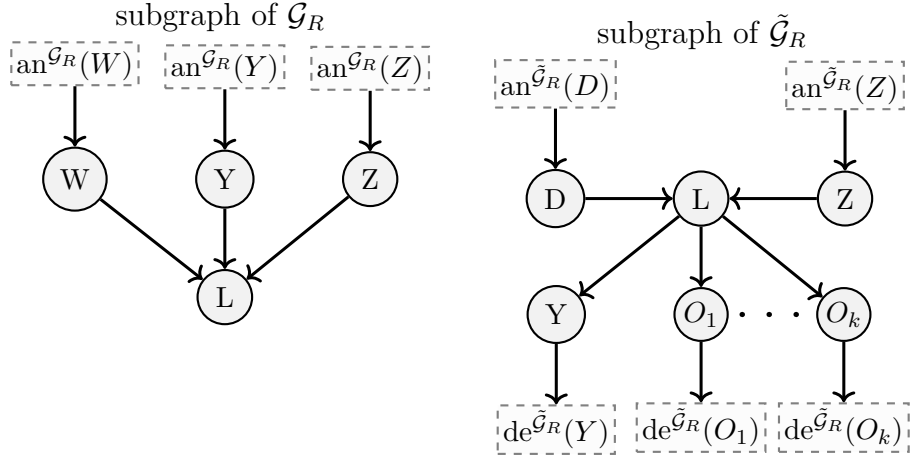

Applying 
this iteration scheme,
until no such node can be found,
results in two 
reduced 
graphs $\cG_R=(V_R,\cE_R)$ and $\tilde \cG_R=(V_R,\tilde \cE_R)$. 
These reduced graphs cannot be empty, for that would only happen if $\tilde \cG = \cG$.
Further, they have identical vertices 
but different edges. 
And they can be categorized into one of three cases.
To do so, consider a node $L$ that is a sink node,  i.e., a node without children, in $\cG_R$ and consider its parent in $\cG_R$.
Now, considering $\tilde{\cG}_R$, one of the following conditions must hold: 
the parent is also a parent of $L$ in $\tilde{\cG}_R$ (we then call it $Z$),
the parent is not connected to $L$ in $\tilde{\cG}_R$ (we then call it $W$), or
the parent is a child of $L$ in $\tilde{\cG}_R$ (we then call it $Y$). 
Figure~\ref{fig:reducedSubgraphs} visualizes these three scenarios. 

We can now obtain bounds for each of the three case individually.
For the case with a node $Z$ (a `staying parent'), define 
$$
\Pi_Z(\cG) := \left\{(z,l,o)\in V^3 \text{ s.t.\ }  (z\to l)\in \cE \text{ and }
o\in \NDg{\cG}{l}\setminus\{z,l\} \right\}.
$$
The score gap can then be lower bounded by 
$\min_{(z,l,o)\in \Pi_Z(\cG)} I(X_z;X_o|X_l)$
(see \Cref{lm:ScoreGapCaseOne}).
Intuitively, 
$I(X_z;X_o|X_l)$ quantifies the strength of the connection between $z$ and $o$, when conditioning on $l$ (which does not lie on the path between $z$ and $o$). This is a non-local bound in that it does not constrain the length of the path connecting $z$ and $o$. Analyzing or bounding this term might be difficult.
We will see in \ref{sec:GaussianLocalization} that this part is not needed
in the Gaussian case.

For the case with a node $W$ (`removing parent'), define
$$
\Pi_W(\cG) := \left\{(w,l,o)\in V^3 \text{ s.t.\ }  (w\to l)\in \cE \text{ and } 
o\in (\CHg{\cG}{w} \setminus \{l\})\cup \PAg{\cG}{w}\right\}.
$$
This case results in the lower bound
$\min_{(w,l,o)\in \Pi_W(\cG)} I(X_w;X_l|X_o)$
(see \Cref{lm:ScoreGapCaseTwo}).
Here, $w$ is a parent of $l$ and $o$ is directly connected to $w$.
Intuitively, 
$I(X_w;X_l|X_o)$ quantifies the strength of the edge $w \rightarrow l$. We condition on $o$ but that node is not directly connected to $l$ (only via $w$).
For the first two cases, faithfulness \citep{spirtes2000causation}
implies that these terms are non-zero
and bounding them away from zero reminds of strong faithfulness \citep{ZhangStrongFaithful}. 
However, in the second case, one considers individual edges, which reminds more  of
a strong version of causal minimality
\citep{spirtes2000causation, Peters2017}.

For the case with a node $Y$ (`parent to child'), 
a lower bound is given by the minimal edge reversal score gap
$\min_{(j\to i) \in \cE} \Delta \lE(j \lra i)$
(see \Cref{lm:ScoreGapCaseThree}).
The term $\Delta \lE(j \lra i)$ measures the identifiability of the direction of an individual edge. It is zero in the linear Gaussian case, for example. 
We provide more details on the reduced graphs and on the arguments in the three cases in Section~\ref{sec:moredetailsgraphreduction} of \Cref{app:treesproofs}.

Combining the three bounds from above, we obtain the following theorem. 
\begin{restatable}[]{theorem}{thm:scoreGapEntropy}
	\label{lm:thm:scoreGapEntropy}
	It holds that
	\begin{align} \notag
		\min_{\tilde \cG \in \cT_p\setminus \{\cG\}}\lE(\tilde \cG) -\lE(\cG) \geq \min\bigg\{
		& \min_{(z,l,o)\in \Pi_Z(\cG)} I(X_z;X_o|X_l), \\
		& \min_{(w,l,o)\in \Pi_W(\cG)} I(X_w;X_l|X_o), \notag \\
		& \min_{(j\to i) \in \cE} \Delta \lE(j \lra i) \bigg\}. \label{eq:EntropyScoreGapUniformLowerBound}
	\end{align}
\end{restatable}
This result lower bounds the identifiability gap using information-theoretic quantities. 
Corresponding results for the Gaussian score follow immediately 
by \Cref{lm:ScoreOrderings}. 
The last two terms are local properties of the underlying structural causal model; the first term is not.
As seen in Section~\ref{sec:ScoreGapMarkovEquivalentGraphs}, the last term on the right-hand side is required when considering only Markov equivalent trees; if it is non-zero, it allows us to orient all edges in the skeleton. The first two terms (non-zero under faithfulness) are additionally required when the considered trees are not Markov equivalent.

We now turn to the case of Gaussian trees. Here, the first term is not needed; the bound then depends only on local properties of the structural causal model.

\subsection{Gaussian Multivariate Trees} \label{sec:GaussianLocalization}
The score gap lower bound in  \Cref{eq:EntropyScoreGapUniformLowerBound} consists of local dependence properties except for the node tuples $\Pi_Z(\cG)$ (\Cref{lm:ScoreGapCaseOne}) that arise when considering alternative graphs 
that result in reduced graphs with a node $Z$ (`staying parents').
However, we show that in the Gaussian case, 
the score gap for such 
alternative graphs 
can be lower bounded by the score gaps already considered in 
alternative graphs with a node $Y$ (`parent to child') and a node $W$ (`removing parent').
Thus, we have the following theorem, 
with a bound 
consisting only of local %
properties of the model.
\begin{restatable}[Gaussian localization of the identifiability gap]{theorem}{GaussScoreGapCaseOne}
	\label{thm:GaussScoreGapCaseOne}
	In a Gaussian setting (see \Cref{sec:scorefunctions}), we have
	\begin{align*}
		\min_{\tilde \cG \in \cT_p\setminus \{\cG\}}\lG(\tilde \cG) -\lG(\cG)&\geq \min\left\{ \min_{(w,l,o)\in \Pi_W(\cG)} I(X_w; X_l\, | \, X_o),\min_{(j\to i) \in \cE} \Delta \lE( j \lra i)\right\}.
	\end{align*}	
\end{restatable}

\section{Simulation Experiments} \label{sec:Simulations}
In this section, we investigate the finite-sample performance of CAT and perform simulation experiments investigating the identifiability gap and its lower bound. In \Cref{sec:StructureLearningForTrees} we compare the performance of CAT to CAM of \cite{CAM} for Gaussian and non-Gaussian additive noise models with causal graphs given by directed trees.  In \Cref{sec:ExperimentIdentifiabilityConstant} we perform simulation experiments that highlight the behavior of the identifiability gap and its corresponding lower bound derived in \Cref{sec:ScoreGap}. In \Cref{sec:ExperimentCATonDAGs} we compare the CAT and CAM for causal discovery on non-tree DAG models (CAT always outputs a directed tree).
The code scripts (R) for the simulation experiments and an implementation of CAT is available on GitHub.\footnote{\url{https://github.com/MartinEmilJakobsen/CAT}}

\subsection{Causal Structure Learning for Trees}\label{sec:StructureLearningForTrees} 
In this section, we compare the performance of the structure learning methods CAT and CAM when employed on additive noise  models with causal graphs given by directed trees. 
\subsubsection{Tree Generation Schemes}
\label{sec:TreeGenerationSchemes}
\begin{figure}
	\includegraphics[width=\textwidth-20pt]{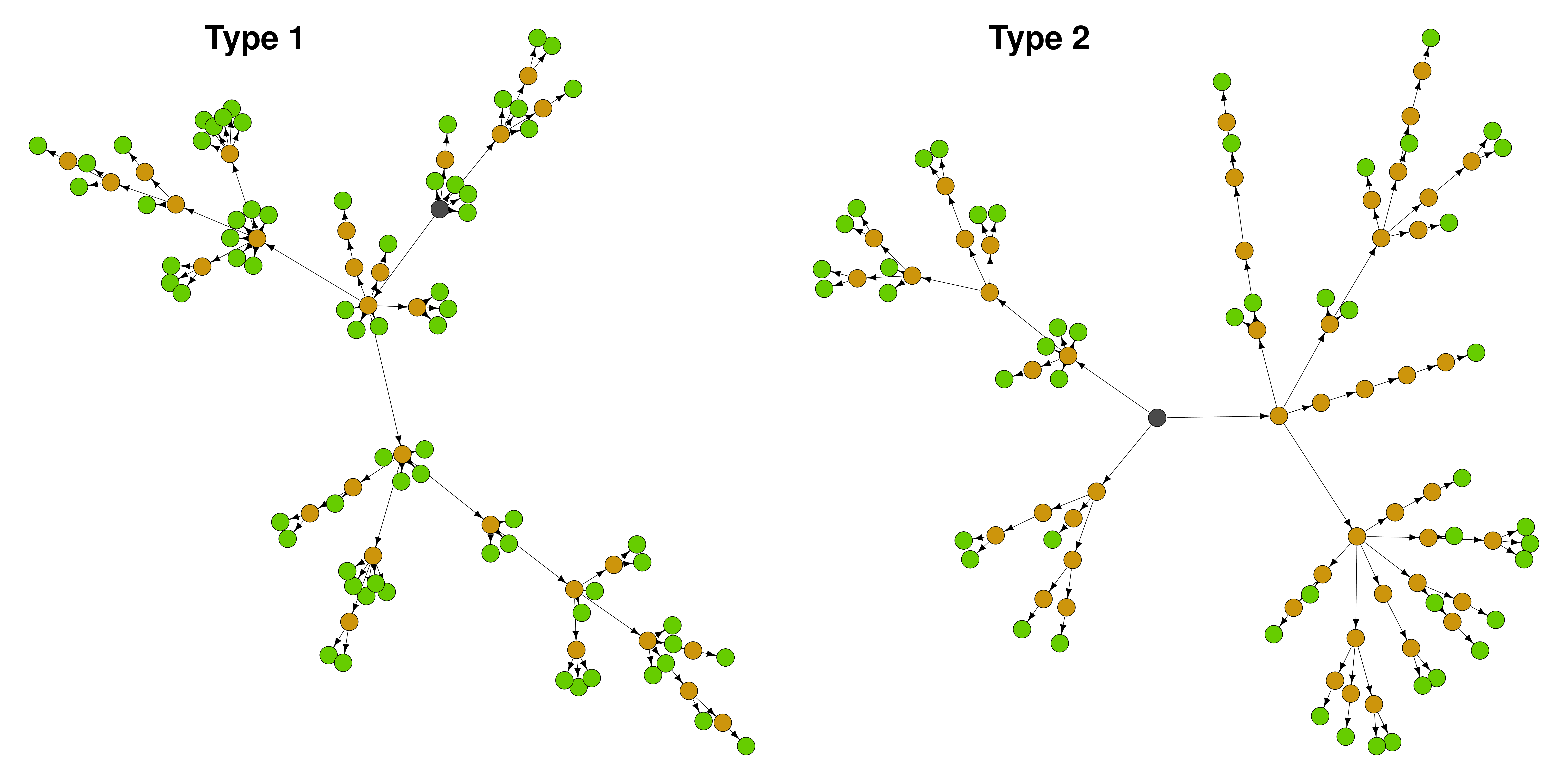}
	\caption{Illustration of Type 1 (many leaf nodes) and Type 2 (many branch nodes) directed trees over $p=100$ nodes. The green nodes are leaf nodes, the brown nodes are branch nodes, and the black nodes are root nodes. The Type 1 tree contains 70 leaf nodes, while the Type 2 tree only contains 49 leaf nodes.} \label{fig:TreeTypes}
\end{figure}

We employ two different random directed tree generation schemes: Type 1 (many leaf nodes) and Type 2 (many branch nodes). 
In \Cref{fig:TreeTypes} we have illustrated two directed trees generated in accordance with the two generation schemes. 
For more details, see   \Cref{alg:type1tree,alg:type2tree} in Section \ref{app:TreeGeneration} of \Cref{app:Experiments}.

\subsubsection{Gaussian Experiment} \label{sec:ExperimentGaussianTrees}
In this experiment, we generate data similarly to the experimental setup of \cite{CAM}. For any given directed tree we generate causal functions by sample paths of Gaussian processes with radial basis function (RBF) kernel and bandwidth parameter of one. Sample paths of Gaussian processes with radial basis function kernels are almost surely infinitely continuous differentiable \citep[e.g.,][]{kanagawa2018gaussian}, non-constant and nonlinear, so they satisfy the requirements of \Cref{lm:RestrictedModelConditionGaussian}. See \Cref{fig:samplepaths} in Section \ref{sec:additionalIllustrations} of \Cref{app:Experiments} for illustrations of random draws of such functions. Root nodes are mean zero Gaussian variables with standard deviation sampled uniformly on $(1,2)$. Furthermore, for each fixed tree and set of causal functions, we introduce at each non-root node additive Gaussian noise with mean zero and standard deviation sampled uniformly on $(1/5,\sqrt{2}/5)$.

We first compare our method CAT with Gaussian score function (CAT.G) against the method CAM of \cite{CAM} on the previously detailed nonlinear additive Gaussian noise tree setup. 
We implement CAT.G without sample-splitting and use the R-package \verb!GAM! \citep[Generalized Additive Models,][]{hastie2020package} with default settings to construct a thin plate regression spline estimate of the conditional expectations. We use the implementation of  Chu--Liu--Edmonds' algorithm from the R-package \verb!RBGL!.\footnote{The RBGL implementation finds maximum edge weight directed trees and requires all positive edge weights. As such, we take the negative of our edge weights  and shift them all by the absolute value of smallest edge-weight. If an edge weight is set to zero this edge can not be chosen. \label{footnode:EdmondsRBGL}}  
CAM is employed with a maximum number of parents set to one (restricting the output to directed trees), without preliminary neighborhood selection and subsequent pruning. We measure the performance of the methods by computing the Structural Hamming Distance \citep[SHD,][]{tsamardinos2006max} and Structural Intervention Distance \citep[SID,][]{peters2015structural} to the causal tree. 
\begin{figure}[t]
	\begin{center}
		\includegraphics[width=\textwidth-20pt]{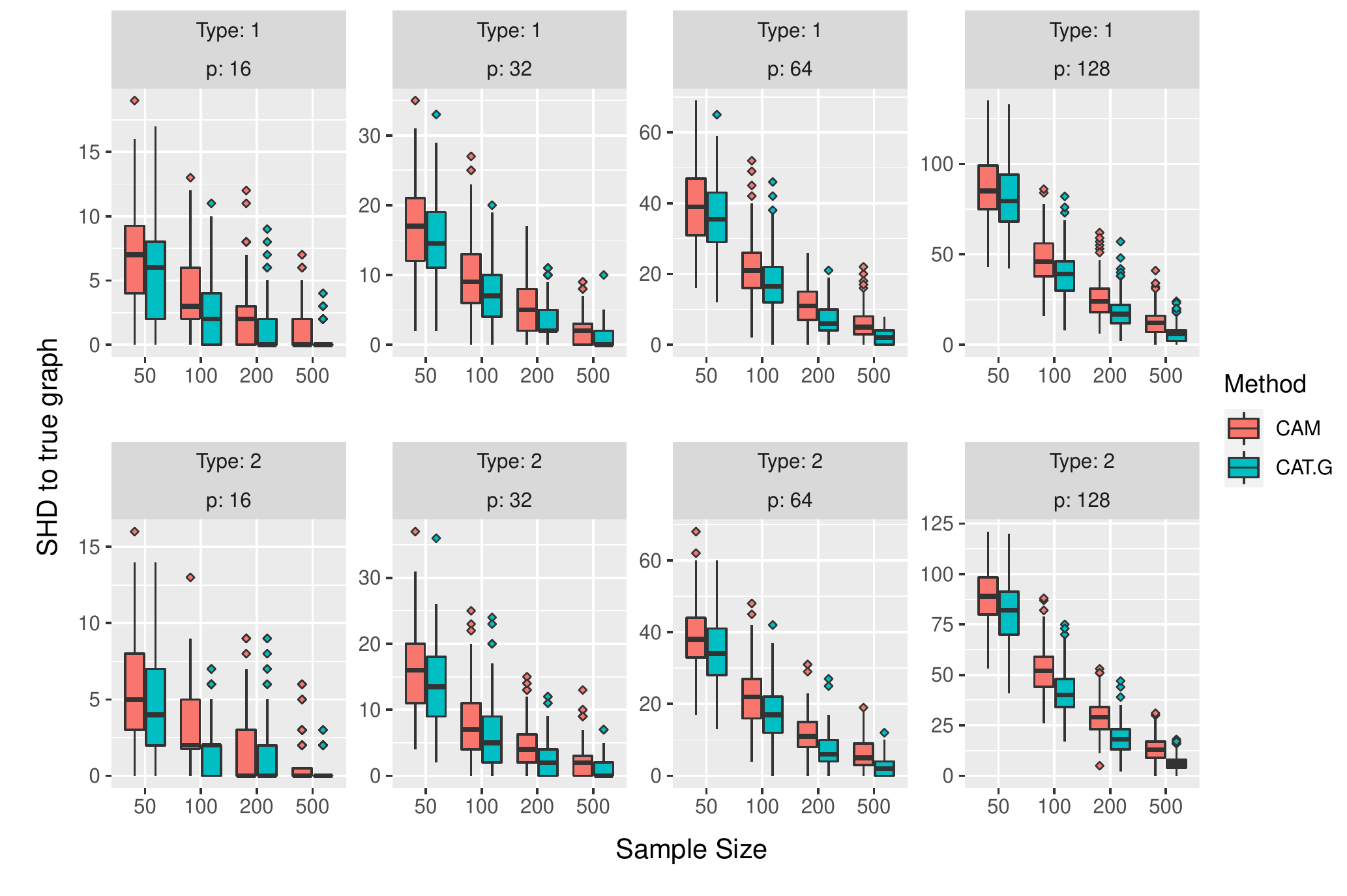}
	\end{center}
	\caption{Gaussian setting: Boxplots of the SHD performance of CAM and CAT.G (Gaussian score) for varying sample sizes, system sizes, and tree types. 
		CAT.G outperforms CAM in a wide range of scenarios.
	}
	\label{fig:boxplot_Gaussian_SHD_Type1and2}
\end{figure}

For each system size $p\in\{16,32,64,128\}$ we generate a causal tree, corresponding causal functions and noise variances and sample  $n\in\{50,100,200,500\}$ observations. This is repeated 200 times and the SHD results are summarized in the boxplot of \Cref{fig:boxplot_Gaussian_SHD_Type1and2}. 
Both methods perform better on trees of Type 2  than on trees of Type 1. 
CAT.G outperforms CAM in terms of SHD to the true graph both in median distance and IQR length and position for all sample sizes, system sizes and tree types. 
Considering the SID to the causal tree yields similar conclusions; see \Cref{fig:boxplot_Gaussian_SID_Type1and2} in Section \ref{sec:additionalIllustrations} of \Cref{app:Experiments}.
In their default versions, CAM and CAT.G use different estimation techniques of the conditional expectations, but this does not seem to be the source of the performance difference: \Cref{fig:boxplot_Gaussian_CamScores} in Section \ref{sec:additionalIllustrations} of \Cref{app:Experiments} illustrates a similar SHD performance difference when forcing CAT.G to use the edge weights  produced by the CAM implementation.

\subsubsection{Non-Gaussian Experiment}
\label{sec:NonGaussianExperiment}

\begin{figure}[t]
	\begin{center}
		\includegraphics[width=\textwidth-20pt]{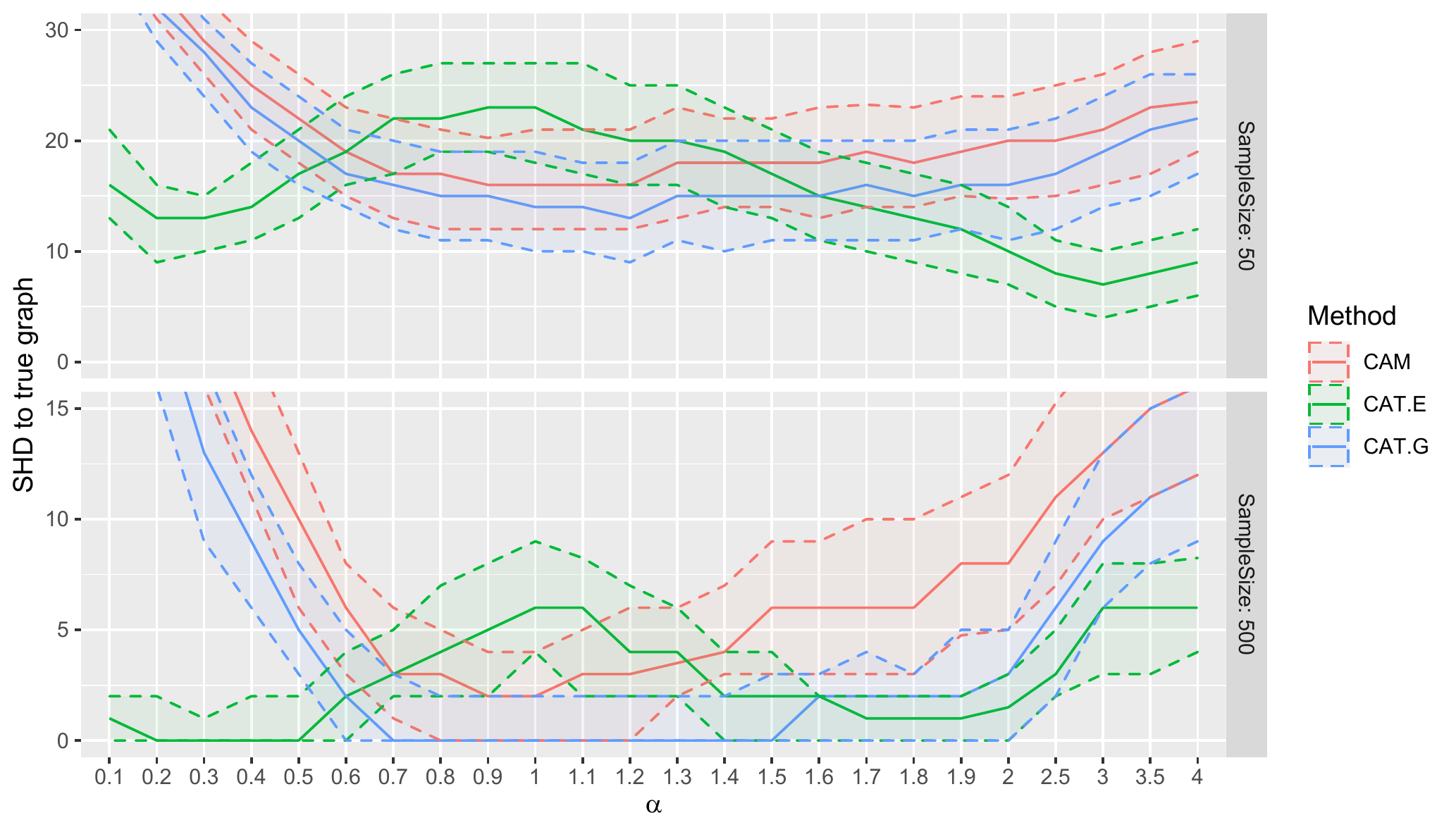}
	\end{center}
	\caption{Deviations from Gaussianity: 
		The parameter $\alpha$ controls the noise deviation from the Gaussian distribution. 
		CAT.G and CAT.E are instances of CAT with edge weights  derived from Gaussian and entropy score functions, respectively. 	
		The solid lines represent the median SHD and the shaded (dashed) region represents the interquartile range. Using the entropy score yields better results for noise distributions that deviate strongly from Gaussian noise.}
	\label{fig:boxplot_NonGaussian_SHD}
\end{figure}

We  now compare the performance of CAM and CAT with Gaussian (CAT.G) and entropy (CAT.E) score functions in a setup with varying noise distributions. 
The entropy edge weights  used by CAT.E are estimated with the differential entropy estimator of \cite{10.1214/18-AOS1688} as implemented in the CRAN R-package \verb!IndepTest!  \citep[][]{IndepTest}. 
We use the same simulation setup as in  \Cref{sec:ExperimentGaussianTrees}  but now we only consider trees of Type~1 and parameterize the setup by $\alpha >0$, which controls the deviation of the additive noise innovations from a Gaussian distribution. More precisely, we generate the additive noise variables $N_i(\alpha)$ 
as
\begin{align*}
	N_i(\alpha) = \mathrm{sign}(Z_i)|Z_i|^\alpha,
\end{align*}
where $Z_i \sim \cN(0,\sigma_i^2)$ with 
$\sigma_i$ sampled uniformly on $(1/5,\sqrt{2}/5)$ or uniformly on $(1,2)$ if $i=\root{\cG}$. For $\alpha=1$ this yields
Gaussian noise, while for alpha $\alpha\not= 1$ the noise is non-Gaussian. We conduct the experiment for all  $\alpha \in \{0.1,0.2,\ldots,2,2.5,3,3.5,4\}$ and sample sizes $n\in\{50,500\}$ for a fixed system size of $p=32$. Each setting is repeated 500 times and the results are illustrated in \Cref{fig:boxplot_NonGaussian_SHD}.

For Gaussian noise, both CAM and CAT.G outperform CAT.E. This can (at least) be attributed to two factors: (i) CAT.E does not, unlike CAM and CAT.G, explicitly use the Gaussian noise specification and (ii) differential entropy estimation is a difficult statistical problem \citep[see, e.g.,][]{paninski2003estimation,Yanjun2020} 
For small and moderate deviations from Gaussianity,
CAT.G outperforms both CAM and CAT.E.
For larger  deviations, 
CAT.E outperforms both CAT.G and CAM in terms of median SHD.  Finally, we note that CAT.G always outperforms CAM in terms of median SHD.

\subsection{Identifiability Gap} \label{sec:ExperimentIdentifiabilityConstant}
We now investigate the behavior of the identifiability gap 
in bivariate models
(\Cref{sec:ExperimentBivariateIdentifiabilityConstant})
and 
evalute the lower bound derived in \Cref{sec:ScoreGap} empirically for multivariate models (\Cref{sec:ExperimentMultiVariateIdentifiabilityConstant}). 
\subsubsection{Bivariate Identifiability Gap} \label{sec:ExperimentBivariateIdentifiabilityConstant}
In this experiment, we 
investigate the behavior of the bivariate identifiability gap and 
analyze both
a Gaussian and a non-Gaussian setup. Let us consider an additive noise model over $(X,Y)$ with causal graph $X\to Y$. 
The causal functions will be chosen from the following function class. 
For any $\lambda\in[0,1]$,
define 
$f_{\lambda}:\R\to \R$ 
as
\begin{align*}
	f_{\lambda}(x) = (1-\lambda)x^3 + \lambda x.
\end{align*}
That is,  $\lambda \mapsto f_\lambda$ interpolates between a cubic function $x\mapsto x^3$ and a linear function $x\mapsto x$.  For any $(\alpha,\lambda)\in(0,\infty)\times [0,1]$ we consider the following bivariate structural causal additive model
\begin{align*}
	X:= \mathrm{sign}(N_X)|N_X|^\alpha, \quad Y:= f_{\lambda}(X)+ N_Y,
\end{align*}
where $N_X,N_Y$ are independent standard normal distributed random variables. Recall that the bivariate identifiability gap is given by
\begin{align}
	\lE(Y\to X)- \lE(X\to Y) =&\, h(X-\E[X|Y])+h(Y)-h(X-\E[X|Y],Y) \notag\\
	=&\,I(X-\E[X|Y];Y), \label{eq.IG}
\end{align}
by \Cref{lm:EdgeReversal}. 
Thus, the causal graph $X\to Y$ is identified by the entropy score function if $I(X-\E[X|Y];Y)>0$.

For any fixed $\lambda$ and $\alpha$ we 
now
estimate the identifiability gap; we also 
calculate the $p$-value associated with the null hypothesis that the identifiability gap is zero (based on 50000 observations). 
Similarly to the previous experiment, we estimate the conditional expectations using GAM. We estimate (without sample splitting) the identifiability gap  and construct $p$-values using the CRAN R-package \verb!IndepTest!
\citep[][]{IndepTest}. More specifically, we use the differential entropy estimator of \cite{10.1214/18-AOS1688}  and the mutual information based independence test of \cite{10.1093/biomet/asz024}, respectively.

\begin{figure}
	\begin{center}
		\includegraphics[width=\textwidth-20pt]{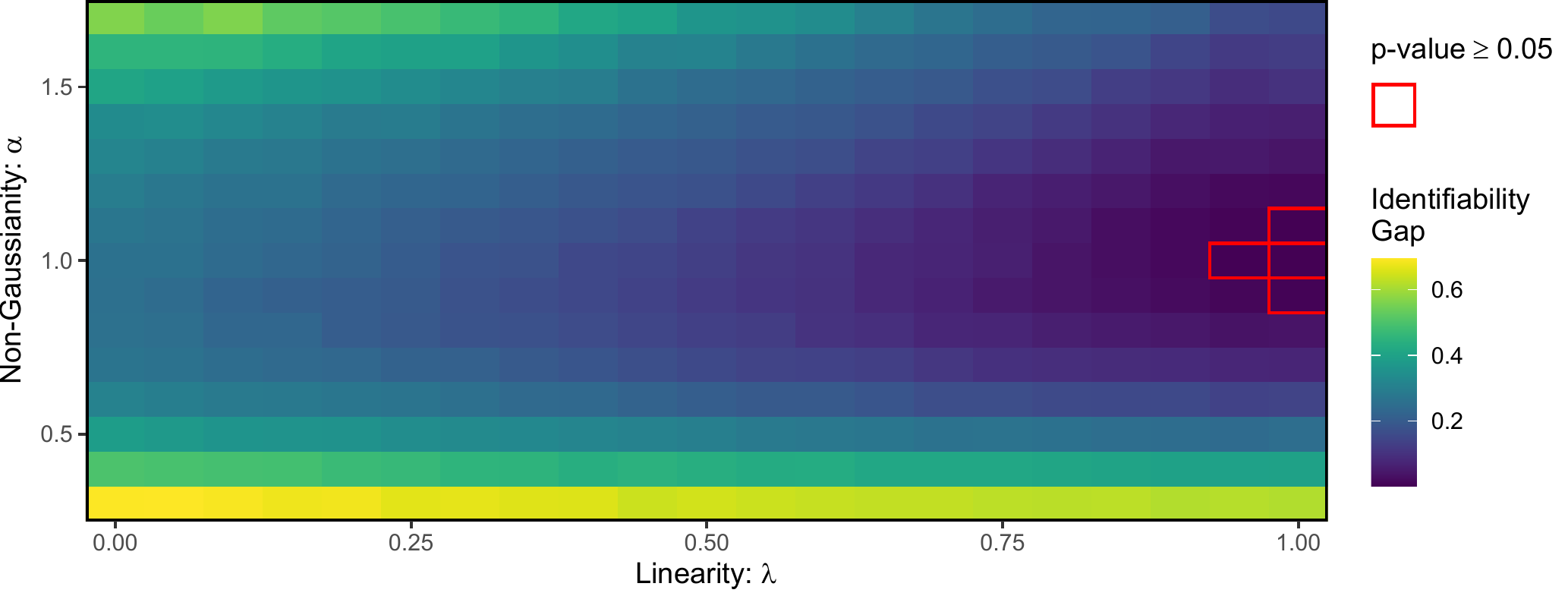}
	\end{center}
	\caption{Heatmap of the identifiability gap for varying $\lambda$ and $\alpha$. Tiles with a red boundary correspond to the models for which the mutual information based independence test cannot reject the null hypothesis of a vanishing identifiability gap.} \label{fig:IdentifiabilityConstant_Heatmap}
\end{figure}

The heatmap of \Cref{fig:IdentifiabilityConstant_Heatmap} illustrates the behavior of the identifiability gap for all combinations of $\lambda\in\{0,0.05,\ldots,1\}$ and $\alpha \in \{0.3,0.4,\ldots,1.7\}$. It suggests that the identifiability gap only tends to zero when we approach the linear Gaussian setup. Only in the models closest to the linear Gaussian setup are we unable to reject the null-hypothesis of a vanishing identifiability gap.

This is also what the theory predicts, namely that for bivariate linear Gaussian additive models, the causal direction is not identified. 
It is known that for linear models, non-Gaussianity is helpful for identifiability. The empirical results indicate that the same holds for nonlinear models, i.e., that the identifiability gap increases with the degree of non-Gaussianity.
\subsubsection{Multivariate Identifiability Gap}
\label{sec:ExperimentMultiVariateIdentifiabilityConstant}
In this experiment, we investigate the identifiability gap and its relation to the lower bounds established in \Cref{thm:GaussScoreGapCaseOne}. For a Gaussian additive noise tree model, it holds that
\begin{align*}
	\min_{\tilde \cG \in \cT_p\setminus \{\cG\}}\lG(\tilde \cG) -\lG(\cG)&\geq \min\left\{\min_{(w,l,o)\in \Pi_W(\cG)} I(X_w; X_l\, | \, X_o),\min_{i\to j \in \cE} \Delta \lE( i \lra j)\right\}.
\end{align*}
In other words, the identifiability gap is lower bounded by the minimum of the smallest local faithfulness measures and the smallest edge-reversal score difference. 
We now investigate empirically
how important the first term is 
for the inequality to hold.  
More specifically, for a given model generation scheme, we quantify how often the minimum  edge reversal is sufficiently small to establish the lower bound without the conditional mutual information term, that is, how often the identifiability  constant $\min_{\tilde \cG \in \cT_p\setminus \{\cG\}} \lG(\tilde{\cG})-\lG(\cG)$
is larger than the minimum edge reversal. 

The minimum edge reversal can be estimated using the same conditional expectation and entropy estimators of the experiment in \Cref{sec:ExperimentBivariateIdentifiabilityConstant}. However, estimating the identifiability gap between the second-best scoring tree and the causal tree needs further elaboration. We know that the best scoring (causal) tree  can be found by Chu--Liu--Edmonds' (a directed MWST) algorithm. The second-best scoring tree differs from the best scoring tree in at least one edge. Thus, given the best scoring graph, we remove one of the $p-1$ edges of the best scoring tree from the pool of possible edges and rerun Chu--Liu--Edmonds' algorithm. We do this for each of the $p-1$ edges in the best scoring tree which leaves us with $p-1$ possibly different sub-optimal trees of which the minimum score is attained by the second-best scoring graph.

\begin{figure}
	\begin{center}
		\includegraphics[width=\textwidth-20pt]{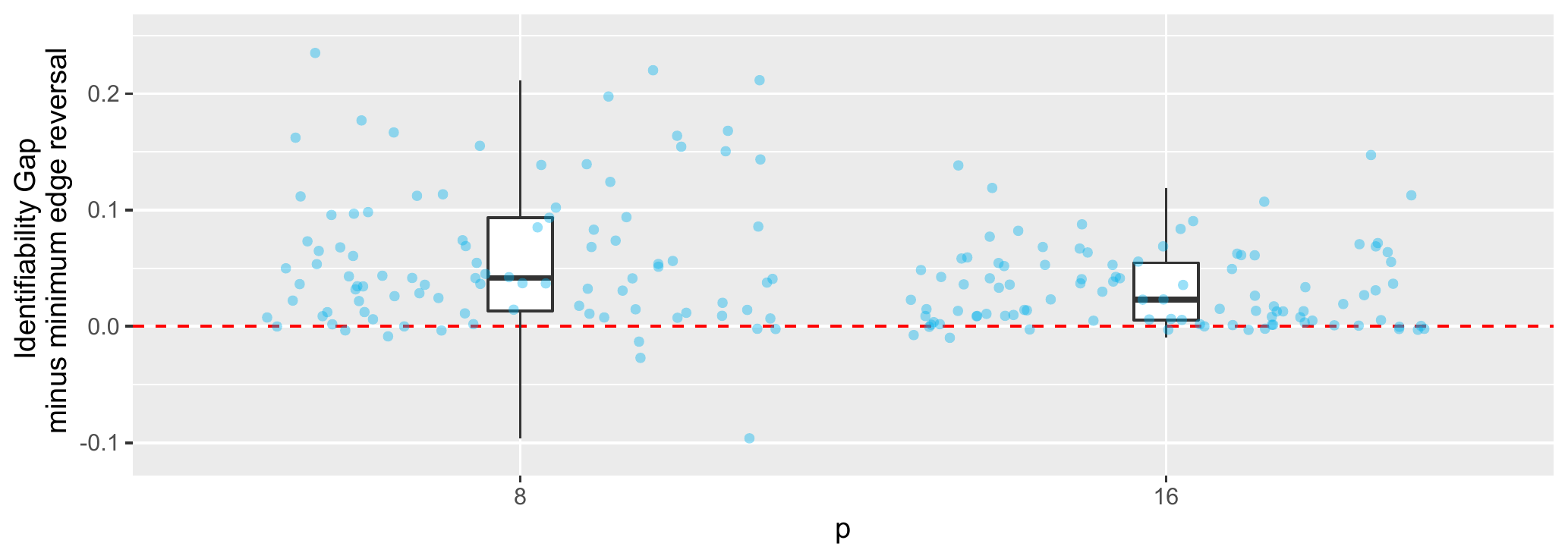}
	\end{center}
	\caption{Empirical analysis of the lower bound on the identifiability gap, see
		\Cref{sec:ExperimentMultiVariateIdentifiabilityConstant}. 
		In most of the simulated settings, we see that the estimated identifiability gap is larger than the smallest edge-reversal score difference. This suggests that in many cases, the latter term is sufficient for establishing a lower bound on the identifiability gap.
	} \label{fig:IdentifiabilityConstant_Multivariate}
\end{figure}

For the experiment,
we randomly sample data generating models similarly to the experiment in \Cref{sec:ExperimentGaussianTrees}. However, we change the causal functions from explicit sample paths of a Gaussian process to a GAM model estimating the sample paths due to memory constraints when generating large sample sizes.
\Cref{fig:IdentifiabilityConstant_Multivariate} illustrates, for  $p\in\{8,16\}$, boxplots of the difference between the  identifiability gap and the minimum edge reversal for 100 randomly generated Gaussian additive noise tree models. For each model, the identifiability gap and corresponding minimum edge reversal is estimated from 200000 independent and identically distributed observations. The illustration suggests that it is in general necessary to also consider the conditional mutual information term  in order to establish a lower bound. However, it also shows that in the majority (90\%) of the models, the minimum edge reversal is indeed a lower bound for the identifiability gap. 

\subsection{Robustness: CAT on DAGs} \label{sec:ExperimentCATonDAGs}
This experiment analyzes how CAT performs compared to CAM when applied to data generated from a Gaussian additive model with a non-tree DAG as a causal graph. More specifically, we analyze the behavior on single-rooted DAGs. For any fixed $p\in \N$ we generate a directed tree of Type 1 and for each zero in the upper triangular part of the adjacency matrix we add an edge with 5\% probability. 
The causal functions and Gaussian noise innovations are generated according to the specifications given in the experiment of \Cref{sec:ExperimentMultiVariateIdentifiabilityConstant}. 
The structural assignment for each node is 
additive in each causal parent, 
i.e., for all $i\in\{1,\ldots,p\}$, $X_i := \sum_{j\in \PAg{\cG}{i}} f_{ji}(X_j) + N_i$, 
with $(N_1,\ldots,N_p)$ mutually independent Gaussian distributed noise innovations. For each $p\in\{16,32,64\}$ and sample size $n\in \{50,250,500\}$ we randomly generate 100 single-rooted Gaussian additive models according to the above specifications.

\begin{figure}[t]
	\begin{center}
		\includegraphics[width=\textwidth-20pt]{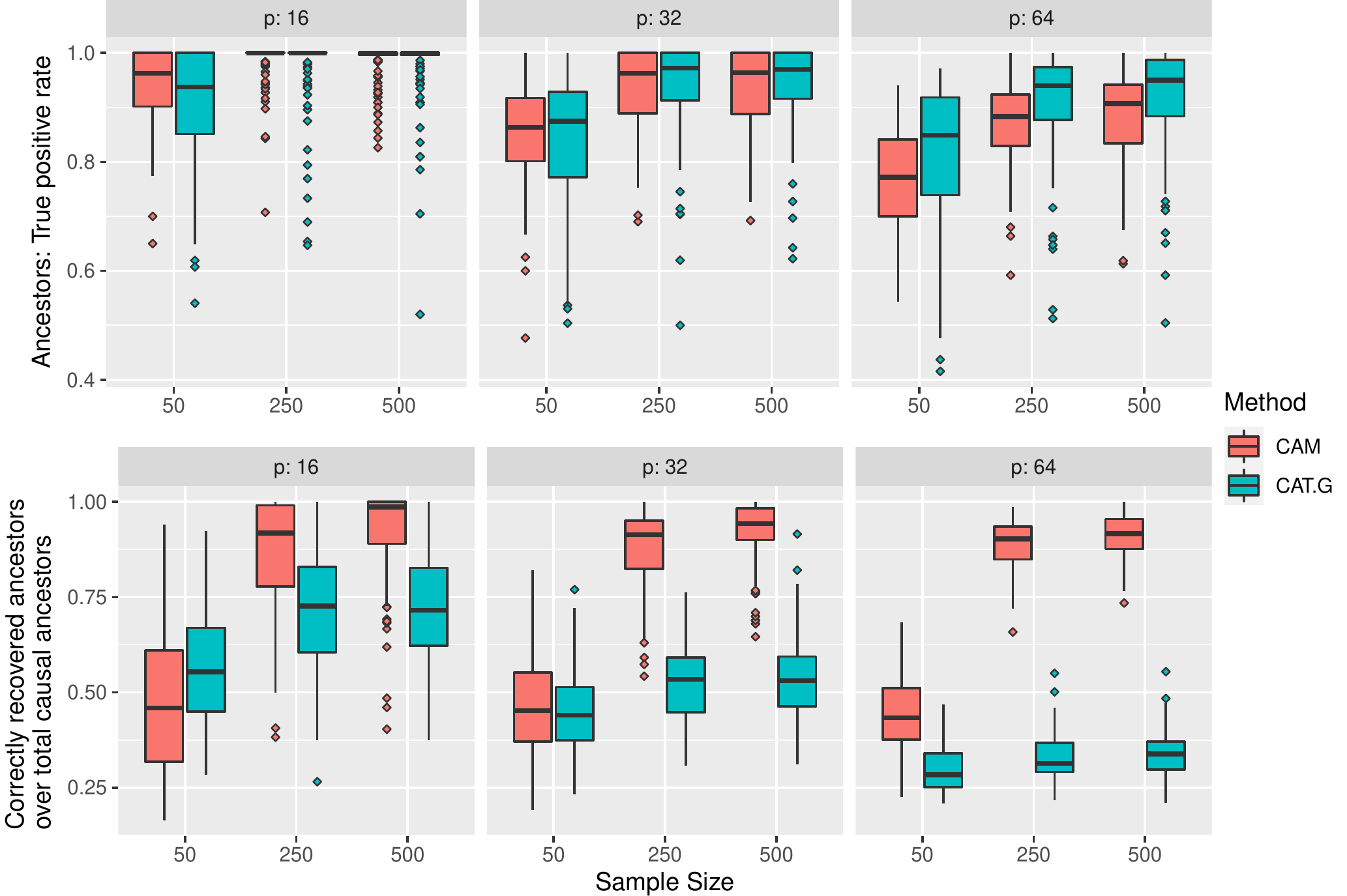}
	\end{center}
	\caption{Estimating ancestor relations in non-tree DAGs, see \Cref{sec:ExperimentCATonDAGs}. 
		CAT.G slightly outperforms CAM in terms of true positive rates for large graphs (top) but finds less ancestor relationships (bottom) due to fitting a tree.}
	\label{fig:SingleRootedDagsAncestorRelations}
\end{figure}
As CAT.G outputs trees, we do not expect it to output the correct graph. 
\Cref{fig:SingleRootedDagsAncestorRelations} illustrates
the performance of CAT.G and CAM in terms of ancestor relations. For this experiment, we employ CAM with preliminary neighborhood selection and subsequent pruning. For small systems, CAM slightly outperforms CAT.G in terms of true positive rate (TPR) when classifying causal ancestors. However, for large systems and large sample sizes, CAT.G outperforms CAM in that metric. On the other hand, CAM is not limited to trees which allows it to find a more significant proportion of the true ancestor, as seen by the fraction of correctly classified ancestors over actual ancestors.
CAT.G seems to be a viable alternative for practical non-tree applications where the true positive rate of estimated ancestors is more important than finding all
ancestor relations.

In \Cref{fig:SingleRootedDagsEdgeRelations} of Section \ref{sec:additionalIllustrations} of \Cref{app:Experiments} we have illustrated similar comparisons when focusing on recovered edges. The true positive rate of the recovered edges for CAT.G is larger than CAM  only for small sample sizes, while the opposite is true for large sample sizes. 
As expected, and as for the ancestor relationships, 
the fraction of correctly predicted edges over total causal edges is significantly higher for CAM.

Finally, while both methods are relatively efficient, CAT has a slightly lower runtime than the greedy search algorithm of CAM. The average runtime of CAM and CAT.G in this experiment for $p=64$ and $n=500$ was 193 and 139 seconds, respectively. For both methods, 
the most time consuming part is 
estimating of the conditional expectations that are used to compute the edge weights.

\section{Summary and Future Work}
This paper shows that exact structure learning is possible for systems of lesser complexity, i.e., for restricted structural causal models with additive noise and causal graphs given by directed trees. 
We propose the method CAT, which is guaranteed to consistently recover the causal directed tree in a Gaussian noise setting under mild assumptions on the  regression methods used to estimate conditional means.
Furthermore, we argue that CAT is consistent in an asymptotic setup with vanishing identifiability. 
We present a computationally feasible procedure to test substructure hypotheses and provide an analysis of the identifiability gap. Simulation experiments show that CAT outperforms other (more general) structure learning methods for the specific task of recovering the causal graph in additive noise structural causal models when the causal structure is given by directed trees.

The proof of \Cref{thm:UniqueGraph} is based on
the fact that the causal functions of alternative models 
are differentiable and that the noise  densities are continuous. We conjecture that it is possible to get even stronger identifiability statements under weaker assumptions; 
proving such a result necessitates new proof strategies. Furthermore, it should be possible to bootstrap a unbiased simultaneous hypercube confidence region for the Gaussian edge weights. This, however, requires a sufficiently fast convergence rate of the estimation error of the conditional expectations corresponding to non-causal edges. Compared to the Bonferroni correction, this approach could increase the power of the test.  

\section*{Acknowledgments}
We thank Phillip Bredahl Mogensen and Thomas Berrett for helpful discussions on the entropy score and its estimation. PB and JP thank David Bürge and Jan Ernest for helpful discussions on exploiting Chu–Liu–Edmonds’ algorithm for causal discovery during the early stages of this project. MEJ and JP were supported by the Carlsberg Foundation; JP was, in addition, supported by a research grant
(18968) from VILLUM FONDEN. RDS was supported by EPSRC grant EP/N031938/1. PB received funding from the European Research Council (ERC) under the European Union's Horizon 2020 research and innovation programme (grant agreement No. 786461).

\chapter[Learning Summary Graphs of Time Series]{Learning Summary Graphs of Time Series and Artifacts in DAG Models} \label{ch:pmlr}

{\small \textsc{Joint work with
		\begin{quote}
			Sebastian Weichwald,  Phillip Bredahl Mogensen, Lasse Petersen, Nikolaj Thams and Gherardo Varando
\end{quote}}}
\vspace{0.75cm}

\begin{quoting}[leftmargin=0.5cm]
	\begin{center}
		\textbf{Abstract}
	\end{center}
	
	{\small 
		In this article, we describe the algorithms for causal structure learning from time series data that won the Causality 4 Climate competition at the Conference on Neural Information Processing Systems 2019 (NeurIPS).
		We examine how our combination of established ideas achieves competitive performance on semi-realistic and realistic time series data exhibiting common challenges in real-world Earth sciences data.
		In particular, we discuss
		a) a rationale for leveraging linear methods to identify causal links in non-linear systems,
		b) a simulation-backed explanation as to why large regression coefficients may predict causal links better in practice than small p-values and
		thus why normalising the data
		may sometimes hinder causal structure learning.
		
		For benchmark usage, we detail the algorithms here and provide implementations at \href{https://github.com/sweichwald/tidybench}{github.com/sweichwald/\textcolor{TIDY}{tidybench}}.
		We propose the presented competition-proven methods for baseline benchmark comparisons to guide the development of novel algorithms for structure learning from time series.}
\end{quoting}
\textbf{Keywords:} 	Causal discovery, structure learning, time series, scaling.

\section{Introduction}\label{sec:introduction}






Inferring causal relationships from large-scale observational studies is an
essential aspect of modern climate science \citealp[][]{runge2019inferring,runge2019detecting}.
However, randomised studies and controlled interventions
cannot be carried out, due to both ethical and practical reasons.
Instead, simulation studies based on climate models are state-of-the-art to study
the complex patterns present in Earth climate systems~\citep{ipcc2013}.

Causal inference methodology can integrate and
validate current climate models and can be used to probe cause-effect relationships between
observed variables.
The Causality 4 Climate (C4C) NeurIPS competition~\citep{runge2020neurips} aimed to further the understanding and development of methods
for structure learning from time series data exhibiting common challenges in and properties of realistic weather and climate data.

\paragraph*{Structure of this work}
Section~\ref{sec:structurelearning} introduces the structure learning task considered.
In Section~\ref{sec:winningalgorithms}, we describe our winning algorithms.
With a combination of established ideas, our algorithms achieved competitive performance on semi-realistic data across all $34$ challenges in the C4C competition track.
Furthermore, at the time of writing, our algorithms lead the rankings for all hybrid and realistic data set categories available on the \href{http://causeme.net}{CauseMe.net} benchmark platform which also offers additional synthetic data categories~\citep{runge2019inferring}.
These algorithms---which can be implemented in a few lines of code---are built on simple methods, are computationally efficient, and exhibit solid performance across a variety of different data sets.
We therefore encourage the use of these algorithms as baseline benchmarks and guidance of future algorithmic and methodological developments for structure learning from time series.

Beyond the description of our algorithms, we aim at providing intuition that can explain the phenomena we have observed throughout solving the competition task.
First, if we \emph{only} ask whether a causal link exists in some non-linear time series system, then we may sidestep the extra complexity of explicit non-linear model extensions (cf.\ Section~\ref{sec:nonlinearbylinear}).
Second, when data has a
meaningful natural scale, it may---somewhat unexpectedly---be advisable to forego data normalisation and to use raw (vector auto)-regression coefficients instead of p-values to assess whether a causal link exists or not (cf.\ Section~\ref{sec:coeffvspval}).

\section{Causal Structure Learning from Time-discrete Observations}\label{sec:structurelearning}
The task of inferring the causal structure from observational data is often referred to as `causal discovery' and was pioneered by \citet{Pearl2009} and \cite{spirtes2000causation}.
Much of the causal inference literature is concerned with structure learning from
independent and identically distributed (iid) observations.
Here, we briefly review some aspects and common assumptions for causally modelling time-evolving systems.
More detailed and comprehensive information can be found in the provided references.

\paragraph*{Time-discrete observations}
We may view the discrete-time observations as arising from an underlying continuous-time causal system~\citep{Peters2020}.
While difficult to conceptualise, the correspondence between structural causal models and differential equation models can be made formally precise~\citep{mooij2013from,rubenstein2016from,bongers2018random}.
Taken together, this yields some justification for modelling dynamical systems by discrete-time causal models.


\paragraph*{Summary graph as inferential target}
It is common to assume a time-homogeneous causal structure such that the dynamics of the observation vector $X$ are governed by $X^{t} := F(X^{{\operatorname{past}(t)}},N^t)$
where the function $F$
determines the next observation based on past values $X^{\operatorname{past}(t)}$ and the noise innovation $N^t$.
Here, structure learning amounts to identifying the summary graph with adjacency matrix $A$ that summarises the causal structure in the following sense:
the $(i,j)^\text{th}$ entry of the matrix $A$ is $1$ if $X_i^{\operatorname{past}(t)}$ enters the structural equation of $X_i^t$ via the $i^{th}$ component of $F$ and $0$ otherwise.
If $A_{ij}=1$, we say that ``$X_i$ causes $X_j$''.
While summary graphs can capture the existence and non-existence of cause-effect relationships, they do in general not correspond to a time-agnostic structural causal model that admits a causal semantics consistent with the underlying time-resolved structural causal model~\citep{rubenstein2017causal,janzing2018structural}.

\paragraph*{Time structure may be helpful for discovery}

In contrast to the iid setting, the Markov equivalence class of the summary graph induced by the structural equations of a dynamical system is a singleton when assuming causal sufficiency and no instantaneous effects~\citep{Peters2017,wengel2019markov}.
This essentially yields a justification and a constraint-based causal inference perspective on Wiener-Granger-causality \citep{wiener1956theory,granger1969investigating,Peters2017}


\paragraph*{Challenges for causal structure learning from time series data}

Structure learning from time series is a challenging task hurdled by further problems such as time-aggregation, time-delays, and time-subsampling.
All these challenges were considered in the C4C competition and are topics of active research~\citep{danks2013learning,hyttinen2016causal}.

\section[Winning Algorithms]{The \textcolor{TIDY}{Ti}me-series \textcolor{TIDY}{D}iscover\textcolor{TIDY}{y} \textcolor{TIDY}{Bench}mark (\textcolor{TIDY}{tidybench}): Winning Algorithms}
\label{sec:winningalgorithms}

We developed four simple algorithms,
\begin{enumerate}[leftmargin=2cm,rightmargin=2cm]
\setlength\itemsep{0em}
\item[\texttt{SLARAC}] Subsampled Linear Auto-Regression Absolute Coefficients (cf.\ Alg.~1)

\item[\texttt{QRBS}] Quantiles of Ridge regressed Bootstrap Samples~(cf.\ Alg.~2)

\item[\texttt{LASAR}] LASso Auto-Regression

\item[\texttt{SELVAR}] Selective auto-regressive model

\end{enumerate}
which came in first in 18 and close second in 13 out of the 34 C4C competition categories and won the overall competition~\citep{runge2020neurips}.
Here, we provide detailed descriptions of the \texttt{SLARAC} and \texttt{QRBS} algorithms.
DYAnalogous descriptions for the latter two algorithms and implementations of all four algorithms are available at \href{https://github.com/sweichwald/tidybench}{github.com/sweichwald/\textcolor{TIDY}{tidybench}}.

All of our algorithms output an edge score matrix that contains
for each variable pair $(X_i,X_j)$ a score that reflects how likely it is that the edge $X_i \to X_j$ exists.
Higher scores correspond to edges that are inferred to be more likely to exist than edges with lower scores, based on the observed data.
That is, we rank edges relative to one another but do not perform hypothesis tests for the existence of individual edges.
A binary decision can be obtained by choosing a cut-off value for the obtained edge scores.
In the C4C competition, submissions were compared to the ground-truth cause-effect adjacency matrix and assessed based on the achieved ROC-AUC when predicting which causal links exist.

The idea behind our algorithms is the following:
regress present on past values and inspect the regression coefficients to decide whether one variable is a Granger-cause of another.
\texttt{SLARAC} fits a VAR model on bootstrap samples of the data each time choosing a random number of lags to include;
\texttt{QRBS} considers bootstrap samples of the data and Ridge-regresses time-deltas $X(t) - X(t-1)$ on the preceding values $X(t-1)$;
\texttt{LASAR} considers bootstrap samples of the data and iteratively---up to a maximum lag---LASSO-regresses the residuals of the preceding step onto values one step further in the past and keeps track of the variable selection at each lag to fit an OLS regression in the end with only the selected variables at selected lags included;
and \texttt{SELVAR} selects edges employing a hill-climbing procedure based on the
leave-one-out residual sum of squares and finally scores the selected
edges with the absolute values of the regression coefficients.
In the absence of instantaneous effects and hidden confounders, Granger-causes are equivalent to a variable's causal parents~\cite[Theorem 10.3]{Peters2017}.
%
%
In Section~\ref{sec:coeffvspval}, we argue that the size of the regression coefficients may in certain scenarios be more informative about the existence of a causal link than standard test statistics for the hypothesis of a coefficient being zero.
It is argued that for additive noise models, information about the causal ordering may be contained in the raw marginal variances. In test statistics such as the F- and T-statistics, this information is lost when normalising by the marginal variances.

\begin{figure}[ht!]
\begin{center}
	\includegraphics[width=1\textwidth-1.75cm]{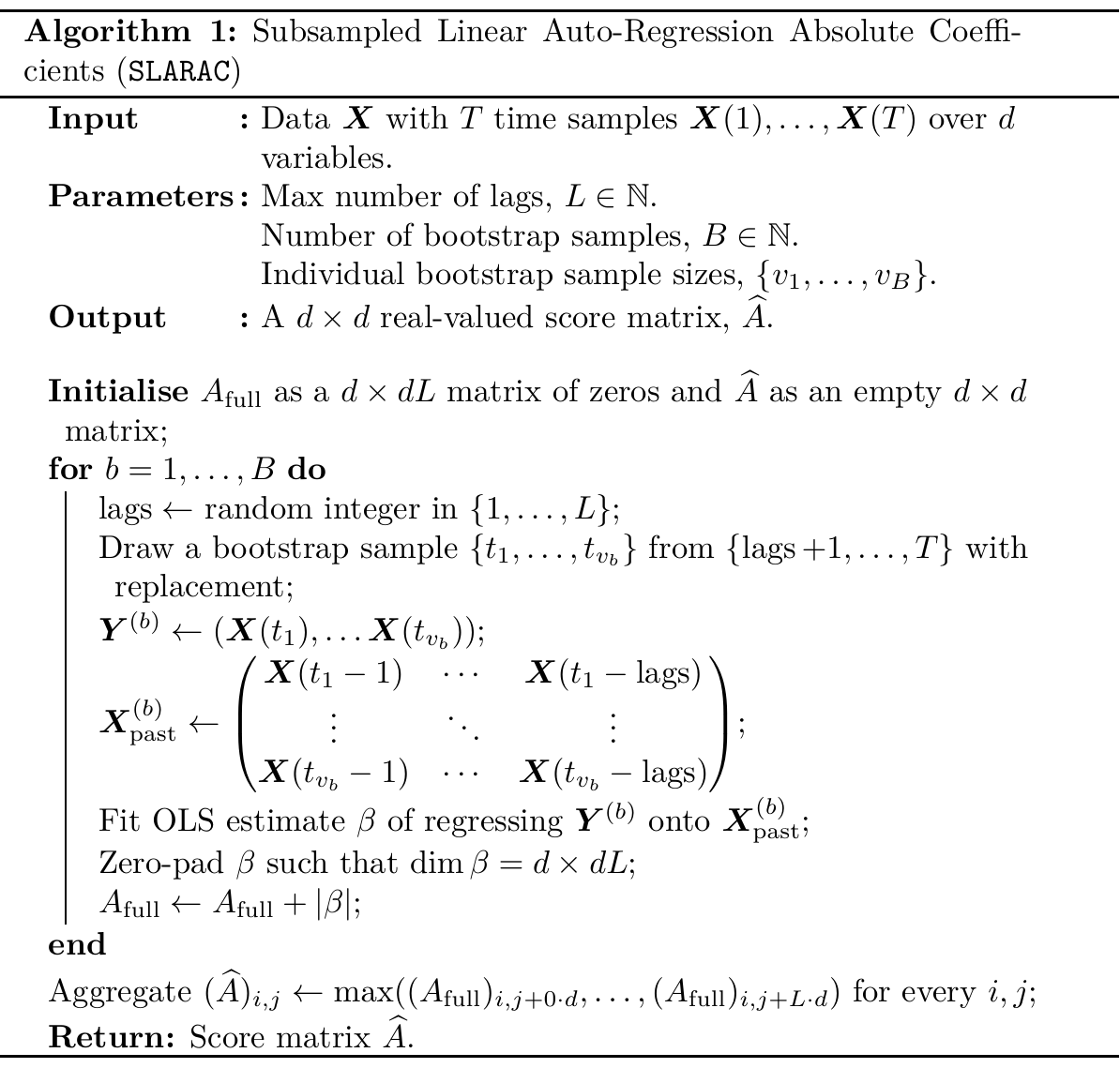}
\end{center}
\end{figure}
\begin{figure}[ht!]
\begin{center}
	\includegraphics[width=1\textwidth-1.75cm]{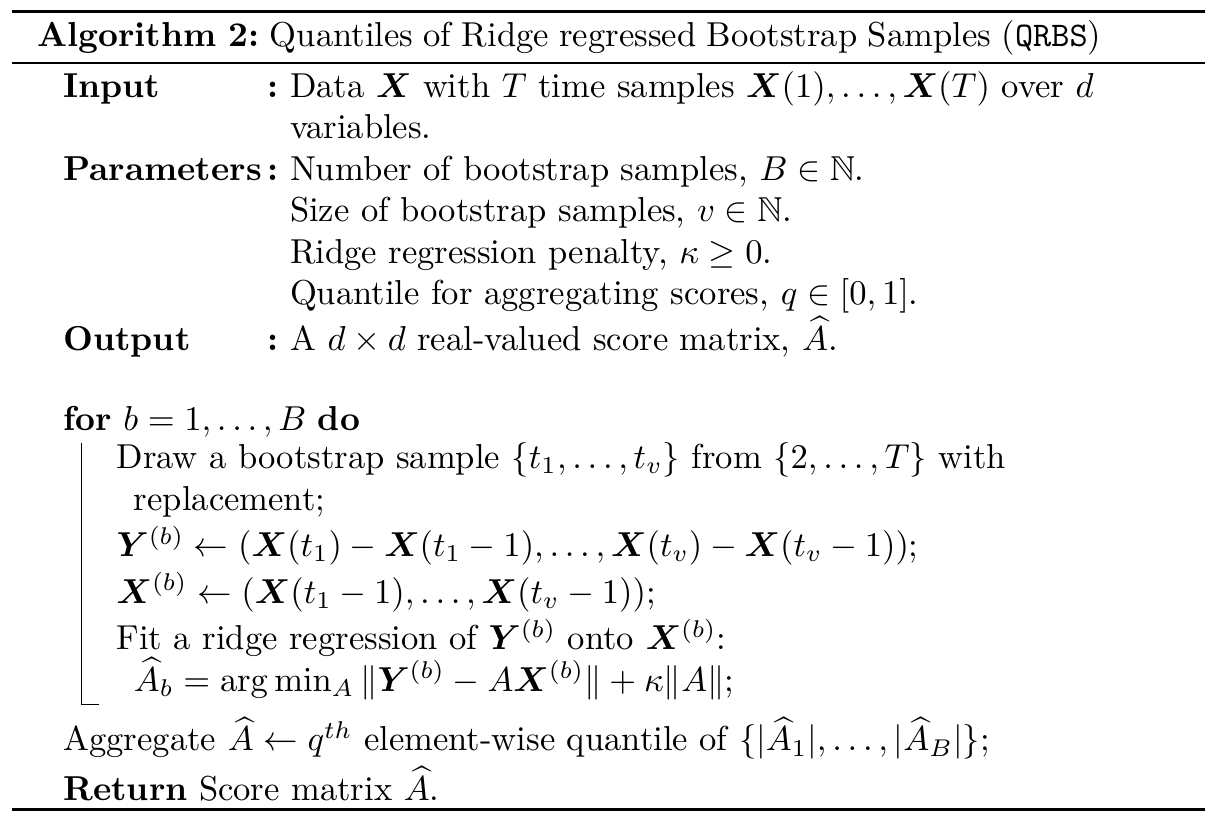}
\end{center}
\end{figure}

\section{Capturing Nonlinear Cause-Effect Links by Linear Methods}\label{sec:nonlinearbylinear}

We explain the rationale behind our graph reconstruction algorithms and how they may capture non-linear dynamics despite being based on linearly regressing present on past values.
For simplicity we will outline the idea in a multivariate regression setting with additive noise, but it
extends to the time series setting by assuming time homogeneity.

Let $N, X(t_1),X(t_2) \in \mathbb{R}^d$ be random variables such that $$X(t_2) := F(X(t_1)) + N$$ for some differentiable function
$F = (F_1, \dots, F_d) : \mathbb{R}^d \to \mathbb{R}^d$. Assume that $N$ has mean zero, that
it is independent from $X(t_1)$, and that it has mutually independent components.
For each $i, j = 1, \dots, d$ we define the quantity of interest
  \begin{align*}
  \theta_{ij} = \mathbb{E} \left|\partial_i F_j\left(X(t_1)\right)\right|,
  \end{align*}
such that $\theta_{ij}$ measures the expected effect from $X_i(t_1)$ to $X_j(t_2)$.
We take the matrix $\Theta = \left(\bm{1}_{\theta_{ij}>0}\right)$ as the adjacency matrix of the summary graph
between $X(t_1)$ and $X(t_2)$.


In order to detect regions with non-zero gradients of $F$ we create bootstrap samples $\mathcal{D}_1, \dots, \mathcal{D}_B$.
On each bootstrap sample $\mathcal{D}_b$ we obtain the regression coefficients $\widehat{A}_b$ as estimate of the directional derivatives by a (possibly penalised) linear
regression technique.
Intuitively, if $\theta_{ij}$ were zero, then on any bootstrap sample we would obtain a small non-zero contribution.
Conversely, if $\theta_{ij}$ were non-zero, then we may for some bootstrap samples obtain a linear fit of $X_j(t_2)$ with large absolute regression coefficient for $X_i(t_1)$.
The values obtained on each bootstrap sample are then aggregated by, for example, taking the average of the absolute regression coefficients $\widehat{\theta}_{ij} = \frac{1}{B} \sum_{b=1}^B \left|(\widehat{A}_b)_{ij}\right|$.

This amounts to searching the predictor space for an effect from $X_i(t_1)$ to $X_j(t_2)$, which
is approximated linearly.
It is important to aggregate the absolute values of the coefficients to avoid cancellation
of positive and negative coefficients. The score $\widehat{\theta}_{ij}$ as such contains no information about whether the effect from $X_i(t_1)$ to $X_j(t_2)$
is positive or negative and it cannot be used to predict $X_j(t_2)$ from $X_i(t_1)$.
It serves as a score for the existence of a link between the two variables.
This rationale explains how linear methods may be employed for edge detection in non-linear settings without requiring extensions of Granger-type methods that explicitly model the non-linear dynamics and hence come with additional sample complexity~\citep{marinazzo2008kernel,marinazzo2011nonlinear,stramaglia2012expanding,stramaglia2014synergy}.

\section[Regression Coefficients and Artifacts in DAGs]{Large Regression Coefficients May Predict Causal Links Better in Practice Than Small P-values}\label{sec:coeffvspval}

This section aims at providing intuition behind two phenomena:
We observed a considerable drop in the accuracy of our edge predictions whenever
1) we normalised the data or
2)~used the T-statistics corresponding to testing the hypothesis of regression coefficients being zero to score edges instead of the coefficients' absolute magnitude.
While one could try to attribute these phenomena to some undesired artefact in the competition setup,
it is instructive to instead try to understand when exactly one would expect such behaviour.

We illustrate a possible explanation behind these phenomena and do so in an iid setting in favour of a clear exposition, while the intuition extends to settings of time series observations and our proposed algorithms.
The key remark is, that under comparable noise variances, the variables' marginal variances tend to increase along the causal ordering.
If data are observed at comparable scales---say sea level pressure in different locations measured in the same units---or at scales that are in some sense naturally relative to the true data generating mechanism, then  absolute regression coefficients may be preferable to T-test statistics.
Effect variables tend to have larger marginal variance than their causal ancestors.
This helpful signal in the data is diminished by normalising the data or the rescaling when computing the T-statistics corresponding to testing the regression coefficients for being zero.
This rationale is closely linked to the identifiability of Gaussian structural equation models under equal error variances~\citet{peters2014identifiability}.
Without any prior knowledge about what physical quantities the variables correspond to and their natural scales, normalisation remains a reasonable first step.
We are not advocating that one should use the raw coefficients and not normalise data, but these are two possible alterations of existing structure learning procedures that may or may not, depending on the concrete application at hand, be worthwhile exploring.
Our algorithms do not perform data normalisation, so the choice is up to the user whether to feed normalised or raw data, and one could easily change to using p-values or T-statistics instead of raw coefficients for edge scoring.

\subsection{Instructive IID Case Simulation Illustrates Scaling Effects}\label{sec:iidsimulation}
We consider data simulated from a standard acyclic linear Gaussian model.
Let $N \sim \mathcal{N}\left(0, \operatorname{diag}(\sigma_1^2, \dots, \sigma_d^2)\right)$ be a $d$-dimensional random variable and let $\boldsymbol{B}$ be a $d\times d$ strictly lower-triangular matrix.
Further, let $X$ be a $d$-valued random variable constructed according to the structural equation $X = \boldsymbol{B} X + N$, which induces
a distribution over $X$ via $X = (I - \bm{B})^{-1}N$.
We have assumed, without loss of generality, that the causal order is aligned such that $X_i$ is further up in the causal order than $X_j$ whenever $i<j$.
We ran $100$ repetitions of the experiment, each time sampling a random lower triangular $50\times 50$-matrix $\bm{B}$ where each entry in the lower triangle is drawn from a standard Gaussian with probability $\sfrac{1}{4}$ and set to zero otherwise.
For each such obtained $\bm{B}$ we sample $n=200$ observations from $X=\bm{B}X+N$ which we arrange in a data matrix $\bm{X}\in\mathbb{R}^{200 \times 50}$ of zero-centred columns denoted by $\bm{X}_{j}$.

We regress each $X_j$ onto all remaining variables $X_{\neg j}$ and compare scoring edges $X_i\to X_j$ by the absolute values of a) the regression coefficients $|\widehat{b}_{i\to j}|$, versus b) the T-statistics $|\widehat{t}_{i\to j}|$ corresponding to testing the hypothesis that the regression coefficient $\widehat{b}_{i\to j}$ is zero. That is, we consider
\[
|\widehat{b}_{i\to j}| = \left|(\bm{X}_{\neg j}^\top \bm{X}_{\neg j})^{-1}\bm{X}_{\neg j}^\top \bm{X}_j\right|_i
\]
versus
\begin{equation}
|\widehat{t}_{i\to j}|  =
|\widehat{b}_{i\to j}|
{\color{FACTOR}\sqrt{
\frac{
\widehat{\operatorname{var}}(X_i|X_{\neg i})
}{
\widehat{\operatorname{var}}(X_j|X_{\neg j})
}}}
\sqrt{\frac{
(n - d)
}{
\left(1 - \widehat{\operatorname{corr}}^2({X_i,X_j|X_{\neg \{i, j\}}})\right)
}}\label{eq:t_mulit}
\end{equation}
where
$
\widehat{\operatorname{var}}(X_j|X_{\neg j})$
is the residual variance after regressing $X_j$ onto the other variables $X_{\neg j}$,
and
$
\widehat{\operatorname{corr}}({X_i,X_j|X_{\neg \{i, j\}}})
$
is the residual correlation between $X_i$ and $X_j$ after regressing both onto the remaining variables.
%
%

\begin{figure}[h]
\resizebox{\textwidth}{!}{\input{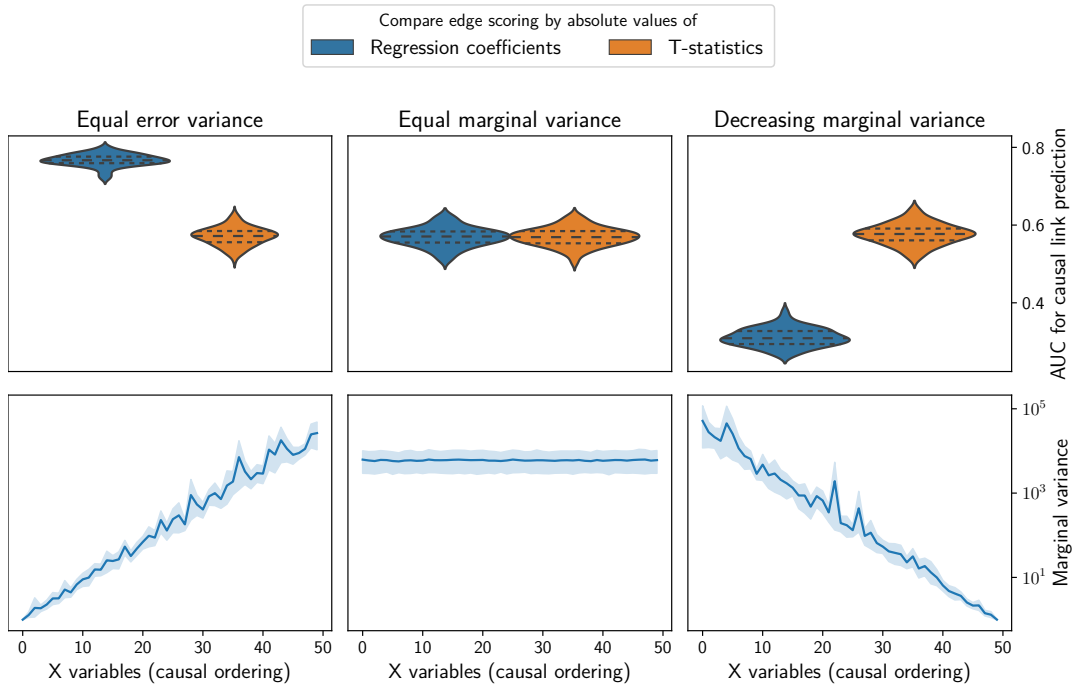}}
\caption{Results of the simulation experiment described in Section~\ref{sec:iidsimulation}.
Data is generated from an acyclic linear Gaussian model, in turn each variable is regressed onto all remaining variables and either the raw regression coefficient $|\widehat{b}_{i\to j}|$ or the corresponding T-statistics $|\widehat{t}_{i\to j}|$ is used to score the existence of an edge $i\to j$. 
The top row shows the obtained AUC for causal link prediction and the bottom row the marginal variance of the variables along the causal ordering.
The left panel shows naturally increasing marginal variance for equal error variances, for the middle and right panel the model parameters and error variances are rescaled to enforce equal and decreasing marginal variance, respectively.
}\label{fig:simulation}
\end{figure}

We now compare, across three settings, the AUC obtained by either using the absolute value of the regression coefficients $|\widehat{b}_{i\to j}|$ or the absolute value of the corresponding T-statistics $|\widehat{t}_{i\to j}|$ for edge scoring.
Results are shown in the left, middle, and right panel of Figure~\ref{fig:simulation}, respectively.

\paragraph*{In the setting with equal error variances $\sigma_i^2 = \sigma^2_j\ \forall i,j$,}
we observe that
i) the absolute regression coefficients beat the T-statistics for edge predictions in terms of AUC, and
ii) the marginal variances naturally turn out to increase along the causal ordering.

When moving from $|\widehat{b}_{i\to j}|$ to $|\widehat{t}_{i\to j}|$ for scoring edges, we multiply by a term that compares the relative residual variance of $X_i$ and $X_j$.
If $X_i$ is before $X_j$ in the causal ordering it tends to have both smaller marginal and---in our simulation set-up---residual variance than $X_j$ as it becomes increasingly more difficult to predict variables further down the causal ordering.
In this case, the fraction of residual variances will tend to be smaller than one and consequently the raw regression coefficients $|\widehat{b}_{i\to j}|$ will be shrunk when moving to $|\widehat{t}_{i\to j}|$.
This can explain the worse performance of the T-statistics compared to the raw regression coefficients for edge scoring as scores will tend to be shrunk when in fact $X_i \to X_j$.


\paragraph*{Enforcing equal marginal variances by rescaling the rows of $\bm{B}$ and the $\sigma_i^2\text{'s}$,}
we indeed observe that regression coefficients and T-statistics achieve comparable performance in edge prediction in this somewhat artificial scenario.
Here, neither the marginal variances nor the residual variances appear to contain information about the causal ordering any more and the relative ordering between regression coefficients and T-statistics is preserved when multiplying by the factor {\color{FACTOR}highlighted} in Equation~\ref{eq:t_mulit}.



\paragraph*{Enforcing decreasing marginal variances by rescaling the rows of $\bm{B}$ and the $\sigma_i^2\text{'s}$,}
we can, in line with our above reasoning, indeed obtain an artificial scenario in which the T-statistics will outperform the regression coefficients in edge prediction, as now, the factors we multiply by will work in favour of the T-statistics.




\section{Conclusion and Future Work}

We believe that competitions like the C4C competition~\citep{runge2020neurips} and causal discovery benchmark platforms like \href{http://causeme.net}{CauseMe.net}~\citep{runge2019inferring} are important for bundling and informing the community's joint research efforts into methodology that is readily applicable to tackle real-world data.
In practice, there are fundamental limitations to causal structure learning that ultimately require us to employ untestable causal assumptions to proceed towards applications at all.
Yet, both these limitations and assumptions are increasingly well understood and characterised by methodological research and time and again need to be challenged and examined through the application to real-world data.

Beyond the algorithms presented here and proposed for baseline benchmarks, different methodology as well as different benchmarks may be of interest.
For example, our methods detect causal links and are viable benchmarks for the structure learning task but they do not per se enable predictions about the interventional distributions.

%
%

\section*{Acknowledgments}
The authors thank Niels Richard Hansen, Steffen Lauritzen, and Jonas Peters for insightful discussions.
Thanks to the organisers for a challenging and insightful Causality 4 Climate NeurIPS competition.
NT was supported by a research grant (18968) from VILLUM FONDEN.
LP and GV were supported by a research grant (13358) from VILLUM FONDEN.
MEJ and SW were supported by the Carlsberg Foundation.


\begin{appendices}
\crefalias{section}{appsec}
\renewcommand{\thesection}{\Alph{chapter}.\arabic{section}}
\renewcommand{\theequation}{\Alph{chapter}.\arabic{equation}}
\renewcommand{\thetheorem}{\Alph{chapter}.\arabic{theorem}}
\renewcommand{\theassumption}{\Alph{chapter}.\arabic{assumption}}
\renewcommand{\theproposition}{\Alph{chapter}.\arabic{proposition}}
\renewcommand{\thecorollary}{\Alph{chapter}.\arabic{corollary}}
\renewcommand{\thelemma}{\Alph{chapter}.\arabic{lemma}}
\renewcommand{\theexample}{\Alph{chapter}.\arabic{example}}
\renewcommand{\theremark}{\Alph{chapter}.\arabic{remark}}
\renewcommand{\thedefinition}{\Alph{chapter}.\arabic{definition}}
\renewcommand{\thefigure}{\Alph{chapter}.\arabic{figure}}
\renewcommand{\thetable}{\Alph{chapter}.\arabic{table}}
\renewcommand{\thealgorithm}{\Alph{chapter}.\arabic{algorithm}}

\crefname{section}{appendix}{appendices}

\csname @removefromreset\endcsname{equation}{section}
\numberwithin{equation}{chapter}
\csname @removefromreset\endcsname{figure}{section}
\numberwithin{figure}{chapter}	
\csname @removefromreset\endcsname{algorithm}{section}
\numberwithin{algorithm}{chapter}
	
\chapter[Distributional Robustness of K-class Estimators and the PULSE]{Distributional Robustness of K-class Estimators and the PULSE}
  
  \vspace*{1cm}    
  
  \begin{quote}
    \begin{enumerate}
 \item[\textbf{\ref{sec:simsem}}] \nameref{sec:simsem} 
 \item[\textbf{\ref{app:algo}}] \nameref{app:algo} 
 \item[\textbf{\ref{sec:RobustnessProofs}}] \nameref{sec:RobustnessProofs} 
 \item[\textbf{\ref{sec:SomeProofsOfSecPULSE}}] \nameref{sec:SomeProofsOfSecPULSE} 
 \item[\textbf{\ref{sec:RemainingProofsOfSecPULSE}}] \nameref{sec:RemainingProofsOfSecPULSE} 
 \item[\textbf{\ref{sec:AuxLemmas}}] \nameref{sec:AuxLemmas} 
 \item[\textbf{\ref{app:AddRemarks}}] \nameref{app:AddRemarks} 
 \item[\textbf{\ref{sec:Experiments}}] \nameref{sec:Experiments} 
 \item[\textbf{\ref{app:EmpricalApp}}] \nameref{app:EmpricalApp} 
 \item[\textbf{\ref{sec:WeakInst}}] \nameref{sec:WeakInst} 
 \item[\textbf{\ref{sec:AppFigs}}] \nameref{sec:AppFigs} 
    \end{enumerate}
  \end{quote}
  
\section{Structural Equation Models and Interventions} 
\label{sec:simsem}
Structural equation models and simultaneous equation models are causal models. That is, they contain more information than the description of an observational distribution. 
We first
introduce the notion of structural equation models (also called structural causal models)
and use an example to show how they can be written as in the form of
simultaneous equation models (SIM) commonly used in econometrics, see \Cref{sec:LinearModels}.

\subsection{Structural equaton models and interventions}
A structural equation model (SEM)
(e.g.\ \citealp{Bollen1989}, and \citealp{Pearl2009})
over variables $X_1, \ldots, X_p$ consists of
$p$ assignments of the form
$$
X_j := f_j(X_{\PA{j}}, \ep_j), \qquad j = 1, \ldots, p,
$$
where $\PA{j} \subseteq \{1, \ldots, p\}$
are called the parents of $j$, 
together with a distribution over 
the noise variables $(\ep_1, \ldots, \ep_p)$, which is assumed to have jointly independent marginals. 
The corresponding graph 
over $X_1, \ldots, X_p$ 
is obtained by drawing directed edges from the variables on the right-hand side to 
the variables on the left-hand side. 
If the corresponding 
graph is acyclic, the SEM induces a unique distribution over 
$(X_1, \ldots, X_p)$, which is often called the observational distribution.
Section~\ref{sec:LinearModels} below discusses an example of linear assignments,
which also allows for a cyclic graph structure.
The framework of SEMs also models the effect of 
interventions:
An intervention on variable $j$ corresponds to 
replacing the $j$th assignment.
For example, 
replacing it by
$X_j = 4$, called a hard intervention, or, more generally, by
$X_j = g(X_{\tPA{j}}, \tilde{\ep}_j)$
induces yet another distribution over $X$ that is called an interventional distribution 
and that we denote by
$P^{\mathrm{do}(X_j = 4)}$ or 
$P^{\mathrm{do}(X_j = g(X_{\tPA{j}}, \tilde{\ep}_j))}$, respectively.
A formal introduction to SEMs, 
in the general case of cyclic assignments is provided by \citet{bongers2021foundations}, for example.
In an SEM, 
we call all $X$ variables endogenous and,
in addition, all variables $X_j$, for which we have
$\PA{j} = \emptyset$, will be called exogenous.
A subset of variables is called exogenous relative to another subset
if it does not contain a variable 
that has a parent belonging to the other set. 

In the  paper, 
we are mostly interested in one of these equations and we denote the corresponding target variable as $Y$.
Furthermore, some of the other $X$ variables  may be unobserved, which we indicate by using the notation $H$ (denoting a vector of variables). In linear models, hidden variables can 
equivalently
be represented 
as correlation in the noise variables; see e.g.\ \citet{bongers2021foundations}, and \citet{Hyttinen2012}.
Finally, we let $A$ denote a collection of variables that are known 
to enter the system as exogenous variables, relative to $(Y,X,H)$.

\subsection{Example of a Linear Structural Equation Model} \label{sec:LinearModels}
Let the distribution of $(Y,X,H,A)$ be generated according to the possibly cyclic 
SEM,
\begin{align} \label{ARModelsupplement}
	\begin{bmatrix}
		Y & X^\t & H^\t \end{bmatrix} 
	:=  \begin{bmatrix}
		Y &X^\t  & H^\t  \end{bmatrix} B   +  A^\t M+ \ep^\t.
\end{align}
Here, $B$ is a square matrix
with eigenvalues whose absolute value is strictly smaller than one. This implies that $I-B$ is invertible ensuring that 
the distribution of
$(Y,X,H)$ is well-defined since 
$(Y,X,H)$
can be expressed in terms of $B,M,A$ and $\ep$ as $(I-B^\t)^{-1}(M^\t A+ \ep)$. 
We denote the random vectors $Y\in \R, X\in \R^{d}, A\in \R^{q},H\in \R^r$  and $\ep\in \R^{d+1+r}$ 
by target, endogenous regressor, anchor, hidden and noise variables, respectively. 
We assume that $\ep\independent A$ rendering the so-called anchors as exogenous variables but the coordinate components of $A$ may be dependent on each other. 
As above,
we assume joint independence of the noises 
$\ep_1, \ldots, \ep_{1+d+r}$.
Let 
$\fY \in \R^{n\times 1},\fX \in \R^{n\times d},\fA \in \R^{n\times q}, \fH \in \R^{n\times r}$ and $\bm{\ep}\in \R^{n\times(1+d+r)}$ be data-matrices with $n\in \N$ row-wise i.i.d.\ copies of the variables solving the system in \Cref{ARModel}. Transposing the structural equations and stacking them vertically by row-wise observations, we can represent all structural equations by $$[
\fY  \, \, \fX  \, \,  \fH
] := [
\fY  \,  \, \fX \,  \,  \fH
] B + \fA M + \bm{\ep}.$$ We can solve the structural equations for the endogenous variables and get the so-called structural and reduced form equations, commonly seen in econometrics, 
\begin{align} \label{Eq:FullStructuralAndReducedFormOfSEM}
	\begin{bmatrix}
		\fY  & \fX &  \fH
	\end{bmatrix} \Gamma  = \fA M + \bm{\ep} \quad \text{and} \quad \begin{bmatrix}
		\fY  & \fX &  \fH
	\end{bmatrix}   = \fA \Pi + \bm{\ep}\Gamma^{-1},
\end{align}
respectively, where $\Gamma := I-B$ and  $\Pi := M \Gamma^{-1}$.
Note that the equations in \Cref{Eq:FullStructuralAndReducedFormOfSEM} differ from the standard representations of simultaneous equation models as we have unobserved endogenous variables $\fH$ in the system. In this setup,
identifiability of the full system parameters $\Gamma$ and $M$ in general breaks down due to the dependencies generated by the unobserved endogenous variables. 
We now assume without loss of generality that $\Gamma$ has a unity diagonal, such that the target equation of interest, corresponding to the first column of \Cref{Eq:FullStructuralAndReducedFormOfSEM}, is given by
\begin{align} \label{eq:StructuralEquationOfInterestSupp}
	\fY = \fX \gamma_0 + \fA \beta_0  + \fH \eta_0 + \bm{\ep}_Y = %
	\fZ \alpha_0 + \tilde{\fU}_Y,
\end{align}
where $(1, -\gamma_0,- \eta_0)\in \R^{(1+d+r)}$, $ \beta_0\in \R^q$ and $\bm{\ep}_Y$  are the first columns of $\Gamma$, $M$ and $\bm{\ep}$ respectively, $\fZ := [
\fX \, \, \fA]$, $\alpha_0 = (\gamma_0, \beta_0)\in \R^{d+q}$ and $\tilde{\fU}_Y := \fH\eta_0 + \bm{\ep}_Y$.

The parameter of interest, $\alpha_0$, 
can be derived directly from the corresponding entries in the matrices $B$ and $M$.
It carries causal information in that, for example,
after intervening on all variables except for $Y$, that 
is, considering an intervention 
$Z := z$, 
and 
$H := h$, 
$Y$ has the mean 
$z \alpha_0 + h \eta_0 + E \epsilon_1$, see \Cref{ARModel}. 
%

In \Cref{eq:StructuralEquationOfInterestSupp} we have represented the target variable in terms of a linear combination of the observable variables $Z=(X^\t,A^\t)^\t$ and some unobservable noise term $\tilde{U}_Y$. 
In contrast to \Cref{ARModel},   
\Cref{eq:StructuralEquationOfInterestSupp}, 
which is more commonly used in the econometrics literature, 
models 
the influence of the latent variables using a dependence between endogenous 
variables and the noise term $\tilde U_Y$;
this equivalence is well-known and described by \citet{bongers2021foundations} and \citet{Hyttinen2012}, for example.
The construction in \Cref{ARModel} can be seen as a manifestation of Reichenbach's common cause principle (\citealp{reichenbach1956direction}). This principle stipulates that if two random variables are dependent then either one causally influences the other or there exists a third variable which causally influences both. 

\section{Algorithms} \label{app:algo}
In this section we present two algorithms. Algorithm 1 details a binary search procedure for the dual PULSE parameter $\lambda^\star_n(p_{\min})$ and Algorithm 2 
details the algorithmic construction and output messages of the PULSE estimator.
		\begin{algorithm}[H]
	\caption{Binary.Search with precision $1/N$.\label{Binary.Search.Lambda.Star}}
	\begin{algorithmic}[1]
		\State \textbf{input} $p_{\min}, N$
		%
		\State {\bf if} {$T_n(\hat{\alpha}_{\text{TSLS}}^n) \geq  Q_{\chi^2_{q}}(1-p_{\min})$} 
		{\bf then} terminate procedure {\bf end if}
		%
		\State $\ell_{\min}\leftarrow 0$; 
		$\ell_{\max} \leftarrow 2$
		\While{$T_n(\hat{\alpha}_\text{K}^n(\ell_{\max} )) >  Q_{\chi^2_{q}}(1-p_{\min}) $}
		\State $\ell_{\min} \leftarrow \ell_{\max}$; 
		$\ell_{\max} \leftarrow \ell_{\max}^2$
		\EndWhile
		\State $\Delta \leftarrow \ell_{\max}-\ell_{\min}$
		\While{$\Delta > 1/N$ } 
		\State $\ell \leftarrow (\ell_{\min}+ \ell_{\max})/2$
		\State {\bf if} $T_n(\hat{\alpha}_\text{K}^n (\ell)) >  Q_{\chi^2_{q}}(1-p_{\min}) $ {\bf then}
		$\ell_{\min} \leftarrow \ell$
		{\bf else}
		$\ell_{\max} \leftarrow \ell$
		{\bf end if}
		\State $\Delta \leftarrow \ell_{\max}-\ell_{\min}$
		\EndWhile
		\State return($\ell_{\max}$)
	\end{algorithmic}
\end{algorithm}

	\begin{algorithm}[H] \label{algo:pulseplus}
	
	\caption{PULSE$+$}
	\begin{algorithmic}[1]
		\State \textbf{input} $p_{\min}$, precision $1/N$, $\hat{\alpha}_{\text{ALT}}^n$
		%
		\If{$T_n(\hat{\alpha}_{\text{TSLS}}^n) \geq  Q_{\chi^2_{q}}(1-p_{\min})$}
		\State Warning: TSLS outside interior of acceptance region.
		\State $\hat{\alpha}_{\text{PULSE}+}^n(p_{\min}) \leftarrow \hat{\alpha}_{\text{ALT}}^n$
		\Else 
		\If{$T_n(\hat{\alpha}_{\text{OLS}}^n) \leq Q_{\chi^2_{q}}(1-p_{\min})$}
		\State Warning: The OLS is accepted.
		\State $\lambda^\star_n(p_{\min}) \leftarrow 0$
		
		\Else
		\State $\lambda^\star_n(p_{\min}) \leftarrow \text{Binary.Search}(N,p_{\min})$
		\EndIf
		\State $\hat{\alpha}_{\text{PULSE}+}^n(p_{\min}) \leftarrow (\fZ^\t (\fI+\lambda_n^\star(p_{\min}) P_\fA)\fZ)^{-1} \fZ^\t(\fI+\lambda_n^\star(p_{\min}) P_\fA)\fY$
		\EndIf
		\State return($\hat{\alpha}_{\text{PULSE}+}^n(p_{\min})$)
	\end{algorithmic}
	\label{alg:2}
\end{algorithm}

\section{Proofs of Results in Section~\ref{SEC:ROBUSTNESS}}
\label{sec:RobustnessProofs}
\medskip


\noindent\begin{proofenv}{\textbf{\Cref{lm:PenalizedKClassSolutionUniqueAndExists}}}
	The minimizations of \Cref{KclassLossFunctionPop}  and \Cref{KclassLossFunctionEmp} are unconstrained optimization problems. We  know that there exists a unique solution if the problems are strictly convex. Thus, it suffices to verify the second order condition for strict convexity of the objective functions, i.e., $D^2	l_{\mathrm{K}}^n(\alpha;\kappa)\succ 0$. To this end, note that
	$
	D l_{\mathrm{OLS}}^n(\alpha;\fZ_*,\fX) =  2(\alpha^\t \fZ_*^\t \fZ_* - \fY^\t\fZ_*)/n$  and $ 
	D l_{\mathrm{IV}}^n(\alpha;\fY,\fZ_*,\fA) =  2(\alpha^\t \fZ_*^\t P_\fA \fZ_* -  \fY^\t P_\fA \fZ_*)/n$. 
	Thus, the first order derivative of the K-class regression loss function is given by the $\kappa$-weighted affine combination of these two, that is,
	\begin{align*}
	&D l_{\mathrm{K}}^n(\alpha;\kappa,\fY,\fZ_*,\fA) \\
	&= 2n^{-1}\lp (1-\kappa)\lp \alpha^\t \fZ_*^\t \fZ_* - \fY^\t\fZ_*\rp + \kappa \lp \alpha^\t \fZ_*^\t P_\fA \fZ_* -  \fY^\t P_\fA \fZ_*  \rp \rp \\
	&= 2n^{-1} \lp \alpha^\t \lp \fZ_*^\t \lp (1-\kappa)\fI +\kappa   P_\fA \rp \fZ_*  \rp - \lp \fY^\t \lp (1-\kappa) \fI + \kappa P_\fA\rp  \fZ_* \rp \rp  \\
	&= 2n^{-1} \lp \alpha^\t \lp \fZ_*^\t \lp \fI - \kappa\lp \fI- P_\fA \rp  \rp \fZ_*  \rp - \lp \fY^\t \lp \fI - \kappa\lp \fI- P_\fA \rp \rp  \fZ_* \rp \rp \\
	&= 2n^{-1} \lp \alpha^\t \lp \fZ_*^\t \lp \fI - \kappa P_\fA^\perp   \rp \fZ_*  \rp - \lp \fY^\t \lp \fI - \kappa P_\fA^\perp \rp  \fZ_* \rp \rp ,
	\end{align*}
	where $P_\fA^\perp = \fI - P_\fA$. 	The second order derivative is given by
	\begin{align*}
	D^2 l_{\mathrm{K}}^n(\alpha;\kappa,\fY,\fZ_*,\fA) 
= 2n^{-1}\fZ_*^\t(I-\kappa P_\fA^\perp)\fZ_* ,
	\end{align*}
	The second derivative is
	and is proportional to the matrix we need to invert in order to solve the normal equation that yields the K-class estimator. As a consequence, we have that the K-class estimator is guaranteed to exist and 
	 be unique if the second derivative is strictly positive definite, i.e., invertible.

Let us first consider $\kappa < 1$.	
	To see that $D^2 l_{\mathrm{K}}^n(\alpha;\kappa,\fY,\fZ_*,\fA)\succ0$, 
	take any $y \in \R^{d_1+q_1}\setminus \{0\}$ and assume that Assumption \Cref{ass:ZtZfullRank} holds. That is, we assume that $\text{rank}(\fZ_*^\t \fZ_*)=\text{rank}(\fZ_*)=d_1+q_1$ almost surely such that  $ z= \fZ_* y \in \R^{n}\setminus \{0\}$ almost surely. 
	Without Assumption \Cref{ass:ZtZfullRank}, choosing $y\in \ker(\fZ_*)\setminus \{0\}$ yields a zero in the following quadratic form with positive probability. However, with this assumption (disregarding $2n^{-1}$) we get that
	\begin{align*}
	y^\t 	D^2 l_{\mathrm{K}}^n(\alpha;\kappa) y & \propto (1-\kappa)y^\t \fZ_*^\t \fZ_* y +  \kappa y^\t \fZ_*^\t P_\fA \fZ_* y  = (1-\kappa)\| z \|_2^2 + \kappa \|P_\fA z\|_2^2 \\
	&\geq \left\{\begin{array}{ll}
	(1-\kappa)\|z\|_2^2+\kappa \|z\|_2^2 = \|z\|_2^2, & \text{if } \kappa \in (-\infty,0) , \\
	(1-\kappa)\|z\|_2^2, & \text{if } \kappa \in [0,1), 
	\end{array}\right. >0.	\end{align*}
    Here, we used that $P_\fA = P_\fA^\t = \fA(\fA^\t\fA)^{-1} \fA^\t$ is an orthogonal projection matrix, hence $P_\fA=P_\fA^\t P_\fA$ and $0 \leq \|P_\fA w\|_2^2 \leq \|w\|_2^2$ for any $w\in \R^q$.

	Let us now consider the case $\kappa = 1$. The quadratic form is now given by
	$$
	y^\t 	D^2 l_{\mathrm{K}}^n(\alpha;\kappa)  y =\|P_\fA z\|_2^2 = y^\t \fZ_*^\t \fA (\fA^\t \fA)^{-1} \fA^\t \fZ_* y.
	$$
	If $\text{rank}(\fA^\t \fZ_*)< d_1+q_1$ with positive probability, then any $y\in \ker(\fA^\t \fZ_*)\setminus \{0\}\not = \emptyset$ yields a zero quadratic value, 	showing that $l_{\mathrm{K}}^n(\alpha;\kappa)$ is not strictly convex with positive probability. However, if Assumption \Cref{ass:AtZfullRank} holds, i.e., that $\fA^\t \fZ_*\in \R^{q\times(d_1+q_1)}$ satisfies $\text{rank}(\fA^\t \fZ_*)=d_1+q_1$ almost surely, then $D^2 l_{\mathrm{K}}^n(\alpha;\kappa)$ is also guaranteed to be positive definite almost surely. 
	
Thus, we have shown sufficient conditions for $D^2 l_{\mathrm{K}}^n(\alpha;\kappa)$ to be almost surely positive definite, ensuring strict convexity of the $l_{\mathrm{K}}^n(\alpha;\kappa)$, hence almost sure uniqueness of a global minimum. The unique global minimum is then found as a solution to the normal equation $
	Dl_{\text{K}}^n(\alpha;\kappa) =0  
$
	which is given by $\hat{\alpha}_{\text{K}}^n (\kappa) = (\fZ_*^\t (\fI-\kappa P_\fA^\perp)\fZ_*)^{-1} \fZ_*^\t(\fI-\kappa P_\fA^\perp)\fY$. We conclude that under the above conditions the K-class estimator $\hat{\alpha}_{\text{K}}^n (\kappa)$ solves the unconstrained minimization problem $
	\argmin_{\alpha\in \R^{d_1+q_1}} l_{\mathrm{K}}^n(\alpha;\kappa)$ 
	almost surely.
\end{proofenv}

 \noindent 
\noindent\begin{proofenv}{\textbf{\Cref{lm:PopulationPenalizedKClassSolutionUniqueAndExists}}}
We first prove that the population estimand that minimizes the population loss function is well-defined. It suffices to show strict convexity of the population loss function. Let Assumption \Cref{ass:VarianceOfZPositiveDefinite} hold, i.e., that $\text{Var}(Z_*)$ is positive definite, and consider $\kappa \in[0,1)$. For any $y\in \R^{d_1+q_1}\setminus \{0\}$ we see that
\begin{align}
y^\t 	D^2 l_{\mathrm{K}}(\alpha;\kappa) y &= (1-\kappa)y^\t E(Z_* Z_*^ \t) y +  \kappa y^\t E(Z_* A^\t) E(A A^\t)^{-1} E(A Z_*^\t) y \notag \\
&\geq (1-\kappa)y^\t E(Z_* Z_*^\t) y = (1-\kappa) \lp y^\t  \text{Var}(Z_*)y + y^\t E(Z_*)E(Z_*)^\t y \rp  \notag\\
& \geq  (1-\kappa)  y^\t  \text{Var}(Z_*)y   > 0,\label{eq:EZZtIsPositiveDefinite} 
\end{align}
proving strict convexity of the K-class penalized loss function. Now let $\kappa=1$ and let Assumption  \Cref{ass:VarianceOfAPositiveDefinite} and Assumption \Cref{ass:EAZtFullColumnRank} hold, i.e., $\text{Var}(A)$ is positive definite and  $E(AZ_*^\t)$ is of full column rank (which implicitly assumes we are in the just- or over-identified case). First note that by the above considerations this implies that $E(AA^\t)$ and its inverse $E(AA^\t)^{-1}$ are positive definite. For any $y\in \R^{d_1+q_1}\setminus \{0\}$ we note that $z:=E(A Z_*^\t) y \not = 0 $ by injectivity of $E(A Z_*^\t)$, and hence $$
y^\t 	D^2 l_{\mathrm{K}}(\alpha;\kappa) y =   z^\t E(A A^\t)^{-1} z > 0,
$$
by the positive definiteness of $E(AA^\t)^{-1}$. Proving strict convexity.

In both setups the minimization estimator of the population loss function solves the normal equation $0 = Dl_{\text{K}}(\alpha;\kappa) = (1-\kappa)Dl_{\text{OLS}}(\alpha) + \kappa Dl_{\text{IV}}(\alpha) $ which by rearranging the terms
yields that
\begin{align*}
\alpha_{\text{K}}(\kappa) =&  \lp (1-\kappa)E(Z_*Z_*^\t) + \kappa E(Z_*A^\t)E(AA^\t)^{-1}E(AZ_*^\t)\rp^{-1} \\
&\cdot \lp (1-\kappa) E(Z_*Y) + \kappa E(Z_*A^\t)E(AA^{\t})^{-1}E(AY)\rp.
\end{align*}
We now prove that the estimators are asymptotically well-defined if
	the population conditions of Assumption \Cref{ass:VarianceOfZPositiveDefinite} and Assumption \Cref{ass:EAZtFullColumnRank}	
	hold. 
	For $\kappa\in[0,1)$, we know from \Cref{lm:PenalizedKClassSolutionUniqueAndExists} that
	\begin{align*}
		P\left[ \argmin_{\alpha\in \R^{d_1+q_1}} l_{\mathrm{K}}^n(\alpha;\kappa,\fY,\fZ_*,\fA) \textit{ is well-defined} \right] \geq  P\left[ \fZ_*^\t \fZ_* \textit{ is positive definite} \right],
	\end{align*}
	So it suffices to show that the lower converges to one in probability. By the weak law of large numbers we have, for any $\ep >0$ that $
	P(\|\fZ_*^\t \fZ_* - E(Z_*Z_*^\t)\| < \ep )  \to 1 \label{eq:ZtZConvp}$. 
	Note that by Assumption \Cref{ass:VarianceOfZPositiveDefinite}, i.e., that $\text{Var}(Z_*)$ is positive definite, we also have that $E(Z_*Z_*^\t)$ is positive definite; see \Cref{eq:EZZtIsPositiveDefinite} above. Note that the set of positive definite matrices $S_+$ is an open set in the space of symmetric matrices $S$ of the same dimensions. Hence, there must exist an open ball $B(E(Z_*Z_*^\t),c)\subseteq S_+$ with center $E(Z_*Z_*^\t)$ and radius $c>0$, fully contained in the set of positive definite matrices. By virtue of the above convergence in probability, we have that
\begin{align*}
		P\left[ \fZ_*^\t \fZ_* \textit{ is positive definite} \right] &\geq  P \lp \fZ_*^\t \fZ_* \in B(E(Z_*Z_*^\t),c)\rp \\& 
	\geq  P(\|\fZ_*^\t \fZ_* - E(Z_*Z_*^\t)\| < c )  \to 1,
\end{align*}
	proving that the estimator minimizing the K-class penalized regression function is asymptotically well-defined. In the case of $\kappa=1$ the argument for asymptotic well-definedness follows by almost the same arguments. Arguing that $\fA^\t\fA$ is positive definite with probability converging to one since $\text{Var}(A)$ is assumed positive definite follows from the same arguments as above.
	To see that $\fA^\t\fZ_*$ is of full column rank with probability converging to one, we use that $E(AZ_*^\t)$ is assumed full column rank. If $q=d_1+q_2$, then follows from the above arguments. Otherwise, if $q> d_1+q_1$, then we modify the above arguments using that the set of injective linear maps from $\R^{d_1+q_1}$ to $\R^q$ is an open set of all linear maps from $\R^{d_1+q_1}$ to $\R^q$. 
	
Finally, by the law of large numbers, Slutsky's theorem and the continuous mapping theorem, one can easily realize that $\hat{\alpha}_{\text{K}}^n (\kappa) \convp \alpha_{\text{K}}(\kappa)$.
\end{proofenv}

 \noindent
 \noindent\begin{proofenv}{\textbf{\Cref{sthm:TheoremIntRobustKclas}}}
	Let $(Y,X,H,A)$ be generated by the SEM given by
	\begin{align} \label{KclassRobustnessEquationSEM}
	[
	Y \, \,  X^\t \, \, H^\t 
	]^\t := B [
	Y \, \, X^\t \, \, H^\t 
	]^\t + MA + \ep,
	\end{align}
	where $\ep$ satisfies $\ep\independent A$ 
	and has jointly independent marginals $\ep_1 \independent \cdots \independent \ep_{d+1+r}$ with finite second moment $E\|\ep\|_2^2<\i$ and mean zero $E(\ep)=0$. The distribution of $A$ is determined independently of \Cref{KclassRobustnessEquationSEM} and with the only requirement that $E\|A\|_2^2<\i$. Note that we have transposed $B$ and $M$ for ease of notation. This implies that $(Y,X,H)$ satisfies the  reduced form equations given by $
	 [
	Y \, \, X^\t \, \, H^\t 
	]^\t = \Pi A + \Gamma^{-1}\ep$, where $\Gamma = I-B$ and $\Pi = \Gamma^{-1} M$. 
	
	Now let $X_*\subset X$ and $A_* \subset A$ be our candidate predictors of $Y$, regardless of which variables directly affect $Y$ and let $Z_* = [
	X_*^\t \, \, A_*^\t
	]^\t $. By the reduced form structural equations we  derive the marginal reduced forms as
	\begin{align} \label{Eq:MarginalReducedFormOfX}
	Y = \Pi_Y A + \Gamma^{-1}_Y \ep \quad \text{and} \quad X_* = \Pi_{X_*} A + \Gamma^{-1}_{X_*} \ep ,
	\end{align}
	where $\Pi_Y,\Pi_{X_*},\Gamma^{-1}_Y,\Gamma^{-1}_{X_*}$ are the relevant sub-matrices of rows from $\Pi$ and $\Gamma^{-1}$.  Furthermore, let $(Y^v,X^v,H^v)$ 
	be generated as a solution to the SEM of \Cref{KclassRobustnessEquationSEM} under the intervention $\text{do}(A:=v)$, where $v\in \mathcal{L}^2(\Omega,\cF,P)$ is any fixed stochastic element uncorrelated with $\ep$. Under the intervention and by  similar manipulations as above, we arrive at the following marginal reduced forms $
	Y^v = \Pi_Y v + \Gamma^{-1}_Y \ep$ and $X_*^v = \Pi_{X_*} v + \Gamma^{-1}_{X_*} \ep.$
For a fixed $\gamma$ and $\beta$, with $A_{-*}$ being $A\setminus A_{*}$, we have that
	\begin{align*}
Y-\gamma^\t X_* - \beta^\t A_* &= (\Pi_Y - \gamma^\t \Pi_{X_*})A +(\Gamma_Y^{-1}-\Gamma_{X_*}^{-1})\ep - \beta^\t A_* \\
&= (\delta_1^\t-\beta^\t)A_* + \delta_2^\t A_{-*} + w^\t \ep = \xi^\t A + w^\t \ep,
\end{align*}
where $\delta_1$, $\delta_2$ are such that
$(\Pi_Y - \gamma^\t \Pi_{X_*})A = \delta_1^\t A_* + \delta_2^\t A_{-*}$, $\xi$  is such that
$\xi^\t A = (\delta_1^\t-\beta^\t)A_* + \delta_2^\t A_{-*} $  and $w^\t := (\Gamma_Y^{-1}-\Gamma_{X_*}^{-1})$.  Similar manipulations yield that 
the regression residuals 
under the intervention are given by
$
Y^v-\gamma^\t X_*^v - \beta^\t v_* = \xi^\t v + w^\t \ep. %
$
Since $A \independent \ep$ and $\ep$ has mean zero, we have that
\begin{align}
E(Y-\gamma^\t X_* - \beta^\t A_*| A) &= \xi^\t A + w^\t E(\ep) = \xi^\t A, \label{KclassRobustnessEquationRegResidualsCondMeanGivenA} \\
Y-\gamma^\t X_* - \beta^\t A_* - E(Y-\gamma^\t X_* - \beta^\t A_*| A) &= w^\t \ep. \label{KclassRobustnessEquationRegResidualsMinusRegResidualsCondMeanOnA}
\end{align}
By construction $E(v\ep^\t)=0$, so
\begin{align}
E^{\text{do}(A:=v)}\left[ \lp Y-\gamma^\t X_* - \beta^\t A_* \rp^2  \right] &= E\left[ \lp \xi^\t v + w^\t \ep \rp^2  \right] \notag \\
&= E\left[ \lp\xi^\t v \rp^2 \right] + E\left[ \lp w^\t \ep \rp^2 \right] + \xi^\t E(v\ep^\t)w \notag \\
&= E\left[ \lp\xi^\t v \rp^2 \right] + E\left[ \lp w^\t \ep \rp^2 \right]. \label{KclassRobustnessEquationEqExpectedResidualsUnderInt}
 \end{align}
We investigate the terms of \Cref{KclassRobustnessEquationEqExpectedResidualsUnderInt} and note by \Cref{KclassRobustnessEquationRegResidualsMinusRegResidualsCondMeanOnA} that
\begin{align}
 E\left[ \lp w^\t \ep \rp^2 \right]  &=  E\left[ \lp Y-\gamma^\t X_* - \beta^\t A_* - E(Y-\gamma^\t X_* - \beta^\t A_*| A)\rp^2 \right]  \notag \\
& =  E\left[ \lp Y-\gamma^\t X_* - \beta^\t A_* \rp^2 \right] +E\left[   E \lp Y-\gamma^\t X_* - \beta^\t A_*| A\rp^2 \right] \label{KclassRobustnessEquationSecondTerm}\\
 &\qquad - 2 E\left[ \lp Y-\gamma^\t X_* - \beta^\t A_*\rp  E(Y-\gamma^\t X_* - \beta^\t A_*| A) \right]. \notag 
\end{align}
In \Cref{KclassRobustnessEquationRegResidualsCondMeanGivenA} we established that $E(Y-\gamma^\t X_* - \beta^\t A_*| A)$ is a linear function of $A$, so it must hold that
\begin{align*}
E(Y-\gamma^\t X_* - \beta^\t A_*| A) &= \argmin_{Z\in\sigma(A)}\|Y-\gamma^\t X_* - \beta^\t A_*-Z\|_{L^2(P)}^2 \\
&=A^\t \argmin_{c \in \R^{q}}\|Y-\gamma^\t X_* - \beta^\t A_*-  A^\t c\|_{L^2(P)}^2  \\
&= A^\t E(AA^\t)^{-1} E\left[ A\lp Y-\gamma^\t X_* - \beta^\t A_*\rp \right],
\end{align*}
almost surely. In the first equality we used that the conditional expectation is the best predictor under the $L^2(P)$-norm and in the third equality we used that the minimizer is given by the population ordinary least square estimate. 
An immediate consequence of this is that the second term of \Cref{KclassRobustnessEquationSecondTerm} equals 
\begin{align*}
		E\left[  E \lp Y-\gamma^\t X_* - \beta^\t A_*|A  \rp^2 \right] &= E[(Y-\gamma^\t X_* - \beta^\t A_*)A^\t]E(AA^\t)^{-1}\\
	&\quad \quad \cdot  E[A(Y-\gamma^\t X_*- \beta^\t A_*)],
\end{align*}
	which is seen to be of the same form of the third term in \Cref{KclassRobustnessEquationSecondTerm},
	\begin{align*}
	&E \left[ \lp Y-\gamma^\t X_* - \beta^\t A_*\rp  E(Y-\gamma^\t X_* - \beta^\t A_*|A)  \right] \\	= &E \left[ \lp Y-\gamma^\t X_* - \beta^\t A_*\rp A^\t \right] E(AA^\t)^{-1}  E\left[A(Y-\gamma^\t X_* - \beta^\t A_*)\right].
	\end{align*}
	Thus, we conclude that the second term of \Cref{KclassRobustnessEquationEqExpectedResidualsUnderInt} is given by
	\begin{align*}
	E \left[\lp w^\t \ep \rp^2 \right] &= E \left[\lp Y-\gamma^\t X_* - \beta^\t A_* \rp^2 \right]  - E\left[  E \lp Y-\gamma^\t X_* - \beta^\t A_*|A \rp^2 \right] \\
	&= l_{\mathrm{OLS}}(\alpha;Y,Z_*)- l_{\mathrm{IV}}(\alpha;Y,Z_*,A). 
	\end{align*}
Taking the supremum over all $v\in C(\kappa)$ of the first term of \Cref{KclassRobustnessEquationEqExpectedResidualsUnderInt} we obtain
	\begin{align*}
	\sup_{v\in C(\kappa)}E \left[\lp \xi^\t v \rp^2 \right] &= \sup_{v\in C(\kappa)}\xi^\t E \left[   vv^\t  \right] \xi = \frac{1}{1-\kappa}\xi^\t E \left[  AA^\t \right] \xi = \frac{1}{1-\kappa} E \left[    \lp \xi^\t A \rp^2 \right] \\
	&= \frac{1}{1-\kappa} E \left[    E \lp Y-\gamma^\t X_* - \beta^\t A_*|A \rp^2 \right] = \frac{1}{1-\kappa} l_{\mathrm{IV}}(\alpha;Y,Z_*,A),
	\end{align*}
where the 
second last equation follows from 
\Cref{KclassRobustnessEquationRegResidualsCondMeanGivenA}
and the
second equation follows from the following argument.	For any $v\in C(\kappa)$ we have that $E(vv^\t) \preceq \frac{1}{1-\kappa} E(AA^\t)$, that is,  for all $x\in \R^q$ it holds that $\frac{1}{1-\kappa} x^\t E(AA^\t) x \geq x^\t E(vv^\t)x$, which implies that the upper bound is attained for any $v$ such that $E(vv^\t) = \frac{1}{1-\kappa}E(AA^\t)$.
	Thus, we have that 
	\begin{align*}
	\sup_{v\in C(\kappa)}  E^{\mathrm{do}(A:=v)}\left[ (Y - \gamma^\t X_*  - \beta^\t A_*  )^2 \right] =&  \sup_{v\in C(\kappa)}  E\left[ \lp\xi^\t v \rp^2 \right] + E\left[ \lp w^\t \ep \rp^2 \right]  \\
	=& l_{\mathrm{OLS}}(\alpha;Y,Z_*)+ \frac{\kappa }{1-\kappa} l_{\mathrm{IV}}(\alpha;Y,Z_*,A).
	\end{align*}
	By the representation in  \Cref{eq:KclassLossFunctionAsPenalizedOLS} it therefore follows that the population K-class estimate with parameter $\kappa\not = 1$ is given as the estimate that minimizes the worst case mean squared prediction error over all interventions contained in $C(\kappa)$, that is,
	\begin{align*}
	\alpha_{\mathrm{K}}(\kappa;Z_*,A) &= 	\argmin_{\gamma\in \R^d,\beta\in \R^{q_1}}\sup_{v\in C(\kappa)}  E^{\mathrm{do}(A:=v)}\left[ (Y - \gamma^\t X_*  - \beta^\t A_*  )^2 \right] .
	\end{align*}
\end{proofenv}

\section{Proofs of Selected Results in Section~\ref{SEC:PULSE}}
\label{sec:SomeProofsOfSecPULSE}
\medskip

\begin{restatable}[K-class estimators differ]{corollary}{KclassSolutionsDistinct}
	\label{cor:KclassSolutionsDistinctApp}
	Let 
	\Cref{ass:KclassNotInIV,ass:ZtZfullrankandAtZfullrank} hold.
	If $\lambda_1,\lambda_2\geq 0$ with $\lambda_1\not = \lambda_2$, then $\hat{\alpha}_{\mathrm{K}}^n(\lambda_1) \not = \hat{\alpha}_{\mathrm{K}}^n(\lambda_2)$.
\end{restatable}
\noindent\begin{proofenv}{\Cref{cor:KclassSolutionsDistinctApp}}
	Let \Cref{ass:ZtZfullrankandAtZfullrank,ass:KclassNotInIV} hold. $\hat{\alpha}_{\text{K}}^n(\lambda)$ is well-defined for all $\lambda \geq 0$ by \Cref{lm:PenalizedKClassSolutionUniqueAndExists}. Let $\lambda_1,\lambda_2\geq 0$
	with $\lambda_1 \not = \lambda_2$ and note that the orthogonality condition derived in the proof of \Cref{lm:KclassNotEqualToTwoSLS} also applies here. That is, $
	\la \fY- \fZ \hat{\alpha}_{\text{K}}^n(\lambda_i), (\fI + \lambda_i P_\fA) z\ra =0, $ for all $z\in \cR(\fZ)$ and  $i=1,2$. Assume for contradiction that $\hat{\alpha}_{\text{K}}^n(\lambda_1) = \hat{\alpha}_{\text{K}}^n(\lambda_2)$. This implies that
	\begin{align*}
	0&=\la \fY - \fZ \hat{\alpha}_{\mathrm{K}}^n(\lambda_1),  (\fI + \lambda_1 P_\fA) z - (\fI + \lambda_2 P_\fA) z\ra \\
	&= \la \fY - \fZ \hat{\alpha}_{\mathrm{K}}^n(\lambda_1),   (\lambda_1-\lambda_2) P_\fA  z\ra = (\lambda_1-\lambda_2) \la \fY - \fZ \hat{\alpha}_{\mathrm{K}}^n(\lambda_1),   P_\fA  z\ra,
	\end{align*}
	for any $z\in \cR(\fZ)$. Thus, by symmetry and idempotency of $P_\fA$ we have that for all $z\in \cR(\fZ)$, $$
	\la P_\fA \fY - P_\fA \fZ \hat{\alpha}_{\mathrm{K}}^n(\lambda_1),   P_\fA  z\ra = \la \fY - \fZ \hat{\alpha}_{\mathrm{K}}^n(\lambda_1),   P_\fA  z\ra =0.$$
	That is, $P_\fA \fZ \hat{\alpha}_{\mathrm{K}}^n(\lambda_1)$ is the orthogonal projection of $P_\fA \fY$ onto $\cR(P_\fA \fZ)$.
	This is equivalent with saying that $\hat{\alpha}_{\mathrm{K}}^n(\lambda_1) \in \cM_{\text{IV}}$ as the space of minimizers of $l_{\text{IV}}^n$ are exactly the coefficients in $\R^{d_1+q_1}$ which mapped through $P_\fA \fZ$ yields this orthogonal projection. See the proof of \Cref{lm:KclassNotEqualToTwoSLS} for further elaboration on this equivalence.
	This is a contradiction to \Cref{ass:KclassNotInIV}, hence $\hat{\alpha}_{\text{K}}^n(\lambda_1) \not =  \hat{\alpha}_{\text{K}}^n(\lambda_2)$.
\end{proofenv}

\begin{restatable}[Monotonicity of the losses and the test statistic]{lemma}{MonotonicityOfTestOfLambda} \textcolor{white}{,}\label{lm:OLSandIV_Monotonicity_FnctOfPenaltyParameterLambdaApp} \\When Assumption \Cref{ass:ZtZfullrank} holds the maps $
	[0,\i)\ni \lambda \mapsto  l_{\mathrm{OLS}}^n(\hat{\alpha}_{\mathrm{K}}^n (\lambda) )$ and $ [0,\i)\ni  \lambda \mapsto l_{\mathrm{IV}}^n(\hat{\alpha}_{\mathrm{K}}^n (\lambda) ) $
	are monotonically increasing and monotonically decreasing, respectively. Consequently, if \Cref{ass:ZYfullcolrank} holds, we have that the map 
	$
	[0,\i)\ni \lambda \longmapsto T_n (\hat{\alpha}_{\mathrm{K}}^n (\lambda) ) 
	$
	is monotonically decreasing. Furthermore, if \Cref{ass:KclassNotInIV} also holds, these monotonicity statements can be strengthened to strictly decreasing and strictly increasing.
\end{restatable}
\noindent\begin{proofenv}{\Cref{lm:OLSandIV_Monotonicity_FnctOfPenaltyParameterLambdaApp}}
Let Assumption \Cref{ass:ZtZfullrank} hold, such that $\hat{\alpha}_{\text{K}}^n(\lambda)$ is well-defined for all $\lambda\geq 0$; see \Cref{lm:PenalizedKClassSolutionUniqueAndExists}.
	Let $\lambda_2 > \lambda_1 \geq 0$ and note that
	\begin{align*}
	l_{\mathrm{OLS}}^n(\hat{\alpha}_{\mathrm{K}}^n (\lambda_1) ) &+ \lambda_1 l_{\mathrm{IV}}^n(\hat{\alpha}_{\mathrm{K}}^n (\lambda_1) )  \leq l_{\mathrm{OLS}}^n(\hat{\alpha}_{\mathrm{K}}^n (\lambda_2) ) + \lambda_1 l_{\mathrm{IV}}^n(\hat{\alpha}_{\mathrm{K}}^n (\lambda_2) ) \\
	&= l_{\mathrm{OLS}}^n(\hat{\alpha}_{\mathrm{K}}^n (\lambda_2) ) + \lambda_2 l_{\mathrm{IV}}^n(\hat{\alpha}_{\mathrm{K}}^n (\lambda_2) ) + (\lambda_1-\lambda_2)l_{\mathrm{IV}}^n(\hat{\alpha}_{\mathrm{K}}^n (\lambda_2) ) \\
	&\leq l_{\mathrm{OLS}}^n(\hat{\alpha}_{\mathrm{K}}^n (\lambda_1) ) + \lambda_2 l_{\mathrm{IV}}^n(\hat{\alpha}_{\mathrm{K}}^n (\lambda_1) ) + (\lambda_1-\lambda_2)l_{\mathrm{IV}}^n(\hat{\alpha}_{\mathrm{K}}^n (\lambda_2) ),
	\end{align*}
	where we used that $\hat{\alpha}_{\mathrm{K}}^n (\lambda)$ minimizes the expressions with penalty factor $\lambda$. Thus, $$
	(\lambda_1-\lambda_2)l_{\mathrm{IV}}^n(\hat{\alpha}_{\mathrm{K}}^n (\lambda_1) ) \leq (\lambda_1-\lambda_2)l_{\mathrm{IV}}^n(\hat{\alpha}_{\mathrm{K}}^n (\lambda_2) ),$$ which is equivalent with $$ l_{\mathrm{IV}}^n(\hat{\alpha}_{\mathrm{K}}^n (\lambda_1) ) \geq l_{\mathrm{IV}}^n(\hat{\alpha}_{\mathrm{K}}^n (\lambda_2) ),$$ 
	proving that $\lambda \mapsto l_{\mathrm{IV}}^n(\hat{\alpha}_{\mathrm{K}}^n (\lambda) )$ is monotonically decreasing. 
	
	If $\lambda_2 > \lambda_1 =0$, then we note that $$
	l_{\mathrm{OLS}}^n(\hat{\alpha}_{\mathrm{K}}^n (\lambda_1) ) = \min_{\alpha} \{l_{\mathrm{OLS}}^n( \alpha) \}  \leq l_{\mathrm{OLS}}^n(\hat{\alpha}_{\mathrm{K}}^n (\lambda_2) ).$$	For any $\lambda >0$, $$
	\hat{\alpha}_{\mathrm{K}}^n (\lambda) = \argmin_{\alpha} \{ l_{\mathrm{OLS}}^n(\alpha) + \lambda l_{\mathrm{IV}}^n(\alpha) \} 
	= \argmin_{\alpha} \{ \lambda^{-1}l_{\mathrm{OLS}}^n(\alpha) +  l_{\mathrm{IV}}^n(\alpha) \}.$$ 
	Thus, if $\lambda_2> \lambda_1 > 0,$ we have that
	\begin{align*}
	\lambda_1^{-1}l_{\mathrm{OLS}}^n(\hat{\alpha}_{\mathrm{K}}^n (\lambda_1) ) +&  l_{\mathrm{IV}}^n(\hat{\alpha}_{\mathrm{K}}^n (\lambda_1) ) 
	 \leq \lambda_1^{-1}l_{\mathrm{OLS}}^n(\hat{\alpha}_{\mathrm{K}}^n (\lambda_2) ) +  l_{\mathrm{IV}}^n(\hat{\alpha}_{\mathrm{K}}^n (\lambda_2) ) \\
	= &\lambda_2^{-1}l_{\mathrm{OLS}}^n(\hat{\alpha}_{\mathrm{K}}^n (\lambda_2) ) +  l_{\mathrm{IV}}^n(\hat{\alpha}_{\mathrm{K}}^n (\lambda_2) ) + (\lambda_1^{-1}-\lambda_2^{-1})l_{\mathrm{OLS}}^n(\hat{\alpha}_{\mathrm{K}}^n (\lambda_2) ) \\
	\leq &\lambda_2^{-1}l_{\mathrm{OLS}}^n(\hat{\alpha}_{\mathrm{K}}^n (\lambda_1) ) +  l_{\mathrm{IV}}^n(\hat{\alpha}_{\mathrm{K}}^n (\lambda_1) ) + (\lambda_1^{-1}-\lambda_2^{-1})l_{\mathrm{OLS}}^n(\hat{\alpha}_{\mathrm{K}}^n (\lambda_2)),
	\end{align*}
	hence $
	l_{\mathrm{OLS}}^n(\hat{\alpha}_{\mathrm{K}}^n (\lambda_1) ) \leq l_{\mathrm{OLS}}^n(\hat{\alpha}_{\mathrm{K}}^n (\lambda_2) ),$
	so $\lambda \mapsto l_{\mathrm{OLS}}^n(\hat{\alpha}_{\mathrm{K}}^n (\lambda) )$ is monotonically increasing. 	

	When \Cref{ass:ZYfullcolrank} holds, the map $$
	\lambda \mapsto T_n (\hat{\alpha}_{\mathrm{K}}^n (\lambda) ) = n  \frac{l_{\text{IV}}^n(\hat{\alpha}_{\mathrm{K}}^n (\lambda))}{l_{\mathrm{OLS}}^n(\hat{\alpha}_{\mathrm{K}}^n (\lambda))} ,$$
	is well-defined and monotonically decreasing, as it is given by a positive,  monotonically decreasing function over a strictly positive and monotonically increasing function.
	
	Furthermore, when \Cref{ass:KclassNotInIV} holds, 
	\Cref{cor:KclassSolutionsDistinct} yields that for $\lambda_1,\lambda_2\geq0 $ with $\lambda_1 \not = \lambda_2$ it holds that $\hat{\alpha}_{\text{K}}^n(\lambda_1) \not = \hat{\alpha}_{\text{K}}^n(\lambda_2)$. 
	As a consequence, the above inequalities become strict, 
	since
	otherwise (Dual.$\lambda.n$) has two distinct solutions which contradicts
	\Cref{lm:PenalizedKClassSolutionUniqueAndExists}. Replacing the above inequalities with strict inequalities yields that the functions are strictly increasing and decreasing, respectively.
\end{proofenv}

\begin{restatable}[]{lemma}{TestInAlphaLambdaStarEqualsQuantile}
	\label{lm:TestInAlphaLambdaStarEqualsQuantileApp}
	Let $p_{\min}\in(0,1)$ and let Assumption \Cref{ass:ZtZfullrank} and \Cref{ass:ZYfullcolrank} hold. If $\lambda_n^\star(p_{\min})<\infty$, it holds that \begin{align} \label{eq:TestInAlphaLambdaStarLessQ}
	T_n(\hat{\alpha}_{\mathrm{K}}^n (\lambda_n^\star(p_{\min}))) \leq  Q_{\chi^2_{q}}(1-p_{\min}).
	\end{align}
	If the ordinary least square estimator  satisfies
	$T_n(\hat{\alpha}_{\mathrm{OLS}}^n)< Q_{\chi^2_{q}}(1-p_{\min})$, then \Cref{eq:TestInAlphaLambdaStarLessQ} holds with strict inequality, otherwise it holds with equality.
\end{restatable}
\noindent\begin{proofenv}{\Cref{lm:TestInAlphaLambdaStarEqualsQuantileApp}}
	Let $p_{\min}\in(0,1)$ and let let Assumption \Cref{ass:ZtZfullrank} and \Cref{ass:ZYfullcolrank} hold, such that $\hat{\alpha}_{\text{K}}^n(\lambda)$ for all $\lambda\geq 0$ and $T_n(\alpha)$ for all $\alpha\in \R^{d_1+q_1}$ are well-defined, by \Cref{lm:PenalizedKClassSolutionUniqueAndExists}.

	Assume that $\lambda_n^\star(p_{\min})< \i$, so we know that $T_n(\hat{\alpha}_{\text{K}}^n(\lambda)) \leq Q_{\chi^2_{q}}(1-p)$ for all $\lambda > \lambda_n^\star(p_{\min})$ by the monotonicity of  \Cref{lm:OLSandIV_Monotonicity_FnctOfPenaltyParameterLambda}. 
	Thus, the first statement follows
	if we can show that $\lambda \mapsto T_n(\hat{\alpha}_{\text{K}}^n(\lambda))$ is a continuous function. Since $\alpha \mapsto T_n(\alpha)$ is continuous it suffices to show that $[0,\infty) \ni \lambda \mapsto \hat{\alpha}_{\text{K}}^n(\lambda)$ is continuous. Recall that $
	\hat{\alpha}_{\text{K}}^n(\lambda) =(\fZ^\t (\fI+\lambda P_\fA)\fZ)^{-1} \fZ^\t(\fI+\lambda P_\fA)\fY, $	for any $\lambda \geq 0$. Note that the functions
	$\text{Inv}:S_{++}^{d_1+q_1} \to S_{++}^{d_1+q_1}$ given by $\fM\stackrel{\text{Inv}}{\mapsto} \fM^{-1}$,  $\lambda \mapsto \fZ^\t (\fI+\lambda P_\fA)\fZ$, $\lambda \mapsto  \fZ^\t(\fI+\lambda P_\fA)\fY$ and $(\fB,\fC) \mapsto \fB \fC$ 
	are all continuous maps, where $S_{++}^{d_1+q_1}$ is the set of all positive definite $(d_1+q_1)\times(d_1+q_1)$ matrices. We have that $\lambda \mapsto \hat{\alpha}_{\text{K}}^n(\lambda)$ is a composition of these continuous maps, hence it itself is continuous. This proves the first statement.

	In the case that OLS is strictly feasible in the PULSE problem, $T_n(\hat{\alpha}_{\text{OLS}}^n)<  Q_{\chi^2_{q}}(1-p_{\min}) $, we have that
	$$
	\lambda^\star(p_{\min})=\inf \left\{  \lambda \geq 0  :   T_n ( \hat{\alpha}_{\text{K}}^n (\lambda) )\leq Q_{\chi^2_{q}}(1-p_{\min}) \right\} = 0,
	$$
	since $\hat{\alpha}_{\text{K}}^n(0) = \hat{\alpha}_{\text{OLS}}^n$,	hence
	$$
	T_n(\hat{\alpha}_{\mathrm{K}}^n (\lambda_n^\star(p_{\min}))) =T_n(\hat{\alpha}_{\mathrm{K}}^n (0)) = T_n(\hat{\alpha}_{\text{OLS}}^n) <  Q_{\chi^2_{q}}(1-p_{\min}).
	$$
	Similar arguments show that, if the OLS is just-feasible in the PULSE problem, $T_n(\hat{\alpha}_{\text{OLS}}^n)= Q_{\chi^2_{q}}(1-p_{\min})$, then $T_n(\hat{\alpha}_{\mathrm{K}}^n (\lambda_n^\star(p_{\min})))= Q_{\chi^2_{q}}(1-p_{\min})$.

	In the case that the OLS estimator is infeasible in the PULSE problem, $Q_{\chi^2_{q}}(1-p_{\min}) < T_n(\hat{\alpha}_{\text{OLS}}^n)$, continuity and monotonicity of $\lambda \mapsto T_n(\hat{\alpha}_{\text{K}}^n(\lambda))$ 
	entail it must hold that $T_n(\hat{\alpha}_{\text{K}}^n(\lambda_n^\star(p_{\min})))=Q_{\chi^2_{q}}(1-p_{\min})$, 
	as otherwise $$T_n(\hat{\alpha}_{\text{K}}^n(\lambda_n^\star(p_{\min})))<Q_{\chi^2_{q}}(1-p_{\min}) < T_n(\hat{\alpha}_{\text{K}}^n(0)),$$
	implying that there exists $\tilde{\lambda} < \lambda_n^\star(p_{\min})$ such that
	$T_n(\hat{\alpha}_{\text{K}}^n(\tilde{\lambda}))\leq Q_{\chi^2_{q}}(1-p_{\min})$, contradicting $\lambda_n^\star(p_{\min}) = \inf\{\lambda \geq 0 : T_n(\hat{\alpha}_{\mathrm{K}}^n (\lambda))\leq Q_{\chi^2_{q}}(1-p_{\min}) \}$.
\end{proofenv}

\section{Proofs of Remaining Results in Section~\ref{SEC:PULSE}}
\label{sec:RemainingProofsOfSecPULSE}

\medskip

\noindent
\noindent\begin{proofenv}{\textbf{\Cref{prop:TestingVanishingCorr}}}
	We want to show an asymptotic guarantee that type I errors (rejecting a true hypothesis) occur with probability $p$. That is,  if $\cH_0(\alpha)$ is true, then $P(T_n^c(\alpha) >  Q_{\chi^2_{q}}(1-p) ) \stackrel{n\to\i }{\longrightarrow} p$. Furthermore, we want to show that for any fixed alternative, the probability of	type II errors (failure to reject) converges to zero. That is, if $P$ is such that	
	$\cH_0(\alpha)$ is false, then $P(T_n^c(\alpha) \leq  Q_{\chi^2_{q}}(1-p)) \stackrel{n\to\i }{\longrightarrow} 0$.

	Fix any $\alpha\in \R^{d_1+q_1}$. It suffices to show that under the null-hypothesis
	$T_n^c(\alpha)$ is asymptotically Chi-squared distributed with $q$ degrees of freedom and that $T_n^c(\alpha)$ tends to infinity under any fixed alternative. Without loss of generality assume that $c(n)=n$ for all $n\in \N$ and recall that
$$
	T_n^c(n)= T_n(\alpha) = n \frac{l_{\text{IV}}^n(\alpha)}{l_{\text{OLS}}^n(\alpha)} 
	=  n \frac{\|P_\fA (\fY-  \fZ \alpha) \|_2^2}{ \|\fY- \fZ \alpha\|_2^2} .
$$
	By the idempotency of $P_\fA$ the 
numerator can be rewritten as $$
	\|P_\fA (\fY-  \fZ \alpha) \|_2^2 =   \|(\fA^\t \fA)^{-1/2} \fA^\t  \fR(\alpha) \|_2^2,
$$
	while the denominator takes the form $\|\fR(\alpha)\|_2^2$. Here, $\fR(\alpha) := \fY - \fZ \alpha$ and $R(\alpha):= Y-Z^\t\alpha$ denotes the empirical and population regression residuals, respectively. \Cref{ass:ZYfullcolrank} ensures that $T_n$ is well-defined on the entire domain of $\R^{d_1+q_1}$ as the denominator is never zero. Furthermore, note that both $R(\alpha)$ for any $\alpha\in \R^{d_1+q_1}$ and $A_i$ for any $i=1,...,q$ have finite second moments
by virtue of Assumption \Cref{ass:SecondMomentA}. %

Assume that the null hypothesis of zero correlation between the components of
$A$ and the regression residuals $R(\alpha)$ holds. First we show that the null hypothesis, under the stated assumptions, implies independence between the exogenous variables $A$ and the regression residuals $R(\alpha)$.
It holds that $E(AR(\alpha)) = E(A)E(R(\alpha))=0$ by Assumption \Cref{ass:MeanZeroA}, the mean zero assumption of $A$. Assumption \Cref{ass:AIndepUy}, i.e., $A\independent U_Y$, yields that
\begin{align} \label{Eq:TempMomentOfProductAandResiduals}
0 &= E(AR(\alpha)) 
= E(AZ^\t)(\alpha_0-\alpha) + E(AU_Y) = E(AZ^\t)(\alpha_0-\alpha),
\end{align}
proving that $\alpha - \alpha_0 = w$ 
for some $w \in \text{kern}(E(AZ^\t))$. Recall that
the marginal structural equation of \Cref{Eq:MarginalReducedFormOfX} states that $X_* = \Pi_{X_*} A + \Gamma^{-1}_{X_*} \ep$. Thus, $Z$ has 
the following representation
\begin{align*}
Z = \begin{bmatrix}
X_* \\ A_*
\end{bmatrix} =\begin{bmatrix}
\Pi_{X_*} A + \Gamma^{-1}_{X_*} \ep\\ A_*
\end{bmatrix}  = \begin{bmatrix}
\Pi_{X_*}^{(*)} & \Pi_{X_*}^{(-*)} \\
I & 0
\end{bmatrix} \begin{bmatrix}
A_* \\ A_{-*}
\end{bmatrix} + \begin{bmatrix}
\Gamma_{X_{*}}^{-1} \\ 0 
\end{bmatrix} \ep =: \Lambda A + \Psi \ep,
\end{align*}
where $\Pi_{X_*} = [
\Pi_{X_*}^{(*)} \, \, \Pi_{X_*}^{(-*)} ] \in \R^{d_1 \times(q_1+q_2)}$ and $\Lambda$, $\Psi$ are the 
conformable block-matrices. Since $A\independent \ep$ by Assumption \Cref{ass:AindepEp} we have that $E(A\ep^\t)=0$, hence $$
0=E(AZ^\t) w = E(AA^\t)\Lambda^\t w + E(A\ep^\t)\Psi^\t w = E(AA^\t)\Lambda^\t w.$$ 
This proves that $\Lambda^\t w =0$  as $E(AA^\t)$ is of full rank by Assumption \Cref{ass:VarianceOfAPositiveDefinite}. Hence, 
\begin{align*}
	R(\alpha) &= Y-Z^\t \alpha = Z^\t(\alpha_0-\alpha)+ U_Y 
= Z^\t w +U_Y \\
&= A^\t \Lambda^\t w + \ep^\t \Psi^\t w +U_Y =\ep^\t \Psi^\t w +U_Y.
\end{align*}
Furthermore, $U_Y=  \alpha_{0,-*}^\t Z_{-*}+\eta^\t_0  H + \ep_Y$ can be written as a linear function of $A$ plus a linear function of $\ep$. To realize this, simply express $Z_{-*}$ and $H$ by their marginal reduced form structural equations. Hence, the assumptions that $A \independent U_Y$ must entail that $A$ vanishes from the expression of $U_Y$. As a consequence we have that $R(\alpha)$ is a linear function only of $\ep$, from which the assumption that $A\independent \ep$ yields that $A \independent R(\alpha)$. That is, the null hypothesis of zero correlation implies independence in the linear structural equation model, 
under the given assumptions. Thus, $E\|AR(\alpha)\|_2^2 = E\|A\|_2^2 E\|R(\alpha)\|_2^2<\i$, so the covariance matrix of $AR(\alpha)$ is well-defined.

By the established independence and \Cref{Eq:TempMomentOfProductAandResiduals}, the covariance matrix of $AR(\alpha)$ has the following representation
	$$
	\Cov( AR(\alpha) )=  
	E(AA^\t)E(R(\alpha)^2) \succ 0.$$
	The positive definiteness follows from the facts that $E(AA^\t)\succ 0$ and $E(R(\alpha)^2)>0$ for any $\alpha\in \R^{d_1+q_1}$. $E(AA^\t)\succ 0$ follows by Assumption \Cref{ass:VarianceOfAPositiveDefinite} and $E(R(\alpha)^2)>0$ for any $\alpha\in \R^{d_1+q_1}$ follows by  Assumption \Cref{ass:SpectralRadiusOfBLessThanOne}, Assumption \Cref{ass:epIndependentMarginals} and \Cref{ass:NonDegenYNoise};  non-degeneracy and mutual independence of the marginal noise variables in $\ep$. To see this, expand $R(\alpha)$ in terms of the marginal reduced form structural equations of $Y$ and $Z$ and use that $(I-B^\t)$ is invertible to see that $\ep$ does not vanish in the expression $R(\alpha)$. The multi-dimensional Central Limit Theorem yields that
	$$
	\frac{1}{\sqrt{n}}\fA^\t \fR(\alpha) = \sqrt{n} \lp \frac{1}{n}\sum_{i=1}^n \begin{pmatrix}	A_{i,1}R(\alpha)_{i} \\ 	\vdots \\	A_{i,q} R(\alpha)_{i}	\end{pmatrix}\rp  
	\convd \cN(0,\Cov(AR(\alpha))).$$
	Furthermore, note that regardless of whether or not the null-hypothesis is true, we have that $$
	\sqrt{n}(\fA^\t \fA)^{-1/2} 
	 \convp E(AA^\t)^{-1/2},
$$
	 and
	$$
	\frac{1}{\sqrt{n}}\|\fR(\alpha)\|_2 = \sqrt{\frac{1}{n}\sum_{i=1}^n R(\alpha)_i^2} 
	\convp \sqrt{E(R(\alpha)^2)}>0,
	$$
	by the law of large numbers and the continuity of the matrix square root operation on the cone of symmetric positive-definite matrices.	
	We can represent the test-statistic as $
	T_n(\alpha):= \| \sqrt{n}W_n(\alpha)\|_2^2$ with $$W_n(\alpha):= (\fA^\t \fA)^{-1/2}\fA^\t \fR(\alpha)/\|\fR(\alpha)\|_2,$$ 
	and have that $
	\sqrt{n}W_n(\alpha) 
	\convd W \sim  \cN\lp 0, I \rp$, 
	by Slutsky's theorem and linear transformation rules of multivariate normal distributions.	Hence, 
	the continuous mapping theorem yields that $$
	T_n(\alpha) = \left\|\sqrt{n}W_n(\alpha) \right\|_2^2 \convd \|W\|_2^2 = \sum_{i=1}^qW_i^2 \sim  \chi^2_{q},$$ 
	where $\chi^2_{q}$ is the Chi-squared distribution with $q$ degrees of freedom, since $W_1\independent \cdots \independent W_q$. This proves that the test-statistic $T_n$ has the correct asymptotic distribution under the null-hypothesis.

	Now fix a distribution $P$, for which the null hypothesis of simultaneous zero correlation between the components of $A$ and the residuals $R(\alpha)$ does not hold. That is, there exists an $j\in\{1,...,q\}$ such that $E(A_{j}R(\alpha)) \not = E(A_j)E(R(\alpha))=0$.  Note that 
	\begin{align*}
	\left\| n^{-1/2}(\fA^\t \fA)^{1/2} \right \|_{\text{op}}^2 T_n(\alpha) &= \left\| n^{-1/2}(\fA^\t \fA)^{1/2} \right \|_{\text{op}}^2 \left\| \frac{\sqrt{n}(\fA^\t \fA)^{-1/2}\frac{1}{\sqrt{n}}\fA^\t \fR(\alpha)}{\frac{1}{\sqrt{n}}\|\fR(\alpha)\|_2} \right\|_2^2 \\
	&\geq \left\| \frac{\frac{1}{\sqrt{n}}\fA^\t \fR(\alpha)}{\frac{1}{\sqrt{n}}\|\fR(\alpha)\|_2} \right\|_2^2 \geq \left| \frac{\frac{1}{\sqrt{n}}\fA_j^\t \fR(\alpha)}{\frac{1}{\sqrt{n}}\|\fR(\alpha)\|_2} \right|^2= n \left| \frac{ \frac{1}{n} \fA_j^\t \fR(\alpha)}{\frac{1}{\sqrt{n}}\|\fR(\alpha)\|_2} \right|^2, %
	\end{align*}
	where $\fA_j^\t := (\fA_j)^\t$ and $\fA_j$ is the \textit{j}'th column
	of $\fA$ corresponding to the i.i.d.\ vector consisting of $n$ copies of the \textit{j}'th exogenous variable $A_j$ and $\|\cdot\|_{\text{op}}$ is the operator norm.
The lower bound diverges to infinity in probability
as the latter factor tends to $|E(A_jR(\alpha)) / \sqrt{E(R(\alpha)^2)}|^2 > 0$ in probability by the law of large numbers and Slutsky's theorem. Hence, it holds that $T_n(\alpha) \convp \i,$
as $$\left\| n^{-1/2}(\fA^\t \fA)^{1/2} \right \|_{\text{op}} \to \left\| E(A A^\t )^{1/2} \right\|_{\text{op}} \in(0,\i).$$ 
  This concludes the proof.
\end{proofenv}

\noindent
\noindent\begin{proofenv}{\textbf{\Cref{lm:PrimalUniqueSolAndSlatersConditions}}}
	Let \Cref{ass:ZtZfullrankandAtZfullrank} hold, i.e., that $\fZ^\t \fZ$ and $\fA^\t \fZ$ are of full rank.
	That  $\alpha \mapsto l_{\mathrm{IV}}^n(\alpha;\fY,\fZ,\fA)$ is a convex function and $\alpha \mapsto l_{\text{OLS}}^n(\alpha)$ is a strictly convex function can be seen from the quadratic forms of their second derivatives, i.e., $$
	y^\t D^2l_{\text{IV}}^n(\alpha)y = 2n^{-1} y^\t \fZ^\t \fA (\fA^\t \fA)^{-1} \fA^\t \fZ y = 
	2n^{-1}\|(\fA^\t \fA)^{-1/2} \fA^\t \fZ y\|_2^2\geq 0,$$ and $$	y^\t D^2l_{\text{OLS}}^n(\alpha) y = 2n^{-1} y^\t \fZ^\t \fZ y =2
	n^{-1}\|\fZ y\|_2^2 >0,$$ 
	for any $y\in \R^{d_1+q_1}\setminus \{0\}$.  Here, we also used that $\fA^\t \fA$ is of full rank by Assumption \Cref{ass:AtAfullRank} and that $\fZ\in \R^{n\times(d_1+q_1)}$ is an injective linear transformation as $d_1+q_1=\text{rank}(\fZ^\t \fZ)=\text{rank}(\fZ)$.

Suppose that there exists two optimal solutions $\alpha_1,\alpha_2$ to the (Primal$.t.n$) problem. By the convexity of the feasibility set any convex combination is also feasible. However, $$l_{\mathrm{OLS}}^n \lp \alpha_1/2 + \alpha_2/2\rp < l_{\mathrm{OLS}}^n(\alpha_1)/2 + l_{\mathrm{OLS}}^n(\alpha_2)/2 = l_{\mathrm{OLS}}^n(\alpha_1),$$ since $l_{\mathrm{OLS}}^n(\alpha_1)=l_{\mathrm{OLS}}^n(\alpha_2)$.	This means that $\alpha_1/2+\alpha_2/2$ has a strictly better objective value than the optimal point $\alpha_1$, which is a contradiction. 
	Hence, there cannot exist multiple solutions to the optimization problem (Primal$.t.n$).
	
Regarding the claim of solvability, note that $\fZ^\t \fZ$ is positive definite and as a consequence the smallest eigenvalue $\lambda_{\min}(\fZ^\t \fZ)$ is strictly positive. Thus, using the lower bound of the symmetric quadratic form $\alpha^\t \fZ^\t \fZ \alpha \geq\lambda_{\min}(\fZ^\t \fZ) \|\alpha\|_2^2$, we get that
	\begin{align}
	l_{\text{OLS}}^n(\alpha) &= \fY^\t \fY + \alpha^\t \fZ^\t  \fZ \alpha - 2\fY^\t \fZ \alpha \notag \geq \fY^\t \fY + \lambda_{\min}(\fZ^\t \fZ)\| \alpha\|_2^2 - 2|\fY^\t \fZ \alpha|  \notag\\
	& \geq  \fY^\t \fY + \lambda_{\min}(\fZ^\t \fZ)\| \alpha\|_2^2 - 2\|\fY^\t \fZ\|_{\text{op}} \|\alpha\|_2  \to \i, \label{eq:lOLSTendsToInfinityAsAlphaGrowsLarge}
	\end{align}
	as $\|\alpha\|_2\to\i$, where we used that for the linear operator $\fY^\t \fZ:\R^{d_1+q_1}\to \R$ the operator
	 norm is given by $\|\fY^\t \fZ\|_{\text{op}} := \inf\{c\geq 0 : |\fY \fZ v|\leq c\|v\|_2, \forall v\in \R^{d_1+q_1}\}$, obviously satisfying $|\fY^\t \fZ v|\leq  \|\fY^\t \fZ\|_{\text{op}}\|v\|_2$ for any $v\in \R^{d_1+q_1}$. 

	 Now assume that $t> \inf_\alpha l_{\text{IV}}^n(\alpha)$.
	 This implies that there exists at least one point $\tilde{\alpha}\in\R^{d_1+q_1}$ such that $l_{\text{IV}}^n (\tilde{\alpha}) \leq t$, hence 
	 we only need to consider points $\alpha$ such that $l_{\text{OLS}}^n(\alpha) \leq l_{\text{OLS}}^n(\tilde{\alpha})$ 
as possible solutions of the optimization problem.	
By the considerations 
in \Cref{eq:lOLSTendsToInfinityAsAlphaGrowsLarge}
above, there exists
	 $c\geq 0$ such that is suffices to search over the closed ball $\overline{B(0,c)}$. 
Indeed, 
	 for a sufficiently large $c\geq 0$ we know that $\alpha \not \in \overline{B(0,c)}$	 
	 implies that $l_{\text{OLS}}^n(\alpha)> l_{\text{OLS}}^n(\tilde{\alpha})$ by \Cref{eq:lOLSTendsToInfinityAsAlphaGrowsLarge}. Furthermore, as the inequality constraint function $\alpha \mapsto l_{\text{IV}}^n(\alpha)$ is continuous, the set of feasible points $(l_{\text{IV}}^n)^{-1}((-\infty,t])$ is closed. Hence, our minimization problem is equivalent with the minimization of the continuous function $\alpha\mapsto l_{\text{OLS}}^n(\alpha)$ over the convex and compact set $\overline{B(0,c)} \cap (l_{\text{IV}}^n)^{-1}((-\infty,t])$. By the extreme value theorem, the minimum exist and is attainable. We conclude that the primal problem is solvable if $t > \inf_\alpha l_{\text{IV}}^n(\alpha)$.  
	
	By definition, Slater's condition is satisfied if there exists a point in the relative interior of the problem domain where the constraint inequality is strict \citep[][]{boyd2004convex}. Since the problem domain is 
	$\R^{d_1+q_1}$, we need the existence of $\alpha\in \R^{d_1+q_1}$ such that $l^n_{\text{IV}}(\alpha)<t$. This is clearly satisfied if $t > \inf_{\alpha} l_{\text{IV}}^n(\alpha)$. Let us now specify the exact lower bound for the constraint bound as a function of the over-identifying restrictions. 	\textit{Under- and just-identified case: $q_2 \leq d_1$ ($q\leq d_1+q_1$).}  %
	Assumption \Cref{ass:AtZfullrank} yields that $\fA^\t \fZ\in \R^{q\times (d_1+q_1)}$  satisfies $\text{rank}(\fA^\t \fZ) = q$. That is, $\fA^\t \fZ $ is of full row rank, hence surjective. Thus, we are guaranteed the existence of a $\tilde{\alpha}\in \R^{d_1+q_1}$ such that $	\fA^\t \fZ \tilde{\alpha} = \fA ^\t \fY$, implying that  $l_{\mathrm{IV}}^n(\tilde{\alpha}) =0$. \textit{Over-identified case: $d_1 < q_2$ ($d_1+q_1< q$).} Note that the constraint function $l_{\mathrm{IV}}^n(\alpha):\R^{d_1+q_1} \to \R$  is strictly  convex as the  second derivative $D^2l_{\mathrm{IV}}^n(\alpha;\fY,\fZ,\fA) \propto \fZ^\t \fA(\fA^\t \fA)^{-1}\fA^\t \fZ$ is 
	positive definite by the assumption that $\fA^\t \fZ\in \R^{q\times (d_1+q_1)}$ has full (column)
	 rank. 	The global minimum of $l_{\text{IV}}$ is 
therefore	 
	 attained in the unique stationary point. Furthermore,  the stationary point is found by solving the normal equation $
	Dl_{\mathrm{IV}}^n(\alpha;\fY,\fZ,\fA) =0$. The solution to the normal equation is given by $\hat{\alpha}_{\text{TSLS}}^n = (\fZ^\t P_\fA \fZ)^\t \fZ^\t P_\fA\fY,
$
	which is the standard TSLS estimator. 
\end{proofenv}

\noindent
\noindent\begin{proofenv}{\textbf{\Cref{thm:pPULSESolvesPULSE}}}
	Let $p_{\min}\in(0,1)$ and let  \Cref{ass:ZtZfullrankandAtZfullrank} and \Cref{ass:ZYfullcolrank} hold. That is, $\fA^\t\fZ$ and $\fZ^\t \fZ$ are of full rank and $[
	\fZ \, \, \fY]$ is of full column rank. Furthermore, assume that $t_n^\star(p_{\min})>-\i$ and  $T_n(\hat{\alpha}_{\text{Pr}}^n(t_n^\star(p_{\min})))\leq Q_{\chi^2_q}(1-p_{\min})$. 
	First assume that $\hat{\alpha}_\text{Pr}^n(t_n^\star(p_{\min})) = \hat{\alpha}_{\text{OLS}}^n$. We note that
	$$
	T_n(\hat{\alpha}_{\text{OLS}}^n) = T_n(\hat{\alpha}_\text{Pr}^n(t_n^\star(p_{\min})) ) %
	\leq Q_{\chi^2_q}(1-p_{\min}),
	$$
	hence the global minimizer $\hat{\alpha}_{\text{OLS}}^n$ of $\alpha \mapsto l_{\text{OLS}}^n(\alpha)$ is unique, feasible and necessarily optimal in the PULSE problem, so $\hat{\alpha}_\text{Pr}^n(t_n^\star(p_{\min}))  = \hat{\alpha}_{\text{OLS}}^n = \hat{\alpha}^n_{\text{PULSE}}$ and we are done.

	Now assume that $\hat{\alpha}_\text{Pr}^n(t_n^\star(p)) \not = \hat{\alpha}_{\text{OLS}}^n$. Consider the PULSE problem of interest
	\begin{align} \tag{PULSE}
	\begin{array}{ll}
	\mathrm{min}_\alpha & l_{\mathrm{OLS}}^n(\alpha)  \\
	\mathrm{subject \, to} & T_n(\alpha) \leq Q_{\chi^2_{q}}(1-p_{\min}),
	\end{array} %
	\end{align}
	which is, in general, a non-convex quadratically constrained quadratic program. First we argue that the problem is solvable, i.e., the optimum is attainable.
	\begin{quote} \normalsize
To see this, let $p=p_{\min}$, $Q=Q_{\chi^2_{q}}(1-p_{\min})$ and note that by the assumption  $t_n^\star(p_{\min})>-\i$ we have that the feasible set of the PULSE problem is non-empty. By the assumptions that $[ \fZ \, \, \fY]$ is of full column rank we have that $T_n(\alpha)$ is well-defined for any $\alpha \in \R^{d_1+q_1}$, as the denominator is never zero. By continuity of $\R^{d_1+q_1} \ni \alpha \mapsto T_n(\alpha)$ we have that the feasible set $
	\cF := T_n^{-1} \lp (-\i,Q] \rp,$
	is closed and non-empty, since it is the continuous preimage of a closed set. Applying the same arguments as seen earlier in the proof of \Cref{lm:PrimalUniqueSolAndSlatersConditions}, we know that $l_{\text{OLS}}^n(\alpha)\to \i$ when $\|\alpha\| \to \i$. Hence, for a sufficiently large $c>0$ we know that if $\alpha \not \in \overline{B(0,c)}$, where $\overline{B(0,c)}\subset \R^{d_1+q_1}$ is the closed ball with centre 0 and radius $c$, then we only get suboptimal objective values $l_{\text{OLS}}^n(\alpha) > l_{\text{OLS}}^n(\hat{\alpha}_{\text{Pr}}^n (t_n^\star(p)))$. That is, we can without loss of optimality or loss of solutions restrict the feasible set to $\cF'= T_n^{-1} ( (-\i, Q] ) \cap \overline{B(0,c)}$ a closed and bounded set in $\R^{d_1+q_1}$. Hence, by the extreme value theorem the minimum over $\cF'$ is guaranteed to be attained. That is, the PULSE problem is solvable. 	
	\end{quote}	
	However, by the non-convexity of $T_n$, the preimage $T_n^{-1} ( (-\i, Q] )$ is in general not convex, so the minimum is not yet guaranteed to be attained in a unique point.	
	 We will show that the minimum of the PULSE problem is attained in a unique point, that exactly coincides with the primal PULSE solution. Fix any solution $\hat{\alpha}$ to 
the
	 PULSE problem and realize that the PULSE constraint is active in  $\hat{\alpha}$,  \begin{align}
	 T_n(\hat{\alpha})= Q. \label{Eq:PULSEsolActiveInConstraint}
	 \end{align}
\begin{quote} \normalsize
	 This is seen by noting that $\hat{\alpha}_\text{Pr}^n(t_n^\star(p)) \not = \hat{\alpha}_{\text{OLS}}^n$ by assumption, so $\hat{\alpha}_{\text{OLS}}^n$ is not feasible in the PULSE problem, that is,  $
	 \hat{\alpha}_{\text{OLS}}^n\not \in \cF$. 
	 If $\hat{\alpha}_{\text{OLS}}^n$ was feasible, then $t_n^\star(p) =  \sup \{ t \in D_{\text{Pr}} : T_n(\hat{\alpha}_{\mathrm{Pr}}^n(t))\leq Q_{\chi^2_{q}}(1-p_{\min})\} = l_{\text{IV}}^n(\hat{\alpha}^n_{\text{OLS}}),$	since $T_n(\hat{\alpha}_{\text{Pr}}^n(l_{\text{IV}}^n(\hat{\alpha}^n_{\text{OLS}})) = T_n(\hat{\alpha}_{\text{OLS}}^n)\leq Q$,  hence  $$
	 \hat{\alpha}_\text{Pr}^n(t_n^\star(p))  =   \argmin_{\alpha: l_{\text{IV}}^n(\alpha) \leq l_{\text{IV}}^n(\hat{\alpha}_{\text{OLS}}^n)} 
 l_{\text{OLS}}^n(\alpha)   =  \hat{\alpha}_{\text{OLS}}^n,
$$
which is a contradiction. That the optimum must be attained in a point, where the PULSE inequality constraint is active then	 
	  follows from \Cref{lm:SolutionIsTightIfNotStationary} of \Cref{sec:AuxLemmas}
	  and the conclusion above that the only stationary point of $l_{\text{OLS}}^n$, $\hat{\alpha}_{\text{OLS}}^n$, is not feasible.
	 \end{quote}	 
 Thus,
	\begin{align} \label{eq:PULSEsolutionTIGHTconstraint}
	T_n(\hat{\alpha}) = n \frac{l_{\text{IV}}^n(\hat{\alpha})}{l_{\text{OLS}}^n(\hat{\alpha})} = Q \iff   l_{\text{IV}}^n(\hat{\alpha}) = \frac{Q}{n} l_{\text{OLS}}^n(\hat{\alpha}).
	\end{align}
	Furthermore, the assumption that $T_n(\hat{\alpha}_{\text{Pr}}^n(t_n^\star(p)))\leq Q$ means that the solution to the primal PULSE, $\hat{\alpha}_{\text{Pr}}^n(t_n^\star(p))$, is feasible in the PULSE problem. That is, $\hat{\alpha}_\text{Pr}^n(t_n^\star(p)) \in \cF$. As a consequence of this we have that
	\begin{align} \label{eq:TheoremPULSEequalpPULSE_pPULSEfeasibleInPULSE}
	l_{\text{OLS}}^n(\hat{\alpha}_\text{Pr}^n(t_n^\star(p)))  \geq \min_{\alpha\in \cF} l_{\text{OLS}}^n(\alpha) = l_{\text{OLS}}^n(\hat{\alpha}).
	\end{align}
	Now we show that the PULSE solution $\hat{\alpha}$ is feasible in the primal PULSE problem (Primal$.t_n^*(p).n$).   
\begin{quote} \normalsize
	To see this, Note that the feasibility set of the PULSE problem can be shrunk in the following manner
	\begin{align*}
	\cF &= \lb \alpha \in \R^{d_1+q_1} :  l_{\text{IV}}^n(\alpha )\leq \frac{Q}{n}l_{\text{OLS}}^n(\alpha) \rb \\
	&=\lb \alpha \in \R^{d_1+q_1} :  l_{\text{IV}}^n(\alpha )\leq \frac{Q}{n}l_{\text{OLS}}^n(\alpha) , l_{\text{OLS}}^n(\alpha) \geq  l_{\text{OLS}}^n(\hat{\alpha})\rb \\
	&\supseteq \lb \alpha \in \R^{d_1+q_1} :  l_{\text{IV}}^n(\alpha )\leq \frac{Q}{n}l_{\text{OLS}}^n(\hat{\alpha}) , l_{\text{OLS}}^n(\alpha) \geq  l_{\text{OLS}}^n(\hat{\alpha})\rb \\ 
	&= \lb \alpha \in \R^{d_1+q_1} :  l_{\text{IV}}^n(\alpha )\leq  l_{\text{IV}}^n(\hat{\alpha} ) , l_{\text{OLS}}^n(\alpha) \geq  l_{\text{OLS}}^n(\hat{\alpha})\rb  \\
	&= \lb \alpha \in \R^{d_1+q_1} :  l_{\text{IV}}^n(\alpha )\leq  l_{\text{IV}}^n(\hat{\alpha} ) \rb  =: \hat{\cF}(\hat{\alpha}),
	\end{align*}
	where the third equality follows from \Cref{eq:PULSEsolutionTIGHTconstraint}. 	The only claim above that needs justification is that: 
	\begin{equation} \label{eq:toprove}
	 l_{\text{IV}}^n(\alpha )\leq  l_{\text{IV}}^n(\hat{\alpha} ) \implies l_{\text{OLS}}^n(\alpha) \geq  l_{\text{OLS}}^n(\hat{\alpha}).
	\end{equation}
	 For now we assume that this claim holds and provide a proof later. Thus, we have that $\hat{\cF}(\hat{\alpha}) \subseteq \cF$ and we note that $\hat{\alpha}\in \hat{\cF}(\hat{\alpha})$.
	An important consequence of this is that the PULSE solution $\hat{\alpha}$ is also the unique solution to the primal problem (Primal$.l_{\text{IV}}^{n}(\hat{\alpha}).n$). That is,
	\begin{align*}
	\hat{\alpha}  = \hat{\alpha}_{\text{Pr}}^n(l_{\text{IV}}^n(\hat{\alpha})) = \begin{array}{ll}
	\mathrm{argmin}_\alpha & l_{\mathrm{OLS}}^n(\alpha)  \\
	\mathrm{subject \, to} & l_{\text{IV}}^n(\alpha) \leq l_{\text{IV}}^n(\hat{\alpha}).
	 \end{array}
	\end{align*}
We will now prove that
$
l_\text{IV}^n(\hat{\alpha}) \in \cE:=
\{ t\in [ \min_{\alpha}l_{\mathrm{IV}}^n(\alpha),l_{\text{IV}}^n(\hat{\alpha}_{\text{OLS}}^n) ] : T_n(\hat{\alpha}_{\text{Pr}}^n(t)) \leq  Q_{\chi^2_{q}}(1-p)\}.
$
This follows from the following two observations: 
(1)
$\min_{\alpha}l_{\mathrm{IV}}^n(\alpha) \leq l_{\text{IV}}^n(\hat{\alpha}) < 
	l_{\text{IV}}^n(\hat{\alpha}_{\text{OLS}}^n)$ 
	and (2) $T_n(\hat{\alpha}_{\text{Pr}}^n(l_\text{IV}^n(\hat{\alpha}))) \leq  Q_{\chi^2_{q}}(1-p)$.
(1) follows 	
	because 
 $\hat{\alpha}_{\text{OLS}}^n\not \in \cF$, 
 which implies, by the above inclusion, that $\hat{\alpha}_{\text{OLS}}^n\not \in \hat{\cF}(\hat{\alpha})$. 
(2)	follows because $\hat \alpha$ solves
(Primal$.l_{\text{IV}}^{n}(\hat{\alpha}).n$) and thus
$\hat{\alpha}_{\text{Pr}}^n(l_\text{IV}^n(\hat{\alpha})) = \hat{\alpha}$;
$T_n(\hat{\alpha}) \leq Q_{\chi^2_q}(1-p)$
holds because $\hat{\alpha}$ is feasible for the PULSE problem.

	Now, since $t_n^\star(p)=\sup(\cE\setminus \{\min_{\alpha}l_{\mathrm{IV}}^n(\alpha)\})\in \R$ implies $t_n^\star(p)= \sup(\cE)$, it follows that	 $l^n_{\text{IV}}(\hat{\alpha}) \leq t_n^\star(p)$. In other words, any solution $\hat{\alpha}$ to the PULSE problem is feasible in the primal PULSE problem (Primal$.t_n^*(p).n$). 
\end{quote}		
	Hence,
	\begin{align} \label{eq:TheoremPULSEequalpPULSE_PULSEfeasibleInpPULSE}
	l_{\text{OLS}}^n (\hat{\alpha}) \geq l_{\text{OLS}}(\hat{\alpha}_{\text{Pr}}^n(t_n^\star(p))).
	\end{align}
	\Cref{eq:TheoremPULSEequalpPULSE_pPULSEfeasibleInPULSE} and \Cref{eq:TheoremPULSEequalpPULSE_PULSEfeasibleInpPULSE}  now yield that $l_{\text{OLS}}^n(\hat{\alpha}_{\text{Pr}}^n(t_n^\star(p))) =  l_{\text{OLS}}^n (\hat{\alpha})$ for any PULSE solution $\hat{\alpha}$. 
	Thus, any solution $\hat{\alpha}$ to the PULSE problem  is feasible in the primal PULSE problem (Primal$.t_n^*(p).n$) and it attains the optimal primal PULSE objective value. We conclude that $\hat{\alpha}$ solves the primal PULSE problem. Furthermore, it must hold that $
	\hat{\alpha} = \hat{\alpha}_{\text{Pr}}^n(t_n^\star(p))$, 
	by uniqueness of solutions to the primal PULSE problem(see \Cref{lm:PrimalUniqueSolAndSlatersConditions}). This implies two things: solutions to the PULSE problem are unique and the PULSE solution coincides with the primal PULSE solution.
	
It only remains to prove the claim of \Cref{eq:toprove}, which ensures
 $\hat{\cF}(\hat{\alpha}) \subseteq \cF$. Assume for contradiction that there exists an
$\alpha$ such that $l_{\text{IV}}^n(\alpha )\leq  l_{\text{IV}}^n(\hat{\alpha} )$ and $l_{\text{OLS}}^n(\alpha) <  l_{\text{OLS}}^n(\hat{\alpha})$, that is, we assume that
	\begin{align*}
	\cA := \underset{=:\cB}{\underbrace{\{\alpha \in \R^{d_1+q_1} : l_{\text{OLS}}^n(\alpha) < l_{\text{OLS}}^n(\hat{\alpha})\}}} \cap \underset{=:\cC}{\underbrace{\{\alpha \in \R^{d_1+q_1}: l_{\text{IV}}^n(\alpha) \leq l_{\text{IV}}^n(\hat{\alpha})\}}}   \not = \emptyset.
	\end{align*}
Define $\cM_{\text{IV}} := \{\alpha : l_{\text{IV}}^n(\alpha) = \min_{\alpha'} l_{\text{IV}}^n(\alpha')\}$ as the solution space to the generalized method of moments formulation of the instrumental variable minimization problem.
	 We now prove that  $
	\cM_{\text{IV}} \cap \cA = \emptyset$. 
\begin{quote} \normalsize
	That is, we claim that
	in the just- and over-identified setup  $\hat{\alpha}_{\text{TSLS}}^n\not \in \cA$ and in the under-identified setup none of the infinitely many solutions in the solution space of the instrumental variable minimization problem  lies in $\cA$. These statements follow by first noting that $\cM_{\text{IV}} \subset \cF$ in any identification setting. In the under- and -just identified setup this is seen by noting that $l_{\text{IV}}^n(\alpha)=0$ for any $\alpha \in \cM_{\text{IV}}$, which implies $T_n(\alpha)=0\leq  Q$, hence $\cM_{\text{IV}}\subset \cF$. 
In the over-identified setup, where $\cM_{\mathrm{IV}}=\{\hat{\alpha}_{\text{TSLS}}^n\}$, we will now  argue that $\cM_{\text{IV}} \subset \cF$
follows from the assumption that $t_n^\star(p) <\infty$. 
We first prove
that $D_{\text{Pr}}\ni t \mapsto T_n(\hat{\alpha}^n_{\text{Pr}}(t))$ is weakly increasing. If $t_1<t_2$ are two constraint bounds for which the primal problem is solvable, then $l_{\text{OLS}}^n(\hat{\alpha}^n_{\text{Pr}}(t_1)) \geq l_{\text{OLS}}^n(\hat{\alpha}^n_{\text{Pr}}(t_2))$ as the feasibility set
for $t_2$ is larger than the 
one for $t_1$.
Furthermore, the solution $\hat{\alpha}^n_{\text{Pr}}(t_2)$ either equals
$\hat{\alpha}^n_{\text{Pr}}(t_1)$ or is contained in the set $\{\alpha\in \R^{d_1+q_1}: t_1< l_{\text{IV}}^n(\alpha) \leq t_2 \}$; in the latter case we have
$l_{\text{IV}}^n (\hat{\alpha}^n_{\text
	{Pr}}(t_1)) \leq t_1 < l_{\text{IV}}^n (\hat{\alpha}^n_{\text
	{Pr}}(t_2))  \leq t_2$. Thus, we have in both cases that
	$l_{\text{IV}}^n (\hat{\alpha}^n_{\text
	{Pr}}(t_1)) \leq  l_{\text{IV}}^n (\hat{\alpha}^n_{\text
	{Pr}}(t_2))$. Combining 
	the
 two observations above we have that $$
T_n(\hat{\alpha}^n_{\text{Pr}}(t_1)) = n \frac{l_{\text{IV}}^n(\hat{\alpha}^n_{\text{Pr}}(t_1))}{l_{\text{OLS}}^n(\hat{\alpha}^n_{\text{Pr}}(t_1))} \leq n \frac{l_{\text{IV}}^n(\hat{\alpha}^n_{\text{Pr}}(t_2))}{l_{\text{OLS}}^n(\hat{\alpha}^n_{\text{Pr}}(t_2))} = T_n(\hat{\alpha}^n_{\text{Pr}}(t_2)).$$ 
Hence, as $-\infty<\min_{\alpha}l_{\text{IV}}^n(\alpha) = l_{\text{IV}}^n(\hat{\alpha}_{\text{TSLS}}^n) < t_n^\star(p) < \i$ are two points for which the primal problem is solvable we get that $$
T_n(\hat{\alpha}_{\text{TSLS}}^n) = T_n(\hat{\alpha}^n_{\text{Pr}}(l_{\text{IV}}^n(\hat{\alpha}_{\text{TSLS}}^n))) \leq T_n(\hat{\alpha}^n_{\text{Pr}}(t_n^\star(p))) \leq Q.$$ 
This proves that $\cM_{\text{IV}} \subset \cF$ in the over-identified setup.
	Now, if $\cM_{\text{IV}}\cap \cA \not = \emptyset$, there exists an $\alpha\in \cM_{\text{IV}}\cap \cA\subseteq \cF\cap \cA$ such that $\alpha$ is feasible in the PULSE problem $(\alpha\in \cF)$ and $\alpha$ is super-optimal compared to $\hat{\alpha}$, $l_{\text{OLS}}^n(\alpha) < l_{\text{IV}}^n(\hat{\alpha})$ ($\alpha\in \cA$), contradicting that $\hat{\alpha}$ is an solution to the PULSE problem. We can thus conclude that
$\cM_{\mathrm{IV}} \cap \cA = \emptyset$. 
\end{quote}
This allows us to fix two distinct points $\bar{\alpha}\not = \alpha'$ such that  $\bar{\alpha} \in \cA $ and $\alpha'\in \cM_{\text{IV}}$. Consider the proper line segment function between $\bar{\alpha}$ and  $\alpha'$, $f(t):[0,1]\to \R^{d_1+q_1}$ given by
	$
	f(t) := t\alpha' + (1-t) \bar{\alpha}.
	$
	A multivariate convex function is convex in any direction from any given starting point in its domain, so both $l_{\text{IV}}^n\circ f:[0,1]\to \R_+$ and $l_{\text{OLS}}^n\circ f:[0,1]\to \R_+$ are convex. Since $\cM_{\mathrm{IV}} \cap \cA = \emptyset$ it is obvious that the function $f$ will for sufficiently large $t$ 'leave' the set $\cA$.
We will now prove that $f$ actually leaves the superset $ \cB\supset \cA$.
More precisely, we will prove that
there exists a $t_1 \in (0,1]$ such that 
for all $t'\in[0,t_1)$ it holds that $f(t') \in \cB$
and
for all $t'\in[t_1,1]$ it holds that 
$f(t') \notin \cB$
(which implies
$f(t') \notin \cA$).
\begin{quote} \normalsize
Because $l_{\text{IV}}^n(\alpha')=\min_{\alpha}l_{\text{IV}}^n(\alpha)$ we have that $\alpha'\in \cC=\{\alpha : l_{\text{IV}}^n(\alpha) \leq l_{\text{IV}}^n(\hat{\alpha})\}$. 
	By convexity of $l_{\text{IV}}^n$ 	(Lemma~\ref{lm:PrimalUniqueSolAndSlatersConditions}) the sublevel set $\cC$ is convex and thus contains the entire line segment between $\bar{\alpha}$ and $\alpha'$. As a consequence $a'\not\in \cB$.
It therefore suffices to
construct a $t_1 \in (0,1]$ such that 
for all $t'\in[0,t_1)$ it holds that $f(t') \in \cB$
and
for all $t'\in[t_1,1]$ it holds that 
$f(t') \notin \cB$. 
We now consider the set
$
\{t \in [0,1] \,:\,
l_{\text{OLS}}^n(f(t)) < l_{\text{OLS}}^n(\hat{\alpha})
 \} = f^{-1}(\cB).
$
This set 
contains $0$ 
because $\bar{\alpha} \in \cA\subset \cB$;
it does not contain $1$ because
$\alpha'\not \in \cB$;
it
is convex, as it is a sublevel set of a convex function ($l_{\text{OLS}}^n \circ f$);
it is relatively open in $[0,1]$ because 
it is a pre-image of an open set under a continuous function ($l_{\text{OLS}}^n \circ f$).
Thus, the set must be of the form
$[0,t_1)$ for some $t_1 \in (0,1]$. This $t_1$ satisfies the desired criteria.
\end{quote}
We constructed $t_1$ above such that for all $t'\in[0,t_1)$ it holds that 
$l_{\text{OLS}}^n(f(t'))< l_{\text{OLS}}^n(\hat{\alpha})$
and for all $t'\in[t_1,1]$ it holds that 
$l_{\text{OLS}}^n(f(t'))\geq l_{\text{OLS}}^n(\hat{\alpha})$. By continuity of  $l_{\text{OLS}}^n \circ f$ we must therefore have that $l_{\text{OLS}}^n(f(t_1))= l_{\text{OLS}}^n(\hat{\alpha})$. 
	Since $f(1)= \alpha'$ is a global minimum for $l_{\text{IV}}^n$, we have that 1 must also be a global minimum for $l_{\text{IV}}^n\circ f$, implying that the convex the function $l_{\text{IV}}^n\circ f:[0,1]\to \R_+$ is monotonically decreasing. 
	It must therefore hold that $$
	l_{\text{IV}}^n(f(t_1)) <  l_{\text{IV}}^n(f(0)) =  l_{\text{IV}}^n(\bar{a}) \leq l_{\text{IV}}^n(\hat{\alpha})
.$$
The first inequality is strict because if $l_{\text{IV}}^n(f(t_1)) = l_{\text{IV}}^n(f(0)) =l_{\text{IV}}^n(\bar{\alpha})$, then convexity of $l_{\text{IV}}^n$ implies that
	$$
	l_{\text{IV}}^n(f(t_1)) = l_{\text{IV}}^n(t_1 \alpha' + (1-t_1) \bar\alpha)  \leq t_1 l_{\text{IV}}^n(\alpha') + (1-t_1)l_{\text{IV}}^n(\bar{\alpha}),$$ which happens if and only if $
	l_{\text{IV}}^n(\bar{\alpha})\leq l_{\text{IV}}^n(\alpha')
	$
contradicting the already established fact that $l_{\text{IV}}^n(\bar{\alpha})> l_{\text{IV}}^n(\alpha')$, which holds since $\alpha'\in \cM_{\text{IV}}$ but $\bar{\alpha} \not \in \cM_{\mathrm{IV}}$. We conclude that $l_{\text{IV}}^n(f(t_1))< l_{\text{IV}}^n(\hat{\alpha})$.

	Thus, we have argued that $\cM_{\mathrm{IV}} \cap \cA = \emptyset$ implies the existence of an $\tilde{\alpha} := f(t_1) = t_1 \alpha ' + (1-t_1) \bar{\alpha}$ such that $l_{\text{IV}}^n(\tilde{\alpha}) < l_{\text{IV}}^n(\hat{\alpha})$ and $l_{\text{OLS}}^n( \tilde{\alpha}) = l_{\text{OLS}}^n(\hat{\alpha})$. We have illustrated the above considerations  in \Cref{fig:pPULSEequalsPULSEProofInsersectionPlot}.
	\begin{figure}[!ht] 
		\centering		\includegraphics[width=\textwidth-150pt]{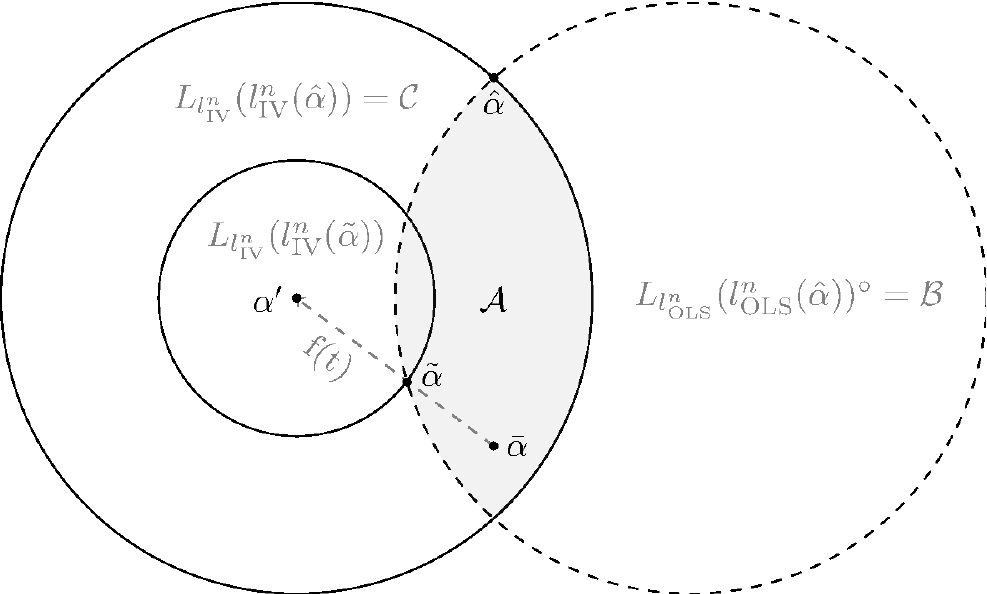}
		\caption{	
		Illustration of the described procedure in the
		just- or over-identified setup with $d_1+q_1=2$, where we show that  $\cA \not = \emptyset$ leads to a contradiction. Here, $L_g(c) := \{\alpha: g(\alpha) \leq c\}$ 
		is the $c$ sublevel set of the function $g$ and $A^\circ$ denotes the interior of a set $A$. 
		The illustration is simplified, e.g., because 
		the sublevel sets are convex but not necessarily Euclidean balls.		
		 Note that the position of $\hat{\alpha}_{\text{OLS}}^n$ is not specified, as it can possibly be in either $\cA$ or $L_{l_{\text{OLS}}^n}(l_{\text{OLS}}^n(\hat{\alpha}))\setminus \cA$. In the under-indentified setup $\alpha'$ would lie in the $d-q_2=1$ dimensional subspace $\cM_{\text{IV}}$ and the level sets would be slabs around this line. } \label{fig:pPULSEequalsPULSEProofInsersectionPlot}
	\end{figure}
It follows that $$
	T_n(\tilde{\alpha}) = n \frac{l_{\text{IV}}^n(\tilde{\alpha})}{l_{\text{OLS}}^n(\tilde{\alpha})} =  n \frac{l_{\text{IV}}^n(\tilde{\alpha})}{l_{\text{OLS}}^n(\hat{\alpha})} < n \frac{l_{\text{IV}}^n(\hat{\alpha})}{l_{\text{OLS}}^n(\hat{\alpha})}  = Q,$$ 
	implying that $\tilde{\alpha}$ is 	
	strictly feasible in the PULSE problem and, in fact, a solution  
		 as the objective value is optimal.
We argued earlier in \Cref{Eq:PULSEsolActiveInConstraint} that any solution to the PULSE problem must be tight in the inequality constraint, hence we have arrived at a contradiction.  We conclude that $\cA =\emptyset$, which implies 
that \Cref{eq:toprove} must hold.
\end{proofenv}

\noindent
\noindent\begin{proofenv}{\textbf{\Cref{lm:KclassNotEqualToTwoSLS}}}
	Assume that we are in the just- or over-identified setup and that \Cref{ass:ZtZfullrankandAtZfullrank} are satisfied. That is, $\fZ^\t\fZ$, $\fA^\t \fZ$ and $\fA^\t \fA$ are almost surely of full rank. In particular we have that 	
	$\fZ$, $\fA^\t \fZ$ and $P_\fA \fZ$ are almost surely of full column rank (injective linear maps).  Furthermore, let $\ep$ have density with respect to the Lebesgue measure and let $B$ be lower triangular.
	Fix $\lambda \geq 0$ and $\omega\in W_\lambda$, where 
\begin{align*}
		W_\lambda:=& (\hat{\alpha}_{\mathrm{K}}^n(\lambda)= \hat{\alpha}_{\text{TSLS}}^n) \cap (\text{rank}(\fZ^\t \fZ)=d_1+q_1) \\&\quad \cap  (\text{rank}(\fA^\t \fZ)=d_1+q_1) \cap (\text{rank}(\fA^\t \fA)=q),
\end{align*}
	satisfying
	$P(\hat{\alpha}_{\mathrm{K}}^n(\lambda)= \hat{\alpha}_{\text{TSLS}}^n) = P(W_\lambda)$. 	By \Cref{eq:KclassLossFunctionAsPenalizedOLS} we have that 
\begin{align*}
		\hat{\alpha}_{\mathrm{K}}^n(\lambda) &= \argmin_{\alpha} \{l_{\mathrm{OLS}}^n(\alpha) + \lambda l_{\mathrm{IV}}^n(\alpha)\}  
	\\
	&=\argmin_{\alpha} \{(\fY-\fZ\alpha)^\t (\fY-\fZ\alpha) + \lambda (\fY-\fZ\alpha)^\t P_\fA (\fY-\fZ\alpha)\} \\
	&=\argmin_{\alpha}  (\fY-\fZ\alpha)^\t(\fI+  \lambda P_\fA) (\fY-\fZ\alpha) \\
	&= \argmin_{\alpha} \| (\fI + \lambda P_\fA)^{1/2} (\fY - \fZ \alpha ) \|_2^2 \\
	&= \argmin_{\alpha} \| \fY - \fZ \alpha \|_{(\fI+\lambda P_\fA)}^2 ,
\end{align*}
	where $\| \cdot \|_{(\fI+\lambda P_\fA)}$ is the norm induced by the inner product $
	\la x , y \ra_{(\fI+\lambda P_\fA)} = x^\t (\fI+\lambda P_\fA) y$.
	The solution $\fZ \hat{\alpha}_{\mathrm{K}}^n(\lambda)$ is well-known to coincide with the orthogonal projection of $\fY$ onto $\cR(\fZ)$, the range of $\fZ$, with respect to the inner product $\la \cdot , \cdot \ra_{(\fI+\lambda P_\fA)}$. Hence, $\fZ \hat{\alpha}_{\mathrm{K}}^n(\lambda)$ is the unique element in this closed linear subspace such that for all $z\in \cR(\fZ)$ it holds that
	$$
	\la \fY - \fZ \hat{\alpha}_{\mathrm{K}}^n(\lambda),  z \ra_{(\fI+\lambda P_\fA)} = \la \fY - \fZ \hat{\alpha}_{\mathrm{K}}^n(\lambda), (\fI+\lambda P_\fA) z \ra = 0  ,$$ or equivalently,  
	\begin{align} \label{eq:LemmaKclassDifferentFromIVSolutionsKclassPredictionUniqueEquality}
	\la \fY - \fZ \hat{\alpha}_{\mathrm{K}}^n(\lambda),  z \ra = - \lambda \la \fY - \fZ \hat{\alpha}_{\mathrm{K}}^n(\lambda),  P_\fA z \ra, \quad \forall z\in \cR(\fZ).
	\end{align}
	We note that if $\lambda = 0$ then $\hat{\alpha}_{\mathrm{K}}^n(\lambda) = \hat{\alpha}_{\text{OSL}}^n$, seen either by directly inspecting the closed form solution of $\hat{\alpha}_{\mathrm{K}}^n(\lambda)$ or concluding the same from \Cref{eq:LemmaKclassDifferentFromIVSolutionsKclassPredictionUniqueEquality}. 
	
	Furthermore, when $\lambda > 0$ 
	we have that  $\hat{\alpha}_{\mathrm{K}}^n(\lambda)=\hat{\alpha}_{\text{TSLS}}^n$ implies that, again, 	$\hat{\alpha}_{\mathrm{K}}^n(\lambda) = \hat{\alpha}_{\text{OSL}}^n$. 
		To see this, we note that $$
		\hat{\alpha}_{\text{TSLS}}^n =\argmin_\alpha l_{\text{IV}}^n(\alpha) =  \argmin_\alpha \| P_\fA \fY - P_\fA \fZ \alpha \|_2^2,$$
		so $P_\fA \fZ\hat{\alpha}_{\text{TSLS}}^n$ is the orthogonal projection of $P_\fA \fY$ onto $\cR(P_\fA\fZ)$. That is, $P_\fA \fZ\hat{\alpha}_{\text{TSLS}}^n$ is the unique element in $\cR(P_\fA \fZ )$ such that $\la P_\fA \fY - P_\fA \fZ \hat{\alpha}_{\text{TSLS}}^n , s \ra = 0$ for all $s\in \cR(P_\fA \fZ )$, i.e.,  
		$$\la P_\fA \fY - P_\fA \fZ \hat{\alpha}_{\text{TSLS}}^n, P_\fA z \ra  = 0,$$ for all  $z\in \cR(\fZ)$. Thus, if $\hat{\alpha}_{\mathrm{K}}^n(\lambda)= \hat{\alpha}_{\text{TSLS}}^n$ for some $\lambda >0$ we have that
\begin{align*}
			0 &= 	\la P_\fA \fY - P_\fA \fZ \hat{\alpha}_{\text{TSLS}}^n , P_\fA z \ra  \\
			&= 	\la P_\fA \fY - P_\fA \fZ \hat{\alpha}_{\mathrm{K}}^n(\lambda) , P_\fA z \ra 
		\\
		&= \la \fY -  \fZ \hat{\alpha}_{\mathrm{K}}^n(\lambda), P_\fA z \ra \\
		&= -\lambda^{-1}\la \fY - \fZ \hat{\alpha}_{\mathrm{K}}^n(\lambda),  z \ra,  
\end{align*}
		hence $\la \fY - \fZ \hat{\alpha}_{\mathrm{K}}^n(\lambda),  z \ra = 0 $ for all $z\in \cR(\fZ)$, where we used \Cref{eq:LemmaKclassDifferentFromIVSolutionsKclassPredictionUniqueEquality} and in the third equality we used	
		that $P_\fA$ is idempotent and symmetric. This implies that $\hat{\alpha}_{\mathrm{K}}^n(\lambda) = \hat{\alpha}_{\text{OLS}}^n$, as it satisfies the uniquely determining condition for the ordinary least square estimator. 

	Hence, for any $\lambda \geq 0$, whenever $\hat{\alpha}_{\mathrm{K}}^n(\lambda)= \hat{\alpha}_{\text{TSLS}}^n$ we know that $
	\hat{\alpha}_{\text{TSLS}}^n =\hat{\alpha}_{\mathrm{K}}^n(\lambda) = \hat{\alpha}_{\text{OLS}}^n$. 
	Thus, for any $\lambda \geq 0$ it holds that $$
	P(\hat{\alpha}_{\mathrm{K}}^n(\lambda)= \hat{\alpha}_{\text{TSLS}}^n) \leq P(\hat{\alpha}_{\text{TSLS}}^n = \hat{\alpha}_{\text{OLS}}^n).$$
	Recall that the reduced form equations of our system are given by
$	[\fY  \, \, \fX \, \,  \fH ]
= \fA \Pi + \bm{\ep}\Gamma^{-1}$ where $\Gamma := I-B$. When $B$ is lower triangular, so is $\Gamma$ and $\Gamma^{-1}$. %
	By selecting the relevant columns of $\Pi$ and $\Gamma^{-1}$ we may express the marginal reduced form structural equations of $\fS$  that consist of any collection of columns from $[
	\fY \, \,  \fX \, \, \fH
	]$ by $\fS = \fA \Pi_{S} + \bm{\ep} \Gamma^{-1}_S$ for conformable matrices $\Pi_{S}$ and $\Gamma^{-1}_S$. In particular, we have that the marginal reduced form structural equations for $\fY$ and $\fX_{*}$ are given by $
	\fY = \fA \Pi_Y + \bm{\ep}\Gamma^{-1}_{Y}$ and $
	\fX_{*} = \fA \Pi_{X_*} + \bm{\ep}\Gamma^{-1}_{X_*}$, 
	where $\Pi_Y, \Pi_{X_*},\Gamma^{-1}_{Y}$ and $\Gamma^{-1}_{X_{*}}$ are matrices conformable with the following block representation
	\begin{align*}
	\Pi &= [ \, 
	\underbrace{\Pi_{Y_{}}}_{q\times 1} \,\,  \underbrace{\Pi_{X_*}}_{q\times d_1} \, \, \underbrace{\Pi_{X_{-*}}}_{d\times q_2} \, \, \underbrace{\Pi_{H_{}}}_{q\times r} \,
	] \in \R^{q \times l}, 
	\quad \text{and} \quad	
	\Gamma^{-1} = 
	[ \,\underbrace{\Gamma^{-1}_{Y_{}}}_{l\times 1}  \,\,
	\underbrace{\Gamma^{-1}_{X_*}}_{l\times d_1} \,\,
	\underbrace{\Gamma^{-1}_{X_{-*}}}_{l \times d_2} \,\,
	\underbrace{\Gamma^{-1}_{H_{}}}_{l \times r}  \,]  \in \R^{l\times l},
	\end{align*}
where $l:=1+d+r$. 	Note that by the lower triangular structure of $\Gamma^{-1}$ we have that the only matrix among $ \Gamma^{-1}_{Y}$, $\Gamma^{-1}_{X_*}$, $\Gamma^{-1}_{X_{-*}}$ and $\Gamma^{-1}_{H}$ that has a non-zero first row is $\Gamma^{-1}_Y$.
	
	Now assume without loss of generality that the first row of $\Gamma^{-1}$ is given by the first canonical Euclidean basis vector $(1,0,...,0)\in \R^{1\times l}$ such that we have the following partitionings
	\begin{align*}
	\bm{\ep} &= [ \,
	\underbrace{\bm{\ep}_Y}_{n\times 1}  \, \,  \underbrace{\bm{\ep}_{-Y}}_{n\times(d+r)}
	\, ]\in \R^{n\times l}, &
	\Gamma_{Y}^{-\t} &= [ \, 1  \, \, 
	\underbrace{\Gamma_{-Y,Y}^{-\t}}_{1\times (d+r)}
	\,] \in \R^{1 \times l},\\
	\Gamma_{X_{*}}^{-\t} &= [ \,
	\bm{0}_{d_1\times 1}  \, \, 
	\underbrace{\Gamma_{-Y,X_{*}}^{-\t}}_{d_1\times(d+r)}
	\, ] \in \R^{d_1 \times l}, &
	\Gamma_{X_{1}}^{-\t} &= [ \,
	0  \, \,
	\underbrace{\Gamma_{-Y,X_{1}}^{-\t}}_{1\times (d+r)}
	\,] \in \R^{1 \times l},
	\end{align*}
	where $\fX_1$ is the first column of $\fX$.
	Hence, we note that $\bm{\ep}\Gamma^{-1}_{Y} =\bm{\ep}_Y + \bm{\ep}_{-Y}\Gamma_{-Y,Y}^{-1}$, such that the marginal reduced form structural equation for $\fY$ has the following representation $$
	\fY = \fA \Pi_Y + \bm{\ep}\Gamma^{-1}_{Y} 
	= \fA \Pi_Y + \bm{\ep}_{-Y}\Gamma_{-Y,Y}^{-1}+ \bm{\ep}_Y  =: f_y(\fA,\bm{\ep}_{-Y}) + \bm{\ep}_{Y}.$$
	We can also represent $\fZ$ in terms of these structural coefficient block matrices by
	\begin{align*}
	\fZ &=  \begin{bmatrix}
	\fX_{*} & \fA_{*} 
	\end{bmatrix} =  \begin{bmatrix}
	\fA \Pi_{X_{*}} + \bm{\ep}\Gamma^{-1}_{X_{*}} & \fA_{*}
	\end{bmatrix}  = \begin{bmatrix}
	\fA \Pi_{X_{*}}  & \fA_{*}
	\end{bmatrix} + \bm{\ep} \begin{bmatrix}
	\Gamma^{-1}_{X_{*}} & \bm{0}_{l \times q_1}
	\end{bmatrix}  \\
	&= \fA \begin{bmatrix}
	\Pi_{X_{*}}  & \begin{bmatrix}
	\fI_{q_1\times q_1} \\ \bm{0}_{q_2\times q_1}
	\end{bmatrix}
	\end{bmatrix} +  \begin{bmatrix}
	\bm{\ep}_{-Y} \Gamma_{-Y,X_{*}}^{-1} & \bm{0}_{l \times q_1}
	\end{bmatrix} =:f_z(\fA,\bm{\ep}_{-Y}).
	\end{align*}
	Assumption \Cref{ass:AindepEp} and Assumption \Cref{ass:epIndependentMarginals} together  with the assumption that the data matrices consist of row-wise i.i.d.\ copies of the system variables, yield that $\fA \independent \bm{\ep}_{Y}$ and $\bm{\ep}_{-Y}\independent \bm{\ep}_{Y} $. This implies that the conditional distribution of $\bm{\ep}_{Y}$ given $\fA$ and $\bm{\ep}_{-Y}$ satisfies $P_{\bm{\ep}_{Y} |\fA = A,\bm{\ep}_{-Y}=e} = P_{\bm{\ep}_{Y}}$ for $P_{\fA,\bm{\ep}_{-Y}}$-almost all $(A,e)\in \R^{n\times q}\times \R^{n\times (d+r)}$. Hence, conditional on $\fA = A$ and $\bm{\ep}_{-Y} =e$ we have that
	$
	\fY|(\fA = A,\bm{\ep}_{-Y} =e) \stackrel{a.s.}{=} f_y(A,e) + \bm{\ep}_Y,$ 
	and 
	$ \fZ |(\fA = A,\bm{\ep}_{-Y} =e) \stackrel{a.s.}{=} f_z(A,e).$
	Now let $(P_\fA\fZ)^+ = (\fZ^\t P_\fA \fZ)^{-1}\fZ^\t P_\fA$ and $
	\fZ^+ =(\fZ^\t \fZ)^{-1}\fZ^\t$ denote the pseudo-inverse matrices of the almost surely full column rank matrices $P_\fA \fZ$ and $\fZ$. Furthermore, note that the pseudo-inverses are unique for all matrices, 	
	i.e., if $P_\fA \fZ \not = \fZ$, then $(P_\fA\fZ)^+ \not = \fZ^+$.   We realize that $
	\hat{\alpha}^n_{\text{TSLS}} =(\fZ^\t P_\fA \fZ)^{-1}\fZ^\t P_\fA \fY = (P_\fA \fZ)^+\fY $ and $ \hat{\alpha}^n_{\text{OLS}} = (\fZ^\t\fZ)^{-1} \fZ^\t \fY =\fZ^+\fY$. 
	Thus, with slight abuse of notation we let $Z:=f_z(A,e)$ for any $A,e$, and note that
	\begin{align} \notag
	&P(\hat{\alpha}^n_{\text{TSLS}}= \hat{\alpha}^n_{\text{OLS}})\\ \notag
	&= P((P_\fA\fZ)^+ \fY =\fZ^+\fY) \\\notag
	&= \int P\lp [(P_\fA \fZ)^+- \fZ^+] \fY =0 | \fA = A,\bm{\ep}_{-Y} =e \rp \, \mathrm{d} P_{\fA,\bm{\ep}_{-Y}} (A,e) \\\notag
	&= \int P\lp [(P_A Z)^+- Z^+](f_y(A,e) + \bm{\ep}_Y)  =0 \rp \, \mathrm{d} P_{\fA,\bm{\ep}_{-Y}} (A,e) \\ \label{Eq:KclassNotEqTSLSTempEq1}
	&= \int \mathbbm{1}_{(P_A Z \not = Z)}P\lp [(P_A Z)^+- Z^+](f_y(A,e) + \bm{\ep}_Y)  =0 \rp \, \mathrm{d} P_{\fA,\bm{\ep}_{-Y}} (A,e),
	\end{align}
	where $P_A = A(A^\t A)^{-1}A^\t\in \R^{n\times n}$. The last equality is due to the claim that $\mathbbm{1}_{(P_A Z \not = Z)}=1$ for $P_{\fA,\bm{\ep}_{-Y}}$ almost all $(A,e)$, or equivalently  $$
	\int \mathbbm{1}_{(P_A Z  = Z)} \, \mathrm{d} P_{\fA,\bm{\ep}_{-Y}} (A,e) 
	= \int \mathbbm{1}_{(P_\fA \fZ  = \fZ)} \, \mathrm{d} P 
	=  P(P_\fA \fZ =\fZ)=0.$$
	We prove this claim now.
	\begin{quote} \normalsize
		We now prove that $P(P_\fA \fZ =\fZ)=0$. First we note that $P_\fA \fZ =\fZ$ implies that $\cR(\fZ) \subseteq \cR(\fA)$.  
		Since $ \fZ = [ \fX_* \, \, \fA_* ]$ with $\fA = [\fA_*\,\,  \fA_{-*} ]$ it must hold that $\cR(\fX_*) \subseteq \cR(\fA)$. 
		Assume without loss of generality that $\fX_1$, the first column of $\fX$, is also a column of $\fX_*$. 
		Note that $\cR(\fX_*)\subseteq \cR(\fA)$ implies that $\fX_1$ can be written as a linear combination of the columns in $\fA$, i.e., there exists a $b=(b_1,...,b_q)\in \R^{q}$ such that $
		\fX_1 = b_1 \fA_1 + \cdots + b_{q} \fA_q = \fA b$, 
		namely $b=(\fA^\t \fA)^{-1}\fA \fX_1$. The marginal reduced form structural equation for $\fX_1$ is given by $		\fX_1 = \fA \Pi_{X_1} + \bm{\ep}\Gamma^{-1}_{X_1} = \fA \Pi_{X_1} + \bm{\ep}_{-Y} \Gamma_{-Y,X_{1}}^{-1}=\fA \Pi_{X_1} + \tilde{\bm{\ep}}$, 
		where $\tilde{\bm{\ep}}:= \bm{\ep}_{-Y}\Gamma_{-Y,X_{1}}^{-1}$. These two equalities are only possible if $\tilde{\bm{\ep}}\in \cR(\fA)$. 
		Note that $\tilde{\bm{\ep}}$ has jointly independent marginals (i.i.d.\ observations). Each coordinate is an independent copy of a linear combination of $1+d+r$ independent random variables $\ep_1,...,\ep_{1+d+r}$ all with density with respect to Lebesgue measure. 
		We conclude that $\tilde{\bm{\ep}}$ has density with respect to the $n$-dimensional Lebesgue measure as the linear 
		combination is non-vanishing. This holds because
		$\Gamma_{-Y,X_{1}}^{-1}\not=0$ by virtue of being a column of the invertible matrix $
		\Gamma^{-1}$, where we have removed the first entry (which was 
		a zero element). Furthermore, since $\fA \independent \bm{\ep}$, we also have that $\fA 
		\independent \tilde{\bm{\ep}}$. Hence, the conditional distribution of $\tilde{\bm{\ep}}$ given $\fA$ satisfies 
		$P_{\tilde{\bm{\ep}} |\fA = A} = P_{\tilde{\bm{\ep}}}$ for $P_{\fA}$-almost all $A\in \R^{n\times q}$. We conclude that 
		\begin{align*}
			P(P_\fA \fZ =\fZ)  &\leq P(\tilde{\bm{\ep}} \in \cR(\fA)) \\
			&= \int P(\tilde{\bm{\ep}} \in \cR(\fA)|\fA=A) \, \mathrm{d} P_{\fA}(A) 
		\\
		&= \int P(\tilde{\bm{\ep}} \in \cR(A)) \, \mathrm{d} P_{\fA}(A) \\
		&=0.
		\end{align*}
		The last equality follows from the fact that $q= \text{rank}(\fA^\t\fA) = \text{rank}(\fA)< n$ implies that $\cR(\fA)$ is a $q$-dimensional subspace of $\R^n$. Hence, for $P_\fA$-almost all $A\in \R^{n\times q}$ it
		holds that $\cR(A)$ is a $q$-dimensional subspace of $\R^n$. The probability that $\tilde{\bm{\ep}}$ lies in a $q$-dimensional subspace of $\R^n$ is zero, since it has density with respect to the $n$-dimensional Lebesgue measure.
	\end{quote}		 
	Thus, it suffices to show that $$ \label{Eq:EpYInaffineSubspaceWithProbZero}
	P\lp [(P_A Z)^+- Z^+](f_y(A,e) + \bm{\ep}_Y)  =0 \rp =0,$$
	for any $A\in \R^{n\times q}$
	and $Z=f_z(A,e)\in \R^{n\times (d_1+q_1)}$ with $P_AZ \not = Z$. 
		Therefore, let $A\in \R^{n\times q}$ 
		and $Z=f_z(A,e)\in \R^{n\times (d_1+q_1)}$ with $P_AZ \not = Z$. It holds that $(P_AZ)^+\not = Z^+$, which implies that $(P_AZ)^+-Z^+\not = 0$. Furthermore, we have that $$
		[(P_A Z)^+- Z^+](f_y(A,e) + \bm{\ep}_Y)  =0 ,$$ if and only if $$\bm{\ep}_\fY \in \text{ker}((P_A Z)^+- Z^+)-[(P_A Z)^+- Z^+]f_y(A,e),$$
		so it suffices to show that $\bm{\ep}_Y$ has zero probability to be in the affine (translated) subspace $$\text{ker}((P_A Z)^+- Z^+)-[(P_A Z)^+- Z^+]f_y(A,e)\subset \R^n.$$ This affine subspace has dimension $n$ if and only if $(P_AZ)^+-Z^+ = 0$, which we know is false. Hence, the dimension of the affine subspace is strictly less than $n$. As $\bm{\ep}_Y$ has density with respect to the $n$-dimensional Lebesgue measure, we know that the probability of being in a $N<n$ dimensional affine subspace is zero. 

	Hence, we have shown that $
	P(\hat{\alpha}^n_{\text{TSLS}}= \hat{\alpha}^n_{\text{OLS}}) = 0$. 
	Combining all of our observations we get that $
	P(\hat{\alpha}_{\mathrm{K}}^n(\lambda)= \hat{\alpha}_{\text{TSLS}}^n) \leq P(\hat{\alpha}_{\text{TSLS}}^n = \hat{\alpha}_{\text{OLS}}^n) = 0.$
	We conclude that $$
	P \lp   \hat{\alpha}_{\text{TSLS}}^n \not = \hat{\alpha}_{\mathrm{K}}^n(\lambda) \rp = 1,\quad \text{for all }\lambda \geq 0.$$ However, we can easily strengthen this to  
	$P ( \cap_{\lambda \geq 0}   (\hat{\alpha}_{\text{TSLS}}^n \not = \hat{\alpha}_{\mathrm{K}}^n(\lambda) ) ) = 1.
	$
	To this end, let $\omega$ be a realization in the almost sure set $\cap_{\lambda \in \bQ_+}	W_\lambda.
	$
	Then, 
	$\omega \in \bigcap_{\lambda \geq 0 } \lp \hat{\alpha}_{\text{TSLS}}^n \not = \hat{\alpha}_{\mathrm{K}}^n(\lambda) \rp$.
	Otherwise,
	there exists an $\tilde{\lambda} \in \R_+ \setminus \bQ_+$ such that $\hat{\alpha}_{\text{TSLS}}^n = \hat{\alpha}_{\mathrm{K}}^n(\tilde{\lambda})$. By \Cref{lm:OLSandIV_Monotonicity_FnctOfPenaltyParameterLambda} we have that $\lambda \mapsto l_{\text{IV}}^n( \hat{\alpha}_{\text{IV}}^n (\lambda))$ is monotonically decreasing, but since  $\hat{\alpha}_{\mathrm{K}}^n(\tilde{\lambda})$ already minimizes the $l_{\text{IV}}^n$ function, so will all $\hat{\alpha}_{\mathrm{K}}^n(\lambda )$ for all $\lambda \geq \tilde{\lambda}$. As $\hat{\alpha}_{\text{TSLS}}^n$ is the unique point that minimizes $l_{\text{IV}}^n$ we conclude that $\hat{\alpha}_{\text{TSLS}}^n = \hat{\alpha}_{\mathrm{K}}^n(\lambda )$ for all $\lambda \geq \tilde{\lambda}$, which yields a contradiction.
	We conclude that $
	P \lp \cap_{\lambda \geq 0 } \lp \hat{\alpha}_{\text{TSLS}}^n \not = \hat{\alpha}_{\mathrm{K}}^n(\lambda) \rp \rp =1$.

\end{proofenv}

\noindent
\noindent\begin{proofenv}{\textbf{\Cref{lm:EquivalenceBetweenKlikeAndPrimal}}}
	Let \Cref{ass:ZtZfullrankandAtZfullrank} and \Cref{ass:ZYfullcolrank} hold, i.e., that $\fZ^\t \fZ$ and $\fA^\t \fZ$ are  of full rank and  $[ \fZ \,\, \fY]$ is of full column rank. Furthermore, let \Cref{ass:KclassNotInIV} hold, i.e., that $\hat{\alpha}_{\mathrm{K}}^n(\lambda)\not \in \cM_{\mathrm{IV}}$ for all $\lambda\geq 0$. It holds that (Primal$.t.n$) has a unique solution and satisfies Slater's condition for all  $t>  \min_{\alpha}l_{\mathrm{IV}}^n(\alpha)$ (\Cref{lm:PrimalUniqueSolAndSlatersConditions}). Furthermore,  (Dual$.\lambda.n$) has a unique solution for all $\lambda \geq 0$ (\Cref{lm:PenalizedKClassSolutionUniqueAndExists}).

	First consider an arbitrary $t\in D_{\text{Pr}}$ and note that the dual problem of (Primal$.t.n$), 	not to be confused with the problem (Dual$.\lambda.n$), is given by
	\begin{align} \label{Eq:DualOfPrimal}
	\begin{array}{ll}
	\text{maximize}_\lambda & g_t(\lambda)  \\
	\text{subject to} & \lambda \geq 0.
	\end{array}
	\end{align}
	However, (Dual$.\lambda.n$) is equivalent with the infimum problem in the definition of  $g_t:\R_+ \to \R$ given by $$
	g_t(\lambda) := \inf_\alpha \{l_{\text{OLS}}^n(\alpha)+ \lambda (l_{\text{IV}}^n(\alpha) - t)\}.$$ 
	Now consider $\hat{\alpha}_{\text{Pr}}^n(t)$ solving the primal (Primal$.t.n$). 
	Slater's condition is satisfied, so there exists a $\lambda(t)\geq 0$ solving the dual problem and strong duality holds, $l_{\text{OLS}}^n(\hat{\alpha}_{\text{Pr}}^n(t)) = g_t(\lambda(t))$. 
	We will now show that $\hat{\alpha}_{\text{Pr}}^n(t)$ also solves to the K-class penalized regression problem (Dual.$\lambda(t).n$). That is, we will show that $
	\hat{\alpha}_{\text{Pr}}^n(t) = \begin{array}{ll}
	\underset{\alpha}{\text{argmin}} & l_{\text{OLS}}^n(\alpha) + \lambda(t) l_{\text{IV}}^n(\alpha).
	\end{array}
	$
	To that end, note that
	\begin{align*}
	g_t(\lambda(t)) &=  \inf_\alpha \{ {l_{\text{OLS}}^n(\alpha)+ \lambda(t) (l_{\text{IV}}^n(\alpha) - t)} \} = \inf_\alpha \{ {l_{\text{OLS}}^n(\alpha)+ \lambda(t)l_{\text{IV}}^n(\alpha) } \}- \lambda(t) t\\
	&\leq l_{\text{OLS}}^n(\hat{\alpha}_{\text{Pr}}^n(t))+ \lambda(t) (l_{\text{IV}}^n(\hat{\alpha}_{\text{Pr}}^n(t)) - t) 
	 =  l_{\text{OLS}}^n(\hat{\alpha}_{\text{Pr}}^n(t))  =g_t(\lambda(t)),
	\end{align*}
	where in the last equality
we used strong duality 
	and  the second last equality we used that for any constraint bound $t\in D_{\text{Pr}}$ the inequality constraint of (Primal$.t.n$) is active in the solution $\hat{\alpha}_{\text{Pr}}^n(t)$, i.e., $l_{\text{IV}}^n(\hat{\alpha}_{\text{Pr}}^n(t))=t$; see \Cref{lm:SolutionIsTightIfNotStationary} of \Cref{sec:AuxLemmas}.
	Thus, it holds that 
	\begin{align*}
	& \inf_\alpha \{ {l_{\text{OLS}}^n(\alpha)+ \lambda(t) (l_{\text{IV}}^n(\alpha) - t)} \} = l_{\text{OLS}}^n(\hat{\alpha}_{\text{Pr}}^n(t))+ \lambda(t) (l_{\text{IV}}^n(\hat{\alpha}_{\text{Pr}}^n(t)) - t) \\
	\iff & 
	\inf_\alpha \{ l_{\text{OLS}}^n(\alpha)+ \lambda(t) l_{\text{IV}}^n(\alpha) \} = l_{\text{OLS}}^n(\hat{\alpha}_{\text{Pr}}^n(t))+ \lambda(t) l_{\text{IV}}^n(\hat{\alpha}_{\text{Pr}}^n(t))
	,\end{align*}
	proving that $\hat{\alpha}_{\text{Pr}}^n(t)$ coincides with the unique solution $\hat{\alpha}_{\text{K}}^n(\lambda(t))$ to the K-class problem (Dual.$\lambda(t).n$) as it attains the same objective. Furthermore, there can only be one $\lambda(t)$ solving the dual problem in \Cref{Eq:DualOfPrimal}.
	 If there are two distinct solutions $\lambda',\lambda''\geq 0$ with $\lambda'\not = \lambda''$, then by the above observations we get that $
	\hat{\alpha}_{\text{Pr}}^n(t) = \hat{\alpha}_{\text{K}}^n(\lambda') = \hat{\alpha}_{\text{K}}^n(\lambda''),$
	in contradiction to \Cref{cor:KclassSolutionsDistinct}.

	Conversely, fix $\lambda \geq 0$ and recall that $\hat{\alpha}_{\mathrm{K}}^n(\lambda)$ solves the
	penalized K-class regression problem (Dual.$\lambda.n$), that is, $
	\hat{\alpha}_{\mathrm{K}}^n(\lambda) = \argmin_{\alpha} \{ l_{\text{OLS}}^n(\alpha) + \lambda l_{\text{IV}}^n(\alpha) \}$.
	Now consider a primal constraint bound $t(\lambda) :=  l_{\text{IV}}^n(\hat{\alpha}_{\text{K}}^n(\lambda)) $ and consider the corresponding primal optimization problem (Primal.$t(\lambda).n$) and its dual form given by
	\begin{align} \label{Eq:PrimalKclassEquivalence_PrimalDualSpecificToTheorem}
	\textit{Primal}:\begin{array}{lr}
	\begin{array}{ll}
	\text{minimize} & l_{\text{OLS}}^n(\alpha)  \\
	\text{subject to} &  l_{\text{IV}}^n(\alpha) \leq  t(\lambda)
	\end{array}&  \quad \quad \textit{Dual}:	\begin{array}{ll}
	\text{maximize} & g_{t(\lambda)}(\gamma)  \\
	\text{subject to} & \gamma \geq 0,
	\end{array}
	\end{array}
	\end{align}
	where  $g_{t(\lambda)}:[0,\i) \to \R$ is given by
	$
	g_{t(\lambda)}(\gamma) = \inf_\alpha \{ l_{\text{OLS}}^n(\alpha) + \gamma [l_{\text{IV}}^n(\alpha)-t(\lambda)] \} .
	$
	Here we note that the proposed primal problem satisfies Slater's condition. To see this note that $\hat{\alpha}_{\text{K}}^n(\lambda)\not \in \cM_{\text{IV}}$, by \Cref{ass:KclassNotInIV}, 
	hence $\inf_{\alpha}l_{\text{IV}}^n(\alpha)=\min_{\alpha}l_{\text{IV}}^n(\alpha) < t(\lambda)= l_{\text{IV}}^n( \hat{\alpha}_{\text{K}}^n(\lambda))$. 
	Furthermore, we conclude that $t(\lambda) \in (\min_{\alpha}l_{\text{IV}}^n(\alpha),l_{\text{IV}}(\hat{\alpha}_{\text{OLS}}^n)]=D_{\text{Pr}}$ as $\lambda \mapsto l_{\text{IV}}(\hat{\alpha}_{\text{K}}^n(\lambda))$ 
	is monotonically decreasing and $\hat{\alpha}_{\text{K}}^n(0) = \hat{\alpha}_{\text{OLS}}^n$; see \Cref{lm:OLSandIV_Monotonicity_FnctOfPenaltyParameterLambda}.

	Let $p^\star$ and $d^\star$ denote the optimal objective values for the above primal and dual problem in \Cref{Eq:PrimalKclassEquivalence_PrimalDualSpecificToTheorem}, respectively. It 
	holds that $\hat{\alpha}_{\text{K}}^n(\lambda)$  is primal feasible since it satisfies the inequality constraint of the primal problem in	 \Cref{Eq:PrimalKclassEquivalence_PrimalDualSpecificToTheorem}. This implies that $p^\star \leq l_{\text{OLS}}^n(\hat{\alpha}_{\text{K}}^n(\lambda))$ 
	since $p^\star$ is the infimum of all attainable objective values. By the non-negative duality gap we also have that
	\begin{align*}
	p^\star & \geq d^\star = \sup_{\gamma\geq 0} g_{t(\lambda)}(\gamma)  \geq  g_{t(\lambda)}(\lambda) =\inf_\alpha \{ l_{\text{OLS}}^n(\alpha) + \lambda [l_{\text{IV}}^n(\alpha)-t(\lambda)] \} \\
	&=\inf_\alpha \{ l_{\text{OLS}}^n(\alpha) + \lambda l_{\text{IV}}^n(\alpha) \} - \lambda t(\lambda)=   l_{\text{OLS}}^n(\hat{\alpha}_{\text{K}}^n(\lambda)) + \lambda [l_{\text{IV}}^n(\hat{\alpha}_{\text{K}}^n(\lambda))-l_{\text{IV}}^n(\hat{\alpha}_{\text{K}}^n(\lambda))] \\
	&= l_{\text{OLS}}^n(\hat{\alpha}_{\text{K}}^n(\lambda)),
	\end{align*}
	implying that $l_{\text{OLS}}^n(\hat{\alpha}_{\text{K}}^n(\lambda))= p^\star$. This proves that strong duality holds and that the solution 
	to the K-class regression problem $\alpha_{\text{K}}^n(\lambda)$ solves the primal optimization problem (Primal.$t(\lambda).n$), since it attains the unique optimal objective value while also satisfying the inequality constraint. 
\end{proofenv}

\noindent
\noindent\begin{proofenv}{\textbf{\Cref{thm:PULSEpPULSEdPULSEEequivalent}}}
	Fix any $p_{\min}\in(0,1)$ and let \Cref{ass:ZtZfullrankandAtZfullrank,ass:ZYfullcolrank} hold, 
	i.e., that $\fZ^\t \fZ$ and $\fA^\t \fZ$ are  of full rank and  $[ \fZ \,\, \fY]$ is of  full column rank. Furthermore, let \Cref{ass:KclassNotInIV} hold, i.e., that $\hat{\alpha}_{\mathrm{K}}^n(\lambda)\not \in \cM_{\mathrm{IV}}$ for all $\lambda \geq 0$. It holds that (Primal$.t.n$) has a unique solution and satisfies Slater's condition for all  $t>  \min_{\alpha}l_{\mathrm{IV}}^n(\alpha)$ (\Cref{lm:PrimalUniqueSolAndSlatersConditions}), that (Dual$.\lambda.n$) has a unique solution for all $\lambda \geq 0$ (\Cref{lm:PenalizedKClassSolutionUniqueAndExists}) and that $\{\hat{\alpha}_{\text{Pr}}^n(t):t\in D_{\text{Pr}}\}=\{\hat{\alpha}_{\text{K}}^n(\lambda):\lambda \geq 0\}$ (\Cref{lm:EquivalenceBetweenKlikeAndPrimal}). 
	Finally, we assume that $\lambda_n^\star(p_{\min})<\i$. 
To simplify notation, we write $Q=Q_{\chi^2_{q}}(1-p_{\min})$.	
	
		We claim that the PULSE estimator can be represented in the dual form of the primal PULSE problem. That is, as a K-class estimator $\hat{\alpha}_{\text{PULSE}}^n(p_{\min}) = \hat{\alpha}^n_{\text{K}}(\lambda_n^\star(p_{\min}))$ with stochastic penalty parameter given by $
	\lambda_n^\star(p_{\min}) := \inf\{\lambda \geq 0 : T_n(\hat{\alpha}_{\text{K}}^n (\lambda))\leq Q_{\chi^2_{q}}(1-p_{\min}) \}.
$
We show this by proving that $\hat{\alpha}_{\text{K}}^n(\lambda_n^\star(p_{\min}))= \hat{\alpha}_{\text{Pr}}^n(t_n^\star(p_{\min})))$, which by \Cref{thm:pPULSESolvesPULSE} implies that the claim is true, if the conditions $t_n^\star(p_{\min})>-\i$ and 
$T_n(\hat{\alpha}_{\text{Pr}}^n(t_n^\star(p_{\min})))\leq Q$ can be verified from the assumption that $\lambda_n^\star(p_{\min})<\i$. First, we note that if $\lambda_n^\star(p_{\min})<\i$, then $t_n^\star(p_{\min})>-\i$. 
	\begin{quote} \normalsize
			This follows by noting that, with $t(\lambda) := l_{\text{IV}}^n(\hat{\alpha}_{\text{K}}^n(\lambda))$,  proof of \Cref{lm:EquivalenceBetweenKlikeAndPrimal} \textit{ii)} yields that $\hat{\alpha}_{\text{K}}^n(\lambda) = \hat{\alpha}_{\text{Pr}}^n(t(\lambda))$ for any $\lambda \geq 0$ 
			which yields $
		\lambda_n^\star(p_{\min}) = \inf \left\{  \lambda  \geq 0:   T_n ( \hat{\alpha}_{\text{Pr}}^n \circ t(\lambda) )\leq Q \right\}.
$
		Hence, if $\lambda_n^\star(p_{\min}) <\i$ we know there exists a $\lambda' \geq 0$ such that $T_n ( \hat{\alpha}_{\text{Pr}}^n \circ t(\lambda') )\leq Q$, i.e., there exists a $t'=t(\lambda')\in(\min_{\alpha'} l_{\text{IV}}^n(\alpha'),\i)$ such that $T_n ( \hat{\alpha}_{\text{Pr}}^n (t') )\leq Q$. 
		We have excluded that $t'=\min_{\alpha'} l_{\text{IV}}^n(\alpha')$ as $t'=l_{\text{IV}}^n(\hat{\alpha}_{\text{K}}^n(\lambda')) > \min_{\alpha'} l_{\text{IV}}^n(\alpha')$ since $\hat{\alpha}_{\text{K}}(\lambda') \not \in \cM_{\text{IV}}$. 
		Furthermore, we can without loss of generality assume that $t' \in (\min_{\alpha'} l_{\text{IV}}^n(\alpha'),l_{\text{IV}}^n(\hat{\alpha}^n_{\text{OLS}})]$ because if $t' > l_{\text{IV}}^n(\hat{\alpha}^n_{\text{OLS}})$, then it holds that $T_n ( \hat{\alpha}_{\text{Pr}}^n (l_{\text{IV}}^n(\hat{\alpha}^n_{\text{OLS}})) )\leq Q$ as $\hat{\alpha}_{\text{Pr}}^n (l_{\text{IV}}^n(\hat{\alpha}^n_{\text{OLS}})) = \hat{\alpha}_{\text{Pr}}^n (t')$ since the ordinary least square solution solves 
		all (Primal$.t.n$) with constraints bounds larger than
		 $l_{\text{IV}}^n(\hat{\alpha}^n_{\text{OLS}})$. As a consequence, the set for which we take the supremum over in the definition of $t_n^\star(p_{\min})$ is non-empty, such that $t_n^\star(p_{\min}) >-\i$.
	\end{quote}
	Next we show that $\hat{\alpha}_{\text{K}}^n(\lambda_n^\star(p_{\min}))= \hat{\alpha}_{\text{Pr}}^n(t_n^\star(p_{\min})))$. 
	When this equality is shown, then the remaining condition that 
	$T_n(\hat{\alpha}_{\text{Pr}}^n(t_n^\star(p_{\min})))\leq Q$ 
	follows by \Cref{lm:TestInAlphaLambdaStarEqualsQuantileApp} 
	and we are done. For any constraint bound $t \in D_{\text{Pr}}= ( \min_{\alpha}l_{\mathrm{IV}}^n(\alpha),l_{\text{IV}}^n(\hat{\alpha}_{\text{OLS}}^n) ]$, consider the primal and corresponding dual optimization problems
	\begin{align} \label{eq:dPULSEequalpPULSEPrimalDualProblems}
	\textit{Primal}:\begin{array}{lr}
	\begin{array}{ll}
	\text{minimize} & l_{\mathrm{OLS}}^n(\alpha)  \\
	\text{subject to} & l_{\mathrm{IV}}^n(\alpha)\leq t
	\end{array} &  \quad \quad \textit{Dual}:	\begin{array}{ll}
	\text{maximize} & g_t(\lambda)  \\
	\text{subject to} & \lambda \geq 0,
	\end{array}
	\end{array}
	\end{align}
	with dual function $g_t:\R_{\geq 0} \to \R$ given by $
	g_t(\lambda) := \inf_\alpha\{l_{\mathrm{OLS}}^n(\alpha)+ \lambda (l_{\mathrm{IV}}^n(\alpha) - t)\}.$
	The proof of \Cref{lm:EquivalenceBetweenKlikeAndPrimal} yields that there exists a unique $\tilde{\lambda}(t)\geq 0$ solving the dual problem of \Cref{eq:dPULSEequalpPULSEPrimalDualProblems} such that   $ \label{eq:TheoremPrimalDualConnectionPvalConstraintEqThatPrimalEqualKclass}
	\hat{\alpha}_{\text{Pr}}^n(t) = \hat{\alpha}_{\text{K}}^n(\tilde{\lambda}(t)).
	$
We now prove that $D_{\text{Pr}}\ni t \mapsto \tilde{\lambda}(t)$ is strictly decreasing.
\begin{quote} \normalsize
	Note that by the definition of $g_t$ and \Cref{lm:PenalizedKClassSolutionUniqueAndExists} (or  equivalently the discussion in the beginning of \Cref{sec:DualPULSE}) we have that
	\begin{align} 
	g_t(\lambda) &= \inf_\alpha\{l_{\mathrm{OLS}}^n(\alpha)+ \lambda l_{\mathrm{IV}}^n(\alpha) \} -\lambda t  \label{Eq:RepresentationOfGtInTermsOfKclass}
	= l_{\mathrm{OLS}}^n(\hat{\alpha}_{\text{K}}^n (\lambda))+ \lambda l_{\mathrm{IV}}^n(\hat{\alpha}_{\text{K}}^n (\lambda)) -\lambda t.
	\end{align}
	For any $t_1$, $t_2$ with
	$0\leq  \min_{\alpha}l_{\mathrm{IV}}^n(\alpha)<t_1<t_2 \leq l_{\text{IV}}^n(\hat{\alpha}_{\text{OLS}}^n) $ we have that $g_{t_1}(\tilde{\lambda}(t_1)) \geq g_{t_1}(\tilde{\lambda}(t_2))$ and $g_{t_2}(\tilde{\lambda}(t_2)) \geq g_{t_2}(\tilde{\lambda}(t_1))$ as $\tilde{\lambda}(t)$ maximizes $g_t$.
Hence, by bounding the first term we get that
	\begin{align} \label{eq:TheoremLambdaStarFiniteTStarFinite_Ineq1}
	g_{t_1}(\tilde{\lambda}(t_1)) - g_{t_2}(\tilde{\lambda}(t_2)) & \geq g_{t_1}(\tilde{\lambda}(t_2)) - g_{t_2}(\tilde{\lambda}(t_2)) 
	= \tilde{\lambda}(t_2)(t_2-t_1), 
	\end{align}
where the last equality follows from the representation in \Cref{Eq:RepresentationOfGtInTermsOfKclass}.
 Similarly, by bounding the other term we get that
	\begin{align} \label{eq:TheoremLambdaStarFiniteTStarFinite_Ineq2}
	g_{t_1}(\tilde{\lambda}(t_1)) - g_{t_2}(\tilde{\lambda}(t_2)) & \leq  g_{t_1}(\tilde{\lambda}(t_1)) - g_{t_2}(\tilde{\lambda}(t_1)) 
	= \tilde{\lambda}(t_1)(t_2-t_1). 
	\end{align}
	Combining the inequalities from \Cref{eq:TheoremLambdaStarFiniteTStarFinite_Ineq1,eq:TheoremLambdaStarFiniteTStarFinite_Ineq2} we conclude that $
	\tilde{\lambda}(t_2) (t_2-t_1) \leq \tilde{\lambda}(t_1) (t_2-t_1)$ which implies $\tilde{\lambda}(t_2) \leq \tilde{\lambda}(t_1),
$
	proving that $D_{\text{Pr}} \ni t \mapsto \tilde{\lambda}(t)$, the dual solution as a function of the primal problem constraint bound, is weakly decreasing. We now  strengthen this statement to strictly decreasing. 
	For any constraint bound $t\in D_{\text{Pr}}= ( \min_{\alpha}l_{\mathrm{IV}}^n(\alpha),l_{\text{IV}}^n(\hat{\alpha}_{\text{OLS}}^n)]$ we have that the solution $\hat{\alpha}^n_{\text{Pr}}(t)$ yields an active inequality constraint in the (Primal.$t.n$) problem, i.e., $l_{\text{IV}}^n(\hat{\alpha}_{\text{Pr}}(t))=t$; see \Cref{lm:SolutionIsTightIfNotStationary} of \Cref{sec:AuxLemmas}.
	Therefore, for any $\min_{\alpha}l_{\mathrm{IV}}^n(\alpha) < t_1 < t_2 \leq  l_{\text{IV}}^n(\hat{\alpha}_{\text{OLS}}^n)$ we get that
	$
	l_{\text{IV}}^n(\hat{\alpha}_{\text{K}}^n(\tilde{\lambda}(t_1))) = l_{\text{IV}}^n(\hat{\alpha}_{\text{Pr}}^n(t_1)) = t_1 < t_2 = l_{\text{IV}}^n(\hat{\alpha}_{\text{Pr}}^n(t_2)) = l_{\text{IV}}^n(\hat{\alpha}_{\text{K}}^n(\tilde{\lambda}(t_2))),
	$
	proving that $\tilde{\lambda}(t_1) \not = \tilde{\lambda}(t_2) $, which implies that $D_{\text{Pr}} \ni t\mapsto \tilde{\lambda}(t)$ is strictly increasing.
\end{quote}
Recall, by \Cref{lm:EquivalenceBetweenKlikeAndPrimal} that the K-class estimators for $\kappa\in[0,1)$ coincides with the collection of solutions to every primal problem satisfying Slater's condition. 
That is,
	\begin{align} \label{Eq:EqualityOfPrimalSoltuionsAndKclassSolutions}
	\{\hat{\alpha}_{\text{K}}^n(\lambda):\lambda \geq 0\} 	
	= \{\hat{\alpha}_{\text{Pr}}^n(t): t \in D_{\text{Pr}} \}	
	= \{\hat{\alpha}_{\text{K}}^n(\tilde{\lambda}(t)):t \in D_{\text{Pr}}\},
	\end{align}
	where $\tilde{\lambda}$ is as introduced above.

	It now only remains to show	
	that $\tilde{\lambda}( t_n^\star(p_{\min}))= \lambda_n^\star(p_{\min})$, which implies the 
	wanted 
	conclusion as $\hat{\alpha}_{\text{Pr}}^n(t_n^\star(p_{\min})) = \hat{\alpha}_{\text{K}}^n(\tilde{\lambda}(t_n^\star(p_{\min}))) = \hat{\alpha}_{\text{K}}^n(\lambda_n^\star(p_{\min}))$. 
	We know that $\hat{\alpha}_{\text{K}}^n \circ \tilde{\lambda} (t) = \hat{\alpha}_{\text{Pr}}^n (t)$, hence for all $t\in D_{\text{Pr}}$, $
	(T_n \circ \hat{\alpha}_{\text{K}}^n \circ \tilde{\lambda}) (t) = (T_n \circ \hat{\alpha}_{\text{Pr}}^n) (t)$,
	and that for any $A\subset [0,\infty)$ it holds that  $\tilde{\lambda} ( \tilde{\lambda}^{-1}(A)) = A \cap \cR (\tilde{\lambda})$, 
	where $\cR (\tilde{\lambda}) = \{\tilde{\lambda}(t): t\in D_{\text{Pr}}\}\subseteq [0,\i)$ is the range of the reparametrization function $\tilde{\lambda}:D_{\text{Pr}} \to [0,\i)$. In fact, 
	$\tilde{\lambda}$ is surjective. To see this, note that $[0,\i) \ni \lambda \mapsto \hat{\alpha}_{\text{K}}^n(\lambda)$ is injective by \Cref{cor:KclassSolutionsDistinct}. Thus, $\cR(\tilde{\lambda})=[0,\i)$ must hold, for otherwise 	\Cref{Eq:EqualityOfPrimalSoltuionsAndKclassSolutions} would not hold. Hence, by surjectivity of $\tilde{\lambda}$ we get that for $A\subset [0,\infty)$ it holds that  $\tilde{\lambda} ( \tilde{\lambda}^{-1}(A)) = A$.

	Now consider $\hat{\alpha}_{\text{Pr}}^n: D_{\text{Pr}} \to \R^{d_1+q_1}$, $\hat{\alpha}_{\text{K}}^n: [0,\infty) \to \R^{d_1+q_1}$ and $\tilde{\lambda}:D_{\text{Pr}} \to [0,\infty)$  as measurable (which follows by continuity and monotonicity) mappings  such that $$
	t_n^\star(p_{\min}) = \sup \{ (T_n \circ \hat{\alpha}_{\text{Pr}}^n)^{-1} (-\i,  Q] \} 
	= \sup \{  (T_n \circ \hat{\alpha}_{\text{K}}^n \circ \tilde{\lambda})^{-1} (-\i,  Q] \}.$$ 
	Since $t\mapsto \tilde{\lambda}(t)$ is strictly decreasing, we get that
\begin{align*}
		\tilde{\lambda}(t_n^\star(p_{\min})) &= \tilde{\lambda} (   \sup \{   (T_n \circ \hat{\alpha}_{\text{K}}^n \circ \tilde{\lambda})^{-1} (-\i,  Q] \} ) \\
	&= \inf \left\{ \tilde{\lambda}\lp  (T_n \circ \hat{\alpha}_{\text{K}}^n \circ \tilde{\lambda})^{-1} (-\i,  Q]  \rp \right\} \\
	&= \inf \{ \tilde{\lambda} ( \tilde{\lambda}^{-1} (  (T_n \circ \hat{\alpha}_{\text{K}}^n )^{-1} (-\i,  Q] )) \} \\
	&= \inf \left\{   (T_n \circ \hat{\alpha}_{\text{K}}^n )^{-1} (-\i,  Q]  \right\} \\
&=  \inf \left\{  \lambda \geq 0  :   T_n ( \hat{\alpha}_{\text{K}}^n (\lambda) )\leq Q \right\} 
	\\&= \lambda_n^\star(p_{\min}),
\end{align*}


\end{proofenv}

\noindent
\noindent\begin{proofenv}{\textbf{\Cref{lm:LamdaStarFiniteIFF}}}
	Let $p_{\min}\in(0,1)$ and let \Cref{ass:ZtZfullrankandAtZfullrank,ass:ZYfullcolrank,ass:KclassNotInIV} hold. We have that $$
l_{\mathrm{OLS}}^n(\hat{\alpha}_{\mathrm{K}}^n (\lambda)) \geq l_{\mathrm{OLS}}^n(\hat{\alpha}_{\text{OLS}}^n) 
= n^{-1}\|\fY-\fZ\hat{\alpha}_{\text{OLS}}^n\|_2^2 
= n^{-1}\| \fY - P_\fZ \fY \|_2^2 
>0,$$
as $P_\fZ \fY \not = \fY$ (by \Cref{ass:ZYfullcolrank} we have that $\fY \not \in \text{span}(\fZ)$, such that the projection of $\fY$ onto the column space of $\fZ$ does not coincide with $\fY$ itself). Hence,
 $T_n:\R^{d_1+q_1}\to \R$ is well-defined, and the following upper bound
$$
T_n (\hat{\alpha}_{\text{K}}^n (\lambda) ) = n \frac{l_{\text{IV}}^n(\hat{\alpha}_{\text{K}}^n (\lambda))}{l_{\mathrm{OLS}}^n(\hat{\alpha}_{\text{K}}^n (\lambda))}  \leq n \frac{l_{\text{IV}}^n(\hat{\alpha}_{\text{K}}^n (\lambda)) }{l_{\mathrm{OLS}}^n(\hat{\alpha}_{\text{OLS}}^n)}
,$$
is valid for every $\lambda \geq 0$. In the under- and just-identified setup we know that there exists an $\tilde{\alpha}\in \cM_{\text{IV}} \subset \R^{d_1+q_1}$ such that $
0=l_{\mathrm{IV}}^n(\tilde{\alpha})$. 
Now let $\Lambda>0$ be given by
\begin{align} \label{eq:LambdaEquality}
\Lambda 
:= 
n \frac{  l_{\mathrm{OLS}}^n(\tilde{\alpha}) }{l_{\mathrm{OLS}}^n(\hat{\alpha}_{\text{OLS}}^n) Q_{\chi^2_{q}}(1-p_{\min})}.
\end{align}
For any $\lambda > \Lambda$ we have by the non-negativity of $l_{\text{OLS}}^n(\alpha)/\lambda$ that
\begin{align*}
l_{\text{IV}}^n(\hat{\alpha}_{\text{K}}^n (\lambda))  &\leq  \lambda^{-1}l_{\mathrm{OLS}}^n(\hat{\alpha}_{\text{K}}^n (\lambda)) + l_{\text{IV}}^n(\hat{\alpha}_{\text{K}}^n (\lambda)) = \min_\alpha \{ \lambda^{-1}l_{\mathrm{OLS}}^n(\alpha) + l_{\text{IV}}^n(\alpha)\} \\
&\leq  \lambda^{-1}l_{\mathrm{OLS}}^n(\tilde{\alpha}) + l_{\text{IV}}^n(\tilde{\alpha})  <  \frac{l_{\text{OLS}}^n(\tilde{\alpha})}{\Lambda} = \frac{l_{\mathrm{OLS}}^n(\hat{\alpha}_{\text{OLS}}^n)Q_{\chi^2_{q}}(1-p_{\min})}{n},
\end{align*}
This implies $$
T_n (\hat{\alpha}_{\mathrm{K}}^n (\lambda) )  \leq n \frac{l_{\text{IV}}^n(\hat{\alpha}_{\text{K}}^n (\lambda))}{l_{\mathrm{OLS}}^n(\hat{\alpha}_{\text{OLS}}^n)}  < Q_{\chi^2_{q}}(1-p_{\min}),
$$
whenever 
$\lambda > \Lambda$, proving that $
\lambda_n^\star(p_{\min}) = \inf \{\lambda \geq 0 : T_n(\hat{\alpha}_{\text{K}}^n(\lambda)) \leq Q_{\chi^2_{q}}(1-p_{\min})\}<\i.
$
Now consider the over-identified setup $(q> d_1+q_1)$. We claim that $
\lambda^\star_n(p_{\text{min}}) < \i$ if and only if $T_n(\hat{\alpha}_{\text{TSLS}}^n)< Q_{\chi^2_q}(1-p_{\min})$. 
If $T_n(\hat{\alpha}_{\text{TSLS}}^n)< Q_{\chi^2_q}(1-p_{\min})$, then by continuity of $\lambda \mapsto \hat{\alpha}_{\text{K}}^n(\lambda)$ and $\alpha \mapsto T_n(\alpha)$ it must hold that $\lambda^\star_n(p_{\text{min}}) < \i$. This follows by noting that $$ 
T_n(\hat{\alpha}_{\text{K}}^n(\lambda)) \downarrow T_n(\lim_{\lambda \to \i }\hat{\alpha}_{\text{K}}^n(\lambda))= T_n(\hat{\alpha}_{\text{TSLS}}^n) < Q_{\chi^2_q}(1-p_{\min}),$$
when $\lambda \to \i$, as $\lambda \mapsto T_n(\hat{\alpha}_{\text{K}}^n(\lambda))$ 
is strictly decreasing (\Cref{lm:OLSandIV_Monotonicity_FnctOfPenaltyParameterLambda})
Here, we also used that 
\begin{align*}
	\lim_{\lambda \to \i } \hat{\alpha}_\text{K}^n(\lambda) &= \lim_{\lambda \to \i } (\fZ^\t (\fI+\lambda P_\fA)\fZ)^{-1} \fZ^\t(\fI+\lambda P_\fA)\fY  \\
&=  \lim_{\lambda \to \i } (\fZ^\t (\lambda^{-1}\fI+P_\fA)\fZ)^{-1} \fZ^\t(\lambda^{-1}\fI+ P_\fA)\fY \\ \notag
&= (\fZ^\t P_\fA \fZ)^{-1} \fZ^\t  P_\fA\fY  \\
&= \hat{\alpha}_{\text{TSLS}}^n.
\end{align*}
Hence, there must exist a $\lambda \in [0,\i)$ such that $T_n(\hat{\alpha}_{\text{K}}^n(\lambda))  <Q_{\chi^2_q}(1-p_{\min})$, proving that $\lambda^\star_n(p_{\text{min}}) < \i$.  Furthermore, note that the above arguments also imply that $\label{Eq:TSLSyieldsSmallestTestAmongKclass}
T_n(\hat{\alpha}_{\text{K}}^n(\lambda)) > T_n(\hat{\alpha}_{\text{TSLS}}^n)
$,
for any $\lambda \geq 0$, as $\lambda \mapsto T_n(\hat{\alpha}_{\text{K}}^n(\lambda))$ is strictly decreasing and $T_n(\hat{\alpha}_{\text{TSLS}}^n)$ is the limit as $\lambda \to\i$.

 Conversely, assume that $\lambda^\star_n(p_{\text{min}}) < \i$, which implies that there exists a $\lambda'\in[0,\i)$ such that $T_n(\hat{\alpha}_{\text{K}}^n(\lambda'))\leq Q_{\chi^2_q}(1-p_{\min} )$. Thus, 
 $$
 T_n(\hat{\alpha}_{\text{TSLS}}^n) < T_n(\hat{\alpha}_{\text{K}}^n(\lambda'))\leq Q_{\chi^2_q}(1-p_{\min}),
 $$
 proving that the converse implication also holds. 
 
 We furthermore note that, if the acceptance region is empty, that is $$
 \cA_n(1-p_{\min}) := \{\alpha\in \R^{d_1+q_1}:T_n(\alpha) \leq Q_{\chi^2_q}(1-p_{\min})\} = \emptyset,$$ 
 then it obviously holds that $\lambda^\star_n(p_{\min})=\{\lambda \geq 0 :T_n(\hat{\alpha}_{\text{K}}^n(\lambda) )\leq Q_{\chi^2_q}(1-p_{\min}) \} = \i$.
 The possibility of the acceptance region being empty, follows from the fact that the Anderson-Rubin confidence region can be empty; see \Cref{rm:ConnectionToAndersonRubinCI}. To realize that the Anderson-Rubin confidence region can be empty we refer to the discussions and Monte-Carlo simulations of \citet{davidson2014confidence}.

\end{proofenv}

\noindent
\noindent\begin{proofenv}{\textbf{\Cref{lm:BinarySearchLambdaStarConverges}}}
	Assume that $\lambda_n^\star(p_{\min})<\i$ and that Assumption  \Cref{ass:ZtZfullrank} and \Cref{ass:ZYfullcolrank} hold. Consider \Cref{Binary.Search.Lambda.Star} for any fixed $N\in \N$. 
	The first `while loop' guarantees that $\lambda_{\min}$ and $\lambda_{\max}$ are such that
	$\lambda^{\star}\in(\lambda_{\min},\lambda_{\max}]$. This is seen by noting that $\lambda \mapsto T_n(\hat{\alpha}_\text{K}^n(\lambda))$ is monotonically decreasing
	(\Cref{lm:OLSandIV_Monotonicity_FnctOfPenaltyParameterLambda}) and that $\lambda_n^\star(p_{\min})<\i$. Hence, $T_n(\hat{\alpha}_\text{K}^n(\lambda_{\max}))$
	eventually drops below $Q_{\chi^2_{q}}(1-p)$. 
	 The second `while loop' keeps iterating until the 
	interval $(\lambda_{\min},\lambda_{\max}]$,  which  is guaranteed to contain $\lambda_n^\star(p_{\min})$, has a length less than or equal to $1/N$. Let $\lambda_{\min}$ and $\lambda_{\max}$ denote the last boundaries achieved 
	before the procedure terminates. Then  $
	0\leq \mathrm{Binary.Search}(N,p) - \lambda_n^\star(p_{\min}) = 
	\lambda_{\max} - \lambda_n^\star(p_{\min})  \leq 
		\lambda_{\max}- \lambda_{\min} 	\leq 1/N$. 
	Hence,  $
	\mathrm{Binary.Search}(N,p) - \lambda_n^\star(p_{\min}) \to 0$, as $N\to\i$.
\end{proofenv}

\noindent
\noindent\begin{proofenv}{\textbf{\Cref{thm:ConsistencyOfPULSE}}}
	Consider the just- or over-identified setup $(q\geq d_1+q_1)$, let  \Cref{ass:AIndepUYandMeanZeroA} hold. We furthermore assume that the population rank condition, \Cref{ass:EAZtfullrank}, i.e., $E(AZ^\t)$ is of full rank, are satisfied. We furthermore work under the finite-sample conditions of  \Cref{ass:ZtZfullrankandAtZfullrank,ass:ZYfullcolrank}, i.e., that $\fA^\t \fA$ and  $\fZ^\t\fA$ are of full rank and $\fY\not\in \text{span}(\fZ)$ for all sample-sizes $n\in \N$ almost surely. 
	The  first two 
	of these are not strictly necessary as the population version of these rank assumptions guarantee that $\fA^\t \fA$ and  $\fZ^\t\fA$ are of full rank with probability tending to one; see 
	proof of
\Cref{lm:PopulationPenalizedKClassSolutionUniqueAndExists}.
	Likewise, we can drop the last finite-sample assumption as it is almost surely guaranteed if we assume that the distribution of $\ep_Y$ has density with respect to Lebesgue measure.
	 The proof below is easily modified	 
	 to accommodate these more relaxed assumptions, but for notational simplicity we 
	 prove the statement
	 under the stronger finite-sample assumptions. We also let \Cref{ass:KclassNotInIV} hold which in addition with the previous assumptions guarantees that the dual representation of the PULSE holds whenever $\lambda^\star_n(p_{\min})<\i$; see \Cref{thm:PULSEpPULSEdPULSEEequivalent}.
	 Furthermore, many of the previous theorems and lemmas were shown for a specific realization that satisfies the finite sample assumptions. Hence, we may only invoke the conclusions of these theorems almost surely.	 Note that the assumptions guarantee that the TSLS estimator is consistent, i.e., $\hat \alpha_{\mathrm{TSLS}}^n \convp \alpha_0$.

	Fix any $p_{\min}\in(0,1)$ and let an arbitrary $\ep>0$ be given. We want to prove  that  $
	P(\|\hat{\alpha}_{\text{PULSE}+}^n(p_{\min})-\alpha_0\|>\ep)  \to 0.$
	To that end, define the events $(A_n)_{n\in \N}$ by $
	A_n := (T_n(\hat{\alpha}_{\text{TSLS}}^n)< Q_{\chi^2_{q}}(1-p_{\min})),
	$
	such that
	\begin{align} \label{eq:ConsistencyFirstTerm}
	P(\|\hat{\alpha}_{\text{PULSE}+}^n(p_{\min})-\alpha_0\|>\ep) =& P((\|\hat{\alpha}_{\text{PULSE}}^n(p_{\min})-\alpha_0\|>\ep)\cap A_n) \\
	&+P((\|\hat{\alpha}_{\text{ALT}}^n-\alpha_0\|>\ep)\cap A_n^c), \label{eq:ConsistencySecondTerm}
	\end{align}
	for all $n\in\N$. The last term,  \Cref{eq:ConsistencySecondTerm}, tends to zero as $n\to\i$, $$
	P((\|\hat{\alpha}_{\text{ALT}}^n-\alpha_0\|>\ep)\cap A_n^c) \leq P(\|\hat{\alpha}_{\text{ALT}}^n-\alpha_0\|>\ep) \to 0,
$$
	by the assumption that $\hat{\alpha}_{\text{ALT}}^n\convp \alpha_0$ as $n\to\i$. In regards to the first term, the right-hand side of \Cref{eq:ConsistencyFirstTerm}, we note that $A_n = (\lambda^\star_n(p_{\min})<\i)$, by \Cref{lm:LamdaStarFiniteIFF}. Formally, this event equality only holds when intersecting both sides with the almost sure event that the finite sample rank condition holds. However, we suppress this intersection for ease of notation. Thus, on $A_n$, it holds that $\hat{\alpha}_{\text{PULSE}}^n(p_{\min}) = \hat{\alpha}_{\text{K}}^n(\lambda^\star_n(p_{\min})),$ by \Cref{thm:PULSEpPULSEdPULSEEequivalent}, implying that $$
	P((\|\hat{\alpha}_{\text{PULSE}}^n(p_{\min})-\alpha_0\|>\ep)\cap A_n) = P((\|\hat{\alpha}_{\text{K}}^n(\lambda^\star_n(p_{\min}))-\alpha_0\|>\ep)\cap A_n).$$
	Furthermore, \Cref{lm:TestInAlphaLambdaStarEqualsQuantileApp} yields that on $A_n$, it holds that $$
	T_n(\hat{\alpha}_{\text{K}}^n(\lambda^\star_n(p_{\min}))) \leq Q_{\chi^2_{q}}(1-p_{\min}),$$ or equivalently $$l_{\text{IV}}^n(\hat{\alpha}_{\text{K}}^n(\lambda^\star_n(p_{\min})) \leq n^{-1}Q_{\chi^2_{q}}(1-p_{\min}) l_{\text{OLS}}^n(\hat{\alpha}_{\text{K}}^n(\lambda^\star_n(p_{\min})).$$ 
	On $A_n$, the stochastic factor in the upper bound above, is further bounded from above by 
\begin{align*}
		l_{\mathrm{OLS}}^n(\hat{\alpha}_{\mathrm{K}}^n (\lambda_n^\star(p_{\min}))) &\leq \sup_{\lambda\geq 0 }l_{\mathrm{OLS}}^n(\hat{\alpha}_{\mathrm{K}}^n (\lambda ))  \\
		&= \lim_{\lambda \to \i }l_{\mathrm{OLS}}^n(\hat{\alpha}_{\mathrm{K}}^n (\lambda )) \\
		&= l_{\mathrm{OLS}}^n(\lim_{\lambda \to \i }\hat{\alpha}_{\mathrm{K}}^n (\lambda )) \\
	&= l_{\mathrm{OLS}}^n(\hat{\alpha}_{\text{TSLS}}^n), 
\end{align*}
	where we used continuity of $\alpha \mapsto l_{\mathrm{OLS}}^n(\alpha)$, that $\lambda \mapsto l_{\mathrm{OLS}}^n(\hat{\alpha}_{\mathrm{K}}^n (\lambda ))$ is weakly increasing (\Cref{lm:OLSandIV_Monotonicity_FnctOfPenaltyParameterLambda}) and that	$\lim_{\lambda \to \i } \hat{\alpha}_\text{K}^n(\lambda) = \hat{\alpha}_{\text{TSLS}}^n$.
	Recall that the TSLS estimator is consistent 	$\hat{\alpha}_{\text{TSLS}}^n \convp \alpha_0$, where $\alpha_0$ is the causal coefficient of $Z$ onto $Y$. Hence, Slutsky's theorem and the weak law of large numbers yield that 
\begin{align*}
		l_{\mathrm{OLS}}^n(\hat{\alpha}_{\text{TSLS}}^n)  &= n^{-1} (\fY - \fZ \hat{\alpha}_{\text{TSLS}}^n)^\t (\fY - \fZ \hat{\alpha}_{\text{TSLS}}^n) \\&= 
	n^{-1} \fY^\t \fY + (\hat{\alpha}_{\text{TSLS}}^n)^\t n^{-1} \fZ^\t \fZ\hat{\alpha}_{\text{TSLS}}^n - 2n^{-1}\fY^\t \fZ\hat{\alpha}_{\text{TSLS}}^n  \\&\convp E(Y^2) + \alpha_0^\t E(Z Z^\t ) \alpha_0 - 2 E(YZ^\t) \alpha_0 \\&= E[(Y-Z\alpha_0)^2]. 
\end{align*}
	Thus, on the event $A_n$, we have that $$
	0\leq l_{\text{IV}}^n(\hat{\alpha}_{\text{K}}^n(\lambda^\star_n(p_{\min})) \leq n^{-1}Q_{\chi^2_{q}}(1-p_{\min}) l_{\text{OLS}}^n(\hat{\alpha}_{\text{TSLS}}^n)=:H_n,$$
	where the upper bound $H_n$ converges to zero in probability by Slutsky's theorem. Furthermore, note that
\begin{align*}
		  \notag
    l_\text{IV}^n(\alpha_0) &= \|n^{-1/2} (\fA^\t \fA)^{-1/2} \fA^\t(\fY - \fZ\alpha_0)\|_2^2 \\ \notag
    &=\| (n^{-1}\fA^\t \fA)^{-1/2} n^{-1}\fA^\t\fU_Y\|_2^2 \\
    &\convp \| E(AA^\t)^{-1/2} E(AU_Y)\|_2^2 \\
    &= \| E(AA^\t)^{-1/2} E(A)E(U_Y)\|_2^2 \\\notag 
   & = 
   0,	
\end{align*}
	where we used that $Y=Z^\t \alpha_0 +U_Y$, Assumption \Cref{ass:AIndepUy}: $A\independent U_Y$, and Assumption \Cref{ass:MeanZeroA}: $E(A)=0$ (Alternatively,  $E(U_Y|A)=0$).

	Now define a sequence of (everywhere) well-defined estimators $(\tilde{\alpha}_n)_{n\in\N}$ by $$
	\tilde{\alpha}_n := 
\mathbbm{1}_{A_n} \hat{\alpha}_{\text{K}}^n(\lambda^\star_n(p_{\min})) 
 + \mathbbm{1}_{A_n^c} \alpha_0,$$
	for each $n\in \N$. We claim that the loss function $l_{\text{IV}}^n$ evaluated in this estimator tends to zero in probability, i.e., as $n\to\i$ it holds that
\begin{align} \label{eq:lIVinAlphatildeConvPZero}
	l_{\text{IV}}^n(\tilde{\alpha}_n) = \| (n^{-1}\fA^\t \fA)^{-1/2} n^{-1}\fA^\t(\fY - \fZ\tilde{\alpha}_n)\|_2^2 \convp 0.
\end{align}
	This holds by the above observations as for any $\ep'>0$ we have that
	\begin{align*}
	P(|l_{\text{IV}}^n(\tilde{\alpha}_n)| > \ep') 
	&= P((|l_{\text{IV}}^n(\hat{\alpha}_{\text{K}}^n(\lambda^\star_n(p_{\min})) )| > \ep')\cap A_n) +P((|l_{\text{IV}}^n(\alpha_0)| > \ep')\cap A_n^c) \\
	&\leq    P((|H_n| > \ep')\cap A_n) +P((|l_{\text{IV}}^n(\alpha_0)| > \ep')\cap A_n^c) \\
	&\leq P(|H_n| > \ep') +P(|l_{\text{IV}}^n(\alpha_0)| > \ep') \to 0,
	\end{align*}
	when $n\to\i$. 	Now define the random linear maps $g_n:\Omega\times \R^{d_1+q_1}\to \R^q$ by $$
	g_n(\omega,\alpha) := (n^{-1}\fA^\t(\omega) \fA(\omega))^{-1/2}n^{-1} \fA^\t(\omega)  \fZ(\omega) \alpha,$$
	for all $n\in \N$. 	
	The maps $(g_n)$ converge point-wise, that is, for each $\alpha$, in probability to $g:\R^{d_1+q_1}\to \R^q$, given by $
	g(\alpha) := E(AA^\t)^{-1/2}E(AZ^\t)\alpha$, 
	as $n\to \i$. The map $g$ is injective. This follows by  Assumption \Cref{ass:VarianceOfAPositiveDefinite} and  \Cref{ass:EAZtfullrank}, which implies and state that $E(AA^\t)\in \R^{q\times q}$ and $E(AZ^\t)\in \R^{q\times(d_1+q_1)}$ are of full rank, respectively, hence 
	$
	\text{rank}(E(AA^\t)^{-1/2}E(AZ^\t))=\text{rank}(E(AZ^\t))= d_1+q_1,
	$
	since we are in the just- and over-identified setup, where $q\geq d_1+q_1$. We conclude that $g$ is injective, as its matrix representation is of full column rank.
	Furthermore, by \Cref{eq:lIVinAlphatildeConvPZero} it holds that $
g_n(\tilde{\alpha}_n) \convp E(AA^\t)^{-1/2}E(AY)$. 
Hence, we have that
\begin{align*}	
g_n(\tilde{\alpha}_n)-g_n(\alpha_{0}) &\convp E(AA^\t)^{-1/2}E(AY) -E(AA^\t)^{-1/2}E(AZ^\t)\alpha_{0} \\
&= E(AA^\t)^{-1/2}E(AU_Y) \\
&= 0,
\end{align*}

as $n\to \i$. \Cref{lm:LemmaRandomFunctionOfRandomArgument} of \Cref{sec:AuxLemmas} now yields that $\tilde{\alpha}_n \convp \alpha_0$. Finally, note that as $\hat{\alpha}_{\text{K}}^n(\lambda^\star_n(p_{\min}))= \tilde{\alpha}_n$ on $A_n$ we have that
\begin{align*}
	P((\|\hat{\alpha}_{\text{K}}^n(\lambda^\star_n(p_{\min}))-\alpha_0\|>\ep)\cap A_n) &=P((\|\tilde{\alpha}_n-\alpha_0\|>\ep)\cap A_n)\\ 
&\leq P(\|\tilde{\alpha}_n-\alpha_0\|>\ep) \\
&\to 0,
\end{align*}
proving that  $\hat{\alpha}_{\text{PULSE}+}^n(p_{\min}) \convp \alpha_0$, as $n\to\i$.
\end{proofenv}

\section{Auxiliary Lemmas} 
\label{sec:AuxLemmas}
\medskip

\begin{lemma} \label{lm:LemmaRandomFunctionOfRandomArgument}
	Suppose that $g_n: \mathbb{R}^G \to \mathbb{R}^K$ are random linear maps
	converging point-wise in probability to a non-random  linear map $g:\R^G\to \R^K$ that is injective. If
	$$
	g_n(\hat{\beta}_n - \beta_0) \underset{n\to\i}{\convp}0,
	$$
	then $\hat{\beta}_n$ is a consistent estimator of $\beta_0$. That  is, $	\hat{\beta}_n \underset{n\to\i}{\convp}\beta_0$.
\end{lemma}
\noindent\begin{proofenv}{\Cref{lm:LemmaRandomFunctionOfRandomArgument}}
As $g$ is injective, we have that $\text{rank}(g) = G$, and as such $\text{rank}(g^\t g) = \text(g) = G$ which implies that $g^\t g$ is invertible. Furthermore, by Slutsky's theorem we get that $
g_n\convp g \implies g_n^\t g_n \convp g^\t g$,
as $n\to\i$, that is, for any $\ep>0$, $$
P(||g_n^\t g_n-g^\t g\| \leq \ep) =P(g_n^\t g_n \in \overline{B(g^\t g,\ep)})  \underset{n\to\i}{\to }  1.$$ 
Here $\|\cdot\|$ is any norm on the set of $G\times G$ matrices and $\overline{B(g^\t g,\ep)}$ is the closed ball around $g^\t g$ with radius $\ep$ with respect the this norm.
Now note that the set $\text{NS}_G$ of all non-singular $G\times G$ matrices  is an open subset of all $G\times G$ matrices, which implies that there exists an $\ep>0$, such that  $\overline{B(g^\t g,\ep)} \subset \text{NS}_G.$
Hence, $g_n^\t g_n$ is invertible with probability tending towards 1, that is, $
P(g_n^\t g_n \in \text{NS}_G) \underset{n\to\i}{\to } 1$.
Let $h_n:\Omega\to\text{NS}_G$ be given by $$
h_n(\omega)  := \mathbbm{1}_{(g_n^\t g_n \in \text{NS}_G)} g_n^\t(\omega) g_n(\omega) + \mathbbm{1}_{(g_n^\t g_n \in \text{NS}_G)^c} I.$$
Then $h_n \underset{n\to\i}{\convp} g^\t g$, since for any $\ep >0$
\begin{align*}
	P(\|h_n-g^\t g\| >\ep ) &= P((\|g_n^\t g_n-g^\t g\|> \ep) \cap (g_n^\t g_n \in \text{NS}_G)) \\
	& \qquad + P((\|I-g^\t g\|> \ep )\cap (g_n^\t g_n \in \text{NS}_G)^c) \\
	&\leq  P(\|g_n^\t g_n-g^\t g\|> \ep) + P(g_n^\t g_n \in \text{NS}_G)^c) \\
	&\underset{n\to\i}{\to}  0,
\end{align*}
Continuity of the inverse operator and the continuous mapping theorem, yield that $\|h_n^{-1}\|_{\text{op}}\convp \|(g^\t g)^{-1}\|_{\text{op}} \in \R$ and $\|g_n^\t \|_{\text{op}} \convp \|g^\t\|_{\text{op}}\in\R$ as $n$ tends to infinity, where $\|\cdot\|_{\text{op}}$ is the operator norm induced by the Euclidean norm $\|\cdot\|_2$. Furthermore, 
\begin{align*}
	\|g_n^\t g_n(\hat{\beta}_n-\beta_0) \|_2 &\leq \|g_n^\t \|_{\text{op}} \| g_n(\hat{\beta}_n-\beta_0) \|_2 \\ &\underset{n\to\i}{\convp} \|g^\t \|_{\text{op}}  \cdot  0\\
	&= 0,
\end{align*}

by the assumptions and Slutsky's theorem. Hence, for any  $\ep>0$ 
\begin{align*}
	P(	\|h_n(\hat{\beta}_n-\beta_0) \|_2 > \ep ) &= P(	(\|g_n^\t g_n(\hat{\beta}_n-\beta_0) \|_2 > \ep ) \cap (g_n^\t g_n \in \text{NS}_G) ) \\
	&\qquad + P(	(\|\hat{\beta}_n-\beta_0\|_2 > \ep) \cap  (g_n^\t g_n \in \text{NS}_G)^c) \\
	&\leq  P(	(\|g_n^\t g_n(\hat{\beta}_n-\beta_0) \|_2 > \ep ) ) + P((g_n^\t g_n \in \text{NS}_G)^c)\\
	&\underset{n\to\i}{\to}  0.
\end{align*}
Thus,
\begin{align*}
	\|\hat{\beta}_n-\beta_0\|_2 &= \| h_n^{-1} h_n(\hat{\beta}_n-\beta_0) \|_2 \\& \leq \|h_n^{-1}\|_{\text{op}} \| h_n(\hat{\beta}_n-\beta_0) \|_2 \\ &\underset{n\to\i}{\convp} \|(g^\t g)^{-1}\|_{\text{op}} \cdot 0\\ &=0,
\end{align*}

by Slutsky's theorem, yielding that $\hat{\beta}_n$ is an consistent estimator of $\beta_0$.
\end{proofenv}

\begin{lemma} \label{lm:SolutionIsTightIfNotStationary}
	Let $\hat{\alpha}$ be a solution to a constrained optimization problem of the form
	\begin{align*}
		\begin{array}{ll}
			\underset{\alpha \in \R^{k}}{\mathrm{minimize}} & f(\alpha) \\
			\mathrm{subject \, to} & g(\alpha) \leq c,
		\end{array}
	\end{align*}
	where $f$ is an everywhere differentiable strictly convex function on $\R^k$ for which a stationary point exists, $g$ is continuous and $c\in \R$.  If the stationary point of $f$ is not feasible, then the constraint inequality is tight (active)
	in the solution $\hat{\alpha}$, that is, $
	g(\hat{\alpha}) = c$.
\end{lemma}
\noindent\begin{proofenv}{\Cref{lm:SolutionIsTightIfNotStationary}}
Since	
$\hat{\alpha}$ feasible and the stationary point of $f$ is not feasible,	
we know that $\hat{\alpha}$ is not a stationary point of $f$, hence $Df(\hat{\alpha}) \not = 0$. Now assume that the constraint bound is not tight (active) in the solution $\hat{\alpha}$, that is $g(\hat{\alpha})<c$. By continuity of $g$, we know that there exists an $\epsilon >0$, such that for all $\alpha \in B(\hat{\alpha}, \ep)$, it holds that $g(\alpha)<c$. Furthermore, since $Df(\hat{\alpha}) =0$, we can look at the line segment going through $\hat{\alpha}$ in the direction of the negative gradient of $f$ in $\hat{\alpha}$. That is, $l:\R \to \R^k$ defined by $l(t) = \hat{\alpha} - tDf(\hat{\alpha})$. Note that $$D (f\circ l)(0) = Df(l(0))Dl(0) = -Df(\hat{\alpha}) Df(\hat{\alpha})^\t = -\|Df(\hat{\alpha})\|<0,$$ meaning that the derivative of $f\circ l:\R \to \R$ is negative in zero. Therefore, there exists a $\delta>0$, such that for all $t\in(0,\delta)$ it holds that $f\circ l (t) < f \circ l(0)$, i.e., $$ f(\hat{\alpha}-tDf(\hat{\alpha})) < f(\hat{\alpha}).$$ Thus, for sufficiently small $t'$, is it holds that $t'<\delta$ and $\hat{\alpha}-t'Df(\hat{\alpha}) \in B(\hat{\alpha},\ep)$. We conclude that $\tilde{\alpha}:= \hat{\alpha}-tDf(\hat{\alpha})$ is feasible, $g(\tilde{\alpha})<c$, and super-optimal compared to $\hat{\alpha}$, $f(\tilde{\alpha})< f(\hat{\alpha})$, which contradicts that $\hat{\alpha}$ solves the optimization problem. In words, if the solution is not tight we can take a small step in the negative gradient direction of the objective function and get a better objective value while still being feasible. 
\end{proofenv}

\section{Additional Remarks} 
\label{app:AddRemarks}

\medskip

\begin{remark}[Model misspecification] \label{rm:ModelMispecification}
	\textnormal{Theorem~\ref{sthm:TheoremIntRobustKclas} still holds under
		the following three model misspecifications, which may arise from erroneous non-sample information and unobserved endogenous variables
		(these violations 
		may break the identification of $\alpha_{0,*}$ and 
		generally render 
		the K-class estimators inconsistent even when $\plim\kappa  =1$).}
\begin{itemize}
	\item[\textnormal{(a)}] 	\textnormal{Exclude included endogenous variables. Consider the setup where no hidden variable enters the target equation given by $Y = \gamma_0^\t X + \beta_0^\t A + \ep_Y$, with $\ep_Y \independent A$.  If we 
		erroneously 	
		exclude an  endogenous variable that directly affects $Y$, i.e., $\gamma_{0,-*} \not =0$, this is equivalent to drawing inference from the model $	Y= \gamma_{0,*}^\t X_{*} + \beta_{0,*}^\t A_{*} +U $,
		where $U= \ep_Y + \gamma_{0,-*}^\t X_{-*}$. If $E(A_{-*}U)= E(A_{-*}X_{-*}^\t)\gamma_{0,-*}\not = 0$, we have introduced dependence that renders at least some of the instruments $A_{-*}$ invalid, breaking identifiability.} 
	\item[\textnormal{(b)}] \textnormal{Exclude included exogenous variables. Consider again the setup from $(a)$ where there is no hidden variables entering the target equation. If we 
		erroneously 				
		exclude a exogenous variable that directly affects $Y$, i.e., $\beta_{0,-*} \not =0$, then this is equivalent with drawing inference from the model $	Y= \gamma_{0,*}^\t X_{*} + \beta_{0,*}^\t A_{*} +U $,
		where $U= \ep_Y + \beta_{0,-*}^\t A_{-*}$. It holds that $E(A_{-*}U)= E(A_{-*}A_{-*}^\t)\beta_{0,-*}\not =0$, again rendering the instruments invalid. }%
	\item[\textnormal{(c)}] \textnormal{Possibility of hidden endogenous variables. 
		Consider the case with
		included hidden variables that are directly influenced by the excluded exogenous variables, i.e., $A_{-*}\to H\to Y$. 
		This implies that the excluded exogenous variables $A_{-*}$ are correlated with the collapsed noise variable in the structural equation $Y =  \alpha^\t_{0,*} Z_*  + U$, where $U = \ep_Y + \eta_0^\t H$ with $\eta_0\not =0$. In the case that $E(A_{-*}U) = E(A_{-*}H^\t)\eta_0\not =0$ the instruments are invalid.} 
\end{itemize}

\end{remark}

	\begin{remark}[Connection to the Anderson-Rubin Test] \label{rm:ConnectionToAndersonRubinCI}
		\textnormal{Our acceptance region $
		\cA_n^c(1-p_{\min}) := \{\alpha \in \R^{d_1+q_1}: T_n^c(\alpha) \leq Q_{\chi^2_{q}}(1-p_{\min})\}$, 
		is closely related to the Anderson-Rubin \citep[][]{anderson1949estimation} confidence region of the simultaneous causal parameter $\alpha_0 = (\gamma_0,\alpha_0)$ in an identified model. When the causal parameter $\alpha_0$ is identifiable, i.e., in a just- or over-identified setup $(q\geq d_1+q_2)$ and \Cref{ass:EAZtfullrank} 
		holds, only the causal parameter yields regression residuals $Y-\alpha_0^\t Z$ that are uncorrelated with the exogenous variables $A$. In this restricted setup, our null hypothesis is equivalent with $\tilde{H}_0(\alpha): \alpha = \alpha_0$. 
		The hypothesis $\tilde{H}_0(\alpha)$	
		can be tested by the Anderson-Rubin test and all non-rejected coefficients constitute 
		the Anderson-Rubin confidence region of $\alpha_0$, which is given by $
		\text{CR}_{\text{AR}}^{ex,n}(1-p_{\min}) := \left\{\alpha \in \R^{d_1+q_1}:T^{\text{AR}}_n(\alpha) \leq Q_{F(q,n-q)}(1-p_{\min}) \right\}$, 
		where $Q_{F(q,n-q)}(1-p_{\min})$ is the $1-p_{\min}$ quantile of the $F$ distribution with $q$ and $n-q$ degrees of freedom and the 
		Anderson-Rubin test-statistic $T^{\text{AR}}_n(\alpha)$ is  given by
		\begin{align*}
		T^{\text{AR}}_n(\alpha) :=  \frac{n-q}{q}\frac{(\fY-\fZ\alpha)^\t P_\fA(\fY-\fZ\alpha)}{(\fY-\fZ\alpha)^\t P_\fA^\perp(\fY-\fZ\alpha)} = \frac{n-q}{q}\frac{l_{\text{IV}}^n(\alpha)}{l_{\text{OLS}}^n(\alpha)-l_{\text{IV}}^n(\alpha)}.
		\end{align*}
		The confidence region $\text{CR}_{\text{AR}}^{ex,n}$ is exact whenever several regularity conditions are satisfied, such as
		deterministic exogenous variables and normal distributed errors
		\citep[][Theorem 3]{anderson1949estimation}.
		In a general SEM model the regularity conditions are not fulfilled, but changing the rejection threshold to $Q_{\chi^2_q/q}(1-p_{\min})$, we obtain an asymptotically valid confidence region. That is,
		\begin{align*}
		\text{CR}_{\text{AR}}^{as,n}(1-p_{\min}) := \left\{\alpha \in \R^{d_1+q_1}:T^{\text{AR}}_n(\alpha) \leq Q_{\chi^2_q/q}(1-p_{\min}) \right\},
		\end{align*}
		is an asymptotically valid approximate confidence region
		\citep[][Theorem 6]{anderson1950asymptotic}.	
		This relies on the fact that $T^{\text{AR}}_n(\alpha) \convd \chi^2_q/q$ under the null and $T^{\text{AR}}_n$ diverges to infinity under the general alternative. The test-statistic $T_n^c(\alpha)$ can be seen as a scaled coefficient of determination ($R^2$-statistic) for which $T_n^{\text{AR}}(\alpha)$ is the  corresponding $F$-statistic. That is, one can realize that
		\begin{align*}
		T_n^{\text{AR}}(\alpha) = \frac{n-q}{q} \frac{T_n^c(\alpha)/c(n)}{1-T_n^c(\alpha)/c(n)} \leq Q_{\chi^2_q/q}(1-p_{\min}),
		\end{align*}
		is equivalent to 
		\begin{align*}
		\frac{n-q+ Q_{\chi^2_q}(1-p_{\min})}{c(n)} T_n^c(\alpha) \leq Q_{\chi^2_q}(1-p_{\min}).
		\end{align*}
		Thus, if
		$Q_{\chi^2_q}(1-p_{\min}) \geq q$, then  $\cA_n(1-p_{\min}) \supseteq \text{CR}_{\text{AR}}^{as,n}(1-p_{\min})$ and $\cA_n(1-p_{\min}) \subset \text{CR}_{\text{AR}}^{as,n}(1-p_{\min})$ otherwise, where $\cA_n(1-p_{\min})$ is the acceptance region when using the scaling scheme $c(n)=n$. 
		Furthermore, $\cA_n^c(1-p_{\min})$, the acceptance region under a general scaling scheme $c(n) \sim n$,  is asymptotically equivalent to the Anderson-Rubin approximate confidence region $\text{CR}_{\text{AR}}^{as,n}(1-p_{\min})$. If we choose the specific scaling to be 
		$c(n)=n-q+ Q_{\chi^2_q}(1-p_{\min})$, then 
		they coincide, 
		$\text{CR}_{\text{AR}}^{as,n}(1-p_{\min})=\cA_n^c(1-p_{\min})$ for each $n\in \N$. Whenever the Anderson-Rubin confidence region is exact, we could change the rejection threshold from $Q_{\chi^2_{q}}(1-p_{\min})$ to $c(n)Q_{B(q/2,(n-q)/2)}(1-p_{\min})$ and also get an exact acceptance region, where $B(q/2,(n-q)/2)$ is the Beta distribution with shape and scale parameter $q/2$ and $(n-q)/2$ respectively.}
	\end{remark}
	
\begin{remark}[Connections to pre-test estimators] \label{rm:Pretest}
	\textnormal{
	It has been suggested to use pre-test for choosing	
	 between the TSLS and OLS estimator. When using the Hausman test for endogeneity \citep[][]{hausman1978} one considers the pre-test estimator studied by, e.g., \citet{chmelarova2010} and \citet{guggenberger2010}. If $H$ denotes the Hausman test-statistic that rejects the hypothesis of endogeneity when $H\leq Q$, the pre-test estimator is given by $\alpha_{\mathrm{pretest}}^n=1_{(H\leq Q)}\alpha_{\mathrm{OLS}}^n+1_{(H>Q)}\alpha_{\mathrm{TSLS}}^n$.  The PULSE estimator can  be seen as a 
	 pre-test estimator using the Anderson-Rubin test as a test for endogeneity.  However, PULSE differs from the above in the sense that when endogeneity is not rejected we do not revert to the TSLS estimate but rather to the coefficient within the Anderson-Rubin confidence region that minimizes the mean squared prediction error.}
\end{remark}

	\section{Simulation Study} 
	\label{sec:Experiments}
	
	\medskip

	\subsection{Distributional Robustness} \label{app:DistributionalRobustness}
\begin{figure}[!ht]
		\centering
		\includegraphics[width=\linewidth]{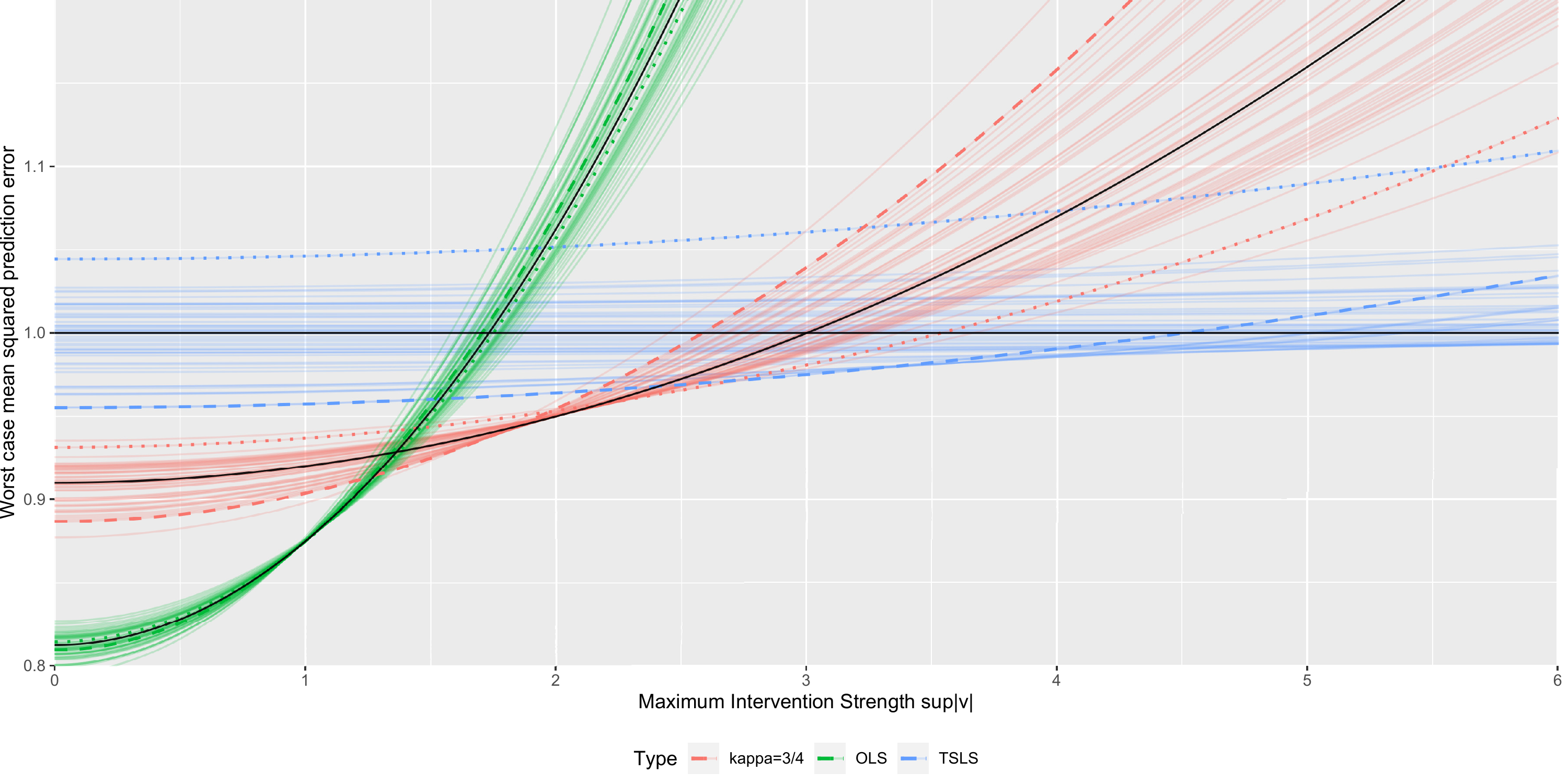}
		\caption{  \normalsize Distributional Robustness of K-class estimators. The plot shows
			the worst case MSPE against the maximum intervention strength condsidered. Each of the 50 repetitions corresponds 
			to three lines (green, red, blue), corresponding to the three
			estimates using $\kappa \in \{0,3/4,1\}$, respectively. The solid black line corresponds to the population coefficients. The OLS is optimal for small interventions but yields a large loss for strong interventions; the TSLS is optimal 
			for large interventions but yields a relatively large loss for small interventions. Choosing a $\kappa$ different from zero and allows us to trade off these two regimes.
			The dashed and the dotted lines correspond
			to the two samples,
			for which 
			the interval on which the 
			$\kappa=3/4$ estimator 
			outperforms 
			TSLS and OLS in terms of worst case MSPE
			is shortest and longest, respectively.
		}
		\label{fig:DistRobustness}
	\end{figure}

	We first illustrate the distributional robustness property of K-class estimators 
	discussed in Section~\ref{sec:intervrobustnessofKclass} in  
	a finite sample setting.
	We consider the model given by
	\begin{align*}
		X := A + U_X, \qquad 
		Y := \gamma  X + U_Y,
	\end{align*}
	where $\gamma=1$ and $A\sim N(0,1)$ independent of $
	\begin{psmallmatrix}
		U_X \\U_Y
	\end{psmallmatrix} \sim \mathcal{N} \left( \begin{psmallmatrix}
		0 \\ 0 
	\end{psmallmatrix}, \begin{psmallmatrix}
		1 & 0.5 \\
		0.5 & 1
	\end{psmallmatrix}\right)$.
	We estimate $\gamma$ from $n=2000$  observations generated by the above system and estimate $\hat \gamma^n_{\text{K}}(\kappa)$ for all $\kappa \in \{0,3/4,1\}$ for which the corresponding population coefficients are given by $\gamma_{\text{K}}(0)=\gamma_{\text{OLS}}=1.25$, $\gamma_{\text{K}}(3/4)=1.1$ and $\gamma_{\text{K}}(1)=\gamma_{\text{TSLS}}=1$.  
	We repeat the simulation 50 times and save the estimated coefficients. 
	\Cref{fig:DistRobustness} 
	illustrates the distributional robustness property of \Cref{sthm:TheoremIntRobustKclas}. For all estimated coefficients $\hat \gamma$ of $\gamma$ we have plotted the analytically computed worst case mean squared prediction error (MSPE) under all hard interventions of absolute strength up to $x$ given by \begin{align} \label{eq:analyticDerivationOfWCMSPE}
		\sup_{|v| \leq x} E^{\mathrm{do}(A:=v)} [ \lp Y-\hat \gamma X\rp^2 ] 
		&=  x^2(1-\hat \gamma)^2 +\hat \gamma^2+3(1-\hat \gamma)
	\end{align}
	against the maximum intervention strength $x$ for the range $x\in[0,6]$. The plot also shows results for the population coefficient as seen in  \citet[Figure~2]{rothenhausler2018anchor}.

	In all 50 repetitions the K-class estimator for $\kappa=3/4$  outperforms both OLS and TSLS in terms of worst case MSPE for maximum intervention strength of $2$.
	This is in line with the theory presented in Section~\ref{sec:intervrobustnessofKclass}.
	In terms of population coefficients  our theoretical results predict
	 that $\kappa=3/4$ is worst case MSPE superior, relative to OLS and TSLS, for all maximum intervention strengths in the range $[1.37,3]$. 
	Among the 50 repetitions we find the outcomes for which the superiority range of $\kappa = 3/4$ has the shortest and longest superiority range length. The shortest superiority range is $[1.27, 2.15]$ and the longest is $[1.46,5.54]$. Clearly, these numbers vary with changing sample size and number of repetitions. For example, with $50$, $200$, $500$, $2000$, $5000$ and $10000$ observations and 50 repetitions, the median lengths of the MSPE superiority range for $\kappa=3/4$ equal 0.82, 1.16, 1.44, 1.74, 1.58 and 1.63, respectively (1.63 is also the length of the theoretically computed interval $[1.37, 3]$).

	\subsection{Estimating causal effects}
	In this subsection we 
	investigate the finite sample behaviour of the PULSE estimator by simulation experiments. We look at how the PULSE estimator fairs in comparison to other well-known single equation estimators in terms of different performance measures.  We generate $n \in \N$ realizations of the SEM in question and construct the estimators of interest based on these $n$ observations. This is repeated $N\in \N$ times, allowing us to estimate different finite sample performance measures of the estimators of interest. The characterization of weak instruments through the minimum eigenvalue of $G_n$, a multivariate analogue to the first stage \textit{F}-statistic, as introduced in \citet{stock2002testing} is important for some of our experimental findings. We refer the reader to \Cref{sec:WeakInst} for a brief introduction.
	\subsubsection{Benchmark Estimators and Performance Measures}
	\label{sec:ExpResultsEstCausPerformanceMeasures}
	We compare the PULSE(5)  estimator, that is PULSE with $p_{\min}=0.05$, to four specific K-class estimators that are well-known to have second moments (in sufficiently over-identified setups). This will allow us to conduct both bias and mean squared error analysis of estimators. 
	Most importantly, we benchmark against  Fuller estimators. 
	The $\kappa$-parameter of the Fuller estimators are given by 
	$\kappa_{\text{FUL}}^n(a) =\kappa_{\text{LIML}}^n -\frac{a}{n-q},$
	where $n-q$ is the degrees of freedom in the first stage regression, $a>0$ is a hyper parameter 
	and $\kappa_{\text{LIML}}^n$ is the stochastic $\kappa$-parameter corresponding the to LIML estimator. One way to represent the $\kappa$-parameter of the LIML estimator is
	$
	\kappa_{\text{LIML}} = \lambda_{\min}(W_1W^{-1})  %
	$
	where $\lambda_{\min}$ denotes the smallest eigenvalue, $W_1$ and $W$ are defined as $
	W = [
	\fY  \, \, \fX
	]^\t P_\fA^\perp [
	\fY  \, \, \fX
	]$ and $W_1 =[
	\fY  \, \, \fX
	]^\t P_{\fA_*}^\perp [
	\fY  \, \, \fX
	],$
	and $P_\fA^\perp = \fI-\fA (\fA^\t \fA)^{-1}\fA^\t$; see, e.g.,  \citet{amemiya1985advanced}.  We choose to benchmark the PULSE estimator against the
	following K-class estimators: OLS ($\kappa=0$), TSLS ($\kappa=1$), Fuller(1) ($\kappa =\kappa_{\text{FUL}}^n(1)$) and Fuller(4) ($\kappa =\kappa_{\text{FUL}}^n(4)$).
	
	The Fuller(1) estimator is approximately unbiased in that the mean bias is zero up to $\mathcal{O}(n^{-2})$ \citep[][Theorem 1]{fuller1977some} and  Fuller(4) exhibits approximate superiority in terms of MSE compared to all other Fuller estimators \citep[][Corollary 2]{fuller1977some}. As we shall see below the PULSE estimator has good MSE performance when instruments are weak and therefore we especially benchmark against Fuller(4) which  has shown better MSE performance than TSLS in simulation studies when instruments are weak; see e.g.\  \citet{hahn2004estimation}. In the over-identified setup we let the PULSE estimator revert to Fuller(4) whenever the dual representation is infeasible.

	We compare the estimators in terms of bias and mean squared error (MSE), which for an $n$-sample estimator $\hat{\alpha}_n$ with target $\alpha\in \R^{d_1+q_1}$ 
	are given by $
	\text{Bias}(\hat \alpha_n) = E(\hat \alpha_n)-\alpha\in \R^{d_1+q_1}$, $
	\text{MSE}(\hat \alpha_n) = E[(\hat \alpha_n-\alpha) (\hat \alpha_n-\alpha)^\t]\in \R^{(d_1+q_1)\times (d_1+q_1)}$. 
	The empirical quantities, estimated 
	from $N$ independent repetitions are denoted by 
	$\widehat{\text{Bias}}(\hat{\alpha}_n)$
	and
	$\widehat{\text{MSE}}(\hat \alpha_n)$.
	In the multivariate setting,
	we compare biases by comparing their Euclidean norms 
	%
	%
	%
	%
	%
	When comparing MSEs, we 
	call
	$\hat{\alpha}_n$ 
	MSE superior to 
	$\tilde{\alpha}_n$ 
	if they are ordered in the partial ordering generated by the proper cone of positive semi-definite matrices (that is, $\widehat{\text{MSE}}(\tilde{\alpha}_n) - \widehat{\text{MSE}}(\hat\alpha_n)$ is
	positive semi-definite).
	We also consider 
	the ordering of its scalarizations given by the determinant and trace (the latter satisfies
	$\text{trace}(\widehat{\text{MSE}}(\hat{\alpha}_n)) = \text{trace}(\widehat{\text{Var}}(\hat{\alpha}_n))+ \|\widehat{\text{Bias}}(\hat{\alpha}_n)\|_2^2$).
	%
	
	We conduct the simulation experiments 
	even though it is not proved that the
	PULSE estimator has finite second moments. 
	In the simulations, the empirical estimates of the mean squared error were stable, possibly
	even more so than for the Fuller estimators for which we know that second moments exists in settings where the noise is Gaussian; see e.g.,\ \citet{fuller1977some,chao2012expository}.
	
Below we describe two multivariate simulation experiments and refer the reader to \Cref{sec:SimUnivariate} in the main paper for a univariate simulation experiment.

	\subsubsection{Varying Confounding Multivariate Experiment.} \label{sec:ExpResultsEstCausalMultidim}
\begin{figure}[h] 
		\centering
		\begin{minipage}{0.40\textwidth}
			\vspace*{\fill}
			
			\begin{align*}
				A &:= N_A \in \mathbb{R}^2, \\[7pt]
				H &:= N_H \in \mathbb{R}^2, \\[7pt]
				X &:=  \xi^\t A +  \delta^\t H + N_X \in \mathbb{R}^2 , \\[7pt]
				Y &:=  \gamma^\t X +  \mu^\t H + N_Y  \in \mathbb{R}.
			\end{align*}
			\vspace*{\fill}
		\end{minipage}	
		\begin{minipage}{0.4\textwidth}
			\centering
			\includegraphics[width=\textwidth]{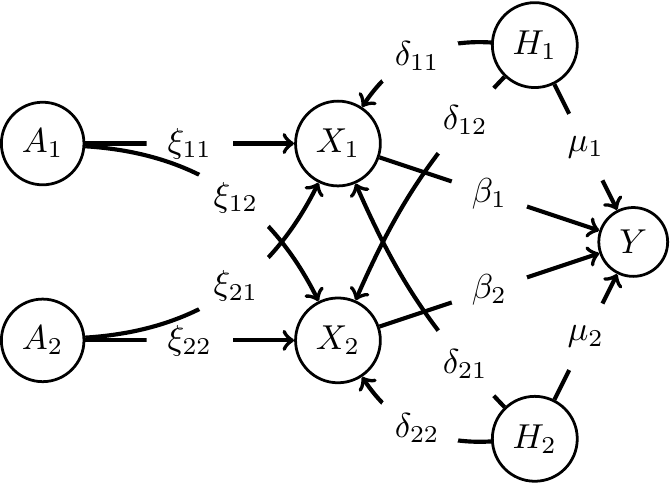}
		\end{minipage}	
		\caption{ The SEM and graph representation used for simulating data in the experiments described in Section~\ref{sec:ExpResultsEstCausalMultidim}.} \label{fig:IV2d_crossdelta}
	\end{figure}

	In this simulation scheme we consider just-identified two-dimensional instrumental variable models with the 
	SEM and causal graph illustrated in \Cref{fig:IV2d_crossdelta}.  Since we want to compare MSE statistics that require estimators with second moments we drop comparisons with the TSLS estimator. 
	
	Here, $\xi, \delta\in \R^{2\times 2}$, $\mu\in \R^2$ and $(N_A,N_X,N_X,N_Y)$ are independent noise innovations. We let $\gamma=(0,0)$ and let the noise innovations for $A,H,Y$ have distribution $(N_A ,N_H, N_Y) \sim  \mathcal{N} (0, I )$. 
	
	We randomly generate 10000 models by letting $
	N_X \sim \mathcal{N} \left( \begin{psmallmatrix}
		0 \\ 0 
	\end{psmallmatrix}, \begin{psmallmatrix}
		\sigma_{1}^2 & 0 \\
		0 & \sigma_{2}^2
	\end{psmallmatrix} \right)$, 
	where the standard deviations is drawn by $\sigma_{1}^2,\sigma_{2}^2\sim \text{Unif}(0.1,1)$ and all other model coefficients are drawn according to $
	\xi_{11},\xi_{12},\xi_{21},\xi_{22},\delta_{11},\delta_{12},\delta_{21},\delta_{22},\mu_1,\mu_2 \sim \text{Unif}(-2,2)$. 
	The hidden confounding induces dependence between the collapsed noise variables $U_X = \delta^\t H + N_X$ and $U_Y = \mu^\t H + N_Y$, which we capture by a normalized cross covariance vector $\rho := \Sigma_{U_X}^{-1/2} \Sigma_{U_XU_Y} \Sigma_{U_Y}^{-1/2}\in \R^2$, where $\Sigma_{U_X} = \text{Var}(U_X)$, $\Sigma_{U_XU_Y}=\text{Cov}(U_X,U_Y)$ and $\Sigma_{U_Y}= \text{Var}(U_Y)$. As such, the degree of confounding can be explained by the norm of $\rho$ given by $
	\| \rho\|_2^2 =  \Sigma_{U_YU_X} \Sigma_{U_X}^{-1}\Sigma_{U_XU_Y}/\Sigma_{U_Y} = \mu^\t \delta (\delta^\t \delta + \mathrm{diag}(\sigma_{1}^2,\sigma_{2}^2))^{-1}{\delta}^\t \mu /(\mu^\t \mu + 1)$. 
	For each of the 10000 generated models we simulate $n=50$ observations and compute the PULSE and benchmark estimators and repeat this $N=5000$ times to estimate the performance measures. 
	
\begin{figure}[ht!] 
		\centering\includegraphics[width=\linewidth]{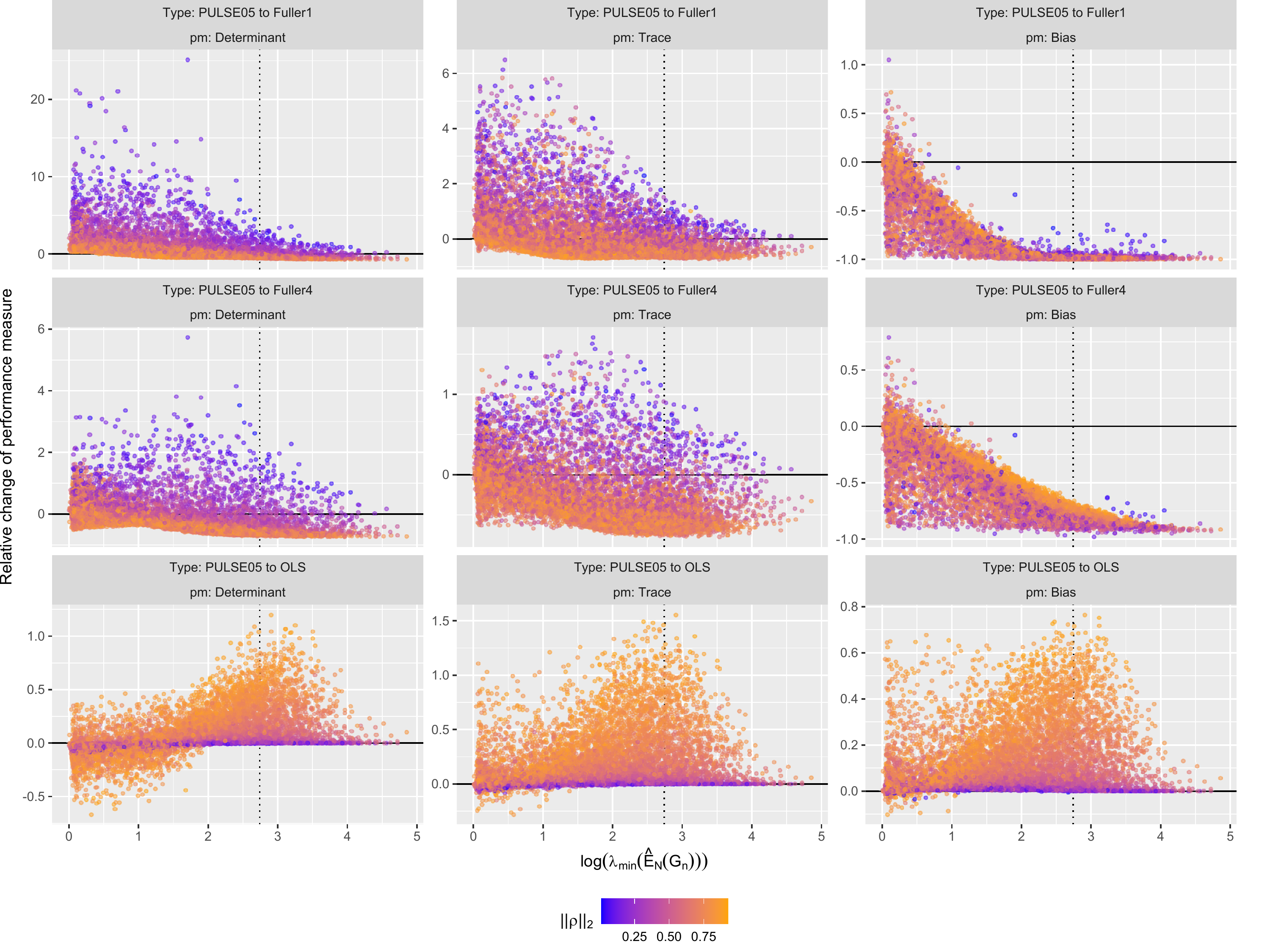}
		\caption{  \normalsize Illustrations of the relative change in the determinant (left) and trace (middle) of the MSE matrix and the Euclidean norm of the bias vector (right) (a positive relative change means that PULSE is better).  Each of the 10000 models corresponds to a point which is color-graded according the the value of $\|\rho\|_2$ (which indicates the strength of confounding), see \Cref{sec:ExpResultsEstCausalMultidim}. PULSE tends to outperform the Fuller estimators for weak instruments and weak confounding.	
		The vertical dotted line at $\log(15.5)$ corresponds to a rejection threshold for weak instruments based on relative change in bias for Fuller estimators \citep[][Table 5.3]{stock2002testing}. Note that the lowest possible negative relative change is $-1$.  }\label{fig:AllRandom_Beta00}
	\end{figure}

	\Cref{fig:AllRandom_Beta00} shows 
	the relative change in the determinant and trace of the MSE matrix and the Euclidean norm of the bias vector.  
	Similarly to the univariate setup, 
	PULSE seems to perform better than Fuller(1) and Fuller(4) in terms of the determinant and trace for 
	settings with weak
	confounding (small $\|\rho\|_2$)
	and weak instruments (small $\lambda_{\min}(\hat E_NG_n)$).
	Most of the MSE matrices do not allow for an ordering:
	PULSE is MSE superior to Fuller(1), 
	Fuller(4), and  
	OLS 
	in 
	9.2\%, 4.6\% and $1\%$ of the cases, while 
	the MSE matrices are not ordered  
	in 
	90.8\%, 95.4\% and 95.8\% of the cases. Note that both Fuller(1) and Fuller(4) is never MSE superior to PULSE.
	In contrast to the univariate setup, there are  models with very  weak instruments for which Fuller outperforms PULSE; these models seems to be exclusively with strong confounding.
	We also see models with strong confounding and moderate to strong instrument strength where the PULSE is superior and models with weak confounding where PULSE is inferior. Hence, the degree of confounding $\|\rho\|_2$  does not completely characterize whether or not PULSE is superior to the Fuller estimators in terms of MSE performance measures in the multi-dimensional setting.
	In regards to the bias we see that both Fuller estimators are less biased than PULSE for all but a few models with very weak instruments. Furthermore, PULSE is for models with strong confounding less biased than OLS but has comparable bias for models with small to moderate confounding.

	We also conducted the above simulation experiment for $\gamma=(1,1)$ and $\gamma=(-1,1)$. The results (not shown but available in the folder 'Plots' in the code repository) are similar to the case $\beta = (0,0)$ and the above observations still apply. \Cref{sec:AppFigs} shows the results of additional experiments,
	where we consider, e.g., PULSE with $p_{\min} = 0.1$.
	
	\subsubsection{Fixed Confounding Multivariate Experiment.}
	In the varying confounding experiment, we saw that when  $\|\rho\|_2$ is small then the majority of the simulated models had PULSE superior to Fuller(1) and Fuller(4) in terms of the determinant and trace of MSE. However, we also saw models with large $\|\rho\|_2$ where PULSE was still superior and models with small $\|\rho\|_2$ where PULSE was inferior. 
	In this experiment, we will investigate this further by fixing the confounding strength $\|\rho\|_2$ and investigating other model aspects that affect which estimator is superior. That is, we consider models with structural assignments given by
	\begin{align*}
		A := N_A \in \mathbb{R}^2, \quad 
		X :=  \xi^\t A +U_X\in \mathbb{R}^2 ,  \quad 
		Y :=  \gamma^\t X + U_Y \in \mathbb{R},
	\end{align*}
	for some $\xi = \in \R^{2\times 2}$ and independent noise innovations $(N_A,(U_X,U_Y))$. We let $\gamma=(0,0)$ and fix the noise innovations for $A$ with distribution $N_A \sim  \mathcal{N} (0, I )$. We let
	\begin{align*}
		\begin{pmatrix}
			U_X \\ U_Y 
		\end{pmatrix} \sim  \cN \lp \begin{pmatrix}
			0 \\ 0 \\ 0 
		\end{pmatrix}, \begin{pmatrix}
			1 & \eta & \phi_1 \\
			\eta & 1 & \phi_2\\
			\phi_1 & \phi_2 & 1
		\end{pmatrix}\rp,
	\end{align*}
	for some $\eta,\phi_1,\phi_2\in[0,1)$. With this noise structure we have that
	$
	\|\rho \|_2^2 = (\phi_1^2 +\phi_2^2-2\eta \phi_1\phi_2)/(1-\eta^2),
	$
	and when $\phi=\phi_1=\phi_2$ it holds that $\|\rho \|_2^2  =  2\phi^2/(1+\eta)$.
	We randomly generate 5000 copies of $\xi$ with each entry drawn by $\text{Unif}(-2,2)$ distribution. For each model, that is, each combination of selected noise-parameter values and $\xi$, we simulate $n=50$ observations and compute the estimators. This is repeated $N=5000$ times to estimate the performance measures.
	
\begin{figure}[!ht] 
		\centering\includegraphics[width=\linewidth]{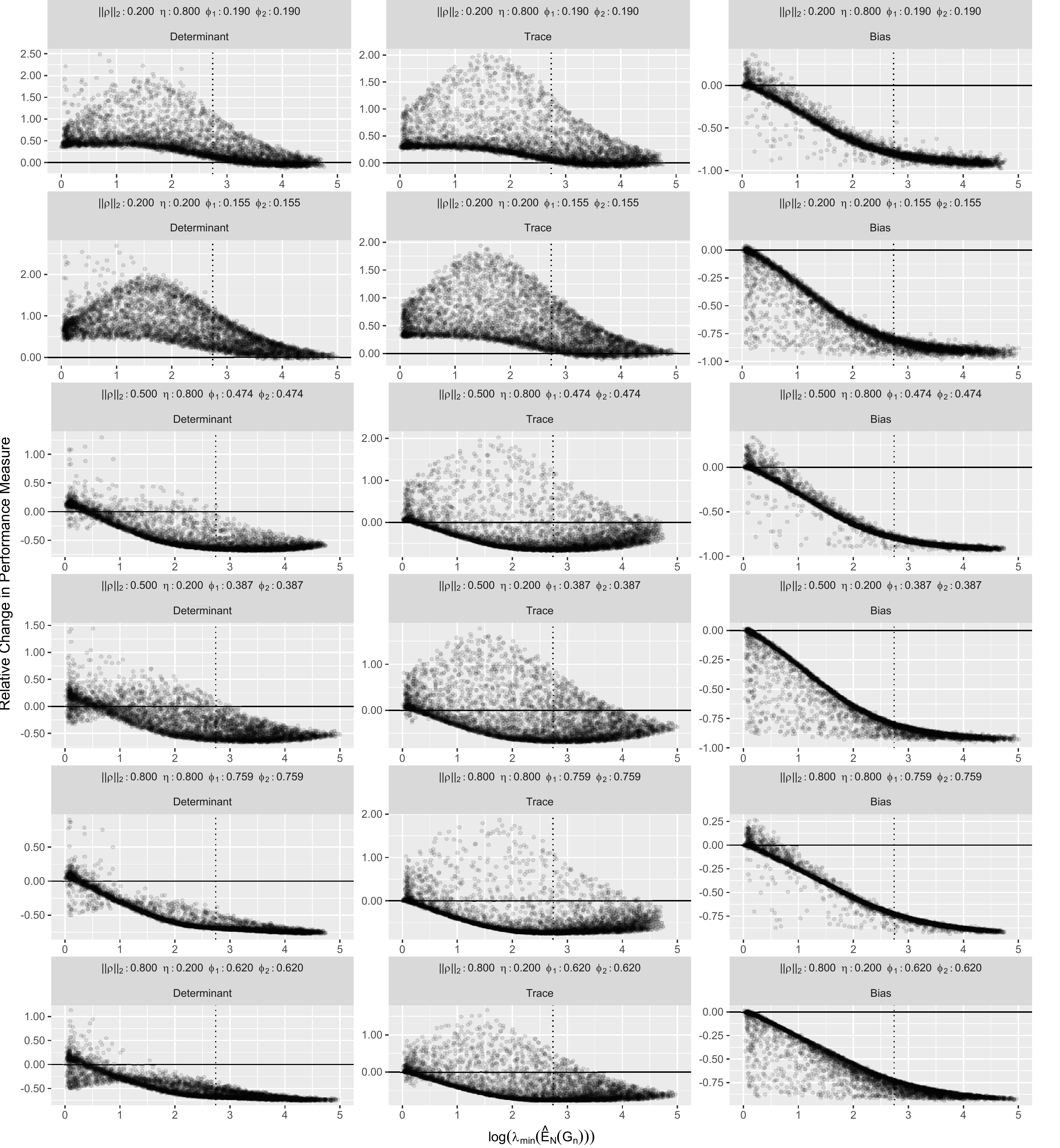}
		\caption{  \normalsize Illustrations of the relative change from PULSE to Fuller(4) in the determinant and trace of the MSE matrix and the Euclidean norm of the bias vector. 	 The vertical dotted line at $\log(15.5)$ corresponds to a rejection threshold for weak instruments based on relative change in bias for Fuller estimators \citep[][Table 5.3]{stock2002testing}.  }\label{fig:AllRandomFixedNoiseCorr}
	\end{figure}

	In \Cref{fig:AllRandomFixedNoiseCorr} we have illustrated the relative change in the performance measures when comparing 
	PULSE to Fuller(4). For setups with weak confounding ($\|\rho\|_2=0.2$), it is seen that if instruments are sufficiently weak ($\lambda_{\min}(\hat{E}_N(G_n))\leq 15.5$), then PULSE is superior to Fuller(4) in terms of both the determinant and trace performance measures. For setups with larger confounding there are still models where PULSE is superior but the characterization of superiority by weakness of instruments is no longer valid.  
	
	In \Cref{tbl:parametervalues} the percentage of models for which PULSE is superior to Fuller(4) in terms of the MSE partial ordering, determinant and trace performance measures is presented. It is seen that setups with identical $\|\rho\|_2$ does 
	not yield
	similar comparisons between PULSE and Fuller(4). 
	
	\begin{table}[ht]
		\caption{MSE superiority} \label{tbl:parametervalues}
		\begin{center}
			\begin{tabu}to \textwidth { X[c] X[c] X[c] X[c] | X[r] X[r] X[c]}
				\toprule \toprule
				\multicolumn{4}{c}{Model Parameters}  & \multicolumn{3}{c}{PULSE Superiority (\%)}  \\ 
				$\|\rho\|_2$ & 	$\eta$ & $\phi_1$ & $\phi_2$ & MSE & determinant & trace\\ \midrule
				0.20 & 0.80 & 0.19 & 0.19 & 48.46 & 85.52 & 86.74 \\ 
				0.20 & 0.20 & 0.15 & 0.15 & 32.34 & 98.66 & 92.16 \\ 
				0.50 & 0.80 & 0.47 & 0.47 & 1.60 & 13.04 & 19.80 \\ 
				0.50 & 0.20 & 0.39 & 0.39 & 0.76 & 19.86 & 27.86 \\ 
				0.80 & 0.80 & 0.76 & 0.76 & 0.14 & 7.48 & 12.80 \\ 
				0.80 & 0.20 & 0.62 & 0.62 & 0.06 & 7.64 & 15.50 \\
				\bottomrule\bottomrule
			\end{tabu}
		\end{center}
		\footnotesize
		\renewcommand{\baselineskip}{11pt}
		\textbf{Note:} The rows show different  noise-parameter values for the different experimental setups. The last three columns describe the percentage of models (out of the 5000 randomly generated models) for which PULSE (with $p_{\min}=0.05$) is superior to Fuller(4) in terms of the MSE partial ordering, determinant and trace performance measures. Whenever PULSE is not superior to Fuller(4) in terms of the MSE partial ordering the MSE matrices are not comparable.
	\end{table}

	For any two setups with identical confounding strength $\|\rho\|_2$ we see that decreasing $\eta$ yields a larger percentage of models for which PULSE is superior in terms of the determinant and trace. Furthermore, we see that decreasing $\|\rho\|_2$ (for fixed $\eta$) has a similar effect. Thus, it seems that both $\rho$ and $\eta$ negatively influences the size of the parameter space of $\xi$ for which PULSE is superior to Fuller(4) in terms of both the determinant and trace performance measures. However, superiority with respect to the partial ordering of the MSE matrices does not exhibit similar behaviour. Decreasing $\|\rho\|_2$ (for fixed $\eta$) still leads to a percentage increase but decreasing $\eta$ (for fixed $\|\rho\|_2$) leads to a percentage decrease, of models for which PULSE is superior to Fuller(4).

\newpage	
\subsection{Under-identified setup} \label{app:underidentifiedexperiment}
In an under-identified setup the causal parameter is not identified by instrumental variable methods. Instead the usual two stage least square procedure, $\argmin_\alpha l_\mathrm{IV}(\alpha)$, yields an entire linear solution space of coefficients that renders the regression residuals uncorrelated with the instruments. The causal coefficient lies within this solution space but we are unable to identify it. In the under-identified setup, the population PULSE coefficient is the point in the solution space which provides the best mean squared prediction error. That is, the population PULSE coefficient is given by
\begin{align*}
	\alpha^*= \argmin_{\alpha: E[A(Y-Z\alpha)]=0} E[(Y-Z\alpha)^2] = \argmin_{\alpha:l_{\mathrm{IV}}(\alpha)=0}l_{\mathrm{OLS}}(\alpha).
\end{align*}
The PULSE estimator in the under-identified setup remains unchanged from the exposition in the main paper. 
Here, the function $l_{\mathrm{IV}}^n$ does not have a unique solution but we can define a modified TSLS estimator 
\begin{align*}
\hat \alpha_{\mathrm{TSLS.mod}}^n := \lim_{\kappa\uparrow 1}\alpha_{\mathrm{K}}^n(\kappa) = \argmin_{\alpha: l_{\mathrm{IV}}^n(\alpha)=0} l_{\mathrm{OLS}}^n (\alpha).
\end{align*}
The modified TSLS estimator is the minimum of a quadratic  
function
subject to a feasible linear constraint, and can be computed efficiently using QP solvers. 
\subsubsection{Under-indentified Example} \label{sec:underIDexample}
\begin{figure}[!ht]
	\centering\includegraphics[scale=1]{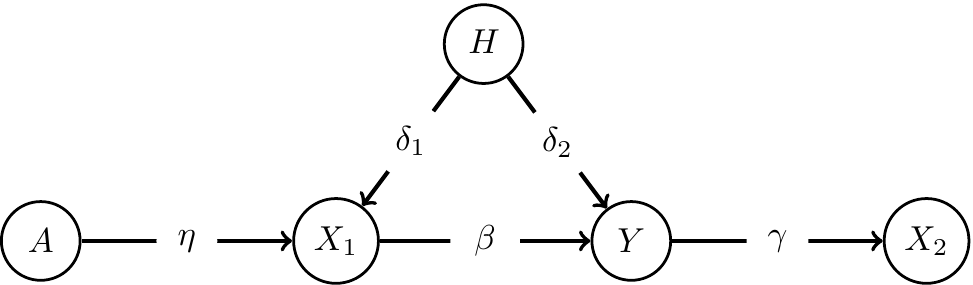}
	\caption{Causal graph of the under-indentified setup in \Cref{sec:underIDexample} Here, $H$ is hidden and the causal parameter $\beta$ is, in general, not identifiable from the distribution over $(A, X_1, X_2, Y)$. 
Existing methods in machine learning try to find invariant sets of covariates (i.e., sets $S$ that, after regressing $Y$ on $X_S$, yield residuals which are uncorrelated with $A$). In this example, no such set exists. PULSE finds a solution and outputs a vector with non-zero coefficients for $X_1$ and $X_2$.}
	\label{fig:underidcausalgraph}
\end{figure}

Consider an under-identified setup with structural assignments given by
\begin{align*}
	A &:= \ep_A, \quad \quad H := \ep_H, \quad \quad X_1 := \eta A + \delta_1H + \ep_1,\\
	Y &:=	\beta X_1 + \delta_2 H + \ep_Y,\quad \quad X_2 := \gamma Y + \ep_2,
\end{align*}
with $(\ep_A,\ep_H,\ep_Y,\ep_1,\ep_2)\sim \cN(0,I_5)$. The causal graph of this structural equation model is illustrated in \Cref{fig:underidcausalgraph}. 
In general, the causal parameter $\beta$ is not identifiable.
Existing methods \citep[e.g.,][]{Peters2016jrssb, rojas2018invariant, pfister2019stabilizing} propose to look for invariant sets that yield residuals which are uncorrelated with $A$ after regressing $Y$ on that set. In general, because of the hidden variable $H$, no such sets exist either. 
The best predictive model under all invariant models, however, is still well-defined. 
To see this, let us derive the population PULSE coefficient 
\begin{align*}
	\alpha^* = \argmin_{\alpha: l_{\mathrm{IV}}(\alpha)=0} E[(Y-\alpha_1X_1-\alpha_2X_2)^2] .
\end{align*}
We know that a necessary and sufficient condition for $l_{\mathrm{IV}}(\alpha)=0$ is that $\Corr(Y-\alpha_1X_1 -\alpha_2X_2,A)=0$. We have
\begin{align*}
	Y-\alpha_1X_1-\alpha_2X_2 &=  Y-\alpha_1X_1-\alpha_2(\gamma Y + \ep_2) \\
	&= (1-\alpha_2 \gamma) (\beta X_1 + \delta_2 H + \ep_Y) -\alpha_1X_1-\alpha_2 \ep_2 \\
	&= (\beta-\alpha_1-\alpha_2\gamma \beta) X_1 + (1-\alpha_2\gamma)\delta_2 H + (1-\alpha_2\gamma)\ep_Y - \alpha_2 \ep_2.
\end{align*}
As $\eta\not = 0$, the regression residuals are uncorreleted with $A$ if and only if $  \alpha_1 = (1-\alpha_2\gamma) \beta$. Hence, 
\begin{align*}
	\alpha^* &= \argmin_{\alpha: \alpha_1 = (1-\alpha_2\gamma)\beta} E[((1-\alpha_2\gamma)\delta_2 H + (1-\alpha_2\gamma)\ep_Y - \alpha_2 \ep_2)^2] \\
	&= \argmin_{\alpha: \alpha_1 = (1-\alpha_2\gamma)\beta} (1-\alpha_2\gamma)^2\delta_2^2 \mathrm{Var}(H) + (1-\alpha_2\gamma)^2 \mathrm{Var}(\ep_Y) + \alpha_2^2 \mathrm{Var}(\ep_2).
\end{align*}
The latter function is convex in $\alpha_2$, so the minimum is attained in a stationary point. We have that
\begin{align*}
	&\frac{\partial}{\partial \alpha_2}   (1-\alpha_2\gamma)^2\delta_2^2 \mathrm{Var}(H) + (1-\alpha_2\gamma)^2 \mathrm{Var}(\ep_Y) +\alpha_2^2 \mathrm{Var}(\ep_2) \\
	&= 2\left[ \alpha_2(\mathrm{Var}(\ep_2)+\gamma^2\delta_2^2 \mathrm{Var}(H) + \gamma^2 \mathrm{Var}(\ep_Y)) - \gamma \delta_2^2 \mathrm{Var}(H) - \gamma\mathrm{Var}(\ep_Y)    \right]=0,
\end{align*}
if and only if 
\begin{align*}
	\alpha_2(\mathrm{Var}(\ep_2)+\gamma^2\delta_2^2 \mathrm{Var}(H) + \gamma^2 \mathrm{Var}(\ep_Y)=   \gamma \delta_2^2 \mathrm{Var}(H) + \gamma\mathrm{Var}(\ep_Y).
\end{align*}
Hence,
\begin{align} \label{eq:alst}
	\alpha_2^* &= \frac{(\mathrm{Var}(\ep_Y)+\delta_2^2\mathrm{Var}(H))\gamma }{\mathrm{Var}(\ep_2)+(\mathrm{Var}(\ep_Y)+\delta_2^2\mathrm{Var}(H))\gamma^2} = \frac{(1+\delta_2^2)\gamma}{1+(1+\delta_2^2)\gamma^2 };
\quad  \alpha_1^* = (1-\alpha_2^*\gamma)\beta.
\end{align}

We now generate 
models by randomly drawing the model coefficients using $\alpha \sim \mathrm{Unif}(1,2)$, $\delta_1 \sim \mathrm{Unif}(1,2), \delta_2 \sim \mathrm{Unif}(1,2), \gamma \sim \mathrm{Unif}(1,2)$ and $\eta \sim \mathrm{Unif}(0.1,1)$
and compute the corresponding population quantities according to Equation~\eqref{eq:alst}.

\begin{figure}[!ht]
	\begin{center}
		\includegraphics[width=\linewidth]{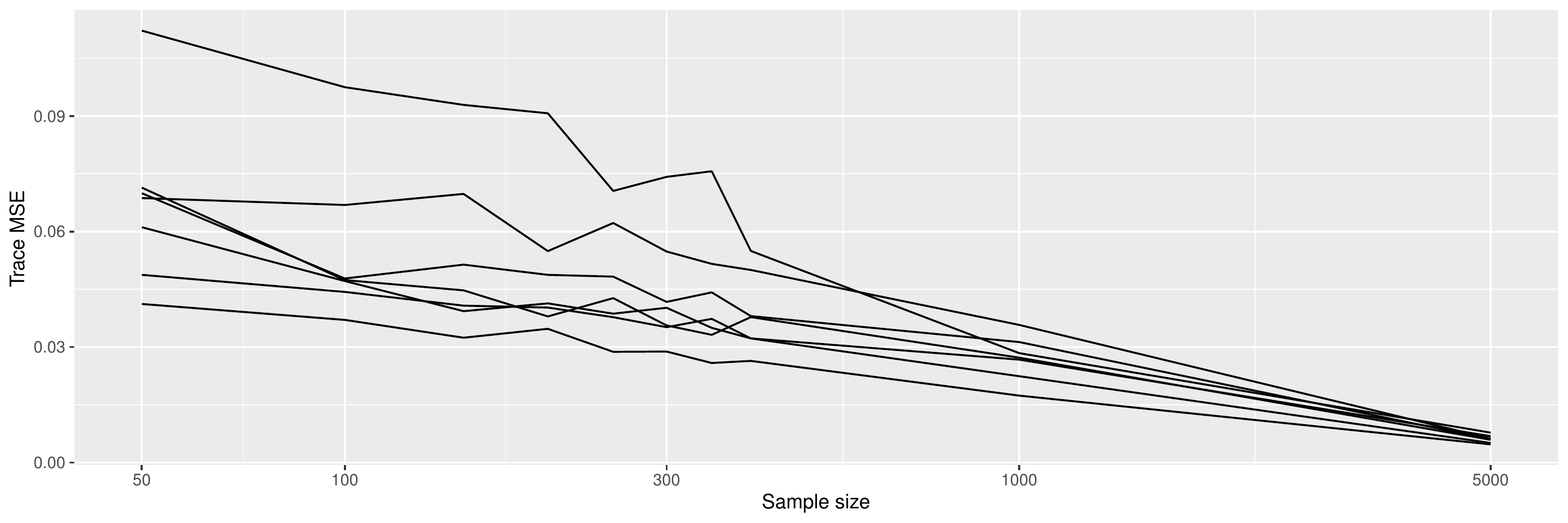}
	\end{center}
	\caption{Illustration of the trace of the estimated MSE matrix of the PULSE estimator in the under-identified setup based on 100 repetitions. PULSE converges towards the population quantities computed in Equation~\eqref{eq:alst}.} \label{fig:underidConvergence}
\end{figure}

For different sample sizes, we then simulate data sets from such models and compute the PULSE estimator.
\Cref{fig:underidConvergence} shows the trace of the estimated MSE of the PULSE estimator (with $p_{\min}=0.05$) 
when comparing to the population quantity derived above.
For each model and sample size, the MSE is estimated based on 100 repetitions. As sample size increases, the MSE indeed approaches the population quantity.

As a comparison, we also implemented the TSLS modification from Equation~\eqref{eq:alst}. 
Similarly to the identified setups, the TSLS modification may come with poor finite sample properties, in particular for weak instruments and small sample size. Indeed, in this example
we observe that PULSE has superior MSE properties for small sample sizes. For example, the trace MSE for the PULSE estimator is on average (over 1000 random models) 50\%  lower than the trace MSE of the modified TSLS estimator for a sample size of 50.

\section{Empirical Applications} 
\label{app:EmpricalApp}
\medskip

We now consider three classical 
instrumental variable applications (see \citet{albouy2012colonial} and \citet{buckles2013season} for discussions on the underlying assumptions). 
	\begin{itemize}
		\item[\ref{sec:QOB}] ``Does compulsory school attendance affect schooling and earnings?'' by \cite{angrist1991does}. This paper investigates the effects of education on wages. The endogenous effect of education on wages are remedied by instrumenting education on quarter of birth indicators.
		\item[\ref{sec:Proximity}] ``Using geographic variation in college proximity to estimate the return to schooling'' by \cite{card1993using}. This paper also investigates the effects of education on wages. In this paper education is instrumented by proximity to college indicator.
		\item[\ref{sec:Colonial}] ``The colonial origins of comparative development: An empirical investigation'' by \cite{acemoglu2001colonial}. This paper investigates the effects of extractive institutions (proxied by protection against expropriation) on the gross domestic product (GDP) per capita. The endogeneity of the explanatory variables are remedied by instrumenting protection against expropriation on early European settler mortality rates.
	\end{itemize}

For each study, we replicate the OLS and TSLS estimates of these studies and provide in addition the corresponding Fuller(4) (see
\Cref{sec:ExpResultsEstCausPerformanceMeasures}) 
and PULSE estimates. 
Since we do not have access to interventional data,
we cannot directly test the distributional 
robustness properties discussed in Section~\ref{sec:RobustnessOfKclass}.
For the third study, however, the exogenous variable is continuous, which allows us to investigate 
distributional robustness empirically 
by holding out data points with extreme values of the exogenous variable and predict on these held-out data.

For the remainder of this section we use the PULSE estimator with $p_{\min}=0.05$ and the test scaling-scheme that renders the test equivalent to the asymptotic version of the Anderson-Rubin test (see \Cref{sec:VanishCorr}). 
Code replicating this analysis is available on GitHub.\footnote{\url{https://github.com/MartinEmilJakobsen/PULSE/tree/master/Empirical_Applications}}

	\subsection{\cite{angrist1991does}} \label{sec:QOB}
The dataset of \cite{angrist1991does} consists, in part, of 1980 US census data of 329,509 men born between 1930--1939. The endogenous target of interest is log weakly wages and the main endogenous regressor is years of education is instrumented on year and quarter of birth indicators. We consider four models M1--M4 corresponding to the models presented in column (1)--(8) in Table 5 of \cite{angrist1991does}. Model M1 is given by the structural reduced form equations
\begin{align*}
	\log \mathrm{weakly \, wage} &= \mathrm{educ}\cdot \gamma + \sum_{i} \mathrm{YR}_i \cdot \beta_i + U_1, \\
	 \mathrm{educ} &= \sum_{i} \mathrm{YR}_i \cdot  \delta_i + \sum_{i,j} \mathrm{YR}_i\cdot \mathrm{QOB}_j \cdot \delta_{i,j} + U_2,
\end{align*}
where $\mathrm{educ}$ is years of education, $(\mathrm{YR}_i)$ is year of birth indicators and $(\mathrm{QOB}_j)$ is quarter of birth indicators. Model M2 is given by M1 with the additional included exogenous regressors of age and age-squared. Models M3 and M4 are given by model M1 and M2, respectively, with additional included exogenous indicators describing race, marital status, metropolitan area and eight regional indicators. All models are over-identified, instrumenting education on a total of 30 binary instruments.

\Cref{tbl:QOB} shows the OLS and TSLS estimates, as well as the Fuller(4) and PULSE estimates for the linear effect of education on log weakly wages.  In all models the PULSE estimates coincide with the OLS estimates. 

\begin{table}[H]
	\caption{\label{tbl:QOB}The estimated return of education on log weakly wage. }
	\begin{center}
		\begin{tabu}to \textwidth {r c c c c c c c}
			\toprule \toprule
			Model & OLS & TSLS & FUL & PULSE & Message & Test & Threshold \\
			\midrule
M1 &0.0711 & 0.0891 & 0.0926 & 0.0711 & OLS Acc. & 26.92 & 55.76\\
M2 &0.0711 & 0.0760 & 0.0739 & 0.0711 & OLS Acc. & 23.15 & 55.76\\
M3 &0.0632 & 0.0806 & 0.0835 & 0.0632 & OLS Acc. & 23.79 & 68.67\\
M4 & 0.0632 & 0.0600 & 0.0555 & 0.0632 & OLS Acc. & 19.59 & 68.67\\
			\bottomrule\bottomrule
		\end{tabu}
	\end{center}
	\footnotesize
	\renewcommand{\baselineskip}{11pt}
	\textbf{Note:} 
Point estimates for the return of education on log weakly wage. The OLS and TSLS values coincide with the ones in
Table V of \citet{angrist1991does}.
The right columns show the values of the test statistic (evaluated in the PULSE estimates) and the test rejection thresholds.
For all models, the OLS is accepted and the PULSE coincides with the OLS.
\end{table}


\subsection{\cite{card1993using}} \label{sec:Proximity}
	The dataset of \cite{card1993using} consists of a US National Longitudinal Survey of Young Men spanning from 1966 to 1981. The subset of interest consists of 3010 observations	
	 for which there is recorded a valid wage and education level in a 1976 interview. 
	The endogenous target of interest is log hourly wages and the main endogenous regressor is years of education. 
	Proximity to a four year college, recorded in 1966, is used as an instrument. 
	We consider two models, M1 and M2, corresponding to models in Panel B, column (5) and (6) of Table 3 \citep[][]{card1993using}
	, respectively. 
	 Model M1 is given by regressing the target, log hourly wages, on included exogenous  indicators of race, metropolitan area and region; the included endogenous regressors are years of education, work-experience and work-experience-squared. The endogenous regressors are instrumented by the excluded exogenous variables age, age-squared and indicator of proximity to college.	In model M2, we have model M1 with the addition of several exogenous indicators of parents education level. 
	
	\Cref{tbl:Proximity} shows the OLS and TSLS estimates, as well as the Fuller(4) and PULSE estimates for the linear effect of education on log hourly wages. 
Again, in all models the OLS estimates are not rejected by the Anderson-Rubin test. Hence, all PULSE estimates coincide with the OLS estimates. 
	\begin{table}[H]
		\caption{\label{tbl:Proximity} The estimated return of education on log hourly wages. }
		\begin{center}
			\begin{tabu}to \textwidth {r c c c c c c c}
				\toprule \toprule
				Model & OLS & TSLS & FUL & PULSE & Message & Test & Threshold \\
				\midrule
M1 &0.0747 & 0.1224 & 0.1156   & 0.0747 & OLS Acc. & 1.22 & 26.30\\
M2& 0.0726 & 0.1324 & 0.1283  & 0.0726 & OLS Acc. & 1.71 & 43.77\\
				\bottomrule\bottomrule
			\end{tabu}
		\end{center}
		\footnotesize
		\renewcommand{\baselineskip}{11pt}
		\textbf{Note:}  
		Point estimates for the return of education on log hourly wage. The OLS and TSLS values coincide with the ones shown in
Table 3 of \citet{card1993using}.
The right columns show the values of the test statistic (evaluated in the PULSE estimates) and the test rejection thresholds.
For all models, the OLS is accepted and the PULSE coincides with the OLS.
	\end{table}

%

\subsection{\cite{acemoglu2001colonial}} \label{sec:Colonial}
In \Cref{sec:mainColonialApplication} of the main paper we describe the data and models of \cite{acemoglu2001colonial}. Furthermore, we replicate the OLS and TSLS estimates and presented the corresponding Fuller(4) and PULSE estimates.

To investigate 
distributional robustness,
we conduct an out-of-sample 
mean squared prediction error (MSPE) analysis on a mean-centered dataset of the just-identified identified model M1.  
This is the simplest model proposed in \cite{acemoglu2001colonial} but the MSPE robustness property of \Cref{sthm:TheoremIntRobustKclas} 
is robust to model misspecifications; see \Cref{rm:ModelMispecification}. 
We do not have access to interventional data.
Instead, 
for different values of $n_{\mathrm{test}} \in \mathbb{N}$, that is, 
for each $n_{\mathrm{test}}\in\{4,8,...,32\}$, 
we remove the data points with the  $n_{\mathrm{test}}/2$ lowest and $n_{\mathrm{test}}/2$ highest settler mortality rates.
We then fit the OLS, TSLS, PULSE and Fuller(4) on the remaining $64-n_{\mathrm{test}}$ observations and compute the out-of-sample MSPE on the $n_{\mathrm{test}}$ held-out observations,
measuring the model's ability to generalize.

The instrument has a larger variance on the held-out data and the population robustness property of K-class estimators (see \Cref{sthm:TheoremIntRobustKclas})
suggests 
that PULSE and Fuller(4) 
might generalize slightly better than OLS or TSLS.\footnote{Here, we consider a just-identified model, so the Fuller(4) K-class parameter $\kappa\in(0,1)$.} 
  The results of this analysis is summarised in  \Cref{tbl:MSPECOLONIAL}. 
 Indeed, we see that the 
OLS is optimal for a small number of held-out data points (when little generalization is required) 
and that for an
increasing 
number of held-out data points, 
 PULSE and FULLER(4) outperform the other estimators in terms of MSPE.

For comparison, we also consider
random sample splits, i.e., taking out a random subset of the dataset rather.
Here, no generalization is required and as expected, 
OLS performs better than the other estimates, see \Cref{tbl:MSPECOLONIALrandomsplits}. 
The MSPE is minimized by 
OLS, PULSE, Fuller(4), and TSLS 
in 
65.9\%, 21.8\%, 6.1\%, and  6.2\% of the cases, respectively. 

\begin{table}[h]
	\caption{\label{tbl:MSPECOLONIALrandomsplits} log GPD MSPE orderings on random sample splits. }
	\begin{center}
			\begin{tabu}to \textwidth {r | r r r r }
				\toprule \toprule
				MSPE&\multicolumn{4}{c}{Outperforms} \\ 
				  &  OLS & PULSE & FUL  & TSLS  \\
				\midrule
OLS & \xmark &65.9\% &  79.7\% & 85.3\% \\
PULSE & 34.1\% &\xmark & 87.7\% & 90.5\% \\
FUL & 20.3\% & 12.3\%&\xmark & 93.8\% \\
TSLS & 14.7\% & 9.5\% & 6.2\% & \xmark \\
				\bottomrule\bottomrule
			\end{tabu}
	\end{center}
	\footnotesize
	\renewcommand{\baselineskip}{11pt}
	\textbf{Note:} 
	The table shows generalization performance for different estimators on model M1 of \cite{acemoglu2001colonial}. 
	The data set is split randomly into a subset of	
	90\% of the data (that is, 58 observations) and the MSPE for the OLS, PULSE, Fuller(4), and TSLS 
	are calculated on the remaining 10\% of the data. 
This procedure is repeated 1000 times. 
The table shows how often the estimators outperform each other. E.g., 
OLS has lower MSPE than TSLS in 85.3\% of the cases. 
Here, no generalziation is needed 
and, as expected, the OLS performs best.
\end{table}	

\begin{landscape}
\begin{table}[h]
\caption{\label{tbl:MSPECOLONIAL} log GPD MSPE on extreme out-of-sample instrument observations. }
\begin{center}
\begin{tabu} to \textwidth {r | c c c c| c c  | c c c c } 
\toprule \toprule 
& \multicolumn{4}{c|}{Estimated coefficient} & \multicolumn{2}{c|}{K-class $\kappa$} & \multicolumn{4}{c}{MSPE}   \\ 
$n_{\mathrm{test}}$&OLS&TSLS&PULSE&FUL&PULSE&FUL&OLS&TSLS&PULSE&FUL\\
\midrule
4 & 0.5015 & 1.1592 & 0.7852 & 0.9509 & 0.8286 & 0.9322  & \bt0.2072 & 2.0358 & 0.3211 & 0.8613 \\
6 & 0.5113 & 0.9441 & 0.6590 & 0.8313 & 0.7075 & 0.9298  & \bt0.8282 & 1.5889 & 0.8692 & 1.2034 \\
8 & 0.5017 & 0.9433 & 0.6287 & 0.8150 & 0.6781 & 0.9273  & 0.7800 & 1.5331 & \bt0.7796 & 1.0961 \\
10 & 0.4978 & 0.8795 & 0.5810 & 0.7717 & 0.5733 & 0.9245 & 0.7018 & 1.0850 & \bt0.6769 & 0.8479 \\
12 & 0.4901 & 0.8693 & 0.5390 & 0.7512 & 0.4407 & 0.9216 & 0.6605 & 1.0346 & \bt0.6357 & 0.7788 \\
14 & 0.4748 & 0.8439 & 0.4748 & 0.7091 & 0.0000 & 0.9184 & \bt0.6562 & 0.8910 & \bt0.6562 & 0.6722 \\
16 & 0.4581 & 0.7655 & 0.4581 & 0.6359 & 0.0000 & 0.9149 & 0.7290 & 0.7581 & 0.7290 & \bt0.6573 \\
18 & 0.4247 & 0.6861 & 0.4247 & 0.5451 & 0.0000 & 0.9111 & 0.7476 & \bt0.6263 & 0.7476 & \bt0.6263 \\
20 & 0.3883 & 0.8604 & 0.3883 & 0.6096 & 0.0000 & 0.9070 & 0.8886 & 0.8354 & 0.8886 & \bt0.6632 \\
22 & 0.3789 & 0.8867 & 0.3789 & 0.6046 & 0.0000 & 0.9024 & 0.8285 & 0.8315 & 0.8285 & \bt0.6072 \\
24 & 0.3784 & 0.7016 & 0.3784 & 0.5450 & 0.0000 & 0.8974 & 0.9152 & \bt0.7251 & 0.9152 & 0.7334 \\
26 & 0.4156 & 0.8753 & 0.5240 & 0.6723 & 0.6682 & 0.8919 & 0.8794 & 1.0333 & \bt0.7957 & 0.8012 \\
28 & 0.4155 & 0.7867 & 0.4676 & 0.6306 & 0.4789 & 0.8857 & 0.8340 & 0.8530 & 0.7880 & \bt0.7468 \\
30 & 0.4016 & 0.8725 & 0.4710 & 0.6278 & 0.5754 & 0.8788 & 0.7989 & 0.9223 & 0.7370 & \bt0.6991 \\
32 & 0.4087 & 0.9103 & 0.4893 & 0.6228 & 0.6344 & 0.8710 & 0.7823 & 0.9880 & 0.7225 & \bt0.7016 \\
\bottomrule\bottomrule
\end{tabu}
\end{center}
\footnotesize
\renewcommand{\baselineskip}{11pt}
\textbf{Note:} 
The table shows generalization performance for different estimators on model M1 of \cite{acemoglu2001colonial}. 
We remove the $n_{\mathrm{test}}$ 
observations with the most extreme values of settler mortality, 
fit 
OLS, TSLS, PULSE and Fuller(4)
on the $64-n_{\mathrm{test}}$ samples, and compute the MSPE on the $n_{\mathrm{test}}$ held-out samples (four right-most columns). 
Indeed, in particular for larger values of $n_{\mathrm{test}}$, where more generalization is needed, PULSE and Fuller(4) outperform OLS and TSLS in the majority of cases. 
The columns ``Estimated coefficient'' show the estimates for the linear effect of average expropriation risk on log GPD of each estimation method. 
The column ``K-class $\kappa$'' shows K-class $\kappa$ parameters for both the PULSE and Fuller(4) estimates; EQ is computed  according to  $\mathrm{Var}(\mathrm{\text{out-of-sample}})=\mathrm{Var}(\mathrm{\text{in-sample}})/(1-\kappa_{\mathrm{EQ}})$.  
	\end{table}

\end{landscape}
\newpage

\newpage
\section{Weak Instruments} 
\label{sec:WeakInst}
\medskip	
There is a wide variety of attempts to quantify weakness of instruments, see e.g.\ \citet{andrews2019weak}  and  \citet{stock2002survey} for an overview.  Heuristically, the presence of weak instruments in a instrumental variable setup refers to the notion that the causal effects of the instruments onto regressors are weak relative to the noise variance of the regressors. This strength of the instruments has direct effects on the finite sample behavior of instrumental variable estimators. For simplicity consider a mean zero collapsed causal structural model with no included exogenous variables entering the equation of interest, that is,
\begin{align} \label{eq:weakInstrumentSetup}
\begin{split}
Y= \gamma^\t X + U_Y, \qquad X= \xi^\t A + U_X,
\end{split}
\end{align}
where $A\in \R^q$ are the collection of exogenous variables and the noise variables $U_X$ and $U_Y$ are possibly correlated. Let $\fA,\fX,\fY$ be a  $n$-sample data matrices of i.i.d. realizations of the system in \Cref{eq:weakInstrumentSetup}. A key statistic used to quantify weakness of instruments is the concentration matrix  given by
$ \mu_n = \Sigma_{U_X}^{-1/2} \xi^\t \fA^\t \fA \xi \Sigma_{U_X}^{-1/2}$,
where $\Sigma_{U_X}$ is the variance matrix of $U_X$. This statistic turns up in numerous different aspect of the finite sample properties of the two-stage least square estimator. 
\Citet{rothenberg1984approximating} argues
that the one-dimensional analogue of $ \mu_n$ under deterministic instruments and normal distributed noise variables directly influences the goodness of approximating a finite sample standardized two-stage least square estimator by its Gaussian asymptotic distribution. 
He argues that for large concentration parameters the Gaussian approximation is good. 
The concentration parameter can also be connected to approximate bias of the two-stage least squares estimator. Under assumptions similar to the above,  \Citet{nagar1959bias} showed that an approximate (to the order of $\mathcal{O}(n^{-1})$) finite sample bias of the two-stage least square estimator is inversely proportional to $ \mu_n$. Note that the concentration matrix $\mu_n$ is not observable, but may be approximated by
$\hat \mu_n = \hat{\Sigma}_{U_X}^{-1/2}\fX^\t P_{\fA} \fX \hat{\Sigma}_{U_X}^{-1/2}$,
where  $\hat{\Sigma}_{U_X} =  \frac{1}{n-q} \fX^\t P_{\fA}^\perp \fX$ is an estimator of the variance matrix of $U_X$ and $P_\fA \fX$ is the ordinary least square prediction of $\fA \xi$.   Now define 
\begin{align*}
G_n := \frac{\hat \mu_n}{q}=\frac{ \hat{\Sigma}_{U_X}^{-1/2}\fX^\t P_{\fA} \fX \hat{\Sigma}_{U_X}^{-1/2}}{q},
\end{align*}
which can be seen
as a multivariate
first-stage $F$-statistic for testing the hypothesis $H_0:\xi=0$. That is, when $X\in \R$, then $G_n = \frac{n-q}{q} \frac{\fX^\t P_\fA \fX }{\fX^\t \fX - \fX^\t P_\fA \fX }$  
is recognized as the F-test for testing $H_0$. \citet{stock2002testing} propose to reject the
hypothesis of a presence of weak instruments if the test-statistic $\lambda_{\min}(G_n)$, the smallest eigenvalue of $G_n$, is larger than a critical value that, for example, depends on 
how much bias you allow your estimator to have.  Prior to this $G_n$ had been  used to test under-identifiability in the sense that the concentration matrix is singular \citep[][]{cragg1993testing}, while the former uses a small minimum eigenvalue of $G_n$  as a proxy for the presence of weak instruments in identified models. From the work of \citet{staiger1997instrumental} a frequently appearing rule of thumb for instruments being non-weak is that the \textit{F}-statistic $G_n$ ($\lambda_{\min}(G_n)$ in higher dimensions) is larger than 10.  A more formal justification of this rule is due to  \citet{stock2002testing} who showed (under weak-instrument asymptotics) that it approximately (in several models) corresponds to a 5\% significance test that the bias of TSLS is at most 10\% of the bias of OLS.

We can, under further model simplification, strengthen the intuition on how the concentration matrix $G_n$ and especially the minimum eigenvalue $\lambda_{\min}(G_n)$ governs the weakness of instruments. To this end assume that $\text{Var}(U_X)= \Sigma_{
	U_X}=I$  and note that $\hat \mu_n$ is approximately proportional to the Hessian of the two-stage least squares objective function. That is, $\hat \mu_n  \approx \Sigma_{U_X}^{-1/2} \fX^\t P_{\fA} \fX \Sigma_{U_X}^{-1/2} = \fX^\t \fA ( \fA^\t \fA)^{-1} \fA^\t \fX \propto H(l_{\text{IV}}^n)$. Hence, we have that $\lambda_{\min}(G_n)$ is approximately proportional to the curvature of two-stage least squares objective function in the direction of least curvature. Thus, if $\lambda_{\min}(G_n)$ is small, 
then, heuristically, 
the objective function $l_{\text{IV}}^n$ has weak identification in the direction of the corresponding eigenvector. That is, changes to the point estimate of $\beta$ away from the two-stage least square solution in this direction does not have a strong effect on the objective value.
Finally, the weak instrument problem is a small sample problem. To this end note that 
$
n^{-1} G_n= n^{-1}\hat{\Sigma}_{U_X}^{-1/2} \fX^\t P_{\fA} \fX \hat{\Sigma}_{U_X}^{-1/2} \convp \text{Var}(U_X)^{-1/2} \xi^\t \text{Var}(A)^{-1} \xi  \text{Var}(U_X)^{-1/2}$, 
hence
by the continuity of the minimum eigenvalue operator, we have that $\lambda_{\min}(G_n) \convp \i$.

	\section{Additional Simulation Experiments} 
	\label{sec:AppFigs}
	
	\medskip
	
	\subsection{Additional Illustrations for the Univariate Experiment}

	\begin{figure}[!h] 
		\centering\includegraphics[width=\linewidth]{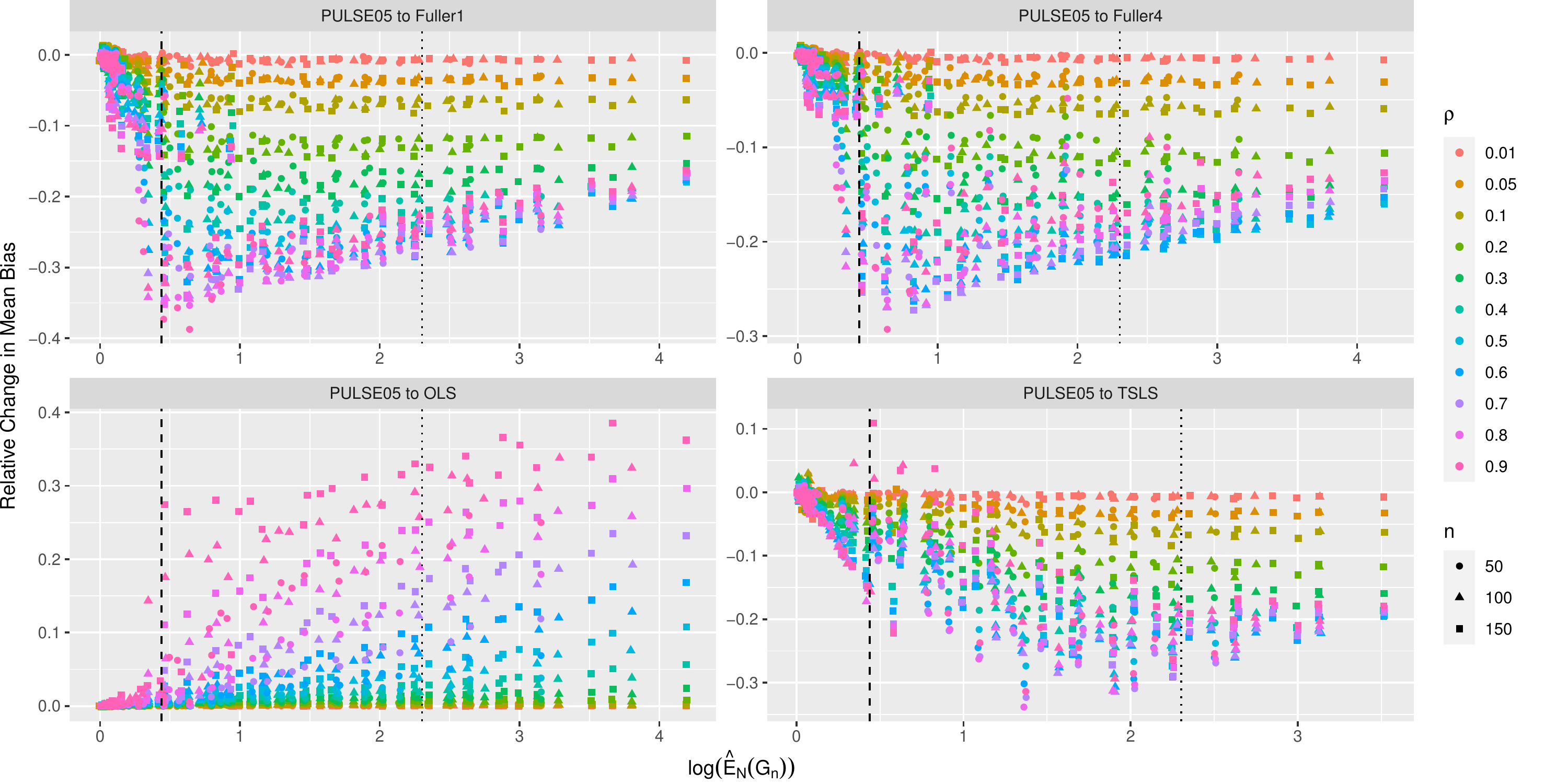}
		\caption{  \normalsize Illustrations of the relative change in the absolute value of the mean bias (a positive relative change means that PULSE is better). The vertical dotted line corresponds to the rule of thumb for classifying instruments as weak, i.e., an  F-test rejection threshold of 10. The first stage F-test for $H_0:\bar \xi=0$, i.e.,  for the relevancy of instruments, 
			at a significance level of 5\%,	
			has different rejection thresholds in the range $[1.55,4.04]$ depending on $n$ and $q$. The vertical dashed line corresponds to the smallest rejection threshold of 1.55.  Note that the lowest possible negative relative change is $-1$.  For the comparison with the TSLS estimator we have removed the case $q=1$ to ensure existence of first moments. TSLS, Fuller(1) and Fuller(4) outperforms PULSE while PULSE outperforms  OLS.  }\label{fig:HahnExpMeanBias}
	\end{figure} 
	
	\begin{figure}[H] 
		\centering\includegraphics[width=\linewidth]{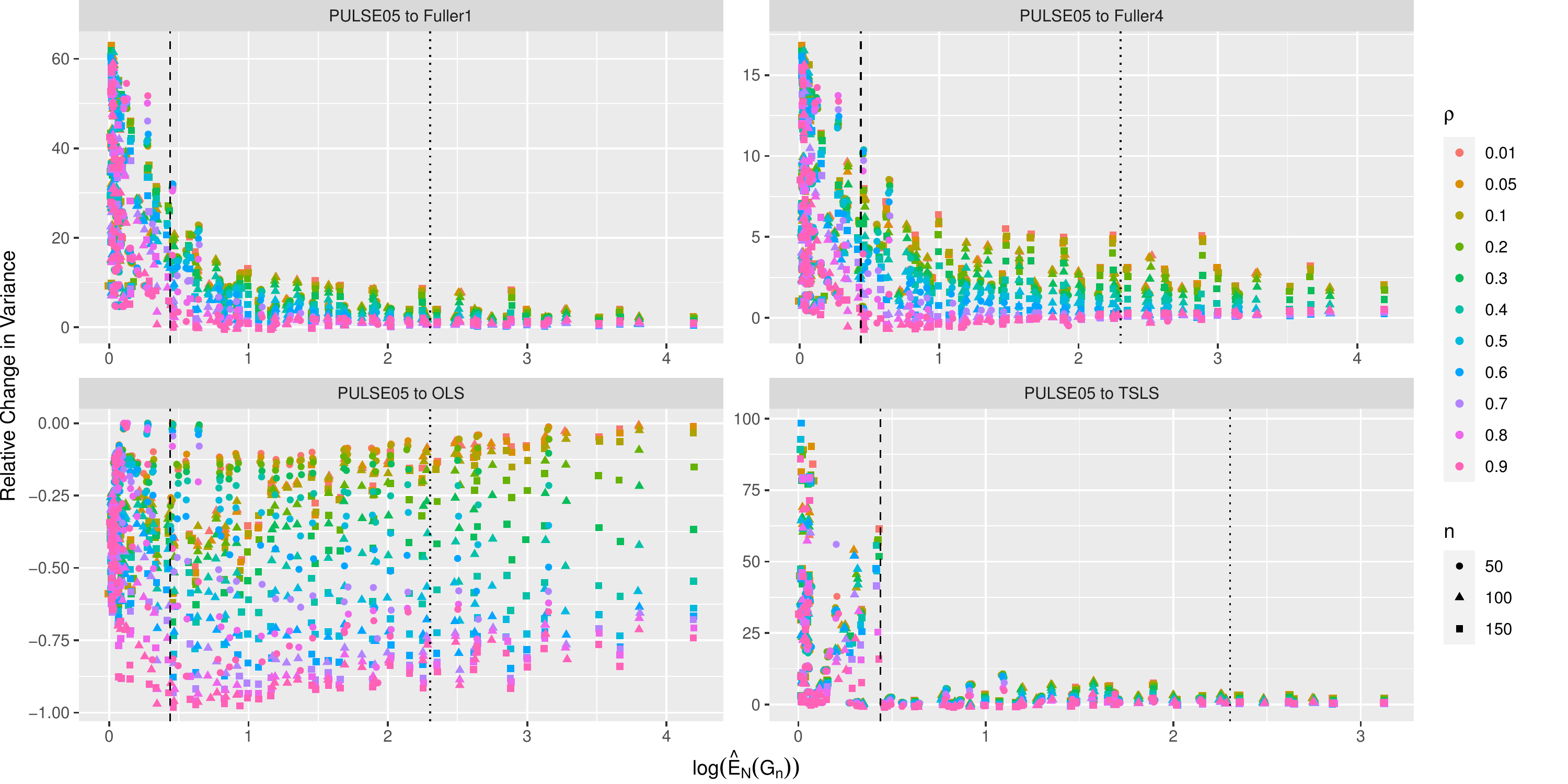}
		\caption{  \normalsize Illustrations of the relative change in variance (a positive relative change means that PULSE is better). The vertical lines are identical to those of \Cref{fig:HahnExpMeanBias}.
			For the comparison with the TSLS estimator we have removed the case $q\in \{1,2\}$ to ensure existence of second moments. We have removed two observations with relative change above 100, in the very weak instrument setting, for aesthetic reasons. PULSE outperforms TSLS, Fuller(1) and Fuller(4), especially for low confounding and weak instruments. We also see that OLS outperforms PULSE with the largest decrease in variance for the large confounding cases. }\label{fig:HahnExpVariance}
	\end{figure} 
	
	\begin{figure}[H] 
		\centering\includegraphics[width=\linewidth]{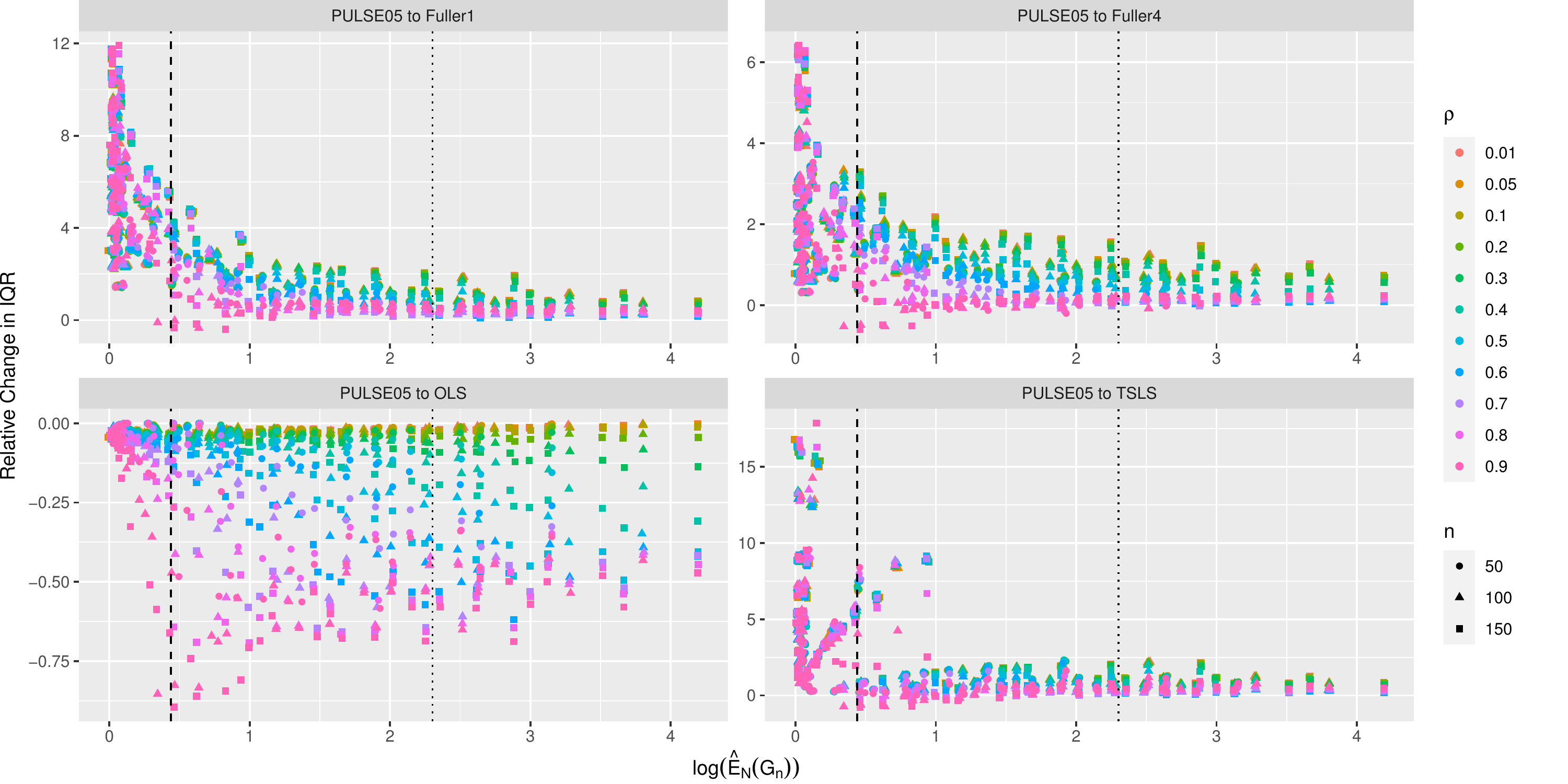}
		\caption{  \normalsize Illustrations of the relative change in interquartile range (a positive relative change means that PULSE is better). 
			The vertical lines are identical to those of \Cref{fig:HahnExpMeanBias}.
			We see that PULSE is superior to Fuller(1), Fuller(4) and TSLS except in very few cases with very large confounding. Furthermore, OLS outperforms PULSE with relatively small difference for low confounding and larger difference for large confounding. }\label{fig:HahnExpIQR}
	\end{figure} 
	\subsection{Additional Illustrations for the Multivariate Experiment}
	\begin{figure}[H] 
		\centering\includegraphics[width=\linewidth-45pt]{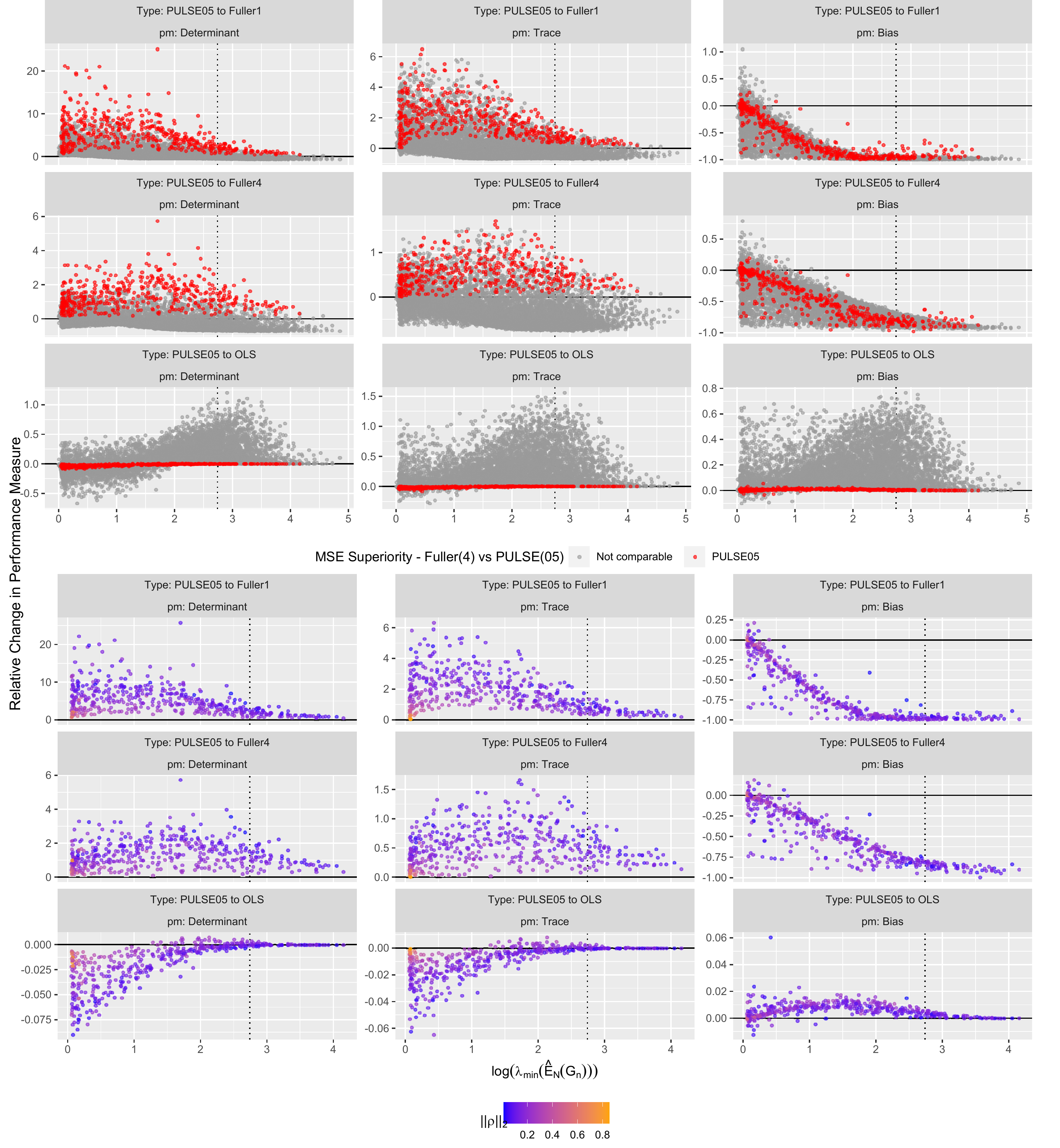}
		\caption{  \normalsize There are two illustrations, both illustrating relative changes in performance measures as in \Cref{fig:AllRandom_Beta00} except that the points are color-graded according to MSE superiority when comparing Fuller(4) and PULSE (top $3\times 3$) and confounding strength $\|\rho\|_2$ (bottom $3\times 3$) . Among the 10000 randomly generated models there are 461 models where PULSE is MSE superior to Fuller(4). In the remaining 9539 models the MSE matrices are not comparable. For the 461 models where PULSE was MSE superior the simulations were repeated with $N=25000$ repetitions to account for possible selection bias.  Of the 461 models 445 were still superior when increasing $N$ from $5000$ to $25000$. The bottom $3\times 3$ grid is an illustration of the relative change in performance measure for the 445 models that remained superior, each model color-graded according to confounding strength. We see that in almost all of these models there is weak to moderate confounding. The exception being a few models in the very weak instrument setting where the confounding is strong.}\label{fig:AllRandom_Beta00_TrueSuperior}
	\end{figure}

	\begin{figure}[H] 
		\centering\includegraphics[width=\linewidth-50pt]{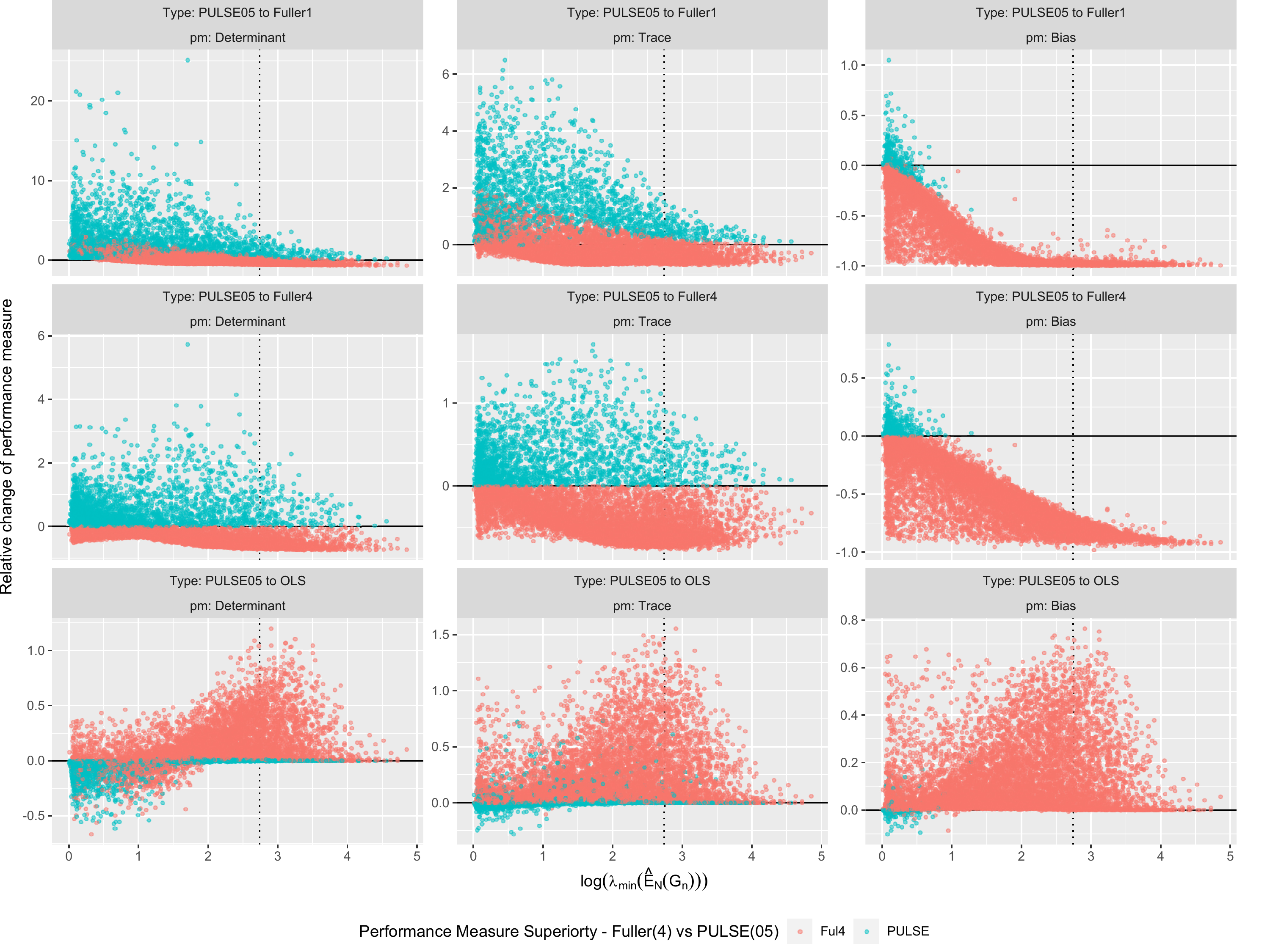}
		\caption{  \normalsize 
			This figure shows the same results as in \Cref{fig:AllRandom_Beta00} except that the points are color-graded according to performance measure superiority when comparing Fuller(4) and PULSE(05). That is, the models have fixed column-wise color-grading according to the comparison between Fuller(4) and PULSE(05).}\label{fig:AllRandom_Beta00_PmSuperior}
	\end{figure} 
\vspace{-0.5cm}
	\begin{figure}[H] 
		\centering\includegraphics[width=\linewidth-50pt]{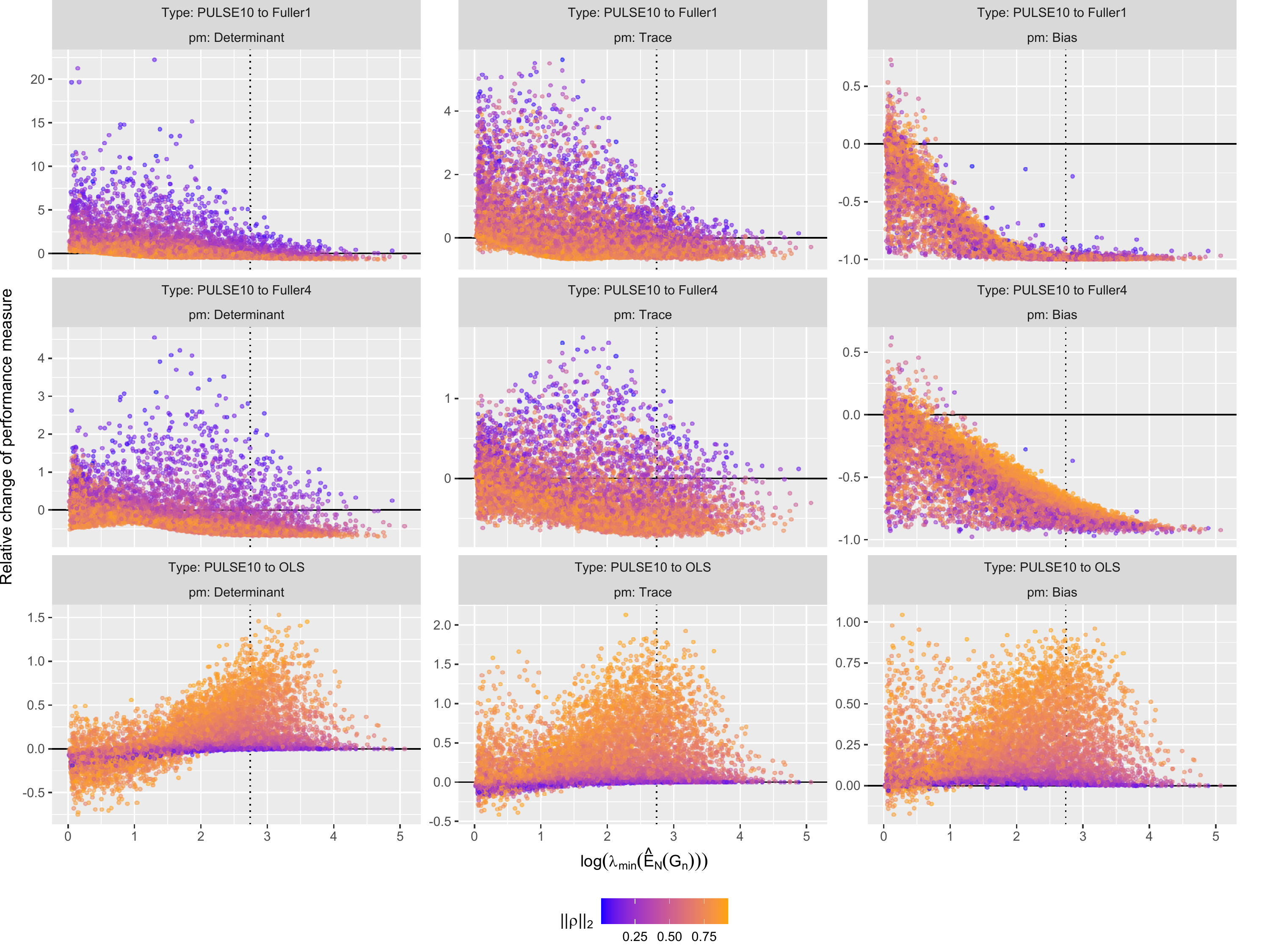}
		\caption{  \normalsize 	This figure shows the same results as in \Cref{fig:AllRandom_Beta00} except that we here compare PULSE with $p_{\min}=0.1$ to the benchmark estimators.}\label{fig:AllRandom_PULSE10}
	\end{figure}

  \chapter[A Causal Framework for Distribution Generalization]{A Causal Framework for Distribution Generalization}
  
  \vspace*{1cm}  
  
  \begin{quote}
    \begin{enumerate}
	\item[\textbf{\ref{sec:causal_relations_X}}] \nameref{sec:causal_relations_X} 
	\item[\textbf{\ref{sec:IVconditions}}] \nameref{sec:IVconditions} 
	\item[\textbf{\ref{sec:test_statistic}}] \nameref{sec:test_statistic} 
	\item[\textbf{\ref{sec:additional_experiments}}] \nameref{sec:additional_experiments} 
	\item[\textbf{\ref{app:proofs}}] \nameref{app:proofs} 
\end{enumerate}
  \end{quote}
 
\section{Transforming Causal Models}
\label{sec:causal_relations_X}

As illustrated in Remark~\ref{rem:model}, our framework can
also be applied 
in situations where training and test distributions are generated from 
an SCM with a different structure than~\eqref{eq:SCMmodelreduced}.
Below, we show that
a general class of SCMs can be transformed into our reduced
setting. 
To this end, assume the true underlying causal
structure is given by the SCM
\begin{align}\label{eq:SCM_full_causal}
	\begin{split}
		A\coloneqq \ep_A\qquad\qquad
		&X\coloneqq w(X, Y) + g(A) + h_2(H, \ep_X)\\
		H\coloneqq \ep_H\qquad\qquad
		&Y\coloneqq f(X) + h_1(H, \ep_Y),
	\end{split}
\end{align}
where, as before, $f,g,w,h_1$ and $h_2$ are measurable functions. 
First, we show how to transform the above SCM into the reduced form \eqref{eq:SCMmodelreduced}
without changing the induced observational distribution. In Appendix~\ref{sec:int_type}, we then 
discuss how to transform interventions in \eqref{eq:SCM_full_causal} to interventions in the reduced 
model.

Throughout this appendix, we assume that~\eqref{eq:SCM_full_causal}
is uniquely solvable in the sense that
there exists a unique function $F$ such that
$(A, H, X, Y)=F(\ep_A,\ep_H,\ep_X,\ep_Y)$ almost surely, see \cite{bongers2021foundations}
for more details.  
Denote by $F_X$ the coordinates of $F$
that correspond to the $X$ variable (i.e., the coordinates from
$r+q+1$ to $r+q+d$). 
We further assume that
there exist functions $\tilde{g}$ and $\tilde{h}_2$ such that
\begin{equation}
	\label{eq:linear_decomposition_inverse}
	F_X(\ep_A,\ep_H,\ep_X,\ep_Y)=\tilde{g}(\ep_A) + \tilde{h}_2((\ep_H, \ep_Y), \ep_X).
\end{equation}
This decomposition 
is not always possible, but 
it exists in the following settings, for example:
(i) \emph{There are no $A$ variables.} 
In these
cases, the additive decomposition
\eqref{eq:linear_decomposition_inverse} becomes trivial.
(ii) \emph{There are further constraints on the original SCM.}
The additive
decomposition \eqref{eq:linear_decomposition_inverse} holds if,
for example, 
$w$ is a linear function or $A$ only enters the
structural assignments of covariates $X$ which have at most $Y$ as
a descendant.

Using the decomposition in \eqref{eq:linear_decomposition_inverse}, we can define the following reduced SCM
\begin{align}\label{eq:SCM_full_causal_simple}
	\begin{split}
		A\coloneqq \ep_A\qquad\qquad
		&X\coloneqq \tilde{g}(A) + \tilde{h}_2(\tilde{H}, \ep_X)\\
		\tilde{H}\coloneqq \ep_{\tilde{H}}\qquad\qquad
		&Y\coloneqq f(X) + h_1(\tilde{H}),
	\end{split}
\end{align}
where $\ep_{\tilde{H}}$ has the same distribution as $(\ep_H, \ep_Y)$
in \eqref{eq:SCM_full_causal}. This model fits the framework from 
Section~\ref{sec:modeling_int_ind_distr}, where the noise term in $Y$ is now taken to
be constantly zero. 
Both SCMs~\eqref{eq:SCM_full_causal}
and~\eqref{eq:SCM_full_causal_simple} induce the same observational
distribution and 
the same function $f$ appears in the assignments of $Y$.

If 
one intends to use 
interventions 
in the original SCM
(i.e., \eqref{eq:SCM_full_causal}) to
model  
the test distributions, one needs  
to also transform these
interventions. 
We discuss how this can be done in the following
subsection. 

\subsection{Transforming Interventions}\label{sec:int_type}
For SCMs of the form \eqref{eq:SCM_full_causal} (and which satisfy \eqref{eq:linear_decomposition_inverse}), 
any distribution arising from an intervention on a 
subset of covariates from $X$ can be equivalently expressed using an intervention on all of $X$
in the corresponding reduced model 
\eqref{eq:SCM_full_causal_simple}. To see this, let $i$ be such an intervention 
in the original SCM, 
and let $\P^i$ be the induced interventional 
distribution over $(X,Y,A)$. We can then generate the same intervention distribution
in 
\eqref{eq:SCM_full_causal_simple} 
using the intervention $X:= \epsilon_X^i$, where the distribution of
$\epsilon_X^i$ coincides with the marginal of $X$ in $\P^i$.  Note,
however, that this type of transformation may fail for some
model classes,
for example,
this may happen  
if the original SCM contains a hidden variable which is a descendant of
some (intervened) $X$ variables and a cause of $Y$.  
Also, even in situations where the above
transform is possible,
the interventions can change
their intervention targets, become non-well-behaved or change their
support. In order to apply the developed methodology, one needs to
check whether the transformed interventions are a well-behaved (this
is not necessarily the case, even if the original intervention was
well-behaved) and how the support of all $X$ variables behaves under
that specific intervention.

\textbf{Intervention type}\quad First, we consider which types of
interventions in~\eqref{eq:SCM_full_causal} translate to well-behaved
interventions in~\eqref{eq:SCM_full_causal_simple}. A simple example
is given by interventions on $A$ in the original SCM, which result in the
same interventions on $A$ also in the reduced SCM. Similarly,
performing hard interventions on all components of $X$ in the original SCM
leads to the same intervention in the reduced SCM, which is in
particular both confounding-removing and confounding-preserving. For
interventions on subsets of the $X$, this is not always the case. To
see that, consider the following example 
\begin{center}
	\begin{minipage}{0.40\columnwidth}
		\centering
		{
			\begin{align*}
				\begin{split}
					A&\coloneqq \ep_A\\
					X_1 &\coloneqq \ep_1\\
					X_2 &\coloneqq Y + \epsilon_2\\
					Y &\coloneqq X_1 + \epsilon_Y
				\end{split}
		\end{align*}}
	\end{minipage}%
	\begin{minipage}{0.18\columnwidth}
		{\Large
			\begin{equation*}
				\xrightarrow{\text{transform}} \, \,
		\end{equation*}}
		\qquad
		$ $
	\end{minipage}%
	\begin{minipage}{0.40\columnwidth}
		\centering
		{
			\begin{align*}
				\begin{split}
					A &\coloneqq \ep_A\\
					H &\coloneqq \ep_Y \\
					X &\coloneqq (\ep_1,  H + \ep_1 + \ep_2)\\
					Y &\coloneqq X_1 + H,
				\end{split}
		\end{align*}}
	\end{minipage}%
\end{center}
with
$\ep_A, \epsilon_1, \epsilon_2, \ep_Y$
i.i.d.\ noise innovations.
Here, the left hand
side represents the original SCM and the right hand side
corresponds to the reduced SCM fitting in our framework. Consider now, in
the original SCM, the intervention $X_1 \coloneqq i$, for some
$i\in\R$. In the reduced SCM, this intervention corresponds to the
intervention $X = (X_1, X_2) \coloneqq (i, H + i + \ep_2)$, which is
neither confounding-preserving nor confounding-removing.\footnote{
	This may not come as a surprise since, without the help of an
	instrument, it is impossible to distinguish whether a covariate is
	an ancestor or a descendant of $Y$.  } On the other hand, any
intervention on $X_2$ or $A$ in the original SCM model corresponds to the
same intervention in the reduced SCM. We can generalize these
observations to the following statements 
\begin{itemize}
	\item \emph{Interventions on $A$:} If we intervene on $A$ in the original
	SCM \eqref{eq:SCM_full_causal} (i.e., by replacing the structural
	assignment of $A$ with $\psi^i(I^i, \ep_A^i)$), then this translates
	to the same intervention on $A$ in the reduced
	SCM \eqref{eq:SCM_full_causal_simple}.
	\item \emph{Shift intervention on
		$X_j$ which are not ancestors of
		$Y$:} If we perform a shift intervention on
	$X_j$ in the original SCM \eqref{eq:SCM_full_causal} (assuming no
	confounding $H$) and $X_j$ is not an ancestor of
	$Y$, then this corresponds to a confounding-preserving intervention
	in the reduced SCM \eqref{eq:SCM_full_causal_simple}.
	\item \emph{Hard interventions on all $X$:} If we intervene on all $X$
	in the original SCM \eqref{eq:SCM_full_causal} by replacing the
	structural assignment of $X$ with an independent random variable
	$I\in\R^d$, then this translates to the same intervention in the
	reduced SCM \eqref{eq:SCM_full_causal_simple} which is
	confounding-removing.
	\item \emph{No $X$ is a descendant of $Y$ and there is no unobserved
		confounding $H$:} If we intervene on $X$ in the original SCM
	\eqref{eq:SCM_full_causal} (i.e., by replacing the structural
	assignment of $X$ with $\psi^i(g, A^i, \ep^i_X ,I^i)$), then this
	translates to a potentially different but confounding-removing
	intervention in the reduced SCM
	\eqref{eq:SCM_full_causal_simple}. This is because the reduced SCM
	\eqref{eq:SCM_full_causal_simple} does not include unobserved
	variables $H$ in this case.
	\item \emph{Hard interventions on a variable $X_j$ which has at most $Y$
		as a descendant:} 
	If we intervene on $X_j$ in the original SCM \eqref{eq:SCM_full_causal}
	by replacing the structural assignment of $X_j$ with an independent
	random variable $I$, then this intervention translates to a
	potentially different but confounding-preserving intervention.
\end{itemize}
Other settings may yield well-behaved interventions, too, but may
require more assumptions on the original SCM model
\eqref{eq:SCM_full_causal} or further restrictions on the intervention
classes.

\textbf{Intervention support}\quad 
A support-reducing intervention in the original SCM can translate to a
support-extending intervention in the reduced SCM. Consider the
following example
\begin{center}
	\begin{minipage}{0.40\columnwidth}
		\centering
		{
			\begin{align*}
				\begin{split}
					X_1 &\coloneqq \ep_1 \\
					X_2 &\coloneqq X_1 + \mathbf{1}\{X_1 = 0.5\}\\
					Y &\coloneqq X_2 + \epsilon_Y
				\end{split}
		\end{align*}}
	\end{minipage}%
	\begin{minipage}{0.18\columnwidth}
		{\Large
			\begin{equation*}
				\xrightarrow{\text{transform}}  \, \,
		\end{equation*}}
		\qquad 
		$ $
	\end{minipage}%
	\begin{minipage}{0.40\columnwidth}
		\centering
		{
			\begin{align*}
				\begin{split}
					X &\coloneqq (\ep_1, \ep_1 + \mathbf{1}\{\ep_1 = 0.5\})\\
					Y &\coloneqq X_2 + \ep_Y,
				\end{split}
		\end{align*}}
	\end{minipage}%
\end{center}
with
$\ep_1, \ep_Y\overset{i.i.d.}{\sim} \mathcal{U}(0,1)$.  As before,
the left hand side
represents the
original SCM, whereas 
the right hand
side
corresponds to the reduced
SCM converted to fit our framework.  Under the observational
distribution, the support of $X_1$ and $X_2$ is equal to the open
interval $(0, 1)$.  Consider now the support-reducing intervention
$X_1:= 0.5$ in original SCM.
Within our framework, such an intervention would correspond to the
intervention $X = (X_1, X_2) := (0.5, 1.5)$, which is
support-extending. This example is rather special in that the SCM
consists of a function that changes on a null set of the observational
distribution. With appropriate assumptions to exclude similar
degenerate cases, it is possible to show that support-reducing
interventions in \eqref{eq:SCM_full_causal} correspond to
support-reducing interventions within our
framework~\eqref{eq:SCM_full_causal_simple}.

\section{Sufficient Conditions for Assumption~1 in IV Settings}
\label{sec:IVconditions}

Assumption~\ref{ass:identify_f} states that $f$ is identified on the
support of $X$ from the observational distribution of $(Y,X,A)$. 
Whether this assumption is satisfied
depends on the structure of
$\cF$ but also on the other function
classes $\cG,\cH_1,\cH_2$ and $\mathcal{Q}$ that
make up the model class $\cM$ from which we assume that the
distribution of $(Y,X,A)$ is generated.

Identifiability of the causal
function in the presence of 
instrumental variables
is a well-studied problem in econometrics
literature. Most prominent is the literature on identification in
linear SCMs
\cite[e.g.,][]{fisher1966identification,greene2003econometric}.
However, identification 
has also been studied for various other
parametric function classes. 
We say that $\mathcal{F}$ is a parametric
function class if it can be parametrized by some finite dimensional
parameter set $\Theta \subseteq \R^p$. 
We here consider classes of the form
\begin{align*}
	\mathcal{F} := \{ f(\cdot ,\theta):\R^d \to \R\,\vert\, \Theta \ni \theta \mapsto f(x,\theta) \in C^2, \forall  x\in \R^d \}.
\end{align*}
Consistent estimation of the 
parameter
$\theta_0$ using instrumental variables in such function classes has
been studied extensively in the econometric literature
\cite[e.g.,][]{amemiya1974nonlinear,jorgenson1974efficient,kelejian1971two}. %
These works also contain rigorous results on how instrumental variable
estimators of $\theta_0$ are constructed and under which conditions
consistency (and thus identifiability) holds. Here, we give an
argument on why the presence of the exogenous variables $A$ yields identifiability under certain
regularity conditions. Assume that $\E[h_1(H, \epsilon_Y)|A]=0$, which implies
that the true causal function $f(\cdot,\theta_0)$ satisfies the
population orthogonality condition
\begin{align} \label{Eq:PopOrthCondNonLinearIV}
	\E[l(A)^\top (Y-f(X,\theta_0))] = \E\big[l(A)^\top \E[h_1(H, \epsilon_Y)|A]\big]= 0,
\end{align}
for some measurable mapping $l:\R^q\to \R^g$,
for some $ g \in \mathbb{N}_{>0}$.
Clearly, $\theta_0$ is identified from the
observational distribution if the map
$\theta \mapsto \E[l(A)^\top (Y-f(X,\theta))]$ is zero if and only
if $\theta=\theta_0$.   Furthermore, since
$\theta\mapsto f(x,\theta)$ is differentiable for all $x\in \R^d$,
the
mean value theorem yields that, for any $\theta\in \Theta$ and
$x\in \R^d$,
there exists an intermediate point
$\tilde{\theta}(x,\theta,\theta_0)$ on the line segment between
$\theta$ and $\theta_0$ such that
\begin{equation*}
	f(x,\theta) - f(x,\theta_0) = D_\theta f(x,\tilde{\theta}(x,\theta,\theta_0))(\theta-\theta_0),
\end{equation*}
where, 
for each
$x\in \R^d$,
$D_\theta f(x,\theta)\in\R^{1\times p}$ is the derivative of
$\theta\mapsto f(x,\theta)$ evaluated in $\theta$.  Composing the above expression with the random vector
$X$, multiplying with $l(A)$ and taking expectations yields that
\begin{align*}
	&\E[l(A)(Y-f(X,\theta_0))] - \E[l(A)(Y-f(X,\theta))] 
	\\
	&=  \E[l(A)D_\theta f(X,\tilde{\theta}(X,\theta,\theta_0))](\theta_0-\theta).
\end{align*}
Hence, if
$ \E[l(A)D_\theta f(X,\tilde{\theta}(X,\theta,\theta_0))]\in
\R^{g\times p}$ is of rank $p$ for all $\theta\in\Theta$
(which implies $g \geq p$), 
then
$\theta_0$ is identifiable as it is the only parameter that satisfies
the population orthogonality condition of
\eqref{Eq:PopOrthCondNonLinearIV}.  As $\theta_0$ uniquely determines
the entire function, we get identifiability of
$f\equiv f(\cdot,\theta_0)$, not only on the support of $X$ but the
entire domain $\R^d$, i.e., both
Assumptions~\ref{ass:identify_f} and~\ref{ass:gen_f} are satisfied.
In the case that $\theta \mapsto f(x,\theta)$ is linear, i.e.\
$f(x,\theta) = f(x)^T \theta$ for all $x\in \R^d$, the above rank
condition reduces to $\E[l(A)f(X)^T]\in \R^{g\times p}$ having rank $p$
(again, implying that $g \geq p$). Furthermore, when
$(x,\theta)\mapsto f(x,\theta)$ is bilinear, a reparametrization of
the parameter space ensures that $f(x,\theta)= x^T \theta$ for
$\theta\in\Theta \subseteq \R^d$. In this case, the rank condition can
be reduced to the well-known rank condition for identification in a
linear SCM, namely that
$\E[AX^T] \in \R^{q\times p}$ is of rank $p$.

Finally, identifiability and methods of consistent estimation of the causal function have also been studied for non-parametric function classes. The conditions for identification are rather technical, however, and we refer the reader to  \cite{newey2013nonparametric,newey2003instrumental} for further details.

\section{Choice of Test Statistic} 
\label{sec:test_statistic}
By considering the variables $$B(X) = (B_1(X), \dots, B_k(X)) \text{ and }C(A) = (C_1(A), \dots, C_k(A)),$$
as vectors of covariates and instruments, respectively, our setting in Section~\ref{sec:nile} reduces to the classical
(just-identified) linear IV setting. 
We could therefore 
use a test statistics similar to the one propsed by the PULSE \citep[][]{jakobsen2020distributional}. 
With a notation that is slightly adapted to our setting, this estimator tests $\tilde{H}_0(\theta)$
using the test statistic
\begin{align*}
	T^1_n(\theta) = c(n) \frac{\norm{\B{P} (\B{Y} - \B{B}\theta)}_2^2}{\norm{\B{Y} - \B{B}\theta}_2^2},
\end{align*}
where $\B{P}$ is the projection onto the columns of $\B{C}$, and $c(n)$ is some function with $c(n) \sim n$ as $n\to\infty$. 
Under the null hypothesis, $T^1_n$ converges in distribution to the $\chi^2_{k}$ distribution, and diverges to infinity in 
probability under the general alternative. Using this test statistic, $\tilde{H}_0(\theta)$ is rejected if and only if $T^1_n(\theta)> q(\alpha)$, 
where $q(\alpha)$ is the $(1-\alpha)$-quantile of the $\chi^2_{k}$ distribution. The acceptance region of this test statistic 
is asymptotically equivalent with the confidence region of the Anderson-Rubin test \cite{anderson1949estimation} for the 
causal parameter $\theta^0$. Using the above test results in a consistent estimator for $\theta^0$ 
\citep[][Theorem~3.12]{jakobsen2020distributional}; the proof exploits the particular form of $T^1_n$ without 
explicitly imposing that assumptions~\ref{ass:ConsistentTestStatistic} and \ref{ass:LambdaStarAlmostSurelyFinite} hold.

If the number $k$ of basis functions is large, however, numerical experiments suggest that the above 
test has low power in finite sample settings. As default, our algorithm  therefore 
uses a different test based on a penalized regression approach. 
This test has been proposed in \cite{chen2014note}
for inference in nonparametric regression models. 
We now introduce this procedure
with a notation that is adapted to our setting. For every $\theta \in \R^k$, let 
$R_\theta = Y - B(X)^\top \theta$ be the residual associated with $\theta$. 
We then test the slightly stronger hypothesis
\begin{equation*}
	\bar{H}_0(\theta): \exists \,  \sigma_\theta^2>0 \text{ s.t. }  \E[R_\theta \given A] \eqas 0  \text{ and } \text{Var}[R_\theta \given A] = \sigma_\theta^2
\end{equation*}
against the alternative that $\E[R_\theta \given A] = m(A)$ for some
smooth function $m$.  To see that the above hypothesis implies
$\tilde{H}_0(\theta)$ (and therefore $H_0(\theta)$, see
Section~\ref{sec:estimation}),
let $\theta \in \R^k$ be such that
$\bar{H}_0(\theta)$ holds true.  Then,
\begin{align*}
	\E[C(A)(Y - B(X)^\top \theta)] 
	= \E[C(A) R_\theta] = \E[\E[C(A) R_\theta \given A]] 
	= \E[C(A) \E[R_\theta \given A]] = 0,
\end{align*}
showing that also $\tilde{H}_0(\theta)$ holds true. 
Thus, if $\tilde{H}_0(\theta)$ is false, then also $\bar H_0(\theta)$ is false.
As a test statistic $T^2_n(\theta)$ for $\bar{H}_0(\theta)$, we use (up to a normalization) 
the squared
norm of a penalized regression estimate of $m$, evaluated at the data $\B{A}$, i.e., 
the TSLS loss $\norm{\B{P}_\delta (\B{Y} - \B{B}\theta)}_2^2$.
In the fixed design case, where $\B{A}$ is non-random, it has been shown that,
under $\bar{H}_0(\theta)$ and certain additional regularity conditions, it holds that 
\begin{align*}
	\frac{\norm{\B{P}_\delta (\B{Y} - \B{B}\theta)}_2^2 - \sigma_\theta^2 c_n}{\sigma_\theta^2 d_n} \stackrel{\text{d}}{\longrightarrow} \mathcal{N}(0,1),
\end{align*}
where $c_n$ and $d_n$ are known functions of $\B{C}$, $\B{M}$ and $\delta$ \citep[][Theorem~1]{chen2014note}.
The authors further state that the above convergence is unaffected by exchanging $\sigma_\theta^2$
with a consistent estimator $\hat{\sigma}_\theta^2$, which motivates our use of the test statistic
\begin{equation*}
	T^2_n(\theta) := \frac{\norm{\B{P}_\delta (\B{Y} - \B{B}\theta)}_2^2 - \hat \sigma_{\theta,n}^2 c_n}{\hat \sigma_{\theta,n}^2 d_n},
\end{equation*}
where $\hat \sigma_{\theta,n}^2 := \frac{1}{n-1} \sum_{i=1}^n \norm{(\B{I}_n - \B{P}_\delta)(\B{Y} - \B{B} \theta)}_2^2$. 
As a
rejection threshold $q(\alpha)$ we use the $1-\alpha$ quantile of a standard normal distribution. 
For results on the asymptotic power of the test defined by $T^2$, we refer to Section~2.3 in \cite{chen2014note}. 

In our software package, both of the above tests are available options.

\section{Addition to Experiments} 
\label{sec:additional_experiments}

\subsection{Sampling of the Causal Function} \label{sec:exp_sampling}
\begin{figure}
	\centering
	\includegraphics[width=\linewidth]{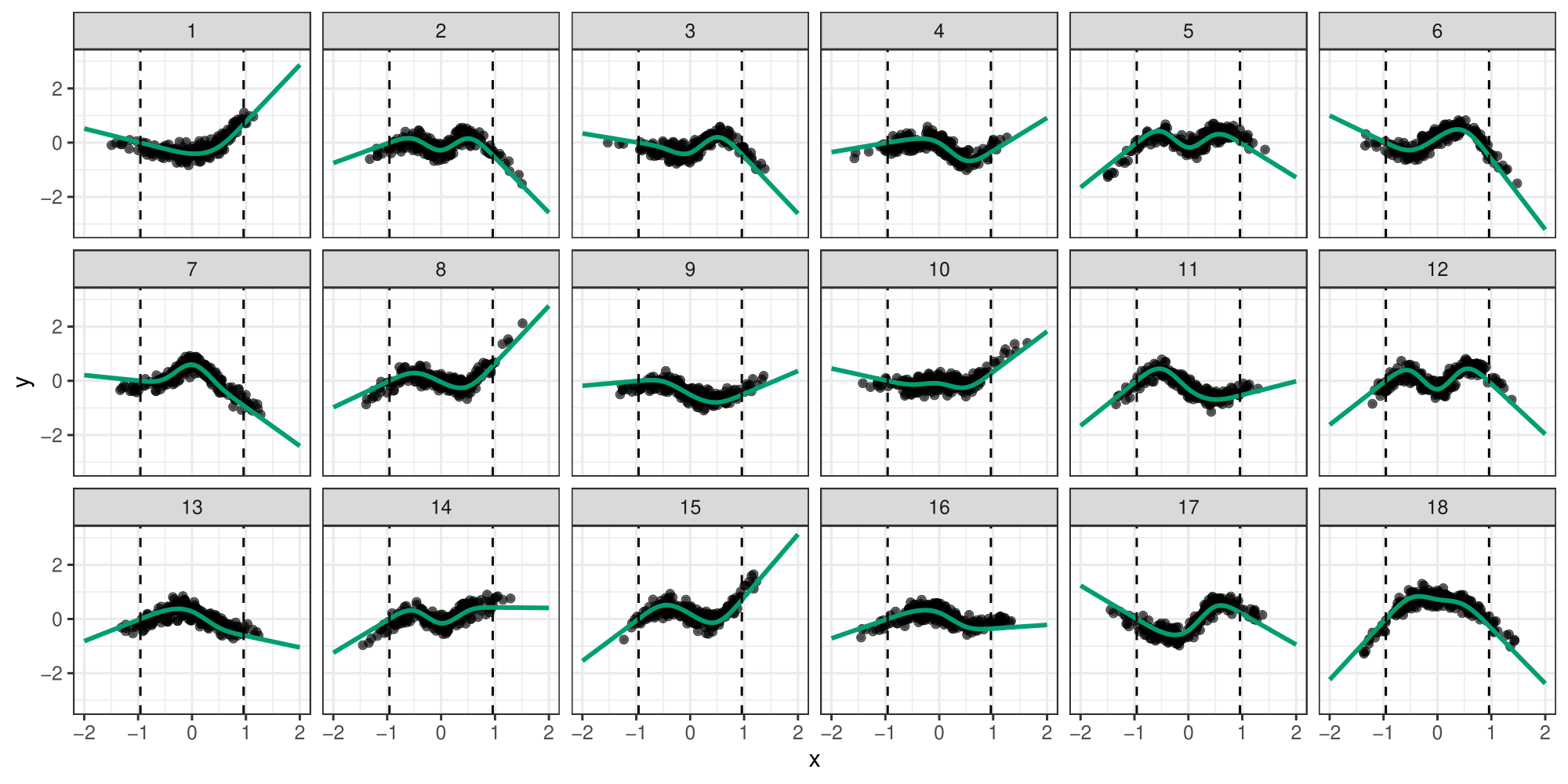}
	\caption{
		The plots show independent realizations of the causal function that
		is used in all our experiments. 
		These are sampled from a 
		linear space of natural cubic splines, as described in
		Appendix~\ref{sec:exp_sampling}. To ensure a fair comparison
		with the alternative method, NPREGIV, the true causal 
		function is chosen from a model class different from 
		the one assumed by the NILE.}
	\label{fig:f_samples}
\end{figure}

To ensure linear extrapolation of the causal function, we have chosen a function
class consisting of natural cubic splines, which, by construction, extrapolate linearly
outside the boundary knots. We now describe in detail how we sample 
functions from this class for the experiments in Section~\ref{sec:experiments}.
Let $q_{\min}$ and $q_{\max}$ be the respective $5\%$- and $95\%$ quantiles of $X$, 
and let $B_1, \dots, B_4$ be a basis of natural cubic splines corresponding to 5 knots 
placed equidistantly between $q_{\min}$ and $q_{\max}$. We then sample coefficients 
$\beta_i \iid \text{Uniform}(-1,1)$, $i = 1, \dots, 4$, and construct $f$ as 
$f = \sum_{i=1}^4 \beta_i B_i$. For illustration, we have included 18 
realizations in Figure~\ref{fig:f_samples}. 

\subsection{Violations of the Linear Extrapolation Assumption} \label{sec:exp_violation_lin_extrap}
\begin{figure}[t]
	\centering
	\includegraphics[width=\linewidth]{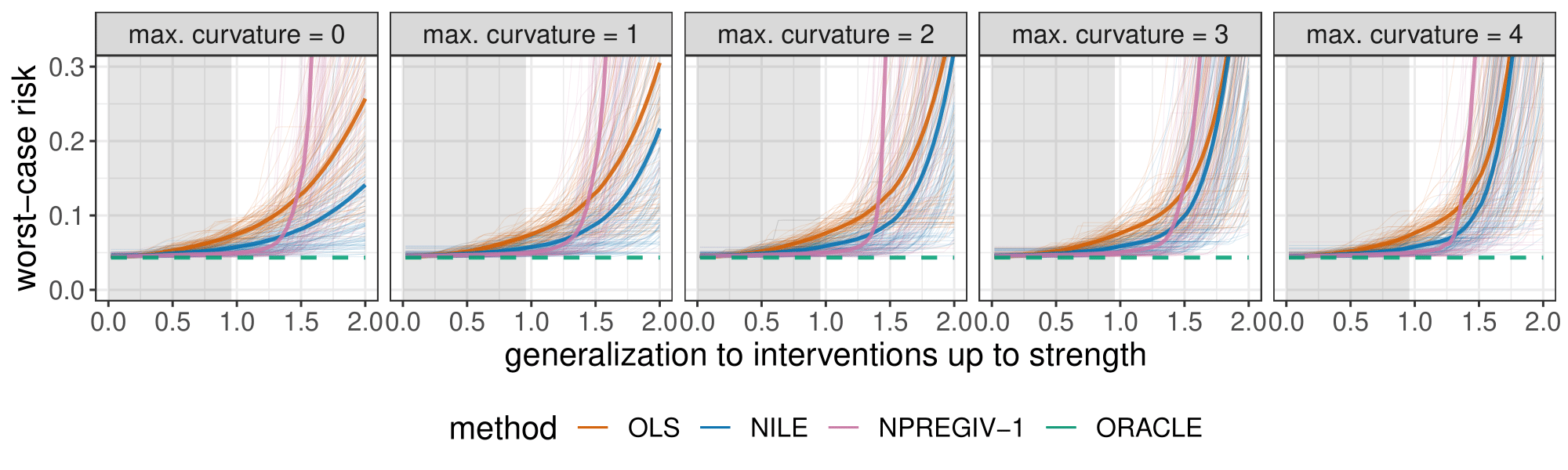}
	\caption{Worst-case risk for increasingly strong violations of the linear extrapolation assumption. 
		The grey area marks the inner 90 \% quantile range of $X$ in the training distribution.
		As the 
		curvature of $f$
		outside the domain of the observed data increases, 
		it becomes 
		difficult 
		to predict the interventional behavior of $Y$ for strong interventions. However, even in situations
		where the linear extrapolation assumption is strongly violated,  it remains beneficial to
		extrapolate linearly. %
	}
	\label{fig:violation_lin_extrap}
\end{figure}

We have assumed that the true causal function extrapolates linearly 
outside the 90\% quantile range of $X$. We now investigate the performance
of our method for violations of this assumption. To do so, we again sample from 
the model \eqref{eq:sim_model}, with $\alpha_A = \alpha_H = \alpha_\epsilon = 1/\sqrt{3}$. 
For each data set, the causal function is sampled as follows. Let $q_{\min}$ and $q_{\max}$
be the $5\%$- and $95\%$ quantiles of $X$. We first generate a function $\tilde{f}$
that linearly extrapolates outside $[q_{\min}, q_{\max}]$ as described in 
Section~\ref{sec:exp_sampling}. For a given threshold $\kappa$, we then draw 
$k_1, k_2 \iid \text{Uniform}(-\kappa, \kappa)$ and construct $f$ for every $x \in \R$ by
\begin{equation*}
	f(x) = \tilde{f}(x) + \frac{1}{2} k_1 ((x-q_{\min})_{-})^2 + \frac{1}{2} k_2 ((x-q_{\max})_{+})^2,
\end{equation*}
such that the curvature of $f$ on $(-\infty, q_{\min}]$ and $[q_{\max}, \infty)$ is $k_1$ and $k_2$, respectively. 
Figure~\ref{fig:violation_lin_extrap} shows results for $\kappa = 0,1,2,3,4$. As the curvature increases, 
the ability to generalize decreases. 

\subsection{Running NILE on Half of the Available Data}
\begin{figure}[t]
	\centering
	\includegraphics[width=.7\columnwidth]{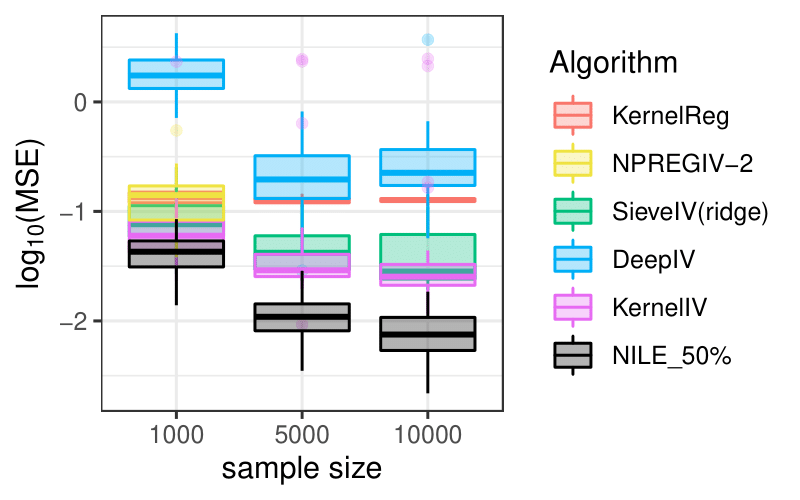}
	\caption{Same results as shown in Figure~\ref{fig:kernelIV}, except that here, NILE is run only on 
		half of the available data.}
	\label{fig:nile50_vs_kernelIV}
\end{figure}
In Section~\ref{sec:experiments}, we compared the NILE to several alternative procedures for 
estimating a non-linear causal function. As mentioned, these procedure use a sample-splitting
strategy, where the two steps of the the two-stage-least-squares procedure are run on 
disjoint data sets. The NILE, on the other hand, uses all of the available data for the model fitting. 
Figure~\ref{fig:nile50_vs_kernelIV} shows that, even when using only half of the available data, 
the NILE still outperforms the other methods considerably. 

\section{Proofs}\label{app:proofs}


\begin{proofenv}{\Cref{prop:minimax_equal_causal}}
	Assume that $\cI$ is a set of interventions on $X$ with at least one
	confounding-removing intervention. Let $i \in \II$ and
	$\fs \in \FF$, then we have the following expansion
	\begin{align} \label{eq:decomp_equation}
		\E_{M(i)}[(Y - \fs(X))^2] &= 
		\E_{M(i)}[(f(X) - \fs(X))^2]
		+\E_{M(i)}[\xi_Y^2]  
		\\ & \qquad +2\E_{M(i)}[\xi_Y(f(X)-\fs(X))], \notag 
	\end{align}
	where $\xi_Y=h_1(H, \epsilon_Y)$. For any intervention $i\in \cI$
	the causal function $f$ always yields an identical loss. In particular,
	it holds that
	\begin{align} \label{eq:SupLossCausalFunction}
		\sup_{i\in\II} \E_{M(i)}[(Y - f(X))^2] =\sup_{i\in\II} \E_{M(i)}[\xi_Y^2] = \E_{M}[\xi_Y^2],
	\end{align}
	where we used that the distribution of $\xi_Y$ is not affected by an
	intervention on $X$. The loss of the causal function can
	never be better than the minimax loss, that is,
	\begin{align} \label{eq:lowerbdd_prop1}
		\inf_{\fs\in\FF}\sup_{i\in\II}\E_{M(i)}[(Y - \fs(X))^2]
		\leq\sup_{i\in\II}\E_{M(i)}[(Y - f(X))^2]   =\E_{M}[\xi_Y^2].
	\end{align}
	In other words, the minimax solution (if it exists) is always better
	than or equal to the causal function. We will now show that when
	$\cI$ contains at least one confounding-removing intervention, then
	the minimax loss is dominated by any such intervention.
	
	Fix $i_0\in \cI$ to be a confounding-removing intervention and let
	$(X,Y,H,A)$ be generated by the SCM $M(i_0)$.  Recall that there
	exists a map $\psi^{i_0}$ such that
	$X:= \psi^{i_0}(g, h_2, A, H, \ep_X ,I^{i_0})$ 
	and that
	$X\independent H$ as $i_0$ is a confounding-removing intervention.
	Furthermore, since 
	the vectors $A$, $H$, $\ep_X$, $\ep_Y$ and
	$I^{i_0}$ are mutually independent, we have that
	$(X,H)\independent \ep_Y$ which together with $X\independent H$
	implies $X, H$ and $\ep_Y$ are mutually independent, and hence
	$X \independent h_1(H,\ep_Y)$. Using this independence we get that
	$\E_{M(i_0)}[\xi_Y(f(X)-\fs(X))]=\E_{M}[\xi_Y]\E_{M(i_0)}[(f(X)-\fs(X))]$. Hence,
	\eqref{eq:decomp_equation} for the intervention $i_0$ together with
	the modeling assumption $\E_{M}[\xi_Y]=0$ implies 
	that
	for all
	$\fs \in \FF$, 
	\begin{align*}
		\E_{M}[\xi_Y^2] &\leq \E_{M(i_0)}[(f(X) - \fs(X))^2]+\E_{M}[\xi_Y^2] =  \E_{M(i_0)}[(Y - \fs(X))^2] .
	\end{align*}
	This proves that the smallest loss at a confounding-removing
	intervention is achieved by the causal function. 
	Denoting the non-empty subset of confounding-removing interventions by $\cI_{\text{cr}}\subseteq \cI$, this implies
	\begin{align} 
		\E_{M}[\xi_Y^2] &=  \inf_{\fs\in\FF} \E_{M(i_0)}[(Y - \fs(X))^2] \notag
		\leq \inf_{\fs\in\FF} \sup_{i\in \cI_{\text{cr}}} \E_{M(i)}[(Y - \fs(X))^2] \notag \\
		&\leq  \inf_{\fs\in\FF} \sup_{i\in\II}\E_{M(i)}[(Y - \fs(X))^2].   \label{eq:upperbdd_prop1}
	\end{align}
	Combining \eqref{eq:lowerbdd_prop1} and \eqref{eq:upperbdd_prop1} it
	immediately follows that
	\begin{align*}
		\inf_{\fs\in\FF} \sup_{i\in\II}\E_{M(i)}[(Y - \fs(X))^2] =  \sup_{i\in\II} \E_{M(i)}[(Y - f(X))^2],
	\end{align*}
	and hence
	\begin{equation*}
		f\in\argmin_{\fs \in \FF} \sup_{i \in \II} \E_{M(i)}[(Y-\fs(X))^2],
	\end{equation*}
	which completes the proof of Proposition~\ref{prop:minimax_equal_causal}.
\end{proofenv}


\begin{proofenv}{\Cref{prop:shift_interventions}}
	Let $\FF$ be the class of all linear functions and let $\II$ denote
	the set of interventions on $X$ that satisfy
	\begin{equation*}
		\sup_{i\in \cI} \lambda_{\min}\big(\E_{M(i)}\big[XX^\top\big]\big) =\infty.
	\end{equation*}
	We claim that the causal function $f(x)=b^\top x$ is the unique
	minimax solution of \eqref{eq:minimax_problem}. We prove the result
	by contradiction. 
	Let $\bar{f}\in\mathcal{F}$ (with $\bar{f}(x)=\bar{b}^{\top}x$) be such that 
	\begin{equation*}
		\sup_{i\in\II}\E_{M(i)}[(Y-\bar{b}^\top X)^2]\leq\sup_{i\in\II}\E_{M(i)}[(Y-b^\top X)^2],
	\end{equation*}
	and assume that $\norm{\bar{b}-b}_2>0$.
	For a fixed $i\in \cI$, we get the following bound
	\begin{align*}
		\E_{M(i)}[(b^\top X-\bar{b}^\top X)^2]
		&=(b-\bar{b})^\top \E_{M(i)}[XX^{\top}](b-\bar{b})
		\geq \lambda_{\min}(\E_{M(i)}[XX^{\top}]) \|b-\bar{b} \|_2^2.
	\end{align*}
	Since we assumed that the minimal eigenvalue is unbounded, this
	means that we can choose $i\in\II$ such that
	$\E_{M(i)}[(b^\top X-\bar{b}^\top X)^2]$ can be arbitrarily large. However,
	applying Proposition~\ref{prop:difference_to_causal_function}, this
	leads to a contradiction since
	$\sup_{i \in \II} \E_{M(i)}[(b^\top X-\bar{b}^\top X)^2]\leq 4\operatorname{Var}_{M}(\xi_Y)$
	cannot be satisfied. Therefore, it must holds that $\bar{b}=b$, which
	moreover implies that $f$ is indeed a solution to the minimax
	problem
	$\argmin_{\fs \in\mathcal{F}}\sup_{i\in\II}\E_{M(i)}[(Y-\fs(X))^2]$,
	as it achieves the lowest possible objective value. This
	completes the proof of Proposition~\ref{prop:shift_interventions}.
\end{proofenv}


\begin{proofenv}{\Cref{prop:difference_to_causal_function}}
	Let $\II$ be a set of interventions on $X$ or $A$ and let
	$\fs\in\FF$ with
	\begin{equation}
		\label{eq:better_than_causal_cond}
		\sup_{i\in\II}\E_{M(i)}[(Y-\fs(X))^2]\leq\sup_{i\in\II}\E_{M(i)}[(Y-f(X))^2].
	\end{equation}
	For any $i\in\II$, the Cauchy-Schwartz inequality implies that
	\begin{align*}
		& \, \E_{M(i)}[(Y-\fs(X))^2]
		= \E_{M(i)}[(f(X)+\xi_Y-\fs(X))^2]\\
		&\,=   \E_{M(i)}[(f(X)-\fs(X))^2]+\E_{M(i)}[\xi_Y^2]
		+2\E_{M(i)}[\xi_Y(f(X)-\fs(X))]\\
		& \,\geq \E_{M(i)}[(f(X)-\fs(X))^2]+\E_{M}[\xi_Y^2] -2\left(\E_{M(i)}[(f(X)-\fs(X))^2]\E_{M}[\xi_Y^2]\right)^{\frac{1}{2}}.
	\end{align*}
	A similar computation shows that the causal function $f$ satisfies
	\begin{equation*}
		\E_{M(i)}[(Y-f(X))^2]=\E_{M}[\xi_Y^2].
	\end{equation*}
	So by condition \eqref{eq:better_than_causal_cond} this 
	implies for any $i\in\II$ that
	\begin{align*}
		\E_{M}[\xi_Y^2] \geq &\,  \E_{M(i)}[(f(X)-\fs(X))^2]+\E_{M}[\xi_Y^2] \\& \qquad
		-2\left(\E_{M(i)}[(f(X)-\fs(X))^2]\E_{M}[\xi_Y^2]\right)^{\frac{1}{2}},
	\end{align*}
	which is equivalent to
	\begin{align*}
		&\E_{M(i)}[(f(X)-\fs(X))^2] \leq 2\sqrt{\E_{M(i)}[(f(X)-\fs(X))^2]\E_{M}[\xi_Y^2]},
	\end{align*}
	i.e. $\E_{M(i)}[(f(X)-\fs(X))^2]\leq 4\E_{M}[\xi_Y^2]$.
	As this inequality holds for all $i\in\II$, we can take the supremum
	over all $i\in\II$, which completes the proof of
	Proposition~\ref{prop:difference_to_causal_function}.  
\end{proofenv}

\begin{proofenv}{\Cref{prop:misspecification_minimax}}
	As argued before, we have that for all $i \in \II_1$, 
	\begin{equation*}
		\E_{M(i)}\big[(Y-f(X))^2\big]=\E_{M(i)}\big[\xi_Y^2\big]=\E_{M}\big[\xi_Y^2\big].
	\end{equation*}
	Let now
	$f_1^*\in\mathcal{F}$ be a minimax solution w.r.t.\ $\II_1$. 
	Then, using that the causal function $f$ lies in $\FF$, it holds that
	\begin{align*}
		\sup_{i\in\II_1}\E_{M(i)}\big[(Y-f_1^*(X))^2\big] &\leq \sup_{i\in\II_1}\E_{M(i)}\big[(Y-f(X))^2\big] = \E_{M}\big[\xi_Y^2\big].
	\end{align*}
	Moreover, if $\II_2\subseteq\II_1$, then it must also hold that
	\begin{align*}
		\sup_{i\in\II_2}\E_{M(i)}\big[(Y-f_1^*(X))^2\big] &\leq \E_{M}\big[\xi_Y^2\big] =\sup_{i\in\II_2}\E_{M(i)}\big[(Y-f(X))^2\big].
	\end{align*}
	To prove the second part, we give a one-dimensional example. Let $\mathcal{F}$ be
	linear (i.e., $f(x)=b x$) and let $\II_1$ consist of shift
	interventions on $X$ of the form
	\begin{equation*}
		X^i\coloneqq g(A^i) + h_2(H^i, \epsilon_X^i)+ c,
	\end{equation*}
	with 
	$c\in [0, K]$. 
	Then, the minimax solution $f^*_1$
	(where $f^*_1(x)=b^*_1 x$) with respect to $\II_1$ is not equal to the
	causal function $f$ as long as $\cov(X, \xi_Y)$ is 
	strictly positive.
	This can be seen by explicitly computing the OLS
	estimator for a fixed shift $c$ and observing that the worst-case
	risk is attained at $c=K$. %
	Now let $\cI_2$ be a set of
	interventions of the same form as $\cI_1$ but including shifts with $c>K$
	such that
	$\cI_2 \not \subseteq \cI_1$. Since $\cF$ consists of linear
	functions, we know that the loss
	$\E_{M(i)}\big[(Y-f_1^*(X))^2\big]$ can become arbitrarily large,
	since
	\begin{align*}
		& \, \E_{M(i)}\big[(Y-f_1^*(X))^2\big]\\
		& \,=(b-b^*_1)^2\E_{M(i)}[X^2]+\E_{M}[\xi_Y^2]+2(b-b^*_1)\E_{M(i)}[\xi_Y X]\\
		& \,=(b-b^*_1)^2(c^2+\E_{M}[X^2]+2c\E_{M}[X])+\E_{M}[\xi_Y^2]
	\\&\qquad 	+2(b-b^*_1)(\E_{M}[\xi_Y X]+\E_{M}[\xi_Y]c),
	\end{align*}
	and $(b-b^*)^2>0$. In contrast, the loss for the causal function
	is always $\E_{M}[\xi_Y^2]$, so the worst-case risk of $f^*_1$
	becomes arbitrarily worse than that of $f$. This completes the
	proof of Proposition~\ref{prop:misspecification_minimax}.
\end{proofenv}


\begin{proofenv}{\Cref{prop:suff_general}}
	Let $\epsilon > 0$. By definition of the infimum, we can find $f^* \in \FF$ such that
	\begin{align*}
		&\left| \sup_{i\in\II}\E_{M(i)}\big[(Y-f^*(X))^2\big]
		- \inf_{\fs\in\FF}\,
		\sup_{i\in\II}\E_{M(i)}\big[(Y-\fs(X))^2\big] \right| \leq \epsilon.
	\end{align*}
	Let now $\tilde{M} \in \MM$ be s.t.\ $\P_{\tilde{M}} = \P_M$. By assumption, the left-hand side 
	of the above inequality is unaffected by substituting $M$ for $\tilde{M}$, and the 
	result thus follows. 
\end{proofenv}


\begin{proofenv}{\Cref{prop:impossibility_interpolation}}
	We first show that the causal parameter $\beta$ is not a minimax solution.
	Let $u := \sup \II < \infty$, since $\II$ is bounded,
	and
	take $b = \beta + 1/(\sigma u)$. 
	By an explicit computation we get that 
	\begin{align*}
		\inf_{\bes \in \R}\,
		\sup_{i\in\II} \E_{M(i)}\big[(Y-\bes X)^2\big]   \leq &\, \sup_{i\in\II} \E_{M(i)}\big[(Y-b X)^2\big]
		\\=&  \sup_{i\in\II} \E_{M(i)}\big[(\ep_Y + \tfrac{1}{\sigma}H - 
		\tfrac{1}{\sigma u}iH)^2\big]\\
		= &\, \sup_{i\in\II} \left[1 + \left(1 - \tfrac{i}{u}\right)^2\right] 
		\\<&  2
		\\=& \sup_{i\in\II}\E_{M(i)}\big[(Y-\beta X)^2\big],
	\end{align*}
	where the last inequality holds because $0 < 1 + (1 - i/u)^2 < 2$ for all
	$i \in \II$, and since $\II\subset \R_{>0}$ is compact with upper bound $u$. Hence,
	\begin{align*}
		\sup_{i\in\II}\E_{M(i)}\big[(Y-\beta X)^2\big]
		- \inf_{\bes \in \R}\,
		\sup_{i\in\II} \E_{M(i)}\big[(Y-\bes X)^2\big] > 0,
	\end{align*}
	proving that the causal parameter is not a minimax solution for model $M$
	w.r.t. $(\cF, \cI)$. Recall that in order to prove that $(\bP_{M},\cM)$ does not generalize with respect to $\cI$ we have to show that there exists an $\ep >0$ such that for all $b\in \R$ it holds that
	\begin{align*}
		&\sup_{\tilde M: \bP_{\tilde{M}}= \bP_M}\big|   \sup_{i\in\II}\E_{\tilde{M}(i)}\big[(Y-b X)^2\big]
		- \inf_{\bes \in \R}\,
		\sup_{i\in\II} \E_{\tilde{M}(i)}\big[(Y-\bes X)^2\big]\big| \geq \ep.
	\end{align*}
	Thus, it remains to show that for all $b \not = \beta$ there exists a model $\tilde M \in \cM$ with $\bP_M = \bP_{\tilde M}$ such that the generalization loss is bounded below uniformly by a positive constant. We will show the stronger statement that 
	for any $b \neq \beta$, there exists a model $\tilde{M}$ with
	$\P_{\tilde{M}} = \P_M$, such that under $\tilde{M}$, 
	$b$ results in arbitrarily large generalization error.
	Let $c > 0$ and $i_0 \in \II$. Define 
	\begin{align*}
		\tilde\sigma \coloneqq \frac{\sign{((\beta - b)i_0)}\sqrt{1 + c} - 1}{
			(\beta - b)i_0} > 0,
	\end{align*}
	and let 
	$\tilde M \coloneqq M(\gamma, \beta, \tilde{\sigma}, Q)$. By construction of the model class $\MM$, 
	it holds that $\P_{\tilde{M}} = \P_{M}$.
	Furthermore, by an explicit computation we get that 
	\begin{align}\label{eq:proof_imp_a1}
		\begin{split}
			\sup_{i\in\II} \E_{\tilde{M}(i)}\big[(Y-bX)^2\big]
			\geq & \, \E_{\tilde{M}(i_0)}\big[(Y-bX)^2\big]
			\\=& \E_{\tilde{M}(i_0)}\big[((\beta-b)i_0H+\ep_Y+
			\tfrac{1}{\tilde\sigma}H)^2\big]\\
			=& \, \E_{\tilde{M}(i_0)}\big[([(\beta-b)i_0\tilde{\sigma}+1] \ep_H+\ep_Y )^2\big]  
			\\=& [(\beta-b)i_0\tilde{\sigma}+1]^2 + 1\\
			= & \, ((\beta - b)i_0\tilde\sigma)^2 + 2(\beta - b)i_0\tilde\sigma +
			2\\
			=& \, (\sign{((\beta-b)i_0)}\sqrt{1+c}-1)^2 
			+ 2\sign{((\beta-b)i_0)}\sqrt{1+c}\\
			=& \,  c + 2.
		\end{split}
	\end{align}
	Finally, by definition of the infimum, it holds that
	\begin{align}\label{eq:proof_imp_b1}
		\begin{split}
			\inf_{\bes \in \R}\,\sup_{i\in\II} \E_{\tilde{M}(i)}\big[(Y-\bes
			X)^2\big] \leq \sup_{i\in\II}\E_{\tilde{M}(i)}\big[(Y-\beta
			X)^2\big] = 2.
		\end{split}
	\end{align}
	Combining~\eqref{eq:proof_imp_a1} and~\eqref{eq:proof_imp_b1} 
	yields that the generalization error is bounded below by $c$. That is,
	\begin{align*}
		&\big|   \sup_{i\in\II}\E_{\tilde{M}(i)}\big[(Y-b X)^2\big] 
		- \inf_{\bes \in \R}\,
		\sup_{i\in\II} \E_{\tilde{M}(i)}\big[(Y-\bes X)^2\big]\big| \geq c.
	\end{align*}
	The above results make no assumptions on $\gamma$, and hold true, 
	in particular, if $\gamma \neq 0$ (in which case Assumption~\ref{ass:identify_f} 
	is satisfied, see Appendix~\ref{sec:IVconditions}).
	This completes the proof of Proposition~\ref{prop:impossibility_interpolation}.
	
\end{proofenv}


\begin{proofenv}{\Cref{prop:genX_intra}}
	Let $\II$ be a well-behaved set of interventions on $X$. We
	consider two cases; (A) all interventions in $\II$ are
	confounding-preserving and (B) there is at least one intervention in
	$\II$ that is confounding-removing.
	
	\textbf{Case (A):} In this case, we prove the result in two steps:
	(i) We show that $(A, \xi_X, \xi_Y)$ is identified from the
	observational distribution $\P_M$. (ii) We show that 
	this 
	implies that the intervention distributions $(X^i, Y^i)$, $i \in \II$, are also
	identified from the observational distribution, and conclude by using
	Proposition~\ref{prop:suff_general}.   
	Some of the details will be slightly technical
	because we allow for a large class of
	distributions (e.g., there is no assumption on the existence of densities).
	
	We begin with step (i). In this case, $\II$ is a set of confounding-preserving
	interventions on $X$, and we have that
	$\supp_{\II}(X)\subseteq\supp(X)$. Fix
	$\tilde{M} =(\tilde{f},\tilde{g},\tilde{h}_1,\tilde{h}_2,\tilde{Q})
	\in \MM$ such that $\P_{\tilde{M}} = \P_M$ and let
	$(\tilde{X}, \tilde{Y},\tilde{H},\tilde{A})$ be generated by the SCM
	of $\tilde{M}$. We have that
	$(X,Y,A) \eqd (\tilde{X}, \tilde{Y},\tilde{A})$ and by
	Assumption~\ref{ass:identify_f}, we have that $f \equiv \tilde{f}$
	on $\supp(X)$, hence $f(X) \eqas \tilde{f}(X)$. Further, fix any
	$B\in \cB(\R^p)$ (i.e., in the Borel sigma-algebra on $\R^p$) and note
	that
	\begin{align*}
		\bE_M[\mathbbm{1}_{B}(A)X|A]
		&=\bE_M[\mathbbm{1}_{B}(A)g(A)
		+\mathbbm{1}_{B}(A)h_2(H,\ep_X)|A] \\
		&= \bE_M[\mathbbm{1}_{B}(A)g(A)|A]
		+ \mathbbm{1}_{B}(A)\bE[h_2(H,\ep_X)]
		= \mathbbm{1}_{B}(A)g(A),
	\end{align*}
	almost surely. Here, we have used our modeling assumption $\E[h_2(H, \epsilon_X)] = 0$. 
	Hence, by similar arguments for
	$\bE_{\tilde{M}}(\mathbbm{1}_{B}(\tilde{A})\tilde{X}|\tilde{A})$ and
	the fact that
	$(X,Y,A) \eqd (\tilde{X}, \tilde{Y},\tilde{A})$ we have
	that
	\begin{align*}
		\mathbbm{1}_{B}(A)g(A) &\eqas\bE_M (\mathbbm{1}_{B}(A)X|A) 
		\\&\eqd \bE_{\tilde{M}}(\mathbbm{1}_{B}(\tilde{A})\tilde{X}|\tilde{A}) \\&\eqas \mathbbm{1}_{B}(\tilde{A})\tilde{g}(\tilde{A}).
	\end{align*}
	We conclude that
	$\mathbbm{1}_{B}(A)g(A)\eqd
	\mathbbm{1}_{B}(\tilde{A})\tilde{g}(\tilde{A})$ for any
	$B\in \cB(\R^p)$.
	Let $\bP$ and
	$\tilde{\bP}$ denote the respective background probability measures on which
	the random elements $(X,Y,H,A)$ and
	$(\tilde{X},\tilde{Y},\tilde{H},\tilde{A})$ are defined. 
	Fix any $F\in \sigma(A)$ (i.e., in the sigma-algebra
	generated by $A$) and note that there exists a $B\in \cB(\R^p)$ such
	that $F=\{A\in B\}$. Since $A \eqd \tilde{A}$, we have that,
	\begin{align*}
		\int_F g(A) \, \mathrm{d} \bP &= \int \mathbbm{1}_{B}(A) g(A) \, \mathrm{d} \bP \\&= \int \mathbbm{1}_{B}(\tilde{A})\tilde{g}(\tilde{A})  \, \mathrm{d} \tilde{\bP}  \\&= \int  \mathbbm{1}_{B}(A)\tilde{g}(A)  \, \mathrm{d}\bP \\&= \int_F \tilde{g}(A)  \, \mathrm{d}\bP.
	\end{align*}
	Both $g(A)$ and $\tilde{g}(A)$ are $\sigma(A)$-measurable and they
	agree integral-wise over every set $F\in \sigma(A)$, so we must have
	that $g(A) \eqas \tilde{g}(A)$. With
	$\eta(a,b,c)= (a,c-\tilde{f}(b),b-\tilde{g}(a))$ we have that
	\begin{align*}
		(A,\xi_Y,\xi_X)   &\eqas (A,Y-\tilde{f}(X),X-\tilde{g}(A)) 
		=\eta(A,X,Y) 
		\eqd\eta(\tilde{A},\tilde{X},\tilde{Y}) 
		= (\tilde{A},\tilde{\xi}_Y,\tilde{\xi}_X),
	\end{align*}
	so $(A,\xi_Y,\xi_X)
	\eqd(\tilde{A},\tilde{\xi}_Y,\tilde{\xi}_X)$. This
	completes step (i).
	
	Next, we proceed with step (ii). Take an arbitrary intervention
	$i \in \II$ and let $\phi^i, I^i, \tilde{I}^i$ with
	$I^i \eqd\tilde{I}^i$,
	$I^i\independent (\ep_X^i,\ep_Y^i,\ep_H^i,\ep_A^i) \sim Q$ and
	$\tilde{I}^i \independent
	(\tilde{\ep}^i_X,\tilde{\ep}^i_Y,\tilde{\ep}^i_H,\tilde{\ep}^i_A)
	\sim \tilde{Q}$ be such that the structural assignments for $X^i$ and
	$\tilde{X}^i$ in $M(i)$ and $\tilde{M}(i)$, respectively,  are given as
	\begin{align*}
		X^i &:= \phi^i(A^i,g(A^i), h_2(H^i, \epsilon_X^i), I^i), \\ \tilde{X}^i &:= \phi^i(\tilde{A}^i,\tilde{g}(\tilde{A}^i), \tilde{h}_2(\tilde{H}^i, \tilde{\epsilon}_X^i), \tilde{I}^i).
	\end{align*}
	Define $\xi_X^i := h_2(H^i, \epsilon_X^i)$,
	$\xi_Y^i := h_1(H^i, \epsilon_Y^i)$,
	$\tilde{\xi}_X^i := \tilde{h}_2(\tilde{H}^i, \tilde{\epsilon}_X^i)$
	and
	$\tilde{\xi}_Y^i := \tilde{h}_1(\tilde{H}^i,
	\tilde{\epsilon}_Y^i)$.
	Then, it holds that
	\begin{equation*}
		(A^i, \xi_X^i, \xi_Y^i) \eqd (A,\xi_X, \xi_Y)  \eqd (\tilde{A},  \tilde{\xi}_X, \tilde{\xi}_Y)  \eqd (\tilde{A}^i, \tilde{\xi}_X^i, \tilde{\xi}_Y^i),
	\end{equation*}
	where we used step (i), that $(A^i, \xi_X^i, \xi_Y^i)$ and
	$(A,\xi_X, \xi_Y)$ are generated by identical functions of the noise
	innovations and that $ (\ep_X,\ep_Y,\ep_H,\ep_A) $ and
	$(\ep_X^i,\ep_Y^i,\ep_H^i,\ep_A^i)$ have identical distributions.
	Adding a random variable with the same distribution, that is
	mutually independent with all other variables, on both sides does
	not change the distribution of the bundle, hence
	\begin{align*}
		(A^i, \xi_X^i, \xi_Y^i,I^i)  \eqd (\tilde{A}^i, \tilde{\xi}_X^i, \tilde{\xi}_Y^i, \tilde{I}^i).
	\end{align*}
	Define
	$\kappa(a,b,c,d) :=
	(\phi^i(a,\tilde{g}(a),b,d),\tilde{f}(\phi^i(a,\tilde{g}(a),b,d))+c)$. As
	shown in step (i) above, we have that $g(A^i)\eqas \tilde{g}(A^i)$. Furthermore,
	since $\supp(X^i) \subseteq\supp(X)$ we have that
	$f(X^i) \eqas \tilde{f}(X^i)$, and hence
	\begin{align*}
		(X^i, Y^i) \eqas &\, (X^i,\tilde{f}(X^i)+\xi_Y^i) \\
		=& \,  (\phi^i(A^i,g(A^i), \xi_X^i , I^i) ,  \, \, \tilde{f}(\phi^i(A^i,g(A^i), \xi_X^i, I^i))+\xi_Y^i) \\
		\eqas  &\,(\phi^i(A^i,\tilde{g}(A^i), \xi_X^i , I^i) , \, \, \tilde{f}(\phi^i(A^i,\tilde{g}(A^i), \xi_X^i, I^i))+\xi_Y^i) \\
		=& \, \kappa(A^i, \xi_X^i, \xi_Y^i,I^i)
		\\\eqd& \kappa(\tilde{A}^i, \tilde{\xi}_X^i, \tilde{\xi}_Y^i, \tilde{I}^i)
		\\=& (\tilde{X}^i, \tilde{Y}^i).
	\end{align*}
	Thus, $\bP_{M(i)}^{(X,Y)} = \bP_{\tilde{M}(i)}^{(X,Y)}$, which
	completes step (ii).  Since $i \in \II$ was arbitrary, the result now
	follows from Proposition~\ref{prop:suff_general}.
	
	\textbf{Case (B):} Assume that the intervention set $\II$
	contains at least one confounding-removing intervention. Let
	$\tilde{M} =(\tilde{f},\tilde{g},\tilde{h}_1,\tilde{h}_2,\tilde{Q})
	\in \MM$ be such that $\P_{\tilde{M}} = \P_M$. Then, by
	Proposition~\ref{prop:minimax_equal_causal}, it follows that the
	causal function $\tilde{f}$ is a minimax solution w.r.t.\
	$(\tilde{M}, \II)$.  By Assumption~\ref{ass:identify_f}, we further
	have that $\tilde{f}$ and $f$ coincide on
	$\supp(X) \supseteq \supp_{\II}(X)$. Hence, it follows that
	\begin{align*}
		\inf_{\fs \in \FF} \sup_{i \in \II} \E_{\tilde{M}(i)}[(Y - \fs(X))^2]&= \sup_{i \in \II} \E_{\tilde{M}(i)}[(Y - \tilde{f}(X))^2] \\&= \sup_{i \in \II} \E_{\tilde{M}(i)}[(Y - f(X))^2],
	\end{align*}
	showing that also $f$ is a minimax solution w.r.t.\
	$(\tilde{M}, \II)$. This completes the proof of
	Proposition~\ref{prop:genX_intra}.  
\end{proofenv}


\begin{proofenv}{\Cref{prop:genX_extra}}
	Let $\tilde{M} \in \MM$ be such that $\P_{\tilde{M}} = \P_M$. By Assumptions~\ref{ass:identify_f}~and~\ref{ass:gen_f}, 
	it holds that $f \equiv \tilde{f}$. The proof now proceeds analogously to that of Proposition~\ref{prop:genX_intra}. 
\end{proofenv}


\begin{proofenv}{\Cref{prop:extrapolation_bounded_deriv_cr}}
	By Assumption~\ref{ass:identify_f}, $f$ is identified on
	$\supp^{M}(X)$ by the observational distribution $\bP_{M}$. Let
	$\cI$ be a set of interventions containing at least one
	confounding-removing intervention.
	For any
	$\tilde{M}=(\tilde{f},\tilde{g},\tilde{h}_1,\tilde{h}_2,\tilde{Q})\in
	\cM$, Proposition~\ref{prop:minimax_equal_causal} yields that the
	causal function is a minimax solution. That is,
	\begin{align} \notag
		\inf_{\fs\in\mathcal{F}}\sup_{i\in\cI}\E_{\tilde{M}(i)}\big[(Y-\fs(X))^2\big]
		&= \sup_{i\in\cI}\E_{\tilde{M}(i)}\big[(Y-\tilde{f}(X))^2\big] \\
		&= \sup_{i\in \cI }\E_{\tilde{M}(i)}[\xi_Y^2] =\E_{\tilde{M}}[\xi_Y^2], \label{eq:propboundedderiv_causalfunctionsolvesminimax}
	\end{align}
	where we used that any intervention $i\in \cI$ does not affect the
	distribution of $\xi_Y=\tilde{h}_2(H,\ep_Y)$. Now, assume that
	$\tilde{M}=(\tilde{f},\tilde{g},\tilde{h}_1,\tilde{h}_2,\tilde{Q})\in
	\cM$ satisfies $\bP_{\tilde{M}} = \bP_M$. Since $(\P_M,\cM)$
	satisfies Assumption~\ref{ass:identify_f}, we have that
	$f \equiv \tilde{f}$ on $\supp^M(X)=\supp^{\tilde{M}}(X)$. Let $f^*$
	be any function in $\FF$ such that $f^*=f$ on $\supp^M(X)$.  We
	first show that
	$\norm{\tilde{f} - f^*}_{\cI,\infty} \leq 2\delta K$, where
	$\|f\|_{\cI,\infty} := \sup_{x\in\supp_{\cI}^M(X)}\|f(x)\|$.  By the
	mean value theorem, for all $\fs \in\FF$ it holds that
	$\abs{\fs(x) - \fs(y)} \leq K\norm{x - y}$, for all
	$x, y\in \mathcal D$.  For any $x \in \supp^M_\cI(X)$ and
	$y \in \supp^M(X)$ we have
	\begin{align*}
		\abs[\big]{\tilde{f}(x) - f^*(x)} 
		&=  \abs[\big]{\tilde{f}(x) - \tilde{f}(y) + f^*(y) - f^*(x)}\\
		&\leq  \abs[\big]{\tilde{f}(x) - \tilde{f}(y)} + \abs[\big]{f^*(y) - f^*(x)}\\
		&\leq 2 K \norm{x - y},
	\end{align*}
	where we used the fact that $\tilde{f}(y)= f(y) =f^*(y)$, for all
	$y\in \supp^M(X)$.  In particular, it holds that
	\begin{align}\label{eq:unif_norm}
		\begin{split}
			\norm{\tilde{f} - f^*}_{\cI,\infty} 
			&=\sup_{x\in \supp^M_\cI(X)} \abs[\big]{\tilde{f}(x) - f^*(x)}\\
			&\leq  2K\sup_{x\in \supp^M_\cI(X)} \inf_{y\in\supp^M(X)}\norm{x - y}\\
			&= 2\delta K.
		\end{split}
	\end{align}
	For any $i\in\II$ we have that
	\begin{align} \notag 
		\E_{\tilde{M}(i)}\big[(Y- f^*(X))^2\big] 
		= & \, \E_{\tilde{M}(i)}\big[(\tilde{f}(X)+\xi_Y-f^*(X))^2\big]\nonumber\\
		=&\, \E_{\tilde{M}}\big[\xi_Y^2\big] +
		\E_{\tilde{M}(i)}\big[(\tilde{f}(X)-f^*(X))^2\big]\nonumber\\
		&\ + 2\E_{\tilde{M}(i)}\big[\xi_Y(\tilde{f}(X)-f^*(X))\big].\label{eq:starting_pt_split}
	\end{align}
	Next, we can use Cauchy-Schwarz,
	\eqref{eq:propboundedderiv_causalfunctionsolvesminimax} and
	\eqref{eq:unif_norm} in \eqref{eq:starting_pt_split} to get that
	\begin{align} \nonumber
		\bigg|\sup_{i\in\II}&\, \E_{\tilde{M}(i)}\big[(Y-f^*(X))^2\big]
		-\inf_{\fs\in\FF}\sup_{i\in\II}\E_{\tilde{M}(i)}\big[(Y-\fs(X))^2\big]\nonumber \bigg|\\
		=& \,  \sup_{i\in\II} \E_{\tilde{M}(i)}\big[(Y-f^*(X))^2\big]-\E_{\tilde{M}}[\xi_Y^2]\nonumber \\ \nonumber
		= & \,  \sup_{i\in\II} \big(
		\E_{\tilde{M}(i)}\big[(\tilde{f}(X)-f^*(X))^2\big] +2\E_{\tilde{M}(i)}\big[\xi_Y(\tilde{f}(X)-f^*(X))\big] \big) \nonumber\\
		\leq &\,  4\delta^2K^2+4\delta K \sqrt{\var_M(\xi_Y)},\label{eq:ineq_prt}
	\end{align}
	proving the first statement.  Finally, if $\II$ consists only of
	confounding-removing interventions, then the bound in
	\eqref{eq:ineq_prt} can be improved by using that $\E[\xi_Y]=0$
	together with $H\independent X$. In that case, we get that
	$\E_{\tilde{M}(i)}\big[\xi_Y(\tilde{f}(X)-f(X))\big]=0$ and hence
	the bound becomes $4\delta^2 K^2$. This completes the proof of
	Proposition~\ref{prop:extrapolation_bounded_deriv_cr}.
\end{proofenv}

\begin{proofenv}{\Cref{prop:extrapolation_bounded_deriv}}
	By Assumption~\ref{ass:identify_f},
	$f$ is identified on $\supp^{M}(X)$ by the observational
	distribution $\bP_{M}$. Let $\cI$ be a set of confounding-preserving
	interventions.  For a fixed $\epsilon>0$, let $f^*\in\mathcal{F}$ be
	a function satisfying
	\begin{align}
		&\big|\sup_{i\in\cI}\E_{M(i)}\big[(Y-f^*(X))^2)\big]
		-\inf_{\fs\in\mathcal{F}}\sup_{i\in\cI}\E_{M(i)}\big[(Y-\fs(X))^2)\big] \big| \leq  \epsilon. \label{eq:fstarDiffEpsilon}
	\end{align}
	Fix any secondary model
	$\tilde{M}=(\tilde{f},\tilde{g},\tilde{h}_1,\tilde{h}_2,\tilde{Q})\in
	\cM$ with $\bP_{\tilde{M}} = \bP_M$.  The general idea is to
	derive an upper bound for
	$\sup_{i\in\II}\E_{\tilde{M}(i)}[(Y-f^*(X))^2]$ and a lower
	bound for
	$\inf_{\fs\in\FF}\,
	\sup_{i\in\II}\E_{\tilde{M}(i)}[(Y-\fs(X))^2]$ which will
	allow us to bound the absolute difference of interest.
	
	Since $(\P_M,\cM)$ satisfies
	Assumption~\ref{ass:identify_f}, we have that
	$f \equiv \tilde{f}$ on
	$\supp^M(X)=\supp^{\tilde{M}}(X)$. We first show that
	$$\norm{\tilde{f} - f}_{\cI,\infty} \leq 2\delta K,$$ where $\|f\|_{\cI,\infty} := \sup_{x\in\supp_{\cI}^M(X)}\|f(x)\|$.  By the mean
	value theorem, for all $\fs \in\FF$ it holds that
	$\abs{\fs(x) - \fs(y)} \leq K\norm{x - y}$, for all
	$x, y\in \mathcal D$.  For any $x \in \supp^M_\cI(X)$ and
	$y \in \supp^M(X)$ we have
	\begin{align*}
		\abs[\big]{\tilde{f}(x) - f(x)} 
		&=  \abs[\big]{\tilde{f}(x) - \tilde{f}(y) + f(y) - f(x)}\\
		&\leq  \abs[\big]{\tilde{f}(x) - \tilde{f}(y)} + \abs[\big]{f(y) - f(x)}\\
		&\leq 2 K \norm{x - y},
	\end{align*}
	where we used the fact that $\tilde{f}(y)=f(y)$, for all
	$y\in \supp_M(X)$.  In particular, it holds that
	\begin{align}\label{eq:unif_norm2}
		\begin{split}
			\norm{\tilde{f} - f}_{\cI,\infty} 
			&= \sup_{x\in \supp^M_\cI(X)} \abs[\big]{\tilde{f}(x) - f(x)}\\
			&\leq  2K\sup_{x\in \supp^M_\cI(X)} \inf_{y\in\supp^M(X)}\norm{x - y}\\
			&= 2\delta K.
		\end{split}
	\end{align}
	Let now $i\in\II$ be fixed. The term $\xi_Y = h_1(H, \epsilon_Y)$ is
	not affected by the intervention $i$. Furthermore,
	$\P^{(X,\xi_Y)}_{M(i)}=\P^{(X,\xi_Y)}_{\tilde{M}(i)}$ since $i$ is
	confounding-preserving (this can be seen by a slight modification to the
	arguments from case (A) in the proof of
	Proposition~\ref{prop:genX_intra}). Thus, for any $\fs\in \cF$ we
	have that
	\begin{align}
		&\E_{\tilde{M}(i)}\big[(Y-\fs(X))^2\big] \nonumber\\
		&=\E_{\tilde{M}(i)}\big[(\tilde{f}(X)+\xi_Y-\fs(X)+f(X)-f(X))^2\big]\nonumber\\
		&=\E_{\tilde{M}(i)}\big[\xi_Y^2\big] +
		\E_{\tilde{M}(i)}\big[(f(X)-\fs(X))^2\big]
		+
		\E_{\tilde{M}(i)}\big[(\tilde{f}(X)-f(X))^2\big]\nonumber\\
		&\qquad\qquad + 2\E_{\tilde{M}(i)}\big[\xi_Y(f(X)-\fs(X))\big]\nonumber\\
		&\qquad\qquad +
		2\E_{\tilde{M}(i)}\big[(\tilde{f}(X)-f(X))(f(X)-\fs(X))\big]\nonumber\\
		&\qquad\qquad +
		2\E_{\tilde{M}(i)}\big[\xi_Y(\tilde{f}(X)-f(X))\big]\nonumber\\
		&=\E_{M(i)}\big[\xi_Y^2\big] +
		\E_{M(i)}\big[(f(X)-\fs(X))^2\big]+
		\E_{M(i)}\big[(\tilde{f}(X)-f(X))^2\big]\nonumber\\
		&\qquad\qquad +2\E_{M(i)}\big[\xi_Y(f(X)-\fs(X))\big]\nonumber\\
		&\qquad\qquad +
		2\E_{M(i)}\big[(\tilde{f}(X)-f(X))(f(X)-\fs(X))\big]\nonumber\\
		&\qquad\qquad + 2\E_{M(i)}\big[\xi_Y(\tilde{f}(X)-f(X))\big]\nonumber\\
		&=\E_{M(i)}\big[(Y-\fs(X))^2\big] + L_1^i(\tilde{f}) +
		L_2^i(\tilde{f},\fs) + L_3^i(\tilde{f}),\label{eq:decomposition_intoLs}
	\end{align}
	where, we have made the following definitions
	\begin{align*}
		L_1^i(\tilde{f})&\coloneqq \E_{M(i)}\big[(\tilde{f}(X)-f(X))^2\big],\\
		L_2^i(\tilde{f},\fs)&\coloneqq
		2\E_{M(i)}\big[(\tilde{f}(X)-f(X))(f(X)-\fs(X))\big],\\
		L_3^i(\tilde{f})&\coloneqq 2\E_{M(i)}\big[\xi_Y(\tilde{f}(X)-f(X))\big].
	\end{align*}
	Using \eqref{eq:unif_norm2} it follows that
	\begin{equation}
		\label{eq:estimate_L1}
		0\leq L_1^i(\tilde{f}) \leq 4\delta^2 K^2,
	\end{equation}
	and by the Cauchy-Schwarz inequality it follows that
	\begin{align} 
		\abs[\big]{L_3^i(\tilde{f})} &\leq 2 \sqrt{\var_M(\xi_Y)4\delta^2K^2} =4\delta K \sqrt{\var_M(\xi_Y)}.	\label{eq:estimate_L3}
	\end{align}
	Let now $\fs\in\FF$ be 
	any function
	such that
	\begin{equation} \label{eq:ConditionBetterThanTildeCausal}
		\sup_{i\in\II}\E_{\tilde M(i)}\big[(Y-\fs(X))^2)\big]\leq\sup_{i\in\II}\E_{\tilde M(i)}\big[(Y-\tilde{f}(X))^2)\big],
	\end{equation}
	then by \eqref{eq:unif_norm2}, the Cauchy-Schwarz inequality and
	Proposition~\ref{prop:difference_to_causal_function}, it holds for all $i\in \cI$ that
	\begin{align} \nonumber
		L_2^i(\tilde{f},\fs) =& \, 2\E_{M(i)}\big[(\tilde{f}(X)-f(X))(f(X)-\fs(X))\big]\\ 
		= & \, 2\E_{\tilde{M}(i)}\big[(\tilde{f}(X)-f(X))(f(X)-\fs(X))\big]\nonumber\\
		= & \,  -2\E_{\tilde{M}(i)}\big[(\tilde{f}(X)-f(X))^2\big]
		+2\E_{\tilde{M}(i)}\big[(\tilde{f}(X)-f(X))(\tilde{f}(X)-\fs(X))\big]\nonumber\\
		\geq &\, -8\delta^2K^2 -
		2\sqrt{4\delta^2K^2}\sqrt{4\var_M(\xi_Y)}\nonumber\\
		= & \, -8\delta^2K^2- 8\delta K \sqrt{\var_M(\xi_Y)}, \label{eq:estimate_L2_BetterThanTildeCausal}
	\end{align}
	where, in the third equality, we have added and subtracted the
	term
	$2\E_{\tilde{M}(i)}\big[(\tilde{f}(X)-f(X))\tilde{f}(X)\big]$.
	Now let
	$$\cS := \{\fs \in \cF :
	\sup_{i\in\II}\E_{\tilde{M}(i)}\big[(Y-\fs(X))^2\big] \leq
	\sup_{i\in\II}\E_{\tilde{M}(i)}\big[(Y-\tilde{f}(X))^2\big]
	\}$$ be the set of all functions satisfying
	\eqref{eq:ConditionBetterThanTildeCausal}. Due to
	\eqref{eq:decomposition_intoLs}, \eqref{eq:estimate_L1},
	\eqref{eq:estimate_L3} and
	\eqref{eq:estimate_L2_BetterThanTildeCausal} we have the
	following lower bound of interest
	\begin{align} \notag
		&\inf_{\fs\in\FF}\,	\sup_{i\in\II}\E_{\tilde{M}(i)}\big[(Y-\fs(X))^2\big] \\\notag
		=& \inf_{\fs\in\cS}\,	\sup_{i\in\II}\E_{\tilde{M}(i)}\big[(Y-\fs(X))^2\big] \\\notag
		=& \inf_{\fs\in \cS}\,
		\sup_{i\in\II}\big\{ \E_{M(i)}\big[(Y-\fs(X))^2\big] + L_1^i(\tilde{f}) +
		L_2^i(\tilde{f},\fs) + L_3^i(\tilde{f}) \big\} \\\notag
		\geq& \inf_{\fs\in \cS}\,
		\sup_{i\in\II} \E_{M(i)}\big[(Y-\fs(X))^2\big]  -8\delta^2K^2- 8\delta K \sqrt{\var_M(\xi_Y)} - 4\delta K \sqrt{\var_M(\xi_Y)} \\ \label{eq:lowerboundoninf}
		\geq & \inf_{\fs\in \cF}\,
		\sup_{i\in\II} \E_{M(i)}\big[(Y-\fs(X))^2\big]  
		- 8 \delta^2 K^2 - 12 \delta K \sqrt{\var_M(\xi_Y)}.
	\end{align}
	Next, we construct the aforementioned upper bound of interest. To that end, note that 
	\begin{align} \notag
		&\sup_{i\in\II} \E_{\tilde{M}(i)}\big[(Y-f^*(X))^2\big] \\ \label{eq:QuantityForUpperBound}
		&\quad = \sup_{i\in\II} \left\{ \E_{M(i)}\big[(Y-f^*(X))^2\big]
		+ L_1^i(\tilde{f}) +
		L_2^i(\tilde{f},f^*) + L_3^i(\tilde{f}) \right\},
	\end{align}	
	by \eqref{eq:decomposition_intoLs}. We have already
	established upper bounds for $L_1^i(\tilde f)$ and
	$L_3^i(\tilde{f})$ in \eqref{eq:estimate_L1} and
	\eqref{eq:estimate_L3}, respectively. In order to control
	$L_2^i(\tilde{f},f^*)$ we introduce an auxiliary function. Let
	$\bar{f}^*\in \cF$ satisfy
	\begin{equation} \label{eq:BarfStarBetterThanCausal}
		\sup_{i\in\II}\E_{M(i)}\big[(Y-\bar{f}^*(X))^2)\big]\leq\sup_{i\in\II}\E_{M(i)}\big[(Y-f(X))^2)\big],
	\end{equation}
	and
	\begin{align}
		&\bigg|\sup_{i\in\II}\E_{M(i)}\big[(Y-\bar{f}^*(X))^2\big] - \inf_{\fs\in\FF}\,
		\sup_{i\in\II}\E_{M(i)}\big[(Y-\fs(X))^2\big]\bigg|\leq \epsilon. \label{eq:epsilonineqforBarfStar}
	\end{align}
	Choosing such a $\bar f^*\in \cF$ is always possible. If $f$
	is an $\ep$-minimax solution, i.e., it satisfies
	\eqref{eq:epsilonineqforBarfStar}, then choose $\bar
	f^*=f$. Otherwise, if $f$ is not a $\ep$-minimax solution,
	then choose any $\bar f^*\in \cF$ that is an $\ep$-minimax
	solution (which is always possible). In this case we have that
	\begin{align*}
		&\sup_{i\in\II}\E_{M(i)}\big[(Y-\bar{f}^*(X))^2\big] - \inf_{\fs\in\FF}\,
		\sup_{i\in\II}\E_{M(i)}\big[(Y-\fs(X))^2\big] \leq \ep ,
	\end{align*}
	and
	\begin{align*}
		&\sup_{i\in\II}\E_{M(i)}\big[(Y-f(X))^2\big]
		- \inf_{\fs\in\FF}\,
		\sup_{i\in\II}\E_{M(i)}\big[(Y-\fs(X))^2\big] \geq \ep ,
	\end{align*}
	which implies that \eqref{eq:BarfStarBetterThanCausal} is
	satisfied. We can now construct an upper bound on
	$L_2^i(\tilde{f},f^*)$ in terms of $L_2^i(\tilde{f},\bar f^*)$
	by noting that for all $i\in\cI$
	\begin{align} \notag
		\abs[\big]{L_2^i(\tilde{f},f^*)}
		= & \,  2\big|\E_{M(i)}\big[(\tilde{f}(X)-f(X))(f(X)-f^*(X))\big] \big| \notag \\
		\notag
		\leq & \,  2\big|\E_{M(i)}\big[(\tilde{f}(X)-f(X))(f(X)-\bar{f}^*(X))\big]\big| \\ \notag
		& \,  +2\E_{M(i)}\big|(\tilde{f}(X)-f(X))(\bar{f}^*(X)-f^*(X))\big| \\ \notag
		= & \, \abs[\big]{L_2^i(\tilde{f},\bar f^*)} +2\E_{M(i)}\big|(\tilde{f}(X)-f(X))(\bar{f}^*(X)-f^*(X))\big| \\  \notag
		\leq& \, 2\sqrt{\E_{M(i)}\left[(\tilde{f}(X)-f(X))^2\right]\E_{M(i)}\left[(\bar{f}^*(X)-f^*(X))^2 \right]} +  \abs[\big]{L_2^i(\tilde{f},\bar f^*)} \\ \label{eq:UpperboundOfL2TildefStarf}
		\leq & \,  \abs[\big]{L_2^i(\tilde{f},\bar f^*)} +4\delta K\sqrt{\E_{M(i)}\left[(\bar{f}^*(X)-f^*(X))^2 \right]},
	\end{align}
	where we used the triangle inequality, Cauchy-Schwarz inequality and  \eqref{eq:unif_norm2}. Furthermore, 
	\eqref{eq:unif_norm2} and \eqref{eq:BarfStarBetterThanCausal} together with Proposition~\ref{prop:difference_to_causal_function} yield the following bound
	\begin{align}
		|L_2^i(\tilde{f},\bar f^*)|
		= &\, 2\big|\E_{M(i)}\big[(\tilde{f}(X)-f(X))(f(X)-\bar f^*(X))\big] \big|\nonumber\\
		=&\, 2 \sqrt{\E_{M(i)}\big[(\tilde{f}(X)-f(X))^2\big] \E_{M(i)}\big[(f(X)-\bar f^*(X))^2\big]}\nonumber\\
		\leq &\,  2\sqrt{4\delta^2K^2} \sqrt{4\var_M(\xi_Y)}\nonumber \\
		=  &\, 8\delta K \sqrt{\var_M(\xi_Y)}, \label{eq:estimate_L2}
	\end{align}
	for any $i\in\cI$. Thus, it suffices to construct an upper
	bound on the second term in the final expression in
	\eqref{eq:UpperboundOfL2TildefStarf}. Direct computation leads
	to
	\begin{align*}
		\E_{M(i)}\big[(Y-f^*(X))^2\big]   = &\, \E_{M(i)}\big[(Y-\bar{f}^*(X))^2\big] \\
		&\,  +\E_{M(i)}\big[(\bar{f}^*(X)-f^*(X))^2\big] \\
		&\,   +2\E_{M(i)}\big[(Y-\bar{f}^*(X))(\bar{f}^*(X)-f^*(X))\big].
	\end{align*}
	Rearranging the terms and applying the triangle inequality and Cauchy-Schwarz results in
	\begin{align*}
		\E_{M(i)}&\, \big[(\bar{f}^*(X)-f^*(X))^2\big] \\
		=&\,\E_{M(i)}\big[(Y-f^*(X))^2\big]-\E_{M(i)}\big[(Y-\bar{f}^*(X))^2\big]\\
		& \,  -2\E_{M(i)}\big[(Y-\bar{f}^*(X))(\bar{f}^*(X)-f^*(X))\big]\\
		\leq &\,\big|\E_{M(i)}\big[(Y-f^*(X))^2\big]- \inf_{\fs\in\FF}\,
		\sup_{i\in\II}\E_{M(i)}\big[(Y-\fs(X))^2\big]\big| \\
		& \, + \big| \inf_{\fs\in\FF}\,
		\sup_{i\in\II}\E_{M(i)}\big[(Y-\fs(X))^2\big] -\E_{M(i)}\big[(Y-\bar{f}^*(X))^2\big] \big| \\
		&  \, + 2\E_{M(i)}\big|(Y-\bar{f}^*(X))(\bar{f}^*(X)-f^*(X))\big|\\
		\leq&\, 2\epsilon
		+2\sqrt{\E_{M(i)}\big[(Y-\bar{f}^*(X))^2\big]}\sqrt{\E_{M(i)}\big[(\bar{f}^*(X)-f^*(X))^2\big]}\\
		\leq &\, 2\epsilon+2\sqrt{\var_{M}(\xi_Y)}\sqrt{\E_{M(i)}\big[(\bar{f}^*(X)-f^*(X))^2\big]},
	\end{align*}
	for any $i\in \cI$. Here, we used that both $f^*$ and $\bar f^*$ are $\ep$-minimax solutions with respect to $M$ and that $\bar f^*$ satisfies \eqref{eq:BarfStarBetterThanCausal} which implies that
	\begin{align*}
		&\, \E_{M(i)}\big[(Y-\bar{f}^*(X))^2)\big]
		\leq\sup_{i\in\II}\E_{M(i)}\big[(Y-f(X))^2)\big] =  \sup_{i\in\II} \E_{M(i)}\big[\xi_Y^2\big] = \var_{M}(\xi_Y),
	\end{align*}
	for any $i\in \cI$, as $\xi_Y$ is unaffected by an intervention on $X$.
	Thus, $\E_{M(i)}\big[(\bar{f}^*(X)-f^*(X))^2\big]$ must satisfy $\ell(\E_{M(i)}\big[(\bar{f}^*(X)-f^*(X))^2\big])\leq 0$, where $\ell:[0,\infty)\to \R$ is given by $\ell(z)=z-2\ep-2\sqrt{\var_{M}(\xi_Y)} \sqrt{z}$.
	The linear term of $\ell$ grows faster than the square
	root term, so the largest allowed value of $\E_{M(i)}\big[(\bar{f}^*(X)-f^*(X))^2\big]$  coincides with the largest root of $\ell(z)$. The largest root is given by
	\begin{align*}
		C^2:=2\ep +2\var_{M}(\xi_Y) + 2\sqrt{\var_{M}(\xi_Y)^2+2\ep\var_{M}(\xi_Y)},
	\end{align*}
	where $(\cdot)^2$ refers to the square of $C$. Hence, for any
	$i\in \cI$ it holds that
	\begin{equation} \label{eq:LastFactorUpperBound}
		\E_{M(i)}\big[(\bar{f}^*(X)-f^*(X))^2\big]\leq C^2.
	\end{equation}
	Hence by \eqref{eq:UpperboundOfL2TildefStarf}, \eqref{eq:estimate_L2} and  \eqref{eq:LastFactorUpperBound} we have that the following upper bound is valid for any $i\in \cI$.
	\begin{align} 
		\abs[\big]{L_2^i(\tilde{f},f^*)} &\leq  8\delta K \sqrt{\var_M(\xi_Y)} + 4\delta K C \label{eq:estimate_L2_fstar}.
	\end{align}
	Thus, using \eqref{eq:QuantityForUpperBound} with
	\eqref{eq:estimate_L1}, \eqref{eq:estimate_L3} and
	\eqref{eq:estimate_L2_fstar}, we get the following upper bound
	\begin{align} \notag
		&\sup_{i\in\II} \E_{\tilde{M}(i)}\big[(Y-f^*(X))^2\big] \\
		&\quad  \leq \sup_{i\in\II}\E_{M(i)}\big[(Y-f^*(X))^2\big] + 4\delta^2K^2+4\delta K C+ 12\delta K \sqrt{\var_M(\xi_Y)}.\label{eq:upperboundsup}
	\end{align}	
	Finally, by combining the bounds \eqref{eq:lowerboundoninf}
	and \eqref{eq:upperboundsup} together with
	\eqref{eq:fstarDiffEpsilon} we get that
	\begin{align}
		\bigg|\sup_{i\in\II}&\,\E_{\tilde{M}(i)} \big[(Y-f^*(X))^2\big]
		-
		\inf_{\fs\in\FF}\,
		\sup_{i\in\II}\E_{\tilde{M}(i)}\big[(Y-\fs(X))^2\big] \bigg|\nonumber\\ \nonumber	
		\leq &\, \sup_{i\in\II}\E_{M(i)}\big[(Y-f^*(X))^2\big] - 
		\inf_{\fs\in\FF}\,
		\sup_{i\in\II}\E_{M(i)}\big[(Y-\fs(X))^2\big]\nonumber\\ \nonumber
		&\, +  4\delta^2K^2+4\delta K C+ 12\delta K \sqrt{\var_M(\xi_Y)}\\ \nonumber
		& \,  + 8 \delta^2 K^2 + 12 \delta K \sqrt{\var_M(\xi_Y)} \\
		\leq &\,  \ep + 12 \delta^2 K^2 + 24 \delta K \sqrt{\var_M(\xi_Y)} + 4 \delta K C. \label{eq:CorrectUpperBound}
	\end{align}
	Using that all terms are positive, we get that
	\begin{align*}
		C &= \sqrt{\var_{M}(\xi_Y)} + \sqrt{\var_{M}(\xi_Y) + 2 \ep}
		\leq 2\sqrt{\var_{M}(\xi_Y)} + \sqrt{2 \ep}
	\end{align*}
	Hence, \eqref{eq:CorrectUpperBound} is bounded above by
	\begin{align*}
		\ep + 12 \delta^2 K^2 + 32 \delta K \sqrt{\var_M(\xi_Y)} +
		4 \sqrt{2} \delta K \sqrt{\ep}.
	\end{align*}
	This
	completes the proof of
	Proposition~\ref{prop:extrapolation_bounded_deriv}.
\end{proofenv}


\begin{proofenv}{\Cref{prop:impossibility_extrapolation}}
	Let $\bar{f}\in \FF$ and $c > 0$. By assumption, 
	$\II$ is a well-behaved set of
	support-extending interventions on $X$.
	Since $\supp_{\II}^{M}(X) \setminus \supp^{M}(X)$ has non-empty
	interior, there exists an intervention $i_0 \in \II$ and $\ep>0$
	such that $\P_{M(i_0)}(X\in B) \geq \epsilon$, 
	for some open subset
	$B \subsetneq \bar{B}$,   such that $\mathrm{dist}(B,\R^d \setminus \bar{B})>0$, where $\bar{B}:=\supp_{\II}^{M}(X) \setminus \supp^{M}(X)$.
	Let $\tilde{f}$ be any continuous 
	function satisfying that, for all $x \in B\cup(\R^d\setminus\bar B)$,
	\begin{align*}
		\tilde{f}(x) =
		\begin{cases}
			\bar{f}(x) + \gamma, \quad & x \in B\\
			f(x), \quad & x \in \R^d \setminus \bar{B},
		\end{cases}
	\end{align*}
	where
	$\gamma := \epsilon^{-1/2} \left\{(2 \E_{\tilde{M}}[\xi_{Y}^2] +
	c)^{1/2} + (\E_{\tilde{M}}[\xi_{Y}^2])^{1/2}\right\}$.
	
	Consider a secondary model
	$\tilde{M} = (\tilde{f}, g, h_1, h_2, Q)\in\cM$. Then, by
	Assumption~\ref{ass:identify_f}, it holds that
	$\P_{M} = \P_{\tilde{M}}$.  Since $\II$ only consists of
	interventions on $X$, it holds that
	$\P_{M(i_0)}(X\in B) = \P_{\tilde{M}(i_0)}(X\in B)$ (this
	holds since all components of $\tilde{M}$ and $M$ are equal, except for the
	function $f$, which is not allowed to enter in the intervention on
	$X$). Therefore,
	\begin{align}\label{eq:proof_imp_a}
		\E_{\tilde{M}(i_0)}\big[(Y-\bar{f}(X))^2\big]
		&\geq   \E_{\tilde{M}(i_0)}\big[(Y-\bar{f}(X))^2 \mathbbm{1}_{B}(X)\big] \nonumber\\
		&= \E_{\tilde{M}(i_0)}\big[(\gamma + \xi_Y)^2 \mathbbm{1}_{B}(X)\big] \nonumber\\
		&\geq   \gamma^2\epsilon + 2 \gamma \E_{\tilde{M}(i_0)}\big[\xi_Y
		\mathbbm{1}_{B}(X) \big]\nonumber\\
		&\geq  \gamma^2\epsilon - 2 \gamma \left(\E_{\tilde{M}}\big[\xi_Y^2\big]
		\epsilon\right)^{1/2} \nonumber\\
		&=  c + \E_{\tilde{M}}[\xi_{Y}^2],
	\end{align}
	where the third inequality follows from Cauchy–Schwarz.
	Further, by the definition of the infimum it holds that
	\begin{align}\label{eq:proof_imp_b}
		\begin{split}
			&\inf_{\fs \in \FF}\,\sup_{i\in\II} \E_{\tilde{M}(i)}\big[(Y-\fs(X))^2\big]
			\leq  \sup_{i\in\II}\E_{\tilde{M}(i)}\big[(Y-\tilde{f}(X))^2\big] 
			= \E_{\tilde{M}}[\xi_{Y}^2].
		\end{split}
	\end{align}
	Therefore, combining~\eqref{eq:proof_imp_a} and~\eqref{eq:proof_imp_b}, the 
	claim follows.
\end{proofenv}


\begin{proofenv}{\Cref{prop:genA}}
	We prove the result by showing that under
	Assumption~\ref{ass:identify_g} it is possible to express
	interventions on $A$ as confounding-preserving interventions on $X$
	and applying
	Propositions~\ref{prop:genX_intra}~and~\ref{prop:genX_extra}.  To
	avoid confusion, we will throughout this proof denote the true model
	by $M^0 = (f^0, g^0, h_1^0, h_2^0, Q^0)$.   Fix an intervention
	$i\in\II$. Since it is an intervention on $A$, there exist $\psi^i$
	and $I^i$ such that for any $M = (f,g,h_1, h_2, Q) \in \MM$, the
	intervened SCM $M(i)$ is of the form
	\begin{align*}
		A^i &:= \psi^i(I^i, \ep_A^i),  \quad H^i := \ep_H^i,\\
		X^i &:= g(A^i) + h_2(H^i, \ep_X^i), \\
		Y^i &:= f(X^i) + h_1(H^i,\ep_Y^i),
	\end{align*}
	where $(\ep^i_X, \ep^i_Y, \ep^i_A, \ep^i_H) \sim Q$.  We now define
	a confounding-preserving intervention $j$ on $X$, such that, for all
	models $\tilde{M}$ with $\P_{\tilde{M}} = \P_M$, the distribution of
	$(X,Y)$ under $\tilde{M}(j)$ coincides with that under
	$\tilde{M}(i)$. To that end, define the intervention function
	\begin{equation*}
		\bar{\psi}^j(h_2, A^j, H^j, \ep^j_X ,I^j) \coloneqq g^0(\psi^i(I^j, A^j)) +
		h_2(H^j, \ep_X^j),
	\end{equation*}
	where $g^0$ is the fixed function corresponding to model $M$, and
	therefore not an argument of $\bar{\psi}^j$.    Let now $j$ be
	the intervention on $X$ satisfying that, for all
	$M = (f,g,h_1, h_2, Q) \in \MM$, the intervened model $M(j)$ is
	given as
	\begin{align*}
		&A^{j} := \ep_A^{j}, \quad  H^{j} := \ep_H^{j}, \\
		&X^{j} := \bar{\psi}^j(h_2, A^{j}, H^{j}, \ep^{j}_X ,I^{j}), \\
		&Y^{j}:= f(X^{j}) + h_1(H^{j},\ep_Y^{j}),
	\end{align*}
	where $(\ep^j_X, \ep^j_Y, \ep^j_A, \ep^j_H) \sim Q$ and where $I^j$
	is chosen such that $I^j\eqd I^i$.  By definition, $j$ is
	a confounding-preserving intervention. Let now
	$\tilde{M} = (\tilde{f}, \tilde{g}, \tilde{h}_1, \tilde{h}_2,
	\tilde{Q})$ be such that $\P_{\tilde{M}} = \P_M$, and let
	$(\tilde{X}^i, \tilde{Y}^i)$ and $(\tilde{X}^j, \tilde{Y}^j)$ be
	generated under $\tilde{M}(i)$ and $\tilde{M}(j)$, respectively. By
	Assumption~\ref{ass:identify_g}, it holds for all
	$a \in \supp(A) \cup \supp_{\II}(A)$ that $\tilde{g}(a) = g^0(a)$.
	Hence, we get that
	\begin{align*}
		(\tilde{X}^i, \tilde{Y}^i)
		\eqd &
		(\tilde{g}(\psi^i(I^i, \tilde{\ep}_A^i)) + \tilde{h}_2(\tilde{\ep}^i_H, \tilde{\ep}_X^i),
		\tilde{f}(\tilde{g}(\psi^i(I^i, \tilde{\ep}_A^i)) \\ &+ \tilde{h}_2(\tilde{\ep}^i_H, \tilde{\ep}_X^i)) + \tilde{h}_1(\tilde{\ep}_H^i, \tilde{\ep}^i_Y)) \\
		= &
		(g^0(\psi^i(I^i, \tilde{\ep}_A^i)) + \tilde{h}_2(\tilde{\ep}^i_H, \tilde{\ep}_X^i),
		\tilde{f}(g^0(\psi^i(I^i, \tilde{\ep}_A^i)) \\&+ \tilde{h}_2(\tilde{\ep}^i_H, \tilde{\ep}_X^i)) + \tilde{h}_1(\tilde{\ep}_H^i, \tilde{\ep}^i_Y)) \\
		\eqd &
		(g^0(\psi^i(I^j, \tilde{\ep}_A^j)) + \tilde{h}_2(\tilde{\ep}^j_H, \tilde{\ep}_X^j),
		\tilde{f}(g^0(\psi^i(I^j, \tilde{\ep}_A^j))\\& + \tilde{h}_2(\tilde{\ep}^j_H, \tilde{\ep}_X^j)) + \tilde{h}_1(\tilde{\ep}_H^j, \tilde{\ep}^j_Y)) \\
		\eqd &
		(\bar{\psi}^j(\tilde{h}_2, \tilde{\ep}_A^j, \tilde{\ep}_H^j, \tilde{\ep}^j_X ,I^j), 
		\tilde{f}(\bar{\psi}^j(\tilde{h}_2, \tilde{\ep}_A^j, \tilde{\ep}_H^j, \tilde{\ep}^j_X ,I^j)) \\&+ \tilde{h}_1(\tilde{\ep}_H^j, \tilde{\ep}^j_Y)) \\
		\eqd &
		(\tilde{X}^j, \tilde{Y}^j),
	\end{align*}
	as desired. Since $i \in \II$ was arbitrary, we have now shown that
	there exists a mapping $\pi$ from $\II$ into a set $\mathcal{J}$ of
	confounding-preserving (and hence a well-behaved set) of
	interventions on $X$, such that for all $\tilde{M}$ with
	$\P_{\tilde{M}} = \P_M$,
	$\P^{(X,Y)}_{\tilde{M}(i)} = \P_{\tilde{M}(\pi(i))}^{(X,Y)}$. Hence,
	we can rewrite Equation~\eqref{eq:def_generalization} in
	Definition~\ref{defi:general} in terms of the set $\mathcal{J}$. The
	result now follows from
	Propositions~\ref{prop:genX_intra}~and~\ref{prop:genX_extra}.
\end{proofenv}


\begin{proofenv}{\Cref{prop:impossibility_intA}}
	Let $b \in \R^d$ be such that $f(x) = b^\top x$ for all
	$x \in \R^d$. We start by characterizing the error
	$\E_{\tilde{M}(i)}\big[(Y-\fs(X))^2\big]$. %
	Let us consider models of the form
	$\tilde{M} = (f, \tilde{g}, h_1, h_2, Q) \in \MM$ for some function
	$\tilde{g} \in \GG$ with $\tilde{g}(a) = g(a)$ for all
	$a \in \supp_M(A)$. Clearly, any such model satisfies that
	$\P_{\tilde{M}} = \P_M$. For every $a \in \mathcal{A}$, let
	$i_a \in \II$ denote the corresponding hard intervention on $A$.
	For every $a \in \mathcal{A}$ and $\bes \in \R^d$, we then have
	\begin{equation} \label{eq:linear_mse}
		\begin{aligned}
			&\ \E_{\tilde{M}(i_a)}\big[(Y - \bes^\top X)^2\big]  \\
			= &\ \E_{\tilde{M}(i_a)}\big[(b^\top X + \xi_Y - \bes^\top X)^2\big] \\
			= &\ (b - \bes)^\top \E_{\tilde{M}(i_a)}[X X^\top] (b - \bes) \\&
			+ 2(b - \bes)^\top \E_{\tilde{M}(i_a)}[X \xi_Y] + \E_{\tilde{M}(i_a)}\big[\xi_Y^2] \\
			= &\ (b - \bes)^\top \underbrace{(\tilde{g}(a) \tilde{g}(a)^\top + \E_{M}[\xi_X \xi_X^\top])}_{=: K_{\tilde{M}}(a)} (b - \bes) \\& + 2(b - \bes)^\top \E_{M}[\xi_X \xi_Y] + \E_{M}\big[\xi_Y^2],
		\end{aligned}
	\end{equation}
	where we have used that, under $i_a$, the distribution of 
	$(\xi_X, \xi_Y)$
	is unaffected.
	We now show that, for any $\tilde{M}$ with the above form, the causal
	function $f$ does not minimize the worst-case 
	risk
	across interventions
	in $\II$.  The idea is to show that the worst-case risk~\eqref{eq:linear_mse}
	strictly decreases at
	$\bes = b$ in the direction
	$u := \E_{M}[\xi_X \xi_Y] / \norm{\E_{M}[\xi_X \xi_Y]}_2$.  For every
	$a \in \mathcal{A}$ and $s \in \R$, define
	\begin{align*}
		\ell_{\tilde{M},a}(s) :&= \E_{\tilde{M}(i_a)}\big[(Y-(b + s u)^\top X)^2\big]
		\\&= u^\top K_{\tilde{M}}(a) u \cdot s^2  - 2 u^\top \E_{M}[\xi_X \xi_Y] \cdot s + \E_{M}\big[\xi_Y^2].
	\end{align*}
	For every $a$,
	$\ell_{\tilde{M},a}^\prime (0) = -2 \norm{\E_{M}[\xi_X \xi_Y]}_2 < 0$,
	showing that $\ell_{\tilde{M},a}$ is strictly decreasing at $s=0$ 
	(with a derivative that is bounded away from 0 across all $a \in \mathcal{A}$).
	By boundedness of $\mathcal{A}$ and by the continuity of $a \mapsto \ell_{\tilde{M},a}^{\prime \prime } (0)  = 2 u^\top K_{\tilde{M}}(a) u$, 
	it further follows that $\sup_{a \in \mathcal{A}} \card{\ell^{\prime \prime}_{\tilde{M},a} (0)} < \infty$.
	Hence, we can find $s_0 > 0$ such that for all $a \in \mathcal{A}$, $\ell_{\tilde{M}, a}(0) > \ell_{\tilde{M}, a}(s_0)$.  
	It now follows by continuity of $(a, s) \mapsto \ell_{\tilde{M},a}(s)$ that
	\begin{align*}
		\sup_{i \in \II} \E_{\tilde{M}(i)}\big[(Y-b^\top X)^2\big] &= \sup_{a \in \mathcal{A}} \ell_{\tilde{M}, a}(0) \\
		&> \sup_{a \in \mathcal{A}} \ell_{\tilde{M}, a}(s_0) \\
		&= \sup_{i \in \II} \E_{\tilde{M}(i)}\big[(Y-(b+s_0 u)^\top X)^2\big],
	\end{align*}
	showing that $b+s_0 u$ attains a lower worst-case risk than $b$. 
	
	We now show that all functions other than $f$ may result in an
	arbitrarily large error.  Let $\bar{b} \in \R^d \setminus \{ b \}$ be
	given, and let $j \in \{1, \dots, d\}$ be such that
	$b_j \neq \bar{b}_j$.  The idea is to construct a function
	$\tilde{g} \in \GG$ such that, under the corresponding model
	$\tilde{M} = (f, \tilde{g}, h_1, h_2, Q) \in \MM$, some hard
	interventions on $A$ result in strong shifts of the $j$th coordinate
	of $X$.  Let $a \in \mathcal{A}$. Let $e_j \in \R^d$ denote the $j$th unit vector, and assume
	that $\tilde{g}(a) = n e_j$ for some $n \in \N$.  Using
	\eqref{eq:linear_mse}, it follows that
	\begin{align*}
		\E_{\tilde{M}(i_a)}\big[(Y-\bar{b}^\top X)^2\big] 
		 =& n^2 (\bar{b}_j - b_j)^2 + (\bar{b} - b)^\top \E_{M}[\xi_X \xi_X^\top] (\bar{b} - b)  \\&+ 
		2
		(\bar{b} - b)^\top \E_{M}[\xi_X \xi_Y] + \E_{M}\big[\xi_Y^2].
	\end{align*}
	By letting $n \to \infty$, we see that the above error may become
	arbitrarily large. Given any $c > 0$, we can therefore construct
	$\tilde{g}$ such that
	$\E_{\tilde{M}(i_a)}\big[(Y-\bar{b}^\top X)^2\big] \geq c
	+\E_{M}\big[\xi_Y^2]$.  By 
	carefully
	choosing
	$a \in \text{int}(\mathcal{A} \setminus \supp_M(A))$, this can be done
	such that $\tilde{g}$ is continuous and $\tilde{g}(a) = g(a)$
	for all $a \in \supp_M(A)$, ensuring that $\bP_{\tilde M}= \bP_M$. It follows that
	\begin{align*}
		c 	&\leq \E_{\tilde{M}(i_a)}\big[(Y-\beb^\top X)^2\big]  - \E_{M}\big[\xi_Y^2] \\
		&= \E_{\tilde{M}(i_a)}\big[(Y-\beb^\top X)^2\big]  -  \sup_{i \in \II}\E_{\tilde{M}(i)} \big[(Y- b^\top X)^2] \\
		&\leq \E_{\tilde{M}(i_a)}\big[(Y-\beb^\top X)^2\big]  - \inf_{\bes \in \R^d} \sup_{i \in \II}\E_{\tilde{M}(i)} \big[(Y-\bes^\top X)^2]  \\
		&\leq \sup_{i \in \II} \E_{\tilde{M}(i)}\big[(Y-\beb^\top X)^2\big]
		- \inf_{\bes \in \R^d} \sup_{i \in \II}\E_{\tilde{M}(i)} \big[(Y-\bes^\top X)^2],
	\end{align*}
	which completes the proof of Proposition~\ref{prop:impossibility_intA}.
\end{proofenv}


\begin{proofenv}{\Cref{thm:consis}}
	By assumption, $ \cI$ is a set of 
	interventions on $X$ or $A$ of which at least one is confounding-removing. 
	Now fix any
	$$
	\tilde M=(f_{\eta_0}(x;\tilde{\theta}),\tilde g,\tilde h_1,\tilde h_2,\tilde Q)\in \cM,
	$$ 
	with $\bP_{M}=\bP_{\tilde M}$.
	By Proposition~\ref{prop:minimax_equal_causal}, we have that a minimax solution is given by the causal function. That is,
	\begin{align*}
		\inf_{\fs\in\FF_{\eta_0}}\,
		\sup_{i\in \II}\E_{\tilde{M}(i)}\big[(Y-\fs(X))^2\big]
		&=\sup_{i\in \II}\E_{\tilde{M}(i)}\big[(Y-
		f_{\eta_0}(X;\tilde{\theta})
		)^2\big] \\&= \E_{M}[\xi^2_Y],
	\end{align*}
	where we used that $\xi_Y$ is unaffected by an intervention on $X$.
	By the support restriction $\supp^M(X) \subseteq (a,b)$ we know that
	\begin{align*}
		f_{\eta_0}(x;\theta^0)&=B(x)^\top \theta^0, \\
		f_{\eta_0}(x;\tilde \theta)&=B(x)^\top \tilde \theta, \\ f_{\eta_0}(x;\hat \theta_{\lambda^\star_n,\eta_0,\mu}^n)&=B(x)^\top  \hat \theta_{\lambda^\star_n,\eta_0,\mu}^n,
	\end{align*}
	for all $x\in \supp^M(X)$. Furthermore, as $Y=B(X)^\top \theta^0+\xi_Y$ $\bP_{M}$-almost surely, we have that
	\begin{align} \notag
		\E_M\left[ C(A) Y \right] &= \E_M\left[ C(A) B(X)^\top \theta^0\right] + \E_M\left[C(A) \xi_Y \right] \\
		&= \E_M\left[ C(A) B(X)^\top \right]\theta^0, \label{eq:ExpansionOfECY_0}
	\end{align}
	where we used the assumptions that $\E\left[ \xi_Y \right] =0$ and $A \independent \xi_Y$ by the exogeneity of $A$. Similarly,
	\begin{align*} %
		\E_{\tilde M}\left[ C(A) Y \right] = \E_{\tilde M}\left[ C(A) B(X)^\top \right]\tilde\theta.
	\end{align*}
	As $\bP_M = \bP_{\tilde M}$, we have that $
	\E_M[ C(A) Y ] = \E_{\tilde M}[ C(A) Y ]$ and $\E_M[ C(A) B(X)^\top ]= \E_{\tilde{M}}[ C(A) B(X)^\top ]$, hence 
	\begin{align*}
		&\E_{ M}\left[ C(A) B(X)^\top \right]\tilde\theta = \E_{M}\left[ C(A) B(X)^\top \right]\theta^0
		\iff \tilde \theta = \theta^0,
	\end{align*}
	by assumption \ref{ass:RankCondition},
	which states that $ \E[ C(A) B(X)^\top ]$ is of full rank
	(bijective). 
	In other words, the causal function parameterized by
	$\theta^0$ is identified from the observational distribution. Assumptions~\ref{ass:identify_f}~and~\ref{ass:gen_f} are therefore satisfied.
	Furthermore, we also have that
	\begin{align*}
		\sup_{i\in  \II}&\E_{\tilde{M}(i)}\big[(Y-f_{\eta_0}(X;\hat{\theta}^n_{\lambda^\star_n,\eta_0,\mu}))^2\big] \\
		= &\, \sup_{i\in\cI} \big\{ \E_{\tilde{M}(i)} \big[(f_{\eta_0}(X;\theta^0)-f_{\eta_0}(X;\hat{\theta}^n_{\lambda^\star_n,\eta_0,\mu}))^2\big]
		+  \E_{\tilde{M}(i)}\big[\xi_Y^2\big] \\
		&\quad   + 2 \E_{\tilde{M}(i)}\big[\xi_Y (f_{\eta_0}(X;\theta^0)-f_{\eta_0}(X;\hat{\theta}^n_{\lambda^\star_n,\eta_0,\mu})) \big]  \big\} \\
		\leq &\,  \sup_{i\in\cI} \big\{ \E_{\tilde{M}(i)} \big[(f_{\eta_0}(X;\theta^0)-f_{\eta_0}(X;\hat{\theta}^n_{\lambda^\star_n,\eta_0,\mu}))^2\big] 
		+  \E_{\tilde{M}(i)}\big[\xi_Y^2\big] \\
		&\quad + 2 \sqrt{ \E_{\tilde{M}(i)}\big[\xi_Y^2 \big] \E_{\tilde{M}(i)} \big[(f_{\eta_0}(X;\theta^0)-f_{\eta_0}(X;\hat{\theta}^n_{\lambda^\star_n,\eta_0,\mu}))^2 \big] } \big\} \\
		\leq &\, \sup_{i\in\cI}  \E_{\tilde{M}(i)} \big[(f_{\eta_0}(X;\theta^0)-f_{\eta_0}(X;\hat{\theta}^n_{\lambda^\star_n,\eta_0,\mu}))^2\big] +  \E_{M}\big[\xi_Y^2\big] \\
		&\quad    + 2 \sqrt{ \E_{M}\big[\xi_Y^2 \big] \sup_{i\in\cI}  \E_{\tilde{M}(i)} \big[(f_{\eta_0}(X;\theta^0)-f_{\eta_0}(X;\hat{\theta}^n_{\lambda^\star_n,\eta_0,\mu}))^2 \big] },
	\end{align*}
	by Cauchy-Schwarz inequality, where we additionally used that
	$ \E_{\tilde M(i)} [ \xi_Y^2 ] = \E_{M} [ \xi_Y^2 ]$ as $\xi_Y$ is
	unaffected by interventions on $X$. Thus,
	\begin{align*}
		\big\vert \sup_{i\in\II}&\E_{\tilde{M}(i)}\big[(Y-f_{\eta_0}(X;\hat{\theta}^n_{\lambda^\star_n,\eta_0,\mu}))^2\big]
		- \inf_{\fs\in\FF_{\eta_0}}\,
		\sup_{i\in\II}\E_{\tilde{M}(i)}\big[(Y-\fs(X))^2\big] \big\vert \\
		&\leq  \sup_{i\in\cI}  \E_{\tilde{M}(i)} \big[(f_{\eta_0}(X;\theta^0)-f_{\eta_0}(X;\hat{\theta}^n_{\lambda^\star_n,\eta_0,\mu}))^2\big]  \\
		&\qquad  + 2 \sqrt{ \E_{M}\big[\xi_Y^2 \big] \sup_{i\in\cI}  \E_{\tilde{M}(i)} \big[(f_{\eta_0}(X;\theta^0)-f_{\eta_0}(X;\hat{\theta}^n_{\lambda^\star_n,\eta_0,\mu}))^2 \big] } .
	\end{align*}
	For the next few derivations let $\hat \theta =\hat{\theta}^n_{\lambda^\star_n,\eta_0,\mu}$ for notational simplicity. Note that, for all $x \in \R$,
	\begin{align*}
		(f_{\eta_0}(x;\theta^0)-f_{\eta_0}(x;\hat \theta ))^2
		\leq &\, (\theta^0 -\hat \theta)^\top B(x)B(x)^\top (\theta^0 -\hat \theta) \\
		&+ (B(a)^\top (\theta^0-\hat \theta ) + B'(a)^\top (\theta^0-\hat \theta )(x-a))^2 \\
		&+ (B(b)^\top (\theta^0-\hat \theta ) + B'(b)^\top (\theta^0-\hat \theta )(x-b))^2.
	\end{align*}
	The second term has the following upper bound
	\begin{align*}
		(B(&a)^\top (\theta^0-\hat \theta ) + B'(a)^\top (\theta^0-\hat \theta )(x-a))^2 \\
		=&\, (\theta^0 -\hat \theta)^\top B(a)B(a)^\top (\theta^0 -\hat \theta) \\
		&  +  (x-a)^2(\theta^0 -\hat \theta)^\top B'(a)B'(a)^\top (\theta^0 -\hat \theta)  \\
		&  + 2(x-a) (\theta^0 -\hat \theta)^\top B'(a)B(a)^\top (\theta^0 -\hat \theta) \\
		\leq &\, \lambda_{\mathrm{m}}(B(a)B(a)^\top) \|\theta^0 - \hat \theta\|_2^2 \\
		&  + (x-a)^2 \lambda_{\mathrm{m}}(B'(a)B'(a)^\top ) \|\theta^0 - \hat \theta\|_2^2 \\
		&  + (x-a) \lambda_{\mathrm{m}}(B'(a)B(a)^\top+B(a)B'(a)^\top  ) \|\theta^0 - \hat \theta\|_2^2,
	\end{align*}
	where $\lambda_{\mathrm{m}}$ denotes the maximum eigenvalue. An analogous upper bound can be constructed for the third term. 
	Thus, by combining these two upper bounds with a similar upper bound for the first term, we arrive at
	\begin{align*}
		\E&_{\tilde{M}(i)} \big[(f_{\eta_0}(X;\theta^0)-f_{\eta_0}(X;\hat{\theta}))^2\big] \\
		\leq &\, \lambda_{\mathrm{m}}(\E_{\tilde{M}(i)} [B(X)B(X)^\top])\|\theta^0 -\hat \theta\|_2^2 \\
		& + \lambda_{\mathrm{m}}(B(a)B(a)^\top) \|\theta^0 - \hat \theta\|_2^2 \\
		&  + \E_{\tilde{M}(i)} [(X-a)^2] \lambda_{\mathrm{m}}(B'(a)B'(a)^\top ) \|\theta^0 - \hat \theta\|_2^2 \\
		&  + \E_{\tilde{M}(i)} [X-a] \lambda_{\mathrm{m}}(B'(a)B(a)^\top +B(a)B'(a)^\top  ) \|\theta^0 - \hat \theta\|_2^2\\
		& + \lambda_{\mathrm{m}}(B(b)B(b)^\top) \|\theta^0 - \hat \theta\|_2^2 \\
		&  + \E_{\tilde{M}(i)} [(X-b)^2] \lambda_{\mathrm{m}}(B'(b)B'(b)^\top ) \|\theta^0 - \hat \theta\|_2^2 \\
		&  + \E_{\tilde{M}(i)} [X-b] \lambda_{\mathrm{m}}(B'(b)B(b)^\top  +B(b)B'(b)^\top   ) \|\theta^0 - \hat \theta\|_2^2.
	\end{align*}
	Assumption \ref{ass:MaximumEigenValueBounded}
	imposes that $\sup_{i\in\cI}\E_{\tilde{M}(i)} [X^2]$ 
	and $\sup_{i\in\cI}\lambda_{\mathrm{m}}(\E_{\tilde{M}(i)} [B(X)B(X)^\top])$ are finite. Hence, the supremum of each of 
	the above terms is finite. That is, there exists a constant $c>0$ such that 
	
	\begin{align*}
		&\left\vert \sup_{i\in\II}\E_{\tilde{M}(i)}\big[(Y-f_{\eta_0}(X;\hat{\theta}^n_{\lambda^\star_n,\eta_0,\mu}))^2\big] 
		- \inf_{\fs\in\FF_{\eta_0}}\,
		\sup_{i\in\II}\E_{\tilde{M}(i)}\big[(Y-\fs(X))^2\big] \right\vert \\
		&\quad \leq  c \|\theta^0 - \hat{\theta}^n_{\lambda^\star_n,\eta_0,\mu} \|_2^2 + 2 \sqrt{ \E_{M}\big[\xi_Y^2 \big] c }\|\theta^0 - \hat{\theta}^n_{\lambda^\star_n,\eta_0,\mu} \|_2 .
	\end{align*}
	It therefore suffices to show that
	\begin{align*}
		\hat{\theta}^n_{\lambda^\star_n,\eta_0,\mu} \underset{n\to\infty}{\stackrel{P}{\longrightarrow}} \theta^0,
	\end{align*}
	with respect to the distribution induced by $M$. To simplify
	notation, we henceforth drop the $M$ subscript in the expectations
	and probabilities.
	Note that by the rank conditions in \ref{ass:RankCondition}, and the law of large numbers, we may assume that the corresponding sample product moments satisfy the same conditions. That is, for the 
	purpose of the following arguments, it suffices that the sample product moment only satisfies these rank 
	conditions asymptotically with probability one. 
	
	Let $B:= B(X)$, $C:= C(A)$, let
	$\fB$ and $\fC$ be row-wise stacked i.i.d. copies of $B(X)^\top$ and
	$C(A)^\top $, and recall the definition
	$\fP_\delta := \fC \left( \fC^\top \fC + \delta \fM \right)^{-1}
	\fC^\top$. By convexity of the objective function we can find a closed form expression for our estimator of $\theta^0$ by solving the corresponding normal equations. The closed form expression is given by
	\begin{align*}
		\hat \theta^n_{\lambda, \eta, \mu}
		: &= \argmin_{\theta \in \R^{k}} 
		\norm{\B{Y} - \B{B} \theta }_2^2 + \lambda \norm{\B{P}_\delta(\B{Y} - \B{B} \theta)}_2^2 + \gamma \theta^\top \B{K} \theta \\
		&=  \left(  \frac{\fB^\top  \fB}{n}   + \lambda^\star_n  \frac{\fB^\top \fP_\delta  \fP_\delta   \fB}{n} + \frac{\gamma \fK}{n} \right)^{-1} 
		\left( \frac{\fB^\top \fY }{n} + \lambda^\star_n \frac{\fB^\top \fP_\delta  \fP_\delta   \fY}{n} \right),
	\end{align*}
	where we used that $\lambda^\star_n \in [0,\infty)$ almost surely by \ref{ass:LambdaStarAlmostSurelyFinite}. 
	Consequently
	(using standard convergence arguments and that $n^{-1} \gamma \fK $ and  $n^{-1} \delta \fM$ converges to zero in 
	probability),
	if $\lambda^\star_n$ diverges to infinity in probability as $n$ tends to infinity, then
	\begin{align*}
		\hat{\theta}^n_{\lambda^\star_n,\eta_0,\mu} %
		&\stackrel{P}{\to} \left( \E\left[ BC^\top   \right]\E\left[ CC^\top   \right]^{-1}\E\left[ CB^\top   \right] \right)^{-1}
		\E \left[ BC^\top \right] \E\left[ CC^\top   \right]^{-1} \E\left[ C Y   \right]  \\
		&= \theta^0.
	\end{align*}
	Here, we also used that the terms multiplied by $\lambda^\star_n$ are the only asymptotically relevant terms. 
	These are the standard arguments that the K-class estimator (with minor penalized regression modifications) is consistent as long as the parameter $\lambda^\star_n$ converges to infinity, or, equivalently, $\kappa_n^\star= \lambda^\star_n/(1+\lambda^\star_n)$ converges to one in probability.
	
	We now consider two cases: \textit{(i)} $\E[B\xi_Y]\not =0$ and \textit{(ii)} $\E[B\xi_Y]=0$, corresponding to the case with unmeasured confounding and without, respectively. For \textit{(i)} we show that $\lambda^\star_n$ converges to infinity in probability and for \textit{(ii)} we show consistency by other means (as $\lambda^\star_n$ might not converge to infinity in this case).
	
	\textbf{Case (i):} The confounded case $\E[B\xi_Y]\not =0$. It suffices to show that 
	$$
	\lambda^\star_n := \inf\{\lambda\geq 0 : T_n(\hat \theta^n
	_{\lambda,\eta_{0},\mu})\leq q(\alpha)\}\underset{n\to\infty}{\stackrel{P}{\longrightarrow}} \infty.
	$$
	To that end, note that for fixed $\lambda \geq 0$ we have
	that
	\begin{align} \label{eq:ThetaLambdaConsistentEstimator}
		\hat{\theta}^n_{\lambda,\eta_0,\mu} & \underset{n\to\infty}{\stackrel{P}{\longrightarrow}} \theta_\lambda,
	\end{align}
	where
	\begin{align} \notag
		\theta_\lambda \, &:= \left( \E\left[ BB^\top \right] + \lambda \E\left[ BC^\top   \right]\E\left[ CC^\top   \right]^{-1}\E\left[ CB^\top   \right] \right)^{-1} \\ 
		& \quad  \times \left( \E \left[ B Y  \right]  + \lambda \E \left[ BC^\top \right] \E\left[ CC^\top   \right]^{-1} \E\left[ C Y   \right] \right).  \label{eq:ThetaLambdaFullRepresentation}
	\end{align}
	Recall that \eqref{eq:ExpansionOfECY_0} states that $
	\E\left[ C Y \right] = \E\left[ C B^\top \right]\theta^0$.
	Using \eqref{eq:ExpansionOfECY_0} and that $Y=B^\top \theta^0 + \xi_Y$ $\bP_{M}$-almost surely, we have that the latter factor of \eqref{eq:ThetaLambdaFullRepresentation} is given by
	\begin{align*}
		&\E \left[ B Y  \right]  + \lambda \E \left[ BC^\top \right] \E\left[ CC^\top   \right]^{-1} \E\left[ C Y   \right] \\
		& \quad = \E \left[ B B^\top   \right]\theta^0 + \E \left[ B \xi_Y   \right] 
		+ \lambda \E \left[ BC^\top \right] \E\left[ CC^\top   \right]^{-1} \E\left[ C B^\top    \right] \theta^0 \\
		&\quad = \left( \E \left[ B B^\top   \right] + \lambda \E \left[ BC^\top \right] \E\left[ CC^\top   \right]^{-1} \E\left[ C B^\top    \right] \right)\theta^0  
		+  \E \left[ B \xi_Y   \right].
	\end{align*}
	Inserting this into  \eqref{eq:ThetaLambdaFullRepresentation} we arrive at the following representation of $\theta_\lambda$
	\begin{align} \label{eq:ThetaLambdaInTermsOfTrueTheta}
		&\theta_\lambda  = \theta^0 
		+  \left( \E\left[ BB^\top \right] + \lambda \E\left[ BC^\top   \right]\E\left[ CC^\top   \right]^{-1}\E\left[ CB^\top   \right] \right)^{-1}
		\E \left[ B \xi_Y  \right].
	\end{align}
	Since $\E \left[ B \xi_Y  \right]\not =0$ by assumption, the above yields that
	\begin{align} \label{eq:ThetaTrueNotEqualThetaLambda}
		\forall \lambda \geq 0 : \quad \quad \theta^0 \not = \theta_\lambda.
	\end{align}
	Now we prove that $\lambda^\star_n$ diverges to infinity in probability as $n$ tends to infinity. That is, for any $\lambda \geq 0$ we will prove that
	\begin{align*}
		\lim_{n \to \infty }\bP (\lambda^\star_n \leq \lambda ) =0. 
	\end{align*}
	We fix an arbitrary $\lambda\geq 0$. By
	\eqref{eq:ThetaTrueNotEqualThetaLambda} we have that
	$\theta^0 \not = \theta_{\lambda}$. This implies that there exists an
	$\ep>0$ such that $\theta^0\not \in \overline{B(\theta_{\lambda},\ep)}$, where $\overline{B(\theta_{\lambda},\ep)}$ is the 
	closed ball in
	$\R^k$ with center $\theta_{\lambda}$ and radius $\ep$. 
	By the consistency result \eqref{eq:ThetaLambdaConsistentEstimator}, we know that 
	the sequence of events $(A_n)_{n\in \N}$, for every $n \in \N$, given by 
	$$A_n:= (|\hat \theta_{\lambda,\eta_0,\mu}^n -\theta_{\lambda}|\leq \ep) = (\hat \theta_{\lambda,\eta_0,\mu}^n \in\overline{B(\theta_{\lambda},\ep)}),$$ 
	satisfies $\bP(A_n)\to 1$ as $n\to \infty$. 
	By assumption \ref{ass:MonotonicityAndContinuityOfTest} we have that
	\begin{align*}
		\tilde \lambda\mapsto T_n(\theta^n_{\tilde \lambda, \eta_0,\mu} ), \qquad \text{and} \qquad \theta \mapsto T_n(\theta ),
	\end{align*}
	are weakly decreasing and continuous, respectively.  
	Together with the continuity of $\tilde \lambda \mapsto \hat \theta_{\tilde \lambda,\eta_0,\mu }^n$,
	this implies that also the mapping $\tilde \lambda \mapsto T_n(\hat \theta_{\tilde \lambda,\eta_0,\mu }^n)$
	is continuous. It now follows from Assumption~\ref{ass:LambdaStarAlmostSurelyFinite} 
	(stating that $\lambda^\star_n$ is almost surely finite) that for all $n \in \N$, 
	$\bP(T_{n}( \hat \theta^{n}_{\lambda^\star_{n},\eta_0,\mu} ) \leq q(\alpha))=1$. Furthermore, since $\tilde \lambda\mapsto T_n(\theta^n_{\tilde \lambda, \eta_0,\mu} )$
	is weakly decreasing, it follows that
	\begin{align*}
	\bP (\lambda^\star_{n} \leq \lambda )
		= &\, \bP( \{\lambda^\star_{n} \leq \lambda \} \cap \{T_{n}( \hat \theta^{n}_{\lambda^\star_{n}, \eta_0,\mu} ) \leq q(\alpha)\} )  \\
		\leq &\, \bP( \{\lambda^\star_{n} \leq \lambda \} \cap \{T_{n}( \hat \theta^{n}_{\lambda, \eta_0,\mu} ) \leq q(\alpha) \}) \\
		= &\, \bP( \{\lambda^\star_{n} \leq \lambda \}\cap \{T_{n}( \hat \theta^{n}_{\lambda, \eta_0,\mu} ) \leq q(\alpha)\} \cap A_{n}) \\
		&\qquad + \, \, \bP( \{\lambda^\star_{n} \leq \lambda \} \cap \{T_{n}( \hat \theta^{n}_{\lambda, \eta_0,\mu} ) \leq q(\alpha)\} \cap A_{n}^c) \\
		\leq &\,  \bP( \{\lambda^\star_{n} \leq \lambda \} \cap  \{ T_{n}( \hat \theta^{n}_{\lambda, \eta_0,\mu} ) \leq q(\alpha)\} 
		\cap \{ |\hat \theta_{\lambda,\eta_0,\mu}^n -\theta_{\lambda}|\leq \ep\} )
		\\
		&\qquad+\, \, \bP(A_n^c).
	\end{align*}
	It now suffices to show that the first term converges to zero, since $\bP(A_{n}^c)\to 0$ as $n\to \infty$. 
	We have
	\begin{align*}
		&\, \bP( \{\lambda^\star_{n} \leq \lambda \} \cap \{ T_{n}( \hat \theta^{n}_{\lambda, \eta_0,\mu} ) \leq q(\alpha)\}
		\cap \{ |\hat \theta_{\lambda,\eta_0,\mu}^n -\theta_{\lambda}|\leq\ep\} ) \\
		&\qquad \leq \bP \Big( \{ \lambda^\star_{n} \leq \lambda \} \cap \Big\{ \inf_{\theta\in \overline{B(\theta_{\lambda},\ep)}}T_{n}( \theta ) \leq q(\alpha) \Big\}
		\cap \{ |\hat \theta_{\lambda,\eta_0,\mu}^n -\theta_{\lambda}|\leq\ep\} \Big)  \\
		&\qquad \leq \bP \Big(  \inf_{\theta\in \overline{B(\theta_{\lambda},\ep)}}T_{n}( \theta ) \leq q(\alpha) \Big)\\ 
		&\qquad \stackrel{P}{\to}0, 
	\end{align*}
	as $n\to \infty$, since $\overline{B(\theta_{\lambda},\ep)}$ is a compact set not containing $\theta^0$. Here, we used that the test statistic $(T_n)$ is assumed to have compact uniform power \ref{ass:ConsistentTestStatistic}.
	Hence, $\lim_{n\to \infty} \bP (\lambda^\star_{n} \leq \lambda ) = 0 $ for any $\lambda \geq 0$,
	proving that $\lambda^\star_n$ diverges to infinity in probability, which ensures consistency. 
	
	\textbf{Case (ii):} the unconfounded case $\E[B(X)\xi_Y]=0$. Recall that
	\begin{align} \notag
		\hat{\theta}^n_{\lambda,\eta_0,\mu} \,&= \argmin_{\theta \in \R^{k}}  
		\norm{\B{Y} - \B{B} \theta }_2^2 + \lambda \norm{\B{P}_\delta(\B{Y} - \B{B} \theta)}_2^2 + \gamma \theta^\top \B{K} \theta  \\ \label{eq:ThetaLambdaMinimizesObjectiveFunction}
		&=\argmin_{\theta \in \R^{k}}   l_{\text{OLS}}^n(\theta) + \lambda l_{\text{TSLS}}^n(\theta) + \gamma l_{\text{PEN}}(\theta) ,
	\end{align}
	where we defined $l_{\text{OLS}}^n(\theta):= n^{-1}\norm{\B{Y} - \B{B} \theta }_2^2$, $l_{\text{TSLS}}^n(\theta) := n^{-1}\norm{\B{P}_\delta(\B{Y} - \B{B} \theta)}_2^2$, and $l_{\text{PEN}}(\theta) := n^{-1} \theta^\top \B{K} \theta$. 
	For any $0\leq \lambda_1 < \lambda_2$ we have
	\begin{align*}
		&l_{\text{OLS}}^n(\hat{\theta}^n_{\lambda_1,\eta_0,\mu}) + \lambda_1 l_{\text{TSLS}}^n(\hat{\theta}^n_{\lambda_1,\eta_0,\mu}) + \gamma l_{\text{PEN}}(\hat{\theta}^n_{\lambda_1,\eta_0,\mu})   \\
		&\quad \leq l_{\text{OLS}}^n(\hat{\theta}^n_{\lambda_2,\eta_0,\mu}) + \lambda_1 l_{\text{TSLS}}^n(\hat{\theta}^n_{\lambda_2,\eta_0,\mu})  + \gamma l_{\text{PEN}}(\hat{\theta}^n_{\lambda_2,\eta_0,\mu}) \\
		&\quad = l_{\text{OLS}}^n(\hat{\theta}^n_{\lambda_2,\eta_0,\mu}) + \lambda_2 l_{\text{TSLS}}^n(\hat{\theta}^n_{\lambda_2,\eta_0,\mu}) + \gamma l_{\text{PEN}}(\hat{\theta}^n_{\lambda_2,\eta_0,\mu}) 
		\\
		& \qquad+ (\lambda_1-\lambda_2) l_{\text{TSLS}}^n(\hat{\theta}^n_{\lambda_2,\eta_0,\mu}) \\
		&\quad \leq l_{\text{OLS}}^n(\hat{\theta}^n_{\lambda_1,\eta_0,\mu}) + \lambda_2 l_{\text{TSLS}}^n(\hat{\theta}^n_{\lambda_1,\eta_0,\mu}) + \gamma l_{\text{PEN}}(\hat{\theta}^n_{\lambda_1,\eta_0,\mu})
		\\ & \qquad + (\lambda_1-\lambda_2) l_{\text{TSLS}}^n(\hat{\theta}^n_{\lambda_2,\eta_0,\mu}),
	\end{align*}
	where we used \eqref{eq:ThetaLambdaMinimizesObjectiveFunction}. Rearranging this inequality and dividing by $(\lambda_1 - \lambda_2)$ yields 
	\begin{align*}
		l_{\text{TSLS}}^n(\hat{\theta}^n_{\lambda_1,\eta_0,\mu}) \geq   l_{\text{TSLS}}^n(\hat{\theta}^n_{\lambda_2,\eta_0,\mu}),
	\end{align*}
	proving that $\lambda \mapsto l_{\text{TSLS}}^n(\hat{\theta}^n_{\lambda,\eta_0,\mu})$ is weakly decreasing. Thus, since $\lambda^\star_n \geq 0$ almost surely, we have that
	\begin{align}
		&l_{\text{TSLS}}^n(\hat{\theta}^n_{\lambda^\star_n,\eta_0,\mu})
		\leq  \, l_{\text{TSLS}}^n(\hat{\theta}^n_{0,\eta_0,\mu})
		= \, n^{-1} (\B{Y} - \B{B} \hat{\theta}^n_{0,\eta_0,\mu})^{\top } \B{P}_\delta\B{P}_\delta(\B{Y} - \B{B} \hat{\theta}^n_{0,\eta_0,\mu}). \label{eq:unconfounded_IVinOLSboundedByConvergenceToZero}
	\end{align}
	Furthermore, recall from \eqref{eq:ThetaLambdaConsistentEstimator} that
	\begin{align} \label{eq:unconfoundedOLSconvergence}
		\hat{\theta}^n_{0,\eta_0,\mu}  \underset{n\to\infty}{\stackrel{P}{\longrightarrow}} \theta_0 = \theta^0,
	\end{align}
	where the last equality follows from \eqref{eq:ThetaLambdaInTermsOfTrueTheta} using that we are in the unconfounded case $\E[B(X)\xi_Y]=0$. By expanding and deriving convergence statements for each term, we get
	\begin{align} \notag
		&(\B{Y} - \B{B} \hat{\theta}^n_{0,\eta_0,\mu})^{\top } \B{P}_\delta\B{P}_\delta(\B{Y} - \B{B} \hat{\theta}^n_{0,\eta_0,\mu}) \\ \notag
		& \underset{n\to\infty}{\stackrel{P}{\longrightarrow}}  (\E[ YC^\top ] - \theta_0\E[B C^\top ]) \E[C^\top C]^{-1}
		(
		\E[CY]
		- \E[CB^\top] \theta_0 ) \\
		&  = 0, \label{eq:unconfoundedTSLSinOLSConvpZero}
	\end{align}
	where we used Slutsky's theorem, the weak law of large numbers, \eqref{eq:unconfoundedOLSconvergence} and \eqref{eq:ExpansionOfECY_0}. Thus, by \eqref{eq:unconfounded_IVinOLSboundedByConvergenceToZero} and \eqref{eq:unconfoundedTSLSinOLSConvpZero} it holds that
	\begin{align*}
		l_{\text{TSLS}}^n(\hat{\theta}^n_{\lambda^\star_n,\eta_0,\mu}) = n^{-1}\| \B{P}_\delta(\B{Y} - \B{B} \hat{\theta}^n_{\lambda^\star_n,\eta_0,\mu})  \|_2^2 \underset{n\to\infty}{\stackrel{P}{\longrightarrow}} 0.
	\end{align*}
	For any $z\in \R^n$ we have that 
	\begin{align*}
		\| \B{P}_\delta z  \|_2^2 &  = 
		z^\top  \fC ( \fC^\top \fC + \delta \fM )^{-1}
		\fC^\top\fC ( \fC^\top \fC + \delta \fM )^{-1}
		\fC^\top z \\
		& =
		z^\top  \fC ( \fC^\top \fC + \delta \fM )^{-1}
		(\fC^\top\fC)^{1/2}(\fC^\top\fC)^{1/2}
		( \fC^\top \fC + \delta \fM )^{-1}
		\fC^\top z \\ 
		&= \| (\fC^\top\fC)^{1/2} ( \fC^\top \fC + \delta \fM )^{-1}
		\fC^\top z  \|_2^2,
	\end{align*}
	hence
	\begin{align}  \notag
		\norm{H_n-G_n \hat{\theta}^n_{\lambda^\star_n,\eta_0,\mu} }_2^2 
		&= \| n^{-1/2}(\fC^\top\fC)^{1/2} ( \fC^\top \fC + \delta \fM )^{-1}
		\fC^\top (\B{Y} - \B{B} \hat{\theta}^n_{\lambda^\star_n,\eta_0,\mu})\|_2^2 \\ 
		&\stackrel{P}{\to } 0, \label{eq:consitUconfoundedDifferenceConvPtoZero}
	\end{align}
	where for each $n \in \N$, $G_n\in \R^{k\times k}$ and $H_n \in \R^{k\times 1}$ are defined as
	\begin{align*}
		G_n &:= n^{-1/2}(\fC^\top\fC)^{1/2} ( \fC^\top \fC + \delta \fM )^{-1} \fC^\top \fB, \text{ and } \\
		H_n &:= n^{-1/2}(\fC^\top\fC)^{1/2} ( \fC^\top \fC + \delta \fM )^{-1} \fC^\top \fY.
	\end{align*}
	Using the weak law of large numbers, the continuous mapping theorem and Slutsky's theorem,
	it follows that, as $n \to \infty$,
	\begin{align*}
		G_n \stackrel{P}{\to}  G 	&:= E[CC^\top]^{1/2} E[CC^\top]^{-1} E[CB^\top], \text{ and }\\
		H_n \stackrel{P}{\to}  H 	&:= E[CC^\top]^{1/2} E[CC^\top]^{-1} E[CY] \\
		& = E[CC^\top]^{1/2} E[CC^\top]^{-1} E[CB^\top ]\theta^0 \\
		&= G\theta^0,
	\end{align*}
	where the second to last equality follows from \eqref{eq:ExpansionOfECY_0}. 
	Together with \eqref{eq:consitUconfoundedDifferenceConvPtoZero}, we now have that
	\begin{align*}
		&\, \norm{G_n \hat{\theta}^n_{\lambda^\star_n,\eta_0,\mu}  - G \theta^0}_2^2
		\leq  \norm{G_n \hat{\theta}^n_{\lambda^\star_n,\eta_0,\mu}  - H_n}_2^2 + 
		\norm{H_n  - G \theta^0}_2^2 
		\underset{n\to\infty}{\stackrel{P}{\longrightarrow}}  0.
	\end{align*}
	Furthermore, by the rank assumptions in \ref{ass:RankCondition} we have that $G_n\in \R^{k\times k}$ is of full rank (with probability tending to one), hence
	\begin{align*}
		\|\hat{\theta}^n_{\lambda^\star_n,\eta_0,\mu} -\theta^0\|_2^2 &= \|G_n^{-1}G_n( \hat{\theta}^n_{\lambda^\star_n,\eta_0,\mu} -\theta^0)\|_2^2 \\
		& \leq \|G_n^{-1}\|_{\text{op}}^2 \|G_n( \hat{\theta}^n_{\lambda^\star_n,\eta_0,\mu} -\theta^0) \|_2^2 \\
		&\stackrel{P}{\to}\|G^{-1}\|_{\text{op}}^2 \cdot 0 \\
		&=0,
	\end{align*}
	as $n\to \infty$, proving the proposition.
	
\end{proofenv}

  \chapter[Structure Learning For Directed Trees]{Structure Learning For Directed Trees}

\vspace*{1cm}  

\begin{quote}
	\begin{enumerate}
		\item[\textbf{\ref{app:graphs}}] \nameref{app:graphs} 
		\item[\textbf{\ref{sec:AppDetails}}] \nameref{sec:AppDetails} 
		\item[\textbf{\ref{app:Experiments}}] \nameref{app:Experiments} 
		\item[\textbf{\ref{app:proofs}}] \nameref{app:proofs} 
	\end{enumerate}
\end{quote}

\section{Graph Terminology} \label{app:graphs}
A \textit{directed graph} $\cG=(V,\cE)$ consists of $p\in \N_{>0}$ vertices (nodes)   $V=\{1,\ldots,p\}$ and a collection of directed edges $\cE\subset \{(j\to i) \equiv (j,i): i,j\in V, i\not = j\}$. For any graph $\cG=(V,\cE)$ we let $\PAg{\cG}{i}:= \{v\in V: \exists (v,i)\in \cE\}$ and $\CHg{\cG}{i} :=\{v\in V : \exists (j,v)\in \cE\}$ denote the \textit{parents} and \textit{children} of node $i\in V$ and we define root nodes $
\mathrm{rt}(\cG) := \{v\in V: \PAg{\cG}{i} =\emptyset \}$ as nodes with no parents (that is, no incoming edges). 
A \textit{path} in $\cG$ between two nodes $i_1,i_k\in V$ consists of a sequence $(i_1, i_2), \ldots, (i_{k-1}, i_k)$ of pairs of nodes such that for all $j \in \{1, \ldots, k-1\}$, we have either 
$(i_j \to i_{j+1}) \in \cE$
or
$(i_{j+1} \to i_{j}) \in \cE$.
A \textit{directed path} in $\cG$ between two nodes $i_1,i_k\in V$ consists of a sequence $(i_1, i_2), \ldots, (i_{k-1}, i_k)$ of pairs of nodes such that for all 
$j \in \{1, \ldots, k-1\}$, we have $(i_j \to i_{j+1}) \in \cE$. Furthermore, we let $\ANg{\cG}{i}$ and $\DEg{\cG}{i}$ denote the \textit{ancestors} and \textit{descendants} of node $i\in V$, consisting of all nodes $j\in V$ for which there exists a directed path to and from $i$, respectively. We let $\NDg{\cG}{i}$ denote the \textit{non-descendants} of $i$. A \textit{directed acyclic graph} (DAG) is a directed graph that does not contain any directed cycles, i.e., directed paths visiting the same node twice.
We say that a graph is \textit{connected} if a (possibly undirected) path exists between any two nodes. A \textit{directed tree} is a connected DAG in which all nodes have at most one parent. More specifically, every node has a unique parent except the root node, which has no parent. The root node $\root{\cG}$ is the unique node such there exists a directed path from $\root{\cG}$ to any other node in the directed tree. In graph theory, a directed tree is also called an \textit{arborescence}, a \textit{directed rooted tree}, and a \textit{rooted out-tree}.   A graph $\cG=(V',\cE')$ is a \textit{subgraph} of another graph $\cG=(V,\cE)$ if $V'\subseteq V$, $\cE'\subseteq \cE$ and for all $(j\to i) \in \cE'$ it holds that $j,i\in V'$. A subgraph is \textit{spanning} if $V'=V$. For any DAG $\cG=(V,\cE)$ and three mutually distinct subsets $A,B,C\subset V$ we let $A\dsep{\cG} B\,|\,C$ denote that $A$ and $B$ are d-separated by $C$ in $\cG$ \citep[see, e.g.,][]{Pearl2009}.

\section{Further Details on Section~\ref{sec:ScoreGap}} \label{sec:AppDetails}

\begin{remark} \label{rmk:ConditionalEntropyRebane}
	The conditional entropy score gap is not strictly positive when considering the alternative graphs $\tilde \cG$ that are Markov equivalent  to the causal graph $\cG$, $\tilde \cG\in \mathrm{MEC}(\cG)$. A simple translation of the conditional entropy score function reveals that
	\begin{align*}
		\lCE(\tilde \cG) + C= \sum_{(j\to i)\in \tilde \cE} h(X_i|X_j) -h(X_i) = - \sum_{(j\to i)\in \tilde \cE} I(X_i;X_j),
	\end{align*}
	for a constant $C\in \R$. By symmetry of the mutual information, it holds that $
	\lCE(\tilde \cG ) =\lCE(\cG )$, 
	for any $\tilde \cG \in \mathrm{MEC}(\cG)$, since 
	$\tilde \cG $ and $\cG$ share the same skeleton. Thus, the conditional entropy score function can, at most, identify the Markov equivalence class of the causal graph. In fact, the polytree causal structure learning method of \cite{DBLP:conf/uai/RebaneP87} uses the above translated conditional entropy score function to recover the skeleton of the causal graph.
\end{remark}

\begin{example}[Negative local Gaussian score gap] \label{ex:negativelocalscoregap}
	Consider two graphs $\cG$ and $\tilde \cG$ with different root nodes, i.e., $\root{\cG}\not = \root{\tilde \cG}$. If $x\mapsto \E[X_{\mathrm{rt}(\cG)}|X_{\PAg{\tilde \cG}{\mathrm{rt}(\cG)}}=x]$ is not almost surely constant, then it holds that
	\begin{align*}
		\lG(\tilde \cG, \mathrm{rt}(\cG)) -\lG(\cG, \mathrm{rt}(\cG)) &=  \E[(X_{\mathrm{rt}(\cG)}-\E[X_{\mathrm{rt}(\cG)}| X_{\PAg{\tilde \cG}{\mathrm{rt}(\cG)}}])^2] - \Var(X_{\mathrm{rt}(\cG)}) \\
		&= \E[\Var(X_{\mathrm{rt}(\cG)}|X_{\PAg{\tilde \cG}{\mathrm{rt}(\cG)}})] - \Var(X_{\mathrm{rt}(\cG)}) \\
		& =- \Var(\E[X_{\mathrm{rt}(\cG)}|X_{\PAg{\tilde \cG}{\mathrm{rt}(\cG)}}])< 0.
	\end{align*}
\end{example}

\section{Further Details on the Simulation Experiments} \label{app:Experiments}
This section contains further details on the simulation experiments. 

\subsection{Tree Generation Algorithms} \label{app:TreeGeneration}
\algnewcommand{\IIf}[1]{\State\algorithmicif\ #1\ \algorithmicthen}
\algnewcommand{\EndIIf}{\unskip\ \algorithmicend\ \algorithmicif}

The following two algorithms, \Cref{alg:type1tree} (many leaf nodes) and \Cref{alg:type2tree} (many branch nodes), details how the Type 1 and Type 2 trees are generated, respectively.
\begin{algorithm}[h] \caption{Generating type 1 trees} \label{alg:type1tree}
	\begin{algorithmic}
		\Procedure{Type1}{$p$}
		\State $A := 0\in \R^{p\times p}$
		\For{$j\in\{1,\ldots,p\}$}
		\For{$i\in\{j+1,\ldots,p\}$}
		\If{$\sum_{k=1}^p A_{ki}=0$}
		\If{$i=j+1$}
		\State $A_{ji}:= 1$
		\Else
		\State $A_{ji} := \mathrm{Binomial}(\mathrm{success}=0.1)$
		\EndIf
		\Else 
		\State $A_{ji}:= 0$
		\EndIf
		\EndFor
		\EndFor
		\State \textbf{return} $A$
		\EndProcedure
	\end{algorithmic} 
\end{algorithm}
\begin{algorithm}[h] \caption{Generating type 2 trees} \label{alg:type2tree}
	\begin{algorithmic}
		\Procedure{Type2}{$p$}
		\For{$i\in\{2,\ldots,p\}$}
		\State $j := \mathrm{sample}(\{1,\ldots,i-1\})$
		\State $A_{ji}:=1$
		\EndFor
		\State \textbf{return} $A$
		\EndProcedure
	\end{algorithmic} 
\end{algorithm}

\subsection{Additional Illustrations}
This section contains some additional illustrations of the simulation experiments.
\begin{figure}[H]
	\begin{center}
		\includegraphics[width=\textwidth-20pt]{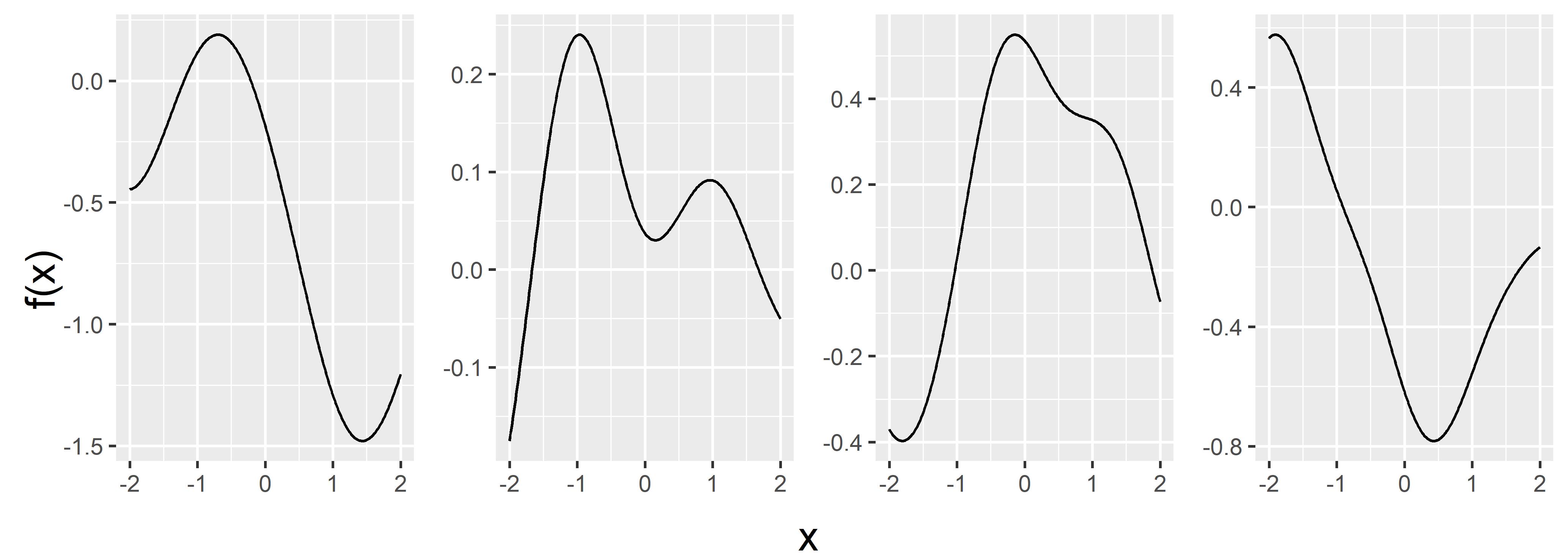}
	\end{center}
	\caption{Four causal functions as modeled by the RBF kernel Gaussian Process. } \label{fig:samplepaths}
\end{figure}

\label{sec:additionalIllustrations}
\begin{figure}[H]
	\begin{center}
		\includegraphics[width=\textwidth-20pt]{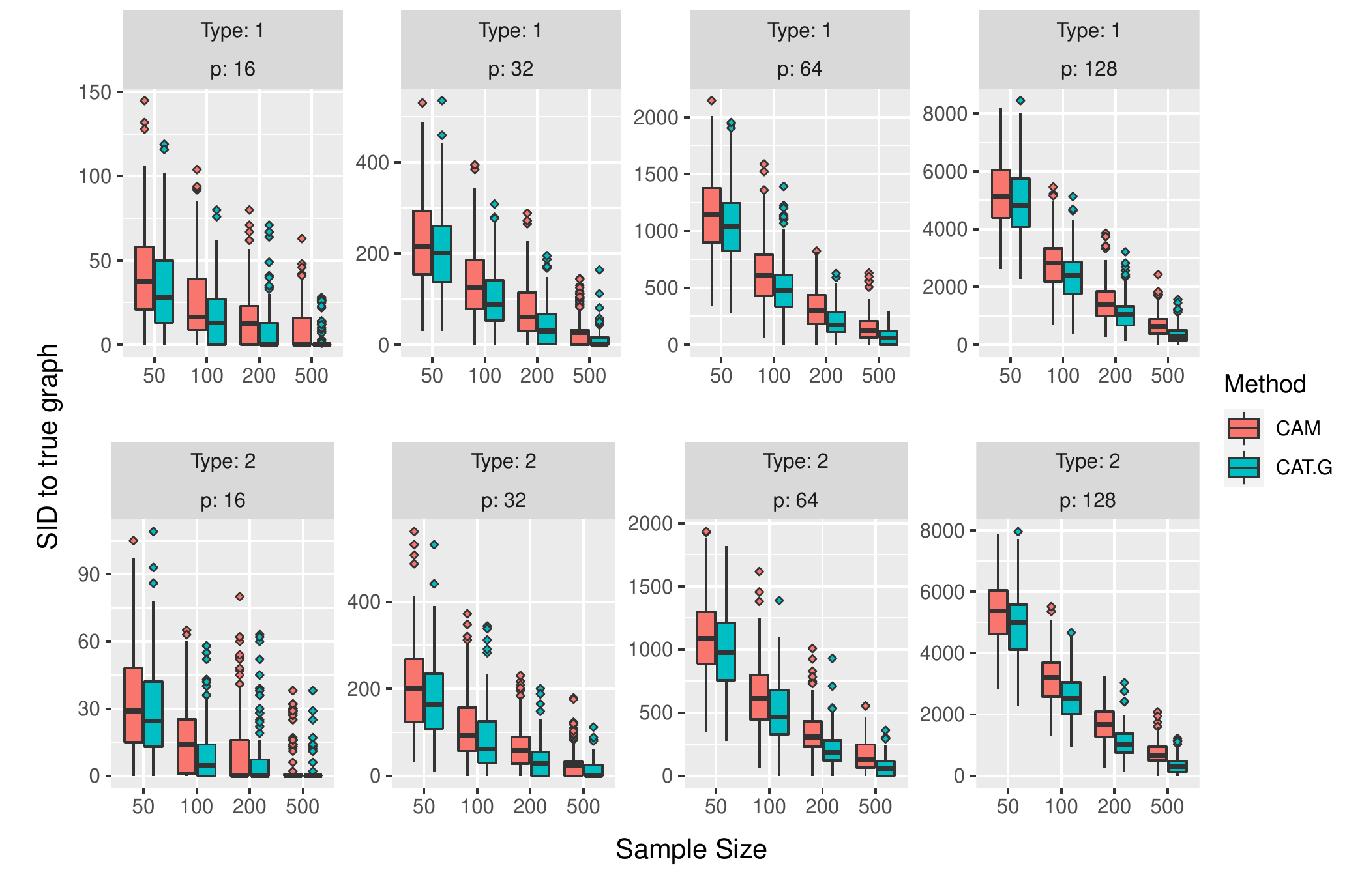}
	\end{center}
	\caption{Boxplot illustrating the SID performance of CAM and CAT for varying sample sizes, system sizes and tree types in the experiment of \Cref{sec:ExperimentGaussianTrees}. CAT.G is CAT with edge weights  derived from the Gaussian score function.}
	\label{fig:boxplot_Gaussian_SID_Type1and2}
\end{figure}
\begin{figure}[H]
	\begin{center}
		\includegraphics[width=\textwidth-20pt]{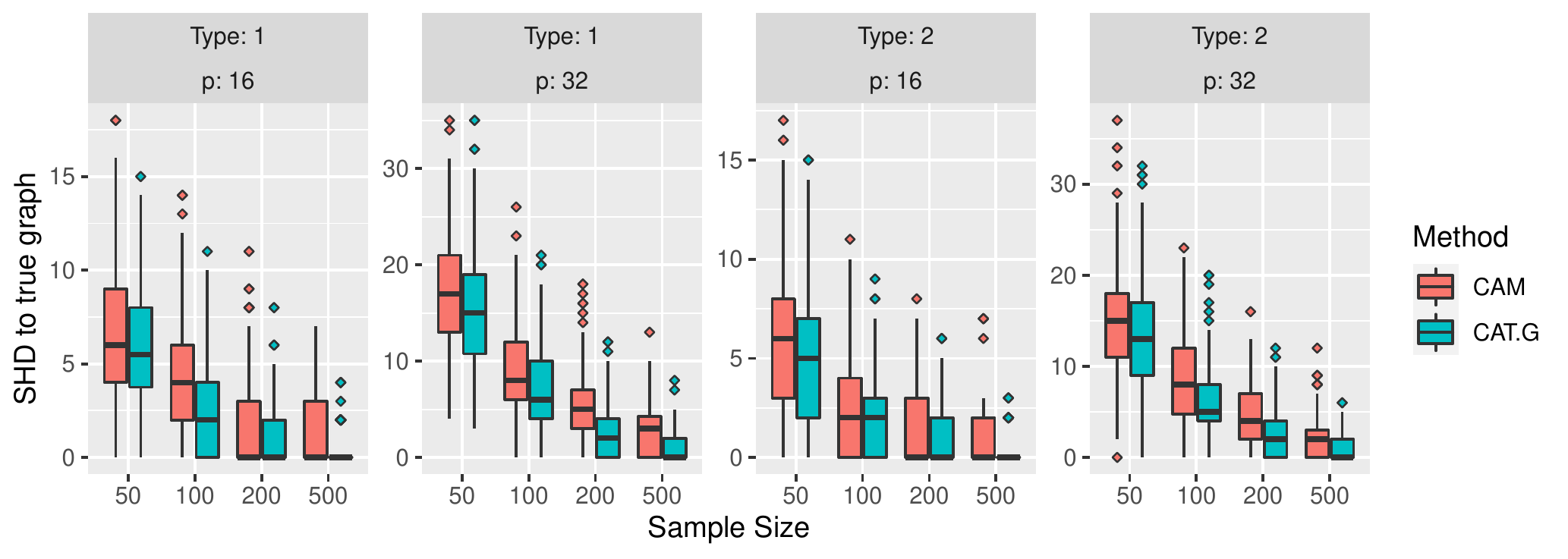}
	\end{center}
	\caption{Boxplot illustrating the SHD performance of CAM and CAT for varying sample sizes, system sizes and tree types in the experiment of \Cref{sec:ExperimentGaussianTrees}. Here CAT.G is run on the CAM edge weights , so that any difference in nonparametric regression technique is ruled out as the source of the performance difference.}
	\label{fig:boxplot_Gaussian_CamScores}
\end{figure}
\begin{figure}[H]
	\includegraphics[width=\textwidth-20pt]{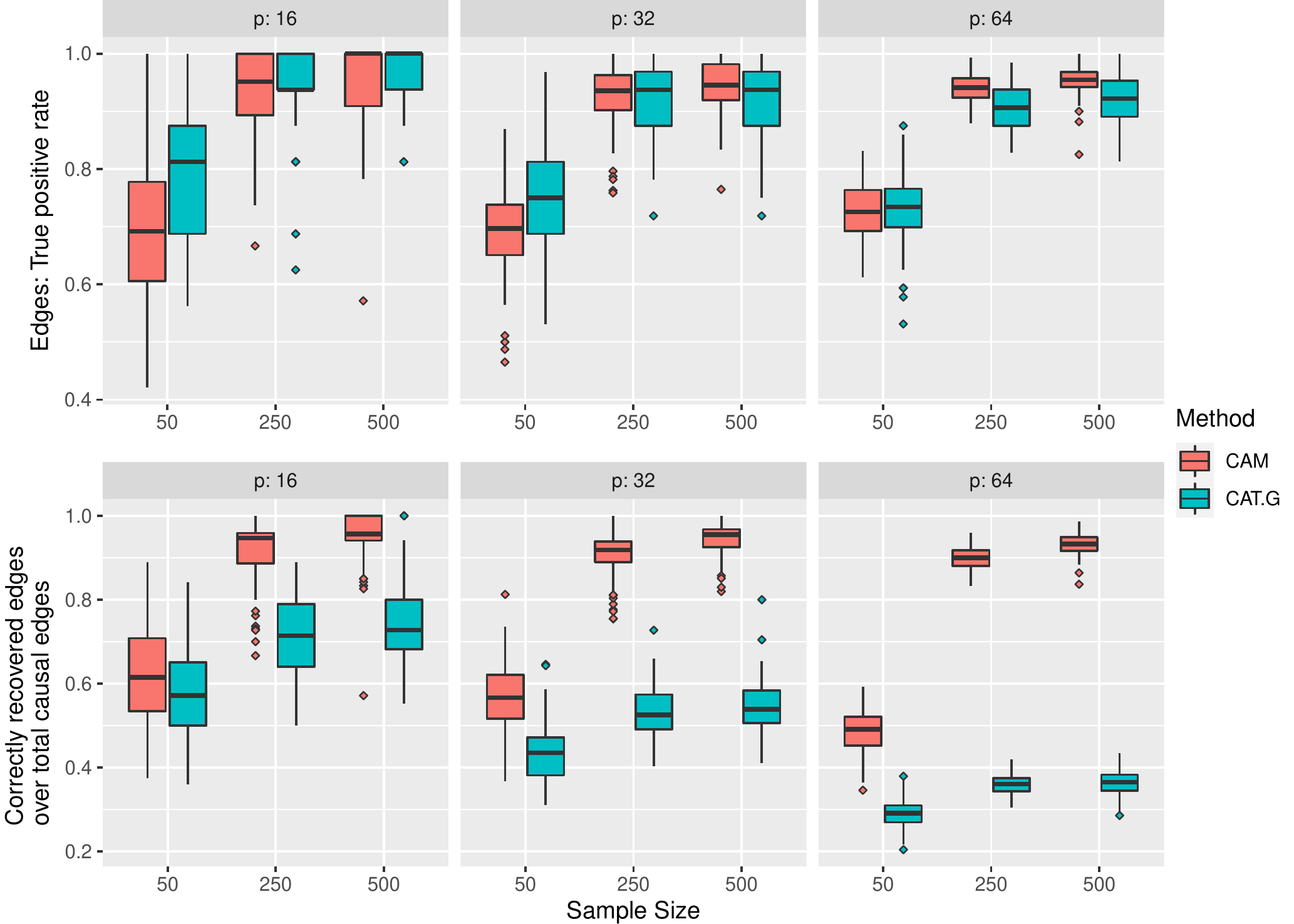}
	\caption{Boxplot of edge relations for the experiment in \Cref{sec:ExperimentCATonDAGs}. }
	\label{fig:SingleRootedDagsEdgeRelations}
\end{figure}
\begin{figure}[H]
	\begin{center}
		\includegraphics[width=\textwidth-20pt]{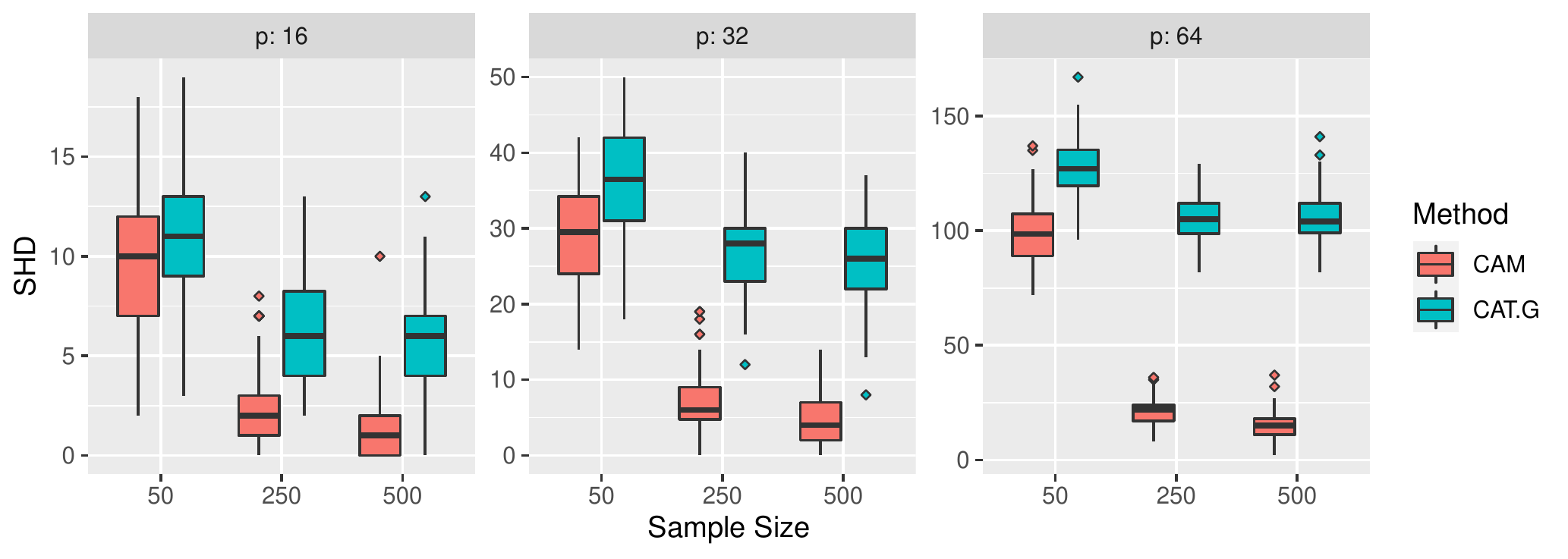}
	\end{center}
	\caption{Boxplot of SHD for the experiment in \Cref{sec:ExperimentCATonDAGs}.}
	\label{fig:SingleRootedDagsSHD}
\end{figure}

\if1
\subsection{When Greedy Searches Fail} \label{sec:WhenGreedySearchFail}
In general, greedy search techniques do not come with theoretical guarantees. The following example is taken from \cite{Peters2022}. As we saw in the previous Gaussian experiment, CAM suffers slightly when there are many leaf nodes and the following setup exemplifies one way a reversed edge may be selected at a leaf node. Consider the following three node Gaussian additive structural causal model with causal graph $(X\to Y \to Z)$
\begin{align} \label{eq:3nodesetup}
	X := N_X, \quad 
	Y := \frac{X^3}{\Var(X^3)} + N_Y, \quad 
	Z := Y+ N_Z,
\end{align}
where $N_X\sim \cN(0,1.5)$, $N_Y \sim \cN(0,0.5)$ and $N_Z \sim \cN(0,0.5)$ are mutually independent. We simulate this setup for sample sizes $n\in\{50,100,200,500,1 000,5 000,10000,20000\}$ and estimate the causal graph by CAT.G, CAT.E and CAM. \Cref{fig:boxplot_3node_SHD} illustrates the results. 
\begin{figure}[h]
	\begin{center}
		\includegraphics[width=\textwidth-20pt]{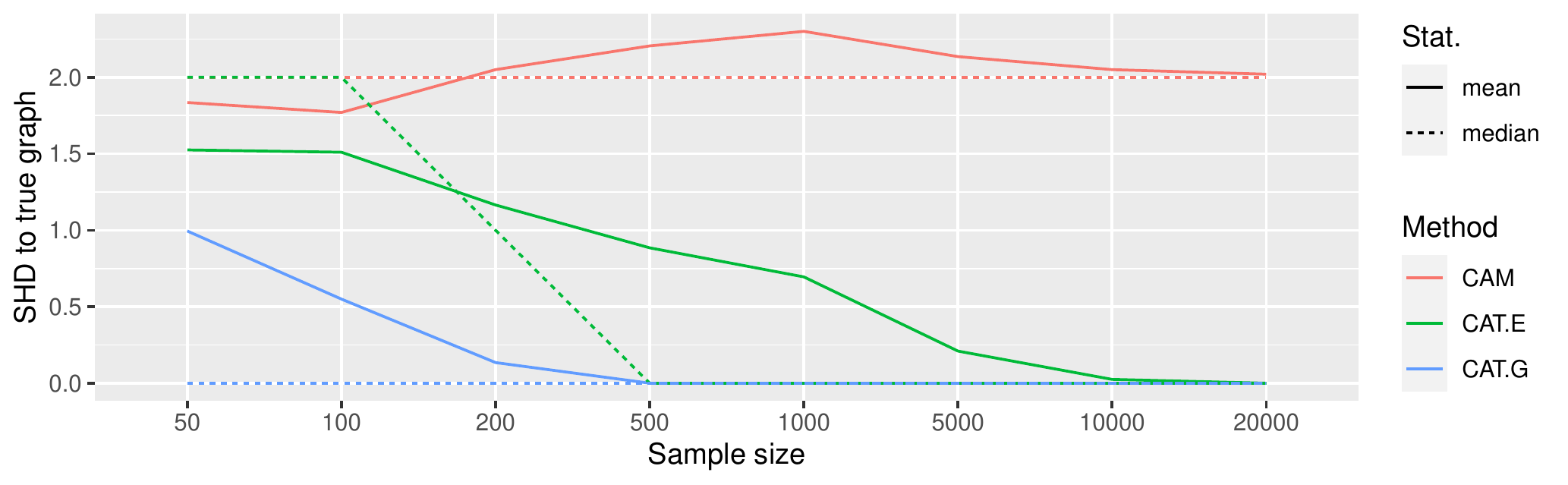}
	\end{center}
	\caption{SHD performance of CAT.G, CAT.E and CAM in the three node setup of \Cref{eq:3nodesetup}. The solid and dashed lines represent the mean and median SHD, respectively.}
	\label{fig:boxplot_3node_SHD}
\end{figure}
Even with increasing sample size, CAM does not converge to the correct answer. The reason is that it selects the wrong edge in the first step.

\Cref{fig:boxplot_3node_weights} shows
the estimated  Gaussian edge weights  based on a sample size of 1000000. 
The smallest edge weight is given by the wrong edge $(Z\to Y)$ so the greedy search erroneously picks this edge. However, Chu--Liu--Edmonds' algorithm used by CAT.G correctly realizes that the full score of the correct graph $X\to Y \to Z$ is smaller than the full score of $Z\to Y \to X$ which is recovered by CAM, i.e., that $-1.41=
\hat 	w_\mathrm{G}(X \to Y) + \hat	w_\mathrm{G}(Y \to Z) < \hat	w_\mathrm{G}(Z \to Y)+ 	\hat w_\mathrm{G}(Y \to X)=-1.28$.
\begin{figure}[h]
	\begin{center}
		\begin{tikzpicture}
			\begin{scope}[every node/.style={circle,thick,draw},
				roundnode/.style={circle, draw=black, fill=gray!10, thick, minimum size=9mm},
				roundnode/.style={circle, draw=black, fill=gray!10, thick, minimum size=7mm,},
				]
				\node[roundnode] (X) at (0,0) {$X$};
				\node[roundnode] (Y) at (5,0) {$Y$};
				\node[roundnode] (Z) at (10,0) {$Z$};
			\end{scope}
			
			\begin{scope}[
				every node/.style={fill=white,circle},
				every edge/.style={draw=red,very thick}]
				\path [-stealth, bend right=15,draw=blue] (Z) edge node [pos=0.5] {$-1.00$} (Y);
				\path [-stealth, bend right=15] (Y) edge node [pos=0.5] {$-0.28$} (X);
			\end{scope}
			
			\begin{scope}[
				every node/.style={fill=white,circle},
				every edge/.style={draw=black,very thick}]
				\path [-stealth, bend left=30] (Z) edge node [pos=0.5] {$-0.17$} (X);
				\path [stealth-, bend right=30] (Z) edge node [pos=0.5] {$-0.26$} (X);
			\end{scope}
			
			\begin{scope}[
				every node/.style={fill=white,circle},
				every edge/.style={draw=blue,very thick}]
				\path [-stealth, bend right=15] (X) edge node [pos=0.5] {$-0.46$} (Y);
				\path [-stealth, bend right=15] (Y) edge node [pos=0.5] {$-0.95$} (Z);
				
			\end{scope}
			
		\end{tikzpicture}
	\end{center}
	\caption{Visualization of the edge weights  of the experiment in Section~\ref{sec:WhenGreedySearchFail}. Each edge label is the estimated Gaussian edge weight  as produced by the CAM scoring method based on 1000000 i.i.d.\ observations generated from the structural causal system of \Cref{eq:3nodesetup}. The red and blue edges are recovered by the greedy search of CAM and CAT, respectively.}
	\label{fig:boxplot_3node_weights}
\end{figure}
\fi

\newpage

\section{Proofs} \label{app:treesproofs}
This section contains the proofs of all results presented in the main text.

\subsection{Proofs of Section~\ref{sec:ScoreBasedStructureLearning}}

\begin{proofenv}{\Cref{lm:RestrictedModelConditionGaussian}}
	Let $\theta = (\cG,(f_i),P_N)\in \cT_p \times \cD_3^p \times \cP_{\mathrm{G}}^p$. Furthermore, let all causal functions $(f_i)$ be nowhere constant and nonlinear. The additive noise is Gaussian, so the log density of $N_i$ for all $i\in\{1,...,p\}$ is given by
	\begin{align*}
		\nu_i(x) = -\frac{1}{2}\log(2\pi \sigma_i^2) -\frac{x^2}{2\sigma_i^2} , \quad  	\nu_i'(x) = -\frac{x}{\sigma_i^2}, \quad  \nu_i''(x)= -\frac{1}{\sigma_i^2}, \quad  \nu_i'''(x) = 0.
	\end{align*}
	By assumption we have that condition (i) of \Cref{def:RestrictedSEMGeneralCase} is satisfied, hence assume for contradiction that condition (ii) of \Cref{def:RestrictedSEMGeneralCase} is not satisfied. That is, we assume that there exists an $i\in \{1,...,p\}\setminus \{\root{\cG}\}$ such that for all \begin{align*}
		(x,y)\in \cJ :=& \, \{(x,y)\in \R^2: \nu_i''(y-f_i(x))f'_i(x)\not = 0\}\\
		=&\,	\{(x,y)\in \R^2: f_i'(x)\not = 0\},
	\end{align*}
	it holds that
	\begin{align} \label{eq:RestrictedGaussianModelTempEq1}
		\xi'''(x) - \xi''(x)  \frac{f_i''(x)}{f_i'(x)}  -\frac{2 f_i''(x)f_i'(x) }{\sigma^2} &=    -\frac{y-f_i(x) }{\sigma^2}\left(  f_i'''(x)-\frac{(f_i''(x))^2}{f_i'(x)}\right).
	\end{align}
	Henceforth, suppress the subscript $i$ of $f_i$ and $\sigma_i$.
	First note that $\{x\in \R: f'(x) = 0\}$ is closed by continuity of $f'$. The complement is open, hence there exists a countable collection of mutually disjoint open intervals $(O_k)_{k\in \bZ}$ such that  $\{x\in \R: f'(x) \not = 0\} = \cup_{k\in \bZ} O_k$. Since $f$ is nowhere constant we know that $\{x\in \R : f'(x)=0\}$ has empty interior which implies that $\overline{\cup_{k\in \bZ} O_k}=\R$. Now let $(O_k)_{k\in\bZ}$ be indexed by $\mathbb{Z}$ such that for any $k,j\in \mathbb{Z}$ with $k<j$ and $x\in O_k,y\in O_j$ it holds that $x<y$.
	As the left-hand side of \Cref{eq:RestrictedGaussianModelTempEq1} is constant in $y$ it must hold that
	\begin{align*}
		0 = 	f'''(x) - \frac{(f''(x))^2}{f'(x)} =   \frac{\frac{\partial f''(x) }{\partial x}f'(x)-  f''(x)\frac{\partial f'(x)}{\partial x}}{(f'(x))^2} = \frac{\partial }{\partial  x} \left(\frac{f''(x)}{f'(x)} \right) ,
	\end{align*}
	i.e., $f''(x)/f'(x)$ is constant, for all $x\in \cup_{k\in \mathbb{Z}} O_k$. On each $O_k$ we have that \begin{align*}
		\partial/ \partial x \log(\text{sign}(f'(x)) f'(x)) = c_{k,1} &\iff \log( \text{sign}(f'(x)) f'(x)) = c_{k,1}x + c_{k,2} \\ &\iff \text{sign}(f'(x)) f'(x)= \exp(c_{k,1}x+c_{k,2}) \\&\iff f'(x) = \pm \exp(c_{k,1}x+c_{k,2}).
	\end{align*}
	Recall that we have assumed continuous differentiability of $f'$. That is, for any $k\in\mathbb{Z}$ and $t_k := \sup(O_k) = \inf(O_{k+1})$ we have $\lim_{x\uparrow t_k} f'(x)  
	= \lim_{x\downarrow  t_k} f'(x)$ and  $\lim_{x\uparrow t_k} f''(x)  
	= \lim_{x\downarrow  t_k} f''(x)$. Assume without loss of generality that $f'(x)=\exp(c_{k,1}x+c_{k,2})$ for all $x\in O_k$ and $k\in\mathbb{Z}$.  These conditions impose the restrictions $
	(c_{k,1}-c_{{k+1},1} ) t_k =  c_{{k+1},2}-c_{k,2}$ and $\log ( c_{k,1}/c_{k+1,1}) + (c_{k,1}-c_{k+1,1}) t_k = c_{k+1,2}- c_{k,2}$ 
	which entails that $c_{k,1}=c_{k+1,1}$ and $c_{k,2}=c_{k+1,2}$. This proves that there exists $c_1,c_2\in \R$ such that  $f'(x) =  \exp(c_1x+c_2)$ for all $x\in \R$. Thus, the differential equation holds for all $x\in \R$,
	\begin{align*}
		0= 	\xi'''(x) - \xi''(x)  \frac{f''(x)}{f'(x)}  -\frac{2 f''(x)f'(x) }{\sigma^2} = \frac{\partial }{\partial x}\left( \frac{\xi''(x)}{f'(x)}\right)  - 2\frac{f''(x)}{\sigma^2},
	\end{align*}
	by division with $f'(x)$. By integration this implies that $0 = \xi''(x)/f'(x) - 2f'(x)/\sigma^2 +c_3$ such that $\xi''(x)= 2\exp(2c_1 x+2c_2)/\sigma^2 -c_3\exp(c_1x+c_2)$ and $\xi'(x) =\exp(2c_1x+2c_2)/c_1\sigma^2 - c_3\exp(c_1x+c_2)/c_1 + c_4$ and 
	\begin{align*}
		\xi(x) = \frac{\exp(2c_1x+2c_2)}{2c_1^2\sigma^2} - \frac{c_3\exp(c_1x+c_2)}{c_1^2} + c_4x +c_5.
	\end{align*}
	We see that $\xi(x) \to \i\iff p_{X_{\PAg{\cG}{i}}}(x) \to \infty$ as $x\to \text{sign} (c_1) \cdot\i$, in contradiction with the assumption that $p_{X_{\PAg{\cG}{i}}}(x)$ is a probability density function if $c_1\not = 0$. Thus, it must hold that $f''(x)/f'(x)=0$ for all $x\in \R$, or equivalently, that $f$ is a linear function,  yielding a contradiction.

	This proves that whenever $f_i\in \cD_3$ is a nowhere constant and nonlinear function and the additive noise is Gaussian then condition (ii) of \Cref{def:RestrictedSEMGeneralCase} is satisfied, so $\theta\in \Theta_R$. %
\end{proofenv}

\begin{proofenv}{\Cref{thm:UniqueGraph}}
	First, we consider the bivariate setting. 	Let $(X,Y)$ be generated by an additive noise SCM $\theta\in \Theta_R\subset \cT_2 \times \cD_3^2\times \cP_{\cC_3}^2$ given by $X:= N_X$ and $Y := f(X)+ N_Y$ 
	with $P_{X} = p_{X}\cdot \lambda$ and $P_{N_Y}=p_{N_Y}\cdot \lambda$ having three times differentiable strictly positive densities and $f$ is a three times differentiable nowhere constant function such that condition (ii) of \Cref{def:RestrictedSEMGeneralCase} holds.

	Assume for contradiction that we do not have observational identifiability of the causal structure $\cG= (V=\{X,Y\},\cE=\{(X\to Y)\})$. That is, there exists $\tilde \theta \in \cT_2 \times \cD_1^p \times \cP_{\cC_0}^p$ with causal graph $\tilde \cG\not = \cG$ or, equivalently, a differentiable function $g$ and noise distributions $P_{\tilde N_X}= p_{\tilde N_X}\cdot \lambda $  and $P_{\tilde N_Y}=p_{\tilde N_Y}\cdot \lambda $ with %
	continuous densities such that the structural assignments $\tilde Y:= \tilde N_Y $ and $\tilde X:= g( \tilde Y)+ \tilde N_X$ induce the same distribution, i.e., 
	\begin{align} \label{eq:EqDistIdentifiabilityProof}
		P_{X,Y} = P_{\tilde X, \tilde Y}.
	\end{align}
	By the additive noise structural assignments we know that both $P_{X,Y}$ and $P_{\tilde X, \tilde Y}$ have densities with respect to $\lambda^2$ given by
	\begin{align*}
		p_{X,Y}(x,y)  &=p_{X}(x)p_{N_Y}(y-f(x)), \\
		p_{\tilde X, \tilde Y}(x,y) &= p_{\tilde N_X}(x-g(y))p_{\tilde Y}(y),
	\end{align*}
	for all $(x,y)\in \R^2$. By the equality of distributions in \Cref{eq:EqDistIdentifiabilityProof} and strict positivity of $p_{X}$ and $p_{N_Y}$ we especially have that  for $\lambda^2$-almost all $(x,y)\in \R^2$ 
	\begin{align}\label{eq:EqDensityIdentifiabilityProof}
		0<p_{X,Y}(x,y)= p_{\tilde X, \tilde Y}(x,y).
	\end{align}
	However, as both $p_{X,Y}$ and $p_{\tilde X, \tilde Y}$ are continuous we realize that the inequality in  \Cref{eq:EqDensityIdentifiabilityProof} holds for all $(x,y)\in \R$ (if they were not everywhere equal there would exists a non-empty open ball in $\R^2$ on which they differ in contradiction with $\lambda^2$-almost everywhere equality).
	Furthermore, by the assumption that $f$ is three times differentiable and $p_{X}$, $p_{N_Y}$ are three times continuously differentiable we have that $\partial^3 \pi /\partial x^3$ and $\partial^3 \pi / \partial x^2 \partial y$ are well-defined partial-derivatives of 
	\begin{align*}
		\pi(x,y):=\log p_{X,Y}(x,y)=  \log  p_{X}(x) + \log p_{N_Y}(y-f(x)) =: \xi(x) + \nu(y-f(x)),
	\end{align*}
	With $\tilde \pi(x,y) := \log p_{\tilde X, \tilde Y}$ we have that
	\begin{align*}
		\tilde \pi(x,y) = \log p_{\tilde N_X}(x-g(y)) +\log p_{\tilde Y}(y) =: \tilde \xi (x-g(y)) +\tilde \nu(y).
	\end{align*}
	Since it holds that $\pi=\tilde \pi$ by \Cref{eq:EqDensityIdentifiabilityProof} the partial-derivatives 
	$\partial^3 \tilde \pi /\partial x^3$ and $\partial^3 \tilde  \pi / \partial x^2 \partial y$ are also well-defined. Now note that for any $x,y\in \R$
	\begin{align*}
		0 =	\lim_{h\to 0} |\tilde \pi(x+h,y) - \tilde \pi(x,y)|/h  = \lim_{h\to 0} |\tilde \xi(x-g(y)+h)-\tilde \xi(x-g(y))|/h,
	\end{align*}
	implying that $\tilde \xi$ is differentiable in $x-g(y)$ for any $x,y\in \R$ or, equivalently, $\tilde \xi$ is everywhere differentiable. Similar arguments yield that $\tilde \xi$ is at least three times differentiable. We conclude that $
	\partial^2\tilde \pi(x,y)/\partial x^2 = \tilde \xi''(x-g(y))$ and $\partial^2 \tilde \pi(x,y) / \partial x \partial y = -\tilde \xi''(x-g(y))g'(y)$ and for any $(x,y)\in \R^2$ such that $\partial^2 \tilde \pi(x,y) / \partial x \partial y \not = 0$ or, equivalently, $$\forall (x,y)\in \cJ:=\left\{(x,y): \frac{\partial^2  \pi(x,y)}{ \partial x \partial y } =-\nu''(y-f(x))f'(x)\not = 0 \right\},$$  it holds that
	\begin{align*}
		\frac{\partial }{\partial x} \left(\frac{\frac{\partial^2}{\partial x^2} \tilde \pi(x,y)}{\frac{\partial^2}{\partial x \partial y} \tilde \pi(x,y)} \right) = \frac{\partial }{\partial x} \left( \frac{-1}{g'(y)} \right) = 0.
	\end{align*}
	It is worth noting that $\cJ\not = \emptyset$ to ensure that the following derivations are not void of meaning. (This can be seen by noting that $f$ is nowhere constant, i.e., $f'(x)\not =0$ for $\lambda$-almost all $x\in \R$. Hence, $\cJ=\emptyset$ if and only if $p_{N_Y}$ is a density such that $\{(x,y)\in \R^2 :f'(x)\not = 0\}\ni (x,y)\mapsto \nu''(y-f(x))$ is constantly zero or, equivalently, $\R \ni y\mapsto\nu''(y)$ is constantly zero. This holds if and only if $p_{N_Y}$ is either exponentially decreasing or exponentially increasing everywhere, which is a contradiction as no continuously differentiable function integrating to one has this property.) For any $(x,y)\in \cJ$ we also have that
	\begin{align*}
		0=\frac{\partial }{\partial x} \left( \frac{\frac{\partial^2}{\partial x^2} \pi(x,y)}{\frac{\partial^2}{\partial x \partial y} \pi(x,y)} \right) &= 	\frac{\partial }{\partial x} \left( \frac{\xi''(x)+ \nu''(y-f(x))f'(x)^2 -\nu'(y-f(x))f''(x) }{-\nu''(y-f(x))f'(x)} \right) \\
		&= -2f'' + \frac{\nu'f'''}{\nu''f'} - \frac{\xi'''}{\nu''f'}+\frac{\nu'''\nu'f''}{(\nu'')^2}\\
		&\quad -\frac{\nu'''\xi''}{(\nu'')^2} -\frac{(f'')^2\nu'}{\nu''(f')^2}+\frac{f''\xi''}{\nu''(f')^2},
	\end{align*}
	which implies that
	\begin{align} \label{eq:DifferentiablEquationIdentifiabilityProof}
		\xi''' =  \xi'' \left( \frac{f''}{f'} -\frac{f' \nu'''}{\nu''} \right) -2\nu''f''f' +  \nu'f''' +\frac{ \nu'''\nu'f''f'}{\nu''}-\frac{\nu'(f'')^2}{f'},
	\end{align}
	in contradiction with the assumption that condition (ii) of \Cref{def:RestrictedSEMGeneralCase} holds. We conclude that $P_{X,Y}\not = P_{\tilde X, \tilde Y}$.
	
	Now consider a multivariate restricted causal model $\theta \in \Theta_R$ over $X=(X_1,\ldots,X_p$) with causal directed tree graph $\cG=(V,\cE)$. Assume for contradiction that there exists an alternative SCM $\tilde \theta  = (\tilde \cG,(\tilde f_i), P_{\tilde N}) \in \cT_p \times \cD_1^p \times \cP_{\cC_0}^p$ inducing $\tilde X = (\tilde X_1,\ldots,\tilde X_p)$ with causal graph $\tilde \cG=(V,\tilde \cE)\not = \cG$,  such that $P_{X}= P_{\tilde X}$. 
	
	Any SCM induced distribution is Markov with respect to the underlying causal graph. As such, we have that $P_X$ is  Markov with respect to both $\cG$ and $\tilde \cG$. Furthermore, since (in $\theta$) the causal functions are non-constant and the noise innovations have strictly positive density, we have, by Proposition 17 of \cite{peters2014causal}, that $P_X$ satisfies causal minimality with respect to causal graph $\cG$ of $\theta$, i.e., it is globally Markov with respect to  $\cG$ but not any proper subgraph of $\cG$. If $P_X$ also satisfies causal minimality with respect to $\tilde \cG$, then, by Proposition 29 of \cite{peters2014causal}, there exist $i,j\in V$ such that $(j\to i)\in \cE$ and $(i\to j)\in \tilde \cE$.  
	
	\begin{quote}
		Assume for contradiction that $P_X$ does not satisfy causal minimality with respect to $\tilde \cG$. By Proposition 4 of \cite{peters2014causal}, we have that there exists $(j'\to i')\in \tilde \cE$ such that $X_{j'}\independent X_{i'}$. Define $A:=\NDg{\tilde \cG}{j'}\cup \{j'\}$ and $B:= \DEg{\tilde \cG}{i'}\cup \{i'\}$. It holds that $A\dsep{\tilde \cG} (B\setminus\{i'\})\,|\, i'$, i.e., $A$ and $B\setminus\{i'\}$ are d-separated by $i'$ in the directed tree $\tilde \cG$. Since $P_X$ is Markov with respect $\tilde \cG$ it holds that $X_{A} \independent X_{B\setminus \{i'\}}\,|\, X_{i'}$, hence $X_{A} \independent X_{B}\,|\, X_{i'}$.
		Similarly, it holds that $X_{A}\independent X_{B}\,|\, X_{j'}$ which implies that $X_{A} \independent X_{i'} \, | \, X_{j'}$.
		By applying the contraction property of conditional independence, we get that
		\begin{align*}
			&X_{A} \, \independent X_{i'} \, |\, X_{j'} \quad \text{and} \quad X_{i'} \, \independent X_{j'} \implies X_{A} \independent X_{i'}, \text{ and }\\
			&X_{A} \, \independent X_{B} \, | \, X_{i'} \quad \text{and}\quad  X_{A} \, \independent X_{i'} \implies X_{A}\, \independent X_{B}.
		\end{align*}		
		Since $A\cup B = V, A\cap B=\emptyset$ and $\cG$ is a directed tree (that spans $V$) there exist either an edge $(j''\to i'')\in \cE$ with $j''\in A$ and $i''\in B$ or $j''\in B$ and $i''\in A$. In either case, we have that $X_{i''}\independent X_{j''}$, which contradicts $P_X$ satisfying causal minimality with respect to $\cG$. We conclude that $P_X$ also satisfies causal minimality with respect to the alternative graph $\tilde \cG$.
	\end{quote}
	Hence, the following structural equations hold for $(X_i,X_j)$ and $(\tilde X_i, \tilde X_j)$ 
	\begin{align*} %
		X_i &= f_i(X_j)+N_i,\quad \text{with}\quad X_j \independent N_i,\\ 
		\tilde X_j &= \tilde f_j(\tilde X_i) + \tilde N_j, \quad \text{with} \quad \tilde X_i \independent \tilde N_j,
	\end{align*}
	with $P_{X_j,X_i} = P_{\tilde X_j, \tilde X_i}$. We can apply the same arguments as in the bivariate setup if we can argue that a density of $X_j$  is three times differentiable and that a density of  $\tilde X_i$ is a %
	continuous density. 
	
	To this end, note that the density $p_{X_j}$ is given by the convolution of two densities
	\begin{align}\label{eq:densityIdentifiabilityProof}
		p_{X_j}(y) = \int_{-\i}^{\i}  p_{f_j(X_{\PAg{\cG}{j}})}(t)p_{N_j}(y-t) \, dt , 
	\end{align}
	as $X_j := f_j(X_{\PAg{\cG}{j}})+N_j$ with $X_{\PAg{\cG}{j}}\independent N_j$. Here we used that $f_j(X_{\PAg{\cG}{j}})$ has density with respect to the Lebesgue measure. 
	\begin{quote}
		To realize this note that $f_j\in \cC_3$ and it is nowhere constant. By arguments similar to those in the proof of  \Cref{lm:RestrictedModelConditionGaussian}, this implies that $f'(x)=0$ at only countably many points $(d_k)$. Now let $(O_k)$ be the countable collection of mutually disjoint open intervals that cover $\R$ except for the points $(d_k)$. By continuity of $f'$ we know that $f'(x)$ is either strictly positive or strictly negative on each $O_k$. That is, $f$ is continuously differentiable and strictly monotone on each $O_k$. Thus, $f$ has a continuously differentiable inverse on each $O_k$ by, e.g., the inverse function theorem. This ensures that $f_j(X_{\PAg{\cG}{j}})$ has a density with respect to the Lebesgue measure whenever $X_{\PAg{\cG}{j}}$ does. By starting at the root node $X_{\root{\cG}}=N_{\root{\cG}}$, which by assumption has a density, we can iteratively apply the above argumentation down the directed path from $\root{\cG}$ to $j$ in order to conclude that any $X_j$ for $j\in \{1,\ldots,p\}$ has a density with respect to the Lebesgue measure.
	\end{quote}	Since  $p_{N_j}$ is assumed strictly positive three times continuous differentiable, the representation in \Cref{eq:densityIdentifiabilityProof} furthermore yields that $p_{X_j}$ is three times differentiable; see, e.g., Theorem 11.4 and 11.5  of \cite{schilling2017measures}. 
	
	Now we argue that $\tilde X_i$ has a continuous density. First note that $P_{X_i}$ at least has a continuous density $p_{X_i}$ by arguments similar to those applied for \Cref{eq:densityIdentifiabilityProof}. By the assumption that $P_X = P_{\tilde X}$ we especially have that $P_{X_i}=P_{\tilde X_i}$ which implies that also $\tilde X_i$  has a continuous density. By virtue of the arguments for the bivariate setup we arrive at a contradiction, so it must hold that $P_{X}\not =P_{\tilde X}$.
\end{proofenv}

\begin{proofenv}{\Cref{lm:EntropyScore}}
	Consider an SCM $\tilde \theta =(\tilde \cG,( \tilde f_i),P_{\tilde N})\in \{\tilde \cG\} \times \cD_1^p \times \cP_{\mathrm{G}}^p$ with $\tilde \cG \not = \cG$ and let $Q_{\tilde \theta}$ be the induced distribution.
	As $Q_{\tilde \theta}$ is Markov with respect to $\tilde \cG$ and generated by an additive noise model the density $q_{\tilde \theta}$ factorizes as
	\begin{align*}
		q_{\tilde \theta}(x) = \prod_{i=1}^p q_{\tilde \theta}(x_i|x_{\PAg{\tilde \cG}{i}}) =  \prod_{i=1}^p q_{\tilde N_i}(x_i- \tilde f_i(x_{\PAg{\tilde \cG}{i}})).
	\end{align*}
	The cross entropy between $P_X$ and $Q_{\tilde \theta}$ is then given by
	\begin{align*}
		h(P_X,Q_{\tilde \theta}) &:= \E\lf -\log\lp q_{\tilde \theta}(X)\rp  \rf \\
		&= \sum_{i=1}^p \E \lf  - \log \lp q_{\tilde N_i}\lp X_i-  \tilde f_i (X_{\PAg{\tilde \cG}{i}} ) \rp  \rp \rf \\
		&=  \sum_{i=1}^p  h\lp  X_i-  \tilde f_i (X_{\PAg{\tilde \cG}{i}} ) , \tilde N_i \rp,
	\end{align*}
	where the latter is a sum of the cross entropies between the distribution of $X_i-  \tilde f_i (X_{\PAg{\tilde \cG}{i}} )$ and the distribution of $ \tilde N_i$.
	As $Q_{\tilde \theta}$ is generated by a Gaussian noise structural causal model, we have for all $1\leq i \leq p$ that $\tilde N_i \sim \cN(0,\tilde \sigma_i^2)$ for some $\tilde \sigma_i^2>0$. Hence for all $1\leq i \leq p$,
	\begin{align*}
		h\lp X_i-  \tilde f_i (X_{\PAg{\tilde \cG}{i}} ), \tilde N_i \rp 
		= &\, \E\lf - \log \lp \frac{1}{\sqrt{2\pi}\sigma_i}\exp\lp - \frac{\lp X_i-  \tilde f_i (X_{\PAg{\tilde \cG}{i}} ) \rp^2 }{2\tilde \sigma_i^2} \rp  \rp  \rf \\
		=&\, \log(\sqrt{2\pi}\tilde \sigma_i) + \frac{\E\lf \lp X_i-  \tilde f_i (X_{\PAg{\tilde \cG}{i}} ) \rp^2 \rf }{2\tilde \sigma_i^2}.
	\end{align*}
	Thus, for given set of causal functions $(\tilde f_i)$ and a fixed $i$, the noise variance that minimizes the  cross entropy is given by
	\begin{align*}
		\tilde \sigma_i = \sqrt{\E\lf \lp X_i-  \tilde f_i (X_{\PAg{\tilde \cG}{i}} ) \rp^2 \rf }.
	\end{align*}
	We thus have
	\begin{align*}
		& \inf_{\tilde \sigma_i > 0} \left\{\log(\sqrt{2\pi}\tilde \sigma_i) + \frac{\E\lf \lp X_i-  \tilde f_i (X_{\PAg{\tilde \cG}{i}} ) \rp^2 \rf  }{2\tilde \sigma_i^2} \right\} \\
		&	=\, \log \lp \sqrt{2\pi} \rp + \frac{1}{2}\log\lp \E\lf \lp X_i-  \tilde f_i (X_{\PAg{\tilde \cG}{i}} ) \rp^2 \rf \rp + \frac{1}{2}.
	\end{align*}
	We conclude that
	\begin{align*}
		&\, \inf_{Q\in \{\tilde \cG\} \times \cD_1^p \times \cP_{\mathrm{G}}^p} h(P_X,Q) \\
		=&\, p\log(\sqrt{2\pi}) +\frac{p}{2} +  \sum_{i=1}^p \frac{1}{2} \log\lp \inf_{\tilde f_i \in \cD_1} \E\lf \lp X_i-  \tilde f_i (X_{\PAg{\tilde \cG}{i}} ) \rp^2 \rf \rp.
	\end{align*}
	Finally, as $\cD_1$ is dense in $\cL^2(P_{X_{\PAg{\tilde \cG}{i}}})$, we have that
	\begin{align*}
		\inf_{\tilde f_i \in \cD_1} \E\lf \lp X_i-  \tilde f_i (X_{\PAg{\tilde \cG}{i}} ) \rp^2 \rf &= \E\lf \lp X_i - \E[X_i|X_{\PAg{\tilde \cG}{i}}]) \rp^2 \rf \\
		&\quad + \inf_{\tilde f_i \in \cD_1} \E\lf \lp \E[X_i|X_{\PAg{\tilde \cG}{i}}]-  \tilde f_i (X_{\PAg{\tilde \cG}{i}} ) \rp^2 \rf \\
		&= \E\lf \lp X_i - \E[X_i|X_{\PAg{\tilde \cG}{i}}]) \rp^2 \rf.
	\end{align*}
	Here we used that $X_{\PAg{\tilde \cG}{i}}$ has density with respect to the Lebesgue measure, $P_{X_{\PAg{\tilde \cG}{i}}} \ll \lambda$, and that the density is differentiable (see proof of \Cref{thm:UniqueGraph}). This concludes the first part of the proof. %
	
	For the second statement, we note that for any $Q\in \{\tilde \cG\} \times \cF(\tilde \cG) \times \cP^p$ there exists some noise innovation distribution $P_{\tilde N} \in\ \cP$ such that $Q$ is the distribution of $\tilde X$ generated by structural assignments 
	\begin{align*}
		\tilde X_i :=  \tilde  f_{i}(X_{\PAg{\tilde \cG}{i}}) +\tilde N_i =  \E[X_i|X_{\PAg{\tilde \cG}{i}}] + \tilde N_i ,
	\end{align*}
	for all $1\leq j\leq p$ and mutually independent  noise innovations $\tilde N=(\tilde N_1,\ldots,\tilde N_p)\sim P_{\tilde N} \in \cP^p$. Let $q$ denote the density of $Q$ with respect to the Lebesgue measure and let $q_{\tilde N_i}$ denote the density of $\tilde N_i$ for all $1\leq i \leq p$. As $Q$ is Markov with respect to $\tilde \cG$ and generated by an additive noise model the density factorizes as
	\begin{align*}
		q(x) = \prod_{i=1}^p q(x_i|x_{\PAg{\tilde \cG}{i}}) =  \prod_{i=1}^p q_{\tilde N_i}(x_i- \E[X_i|X_{\PAg{\tilde \cG}{i}} =x_{\PAg{\tilde \cG}{i}}]).
	\end{align*}
	The cross entropy between $P_X$ and $Q$ is given by
	\begin{align*}
		h(P_X,Q) &= \E\lf -\log\lp q(X)\rp  \rf \\
		&= \sum_{i=1}^p \E\lf  - \log \lp q(X_i|X_{\PAg{ \tilde \cG}{i}}) \rp \rf \\
		&= \sum_{i=1}^p \E \lf  - \log \lp q_{\tilde N_i}\lp X_i-  \E\lf X_i|X_{\PAg{\tilde \cG}{i}} \rf \rp  \rp \rf \\
		&=  \sum_{i=1}^p  h\lp X_i-  \E\lf X_i|X_{\PAg{\tilde \cG}{i}} \rf, \tilde N_i \rp.
	\end{align*}
	Note that $h(P,Q) = h(P) + D_{\mathrm{KL}}(P\|Q) \geq h(P)$ with equality if and only if $Q=P$. Thus, the infimum is attained at noise innovations that are equal in distribution to $X_i-  \E[ X_i|X_{\PAg{\tilde \cG}{i}}]$ (which has a density by assumption). That is,
	\begin{align*}
		\inf_{Q\in \{\tilde \cG\} \times \cF(\tilde \cG) \times \cP^p} h(P_X,Q) &= \sum_{i=1}^p \inf_{\tilde N_j \sim P_{\tilde N_j} \in \cP}  h\lp X_i-  \E\lf X_i|X_{\PAg{\tilde \cG}{i}} \rf, \tilde N_i \rp \\
		&=\sum_{i=1}^p   h\lp X_i-  \E\lf X_i|X_{\PAg{\tilde \cG}{i}} \rf \rp \\
		&= \lE(\tilde \cG).
	\end{align*}
\end{proofenv}

\subsection{Proofs of Section~\ref{sec:Method}}

\begin{proofenv}{\Cref{thm:consistency}} Assume that $\theta = (\cG,(f_i),P_N) \in \Theta_R$ with $\cG=(V,\cE)$. For simplicity of the proof, we assume that $\E[X]=0$ such that the edge weight estimators simplify to
	\begin{align*} 
		\hat w_{ji}:= \hat w_{\mathrm{G}}(j \to i)=\hat w_{ji}(\fX_n,\tilde \fX_n) = \frac{1}{2} \log \left(  \frac{\frac{1}{n}\sum_{k=1 }^n \left( X_{k,i} - \hat \phi_{ji} (X_{k,j}) \right)^2}{\frac{1}{n}\sum_{k=1}^n X_{k,i}^2 } \right),
	\end{align*}
	for all $j\not = i$. Furthermore, define the Gaussian population (for $i\not = j$) and auxiliary (for $(j\to i)\not \in \cE$) edge weights by
	\begin{align*}
		w_{ji}:= \frac{1}{2}\log\lp \frac{\E[(X_i-\phi_{ji}(X_j))^2]}{\E[X_i^2]} \rp , \quad w^*_{ji} :=  \frac{1}{2}\log \left( \frac{\E[ (X_i- \tilde \phi_{ji}(X_j))^2]}{\E[X_i^2]}  \right),
	\end{align*}
	respectively, where $\tilde \phi_{ji}:\R \to \R$ is a fixed function satisfying $$\E[(\hat \phi_{ji}(X_j)-\tilde \phi_{ji}(X_j))^2|\tilde \fX_n] \convp_n 0.$$
	Furthermore, for any $\tilde \cG=(V,\tilde \cE)\in \cT_p$ denote 
	\begin{align*}
		\hat w(\tilde \cG) := \sum_{(j\to i)\in \tilde \cE} \hat w_{ji}, \quad  w(\tilde \cG):= \sum_{(j\to i) \in \tilde \cE} w_{ji},  \quad w^*(\tilde \cG) := \sum_{(j\to i) \in \tilde \cE \setminus \cE} w^*_{ji} + \sum_{(j\to i)\in \tilde \cE \cap \cE} w_{ji},
	\end{align*}
	as the total estimated, population and auxiliary edge weights for $\tilde \cG$.  As the conditional expectation minimizes the MSPE among measurable functions, i.e., $\phi_{ji}=\argmin_{f:\R \to \R} \E[(X_i-f(X_j))^2]$, we especially have, for any $i\not = j$, that 
	$$
	\E[ (X_i- \tilde \phi_{ji}(X_j))^2]\geq \E[ (X_i- \phi_{ji}(X_j))^2].
	$$
	This construction entails, for any $\tilde \cG \in \cT_p$, that
	\begin{align} \label{eq:ineqalitywstar}
		w^*(\tilde \cG)\geq w(\tilde \cG), \quad \text{and} \quad w^*(\cG)=w(\cG).
	\end{align}
	\Cref{ass:identifiabilityOfConditionalMeanScores} implies that there exists an $m>0$ such that
	\begin{align} \label{eq:consistencyIdentifiabilityGap}
		\min_{\tilde \cG\in \cT_p\setminus \{\cG\}} \lG(\tilde \cG) - \lG(\cG) = m>0.
	\end{align}
	Thus, for any $\tilde \cG\in \cT_p\setminus \{\cG\}$ it holds that
	\begin{align} \label{eq:inqualityloss}
		\lG(\cG) +\frac{m}{2} \leq \lG(\tilde \cG)-\frac{m}{2},
	\end{align}
	by the identifiability assumption of \Cref{eq:consistencyIdentifiabilityGap}. Now note that $\lG(\tilde \cG) = w( \tilde \cG) + C$ with $C= \sum_{i=1}^p \log(\E[X_i^2])/2$ for all $\tilde \cG\in \cT_p$. Hence, we have, for all $\tilde \cG \in \cT_p \setminus \{\cG\}$, that
	\begin{align*}
		w^*(\cG) - \frac{m}{2} = 	w(\cG)  + \frac{m}{2} \leq w(\tilde \cG) - \frac{m}{2}\leq w^*(\tilde \cG) - \frac{m}{2},
	\end{align*}
	by the equality and inequalities in \eqref{eq:inqualityloss} and \eqref{eq:ineqalitywstar}. Thus, we have that
	\begin{align*}
		P(\hat \cG = \cG) &=  P\left( \argmin_{\tilde \cG =(V,\tilde \cE) \in \cT_p} \sum_{(j\to i) \in \tilde \cE}\hat w_{\mathrm{G}}(j\to i) = \cG	 \right) \\
		&\geq  P \left(    \bigcap_{\tilde \cG\in \cT_p} \left( |\hat w(\tilde \cG) -   w^*(\tilde \cG)| < \frac{m}{2} \right)  \right).
	\end{align*}
	We conclude that it suffices to show that 
	\begin{align*}
		\sup_{\tilde \cG \in \cT_p}| \hat w(\tilde \cG) -w^*(\tilde \cG)| \convp_n 0.
	\end{align*}
	To this end, let $\cE^*:=\{(j \to i): i,j\in V, i\not = j\}\setminus \cE$ and note that 
	\begin{align} \notag
		& \sup_{\tilde \cG \in \cT_p} |	\hat w(\tilde \cG) -w^*(\tilde \cG) | \\ \notag
		\leq &\sup_{\tilde \cG \in \cT_p} \bigg( \sum_{(j\to i )\in \tilde \cE\setminus \cE} \left| \hat w_{ji} -  \frac{1}{2}\log \left( \frac{\E[ (X_i- \tilde \phi_{ji}(X_j))^2]}{\E[X_i^2]}  \right)  \right| \\ \notag
		& \quad \quad  + \sum_{(j\to i )\in \tilde \cE\cap  \cE} \left| \hat w_{ji} -  \frac{1}{2}\log \left( \frac{\E[ (X_i- \phi_{ji}(X_j))^2]}{\E[X_i^2]}  \right)  \right| \bigg) \\ \notag 
		\leq & \sum_{(j\to i )\in  \cE^*} \left| \hat w_{ji} -  \frac{1}{2}\log \left( \frac{\E[ (X_i- \tilde \phi_{ji}(X_j))^2]}{\E[X_i^2]}  \right)  \right| \\ \label{eq:UpperBoundOnSuppLossDifference}
		& \quad \quad  + \sum_{(j\to i )\in \cE} \left| \hat w_{ji} -  \frac{1}{2}\log \left( \frac{\E[ (X_i- \phi_{ji}(X_j))^2]}{\E[X_i^2]}  \right)  \right| .
	\end{align}
	Now consider a fixed term $(j \to i ) \in \cE$ in the second sum of  \eqref{eq:UpperBoundOnSuppLossDifference}. We can upper bound the absolute difference by
	\begin{align} \notag
		&	\, \left| \hat w_{ji} -  \frac{1}{2}\log \left( \frac{\E[ (X_i- \phi_{ji}(X_j))^2]}{\E[X_i^2]}  \right)  \right| \\ \notag
		&\leq  \frac{1}{2}\left|  \log \left(  \frac{1}{n}\sum_{k=1}^n \left( X_{k,i} - \hat \phi_{ji} (X_{k,j}) \right)^2\right) - \log \left( \E[ (X_i- \phi_{ji}(X_j))^2]  \right)   \right| \\ 
		\label{eq:upperboundinConsistencyTheorem}
		&\quad  +  \frac{1}{2}\left| \log(\E[X_i^2])- \log\left( \frac{1}{n}\sum_{k=1}^n X_{k,i}^2 \right) \right|. 
	\end{align}
	In the upper bound of \eqref{eq:upperboundinConsistencyTheorem}, the last absolute difference vanishes in probability due to the law of large numbers and the continuous mapping theorem. The first absolute difference  also vanishes by the following arguments. Note that,
	\begin{align*}
		0 &\leq    \frac{1}{n}\sum_{k=1 }^n \left( X_{k,i} - \hat \phi_{ji} (X_{k,j}) \right)^2 \\
		&=  \frac{1}{n}\sum_{k=1}^n \left( X_{k,i} -  \phi_{ji} (X_{k,j}) \right)^2 +  \frac{1}{n}\sum_{k=1}^n \left( \phi_{ji} (X_{k,j}) - \hat \phi_{ji} (X_{k,j}) \right)^2\\
		&\quad +  \frac{2}{n}\sum_{k=1}^n \left( X_{k,i} - \phi_{ji} (X_{k,j}) \right) \left( \phi_{ji} (X_{k,j}) - \hat \phi_{ji} (X_{k,j}) \right). %
	\end{align*}
	Hence, it holds that
	\begin{align} \notag
		&\, \left| \frac{1}{n}\sum_{k=1}^n \left( X_{k,i} - \hat \phi_{ji} (X_{k,j}) \right)^2  -  \frac{1}{n}\sum_{k=1}^n \left( X_{k,j} - \phi_{ji} (X_{k,j}) \right)^2 \right|	\\ \notag
		=&\, \bigg|   \frac{1}{n}\sum_{k=1}^n \left( \phi_{ji} (X_{k,j}) - \hat \phi_{ji} (X_{k,j}) \right)^2\\ \notag
		&\quad +  \frac{2}{n}\sum_{k=1}^n \left( X_{k,j} - \phi_{ji} (X_{k,j}) \right) \left( \phi_{ji} (X_{k,j}) - \hat \phi_{ji} (X_{k,j}) \right) \bigg|\\\notag
		\leq &\, \frac{1}{n}\sum_{k=1}^n \left( \phi_{ji} (X_{k,j}) - \hat \phi_{ji} (X_{k,j}) \right)^2 \\ \label{eq:upperboundinconsistencyproof}
		&\quad +2 \sqrt{ \frac{1}{n}\sum_{k=1}^n \left( X_{k,j} - \phi_{ji} (X_{k,j}) \right)^2} \sqrt{ \frac{1}{n}\sum_{k=1}^n \left( \phi_{ji} (X_{k,j}) - \hat \phi_{ji} (X_{k,j}) \right)^2},
	\end{align}
	by Cauchy-Schwarz inequality. 
	By the law of large numbers, we have that the first factor of the second term of \eqref{eq:upperboundinconsistencyproof} converges in probability to a constant,
	\begin{align*}
		\frac{1}{n}\sum_{k=1}^n \left( X_{k,j} - \phi_{ji} (X_{k,j}) \right)^2 \convp_n \E[ X_{1,i}-\phi_{ji}(X_{1,j}))^2].
	\end{align*}
	The first term and latter factor of the second term of \Cref{eq:upperboundinconsistencyproof} vanish in probability by assumption. That is, for any $\ep>0$ we have that
	\begin{align*}
		&P \lp \left|\frac{1}{n}\sum_{k=1} \left( \phi_{ji} (X_{k,j}) - \hat \phi_{ji} (X_{k,j}) \right)^2  \right| > \ep \rp \\
		&=  	P \lp \left|\frac{1}{n}\sum_{k=1} \left( \phi_{ji} (X_{k,j}) - \hat \phi_{ji} (X_{k,j}) \right)^2  \right|\land \ep  > \ep \rp \\
		&\leq \frac{\E\left[ \lp  \frac{1}{n}\sum_{k=1}^n \left( \phi_{ji} (X_{k,j}) - \hat \phi_{ji} (X_{k,j}) \right)^2 \rp \land \ep   \right] }{\ep} \\
		&\leq \frac{\E\left[ \E \left[ \left( \phi_{ji} (X_{1,j}) - \hat \phi_{ji} (X_{1,j}) \right)^2 \big|\tilde \fX_n \right]  \land \ep  \right] }{\ep} \\
		& \to_n 0,
	\end{align*}
	using conditional Jensen's inequality ($x\mapsto \min(x,\ep) = x\land \ep$ is concave) and the dominated convergence theorem.
	This proves that
	\begin{align*}
		\frac{1}{n}\sum_{k=1}^n \left( X_{k,j} - \hat \phi_{ji} (X_{k,j}) \right)^2 \convp_n  \E[ X_{1,i}-\phi_{ji}(X_{1,j}))^2].
	\end{align*}
	Thus, we have shown that the second term of \eqref{eq:UpperBoundOnSuppLossDifference} converges to zero in probability. Finally, the above arguments apply similarly to the first term of \Cref{eq:UpperBoundOnSuppLossDifference} by exchanging every $\phi_{ji}$ with $\tilde \phi_{ji}$. We have shown that $\sup_{\tilde \cG \in \cT_p} |	\hat w(\tilde \cG) -w^*(\tilde \cG) |\convp_n 0$, which concludes the proof.
	
\end{proofenv}

\begin{proofenv}{\Cref{thm:ConsistencyVanishing}}
	Assume that for each sample size $n\in \N$ that $\theta_n = (\cG,...) \in \Theta_R$ with $\cG=(V,\cE)$ and identifiability gap
	\begin{align*}
		\min_{\tilde \cG \in \cT_p\setminus \{\cG\}} \ell_{\mathrm{G}}(\cG) - \ell_{\mathrm{G}}(\tilde \cG)= q_n>0,
	\end{align*}
	with $q_n^{-1}=o(\sqrt{n})$.
	For simplicity of the proof, we assume that $\E_{\theta_n}[X]=0$ such that the edge weight estimators simplify to
	\begin{align*} 
		\hat w_{ji}:=	\hat w_{\mathrm{G}}(j \to i)=\hat w_{ji}(\fX_n,\tilde \fX_n) = \frac{1}{2} \log \left(  \frac{\frac{1}{n}\sum_{k=1 }^n \left( X_{k,i} - \hat \phi_{ji} (X_{k,j}) \right)^2}{\frac{1}{n}\sum_{k=1}^n X_{k,i}^2 } \right).
	\end{align*}
	Furthermore, we continue with the notation and population quantities introduced in the proof of \Cref{thm:consistency}, i.e., $w_{ji}= \log(\E_{\theta_n}[(X_i-\E[X_i|X_j])^2])/\E_{\theta_n}[X_i^2])/2$, where we notionally have suppressed the dependence on $n$. 
	We know that for each SCM $\theta_n$ it holds that
	\begin{align*}
		\ell_{\mathrm{G}} (\cG) +q_n\leq \ell_{\mathrm{G}}(\tilde \cG), \quad  \text{hence} \quad w(\cG) + q_n \leq w(\tilde \cG),
	\end{align*}
	for all $\tilde \cG\in \cT_p\setminus \{\cG\}$. Thus,
	\begin{align*}
		&	P_{\theta_n}\lp  \argmin_{\tilde \cG =(V,\tilde \cE) \in \cT_p} \sum_{(j\to i)\in \tilde \cE} \hat w_{ji} = \cG \rp  \\
		&\geq 	P_{\theta_n}\lp \lp |\hat w(\cG)- w (\cG)| <\frac{q_n}{2}\rp \cap \bigcap_{\tilde \cG \in \cT_p\setminus \{\cG\}} \lp\hat w (\tilde \cG)-w(\tilde \cG) \geq  - \frac{q_n}{2}\rp \rp.
	\end{align*}
	For any $\tilde \cG=(V,\tilde \cE)\in \cT_p$ we have that
	\begin{align*}
		\hat w (\tilde \cG) - w(\tilde \cG) &= \sum_{(j\to i) \in \tilde \cE\cap \cE} \hat w_{ji} - w_{ji} + \sum_{(j\to i)  \in \tilde \cE\setminus \cE} \hat w_{ji} - w_{ji},
	\end{align*}
	where $\hat w_{ji}$ and $w_{ji}$ denote the estimated and population Gaussian weights for the edge $(j\to i)$, respectively. Hence, it suffices to show that 
	\begin{align*}
		&\forall (j\to i) \in \cE,\forall \ep>0 : P_{\theta_n}(|\hat w_{ji}-w_{ji}| < q_n\ep )\to_n 1 ,\\
		& \forall (j\to i)\not \in \cE, \forall \ep>0: P_{\theta_n}\lp \hat w_{ji}-w_{ji}\geq - q_n \ep \rp \to_n 1.
	\end{align*}
	To see this, note that if the above statements hold, then
	\begin{align*}
		P_{\theta_n} \lp |\hat w(\cG)- w (\cG)| <  \frac{q_n}{2} \rp &\geq   P_{\theta_n} \lp \sum_{(j\to i)\in \cE}  |\hat w_{ji}- w_{ji}| < \frac{q_n}{2}   \rp \\
		&\geq   P_{\theta_n} \lp \bigcap_{(j\to i) \in \cE} \lp  |\hat w_{ji}- w_{ji}| < \frac{q_n}{2(p-1)} \rp  \rp \\
		& \to_n 1,
	\end{align*}
	and for any $\tilde \cG= (V,\tilde \cE)\in \cT_p$
	\begin{align*}
		P_{\theta_n} \lp \hat w(\tilde \cG)- w (\tilde \cG) \geq - \frac{q_n}{2} \rp &=  P_{\theta_n} \lp  \sum_{(j\to i) \in \tilde \cE\cap \cE} \hat w_{ji} - w_{ji} + \sum_{(j\to i)  \in \tilde \cE\setminus \cE} \hat w_{ji} - w_{ji} \geq - \frac{q_n}{2} \rp \\
		&  \geq  P_{\theta_n}\left(  \bigcap_{(j\to i)\in \tilde \cE \cap \cE} \lp|\hat w_{ji}-w_{ji}| \leq \frac{q_n}{2(p-1)} \rp \right.\\
		&\quad \quad\quad\quad  \left.  \cap \bigcap_{(j\to i) \in \tilde \cE \setminus \cE}\lp \hat w_{ji}-w_{ji} \geq -\frac{q_n}{2(p-1)}\rp  \right) \\
		&\to_n 1,
	\end{align*}
	hence the probability of the intersections also converges to one.

	\paragraph*{The causal edges:} Now fix $(j\to i)\in \cE$. We want to show that for all $\ep>0$ it holds that
	\begin{align*}
		P_{\theta_n}(|\hat w_{ji}-w_{ji}| < q_n\ep )\to_n 1.
	\end{align*}
	First note that
	\begin{align*}
		\left| \hat w_{ji} - w_{ji} \right| &\leq  \frac{1}{2}\left|  \log \left(  \frac{1}{n}\sum_{k=1}^n \left( X_{k,i} - \hat \phi_{ji} (X_{k,j}) \right)^2\right) - \log \left( \E_{\theta_n}[ (X_i- \phi_{ji}(X_j))^2]  \right)   \right| \\ 
		&\quad  +  \frac{1}{2}\left| \log(\E_{\theta_n}[X_i^2])- \log\left( \frac{1}{n}\sum_{k=1}^n X_{k,i}^2 \right) \right|,
	\end{align*}
	where $\hat \phi_{ji}$ for each $n$ is the estimated conditional expectation $x\mapsto \E_{\theta_n}[X_i|X_j=x]$ based on samples from the auxiliary data set.
	It suffices to show the wanted convergence in probability for each of the above terms. Furthermore,  for all sequences of positive random variables $(Z_n)$ and positive constants $c>0$ and for all $\ep>0$ there exists $\delta>0$ such that
	\begin{align*}
		(q_n^{-1}|\log(Z_n)-\log(c)|\geq \ep) \subseteq (q_n^{-1}|Z_n-c|\geq \delta),
	\end{align*}
	for sufficiently large $n$. To see this, note that if $q_n^{-1}(\log(Z_n)-\log(c))\geq \ep$, then $Z_n> \exp(\log(c)+q_n\ep)=c\exp(q_n\ep)\geq c(1+q_n\ep)$, so $q_n^{-1}(Z_n-c)\geq  c\ep$. On the other hand, if $q_n^{-1}(\log(Z_n)-\log(c))\leq -\ep$, then $Z_n \leq  c\exp(-\ep q_n)\leq c(1-\ep q_n + \ep^2 q_n^2)$, so $q_n^{-1}(Z_n-c)\leq -c\ep + c\ep^2 q_n$. In summary, if $q_n^{-1}|\log(Z_n)-\log(c)|\geq \ep$, then $q_n^{-1}|Z_n-c| \geq  c\ep - c\ep^2 q_n> c \ep(1-M)=:\delta$ where $1>M> \ep q_n$ for sufficiently large $n$. We conclude that it suffices to show that for all $\ep >0$ it holds that 
	\begin{align} \label{eq:unifnum}
		P_{\theta_n} \lp \left| \frac{1}{n}\sum_{k=1}^n \left( X_{k,i} - \hat \phi_{ji} (X_{k,j}) \right)^2 - \E_{\theta_n}[ (X_i- \phi_{ji}(X_j))^2]\right | \geq q_n \ep \rp \to_n 0 
	\end{align}
	and that
	\begin{align} \label{eq:unifden}
		P_{\theta_n} \lp 	\left| \frac{1}{n}\sum_{k=1}^n X_{k,i}^2- \E_{\theta_n}[X_i^2]   \right| \geq q_n\ep \rp \to_n 0 ,
	\end{align}
	\Cref{eq:unifden} is satisfied as the summands are mean zero i.i.d. Therefore, with
	\begin{align*}
		W_n :=  \frac{1}{n}\sum_{k=1}^n X_{k,i}^2- \E_{\theta_n}[X_i^2] ,  \end{align*}
	where $\E_{\theta_n}[q_n^{-1}W_n]=0$, we have that  $
	\E_{\theta_n}[ q_n^{-2}W_n^2 ] = \frac{q_n^{-2}}{n} \E_{\theta_n}[(X_{i}^2- \E_{\theta_n}[X_i^2])^2]$, hence
	\begin{align*}
		P_{\theta_n}(q_n^{-1}W_n \geq \ep) &\leq q_n^{-2}\frac{\E_{\theta_n}[W_n^2]}{\ep^2}\\& \leq \frac{q_n^{-2}}{n} \frac{\sup_{n\in \N} \E_{\theta_n}[(X_{i}^2- \E_{\theta_n}[X_i^2])^2]}{\ep^2}\\
		&\to_n 0,
	\end{align*}
	for any $\ep>0$ as $\sup_{n\in \N}\E_{\theta_n}\|X\|_2^4<\i$ and $q_n^{-1} = o(\sqrt{n})$. 
	
	Now we show \Cref{eq:unifnum}. First, we simplify the notation by letting $Z_k:=X_{k,i}$, $Y_k := X_{k,j}$ $f:=\phi_{ji}$ and $\hat f := \hat \phi_{ji}$ for all $k\in \N$. Note that we have suppressed the dependence of $f=\phi_{ji}$ on $\theta_n$. We have that
	
	\begin{align*}
		\frac{1}{n}\sum_{k=1}^n \left( Z_k - \hat f(Y_k) \right)^2 &= 	\frac{1}{n}\sum_{k=1}^n (Z_k-f(Y_k))^2+ 	\frac{1}{n}\sum_{k=1}^n(f(Y_k) -\hat f(Y_k))^2 \\
		&\quad  + 	\frac{2}{n}\sum_{k=1}^n (Z_k-f(Y_k))(f(Y_k) -\hat f(Y_k))\\
		&=: T_{1,n} + T_{2,n} + T_{3,n}.
	\end{align*}
	It suffices to show that for all $\ep>0$ it holds that
	\begin{enumerate}[label=(\alph*)]
		\item $P_{\theta_n} \lp  |T_{1,n} - \E_{\theta_n}[(Z_1 - f(Y_1))^2]| \geq q_n \ep  \rp  \to_n 0,$
		\item $P_{\theta_n} \lp  |T_{2,n}| \geq q_n \ep  \rp  \to_n 0,$ and 
		\item $P_{\theta_n} \lp  |T_{3,n}| \geq q_n \ep  \rp  \to_n 0.$
	\end{enumerate}
	First we show (a). Each term in the sum of $T_{1,n}-\E_{\theta_n}[(Z_1 - f(Y_1))^2]$ is mean zero and i.i.d., i.e., 
	\begin{align*}
		q_n^{-1}\E_{\theta_n}[ \left( Z_k - f(Y_k) \right)^2  -  \E_{\theta_n}[ (Z_1- f(Y_1))^2]] =0.
	\end{align*}
	Furthermore, 
	\begin{align*}
		&\Var_{\theta_n}(q_n^{-1}(T_{1,n}-\E_{\theta_n}[ (Z_1- f(Y_1))^2] ))\\
		=&\Var_{\theta_n}\lp \frac{q_n^{-1}}{n}\sum_{k=1}^n \left( Z_k - f(Y_k) \right)^2  -  \E_{\theta_n}[ (Z_1- f(Y_1))^2] \rp\\
		= &\frac{q_n^{-2}}{n^2}\sum_{k=1}^n \Var_{\theta_n} \lp  \left( Z_k - f(Y_k) \right)^2  -  \E_{\theta_n}[ (Z_1- f(Y_1))^2] \rp \\
		\leq & \frac{q_n^{-2}}{n} \sup_{n\in \N}\Var_{\theta_n}\lp \left( Z_1 - f(Y_1) \right)^2 \rp \\
		\to_n &0,
	\end{align*}
	since $q_n^{-1}=o(\sqrt{n})$ and $\sup_{n\in \N}\E_{\theta_n}\|X\|_2^4<\i$. Hence,
	\begin{align*}
		P_{\theta_n}\lp |q_n^{-1} (T_{1,n}-\E[ (Z_1- f(Y_1))^2] )| \geq \ep  \rp &\leq \frac{\Var_{\theta_n}(q_n^{-1}(T_{,n}-\E[ (Z_1- f(Y_1))^2] ))}{\ep^2} \\
		&\to_n 0.
	\end{align*}
	by Chebyshev's inequality, proving (a).

	Now we show (b). To that end, note that the terms of $T_{2,n}$ is i.i.d.\ conditional on $\tilde \fX_n$. For a fixed $1>\ep>0$ we have
	\begin{align*}
		P_{\theta_n}\lp |q_n^{-1} T_{2,n} | \geq \ep \rp &= \E_{\theta_n}\lf  P_{\theta_n}\lp q_n^{-1} T_{2,n}  \geq \ep |\tilde \fX_n \rp \land 1\rf \\
		&\leq  \frac{\E_{\theta_n}\lf  \E_{\theta_n}\lf q_n^{-1} T_{2,n}   | \tilde \fX_n  \rf \land 1 \rf}{\ep}\\
		&=  \frac{\E_{\theta_n}\lf q_n^{-1} \E_{\theta_n}\lf  (f(Y_1)-\hat f(Y_1))^2  | \tilde \fX_n  \rf \land 1 \rf}{\ep},
	\end{align*}
	where we used the conditional Markov's inequality. Now fix $1>\delta>0$ and define $A_{n,\delta}:= (q_n^{-1} \E_{\theta_n}\lf  (f(Y_1)-\hat f(Y_1))^2  | \tilde \fX_n  \rf >\delta )$ and note that by assumption there exists an $N_{\delta}\in \N$ such that $\forall n\geq N_\delta:	P_{\theta_n}(A_{n,\delta})< \delta$. Hence, for  $n\geq N_{\delta}$ we have that
	\begin{align} \notag
		&\E_{\theta_n}\lf q_n^{-1} \E_{\theta_n}\lf  (f(Y_1)-\hat f(Y_1))^2  | \tilde \fX_n  \rf \land 1\rf \\
		&= \E_{\theta_n}\lf 1_{A_{n,\delta}} 	q_n^{-1} \E_{\theta_n}\lf  (f(Y_1)-\hat f(Y_1))^2  | \tilde \fX_n  \rf \land 1\rf\\ \notag
		&\quad + \E_{\theta_n}\lf 1_{A_{n,\delta}^c} 	q_n^{-1} \E_{\theta_n}\lf  (f(Y_1)-\hat f(Y_1))^2  | \tilde \fX_n  \rf \land 1\rf \\ \notag
		&\leq  \E_{\theta_n}\lf 1_{A_{n,\delta}} 	q_n^{-1} \E_{\theta_n}\lf  (f(Y_1)-\hat f(Y_1))^2  | \tilde \fX_n  \rf \land 1\rf \\ \notag
		&\quad + \E_{\theta_n}\lf 1_{A_{n,\delta}^c} \delta  \rf \\ \notag
		&\leq \E_{\theta_n}\lf 1_{A_{n,\delta}}  \rf + \delta \\ 
		&=  P_{\theta_n}(A_{n,\delta}) + \delta <2 \delta, \label{eq:boundingprobsplit}
	\end{align}
	hence $
	\limsup_{n\to \i}  P_{\theta_n}\lp |q_n^{-1} T_{2,n} | \geq \ep \rp < 2\delta/\ep$, i.e.,  $P_{\theta_n}\lp |q_n^{-1} T_{2,n} | \geq \ep \rp\to 0$ as $\delta>0$ was chosen arbitrarily, proving (b).

	Now we prove (c). To this end, recall that
	\begin{align*}
		T_{3,n}:=\frac{2}{n}\sum_{k=1}^n (Z_k-f(Y_k))(f(Y_k) -\hat f(Y_k)),
	\end{align*}
	is, conditional on $\tilde \fX$, an i.i.d.\ sum with conditional mean zero
	\begin{align*}
		\E_{\theta_n}[T_{3,n}|\tilde \fX_n] &=  2 \E_{\theta_n}[(Z_k-f(Y_k))(f(Y_k) -\hat f(Y_k))|\tilde \fX_n] \\
		&= 2 \E_{\theta_n}[(\E_{\theta_n}[Z_k| Y_k, \tilde \fX_n] -f(Y_k))(f(Y_k) -\hat f(Y_k))|\tilde \fX_n] \\
		&= 2 \E_{\theta_n}[(f(Y_k) -f(Y_k))(f(Y_k) -\hat f(Y_k))|\tilde \fX_n] =0,
	\end{align*}
	and conditional second moment given by
	\begin{align*}
		\E_{\theta_n}[T_{3,n}^2|\tilde \fX_n] &= \frac{4}{n^2} \sum_{k=1}^n \E_{\theta_n}[(Z_k-f(Y_k))^2(f(Y_k) -\hat f(Y_k))^2|\tilde \fX_n] \\
		&=\frac{4}{n} \E_{\theta_n} \lf (Z_k-f(Y_k))^2(f(Y_k)-\hat f(Y_k))^2 |\tilde \fX_n  \rf \\
		&= \frac{4}{n} \E_{\theta_n} \lf \E_{\theta_n}\lf (Z_k-f(Y_k))^2|\tilde \fX_n, Y_k\rf (f(Y_k)-\hat f(Y_k))^2 |\tilde \fX_n  \rf \\
		&=\frac{4}{n} \E_{\theta_n} \lf \Var_{\theta_n}(Z_k|Y_k) (f(Y_k)-\hat f(Y_k))^2 |\tilde \fX_n  \rf \\
		&\leq \frac{C}{n} \E_{\theta_n} \lf (f(Y_k)-\hat f(Y_k))^2 |\tilde \fX_n  \rf,
	\end{align*}
	$P_{\theta_n}$-almost surely. Hence, w.l.o.g. assume that $0<\ep<1$ and note that the conditional Markov's inequality yields
	\begin{align} \notag
		P_{\theta_n}(|q_n^{-1}T_{3,n}| \geq \ep) &= \E_{\theta_n}[P_{\theta_n}(|q_n^{-1}T_{3,n} |\geq\ep|\tilde \fX_n)\land 1]\\ \label{eq:T3reuse}
		&\leq  \frac{1}{\ep^2}\E_{\theta_n}\lf \E_{\theta_n}\lf q_n^{-2}T_{3,n}^2 |\tilde \fX_n  \rf \land 1  \rf \\
		&\leq  \frac{C}{\ep^2}\E_{\theta_n}\lf  \frac{q_n^{-2}}{n} \E_{\theta_n} \lf (f(Y_k)-\hat f(Y_k))^2 |\tilde \fX_n  \rf \land 1 \rf . \notag
	\end{align}
	By conditional Jensen's inequality, we have that
	\begin{align*}
		\E_{\theta_n} \lf (f(Y_k)-\hat f(Y_k))^2 |\tilde \fX_n  \rf 
		&\leq 1+ \E_{\theta_n} \lf (f(Y_k)-\hat f(Y_k))^2 |\tilde \fX_n  \rf^2 \\
		&\leq 1+ \E_{\theta_n} \lf (f(Y_k)-\hat f(Y_k))^4 |\tilde \fX_n  \rf.
	\end{align*}
	Fix $\delta>0$. Let $A_{n,\delta }:= \lp   \frac{q_n^{-2}}{n} \E_{\theta_n} \lf (f(Y_k)-\hat f(Y_k))^4 |\tilde \fX_n  \rf  > \delta \rp$ and note that $P_{\theta_n}(A_{n,\delta}) \to_n 0$, hence there exists an $N_{\delta}\in \N$ such that $\forall n\geq N_{\delta}: P_{\theta_n}(A_{n,\delta})<\delta$. Furthermore, as $q_n^{-1}=o(\sqrt n)$ there exists an $N\in \N$ such that $q_n^{-2}/n < \delta$ for all $n\geq N$. Similar to the arguments in \Cref{eq:boundingprobsplit} we then have that
	\begin{align*}
		\frac{\ep^2}{C}	P_{\theta_n}(|q_n^{-1}T_{3,n}| \geq \ep)  &\leq  \E_{\theta_n}\lf  \frac{q_n^{-2}}{n}\lp 1+ \E_{\theta_n} \lf (f(Y_k)-\hat f(Y_k))^4 |\tilde \fX_n  \rf \rp  \land 1 \rf \\
		&\leq \frac{q_n^{-2}}{n} +  E_{\theta_n}\lf  \frac{q_n^{-2}}{n} \E_{\theta_n} \lf (f(Y_k)-\hat f(Y_k))^4 |\tilde \fX_n  \rf   \land 1 \rf\\
		&\leq \frac{q_n^{-2}}{n} +  \E_{\theta_n}[1_{A_{n,\delta}}] + \E_{\theta_n}[1_{A_{n,\delta}^c}\delta ] \\
		&< \delta +  P_{\theta_n}(A_{n,\delta}) + \delta < 3\delta, 
	\end{align*}
	for any $n \geq N_\delta\lor N$, so $P_{\theta_n}(q_n^{-1}T_{3,n} \geq \ep)\to_n 0$, proving (c).

	\paragraph*{The non-causal edges:} Now fix $(j\to i) \not \in \cE$, we want to show, for any $\ep>0$ that
	\begin{align*}
		P_{\theta_n}( \hat w_{ji} - w_{ji} \geq - q_n\ep ) \to_n 1,
	\end{align*}
	where
	\begin{align*}
		\hat w_{ji} - w_{ji} &= \frac{1}{2}\lp  \left[   \log \left(  \frac{1}{n}\sum_{k=1}^n \left( X_{k,i} - \hat \phi_{ji} (X_{k,j}) \right)^2\right) - \log \left( \E[ (X_i- \phi_{ji}(X_j))^2]  \right)   \right]\right. \\ 
		&\quad  +  \left.\left[ \log(\E[X_i^2])- \log\left( \frac{1}{n}\sum_{k=1}^n X_{k,i}^2 \right) \right] \right) =: \frac{1}{2}(D_{1,n}+ D_{2,n}).
	\end{align*}
	We have that $
	P_{\theta_n}( \hat w_{ji} - w_{ji} \geq - q_n\ep)  \geq P_{\theta_n}\lp \lp D_{1,n} \geq -  q_n\ep\rp \cap \lp |D_{2,n }|<  q_n\ep \rp \rp$, 
	where the second event has already been shown to have probability converging to one in \Cref{eq:unifden}. Thus, it suffices to show that
	\begin{align*}
		P_{\theta_n}\lp  D_{1,n} \geq - q_n\ep \rp \to_n 1.
	\end{align*}
	By similar arguments as above we have for any sequence of positive random variables $(K_n)_{n\geq 1}$ and a positive constant $K$ that for all $\ep>0$ there exists an $\delta>0$ such that $P_{\theta_n}\lp  \log( K_n) - \log(K) < - q_n\ep \rp \leq P_{\theta_n}(K_n-K< -q_n\delta)$, 
	for sufficiently large $n\in \N$. To see this, note that if $\log(K_n)-\log(K)<-q_n\ep$, then $K_n < K\exp(-\ep q_n)\leq K(1-\ep q_n + \ep^2 q_n^2)$, so $q_n^{-1}(K_n-K)<-K\ep + K\ep^2 q_n<- K\ep(1-M)=:-\delta $ where $1>M> \ep q_n$ for sufficiently large $n$, since $q_n \downarrow 0$. Thus, it suffices to show that for any $\ep>0$ it holds that
	\begin{align*}
		P_{\theta_n} \lp   \frac{1}{n}\sum_{k=1}^n \left( X_{k,i} - \hat \phi_{ji} (X_{k,j}) \right)^2-  \E_{\theta_n}[ (X_i- \phi_{ji}(X_j))^2] \geq - q_n \ep   \rp  \to_n 1.
	\end{align*}
	Again, we simplify the notation $Z_k := X_{k,i}$, $Y_k := X_{k,j}$, $f = \phi_{ji}$ and $\hat f := \hat \phi_{ji}$ for all $k\in \N$.  Now define the following terms
	\begin{align*}
		\frac{1}{n}\sum_{k=1}^n \left( Z_k - \hat f(Y_k) \right)^2   %
		&= 	\frac{1}{n}\sum_{k=1}^n (Z_k-f(Y_k))^2\\
		&\quad + 	\frac{1}{n}\sum_{k=1}^n \{(f(Y_k) -\hat f(Y_k))^2-  \delta_{n,\theta_n}^2\} \\
		&\quad  + 	\frac{2}{n}\sum_{k=1}^n \{(Z_k-f(Y_k))(f(Y_k) -\hat f(Y_k)) + \delta_{n,\theta_n}^2/2\}\\
		&=:  T_{1,n}+\tilde T_{2,n}+\tilde T_{3,n},
	\end{align*}
	where $\delta_{n,\theta_n}^2 := \E_{\theta_n}[(f(Y_1)-\hat f (Y_1))^2|\tilde \fX_n] = \E_{\theta_n}[ (\phi_{ji}(X_j)- \hat \phi_{ji}(X_j))^2 |\tilde \fX_n]$. It suffices to show that for all $\ep>0$ it holds that
	\begin{enumerate}[label=(\alph*)]
		\setcounter{enumi}{3}
		\item $P_{\theta_n} \lp  | T_{1,n} - \E_{\theta_n}[(Z_1 - f(Y_1))^2]| \geq q_n \ep  \rp  \to_n 0,$
		\item $P_{\theta_n} \lp  |\tilde T_{2,n}| \geq q_n \ep  \rp  \to_n 0,$ and
		\item $P_{\theta_n} \lp  \tilde T_{3,n} \geq -q_n \ep  \rp  \to_n 1.$
	\end{enumerate}
	Condition (d) holds by arguments similar  to (a) for the causal edges. 
	
	Now we prove (e). 
	The expansion, conditional on $\tilde \fX_n$, is a sum of mean zero i.i.d. terms, hence
	\begin{align*}
		\E_{\theta_n} \lp \left. q_n^{-2}\tilde T_{2,n}^2 \right| \tilde \fX_n\rp 	&=\frac{q_n^{-2}}{n} \E_{\theta_n}\left[  \{(f(Y_k) -\hat f(Y_k))^2 - \delta_{n,\theta_n}^2 \}^2| \tilde \fX_n \right] \\
		&= \frac{q_n^{-2}}{n} \E_{\theta_n}\left[ (f(Y_k) -\hat f(Y_k))^4 + (\delta_{n,\theta_n}^2 )^2 -2(f(Y_k) -\hat f(Y_k))^2\delta_{n,\theta_n}^2| \tilde \fX_n \right] \\
		&=\frac{q_n^{-2}}{n} \lp\E_{\theta_n}\left[ (f(Y_k) -\hat f(Y_k))^4 | \tilde \fX_n \right] - (\delta_{n,\theta_n}^2)^2 \rp \\ 
		&\leq \frac{q_n^{-2}}{n}  \E_{\theta_n}\left[ (f(Y_k) -\hat f(Y_k))^4 | \tilde \fX_n \right],
	\end{align*}
	using that  $(\delta_{n,\theta_n}^2)^2\geq 0
	$.  Fix $1>\delta>0$ and let $$A_{n,\delta }:= \lp   \frac{q_n^{-2}}{n} \E_{\theta_n} \lf (f(Y_k)-\hat f(Y_k))^4 |\tilde \fX_n  \rf  > \delta \rp,$$ and note that there exists an $N_{\delta}\in \N$ such that $\forall n\geq N_{\delta}: P_{\theta_n}(A_{n,\delta})<\delta$.  Similar to the previous arguments we have for any $1 >\ep > 0$ and $n\geq N_\delta$  that
	\begin{align*}
		P_{\theta_n}\lp \left|  \tilde T_{2,n}  \right| \geq  q_n\ep \rp 
		&= \E_{\theta_n} \left[ P_{\theta_n}\lp\left. \left|   	q_n^{-1}\tilde T_{2,n}  \right|  \geq  \ep \right | \tilde \fX_n \rp \land 1 \right] \\
		&\leq\frac{ 1 }{\ep^2} \E_{\theta_n}\lf \E_{\theta_n}\lf q_n^{-2}\tilde T_{2,n}^2 |\tilde \fX_n\rf \land 1\rf \\
		&\leq \frac{ 1}{\ep^2} \E_{\theta_n}\lf \frac{q_n^{-2}}{n}  \E_{\theta_n}\left[ (f(Y_k) -\hat f(Y_k))^4 | \tilde \fX_n \right] \land 1\rf  \\
		&\leq \frac{1}{\ep^2} \lp \E_{\theta_n}[1_{A_{n,\delta}}] +  \E_{\theta_n}[1_{A_{n,\delta}^c}\delta] \rp < \frac{2\delta }{\ep^2},
	\end{align*}
	by the conditional Markov's inequality.  Since $\delta>0$ was chosen arbitrarily, we conclude that (e) holds.
	
	Finally we show (f). Recall that in the analysis of the causal edges, we defined
	\begin{align*}
		T_{3,n}:=\frac{2}{n}\sum_{k=1}^n (Z_k-f(Y_k))(f(Y_k) -\hat f(Y_k)).
	\end{align*}
	Hence, we have that $\tilde T_{3,n} =   T_{3,n} + \delta_{n,\theta_n}^2$. We realize that for any $0<\ep<1$
	\begin{align*}
		P_{\theta_n}( \tilde T_{3,n}  < -q_n\ep)&\leq P_{\theta_n}(  T_{3,n} + \delta_{n,\theta_n}^2  \leq -q_n\ep) \\&=  P_{\theta_n}\lp   T_{3,n} \leq -\lp q_n\ep +\delta_{n,\theta_n}^2 \rp \rp \\
		& \leq P_{\theta_n}\lp    T_{3,n}^2 \geq \lp q_n\ep +\delta_{n,\theta_n}^2 \rp^2 \rp \\
		&\leq   P_{\theta_n}\lp   T_{3,n}^2 \geq \lp q_n\ep\rp^2 \rp \\
		& = P_{\theta_n}\lp q_n^{-2}   T_{3,n}^2 \geq \ep^2 \rp \\
		&= \E_{\theta_n}\left[ P_{\theta_n}\lp q_n^{-2}  T_{3,n}^2  \geq \ep^2|\tilde \fX_n \rp \land 1 \right]\\
		&\leq \frac{1 }{\ep^2}\E_{\theta_n}\left[ \E_{\theta_n}\left[ \left. q_n^{-2}  T_{3,n}^2   \right| \tilde \fX_n \right] \land 1\right]  \\
		& \to_n 0,
	\end{align*}
	where we used the convergence shown in the proof of (c); see \Cref{eq:T3reuse}. To see that the former arguments apply to non-causal edges, simply note that they did not use any conditions restricted to causal edges.
	This concludes the proof.

\end{proofenv}

\subsection{Proofs of Section~\ref{sec:HypothesisTest}}
\begin{lemma} \label{lm:conditionaltriangularCLT}
	Consider an i.i.d.\ sequence $(X_m)_{m\geq 1}$ of random variables with $X_{m}\in \R^d$ independent from a random infinite sequence $\tilde \fX \in \prod_{i=1}^\i \R^{d}$. Let $(\psi_n)_{n\geq 1}$ be a sequence of measurable functions s.t. for all $n\geq 1$, $\psi_n:\R^d \times (\prod_{i=1}^\i \R^d) \to \R^q$ satisfies the following conditions: 
	\begin{itemize}
		\item[(a)] $\E[\psi_n(X_{m},\tilde \fX)|\tilde \fX]=0$ almost surely, 
		\item[(b)] $\exists \, \Sigma \in \R^{q\times q}:\sum_{m=1}^n \Var(\psi_n(X_{m},\tilde \fX)|\tilde \fX) \convp_n \Sigma$, and 
		\item[(c)]  $\exists\,  \ep>0:\sum_{m=1}^n \E[\|\psi_n(X_{m},\tilde \fX)\|_2^{2+\ep}|\tilde \fX] \convp_n 0$.
	\end{itemize}
	It holds that
	\begin{align*}
		\sum_{m=1}^{n}\psi_{n}(X_{m},\tilde \fX)  \convd_n \cN(0,\Sigma),
	\end{align*}
\end{lemma}
\begin{proofenv}{\Cref{lm:conditionaltriangularCLT}}
	Let the random sequences be defined on a common probability space $(\Omega,\cF,P)$ and define
	\begin{align*}
		A_{nm} &:= \E[\psi_n(X_{m},\tilde \fX)|\tilde \fX], \\
		B_n &:= \Sigma- \sum_{m=1}^n \Var(\psi_n(X_{m},\tilde \fX)|\tilde \fX) , \\
		C_n &:= \sum_{m=1}^n \E[\|\psi_n(X_{m},\tilde \fX)\|_2^{2+\ep}|\tilde \fX]. 	
	\end{align*}
	By assumption we have that $P(\cap_{n,m}(A_{nm} =0))=1, B_n \convp 0$ and $C_n \convp 0$ as $n \to \infty$. First, note that for any subsequence $(n_k)_{k\geq 1}$ of the positive integers, there exists a subsequence $(n_{k_l})_{l\in \N}$ such that 
	\begin{align*}
		P(\lim_{l\to \i} B_{n_{k_l}}=0)=1 \quad \text{for} \quad 	(\lim_{l\to \i} B_{n_{k_l}}=0) &:= \{\omega \in \Omega: \lim_{l\to \i} B_{n_{k_l}}(\omega)=0\},
	\end{align*}
	and
	\begin{align*}
		P(\lim_{l\to \i} C_{n_{k_l}}=0)=1 \quad \text{for} \quad (\lim_{l\to \i} C_{n_{k_l}}=0) &:= \{\omega \in \Omega: \lim_{l\to \i} C_{n_{k_l}}(\omega)=0\}.
	\end{align*}
	Thus, define
	\begin{align*}
		G:=(\cap_{n,m}(A_{nm} =0) \cap (\lim_{l\to \i} B_{n_{k_l}}=0) \cap (\lim_{l\to \i} C_{n_{k_l}}=0)) \subseteq \Omega, \quad \text{with} \quad P(G)=1.
	\end{align*}
	Now fix $\tilde x\in \tilde \fX(G) := \{ \tilde \fX (\omega)\in \prod_{j=1}^\i \R^{d}:  \omega \in G \}$  and note that
	\begin{align*}
		&\forall l\geq 1, \forall 1\leq m \leq n_{k_l}: \E[\psi_{n_{k_l}}(X_{m},\tilde x)] =0,\\
		& \sum_{m=1}^{n_{k_l}} \Var(\psi_{n_{k_l}}(X_{m},\tilde x)) \to_l \Sigma, \text{ and} \\
		& \sum_{m=1}^{n_{k_l}} \E[\|\psi_{n_{k_l}}(X_{m},\tilde x)\|_2^{2+\ep}] \to_l 0.
	\end{align*}
	Furthermore, for any $l\geq1$
	\begin{align*}
		\psi_{n_{k_l}}(X_{1},\tilde x),..., \psi_{n_{k_l}}(X_{n_{k_l}}, \tilde x), \quad \text{are jointly independent},
	\end{align*}
	hence by Lyapunov's central limit theorem  for triangular arrays (see, e.g., \citealp{van2000asymptotic}, Proposition 2.27, and recall that Lyapunov's condition implies the Lindeberg--Feller condition) that
	\begin{align*}
		\sum_{m=1}^{n_{k_l}}\psi_{n_{k_l}}(X_{m},\tilde x)  \convd_l Z\sim  \cN(0,\Sigma).
	\end{align*}
	The above convergence in distribution is equivalent to the following statement: for any continuous bounded function $g:\R^{q}\to \R$ it holds that
	\begin{align*}
		\lim_{l\to \i} \E\left[ g\lp\sum_{m=1}^{n_{k_l}}\psi_{n_{k_l}}(X_{m},\tilde x) \rp
		\right] = \E\left[ g(Z) \right].
	\end{align*}
	Fix a continuous and bounded $g$ and note that the above convergence holds for all $\tilde x \in\tilde \fX(G)$ with $P(G)=1$. Thus, it must hold that
	\begin{align*}
		\E\left[ g\lp \sum_{m=1}^{n_{k_l}}\psi_{n_{k_l}}(X_{m},\tilde \fX) \rp \big| \tilde \fX
		\right] \convas_l  \E\left[ g(Z) \right].
	\end{align*}
	Finally, as $(n_{k_l})_{l\geq 1}$ is a  subsequence of an arbitrary  subsequence of positive integers, we have that
	\begin{align*}
		\E\left[ g\lp \sum_{m=1}^{n}\psi_{n}(X_{m},\tilde x) \rp \big| \tilde \fX
		\right] \convp_n   \E\left[ g(Z) \right],
	\end{align*}
	and since $g$ is bounded the dominated convergence theorem yields that
	\begin{align*}
		&\E\left[ g\lp \sum_{m=1}^{n}\psi_{n}(X_{m},\tilde \fX) \rp  \right]\\
		=&\E\left[\E\left[ g\lp \sum_{m=1}^{n}\psi_{n}(X_{m},\tilde \fX) \rp \big| \tilde \fX
		\right] \right] \to_n   \E\left[ g(Z) \right].
	\end{align*}
	As $g$ was chosen arbitrarily, the above convergence holds for any continuous bounded $g$. We conclude that
	\begin{align*}
		\sum_{m=1}^{n}\psi_{n}(X_{m},\tilde \fX)  \convd_n \cN(0,\Sigma),
	\end{align*}
	proving the theorem.
\end{proofenv}
\begin{lemma}[\citealp{shah2020hardness}, Lemma 19] \label{lm:uniformwlln}
	Let $\mathcal{P}$ be a family of distributions for a random variable $\zeta \in \mathbb{R}$ and suppose $\zeta_{1}, \zeta_{2}, \ldots$ are i.i.d.\ copies of $\zeta$. For each $n \in \mathbb{N}$ let $S_{n}=n^{-1} \sum_{i=1}^{n} \zeta_{i} .$ Suppose that for all $P \in \mathcal{P}$ we have $\mathbb{E}_{P}(\zeta)=0$ and $\mathbb{E}_{P}\left(|\zeta|^{1+\eta}\right)<c$ for some $\eta, c>0 .$ We have that for all $\epsilon>0$,
	$$
	\lim _{n \rightarrow \infty} \sup _{P \in \mathcal{P}} P \left(\left|S_{n}\right|>\epsilon\right)=0.
	$$
\end{lemma}
\begin{lemma}\label{lm:uniformwllntotriangular}
	Let $U$ be a random element and let $(Z_{n})_{n\geq 1}$ be an i.i.d. sequence of random variables such that $U\independent (Z_n)_{n\geq 1}$ and let  $\left((W_{nm})_{m\leq  n}\right)_{ n \geq 1}$ be a triangular array of random variables and $(g_n)_{n\geq 1}$ be measurable mappings with the following properties:
	\begin{enumerate}
		\item[(a)] $\forall n\geq 1,\forall m\leq n:W_{nm}=g_n(Z_{m},U)$,
		\item[(b)] $\exists \eta>0:\mathbb{E}\left(\left|W_{n 1}\right|^{1+\eta} \mid U\right)=O_{p}(1)$, as $n\to \i$.
	\end{enumerate}
	Then, writing $\bar{W}_{n}:=\sum_{m=1}^{n} W_{n m} / n$, we have $$\left|\bar{W}_{n}-\mathbb{E}\left(W_{n1} \mid U\right)\right| \convp_n 0 .$$
\end{lemma}
\begin{proofenv}{\Cref{lm:uniformwllntotriangular}} Denote
	\begin{align*}
		j_n(Z_{m},U) :=	g_n(Z_{m},U)-\E[g_n(Z_{1},U)|U],
	\end{align*}
	for any  $n\geq 1$ and $m\leq n$. Let $\delta>0$ be given. Pick $M>0$ and $N\in \N$ such that the events
	$$
	\Omega_{n}:=\left\{\mathbb{E}\left[\left|g_n(Z_{1},U)\right|^{1+\eta} \mid U\right] \leq M\right\},
	$$
	satisfy $\mathbb{P}\left(\Omega_{n}^{c}\right)<\delta$ for $n\geq N$. Notice that 
	\begin{align*}
		U(\Omega_n) = \left\{\tilde u_n : \mathbb{E}\left[\left|g_n(Z_{1},\tilde u_n)\right|^{1+\eta} \right] \leq M\right\},
	\end{align*}
	since $U \independent (Z_n)_{n\geq 1}$. 	Fix $\ep>0$. Then, for all $n\geq N$
	\begin{align*}
		P\left(\left|\bar{W}_{n}-\mathbb{E}\left(W_{n} \mid U\right)\right|>\epsilon\right) &= P\lp \left| \frac{1}{n}\sum_{m=1}^n j_n(Z_{m},U)\right|> \ep \rp\\
		&<\mathbb{E}\left[P\lp \left| \frac{1}{n}\sum_{m=1}^n j_n(Z_{m},U)\right|> \ep\mid U \rp 1_{\Omega_{n}}\right]+\delta.
	\end{align*}
	By the dominated convergence theorem, the first term on the RHS converges to 0 if
	\begin{align*}
		&\,\sup _{\omega \in \Omega_{n}} P\lp \left| \frac{1}{n}\sum_{m=1}^n j_n(Z_{m},U)\right|> \ep\mid U \rp(\omega)  \\
		&\, = \sup _{\tilde u_n \in U(\Omega_{n})} P\lp \left| \frac{1}{n}\sum_{m=1}^n j_n(Z_{m},\tilde u_n)\right|> \ep \rp \rightarrow_n 0,
	\end{align*}
	which implies the desired statement as $\delta>0$ was chosen arbitrarily. Now note that  for any $n\in \N,\tilde u_n \in U(\Omega_n)$ and all $m\in \N$ it holds that
	\begin{align*}
		\E[|j_n(Z_{m},\tilde u_n)|^{1+\eta}]&= \E[|g_n(Z_{m},\tilde u_n)-\E[g_n(Z_{1},\tilde u_n)]|^{1+\eta}]\\
		&\leq 2^{\eta}\lp \E[|g_n(Z_{m},\tilde u_n)|^{1+\eta}]+|\E[g_n(Z_{1},\tilde u_n)]|^{1+\eta} \rp \\
		&\leq 2^{\eta}\lp \E[|g_n(Z_{m},\tilde u_n)|^{1+\eta}]+\E[|g_n(Z_{1},\tilde u_n)|^{1+\eta}] \rp \\
		&< 2^{\eta+1}M =:c
	\end{align*}
	by the cr and Jensen's inequalities, and
	\begin{align*}
		\E[j_n(Z_{m},\tilde u_n)]=0.
	\end{align*}
	For any $n\in \N$, define the following set of pushforward measures
	\begin{align*}
		\cP_n := \{ P' = (j_n(Z_1,\tilde u_n))(P):\tilde u_n\in U(\Omega_n)\}.
	\end{align*}
	For any $P'\in \cP_n$, let $(Y_{m})_{m\geq 1}$  be a sequence of i.i.d.\ random variables such that  $Y_1 \eqd j_n(Z_{1},\tilde u_n)$ for some $\tilde u_n \in U(\Omega_n)$. Notice that  for all $n\in \N$ and $P'\in \cP_n$ it holds that $\E_{P'}|Y_1|^{1+\eta}< c$ and $\E_{P'}[Y_1]=0$. Thus,
	\begin{align*}
		\sup_{\tilde u_n \in U(\Omega_n)}
		P \lp \left|\frac{1}{n}\sum_{m=1}^n j_n (Z_{m},\tilde u_n)\right|>\ep  \rp  & = 
		\sup_{P'\in \cP_n} P' \lp \left|\frac{1}{n}\sum_{m=1}^n Y_m \right|>\ep  \rp\\
		&\leq
		\sup_{P'\in \cup_k \cP_k} P' \lp \left|\frac{1}{n}\sum_{m=1}^n Y_m \right|>\ep  \rp \\
		&\to_n 0,
	\end{align*}
	by the weak uniform law of large numbers, \Cref{lm:uniformwlln}.
\end{proofenv}
\begin{restatable}[Asymptotic normality of edge weight components]{lemma}{thmasympnormal}
	\label{thm:asymptoticnormalityedgecomponents}
	Let for each sample size $n\in\N$, $\hat \phi^n_{ji}$ denote the estimated conditional mean function $\phi_{ji}$ based on the auxiliary sample $\tilde \fX_n$.
	For any $j\not = i$ and $m\leq n$, define
	\begin{alignat*}{4}
		&\hat R_{nm,ji}
		:= \{X_{m,i}-\hat \phi^n_{ji}(X_{m,j})\},\quad \quad \quad \quad 
		&&\hat \mu_{n,ji}
		:= \frac{1}{n}\sum_{m=1}^n \hat R_{nm,ji}^2, \\
		&R_{m,ji}
		:= \{X_{m,i}- \phi_{ji}(X_{m,j})\},&&
		\mu_{ji} := \E [R_{1,ji}^2], \\
		& \hat V_{m,i} := \left(X_{m,i} - \frac{1}{n}\sum_{k=1}^n X_{k,i} \right)^2 , && \hat \nu_{n,i}:= \frac{1}{n}\sum_{m=1}^n \hat V_{m,i},
		\\
		&  \nu_{i} := \Var(X_{1,i}) ,&&  \delta_{n,ji}^2 :=   \E[(\hat \phi^n_{ji}(X_{1,j})- \phi_{ji}(X_{1,j}))^2|\tilde \fX_n] .
	\end{alignat*}	
	Let
	\begin{align*}
		\widehat \Sigma_n &:= \begin{bmatrix}
			\widehat \Sigma_{n,R} & \widehat \Sigma_{n,RV} \\ \widehat \Sigma_{n,RV}^\t & \widehat \Sigma_{n,V}  
		\end{bmatrix}:=\frac{1}{n}\sum_{m=1}^n \begin{bmatrix}
			\hat R_{nm}^2
			(\hat R_{nm}^2)^\t -\hat \mu_n\hat \mu_n^\t & \hat R_{nm}^2  \hat V_{m}^\t- \hat \mu_n \hat \nu_{n}^\t  \\
			\hat V_{m}(\hat R_{nm}^2)^\t- \hat \nu_{n}\hat \mu_n^\t &  \hat V_{m}\hat V_{m}^\t- \hat \nu_{n} \hat \nu_n^\t
		\end{bmatrix}, %
	\end{align*}
	denote the $p^2\times p^2$ matrix empirical covariance matrix, 
	where the squaring of vectors means that each entry is squared.
	Suppose there exists $\xi > 0$ such that for all $j\not =i$, the following three conditions hold:
	\begin{itemize}
		\item[(i)] $\E \|X\|^{4 + \xi} < \infty$.
		\item[(ii)] $\E[|\hat \phi^n_{ji}(X_{j})- \phi_{ji}(X_{j})|^{4+\xi}|\tilde \fX_n] = O_p(1)$, as $n\to \i$.
		\item[(iii)] $\exists \Sigma\in \R^{p^2\times p^2}:\Var\lp  \begin{bmatrix}
			\hat R_{n1}^2 - \delta_{n}^2 - \mu  \\
			\hat V_{1}  - \nu 
		\end{bmatrix} \bigg| \tilde \fX_n \rp  \convp_n \Sigma$, where $\Sigma$ is  constant.
	\end{itemize} 
	Then we have that $\widehat \Sigma_n \convp \Sigma\in \R^{p^2\times p^2}$ and 
	\begin{align}  \label{eq:AsymptoticNormalityOfEdgeComponents}
		\frac{1}{\sqrt{n}}\sum_{m=1}^n  \begin{bmatrix}
			\hat R_{nm}^2 - \delta_{n}^2 - \mu  \\
			\hat V_{m}  - \nu 
		\end{bmatrix} = \sqrt{n}\begin{bmatrix}
			\hat \mu_n - \delta_n^2 - \mu \\
			\hat \nu_n - \nu
		\end{bmatrix} \convd \cN(0,\Sigma).
	\end{align}
\end{restatable}
\begin{proofenv}{\Cref{thm:asymptoticnormalityedgecomponents}}
	We prove the lemma under the assumption that $\E[X]=0$ under which the variance estimator simplify to $\hat V_{m,i}:= X_{m,i}^2$ and $\hat \nu_{n,i} := \frac{1}{n}\sum_{m=1}^n \hat V_{m,i}$ for all $1\leq i \leq p$. The proof only gets more notionally cumbersome without this assumption. It should follow in all generality by applying  expansion techniques and Slutsky's theorem similar to the standard arguments showing asymptotic normality of the regular sample variance.
	
	Let $\tilde \fX$ denote the auxilliary i.i.d.\ process such that $\tilde \fX_n$ is the first $n$-coordinates of said process.
	Note that conditioning $\hat \phi_{ji}^n$ on $\tilde \fX$ it is equivalent to conditioning on $\tilde \fX_n$ by the i.i.d.\ structure of $\tilde \fX$ and that $\hat \phi_{ji}^n$ only depends on $\tilde \fX_n$. 		First, we define for all $j\not = i$, $n\in \N$   and $m\leq n$ the following conditional expectation regression error $\hat \delta_{nm,ji}  := \{\phi_{ji}(X_{m,j})- \hat \phi_{ji}^n(X_{m,j})\}$.
	Furthermore, for each $n\in \N$ and $m\leq n$  define
	\begin{align*}
		\Psi_{n}(X_{m},\tilde \fX) :=   \begin{bmatrix}
			\hat R_{nm}^2 - \delta_{n}^2 - \mu  \\
			\hat V_{m}  - \nu 
		\end{bmatrix}\in \R^{p^2},
	\end{align*}
	where only $\tilde \fX_n$ (containing the first $n$ coordinates of $\tilde \fX$) is used, and
	\begin{align*}
		\psi_n(X_{m},\tilde \fX) &:= \frac{1}{\sqrt{n}} \Psi_{n}(X_{m},\tilde \fX).
	\end{align*}
	Note that the desired conclusion of  \Cref{eq:AsymptoticNormalityOfEdgeComponents} follows by verifying condition (a), (b) and (c) of \Cref{lm:conditionaltriangularCLT}. First, we show (a), the conditional mean zero condition. To that end, note that for any $i\in\{1,\ldots,p\}$ and $j\in\{1,\ldots,p\}\setminus \{i\}$ it holds that
	\begin{align*}
		\hat R_{nm,ji}^2 &=(X_{m,i}-\phi_{ji}(X_{m,j})+ \phi_{ji}(X_{m,j})-\hat \phi_{ji}^n(X_{m,j}))^2 \\
		&=(R_{m,ji}+ \hat \delta_{nm,ji})^2  \\
		&= R_{m,ji}^2 + \hat \delta_{nm,ji}^2 + 2R_{m,ji} \hat\delta_{nm,ji}.
	\end{align*} 
	Hence, we have that
	\begin{align} \label{eq:asympnormalityexpansion}
		\hat R_{nm,ji}^2- \mu_{ji} - \delta_{n,ji}^2 &=  (R_{m,ji}^2-\mu_{ji}) + (\hat \delta_{nm,ji}^2-\delta_{n,ji}^2) +  2R_{m,ji}\hat \delta_{nm,ji}.
	\end{align}
	The terms of \Cref{eq:asympnormalityexpansion} are mean zero conditionally on $\tilde \fX$, since $\E[R_{m,ji}^2|\tilde \fX]=\E[R_{m,ji}^2]=\mu_{ji}$, $\E[ \hat\delta_{nm,ji}^2|\tilde \fX]=\delta_{n,ji}^2$ and
	\begin{align*}
		\E[R_{m,ji} \hat\delta_{nm,ji}|\tilde \fX] &= \E[\E[R_{m,ji} \hat \delta_{nm,ji}|\tilde \fX,X_{m,j}]| \tilde \fX] \\
		&=  \E[\E[X_{m,i}- \phi_{ji}(X_{m,j})|\tilde \fX,X_{m,j}] \hat \delta_{nm,ji} | \tilde \fX] \\
		&=\E[(\E[X_{m,i}|X_{m,j}]- \phi_{ji}(X_{m,j})) \hat \delta_{nm,ji} |\tilde \fX] \\
		&=0,
	\end{align*}
	as $\phi_{ji}(X_{m,j})=\E[X_{m,i}|X_{m,j}]$ almost surely. Furthermore,
	\begin{align*}
		\E[ X_{m,i}^2-\Var(X_i)|\tilde \fX] = \E[ X_{m,i}^2]-\Var(X_i) =0.
	\end{align*}
	We conclude that
	\begin{align*}
		\E[\psi_n(X_{m},\tilde \fX)|\tilde \fX] = \frac{1}{\sqrt{n}}	\E\left[\begin{bmatrix}
			\hat R_{nm}^2 - \delta_{n}^2 - \mu  \\
			\hat V_{m}  - \nu 
		\end{bmatrix} \bigg|\tilde \fX\right] =0,
	\end{align*}
	almost surely. With respect to (b), convergence of the sum of variances, we have, by assumption, that
	\begin{align} 
		\Sigma_n :=	\begin{bmatrix}
			\Sigma_{n,R} & 	\Sigma_{n,RV}\\
			\Sigma_{n,RV}^\t & \Sigma_{n,V}
		\end{bmatrix} :=	\Var\lp \Psi_{n}(X_{1},\tilde \fX) | \tilde \fX\rp 
		\convp_n \Sigma, \label{eq:claimConvergenceOfInfeasibleCovarianceMatrix}
	\end{align}
	where $\Sigma$ is a positive semi-definite matrix. Furthermore, we have that
	$(X_m)_{m\geq 1}$ is an i.i.d.\ sequence independent of $\tilde \fX$. Therefore, 
	\begin{align*}
		\sum_{m=1}^n \Var(\psi_n(X_{m},\tilde \fX)| \tilde \fX) &= \sum_{m=1}^n \frac{1}{n}  \Var(\Psi_{n}(X_{m},\tilde \fX)|\tilde \fX) \\
		&= \sum_{m=1}^n \frac{1}{n} \Sigma_{n} \\
		&= \Sigma_{n} \\
		& \convp_n \Sigma.
	\end{align*}
	Finally, we show that condition (c), a conditional Lindeberg-Feller condition, is fulfilled. To this end, note that with $\ep:=\xi/2>0$ we have that
	\begin{align}
		& \,\, \E\lf \|\psi_n(X_{m},\tilde \fX)\|^{2+\ep}_2 \big| \tilde \fX \rf \notag \\
		&= \E\lf \left \|  \frac{1}{\sqrt n} \begin{bmatrix}
			\hat R_{nm}^2 - \delta_{n}^2 - \mu  \\
			\hat V_{m}  - \nu 
		\end{bmatrix} \right\|_2^{2+\ep} \bigg| \tilde \fX \rf \\
		& = \frac{1}{n^{\frac{2+\ep}{2}}}  \E\lf \left\|\begin{bmatrix}
			\hat R_{nm}^2 - \delta_{n}^2 - \mu  \\
			\hat V_{m}  - \nu 
		\end{bmatrix}\right\|_2^{2+\ep} \bigg| \tilde \fX \rf\notag \\
		&\leq \frac{1}{n^{\frac{2+\ep}{2}}}  2^{(\frac{2+\ep}{2}-1)}  \bigg( \sum_{i\not = j} \E\lf |\hat R_{nm,ji}^2 -  \mu_{ji}-  \delta_{n,ji}^2|^{2+\ep} | \tilde \fX  \rf \notag\\
		&\quad\quad\quad\quad\quad\quad\quad\quad  + \sum_{i=1}^p \E|X_{m,i}^2-\Var(X_i)|^{2+\ep}\bigg), \label{eq:upperboundforcondmomentpsi}
	\end{align}
	by the cr inequality. We now realize that 
	the second factor of \Cref{eq:upperboundforcondmomentpsi} is stochastically bounded. To see this, note that for any $j\not = i$ it holds that
	\begin{align}
		\E\lf |\hat R_{nm,ji}^2 -  \mu_{ji}-  \delta_{n,ji}^2|^{2+\ep} | \tilde \fX  \rf & \leq 2^{1+\ep} (\E[|\hat R_{nm,ji} |^{4+2\ep}|\tilde \fX] + \mu_{ji}^{2+\ep} + \E[|\delta_{n,ji}^2(\tilde \fX) |^{2+\ep}|\tilde \fX]).
		\label{eq:FellerCondUpperBoundSecondFactor}
	\end{align}
	The first term of the upper bound in \Cref{eq:FellerCondUpperBoundSecondFactor} is $O_p(1)$,
	\begin{align*}
		\E[|\hat R_{nm,ji} |^{4+2\ep}|\tilde \fX]&= 	\E[ |X_{m,i}-\hat \phi_{ji}^n(X_{m,j})|^{4+2\ep}|\tilde \fX]\\ & \leq 2^{3+2\ep} ( \E|X_{m,i}-\phi_{ji}(X_{m,j})|^{4+2\ep}  + \E[ |\phi_{ji}(X_{m,i})-\hat \phi_{ji}^n(X_{m,j})|^{4+2\ep}|\tilde \fX]) \\
		&= 2^{3+2\ep}(\E[|R_{m,ji}|^{4+\xi}]+\E[|\hat \delta_{nm,ji}|^{4+\xi}|\tilde \fX]) = O_p(1),
	\end{align*}
	as $\E\|X\|_2^{4+\xi}<\i$ and $\E[|\hat \delta_{nm,ji}|^{4+\xi}|\tilde \fX]  =O_p(1)$. This holds because $R_{m,ji}= \{X_{m,i}-\E[X_{m,i}|X_{m,j}]\}$ and both terms are in $\cL^{4+\xi}(P)$ if $X_{m,i}\in\cL^{4+\xi}(P)$ which is guaranteed as $\E\|X\|_2^{4+\xi}<\i$. For the third term in the upper bound of \Cref{eq:FellerCondUpperBoundSecondFactor}, we note that by the conditional Jensen's inequality, we have that
	\begin{align*}
		\E[|\delta_{n,ji}^2 |^{2+\ep}|\tilde \fX] \leq \E[ |\phi_{ji}(X_{m,i})-\hat \phi_{ji}^n(X_{m,j})|^{4+2\ep}|\tilde \fX]=\E[|\hat \delta_{nm,ji}|^{4+\xi}|\tilde \fX] = O_p(1),
	\end{align*}
	by assumption. Therefore, we have that
	\begin{align*}
		\sum_{m=1}^n 	\E\lf \|\Psi_n(X_{m},\tilde \fX)\|^{2+\ep}_2 \big| \tilde \fX \rf \leq \frac{n}{n^{\frac{2+\ep}{2}}} O_p(1) = n^{-\ep/2} O_p(1) \convp_n 0,
	\end{align*}
	proving the conditional Lindeberg-Feller condition. By \Cref{lm:conditionaltriangularCLT} it holds that
	\begin{align*} %
		\frac{1}{\sqrt{n}}\sum_{m=1}^{n} \psi_{n}(X_{m},\tilde \fX) \convd_n \cN(0,\Sigma).
	\end{align*}
	Now it only remains to prove that
	\begin{align*}
		\| \widehat \Sigma_{n} -\Sigma_n \| \convp 0,
	\end{align*}
	or, equivalently, that each entry converges to zero in probability. For example, for the entries of the first block matrix with $j\not =i$ and $l\not =r$ we prove that
	\begin{align*}
		|\widehat \Sigma_{n,R,ji,lr}-\Sigma_{n,R,ji,lr} | \convp 0.
	\end{align*}
	Now note that the observable estimated covariance matrix entry is given by
	\begin{align*}
		\widehat \Sigma_{n,R,ji,lr}
		&= \frac{1}{n}\sum_{m=1}^{n} \hat R_{nm,ji}^2 \hat R_{nm,lr}^2 -  \hat \mu_{n,ji} \hat \mu_{n,lr} ,
	\end{align*}
	while the unobservable conditional covariance matrix is given by
	\begin{align*}
		\Sigma_{n,R,ji,lr} &= \E[(\hat R_{nm,ji}^2 -  \mu_{ji}- \delta_{n,ji}^2)(\hat R_{nm,lr}^2 -  \mu_{lr}- \delta_{n,lr}^2)|\tilde \fX]\\
		&= \E[\hat R_{nm,ji}^2 \hat R_{nm,lr}^2| \tilde \fX] - (\mu_{ji}+\delta_{n,ji}^2)(\mu_{lr}+\delta_{n,lr}^2) \\
		&= \E[\hat R_{nm,ji}^2 \hat R_{nm,lr}^2| \tilde \fX] - 
		\E[\hat R_{nm,ji}^2|\tilde \fX] \E[\hat R_{nm,lr}^2 | \tilde \fX],
	\end{align*}
	where we have used that $ \E[\hat R_{nm,ji}^2|\tilde \fX] = \mu_{ji}+\delta_{n,ji}^2$; see \Cref{eq:asympnormalityexpansion} and its discussion.	Note that the second term of the covariance matrix estimator expands to
	\begin{align*}
		\hat \mu_{n,ji} \hat \mu_{n,lr}  &= \lp \frac{1}{n}\sum_{m=1}^n \hat R_{nm,ji}^2 \rp\lp \frac{1}{n}\sum_{m=1}^n\hat R_{nm,lr}^2 \rp	\\
		&= \lp \frac{1}{n}\sum_{m=1}^n \hat R_{nm,ji}^2 -\E[\hat R_{nm,ji}^2]\rp\lp \frac{1}{n}\sum_{m=1}^n\hat R_{nm,lr}^2 -\E[\hat R_{nm,lr}^2] \rp	\\
		&\quad - 
		\E[\hat R_{nm,ji}^2] \E[\hat R_{nm,lr}^2] \\
		&\quad +  \frac{1}{n}\sum_{m=1}^n \hat R_{nm,ji}^2 \E[\hat R_{nm,lr}^2]  \\
		&\quad + \frac{1}{n}\sum_{m=1}^n\hat R_{nm,lr}^2 \E[\hat R_{nm,ji}^2]  ,
	\end{align*}
	Thus
	\begin{align}
		&\, |\widehat \Sigma_{n,R,ji,lr}-\Sigma_{n,R,ji,lr} | \notag \\
		=&\, \bigg|\frac{1}{n}\sum_{m=1}^{n} (\hat R_{nm,ji}^2 \hat R_{nm,lr}^2 -\E[\hat R_{nm,ji}^2 \hat R_{nm,lr}^2|\tilde \fX]) \notag \\
		&\quad - \lp \frac{1}{n}\sum_{m=1}^n \hat R_{nm,ji}^2 -\E[\hat R_{nm,ji}^2|\tilde \fX]\rp\lp \frac{1}{n}\sum_{m=1}^n\hat R_{nm,lr}^2 -\E[\hat R_{nm,lr}^2|\tilde \fX] \rp \notag \\
		&\quad - \frac{1}{n}\sum_{m=1}^n (\hat R_{nm,ji}^2 \E[\hat R_{nm,lr}^2|\tilde \fX]- 	\E[\hat R_{nm,ji}^2|\tilde \fX] \E[\hat R_{nm,lr}^2|\tilde \fX]) \notag \\
		&\quad - \frac{1}{n}\sum_{m=1}^n ( \hat R_{nm,lr}^2 \E[\hat R_{nm,ji}^2|\tilde \fX]-	\E[\hat R_{nm,ji}^2|\tilde \fX] \E[\hat R_{nm,lr}^2| \tilde \fX]) \bigg|. \label{eq:fourtermsexpansion}
	\end{align}
	Each of these terms tends to zero in probability by \Cref{lm:uniformwllntotriangular}. For example, for the first term of \Cref{eq:fourtermsexpansion} it suffices to show that
	\begin{align*}
		\E	\left[|\hat R_{nm,ji}^2 \hat R_{nm,lr}^2|^{1+\ep}|\tilde \fX\right] = O_p(1),
	\end{align*}
	for some $\ep>0$. Fix $\ep = \xi/4$ and note, by the cr-inequality, that
	\begin{align*}
		\hat R_{nm,ji}^2 \hat R_{nm,lr}^2 &= (X_{m,i}-\hat \phi_{ji}^n(X_{m,j}))^2 (X_{m,r}-\hat \phi_{lr}^n(X_{m,l}))^2 \\
		&\leq 4 (R_{m,ji}^2 + \hat \delta_{nm,ji}^2) (R_{m,lr}^2 + \hat\delta_{nm,lr}^2).
	\end{align*}
	Thus, by the cr-inequality and the conditional Cauchy-Schwarz inequality we have, with $c=4^{1+\ep} 2^{2\ep}$, that
	\begin{align*}
		&c^{-1}\E[|\hat R_{nm,ji}^2 \hat R_{nm,lr}^2|^{1+\ep}|\tilde \fX] \\
		\leq &\, c^{-1}4^{1+\ep} \E[|R_{m,ji}^2 + \hat \delta_{nm,ji}^2|^{1+\ep} |R_{m,lr}^2 + \hat\delta_{nm,lr}^2|^{1+\ep}|\tilde \fX]\\
		\leq &\,\E[(|R_{m,ji}|^{2+2\ep} + |\hat \delta_{nm,ji}|^{2+2\ep}) (|R_{m,lr}|^{2+2\ep} + |\hat\delta_{nm,lr}|^{2+2\ep})|\tilde \fX] \\
		\leq&\, \E[|R_{m,ji}|^{2+2\ep}|R_{m,lr}|^{2+2\ep}|\tilde \fX] + \E[|R_{m,ji}|^{2+2\ep}|\hat\delta_{nm,lr}|^{2+2\ep}|\tilde \fX]\\
		&\quad + \E[|\hat \delta_{nm,ji}|^{2+2\ep}|R_{m,lr}|^{2+2\ep}|\tilde \fX] + \E[|\hat \delta_{nm,ji}|^{2+2\ep}|\hat\delta_{nm,lr}|^{2+2\ep}|\tilde \fX]\\
		\leq &\, \E[|R_{m,ji}|^{4+\xi}]\E[|R_{m,lr}|^{4+\xi}] + \E[|R_{m,ji}|^{4+\xi}]\E[|\hat\delta_{nm,lr}|^{4+\xi}|\tilde \fX]\\
		&\quad + \E[|\hat \delta_{nm,ji}|^{4+\xi}|\tilde \fX]\E[|R_{m,lr}|^{4+\xi}] + \E[|\hat \delta_{nm,ji}|^{4+\xi}|\tilde \fX]\E[|\hat\delta_{nm,lr}|^{4+\xi}|\tilde \fX] \\
		=& \,O_p(1),
	\end{align*}
	as $ \E[|\hat \delta_{nm,ji}|^{4+\xi}|\tilde \fX] = O_p(1)$ for all $j\not =i$ by assumption  and $\E[|R_{m,ji}|^{4+\xi}]<\i$ since $\E\|X\|_2^{4+\xi}<\i$.
	
	Similar arguments show convergence in probability of the entries in the other block submatrices of $\widehat \Sigma_n$ less $\Sigma_n$, yielding the desired conclusion.
	
\end{proofenv}

\begin{proofenv}{\Cref{thm:Confidence}}
	We prove the theorem under the simplifying assumption that $\E[X]=0$ for which we can simplify the variance estimator by $\hat V_{m,i}:= X_{m,i}^2$ and $\hat \nu_{n,i} := \frac{1}{n}\sum_{m=1}^n \hat V_{m,i}$ for all $1\leq i \leq p$.
	
	First, note (using the notation introduced in \Cref{thm:asymptoticnormalityedgecomponents}) that $\hat M_1 = \{\hat R_{n1,ji}^2\}_{j\not = i }$, $\hat \mu = \hat \mu_n$, $\hat \nu = \hat \nu_n$ and $\widehat \Sigma = \widehat \Sigma_n$. The conditional mean of $\hat M_1 $ given $\tilde \fX_n$ is given by
	\begin{align*}
		\E[\hat M_1| \tilde \fX_n] = \E[\{\hat R_{n1,ji}^2\}_{j\not = i }|\tilde \fX_n]=  \mu+  \delta_n^2,
	\end{align*}
	see \Cref{eq:asympnormalityexpansion}. Similarly we have that $\E[\hat V_1 | \tilde \fX_n] =\E[\hat V_1] = \nu$. Subtracting a constant (conditional on $\tilde \fX_n$) does not change the conditional variance, hence
	\begin{align*}
		\Var\lp  \begin{bmatrix}
			\hat R_{n1}^2 - \delta_{n}^2 - \mu  \\
			\hat V_{1}  - \nu 
		\end{bmatrix} \bigg| \tilde \fX_n \rp = 	\Var\lp  (\hat M_1^\t, \hat V_1^\t)^\t \bigg| \tilde \fX_n \rp  \convp_n \Sigma.
	\end{align*}
	$\Sigma$ is  constant and positive semi-definite with strictly positive diagonal. As such, the conditions of \Cref{thm:asymptoticnormalityedgecomponents} is satisfied,  which yields that
	\begin{align} \label{eq:ConvergenceInDistFromLem22}
		\frac{1}{\sqrt{n}}\sum_{m=1}^n  \begin{bmatrix}
			\hat R_{nm}^2 - \delta_{n}^2 - \mu  \\
			\hat V_{m}  - \nu 
		\end{bmatrix} = \sqrt{n}\begin{bmatrix}
			\hat \mu - \delta_n^2 - \mu \\
			\hat \nu - \nu
		\end{bmatrix} \convd_n \cN(0,\Sigma),
	\end{align}
	and that
	\begin{align*}
		\widehat \Sigma &= \begin{bmatrix}
			\widehat \Sigma_{M} & \widehat \Sigma_{MV} \\ \widehat \Sigma_{MV}^\t & \widehat \Sigma_{V}  
		\end{bmatrix} \convp \Sigma =: \begin{bmatrix}
			\Sigma_{M} &  \Sigma_{MV} \\  \Sigma_{MV}^\t &  \Sigma_{V}  
		\end{bmatrix}  \in \R^{p^2\times p^2}.
	\end{align*}
	For any $j\not = i$ we denote
	\begin{align*}
		\hat w_{ji} = \frac{1}{2}\log\lp\frac{\hat \mu_{ji}}{\hat \nu_{i}}\rp, \quad \tilde w_{ji} = \frac{1}{2}\log \lp \frac{\hat \mu_{ji}-\delta_{n,ji}^2}{\hat \nu_i} \rp, \quad  w_{ji}= \frac{1}{2} \log \lp \frac{\mu_{ji}}{\nu_i}\rp ,
	\end{align*}
	where the latter is a shorthand notation for the Gaussian edge weight $w_{\mathrm{G}}(j\to i)$. Fix $\alpha\in(0,1)$.  First, consider $(j\to i) \in \cE$ and note that
	
	\begin{align} \label{eq:convdDifference}
		\sqrt{n}\lp\begin{bmatrix}
			\hat \mu_{ji} -  \mu_{ji} \\
			\hat \nu_{i} - \nu_i
		\end{bmatrix}- \begin{bmatrix}
			\hat \mu_{ji} - \delta_{n,ji}^2 - \mu_{ji} \\
			\hat \nu_{i} - \nu_i
		\end{bmatrix}    \rp = \sqrt{n} \begin{bmatrix}
			\delta_{n,ji}^2\\
			0
		\end{bmatrix} = \sqrt{n} \begin{bmatrix}
			\E[ \hat \delta_{nm,ji}^2 |\tilde \fX_n ]\\
			0
		\end{bmatrix} \convp_n 0,
	\end{align}
	by assumption (iv). 
	Hence, \Cref{eq:ConvergenceInDistFromLem22}, \Cref{eq:convdDifference} and the delta method yields that
	\begin{align*}
		\sqrt{n} \lp \hat w_{ji} - w_{ji} \rp &= \sqrt{n}\lp\log\lp\frac{\hat \mu_{ji}}{\hat \nu_i}\rp - \log \lp \frac{\mu_{ji}}{\nu_i}\rp \rp \\
		&= \sqrt{n}(\log(\hat \mu_{ji})-\log(\mu_{ji})- \log(\hat \nu_{i} ) + \log(\nu_i)) \\
		& \convd_n \cN(0,\sigma_{ji}^2),
	\end{align*}
	where 
	\begin{align*}
		\hat \sigma_{ji}^2 := \frac{\widehat \Sigma_{M,ji}}{\hat \mu_{ji}^2} + \frac{\widehat \Sigma_{V,i}}{\hat \nu_{i}^2} - 2 \frac{\widehat \Sigma_{MV,ji,i}}{\hat \mu_{ji} \hat \nu_{i}} \convp  \sigma_{ji}^2 := \frac{\Sigma_{M,ji}}{\mu_{ji}^2} + \frac{\Sigma_{V,i}}{\nu_i^2} - 2 \frac{\Sigma_{MV,ji,i}}{\mu_{ji} \nu_i}\geq 0.
	\end{align*}
	Here $\widehat \Sigma_{M,ji}$ and $\widehat \Sigma_{V,i}$ and their limits use a shorthand notation that denote the corresponding diagonal element, e.g., $\widehat \Sigma_{M,ji}:= \widehat \Sigma_{M,ji,ji}$.
	
	An asymptotically valid marginal confidence interval for $w_{ji}$ with level $\alpha$ is, by virtue of the above convergence in distribution, given by $$
	\hat w_{ji} \pm \hat \sigma_{ji}\frac{q(1-\frac{\alpha}{2})}{2\sqrt{n}},
	$$ 
	where $q(1-\frac{\alpha}{2})$ is the $1-\alpha/2$ quantile of the standard normal distribution. That is, $$P\lp\hat w_{ji} - \hat \sigma_{ji}\frac{q(1-\frac{\alpha}{2})}{2\sqrt{n}} \leq w_{ji}\leq \hat w_{ji} + \hat \sigma_{ji}\frac{q(1-\frac{\alpha}{2})}{2\sqrt{n}}\rp \longrightarrow_n 1-\alpha.$$
	On the other hand, for any $(j\to i)\not \in \cE$ we have, by similar arguments, except that no assumption guarantees that $\sqrt{n}\delta_{n,ji}^2$ vanishes, that
	\begin{align*}
		P\lp\tilde w_{ji} - \tilde \sigma_{ji}\frac{q(1-\frac{\alpha}{2})}{2\sqrt{n}} \leq w_{ji}\leq \tilde w_{ji} + \tilde  \sigma_{ji}\frac{q(1-\frac{\alpha}{2})}{2\sqrt{n}}\rp  \longrightarrow_n 1-\alpha,
	\end{align*}
	where 
	\begin{align*}
		\tilde  \sigma_{ji}^2 &:= \frac{\widehat \Sigma_{M,ji}}{(\hat \mu_{ji}- \delta_{n,ji}^2)^2} + \frac{\widehat \Sigma_{V,i}}{\hat \nu_{i}^2} - 2 \frac{\widehat \Sigma_{MV,ji,i}}{(\hat \mu_{ji} - \delta_{n,ji}^2)\hat \nu_{i}} \\
		&\convp  \sigma_{ji}^2 := \frac{\Sigma_{M,ji}}{\mu_{ji}^2} + \frac{\Sigma_{V,i}}{\nu_i^2} - 2 \frac{\Sigma_{MV,ji,i}}{\mu_{ji} \nu_i}\geq 0,
	\end{align*}
	by the convergence in  \Cref{eq:ConvergenceInDistFromLem22}.
	Note that $\tilde \sigma_{ji}^2$ is not observable since $\delta_{n,ji}^2$ is not observable. 
	Now define 
	\begin{align*}
		\hat u_{\alpha,ji}, \, \hat l_{\alpha,ji} &:= \hat w_{ji} \pm \hat \sigma_{ji}\frac{q\lp1-\frac{\alpha}{2p(p-1)}\rp }{2\sqrt{n}}, \\
		\tilde u_{\alpha,ji}, \,  \tilde l_{\alpha,ji} &:= \tilde w_{ji} \pm \tilde \sigma_{ji}\frac{q\lp1-\frac{\alpha}{2p(p-1)}\rp}{2\sqrt{n}}.
	\end{align*}
	Thus, we have the following Bonferroni corrected simultaneous confidence interval for the Gaussian edge weights 
	\begin{align*}
		&\liminf_{n \to \infty} P \left( \bigcap_{(j\to i)\in \cE} \lp w_{ji}\in \left[ \hat l_{\alpha,ji}, \hat u_{\alpha,ji} \right] \rp  \bigcap_{j\to i \not \in \cE} \lp w_{ji}\in \left[  \tilde l_{\alpha,ji}, \tilde u_{\alpha,ji} \right] \rp  \right) \geq 1-\alpha.
	\end{align*}
	The above confidence region has the correct asymptotic level, but it is infeasible to compute in that $\tilde w_{ji}$, $\tilde \sigma_{ji}$ and $\cE$ are not directly observable from data. 
	Furthermore, define
	\begin{align*}
		C(\hat l_{\alpha},\tilde l_{\alpha}, \hat u_{\alpha}, \tilde u_{\alpha}) := \bigg\{ \argmin_{\tilde \cG=(V,\tilde \cE)\in \cT_p} \sum_{(j \to i )\in \tilde \cE}w_{ji}':  &\forall (j\to i)\in \cE, w_{ji}'\in[\hat l_{\alpha,ji},\hat u_{\alpha,ji}],  \\
		& \forall (j\to i)\not \in \cE,w_{ji}'\in[\tilde  l_{\alpha,ji},\tilde  u_{\alpha,ji}]   \bigg\},
	\end{align*}
	and note that this is an unobservable confidence region for the causal graph. 	
	That is,
	\begin{align*}
		&\, \liminf_{n \to \infty}P(\cG \in C(\hat l_{\alpha},\tilde l_{\alpha}, \hat u_{\alpha}, \tilde u_{\alpha})) \\
		\geq & \,  \liminf_{n \to \infty} P\left(\bigcap_{(j\to i)\in \cE}(w_{ji}\in [\hat l_{\alpha,ji},\hat u_{\alpha,ji}]) \bigcap_{(j\to i)\not \in \cE}(w_{ji}\in [\tilde l_{\alpha,ji},\tilde  u_{\alpha,ji}]) \right) \\
		\geq & \,  1-\alpha.
	\end{align*}
	Our proposed confidence region has the form
	\begin{align*}
		\hat C:=	C(\hat l_{\alpha}, \hat u_{\alpha}) := \bigg\{ \argmin_{\tilde \cG=(V,\tilde \cE)\in \cT_p} \sum_{(j \to i )\in \tilde \cE}w_{ji}':  &\forall j\not = i , w_{ji}'\in[\hat l_{\alpha,ji},\hat u_{\alpha,ji}]\bigg\},
	\end{align*}
	which corresponds to the biased but computable confidence region
	\begin{align*}
		\prod_{j\not = i} [\hat l_{\alpha,ji}, \hat u_{\alpha,ji}]&= \prod_{j\not = i} \left[  \hat w_{ji} \pm \hat \sigma_{ji}\frac{q\lp1-\frac{\alpha}{2p(p-1)}\rp }{2\sqrt{n}} \right].
	\end{align*}
	for the Gaussian edge weights, where the product is over all combinations of possible edges $1\leq j \not = i \leq p$. The biased confidence region $ \prod_{j\not = i} [\hat l_{\alpha,ji}, \hat u_{\alpha,ji}]$ does not necessarily contain the population Gaussian edge weights  with a probability of at least $1-\alpha$ in the large sample limit. However, it can be used to construct a conservative confidence region for the causal graph.
	To see this, note that by further penalizing the wrong (non-causal) edge weights, the causal graph still yields the minimum edge weight spanning directed tree. Hence,
	\begin{align*}
		&\liminf_{n \to \infty}	P(\cG\in  C(\hat l_{\alpha}, \hat u_{\alpha})) \\
		\geq &\liminf_{n \to \infty} P\left(\bigcap_{(j\to i)\in \cE}(w_{ji}\in [\hat l_{\alpha,ji},\hat u_{\alpha,ji}]) \bigcap_{(j\to i)\not \in \cE}(w_{ji}\in [\tilde l_{\alpha,ji},\tilde  u_{\alpha,ji}]) \bigcap_{(j\to i) \not \in \cE}(\tilde u_{\alpha,ji}\leq \hat u_{\alpha,ji}) \right)\\
		\geq  &1-\alpha,
	\end{align*}
	as $P\lp \tilde u_{\alpha,ji}\leq \hat u_{\alpha,ji}\rp \to_n 1$ for all $(j\to i) \not \in \cE$ by \Cref{lm:InequalityWithHighProb} below.
\end{proofenv}

\begin{lemma} \label{lm:InequalityWithHighProb}
	Suppose that the assumptions of \Cref{thm:asymptoticnormalityedgecomponents} hold. It holds that
	\begin{align*}
		\forall (j\to i ) \not \in \cE ,\forall\alpha\in(0,1): 	P\lp \tilde u_{\alpha,ji}\leq \hat u_{\alpha,ji}\rp \to_n 1.
	\end{align*}
\end{lemma}
\begin{proofenv}{\Cref{lm:InequalityWithHighProb}}
	Fix any $(j\to i ) \not \in \cE$ and $\alpha\in(0,1)$ and note that we want to show that
	\begin{align*}
		&\tilde u_{\alpha,ji}\leq \hat u_{\alpha, ji} \\
		\iff & \tilde w_{ji} +c \frac{\tilde \sigma_{ji}}{\sqrt{n}}  \leq \hat  w_{ji} + c\frac{\hat  \sigma_{ji}}{\sqrt{n}} \\
		\iff &0 \leq  \log \lp \hat \mu_{ji} \rp  + c\frac{\hat  \sigma_{ji}}{\sqrt{n}} - \log \lp \hat \mu_{ji}-\delta_{n,ji}^2 \rp -c \frac{\tilde \sigma_{ji}}{\sqrt{n}}
	\end{align*}
	holds with probability converging to one, where $c$ is a strictly positive constant. It suffices to show that an even smaller quantity is non-negative with  probability converging to one. That is, it suffices to show that
	\begin{align*}
		0 \leq  \log \lp \hat \mu_{ji} \rp  + c\frac{\hat  \sigma_{ji}}{\sqrt{n}} - \log \lp \hat \mu_{ji}-\delta_{n,ji}^2 \rp -c \frac{\tilde \sigma_{ji}^*}{\sqrt{n}}  ,
	\end{align*}
	with increasing probability, where
	\begin{align*}
		\tilde \sigma_{ji}^* := \sqrt{\frac{\widehat \Sigma_{M,ji}}{(\hat \mu_{ji}- \delta_{n,ji}^2)^2} + \frac{\widehat \Sigma_{V,i}}{\hat \nu_{i}^2} + 2 \frac{|\widehat \Sigma_{MV,ji,i}|}{(\hat \mu_{ji} - \delta_{n,ji}^2)\hat \nu_{i}}} \geq \tilde \sigma_{ji},
	\end{align*}
	with $P(\tilde \sigma_{ji}^*>0)\to_n 1$.
	Let $d_n(t):[0,\i)\to \R$ denote the random function given by
	\begin{align*}
		d_n(t) :=&\log \lp \hat \mu_{ji} \rp  + c\frac{\hat  \sigma_{ji}}{\sqrt{n}} - \log \lp \hat \mu_{ji}-t \rp \\
		&\quad -\frac{c}{\sqrt{n}} \sqrt{ \frac{\widehat \Sigma_{M,ji}}{(\hat \mu_{ji}- t)^2} + \frac{\widehat \Sigma_{V,i}}{\hat \nu_{i}^2} + 2 \frac{|\widehat \Sigma_{MV,ji,i}|}{(\hat \mu_{ji} - t)\hat \nu_{i}}}.
	\end{align*}
	It holds that $d_n(0)=0$ surely, so by the mean value theorem, the desired conclusion holds if it with probability one (as $n$ tends to infinity) holds, for all $t\in[0,\delta_{n,ji}^2]$,  that $d_n'(t)\geq 0$ . 
	
	Now fix $\eta>0$ and choose $M_\eta,\ep_1,\ldots,\ep_5>0$ such that the constant lower bounds in the following inequalities are strictly positive
	\begin{align*}
		\Omega_n(1) :&= (\hat \mu_{ji}\leq M_\eta),\\
		\Omega_n(2) :&= (\Sigma_{M,ji}- \ep_1 \leq \widehat \Sigma_{M,ji}\leq \Sigma_{M,ji} +\ep_1),\\
		\Omega_n(3) :&= (\Sigma_{V,i}- \ep_2 \leq \widehat \Sigma_{V,i}\leq \Sigma_{V,i} +\ep_2),\\
		\Omega_n(4) :&= ( 0 \leq |\widehat \Sigma_{MV,ji,i}|\leq | \Sigma_{MV,ji,i}|+\ep_3) ,\\
		\Omega_n(5) :&= (\mu_{ji} - \ep_4\leq \hat \mu_{ji}-\delta_{n,ji}^2 \leq \mu_{ji}+ \ep_4 ),\\
		\Omega_n(6) :&= (\nu_i -\ep_5 \leq \hat \nu_{i} \leq \nu_i +\ep_5),
	\end{align*}
	and $\liminf_{n\to \i}P(\Omega_n(1))> 1- \eta$. 
	This is possible as $\hat \mu_{ji}- \delta_{n,ji}^2 \convp_n \mu_{ji}>0$ and 
	\begin{align*}
		\delta_{n,ji}^{2} =	E[|\hat \delta_{nm,ji}|^{2}|\tilde \fX] &= E[|\hat \delta_{nm,ji}|^{\frac{4+\xi}{2+\xi/2}}|\tilde \fX] \\
		&\leq E[|\hat \delta_{nm,ji}|^{4+\xi}|\tilde \fX]^{\frac{1}{2+\xi/2}} = O_p(1),
	\end{align*}
	by the conditional Jensen's inequality and concavity of $[0,\i)\ni x\mapsto x^{\frac{1}{2+\xi/2}}$, which implies that $
	\hat \mu_{ji} = 	(\hat \mu_{ji}- \hat \delta_{n,ji}^{2}-\mu_{ji}) +(\hat \delta_{n,ji}^{2}+\mu_{ji}) = o_p(1) + O_p(1) = O_p(1)$. Furthermore, as
	\begin{align*}
		&\widehat \Sigma_{M,ji} \convp_n \Sigma_{M,ji}>0, \quad \widehat \Sigma_{V,i} \convp_n \Sigma_{V,i}>0, \\
		&|\widehat \Sigma_{MV,ji,i}| \convp_n | \Sigma_{MV,ji,i}|\geq 0 , \quad  \hat \nu_{i} \convp_n \nu_i> 0,
	\end{align*}
	it holds that
	\begin{align*}
		\limsup_{n\to \i }P \lp \bigcup_{1\leq k \leq 6 } \Omega_n(k)^c \rp &\leq \sum_{1\leq k \leq 6} \limsup_{n\to \i } P(\Omega_n(k)^c) \\
		&= \limsup_{n\to \i }P(\Omega_n(1)^c) \leq \eta.
	\end{align*}
	Here we used that the diagonal elements of the limit covariance matrix are assumed strictly positive. That $\mu_{ji},\nu_{i}>0$ follows from the fact that $X_i-\E[X_i|X_j]$ is assumed to have a density (w.r.t. Lebesgue measure) and that the variables are non-degenerate $\nu_i = \text{Var}(X_i)>0$. 
	Thus, we have that
	$$	
	\liminf_{n\to \i} P\lp\bigcap_{1\leq k \leq 6}\Omega_n(k) \rp > 1-\eta.
	$$
	Now consider a fixed $\omega \in \bigcap_{1\leq k \leq 6}\Omega_n(k)$ and note that with $g_n:[0, \delta_{n,ji}^{2}] \to \R$ given by $g_n(t) = \hat \mu_{ji} - t$ we have that  $g_n$ is decreasing and that
	\begin{align*}
		g_n(	[0, \delta_{n,ji}^2]) \subset [\mu_{ji}- \ep_4, \hat \mu_{ji}] \subset (0,M_\eta]
	\end{align*}
	We have for any $t\in[0,\delta_{n,ji}^2]$ that
	\begin{align*}
		d_n'(t) &= \frac{1}{ \hat \mu_{ji}-t} - \frac{c}{\sqrt n} \lp  \frac{\widehat \Sigma_{M,ji}}{(\hat \mu_{ji}- t)^2} + \frac{\widehat \Sigma_{V,i}}{\hat \nu_i^2} +  \frac{2|\widehat \Sigma_{MV,ji,i}|}{(\hat \mu_{ji} - t)\hat \nu_i}\rp^{-1/2} \\
		&\quad \times \lp  \frac{\widehat \Sigma_{M,ji}}{(\hat \mu_{ji}- t)^3}  +  \frac{|\widehat \Sigma_{MV,ji,i}|}{(\hat \mu_{ji} - t)^2\hat \nu_i}\rp,
	\end{align*}
	hence,
	\begin{align*}
		d_n'(t) &= \frac{1}{ \hat\mu_{ji}-t} - \frac{c}{\sqrt n} \lp  \frac{\widehat \Sigma_{M,ji}}{g_n(t)^2} + \frac{\widehat \Sigma_{V,i}}{\hat \nu_{i}^2} +  \frac{2|\widehat \Sigma_{MV,ji,i}|}{g_n(t)\hat \nu_i}\rp^{-1/2} \\
		&\quad \times  \lp  \frac{\widehat \Sigma_{M,ji}}{g_n(t)^3}  +  \frac{|\widehat \Sigma_{MV,ji,i}|}{g_n(t)^2\hat \nu_{i}}\rp \\
		& \geq \frac{1}{ \hat \mu_{ji}} - \frac{c}{\sqrt n} \lp  \frac{\widehat \Sigma_{M,ji}}{\hat \mu_{ji}^2} + \frac{\widehat \Sigma_{V,i}}{\hat \nu_{i}^2} \rp^{-1/2} \\
		&\quad \times \lp  \frac{\widehat \Sigma_{M,ji}}{(\hat \mu_{ji}-\delta_{n,ji}^2)^3}  +  \frac{|\widehat \Sigma_{MV,ji,i}|}{(\hat \mu_{ji}-\delta_{n,ji}^2)^2\hat \nu_{i}}\rp \\
		&\geq \frac{1}{M_\eta} - \frac{c}{\sqrt{n}} \lp \frac{\Sigma_{M,ji}-\ep_1}{M_\eta^2} + \frac{\Sigma_{V,i}-\ep_2}{(\nu_i+\ep_5)^2}  \rp^{-1/2}\\
		&\quad \times  \lp  \frac{\Sigma_{M,ji}+\ep_1}{(\mu_{ji}-\ep_4)^3}  +  \frac{| \Sigma_{MV,ji,i}|+\ep_3}{(\mu_{ji}-\ep_4)^2(\nu_i- \ep_5)}\rp \\
		&=: \frac{1}{M_\eta} - \frac{C_{M_\eta,\ep_1,\ep_2,\ep_3,\ep_4,\ep_5}}{\sqrt{n}}\\
		&\geq 0,
	\end{align*}
	for $n \geq (C_{M_\eta,\ep_1,\ep_2,\ep_3,\ep_4,\ep_5}M_\eta)^2$. We conclude that for $n \geq (C_{M_\eta,\ep_1,\ep_2,\ep_3,\ep_4,\ep_5}M_\eta)^2$
	\begin{align*}
		P \lp  \tilde u_{\alpha,ji}\leq \hat u_{\alpha,ji} \rp &= P \lp 0 \leq  \log \lp \hat \mu_{ji} \rp  + c\frac{\hat  \sigma_{ji}}{\sqrt{n}} - \log \lp \hat \mu_{ji}-\delta_{n,ji}^2 \rp -c \frac{\tilde \sigma_{ji}}{\sqrt{n}}  \rp  \\
		&\geq P\lp \forall t\in[0, \delta_{n,ji}^2]: d_n'(t) \geq 0 \rp\\
		& \geq P \lp \bigcap_{1\leq k \leq 6} \Omega_n(k)\rp.
	\end{align*}
	Hence,
	\begin{align*}
		\liminf_{n\to \i} P \lp  \tilde u_{\alpha,ji}\leq \hat u_{\alpha,ji} \rp  \geq \liminf_{n\to \i} P \lp \bigcap_{1\leq k \leq 6} \Omega_n(k)\rp \geq 1-\eta,
	\end{align*}
	and as $\eta>0$ was chosen arbitrarily, we have the desired conclusion
	\begin{align*}
		P \lp  \tilde u_{\alpha,ji}\leq \hat u_{\alpha,ji} \rp \to 1.
	\end{align*}
\end{proofenv}

\begin{proofenv}{\Cref{thm:testlevel}}
	Consider a collection of arbitrary  and possibly data-dependent substructures $\cR_1,\cR_2,...$ and level $\alpha\in (0,1)$. First, we note that the score associated with two sets of edge weights  $w_1$ and $w_2$ is weakly monotone, that is,  $s(w_1)\leq s(w_2)$ if
	$w_1$ and $w_2$ satisfy the component-wise partial ordering $w_1\leq w_2$.   Furthermore, the restricted score function $w\mapsto s_{\cT(\cR)}(w)$ is also weakly monotone for any set of restrictions $\cR$.
	
	Let $k\in \N$ and suppose that the null hypothesis
	\begin{align*}
		\cH_0(\cR_k) : \cE_{\cR_k} \setminus \cE = \emptyset, \;  \cE\setminus \cE^\text{miss}_{\cR_k}  = \emptyset,\; r_k=\root{\cG},
	\end{align*}
	corresponding to the restriction $\cR_k=(\cE_{\cR_k},\cE_{\cR_k}^{\text{miss}},r_k)$ 	is true. 	
	
	If there is a graph in $\hat C:= \hat C(\hat l_\alpha, \hat u_\alpha)$ satisfying the restrictions imposed by the substructure $\cR_k$, then there exist $\hat l_\alpha \leq w' \leq \hat u_\alpha$ such that $s(w')$ attains its minimum value in a graph satisfying $\cR_k$. Penalizing (or removing) edges that are not present in the minimum edge weight directed tree does not affect the score of the minimum edge weigh directed tree. Hence, it holds that
	$$
	s_{\cT(\cR_k)}(w') = s(w').
	$$ 
	Monotonicity of $s_{\cT(\cR_k)}$ and $s$ in the edge weights  imply that $$
	s_{\cT(\cR_k)}(\hat l_\alpha) \leq s_{\cT(\cR_k)}(w') = s(w') \leq s(\hat u_\alpha).$$ 
	Hence, $s_{\cT(\cR_k)}(\hat l_\alpha) > s(\hat u_\alpha)$ entails that  no graph in $\hat C$  satisfies the restrictions of $\cR_k$. (This is a slightly conservative criterion as  $s_{\cT(\cR_k)}(\hat l_\alpha) \leq  s(\hat u_\alpha)$ does not necessarily guarantee that a graph in $\hat C$ satisfies the restrictions of $\cR_k$.) 
	
	Therefore, if $\psi_{\cR_k}=1$, then we know that there is no graph in $\hat C$  satisfying the restrictions of $\cR_k$. As the causal graph $\cG$ satisfies the restriction $\cR_k$ we conclude that $\cG$ is not contained in $\hat C$. Thus for any true $\cR_k$ we have that
	\begin{align*}
		(\psi_{\cR_k}=1) \subseteq (\cG \not \in \hat C).
	\end{align*}
	Since this holds for any true $\cR_k$, the conclusion follows by noting that
	\begin{align*}
		\limsup_{n\to \i} P\left(\bigcup_{k : \mathcal{H}_0(\cR_k) \text{ is true}} \{\psi_{\cR_k}=1\}\right) \leq 	\limsup_{n\to \i}  P(\cG \not \in \hat C) \leq  \alpha,
	\end{align*}
	where we used \Cref{thm:Confidence}.
	
\end{proofenv}

\subsection{Proofs of Section~\ref{sec:ScoreGap}}

\subsubsection{Proofs of first results in Section~\ref{sec:ScoreGap}}

\begin{proofenv}{\Cref{lm:ScoreOrderings}}
	As conditioning reduces entropy we always have that
	\begin{align*}
		\lCE(\tilde \cG,i) = h(X_i|X_{\PAg{\tilde \cG}{i}}) &= h(X_i - \E[X_i| X_{\PAg{\tilde \cG}{i}}]|X_{\PAg{\tilde \cG}{i}}) \\
		&\leq h(X_i - \E[X_i| X_{\PAg{\tilde \cG}{i}}]) \\
		&= \lE(\tilde \cG,i).
	\end{align*}
	Furthermore, note that when conditioning we `throw out' dependence information captured by the mutual information $I(X_i-\E[X_i|X_{\PAg{ \tilde \cG}{i}}]; X_{\PAg{  \tilde\cG}{i}})$, which is zero if and only if $X_i-\E[X_i|X_{\PAg{ \tilde \cG}{i}}] \independent X_{\PAg{  \tilde\cG}{i}}$. This is especially the case for the true graph, i.e., $X_i-\E[X_i|X_{\PAg{\cG}{i}}] \independent X_{\PAg{\cG}{i}}$, implying that $\lCE(\cG,i)=\lE(\cG,i)$. Consequently, we have that the local conditional entropy score gap lower bounds the local entropy score gap,
	\begin{align*}
		\lCE(\tilde \cG,i)-\lCE( \cG,i) \leq \lE(\tilde \cG,i)-\lE(\cG,i).
	\end{align*}
	Furthermore, from the arguments in the proof of \Cref{lm:EntropyScore} we have that
	\begin{align*}
		\lE(\tilde \cG,i) &= \inf_{\tilde N_i \sim  P_{\tilde N_i}\in \cP} h\lp X_i-  \E\lf X_i|X_{\PAg{ \tilde \cG}{i}} \rf, \tilde N_i \rp \\
		&\leq \inf_{\tilde N_i \sim  P_{\tilde N_i}\in \cP_G} h\lp X_i-  \E\lf X_i|X_{\PAg{ \tilde \cG}{i}} \rf, \tilde N_i \rp \\
		&= \lG(\tilde \cG,i) + \log(\sqrt{2\pi e}).
	\end{align*}
	If $X$ is generated by a Gaussian noise model, i.e., with generating SCM $\theta=(\cG,(f_i),P_N)$ with $P_N\in \cP_{\mathrm{G}}^p$, then $
	\lE(\cG,i) = h(X_i-\E[X_i|X_{\PAg{\cG}{i}}])=h(N_i) = \log(\sqrt{2\pi e} \sigma_i) = \log(\sqrt{2\pi e}) + \frac{1}{2}\log( \E[N_i^2])= \log(\sqrt{2\pi e}) +l_\mathrm{G}(\cG,i)$, in which case the local entropy score gap lower bounds the local Gaussian score gap
	\begin{align*}
		\lE(\tilde \cG,i)-\lE(\cG,i)  \leq \lG(\tilde \cG,i) -\lG(\cG,i).
	\end{align*}	
	
\end{proofenv}

\begin{proofenv}{\Cref{lm:EdgeReversal}}
	Note that $\E[Y|X]=\E[f(X)+N_Y|X]=f(X)+\E[N_Y|X] =f(X)+\E[N_Y]$, since $N_Y\independent N_X=X$. Hence, the score difference can be written as
	\begin{align*}
		\lE(\tilde \cG) - \lE(\cG) =&\, \lE(\tilde \cG,X) - \lE(\cG,X) + \lE(\tilde \cG,Y)-\lE(\cG,Y) \\
		=& \,   h(X-\E(X|Y)) -h(X) +h(Y)- h(Y-E(Y|X))  \\
		=& \,   h(X-\E(X|Y)) -h(X) +h(Y)- h(N_Y+\E[N_Y])  \\
		=& \,   h(X-\E(X|Y)) -h(X) +h(Y)- h(N_Y), 
	\end{align*}
	as the differential entropy is translation invariant.
	Now note that as $N_Y \independent N_X$ it holds that $N_Y\independent f(X)$, so conditioning on $f(X)$ yields that
	\begin{align*}
		h(Y) &= h(Y | f(X)) + I(Y;f(X)) \\
		&= h(f(X)+N_Y | f(X)) + I(Y;f(X)) \\
		&=h(N_Y) + I(Y;f(X)).
	\end{align*}
	Similarly, conditioning on $X$ yields that
	\begin{align*}
		h(Y) &= h(Y | X) + I(Y;X) \\
		&=h(N_Y) + I(Y;X),
	\end{align*}
	which proves that
	\begin{align*}
		I(Y;f(X))= I(Y;X).
	\end{align*}
	This equality is normally derived by restricting $f$ to be bijective, but here it holds regardless by the structural assignment form, as $Y$ only depends on $X$ through  $f(X)$. Furthermore, we have that
	\begin{align*}
		h(X-\E[X|Y]) &= I(X-\E[X|Y];Y) + h(X-\E[X|Y]|Y) \\
		&= I(X-\E[X|Y];Y) + h(X|Y).
	\end{align*}
	Hence,
	\begin{align*}
		h(X-\E[X|Y])- h(X) &= I(X-\E[X|Y];Y)+h(X|Y) - h(X) \\
		&= I(X-\E[X|Y];Y)- I(Y;X).
	\end{align*}
	Thus
	\begin{align*}
		\lE(\tilde \cG) - \lE(\cG) &= h(X-\E[X|Y]) - h(X) +h(Y)-h(N_Y)  \\
		&= I(X-\E[X|Y];Y)- I(Y;X) + h(N_Y) + I(Y;f(X)) -h(N_Y)\\
		&= I(X-\E[X|Y];Y)- I(Y;X)  + I(Y;f(X)) \\
		&= I(X-\E[X|Y];Y),
	\end{align*}
	proving the claim.
\end{proofenv}

\begin{proofenv}{\Cref{lm:EdgeReversalSymmetric}}
	As the conditional mean $\E[X|Y]$ vanishes, we have that
	\begin{align*}
		\lE(\tilde \cG) - \lE(\cG)&=I(X-\E(X|Y);Y) \\
		&=  I(X;Y) \\
		&= I(Y;X) \\
		&= I(Y;f(X)),
	\end{align*}
	where the last equality was derived in the proof of \Cref{lm:EdgeReversal}. 
	Now let $f(X)^\mathrm{G}$ and $N_Y^\mathrm{G}$ be independent normal distributed random variables with the same mean and variance as $f(X)$ and $N_Y$. That is,  $f(X)^\mathrm{G} \sim \cN(\E[f(X)],\Var(f(X)))$, $N_Y^\mathrm{G} \sim \cN(\E[N_Y],\Var(N_Y))$ with $N_Y^\mathrm{G} \independent f(X)^\mathrm{G}$  such that $f(X)^\mathrm{G}+N_Y^\mathrm{G}\sim \cN(\E[f(X)]+\E[N_Y],\Var(f(X))+\Var(N_Y))$.
	
	\begin{itemize}
		\item[(a)] If $D_{\mathrm{KL}}(f(X)\|f(X)^\mathrm{G}) \leq D_{\mathrm{KL}}(N_Y\|N_Y^\mathrm{G})$ then  	by Lemma C.1 of \cite{silva2009unified} we have, since $X\independent N_Y$, that
		\begin{align*}
			I(Y;f(X)) = I(f(X)+N_Y;f(X)) \geq I(f(X)^\mathrm{G}+N_Y^\mathrm{G};f(X)^\mathrm{G}),
		\end{align*}
		Note, we have equality if and only if $f(X)$ and $N_Y$ are jointly Gaussian.  Furthermore,
		\begin{align*}
			I(f(X)^\mathrm{G}+N_Y^\mathrm{G};f(X)^\mathrm{G}) &= h(f(X)^\mathrm{G}+N_Y^\mathrm{G})- h(f(X)^\mathrm{G}+N_Y^\mathrm{G}|f(X)^\mathrm{G}) \\
			&=  h(f(X)^\mathrm{G}+N_Y^\mathrm{G})- h(N_Y^\mathrm{G}) \\
			&=	  \log(\sqrt{2\pi(\Var(f(X))+\Var(N_Y))}) - \log(\sqrt{2\pi\Var(N_Y)})  \\
			&= \frac{1}{2}\log\lp \frac{\Var(f(X))+\Var(N_Y)}{\Var(N_Y)}\rp  \\
			&= \frac{1}{2}\log\lp 1+ \frac{\Var(f(X))}{\Var(N_Y)}\rp.
		\end{align*}
		\item[(b)] If $f(X)+ N_Y$ is log-concave distributed, then by Theorem 3 of \cite{marsiglietti2018lower} we have that
		\begin{align*}
			h(f(X)+N_Y) \geq \frac{1}{2}\log \lp 4\Var(f(X)+N_Y)\rp = \frac{1}{2}\log \lp 4(\Var(f(X))+\Var(N_Y)\rp.
		\end{align*}
		Furthermore, it is well known that for fixed variance, the normal distribution maximizes entropy, hence
		\begin{align*}
			h(N_Y) \leq h(N_Y^\mathrm{G}) = \frac{1}{2} \log \lp 2\pi \Var(N_Y) \rp.
		\end{align*}
		Therefore, we get that 
		\begin{align*}
			I(Y;f(X)) &= I(f(X)+N_Y;f(X)) \\
			&= h(f(X)+N_Y) - h(f(X) + N_Y | f(X)) \\
			&= h(f(X)+N_Y) - h(N_Y) \\
			&\geq \frac{1}{2}\log \lp 4(\Var(f(X))+\Var(N_Y)\rp - \frac{1}{2} \log \lp 2\pi e \Var(N_Y) \rp \\
			&= \frac{1}{2}\log \lp \frac{2}{\pi e} + \frac{2}{\pi e} \frac{\Var(f(X))}{\Var(N_Y)} \rp,
		\end{align*}
		which yields a strictly positive lower bound if and only if
		\begin{align*}
			\frac{2}{\pi e} + \frac{2}{\pi e} \frac{\Var(f(X))}{\Var(N_Y)}> 1 \iff \frac{\Var(f(X))}{\Var(N_Y)} >\frac{\pi e}{2 } -1  \approx 3.27.
		\end{align*}

	\end{itemize}
\end{proofenv}

\begin{restatable}[]{lemma}{MarkovEquivTreesPathReversal}
	\label{lm:MarkovEquivTreesPathReversal}
	Two different but Markov equivalent trees $\tilde \cG$ and $ \hat \cG$ share the exact same edges except for a single reversed directed path between the two root nodes of the graphs,
	\begin{align*}
		\begin{array}{rcccccccccc}
			\hat	\cG: & c_1 & \to & c_2 &  \to & \cdots & \to & c_{r-1} & \to & c_{r}, \\
			\tilde \cG: &c_r & \to & c_{r-1}  & \to & \cdots & \to & c_{2} & \to & c_{1},
		\end{array}
	\end{align*}
	with $c_1=\mathrm{rt} ( \hat \cG)$ and $c_r = \mathrm{rt}(\tilde \cG)$.
\end{restatable}
\begin{proofenv}{\Cref{lm:MarkovEquivTreesPathReversal}} First, note that there always exists a unique directed path in $\hat \cG$ from $\mathrm{rt}(\hat\cG)$  to  $\mathrm{rt}(\tilde \cG)$
	\begin{align*}
		\hat	\cG: \mathrm{rt}(\hat \cG)=c_1  \to \cdots  \to c_{r-1}  \to  c_{r}=\mathrm{rt}(\tilde \cG).
	\end{align*}
	Since $\tilde \cG$ and $\hat\cG$ are Markov equivalent, they share the same skeleton, so in $\tilde \cG$ the above path must be reversed. That is, there exists a unique directed path in $\tilde \cG$ from $\mathrm{rt}(\tilde \cG)$ to $\mathrm{rt}(\hat\cG)$ given by
	\begin{align*}
		\tilde \cG: 	\mathrm{rt}(\tilde \cG) = c_r \to c_{r-1} \to \cdots \to c_1=\mathrm{rt}(\hat\cG),
	\end{align*}
	If $r=p$ we are done, so assume $r<p$. As $\hat \cG$ is a directed tree there must exists a node $z_2$ which is not a part of the above path but is a child of a node in the path. That is, there exists a node $z_1\in\{c_1,\ldots,c_r\}$ such that $\hat\cG$ contains the edge
	\begin{align*}
		\hat\cG: z_1 \to z_2.
	\end{align*}
	Furthermore, by equality of skeleton, this edge must also be present in $\tilde \cG$,
	\begin{align*}
		\tilde 	\cG: z_1 - z_2.
	\end{align*}
	Assume for contradiction that $z_2 \to z_1$ in $\tilde \cG$. As such, it must hold that $z_1=c_r=\root{\tilde \cG}$ for otherwise if $z_1\in\{c_1,\ldots,c_{r-1}\}$ then $z_1$ would have two parents in $\tilde \cG$, a contradiction since $\tilde \cG$ is a directed tree. However, if $z_1=c_r=\root{\tilde \cG}$ then there is an incoming edge into the root node, a contradiction. We conclude that the directed edge $z_1 \to z_2$ also is present in $\tilde \cG$.
	
	Any paths further out on this branch will coincide in both graphs for otherwise there exists nodes with two parents. These arguments show that any paths branching out from the main reversed path will coincide in both $\hat \cG$ and $\tilde \cG$. Thus, the two graphs coincide up to a directed path between root nodes that is reversed.
	
\end{proofenv}

\begin{proofenv}{\Cref{lm:MarkovEquivTreesScoreGap}}
	By \Cref{lm:MarkovEquivTreesPathReversal} there exists a path reversal 
	\begin{align*}
		\begin{array}{rcccccccccc}
			\cG: & \mathrm{rt}(\cG)=c_1 & \to & c_2 &  \to & \cdots & \to & c_{r-1} & \to & c_{r}=\mathrm{rt}(\tilde \cG), \\
			\tilde \cG: & \mathrm{rt}(\tilde \cG)=c_r & \to & c_{r-1}  & \to & \cdots & \to & c_{2} & \to & c_{1}=\mathrm{rt}(\cG),
		\end{array}
	\end{align*}
	while all other edges in $\cG=(V,\cE)$ and $\tilde \cG=(V, \tilde \cE)$ coincide. The entropy score gap is therefore only concerning the root nodes and the reversed edges in the above path. That is
	\begin{align*}
		\lE(\tilde \cG)- \lE(\cG) &= h(X_{\root{\tilde \cG}})  + \sum_{(j,i)\in \tilde \cE} h(X_i-\E[X_i|X_{j}])  \\& \quad \quad -h(X_{\root{ \cG}}) - \sum_{(j,i)\in \cE}  h(X_i-\E[X_i|X_{j}])  \\
		&= h(X_{c_r}) + \sum_{i=1}^{r-1} h(X_{c_i}-\E[X_{c_i}|X_{c_{i+1}}])  \\
		&\quad \quad -h(X_{c_1}) -\sum_{i=2}^{r}  h(X_{c_i}-\E[X_{c_i}|X_{c_{i-1}}]).
	\end{align*}
	Note that
	\begin{align*}
		h(X_{c_r})-h(X_{c_1}) &= \sum_{i=2}^r h(X_{c_i}) - \sum_{i=1}^{r-1} h(X_{c_i}) = \sum_{i=1}^{r-1} h(X_{c_{i+1}})-h(X_{c_i}).
	\end{align*}
	Hence
	\begin{align*}
		&\lE(\tilde \cG)- \lE(\cG) \\
		=\, & \sum_{i=1}^{r-1}  h(X_{c_i}-\E[X_{c_i}|X_{c_{i+1}}]) + h(X_{c_{i+1}}) - h(X_{c_{i+1}}-\E[X_{c_{i+1}}|X_{c_{i}}]) - h(X_{c_i}) \\
		=\, & \sum_{i=1}^{r-1} \Delta \lE (c_i\lra c_{i+1}) \\
		\geq \,&  \min_{1\leq i \leq r-1} \Delta \lE (c_i\lra c_{i+1}),
	\end{align*}
	which concludes the proof.

\end{proofenv}

\subsubsection{Proof of \Cref{lm:thm:scoreGapEntropy}} \label{sec:moredetailsgraphreduction}
We first describe the graphs that result from the reduction 
technique described in \ref{sec:ScoreGapGeneralGraphs}.
To do so, define 
$$
\bL(\cG,\tilde \cG):= \{L\in V_{R}: \CHg{\cG_{R}}{L}=\emptyset \land ( \PAg{\tilde \cG_{R}}{L}\not= \PAg{\cG_{R}}{L} \lor \CHg{\tilde \cG_{R}}{L}\not = \emptyset)\},
$$ 
containing 
the sink nodes in $\cG_R$
that are either not sink nodes in $\tilde \cG_R$ or sink nodes in $\tilde \cG_R$ with different parents: $\PAg{\cG_R}{L}\not = \PAg{\tilde \cG_R}{L}$.
Now fix any $L\in \bL(\cG,\tilde \cG) \subset V_R$ and note that its only parent in $\cG_R$, $\PAg{\cG_R}{L}$, is either also a parent of $L$, a child of $L$ or not adjacent to $L$, in $\tilde \cG_R$. That is, one and only one of the following sets 
is non-empty 
\begin{align*}
	Z(L) :&= \PAg{\cG_R}{L} \cap \PAg{\tilde \cG_R}{L}, &\qquad \text{(`staying parents')}\\
	Y(L) :&= \PAg{\cG_R}{L} \cap \CHg{\tilde \cG_R}{L}, &\qquad \text{(`parents to children')}\\
	W(L) :&= \PAg{\cG_R}{L} \cap (V\setminus \{ L \cup \CHg{\tilde \cG_R}{L} \cup \PAg{\tilde \cG_R}{L}  \}) &\qquad \text{(`removing parents')}
\end{align*}
We define the $\tilde \cG_R$ parent and children of $L$ that are not adjacent to $L$ in $\cG_R$ as
\begin{align*}
	D(L) :&= \PAg{ \tilde \cG_R}{L} \cap (V\setminus \{ L \cup \CHg{ \cG_R}{L} \cup \PAg{ \cG_R}{L}  \}), \text{ and} \\
	O(L):&= \CHg{\tilde \cG_R}{L}  \cap (V\setminus \{ L \cup \CHg{ \cG_R}{L} \cup \PAg{ \cG_R}{L}  \}),
\end{align*}
respectively. 
All such sets contain at most one node and 
by slight abuse of notation, 
we use the same letters to refer to the nodes. 
We will henceforth suppress the dependence on $L$ if the choice is clear from the context.
\Cref{fig:reducedSubgraphs} visualizes the above sets.

Now partition $\cT_p\setminus \{\cG\}$ into the three following disjoint partitions for which there exists a reduced graph sink node $	L\in 	\bL(\cG,\tilde \cG)$  such that $W(L)$, $Y(L)$ and $Z(L)$ is non-empty, respectively. That is, we define
\begin{align*}
	\cT_p(\cG,W):&=\{\tilde \cG\in \cT_p\setminus \{\cG\}: \exists L\in \bL(\cG,\tilde \cG) \text{ s.t. } W(L)\not = \emptyset\}, \\
	\cT_p(\cG,Y):&=\{\tilde \cG\in \cT_p\setminus \{\cG\}: \exists L\in \bL(\cG,\tilde \cG) \text{ s.t. } Y(L)\not = \emptyset\}\setminus\cT_p(\cG,W) ,\\
	\cT_p(\cG,Z):&=\{\tilde \cG\in \cT_p\setminus \{\cG\}: \exists L\in \bL(\cG,\tilde \cG) \text{ s.t. } Z(L)  \not= \emptyset\} \setminus (\cT_p(\cG,W) \cup \cT_p(\cG,Y)).
\end{align*}
Using that $ \cT_p(\cG,W) \cup \cT_p(\cG,Y) \cup \cT_p(\cG,Z) = \cT_p(\cG)$, we can now find a lower bound for the score gap that holds uniformly over all alternative directed tree graphs $\cT_p\setminus \{\cG\}$: 
\begin{align*}
	\min_{\tilde \cG \in \cT_p\setminus \{\cG\}}\lE(\tilde \cG) -\lE(\cG) =&\,  
	\min \bigg\{ \min_{\tilde \cG \in \cT_p(\cG,Z)} \lE(\tilde \cG) -\lE(\cG), \\
	&\quad \quad \quad   \min_{\tilde \cG \in \cT_p(\cG,W)} \lE(\tilde \cG) -\lE(\cG), \min_{\tilde \cG \in \cT_p(\cG,Y)} \lE(\tilde \cG) -\lE(\cG) \bigg\}.
\end{align*}
We now turn to each of these three terms individually and first consider alternative graphs in the partitioning $\cT_p(\cG,Z)$. The following lower bound consists of possibly non-localized conditional dependence properties of the observable distribution $P_X$. (That is, the bound may involve nodes that are not close to each other in the graph $\cG$.)
\begin{restatable}[]{lemma}{ScoreGapCaseOne}
	\label{lm:ScoreGapCaseOne}
	Let $\Pi_Z(\cG)$ denote all tuples $(z,l,o)\in V^3$ of adjacent nodes $(z\to l)\in \cE$ for which there exists a node $o\in \NDg{\cG}{l}\setminus\{z,l\}$. It holds that
	\begin{align*}
		\min_{\tilde \cG \in \cT_p(\cG,Z)} \lE(\tilde \cG) -\lE(\cG)&\geq 	\min_{(z,l,o)\in \Pi_Z(\cG)} I(X_z;X_o|X_l).
	\end{align*}
\end{restatable}
The next result proves a lower bound that holds uniformly over all alternative graphs in $\cT_p(\cG,W)$. The lower bound consists only of local conditional dependence properties. That is, for any subgraph of the causal graph $\cG$ of the form $X_o \to X_w \to X_l$ or $X_o \leftarrow X_w \to X_l$ we measure, by means of conditional mutual information, the conditional dependence of the two adjacent nodes $X_w$ and $X_l$ conditional on $X_o$, $I(X_w;X_l|X_o)$. The lower bound consists of the smallest  of all such local conditional dependence measures.

\begin{restatable}[]{lemma}{ScoreGapCaseTwo}
	\label{lm:ScoreGapCaseTwo}	
	Let $\Pi_W(\cG)$ denote all tuples $(w,l,o)\in V^{3}$ of adjacent nodes $(w\to l)\in \cE$ and $o\in (\CHg{\cG}{w} \setminus \{l\})\cup \PAg{\cG}{w}$. It holds that that\begin{align*}
		\min_{\tilde \cG \in \cT_p(\cG,W)} \lE(\tilde \cG) -\lE(\cG) &\geq \min_{(w,l,o)\in \Pi_W(\cG)} I(X_w;X_l|X_o).
	\end{align*}
\end{restatable}
A uniform lower bound of the score gap over all alternative graphs in the final partition $\cT_p(\cG,Y)$ is given by the smallest edge-reversal of any edge in the causal graph $\cG$.

\begin{restatable}[]{lemma}{ScoreGapCaseThree}
	\label{lm:ScoreGapCaseThree}
	It holds that
	\begin{align*}
		\min_{\tilde \cG \in \cT_p(\cG,Y)} \lE(\tilde \cG) -\lE(\cG) \geq \min_{(j\to i) \in \cE} \Delta \lE( j \lra i).
	\end{align*}	
\end{restatable}
An immediate consequence of \Cref{lm:ScoreGapCaseOne,lm:ScoreGapCaseTwo,lm:ScoreGapCaseThree} is that the entropy identifiability gap is given by the smallest of the lower bounds derived for each partition, see \Cref{lm:thm:scoreGapEntropy}. Thus, it only remains to prove \Cref{lm:ScoreGapCaseOne,lm:ScoreGapCaseTwo,lm:ScoreGapCaseThree}. \\ \\

\begin{proofenv}{\Cref{lm:ScoreGapCaseOne}}
	Let $\tilde \cG\in \Pi_Z(\cG)$ such that $Z\not = \emptyset$. This implies that  $Y=W=\emptyset$ as $L$ can only have one parent in $\cG$. Furthermore, $D=\emptyset$ as $L$ can only have one parent in $\tilde \cG$   and  $O\not = \emptyset$ for otherwise $L$ would have been deleted by the deletion procedure in \Cref{sec:ScoreGap}. Assume without loss of generality that $O=\{O_1,\ldots,O_k\}$ for some $k\in \N$. The two subgraphs are illustrated in \Cref{fig:subgraphsPiZ}.
	\begin{figure}
		\begin{center}
			\begin{tikzpicture}[node distance = 1cm, roundnode/.style={circle, draw=black, fill=gray!10, thick, minimum size=7mm},
				roundnode/.style={circle, draw=black, fill=gray!10, thick, minimum size=7mm},
				outer/.style={draw=gray,dashed,fill=black!1,thick,inner sep=5pt}
				]
				\node[outer,rectangle] (rest) [] {$V_1=\{Z,L\}^c$};
				\node[roundnode] (Z) [above = of rest] {$Z$};
				\node[roundnode] (L) [right = of Z] {$L$};
				\node[above,font=\large\bfseries ] at (current bounding box.north) {$\cG_R$};
				
				\draw[<->, line width=0.4mm] (rest) -- (Z);
				\draw[->, line width=0.4mm] (Z) -- (L);
			\end{tikzpicture} $\quad$	\begin{tikzpicture}[node distance = 1cm, roundnode/.style={circle, draw=black, fill=gray!10, thick, minimum size=7mm},
				roundnode/.style={circle, draw=black, fill=gray!10, thick, minimum size=7mm},
				outer/.style={draw=gray,dashed,fill=black!1,thick,inner sep=5pt}
				]
				\node[outer,rectangle] (rest) [] {$\tilde V_1 =\{Z,L,O,\cA_1,\ldots,\cA_k\}^c$};
				\node[roundnode] (Z) [above = of rest] {$Z$};
				\node[roundnode] (L) [right = of Z] {$L$};
				\node[] (d) [right = of L] {$\vdots$};
				\node[roundnode] (O1) [above = 0.2cm of d] {$O_1$};
				\node[roundnode] (Ok) [below = 0.2cm of d] {$O_k$};
				\node[outer,rectangle] (A1) [right = of O1] {$\cA_1$};
				\node[outer,rectangle] (Ak) [right = of Ok] {$\cA_k$};
				\node[above,font=\large\bfseries ] at (current bounding box.north) {$\tilde \cG_R$};
				
				\draw[<->, line width=0.4mm] (rest) -- (Z);
				\draw[->, line width=0.4mm] (Z) -- (L);
				\draw[->, line width=0.4mm] (L) -- (O1);
				\draw[->, line width=0.4mm] (L) -- (Ok);
				\draw[->, line width=0.4mm] (O1) -- (A1);
				\draw[->, line width=0.4mm] (Ok) -- (Ak);
			\end{tikzpicture}
		\end{center}
		\caption{Illustration of the reduced form graphs $\cG_R$ and  $\tilde \cG_R$ for the case $\tilde \cG \in \Pi_Z(\cG)$. $\cA_1,\ldots,\cA_k$ are possibly empty sets of nodes, and dashed rectangle nodes denotes a possibly multi-node subgraph over the variables enclosed. The bi-directed edges means that the edge can be directed in both directions. An edge pointing into the multi-node subgraph, can possibly be multiple edges into distinct nodes of the subgraph.} \label{fig:subgraphsPiZ}
	\end{figure}
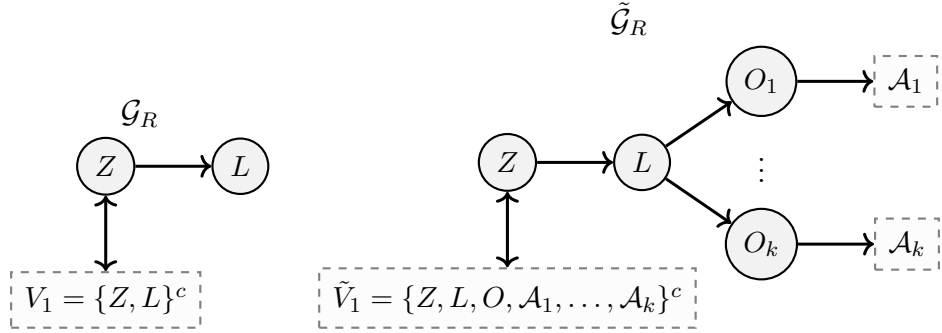
	For ease of notation, fix any $1\leq i\leq k$ and denote $O:=O_i$. We note that in $\tilde \cG$ the following d-separation holds 
	\begin{align*}
		Z\, \dsep{\tilde \cG} O \, |\,  L.
	\end{align*}
	Thus, we have for all probability measures $Q\in  \{\tilde \cG\} \times \cF(\tilde \cG) \times \cP^p$ over nodes $V$ that $Z\independent O\, |\, L$ (as the path between $Z$ and $O$ is blocked by $L$ and all probability measures generated in accordance with an SCM are Markovian with respect to the generating graph $\tilde \cG$). Recall that
	\begin{align*}
		\lE(\tilde \cG)- \lE(\cG) &= \inf_{Q\in  \{\tilde \cG\} \times \cF(\tilde \cG) \times \cP^p}	D_{\mathrm{KL}}(P_X\| Q) \\
		&= \inf_{Q\in  \{\tilde \cG\} \times \cF(\tilde \cG) \times \cP^p} h(P_X,Q) - h(P_X).	
	\end{align*}
	Fix $Q=q\cdot \lambda^{p} \in \{\tilde \cG\} \times \cF(\tilde \cG) \times \cP^p$ and note that it factorizes as $Q=Q_{A|Z,O,L}Q_{Z|L}Q_{O|L}Q_L$, i.e., the density $q$ factorizes as  \begin{align*}
		q(x)&= q_{A|Z,O,L}(a|z,o,l)q_{Z,O,L}(z,o,l)\\
		&= q_{A|Z,O,L}(a|z,o,l)q_{Z|L}(z|l)q_{O|L}(o|l)q_L(l),
	\end{align*}
	for $\lambda^p$-almost all $x=(a,z,o,l)\in \R^p$ where $A=V\setminus \{Z,O,L\}$. Hence, the cross entropy splits additively into
	\begin{align} \notag
		h(P_X,Q) &\geq  \E[-\log(q_{A|Z,O,L}(A|Z,O,L))] \\ \notag
		&\quad \quad +  \E[-\log(q_{Z|L}(Z|L))] \\ \notag
		&\quad \quad  + \E[-\log(q_{O|L}(O|L))] \\
		&\quad \quad  + \E[-\log(q_{L}(L))]. \label{eq:InfCrossEntropyLowerBoundOneCasei}
	\end{align}
	Now note, e.g., that for a conditional distribution (Markov kernel) $Q_{Z|L}$ it holds that
	\begin{align*}
		0\leq 	D_{\mathrm{KL}}(P_{Z|L}P_L\|Q_{Z|L}P_L) &= \E\lf -\log \lp\frac{q_{Z|L}(Z|L)p_L(L)}{p_{Z|L}(Z|L)p_L(L)} \rp    \rf \\
		&= \E[-\log(q_{Z|L}(Z|L))] - \E[-\log(p_{Z|L}(Z|L))],
	\end{align*}
	proving that 
	$$\E[-\log(q_{Z|L}(Z|L))]\geq  \E[-\log(p_{Z|L}(Z|L))].$$
	By similar arguments, we get that the three other terms in the lower bound of \Cref{eq:InfCrossEntropyLowerBoundOneCasei} are bounded below by
	\begin{align*}
		\E[-\log(q_{A|Z,O,L}(A|Z,O,L))] &\geq \E[-\log(p_{A|Z,O,L}(A|Z,O,L))] , \\ 
		\E[-\log(q_{O|L}(O|L))] &\geq \E[-\log(p_{O|L}(O|L))], \\
		\E[-\log(q_L(L))] &\geq \E[-\log(p_L(L))].
	\end{align*}
	This implies that
	\begin{align*}
		\inf_{Q\in \{\tilde \cG\} \times \cF(\tilde \cG) \times \cP^p}	h(P_X,Q) \geq h(P_X,Q^*),
	\end{align*}
	where $Q^* = P_{A|Z,O,L}P_{Z|L}P_{O|L}P_L$. On the other hand, we know that $P_X$ factorizes as  $P_X=P_{A|Z,O,L}P_{Z,O|L}P_{L}$. Thus we have the following entropy score gap lower bound
	\begin{align*}
		\lE(\tilde \cG)- \lE(\cG) & \geq  h(P_X,Q^*) - h(P_X) \\
		&=  	D_{\mathrm{KL}}(P_{X}\|Q^*) \\
		&= D_{\mathrm{KL}}(P_{A|Z,O,L}P_{Z,O|L}P_L\|P_{A|Z,O,L}P_{Z|L}P_{O|L}P_L) \\
		&= D_{\mathrm{KL}}(P_{Z,O|L}P_L\|P_{Z|L}P_{O|L}P_L) \\
		&= D_{\mathrm{KL}}(P_{Z,O|L}\|P_{Z|L}P_{O|L}|P_L) \\
		&=I(Z;O|L).
	\end{align*}
	$\Pi_Z(\cG)$ denotes all tuples $(z,l,o)\in V^3$ of adjacent nodes $(z\to l)\in \cE$ for which there exists a node $o\in \NDg{\cG}{l}\setminus\{z,l\}$. For any graph $\tilde \cG\in \cT_p(\cG,Z)$ we can, by the above considerations, find a tuple $(z,l,o)\in \Pi_Z(\cG)$ such that
	\begin{align*} %
		\lE(\tilde \cG) - \lE(\cG) \geq I(X_o; X_z\, | \, X_l).
	\end{align*} We conclude that
	\begin{align*}
		\min_{\tilde \cG \in \cT_p(\cG,Z)} \lE(\tilde \cG) -\lE(\cG)&\geq 	\min_{(z,l,o)\in \Pi_Z(\cG)} I(X_o;X_z|X_l).
	\end{align*}
	
\end{proofenv}

\begin{proofenv}{\Cref{lm:ScoreGapCaseTwo}}
	Fix any $\tilde \cG\in \cT_p(\cG,W)$ and $L$ with $W \not = \emptyset$ such that $Z=Y=\emptyset$. We have illustrated the subgraph $\cG_R$ in \Cref{fig:GRsubgraphPiW} and the possible subgraphs $\tilde \cG_R$ in \Cref{fig:TildeGrSubgraphPiW}.
	\begin{figure}[h]
		\begin{center}
			\begin{tikzpicture}[node distance = 1cm, roundnode/.style={circle, draw=black, fill=gray!10, thick, minimum size=7mm},
				roundnode/.style={circle, draw=black, fill=gray!10, thick, minimum size=7mm},
				outer/.style={draw=gray,dashed,fill=black!1,thick,inner sep=5pt}
				]
				\node[outer,rectangle] (rest) [] {$\{W,L\}^c$};
				\node[roundnode] (W) [right = of rest] {$W$};
				\node[roundnode] (L) [right = of W] {$L$};
				\node[above,font=\large\bfseries ] at (current bounding box.north) {$\cG_R$};
				
				\draw[<->, line width=0.4mm] (rest) -- (W);
				\draw[->, line width=0.4mm] (W) -- (L);
			\end{tikzpicture} 
		\end{center}
		\caption{Illustrations of the $\cG_R$ subgraph for for $\tilde\cG \in \cT_p(\cG,W)$.} \label{fig:GRsubgraphPiW}
	\end{figure}
	
	\begin{figure}[h]
		\begin{minipage}{.5\textwidth}
			\begin{center}
				\begin{tikzpicture}[node distance = 1cm, roundnode/.style={circle, draw=black, fill=gray!10, thick, minimum size=7mm},
					roundnode/.style={circle, draw=black, fill=gray!10, thick, minimum size=7mm},
					outer/.style={draw=gray,dashed,fill=black!1,thick,inner sep=5pt}
					]
					\node[roundnode] (L) [] {$L$};
					\node[] (d) [right = of L] {$\vdots$};
					\node[roundnode] (O1) [above = 0.2cm of d] {$O_1$};
					\node[roundnode] (Ok) [below = 0.2cm of d] {$O_k$};
					\node[outer,rectangle] (A1) [right = of O1] {$\cA_1$};
					\node[outer,rectangle] (Ak) [right = of Ok] {$\cA_k$};
					\node[above,font=\large\bfseries ] at (current bounding box.north) {$\tilde \cG_R$: $D=\emptyset$, $O\not = \emptyset$};
					
					\draw[->, line width=0.4mm] (L) -- (O1);
					\draw[->, line width=0.4mm] (L) -- (Ok);
					\draw[->, line width=0.4mm] (O1) -- (A1);
					\draw[->, line width=0.4mm] (Ok) -- (Ak);
				\end{tikzpicture} 
			\end{center}
		\end{minipage}
		\begin{minipage}{.5\textwidth} 
			\begin{center}
				\begin{tikzpicture}[node distance = 1cm, roundnode/.style={circle, draw=black, fill=gray!10, thick, minimum size=7mm},
					roundnode/.style={circle, draw=black, fill=gray!10, thick, minimum size=7mm},
					outer/.style={draw=gray,dashed,fill=black!1,thick,inner sep=5pt}
					]
					\node[outer,rectangle] (rest) [] {$\{L,D\}^c$};
					\node[roundnode] (D) [right = of rest] {$D$};
					\node[roundnode] (L) [right = of D] {$L$};
					\node[above,font=\large\bfseries ] at (current bounding box.north) {$\tilde \cG_R$: $D\not = \emptyset$, $O=\emptyset$};
					
					\draw[<->, line width=0.4mm] (rest) -- (D);
					\draw[->, line width=0.4mm] (D) -- (L);
				\end{tikzpicture} 
			\end{center}
		\end{minipage}
		\begin{minipage}{1\textwidth} 
			\begin{center}
				\begin{tikzpicture}[node distance = 1cm, roundnode/.style={circle, draw=black, fill=gray!10, thick, minimum size=7mm},
					roundnode/.style={circle, draw=black, fill=gray!10, thick, minimum size=7mm},
					outer/.style={draw=gray,dashed,fill=black!1,thick,inner sep=5pt}
					]
					\node[outer,rectangle] (rest) [] {$\{D,L,O,\cA_1,\ldots,\cA_k\}^c$};
					\node[roundnode] (D) [right = of rest] {$D$};
					\node[roundnode] (L) [right = of D] {$L$};
					\node[] (d) [right = of L] {$\vdots$};
					\node[roundnode] (O1) [above = 0.2cm of d] {$O_1$};
					\node[roundnode] (Ok) [below = 0.2cm of d] {$O_k$};
					\node[outer,rectangle] (A1) [right = of O1] {$\cA_1$};
					\node[outer,rectangle] (Ak) [right = of Ok] {$\cA_k$};
					\node[above,font=\large\bfseries , yshift=-10mm] at (current bounding box.north) {$\tilde \cG_R$: $D\not = \emptyset$, $O\not = \emptyset$};
					
					\draw[<->, line width=0.4mm] (rest) -- (D);
					\draw[->, line width=0.4mm] (D) -- (L);
					\draw[->, line width=0.4mm] (L) -- (O1);
					\draw[->, line width=0.4mm] (L) -- (Ok);
					\draw[->, line width=0.4mm] (O1) -- (A1);
					\draw[->, line width=0.4mm] (Ok) -- (Ak);
				\end{tikzpicture} 
			\end{center}
		\end{minipage}
		\caption{Illustrations of the possible $\tilde \cG_R$ subgraphs for $\tilde\cG \in \cT_p(\cG,W)$.} \label{fig:TildeGrSubgraphPiW}
	\end{figure}	
	
	Note that for any of the three possible local graph structures presented in \Cref{fig:TildeGrSubgraphPiW} there exists an $A\in\{O_1,\ldots,O_k,D\}$ such that $L\, \dsep{\tilde \cG_R} W\, |\, A$, i.e., A blocks the path between $L$ and $W$. Thus, for all probability measures $Q\in\{\tilde \cG\} \times \cF(\tilde \cG) \times \cP^p$ over nodes $V=\{1,..,p\}$ it always holds that $L\independent W \, | \, A$. By arguments similar to those in the proof of \Cref{lm:ScoreGapCaseOne}, we note that
	\begin{align*}
		\lE(\tilde \cG)- \lE(\cG) &=  \inf_{Q\in\{\tilde \cG\} \times \cF(\tilde \cG) \times \cP^p} h(P_X,Q) - h(P_X),
	\end{align*}
	and  that 
	$$\inf_{Q\in \{\tilde \cG\} \times \cF(\tilde \cG) \times \cP^p}	h(P_X,Q)\geq h(P_X,Q^*),
	$$ for $P_X = P_{K|W,L,A}P_{W,L|A}P_A$ and $Q^*=P_{K|W,L,A}P_{L|A}P_{W|A}P_A$ where $K=V\setminus \{W,L,A\}$. To that end, we now have that
	\begin{align*}
		\lE(\tilde \cG)- \lE(\cG) & \geq  h(P_X,Q^*) - h(P_X) \\
		&=  	D_{\mathrm{KL}}(P_{X}\|Q^*) \\
		&= D_{\mathrm{KL}}(P_{K|W,L,A}P_{W,L|A}P_A\|P_{K|W,L,A}P_{L|A}P_{W|A}P_A) \\
		&= D_{\mathrm{KL}}(P_{W,L|A}P_A\|P_{L|A}P_{W|A}P_A) \\
		&= D_{\mathrm{KL}}(P_{W,L|A}\|P_{L|A}P_{W|A}|P_A) \\
		&=I(W;L|A).
	\end{align*}
	Let $\hat \Pi_W(\cG)$ denote all tuples $(w,l,a)\in V^{3}$ of adjacent nodes $(w\to l)\in \cE$ for which there exists a node $a\in \NDg{\cG}{l}\setminus \{w\}$. Now note that for any graph $\tilde \cG\in \cT_p(\cG,W)$ we can, by the above considerations, find a tuple $(w,l,a)\in \hat\Pi_W(\cG)$ such that
	\begin{align} \label{eq:InequalityCaseTwoSingleGraph}
		\lE(\tilde \cG) - \lE(\cG) \geq I(X_w; X_l\, | \, X_a).
	\end{align}
	(Conversely for any tuple $(w,l,a)\in \hat \Pi_W(\cG)$ we can construct a graph $\tilde \cG\in \cT_p(\cG,W)$ such that \eqref{eq:InequalityCaseTwoSingleGraph} holds. To see this, fix $ (w,l,a)\in \hat \Pi_W(\cG)$ and construct $\tilde \cG$ such that the subtree with root node $l$ is identical in both $\cG$ and $\tilde \cG$ and $a$ blocks the path between $l$ and $w$ in $\tilde \cG$.) Therefore, the following lower bound holds (and it is not unnecessarily small).
	\begin{align*}
		\min_{\tilde \cG\in \cT_p(\cG,W)}\lE(\tilde \cG) - \lE(\cG) \geq \min_{(w,l,a)\in \hat\Pi_W(\cG)} I(X_w; X_l\, | \, X_a).
	\end{align*}
	For any $(w,l,a)\in \hat \Pi_W(\cG)$ it either holds that $a \in (\CHg{\cG}{w} \setminus \{l\})\cup \PAg{\cG}{w} $ or that there exists an $o\in (\CHg{\cG}{w} \setminus \{l\})\cup \PAg{\cG}{w}$ blocking the path between $a$ and $l$ in $\cG$ such that  $X_l\independent X_a | X_o$. Furthermore, we note that as $X_l\independent (X_o,X_a)\, |\, X_w$ we have that
	\begin{align*}
		I(X_w;X_l|X_a) &= h(X_l|X_a)-h(X_l|X_a,X_w) \\
		&= h(X_l|X_a)-h(X_l|X_w) \\
		&= h(X_l|X_a)-h(X_l|X_o,X_w) \\
		&\geq  h(X_l|X_a,X_o)-h(X_l|X_o,X_w) \\ 
		&= h(X_l|X_o)-h(X_l|X_o,X_w) \\
		&= I(X_w;X_l\, | \, X_o),
	\end{align*}
	as further conditioning reduces conditional entropy. Let $\Pi_W(\cG)$ denote all tuples $(w,l,o)\in V^{3}$ of adjacent nodes $(w\to l)\in \cE$ and $o\in (\CHg{\cG}{w} \setminus \{l\})\cup \PAg{\cG}{w}$. By the above considerations we conclude that
	\begin{align*}
		\min_{\tilde \cG\in \cT_p(\cG,W)}\lE(\tilde \cG) - \lE(\cG) \geq \min_{(w,l,o)\in \Pi_W(\cG)} I(X_w; X_l\, | \, X_o).
	\end{align*}
\end{proofenv}

\begin{proofenv}{\Cref{lm:ScoreGapCaseThree}}
	Fix $\tilde \cG\in \cT_p(\cG,Y)$ and $L$ such that $Y\not = \emptyset$. It holds that $W=Z= \emptyset$. We have illustrated the $\cG_R$ in \Cref{fig:GRsubgraphPiY} and the three possible subgraphs $\tilde \cG_R$ in \Cref{fig:TildeGrSubgraphPiY}. 
	\begin{figure}
		\begin{center}
			\begin{tikzpicture}[node distance = 1cm, roundnode/.style={circle, draw=black, fill=gray!10, thick, minimum size=7mm},
				roundnode/.style={circle, draw=black, fill=gray!10, thick, minimum size=7mm},
				outer/.style={draw=gray,dashed,fill=black!1,thick,inner sep=5pt}
				]
				\node[outer,rectangle] (rest) [] {$\{Y,L\}^c$};
				\node[roundnode] (Y) [right = of rest] {$Y$};
				\node[roundnode] (L) [right = of Y] {$L$};
				\node[above,font=\large\bfseries ] at (current bounding box.north) {$ \cG_R$};
				
				\draw[<->, line width=0.4mm] (rest) -- (Y);
				\draw[->, line width=0.4mm] (Y) -- (L);
			\end{tikzpicture} 
		\end{center}
		\caption{Illustrations of the $\cG_R$ subgraph for $\tilde\cG \in \cT_p(\cG,Y)$.} \label{fig:GRsubgraphPiY}
	\end{figure}
	
	\begin{figure}
		\begin{minipage}{.4\textwidth}
			\begin{center}
				\begin{tikzpicture}[node distance = 1cm, roundnode/.style={circle, draw=black, fill=gray!10, thick, minimum size=7mm},
					roundnode/.style={circle, draw=black, fill=gray!10, thick, minimum size=7mm},
					outer/.style={draw=gray,dashed,fill=black!1,thick,inner sep=5pt}
					]
					\node[outer,rectangle] (rest) [] {$\{Y,L,O\}^c$};
					\node[roundnode] (L) [above = of rest] {$L$};
					\node[roundnode] (Y) [left = of L] {$Y$};
					\node[roundnode] (O) [right = of L] {$O$};
					\node[above,font=\large\bfseries ] at (current bounding box.north) {$ \tilde \cG_R$: $D=\emptyset$, $O\not = \emptyset$};
					
					\draw[<->, line width=0.4mm] (rest) -- (L);
					\draw[->, line width=0.4mm] (L) -- (Y);
					\draw[->, line width=0.4mm] (L) -- (O);
				\end{tikzpicture} 
			\end{center}
		\end{minipage}
		\begin{minipage}{.6\textwidth} 
			\begin{center}
				\begin{tikzpicture}[node distance = 1cm, roundnode/.style={circle, draw=black, fill=gray!10, thick, minimum size=7mm},
					roundnode/.style={circle, draw=black, fill=gray!10, thick, minimum size=7mm},
					outer/.style={draw=gray,dashed,fill=black!1,thick,inner sep=5pt}
					]
					\node[outer,rectangle] (rest) [] {$\{Y,L,D\}^c$};
					\node[roundnode] (D) [right = of rest] {$D$};
					\node[roundnode] (L) [right = of D] {$L$};
					\node[roundnode] (Y) [right = of L] {$Y$};
					\node[above,font=\large\bfseries ] at (current bounding box.north) {$ \tilde \cG_R$: $D\not = \emptyset$, $O=\emptyset$};
					
					\draw[<->, line width=0.4mm] (rest) -- (D);
					\draw[->, line width=0.4mm] (D) -- (L);
					\draw[->, line width=0.4mm] (L) -- (Y);
				\end{tikzpicture}
			\end{center}
		\end{minipage}
		\begin{minipage}{1\textwidth} 
			\begin{center}
				\begin{tikzpicture}[node distance = 1cm, roundnode/.style={circle, draw=black, fill=gray!10, thick, minimum size=7mm},
					roundnode/.style={circle, draw=black, fill=gray!10, thick, minimum size=7mm},
					outer/.style={draw=gray,dashed,fill=black!1,thick,inner sep=5pt}
					]
					\node[outer,rectangle] (rest) [] {$\{D,L,Y,O,\cA,\cB\}^c$};
					\node[roundnode] (D) [above = of rest] {$D$};
					\node[roundnode] (L) [right = of D] {$L$};
					\node[] (dummy) [right = of L]{};
					\node[roundnode] (Y) [above = of dummy] {$Y$};
					\node[roundnode] (O) [below = of dummy] {$O$};
					\node[outer,rectangle] (A) [right = of Y] {$\cA$};
					\node[outer,rectangle] (B) [right = of O] {$\cB$};
					\node[above,font=\large\bfseries ] at (current bounding box.north) {$ \tilde \cG_R$: $D\not = \emptyset$,  $O\not = \emptyset$};
					
					\draw[<->, line width=0.4mm] (rest) -- (D);
					\draw[->, line width=0.4mm] (D) -- (L);
					\draw[->, line width=0.4mm] (L) -- (Y);
					\draw[->, line width=0.4mm] (L) -- (O);
					\draw[->, line width=0.4mm] (O) -- (B);
					\draw[->, line width=0.4mm] (Y) -- (A);
				\end{tikzpicture} 
			\end{center}
		\end{minipage}
		\caption{Illustrations of the possible $\tilde \cG_R$ subgraphs for $\tilde\cG \in \cT_p(\cG,Y)$.} \label{fig:TildeGrSubgraphPiY}
	\end{figure}
	Note that for any of the three possible local graph structures of $\tilde \cG_R$ illustrated in \Cref{fig:TildeGrSubgraphPiY} we have that for all probability measures $Q\in\{\tilde \cG\} \times \cF(\tilde \cG) \times \cP^p$ factorizes as $Q_{A|L,Y}Q_{L,Y}$, where $A=V\setminus \{L,Y\}$. It always holds that $Q_{L,Y}$ is the distribution of $(\tilde L, \tilde Y)$ generated in accordance with a structural causal model of the form
	\begin{align} \label{eq:StrucAssignmentTildeYCaseThree}
		\tilde Y:= \tilde f_Y( \tilde L) + \tilde N_Y,
	\end{align}
	where $\tilde f_Y(l) =\E[Y|L=l]$ for all $l\in \R$, and any $\cL(\tilde N_Y),\cL(\tilde L)\in \cP$ with $\tilde N_Y\independent \tilde L$. Now recall that
	\begin{align*}
		\lE(\tilde \cG)- \lE(\cG) &= \inf_{Q\in \{\tilde \cG\} \times \cF(\tilde \cG) \times \cP^p} h(P_X,Q) - h(P_X)	 ,
	\end{align*}
	and notice that by arguments similar to those in the proof of \Cref{lm:ScoreGapCaseOne} we get
	\begin{align*}
		h(P_X,Q) &=  h(P_X,Q_{A|L,Y}Q_{L,Y}) \\
		&=   \E[-\log(q_{A|L,Y}(A|L,Y))]   +  h(P_{L,Y},Q_{L,Y}) \\
		&\geq  \E[-\log(p_{A|L,Y}(A|L,Y))] +  h(P_{L,Y},Q_{L,Y}),
	\end{align*}
	and that $h(P_X) = \E[-\log(p_{A|L,Y}(A|L,Y))] +  h(P_{L,Y})$. Thus, we have that
	\begin{align*}
		\lE(\tilde \cG)- \lE(\cG) \geq \inf_{Q \in\{\tilde \cG\} \times \cF(\tilde \cG) \times \cP^p} h(P_{L,Y},Q_{L,Y}) - h(P_{L,Y}).
	\end{align*}
	For any $Q=Q_{A|L,Y}Q_{L,Y}\in \{\tilde \cG\} \times \cF(\tilde \cG) \times \cP^p$ we have that $Q_{L,Y}$ is uniquely determined by a marginal distribution $Q_{L}\in \cP$ and the noise distribution of $\tilde N_Y\sim q_{\tilde N_Y}\cdot \lambda\in \cP$ from the additive noise structural assignment in \Cref{eq:StrucAssignmentTildeYCaseThree} for $\tilde Y$ and the causal function $\tilde f_Y$. Thus, the density $q_{L,Y}$ of $Q_{L,Y}$ is given by
	\begin{align*}
		q_{L,Y}(l,y)=q_{Y|L}(y|l)q_L(l) = q_{\tilde N_Y}(y-\tilde f_Y(l)) q_L(l) =q_{\tilde N_Y}(y-\E[Y|L=l]) q_L(l) . 
	\end{align*}
	Hence,
	\begin{align*}
		h(P_{L,Y},Q_{L,Y}) 
		&= \E\lf -\log\lp q_{L,Y}(L,Y)\rp  \rf \\
		&= \E\lf -\log\lp q_{Y|L}(Y|L)\rp  \rf + \E\lf -\log\lp q_{L}(L)\rp  \rf \\
		&= \E\lf -\log\lp q_{\tilde N_Y}(Y-\E[Y|L])\rp  \rf +h(P_L,Q_L)  \\
		&= h(Y-\E[Y|L], \tilde N_Y) +h(P_L,Q_L) \\
		&\geq h(Y-\E[Y|L])+ h(L),
	\end{align*}
	where we used that $h(P,Q) = D_{\mathrm{KL}}(P,Q) + h(P) \geq h(P)$.	Thus, we have that \begin{align*}
		\lE(\tilde \cG)- \lE(\cG)& \geq \inf_{Q \in \{\tilde \cG\} \times \cF(\tilde \cG) \times \cP^p} h(P_{L,Y},Q_{L,Y}) - h(P_{L,Y}) \\
		&\geq  h(Y-\E[Y|L])+ h(L) - h(L-\E[L|Y]) - h(Y) \\
		&= \Delta \lE (Y \lra L).
	\end{align*}
	We conclude that
	\begin{align*}
		\min_{\tilde \cG\in \cT_p(\cG,Y)}	\lE(\tilde \cG) - \lE(\cG) \geq \min_{(i\to j)\in \cE}\Delta \lE (j \lra i).
	\end{align*}
	
\end{proofenv}

\subsubsection{Remaining proof of Section~\ref{sec:ScoreGap}}

\begin{proofenv}{\Cref{thm:GaussScoreGapCaseOne}}	Consider a graph $\tilde \cG\in \cT_p(\cG,Z)$ and let $\cG_{R,1}=(\cE_{R,1},V_{R,1})$ and $\tilde \cG_{R,1}=(\tilde \cE_{R,1}, V_{R,1})$ be the reduced graphs after the initial edge and node deletion procedure of \Cref{sec:ScoreGapGeneralGraphs}. The deletion procedure does not change the score gap, that is, $$	\lG(\tilde \cG)- \lG(\cG) = \lG(\tilde \cG_{R,1})- \lG(\cG_{R,1}).$$
	For any $i\geq 1$ and fixed $\cG_{R,i}$ and $\tilde \cG_{R,i}$ we define
	\begin{align*}
		\bL_{R,i}:= \{L\in V_{R,i}: \CHg{\cG_{R,i}}{L}=\emptyset \land ( \PAg{\tilde \cG_{R,i}}{L}\not= \PAg{\cG_{R,i}}{L} \lor \CHg{\tilde \cG_{R,i}}{L}\not = \emptyset)\}.
	\end{align*}
	Now fix $L_{1}\in \bL_{R,1}$ such that $Z_{1}\not = \emptyset$, where $Y_{1},Z_{1},W_{1},D_{1}$ and $O_{1}$ are defined similarly to the variables in  \Cref{sec:ScoreGap}. Let $O_1=\{O_{1,1},\ldots,O_{1,k_1}\}$, for some $k_1\in \N$.

	Assume that there exists an $i\in\{1,\ldots,k_1\}$ such that $(Z_{1}\to O_{1,i})\in \cE_{R,1}$ in which case we have the following two  paths in $\cG_{R,1}$ and $\tilde \cG_{R,1}$ 
	\begin{align*}
		\cG_{R,1}:	O_{1,i} \leftarrow Z_{1} \to L_{1}, \quad \text{and} \quad \tilde \cG_{R,1}: Z_{1}\to L_{1}\to O_{1,i}.
	\end{align*} 
	Since $O_{1,i}\independent_{\tilde \cG_{R,1} } Z_{1}\,|\, L_{1}$, an entropy score gap lower bound is given by
	\begin{align*} \notag
		\lG(\tilde \cG_{R,1})- \lG( \cG_{R,1})&\geq  \lE(\tilde \cG_{R,1})- \lE(\cG_{R,1}) 		\\
		&=  \inf_{Q\in \{\tilde \cG\}\times \cF(\tilde \cG)\times \cP^p} h(P_X,Q) -P_X\\
		&\geq D_{\mathrm{KL}}(P_X \| Q^*)  \\
		&=I(O_{1,i};Z_{1}|L_{1}), 
	\end{align*}
	with $P_X = P_{K|O,Z,L}P_{O,Z|L}P_L$ and $Q^* = P_{K|O,Z,L}P_{Z|L}P_{O|L} P_L$ for $K=V\setminus \{O,Z,L\}$, by arguments similar to those from the proof of \Cref{lm:ScoreGapCaseTwo}. Now note that $ (Z_{1},O_{1,i},L_{1}) \in \Pi_W(\cG_{R,1})\subseteq \Pi_W(\cG)$ as $(Z_1\to O_{1,i})\in \cE_{R,1}$ and $L_1\in \CHg{\cG_{R,1}}{Z_1} \setminus \{O_{1,i}\}\subseteq (\CHg{\cG_{R,1}}{Z_1}\setminus \{O_{1,i}\})\cup \PAg{\cG_{R,1}}{Z_1}$. Hence,
	\begin{align}
		\lG(\tilde \cG_{R,1})- \lG( \cG_{R,1}) \geq \min_{(w,l,o)\in \Pi_W(\cG)}I(X_w;X_l|X_o). \label{eq:LowerBoundIfZToOInfimumOverCombinations}
	\end{align}	
	
	Assume now that for all $i \in \{1,..,k_1\}$ we have $(Z_{1}\to O_{1,i})\not \in \cE_{R,1}$. Let $\hat \cG_{R,1}=(\hat \cE_{R,1}, V_{R,1})$ denote an intermediate graph where $\hat{\cE}_{R,1}$ is identical to $\tilde \cE_{R,1}$ except the edges $\{(L_{1}\to O_{1,i}) :1\leq i \leq k_1\} \subset \tilde \cE_{R,1}$ are replaced by the edges $\{(Z_{1}\to O_{1,i}):1\leq i \leq k_1 \}$. It holds that
	\begin{align*}
		\lG(\tilde \cG_{R,1})- \lG(\cG_{R,1}) 
		&=\lG(\tilde \cG_{R,1}) - \lG(\hat \cG_{R,1}) + \lG(\hat \cG_{R,1}) - \lG(\cG_{R,1}) \\
		&\geq \lG(\hat \cG_{R,1}) - \lG(\cG_{R,1}).
	\end{align*}
	Note that this score gap lower bound is still strictly positive as $\hat \cG_{R,1} \not = \cG_{R,1}$.
	To realize the last inequality (see also \citealp{Peters2022}), simply note that as $O_{1,i}\independent L_{1}\, |\, Z_{1}$ we have for all $i\in \{1,\ldots,k_1\}$ that 
	\begin{align}
		2\lG(\tilde \cG_{R,1},O_{1,i}) &= \log \E[(O_{1,i}-\E[O_{1,i}|L_{1}])^2] \notag\\
		&\geq \log \E[(O_{1,i}-\E[O_{1,i}|Z_{1},L_{1}])^2] \notag \\
		&= \log \E[(O_{1,i}-\E[O_{1,i}|Z_{1}])^2] \notag \\
		&=2\lG(\hat \cG_{R,1},O_{1,i}).\label{eq:intermediategraphineq}
	\end{align}
	Now since all edges in $\tilde \cG_{R,1}$ and $\hat \cG_{R,1}$ coincide except the incoming edges into $O_{1,1},\ldots,O_{1,k_1}$ we get that $$
	\lG(\tilde \cG_{R,1}) -\lG(\hat \cG_{R,1}) = \sum_{i=1}^{k_1} \lG(\tilde \cG_{R,1},O_{1,i})-\lG(\hat \cG_{R,1},O_{1,i}) \geq 0,$$
	where the inequality follows from \Cref{eq:intermediategraphineq}. 
	Now both $\hat \cG_{R,1}$ and $\cG_{R,1}$ have a childless  node $L_{1}$ with the same parent $Z_{1}$, so we let  $\tilde \cG_{R,2}$ and $\cG_{R,2}$ denote these two graphs where the node $L_{1}$ and its incoming edge are deleted. This deletion does not change the graph scores, i.e., 
	\begin{align*}
		\lG(\hat  \cG_{R,1})- \lG(\cG_{R,1})
		&=  \lG(\tilde \cG_{R,2})- \lG(\cG_{R,2}).
	\end{align*}
	Now fix $L_{2}\in \bL_{R,2}$ and define $Y_{2},Z_{2},W_{2},D_{2}$ and $O_{2}=\{O_{2,1},\ldots,O_{2,k_2}\}$ accordingly.

	If either $Y_{2}$ or $W_{2}$ is non-empty, we use the score gap lower bound previously discussed in \Cref{lm:ScoreGapCaseTwo} and \Cref{lm:ScoreGapCaseThree}. If $Z_{2}$ is non-empty, we can repeat the above procedure and iteratively move edges and delete nodes until  we arrive at the first $i\in \N$ with $\tilde \cG_{R,i}$ and $\cG_{R,i}$ being the iteratively reduced graphs and $L_{R,i}\in \bL_{R,i}$ where either
	\begin{itemize}
		\item[i)]  $Y_{i}$ or $W_{i}$ is non-empty, here, we get that $\lG(\tilde \cG) - \lG( \cG)$ is lower bounded by a bound similar to the form of \Cref{lm:ScoreGapCaseTwo} or \Cref{lm:ScoreGapCaseThree}. That is,
		\begin{align*}
			\lG(\tilde \cG_{R,i})- \lG( \cG_{R,i})\geq 	&\,  \lE(\tilde \cG_{R,i})-\lE(\cG_{R,i}) \\
			\geq  &\, \min\left\{\min_{j\to i \in \cE} \Delta \lE( i \lra j),\min_{(w,l,o)\in \Pi_W(\cG)} I(X_w; X_l\, | \, X_o)\right\}.
		\end{align*}
		\item[ii)] $Z_{i}$ is non-empty and there exists a $j\in\{1,\ldots,k_i\}$ such that $(Z_{i} \to O_{i,j})\in \cG_{R,i}$. As previously argued, the score gap lower bound of \Cref{eq:LowerBoundIfZToOInfimumOverCombinations} applies. That is
		\begin{align*}
			\lG(\tilde \cG_{R,i})- \lG( \cG_{R,i}) \geq \lE(\tilde \cG_{R,i})-\lE(\cG_{R,i}) &\geq \min_{(w,l,o)\in \Pi_W(\cG)} I(X_w; X_l\, | \, X_o).
		\end{align*}
	\end{itemize}
	Note that whenever we do not meet scenario i) or ii) we remove a node in both graphs that is a sink node in the reduced true causal graph $\cG_{R,i}$ and the intermediate graph $\hat \cG_{R,i}$. After at most $p-2$ graph reduction iterations of not encountering scenario i) or ii) we are left with two different graphs on two nodes, in which case the score gap is an edge reversal.
	We conclude that
	\begin{align*}
		\lG(\tilde \cG) - \lG(\cG) 
		&\geq \min\left\{\min_{i\to j \in \cE} \Delta \lE( j \lra i),\min_{(w,l,o)\in \Pi_W(\cG)} I(X_w; X_l\, | \, X_o)\right\}.
	\end{align*}
	
\end{proofenv}

\end{appendices}

\bibliographystyle{plainnat-reversed-doi}
\bibliography{
bibfile_merged}

\end{document}